# Survey of Various Fuzzy and Uncertain Decision-Making Methods

## Foundations, Models, and Computational Approaches in Uncertain Decision Science

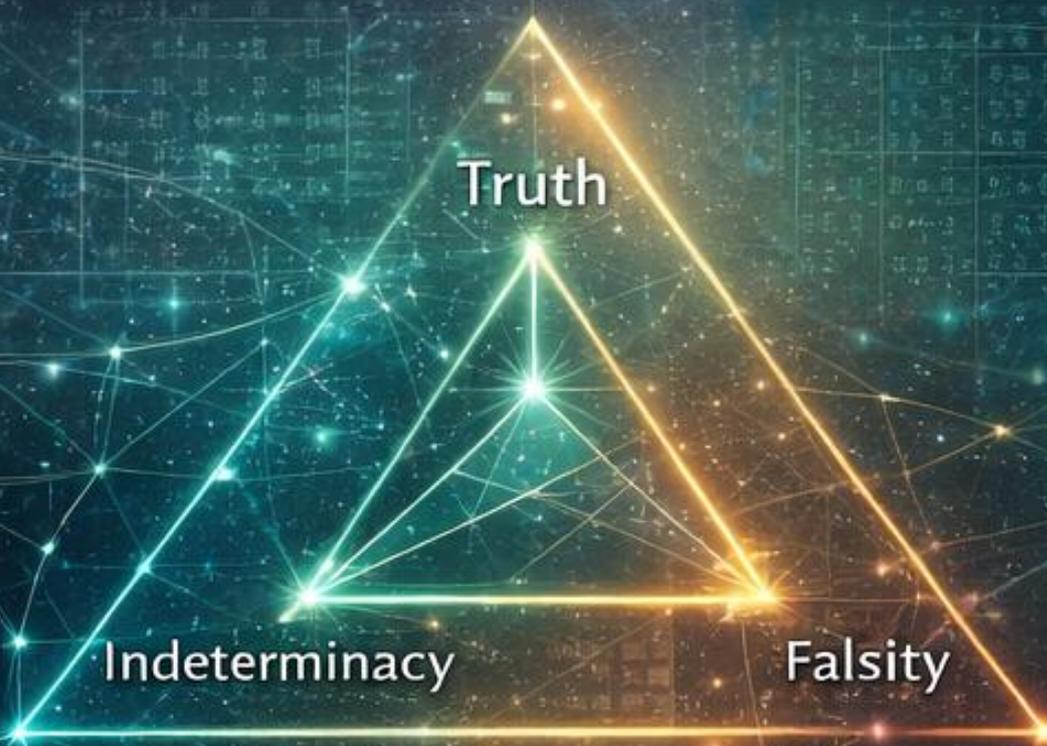

Truth

Indeterminacy

Falsity

Research Monograph in Fuzzy, Neutrosophic, and Multi-Criteria Decision Systems

**Takaaki Fujita, Florentin Smarandache**

# Survey of Various Fuzzy and Uncertain Decision-Making Methods

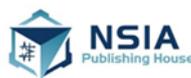

*Editor:*

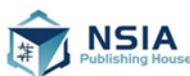

*Peer-Reviewers:*

**Fernando A. F. Ferreira**
ISCTE Business School, BRU-IUL, University Institute of
Lisbon, Avenida das Forças Armadas, 1649-026 Lisbon,
Portugal
Email: fernando.alberto.ferreira@iscte-iul.pt

**Julio J. Valdés**
National Research Council Canada, M-50, 1200 Montreal
Road, Ottawa, Ontario K1A 0R6, Canada
Email: julio.valdes@nrc-cnrc.gc.ca

**Tieta Putri**
College of Engineering, Department of Computer Science
and Software Engineering, University of Canterbury,
Christchurch, New Zealand

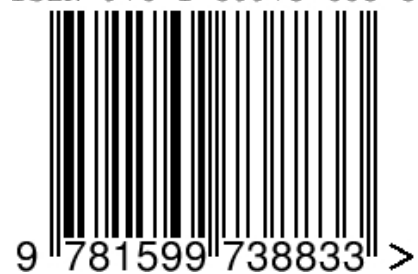

# Contents in this book

The remainder of this book is organized as follows.

# Survey of Various Fuzzy and Uncertain Decision-Making Methods

**Takaaki Fujita [1] * and Florentin Smarandache[2]**
[1] Independent Researcher, Tokyo, Japan.
Email: Takaaki.fujita060@gmail.com
[2] University of New Mexico, Gallup Campus, NM 87301, USA.
Email: fsmarandache@gmail.com

## Abstract

Decision-making in real applications is often affected by vagueness, incomplete information, heterogeneous data, and conflicting expert opinions. This survey reviews uncertainty-aware multi-criteria decision-making (MCDM) and organizes the field into a concise, task-oriented taxonomy.

We summarize problem-level settings (discrete, group/consensus, dynamic, multi-stage, multi-level, multi-agent, and multi-scenario), weight elicitation (subjective and objective schemes under fuzzy/linguistic inputs), and inter-criteria structure and causality modelling. For solution procedures, we contrast compensatory scoring methods, distance-to-reference and compromise approaches, and non-compensatory outranking frameworks for ranking or sorting. We also outline rule/evidence-based and sequential decision models that produce interpretable rules or policies.

The survey highlights typical inputs, core computational steps, and primary outputs, and provides guidance on choosing methods according to robustness, interpretability, and data availability. It concludes with open directions on explainable uncertainty integration, stability, and scalability in large-scale and dynamic decision environments.

# Chapter 1

# Introduction

## 1.1 Uncertain Set

Real-world systems rarely provide perfectly crisp information. Observations may be imprecise, partially reliable, or incomplete. To model such uncertainty in a mathematically disciplined manner, many generalized set-theoretic formalisms have been proposed, including Fuzzy Sets [1], Intuitionistic Fuzzy Sets [2], Neutrosophic Sets [3,4], Vague Sets [5], Hesitant Fuzzy Sets [6], Picture Fuzzy Sets [7], Quadripartitioned Neutrosophic Sets [8], PentaPartitioned Neutrosophic Sets [9], Plithogenic Sets [10], HyperFuzzy Sets [11], and HyperNeutrosophic Sets [12]. Applications of fuzzy models and their extensions—outlined in later parts of this book—have been investigated extensively in decision science, chemistry, control, and machine learning [13]. In practice, the appropriate set model depends on (i) the phenomenon being described and (ii) how many uncertainty components are needed to represent it faithfully.

In the classical fuzzy paradigm, each element $x$ of a universe $X$ is assigned a single membership grade $\mu(x) \in [0,1]$, indicating the extent to which $x$ belongs to a given fuzzy set [1]. For reference, the comparison between classical (crisp) sets and fuzzy sets is presented in Table 1.1.

Table 1.1: Concise comparison between classical (crisp) sets and fuzzy sets.

| Aspect | Classical (Crisp) Set | Fuzzy Set |
|---|---|---|
| Membership model | Indicator function $\chi_A : X \to \{0,1\}$ | Membership function $\mu_A : X \to [0,1]$ [1] |
| Decision boundary | Sharp (in/out) | Gradual (degrees of belonging) |
| Expresses uncertainty | Not directly (binary membership only) | Yes (graded membership captures vagueness) |
| Basic operations | $\cup$, $\cap$, $^c$ defined via Boolean logic | Generalized via t-norms/t-conorms (e.g., max / min as a common choice) |
| Typical use cases | Exact categories, deterministic rules | Imprecise concepts, soft thresholds, human judgments |

For an intuitionistic fuzzy set, each $x \in X$ is characterized by a pair $(\mu(x), \nu(x))$ of membership and non-membership values, where $\mu, \nu : X \to [0,1]$ satisfy

$$0 \le \mu(x) + \nu(x) \le 1,$$

[2, 14] and the residual quantity $1 - \mu(x) - \nu(x)$ is commonly interpreted as hesitation. A neutrosophic set refines this representation by associating to each $x \in X$ a triple

$$(T(x), I(x), F(x)),$$

where $T(x)$, $I(x)$, and $F(x)$ denote degrees of truth, indeterminacy, and falsity, typically in $[0, 1]$. Unlike the intuitionistic fuzzy constraint, neutrosophic components are not required to sum to 1, which enables the encoding of incomplete, inconsistent, or redundant information in a flexible manner [14, 15]. [1] Neutrosophy emphasizes the conceptual significance of neutrality and indeterminacy and has stimulated parallel developments in neutrosophic logic, probability, statistics, measure theory, integration, and related formalisms. These tools now appear in a wide variety of scientific and engineering applications [13].

Plithogenic sets further broaden this landscape by modeling each element through its attribute values together with corresponding degrees of appurtenance, and by introducing a contradiction (dissimilarity) function between distinct attribute values [10, 17, 18]. This added structure supports context-aware aggregation of heterogeneous and potentially conflicting assessments, thereby refining classical fuzzy, intuitionistic fuzzy, and neutrosophic descriptions (see, e.g., [19, 20]). For quick reference, Table 1.2 summarizes the canonical information associated with each element across several representative set extensions (using notation harmonized for this book).

Table 1.2: Representative set extensions and the canonical information stored per element.

| Set Type | Canonical data attached to each element |
| --- | --- |
| Fuzzy Set | Membership mapping $\mu : X \to [0, 1]$. |
| Intuitionistic Fuzzy Set | Membership $\mu$ and non-membership $\nu$ with $\mu(x) + \nu(x) \leq 1$; the gap $1 - \mu(x) - \nu(x)$ is interpreted as hesitation. |
| Neutrosophic Set | Triple $(T, I, F)$ with $T, I, F \in [0, 1]$ (truth, indeterminacy, falsity), treated as independent coordinates. |
| Plithogenic Set | Tuple $(P, v, Pv, \text{pdf}, \text{pCF})$ where $\text{pdf} : P \times Pv \to [0, 1]^s$ encodes $s$-dimensional appurtenance and $\text{pCF} : Pv \times Pv \to [0, 1]^t$ is a symmetric contradiction map taking values in $[0, 1]^t$. |

As a further generalization, Uncertain Sets are also known [20]. Uncertain sets extend classical sets by assigning graded, multi-component membership information (e.g., truth and indeterminacy) to elements, thereby modeling vagueness and incomplete data in a rigorous way. These concepts are also often studied in conjunction with notions such as grey sets, interval sets, rough sets, near sets, soft sets, and hypersoft sets.

## 1.2 Uncertain Decision-Making

Decision-making selects an action among alternatives by evaluating objectives, criteria, constraints, and uncertainty to achieve desired outcomes. These ideas have been extended by incorporating uncertainty-oriented logics, leading to diverse research directions such as Fuzzy Decision-Making [21, 22], Intuitionistic

---

[1] Intuitionistic fuzzy sets do not treat indeterminacy as a primary coordinate, whereas neutrosophic operators incorporate indeterminacy on the same footing as truth-membership and falsehood-nonmembership [14, 16]. It is broadly accepted that the neutrosophic set framework subsumes several well-known models, including intuitionistic fuzzy sets, inconsistent intuitionistic fuzzy sets (covering picture fuzzy and ternary fuzzy sets), Pythagorean fuzzy sets, spherical fuzzy sets, and $q$-rung orthopair fuzzy sets. In a similar spirit, neutrosophication extends a range of decision and uncertainty paradigms, such as regret theory, grey system theory, and three-way decision theory [16].

Fuzzy Decision-Making [23, 24], Neutrosophic Decision-Making [25, 26], and Plithogenic Decision-Making [27–29]. Moreover, decision-making itself admits a wide variety of methodologies; for example, TOPSIS [30], AHP [31], ANP [32], CoCoSo, and DEMATEL are well-known representative methods.

By combining (i) a chosen uncertain-set paradigm (e.g., fuzzy, intuitionistic fuzzy, neutrosophic, plithogenic, or other uncertain sets) with (ii) a selected decision-making technique, one can investigate what novel and practically meaningful outcomes emerge. In particular, research explores how efficiently real-world decision processes can be supported or improved in terms of accuracy, robustness, transparency, and computational cost.

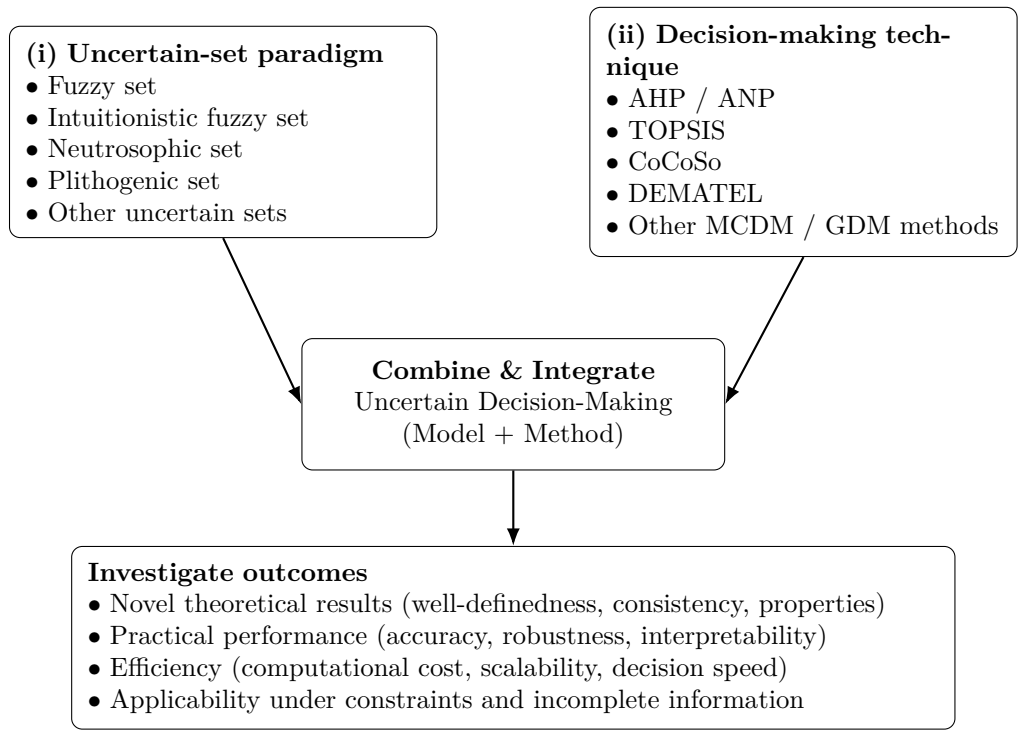

Figure 1.1: Conceptual diagram: combining an uncertain-set paradigm with a decision-making technique to obtain novel and practically meaningful outcomes.

## 1.3 Our Contributions

In view of the discussion above, the paradigms of decision-making and uncertain sets—including fuzzy sets and neutrosophic sets—are of fundamental importance. In this book, we conduct a comprehensive survey of methods for uncertain decision-making. Our primary objective is to provide a useful reference that supports and informs future research by specialists in this area. In addition, we propose several new decision-making methods.

For reference, a compact taxonomy of methods in *Uncertain Decision Science*, classified by task, input structure, and typical outputs, is presented in Table 1.3. Since the table extends across multiple pages, we apologize for any inconvenience. In this book, each of these concepts will be discussed briefly, together with related methods where appropriate.

Table 1.3: Compact taxonomy of methods in *Uncertain Decision Science* (classified by task, input structure, and typical outputs).

| Method | Primary task | Typical inputs | Primary outputs | Notes / when used |
|---|---|---|---|---|
| **A. Problem-level frameworks (problem statements / settings)** | | | | |
| Fuzzy MCDM (general) [33] | Framework | Alternatives × criteria; fuzzy/uncertain evaluations; weights | Ranking / best alternative | Umbrella viewpoint for most items below. |
| Fuzzy MADM [34] | Framework (discrete) | Finite alternatives; attributes; fuzzy numbers / linguistic ratings | Ranking | A common discrete MCDM setting (matrix-based). |
| Fuzzy Group-Decision Making [35] | Framework (group) | Multiple experts; individual fuzzy matrices; expert weights | Collective matrix; ranking | Includes aggregation + (optional) consensus stage. |
| Fuzzy Consensus decision-making | Consensus process | Experts' fuzzy opinions; similarity/consensus threshold | Consensus-reaching + final choice | Explicit iteration until consensus is reached. |
| Fuzzy Multiple Objective Decision-making [36] | Optimization (multi-objective) | Continuous decision vector; multiple (possibly fuzzy) objectives/goals | Compromise solution | Often reduced to a membership-maximization scalarization. |
| Fuzzy ethical decision making | Rule/criteria decision | Ethical dimensions; fuzzy rules or fuzzy scores per action | Action choice / ranking | Domain-specific MCDM (often fuzzy inference). |
| Fuzzy Strategic decision making | Planning decision | Projects/measures; goal achievement + cost/resource satisfaction | Optimal plan / selection set | Often modeled as weighted fuzzy satisfaction under constraints. |
| Fuzzy Dynamic Decision-Making (Dynamic / multi-period MCDA) (cf. [37]) | Framework (dynamic) | Time-indexed evaluations/weights $(\tilde{x}_{ij}(t), w_j(t))$; update/discount rules; (optional) scenario data | Time-aggregated ranking or adaptive policy/plan | Models evolving preferences/performances across periods; aggregates over $t = 1, \ldots, T$ (e.g. discounted, rolling, or state-updated). |

| Method | Primary task | Typical inputs | Primary outputs | Notes / when used |
|---|---|---|---|---|
| Fuzzy Multi-Expert Decision-Making | Framework (group / multi-expert) | Experts' fuzzy evaluations (matrices or preference relations); expert weights; aggregation/consensus rule | Collective fuzzy assessment; ranking/choice | General group MCDA: aggregates multiple experts' fuzzy opinions (optionally with consensus constraints) into a single decision. |
| Fuzzy Multi-Stage Decision-Making | Framework (multi-stage) | Stage-indexed criteria/alternatives; inter-stage constraints; fuzzy transition/feedback rules | Stage-wise decisions; final plan/policy | Sequential decisions across stages; earlier choices affect feasible sets and evaluations at later stages (fuzzy uncertainty propagated). |
| Fuzzy Multi-Level Decision-Making | Framework (hierarchical / bilevel) | Leader–follower objectives/constraints (or hierarchical criteria); fuzzy goals and constraints | Hierarchical solution (compromise) | Models decisions across levels (e.g., bilevel): upper-level choices influence lower-level feasible responses under fuzzy preferences/targets. |
| Fuzzy Multi-Agent Decision-Making | Framework (multi-agent) | Agents' utilities/preferences (fuzzy); interaction protocol (cooperative/competitive); information sharing rules | Joint decision, equilibrium, or negotiated outcome | Multiple agents with possibly conflicting fuzzy preferences; solution via aggregation, bargaining, or game-theoretic negotiation mechanisms. |
| Fuzzy Multi-Scenario Decision-Making | Framework (scenario-based) | Scenario set $\Omega$; scenario weights/probabilities (possibly fuzzy); scenario-wise fuzzy performances | Robust ranking/choice across scenarios | Evaluates alternatives under multiple scenarios; aggregates scenario-wise fuzzy outcomes (e.g., expected, worst-case, regret) for robust selection. |

**B. Weight elicitation (derive criterion weights)**

| Method | Primary task | Typical inputs | Primary outputs | Notes / when used |
|---|---|---|---|---|
| Fuzzy AHP [38] | Weighting (pairwise) | Fuzzy reciprocal pairwise comparison matrices (hierarchy) | Local/global weights; priorities | Tree/hierarchy; geometric mean / eigenvector variants. |
| Fuzzy ANP [39] | Weighting (network) | Pairwise comparisons + dependence network; supermatrix | Global priorities | Generalizes AHP to interdependencies (limit supermatrix). |
| Fuzzy BWM [40] | Weighting (best–worst) | Best-to-others and others-to-worst fuzzy ratios | Weights (optimization) | Consistency-controlled, fewer comparisons than AHP. |

| Method | Primary task | Typical inputs | Primary outputs | Notes / when used |
|---|---|---|---|---|
| Fuzzy CILOS (Criterion Impact Loss) [41] | Objective weighting | Normalized fuzzy decision matrix; impact-loss computation per criterion (fuzzy) | Objective weights (possibly fuzzy $\tilde{w}_j$) | Assigns larger weights to criteria causing greater information/impact loss when omitted or deteriorated (fuzzy loss-based weighting). |
| Fuzzy SWARA [42] | Weighting (stepwise) | Ordered criteria; fuzzy stepwise importance coefficients | Weights (sequential) | Fast elicitation when an importance order is known. |
| Fuzzy FUCOM [43] | Weighting (full consistency) | Ordered criteria; consecutive fuzzy priority ratios | Weights + deviation index | Optimization enforces ratio + transitivity consistency. |
| Fuzzy PIPRECIA [44] | Weighting (pivot/step) | Sequential comparisons to a pivot / neighbor; fuzzy scales | Weights (sequential) | Ordinary + inverse pass often used for robustness. |
| Fuzzy OPA [45] | Weighting (ordinal) | Experts' ordinal rankings (+ optional fuzzy importances) | Weights (max–min model) | No pairwise matrices; uses an optimization model. |
| Fuzzy judgment matrix | Data structure | Fuzzy preference degrees (often complementary) in $[0, 1]$ | Input for weights / consistency | Used in fuzzy AHP-like pipelines and preference modeling. |
| Fuzzy LOPCOW (Logarithmic Percentage Change-driven Objective Weighting) [46] | Objective weighting | Fuzzy decision matrix (defuzzified); benefit/cost types; normalization | Criterion weights $w_j$ (and importance ranking) | Computes $PV_j = 100 \left\|\ln(\mathrm{RMS}_j/\sigma_j)\right\|$ from normalized data, then normalizes $PV_j$ to obtain $w_j$. |
| Fuzzy SIWEC (F-SIWEC) [47] | Weighting (direct rating; dispersion-adjusted) | Experts' linguistic importance ratings → TFNs; normalization; (optional) expert weights | Fuzzy (or defuzzified) criterion weights $\tilde{w}_j$ (and criteria ranking) | Fast weighting without pairwise comparisons; uses dispersion (e.g. standard deviation) to reflect experts' disagreement. |

| Method | Primary task | Typical inputs | Primary outputs | Notes / when used |
|---|---|---|---|---|
| Fuzzy IDOCRIW [48] | Objective weighting (integrated) | Normalized fuzzy decision matrix (benefit/cost); entropy dispersion; CILOS impact-loss (fuzzy/defuzzified); aggregation rule | Integrated objective weights $w_j^{\text{(IDOCRIW)}}$ (and criteria ranking) | Purely data-driven weights combining information dispersion and relative loss; used to reduce subjectivity and stabilize weighting when criteria scales/variability differ. |

**C. Structure / causality modelling (inter-criteria influence)**

| Method | Primary task | Typical inputs | Primary outputs | Notes / when used |
|---|---|---|---|---|
| Fuzzy DEMATEL [49] | Causal analysis | Fuzzy direct-relation (influence) matrix among criteria | Total relation; cause/effect groups | Outputs prominence $(D+R)$ and relation $(D-R)$. |
| Fuzzy ISM (Interpretive Structural Modeling) [50] | Structural modelling (hierarchy) | Expert reachability judgments (binary $\rightarrow$ fuzzy); reachability matrix; transitive closure | Hierarchical directed influence graph | Fuzzy ISM uses graded reachability from linguistic terms; fuzziness is propagated through closure. |
| Fuzzy MICMAC (cross-impact / driving–dependence) [51] | Influence classification | (ISM) reachability matrix (binary/fuzzy); cross-impact strengths | Driving power / dependence indices; clusters (autonomous, dependent, linkage, driving) | Fuzzy MICMAC computes graded driving–dependence strengths and yields robust clustering under ambiguity. |
| Fuzzy Total ISM (TISM) | Interpretable structure model | ISM reachability + link-wise rationales (expert explanations) | Hierarchy + annotated edges (strength + interpretation) | Adds explicit justifications to each link while preserving the ISM-derived hierarchical structure. |
| Fuzzy Cognitive Map (FCM) [52] | Causal network (dynamic) | Signed directed causal graph; fuzzy edge weights; initial concept activations | Scenario dynamics; steady states / trajectories; influence graph | Uses fuzzy causal weights and nonlinear updates to model feedback loops and uncertain propagation. |

**D. Compensatory scoring on a decision matrix ("weighted-sum" families)**

| Method | Primary task | Typical inputs | Primary outputs | Notes / when used |
|---|---|---|---|---|
| Fuzzy SAW [53] | Ranking (additive) | Normalized fuzzy decision matrix + weights | Score and ranking | Prototype weighted-sum method; highly interpretable. |
| Fuzzy SMART [54] | Ranking (additive) | Value scales + swing weights; fuzzy ratings possible | Utility score; ranking | Essentially structured SAW with explicit value scaling. |

| Method | Primary task | Typical inputs | Primary outputs | Notes / when used |
|--------|--------------|----------------|-----------------|-------------------|
| Fuzzy MAUT [55] | Utility aggregation | Single-attribute utilities; weights; fuzzy outcomes possible | Overall utility; ranking | Utility-theoretic (often additive/multiplicative forms). |
| Fuzzy MAC-BETH [56] | Value scale construction | Qualitative pairwise difference-of-attractiveness judgments | Value functions + ranking | Builds numerical value scales from linguistic categories. |
| Fuzzy CO-PRAS [57] | Ranking (benefit/cost) | Decision matrix; weights; benefit/cost partition | Utility degree; ranking | Separates benefit and cost sums in the scoring. |
| Fuzzy MOORA [58] | Ranking (ratio) | Normalized decision matrix; weights; benefit–cost split | Net score; ranking | Often used with Multi-MOORA extensions. |
| Fuzzy ARAS [59] | Ranking (add ideal alt.) | Decision matrix + added optimal alternative; weights | Utility ratio; ranking | Compares each alternative to an explicitly added ideal. |
| Fuzzy WAS-PAS [60] | Ranking (hybrid) | Decision matrix; weights; normalization | Integrated score; ranking | Combines WSM (sum) and WPM (product). |
| Fuzzy Co-CoSo [61] | Ranking (compromise) | Decision matrix; weights; normalized sums/powers | Compromise score; ranking | Combines additive and multiplicative aggregations. |
| Fuzzy REGIME | Pairwise win/loss aggregation | Pairwise comparisons per criterion; weights | Preference matrix; ranking | Aggregates criterion-wise wins/losses across alternatives. |
| Fuzzy TODIM [62] | Prospect-theory ranking | Reference point; gains/losses; attenuation parameter; weights | Dominance values; ranking | Captures loss aversion / asymmetric preference. |
| Fuzzy GRA | Similarity ranking | Reference sequence; normalized data; distinguishing coefficient | Relational grade; ranking | Closeness to ideal pattern via grey relational grades. |
| Fuzzy Preference Selection Index | Ranking (index) | Decision matrix; normalization (often no explicit weights) | Preference index; ranking | Weight-light / weight-free variants are common. |
| Fuzzy Range of Value method | Ranking (interval/value range) | Normalized performances; weights; value ranges | Overall value; ranking | Emphasizes value intervals/ranges in aggregation. |

| Method | Primary task | Typical inputs | Primary outputs | Notes / when used |
|---|---|---|---|---|
| Fuzzy MOOSRA (Multi-Objective Optimization on the basis of Simple Ratio Analysis) [63] | Ratio-based scoring | Normalized fuzzy performances; weights; benefit/cost partition | Ratio score; ranking | Computes a weighted ratio-type utility from fuzzy normalized values (often benefit-over-cost style) to obtain a compensatory ranking. |
| Fuzzy AROMAN (Alternative Ranking Order Method Accounting for two-step Normalization) [64] | Ranking (normalization-based scoring) | Decision matrix; weights; benefit/cost partition; two-step normalization | Overall utility score; ranking | Performs two-step normalization of criterion values, then aggregates weighted normalized performances to obtain a final ranking. |
| Fuzzy RANCOM (e.g., PTF-RANCOM) [65] | Weighting (expert-judgment based) | Criteria; experts' linguistic importance assessments (fuzzy); (optional) expert weights; aggregation operator | Criterion weights (and criteria ranking) | Aggregates fuzzy importance evaluations, scores and ranks criteria, then derives normalized weight coefficients for subsequent MCDA methods. |
| Fuzzy RAFSI [66] | Ranking (functional mapping; rank-reversal resistant) | Fuzzy decision matrix (e.g., TFNs); weights; ideal/anti-ideal reference points; mapping interval $(n_1, n_2)$ | Utility score; ranking | Maps each criterion sub-interval to a common interval, applies max/min normalization (via $A, H$), then aggregates weighted scores. |
| Fuzzy RATMI | Ranking (trace-to-median index) | Fuzzy decision matrix; weights; benefit/cost types; normalization; parameter $v$ | Index score $E_i$; ranking | Aggregates weighted normalized performances via a trace index and a median-similarity term to obtain a final compromise score. |
| **E. Distance-to-reference / border / compromise-index methods** | | | | |
| Uncertain TOPSIS [67] | Distance to (ideal, nadir) | (uncertain/fuzzy) decision matrix; weights; metric | Closeness coefficient; ranking | Ranks by distance to FPIS/FNIS (ideal/anti-ideal). |

| Method | Primary task | Typical inputs | Primary outputs | Notes / when used |
|---|---|---|---|---|
| Uncertain VIKOR [68] | Compromise programming | Decision matrix; weights; best/worst; parameter $v$ | Compromise index $Q$; solution set | Balances group utility and individual regret. |
| Uncertain MABAC [69] | Border approximation | Decision matrix; weights; border area construction | Deviation score; ranking | Ranks by signed distance from border approximation area. |
| Fuzzy EDAS [70] | Distance from average | Decision matrix; average solution; positive/negative distances | Appraisal score; ranking | Uses distances to the average (not to ideal). |
| Fuzzy CODAS [71] | Distance (Euclid + Taxicab) | Decision matrix; negative-ideal; distance measures | Relative assessment; ranking | Often uses Euclidean + Manhattan discrimination. |
| Fuzzy MARCOS [72] | Utility vs ideal/anti-ideal | Extended matrix with ideal/anti-ideal; normalization; weights | Utility functions; ranking | Explicitly includes both ideal and anti-ideal references. |
| Fuzzy MAIRCA (Multi-Attributive Ideal–Real Comparative Analysis) | Ideal–real comparative ranking | Fuzzy decision matrix; weights; theoretical/ideal distribution vs observed (real) performance | Gap-based score; ranking | Compares "theoretical" (ideal) expectations with "real" fuzzy performances; ranks by aggregated fuzzy deviations (gaps). |
| Fuzzy FlowSort | Outranking-based sorting | Preference functions and flows (PROMETHEE); reference profiles; fuzzy evaluations/weights | Category assignment via flows | PROMETHEE-flow sorting: compares alternatives to limiting profiles and assigns classes using positive/negative/net flow rules (fuzzy inputs). |

### F. Outranking (non-compensatory / dominance relations)

| Method | Primary task | Typical inputs | Primary outputs | Notes / when used |
|---|---|---|---|---|
| Uncertain ELECTRE [73] | Outranking / kernel | Concordance–discordance indices; thresholds; weights | Outranking graph; kernel set | Non-compensatory; yields a dominance relation and best set. |
| Uncertain PROMETHEE [74] | Outranking flows | Preference functions on pairwise differences; weights | Positive/negative/net flows | Produces complete or partial ranking via flows. |

| Method | Primary task | Typical inputs | Primary outputs | Notes / when used |
|---|---|---|---|---|
| Fuzzy QUAL-IFLEX [75] | Out-ranking (permutation-based ranking) | Pairwise (possibly fuzzy) preference relations per criterion; criterion weights (optional) | Best ordering (ranking) maximizing concordance | Evaluates permutations; aggregates (fuzzy) concordance of pairwise preferences to select the most consistent ranking. |
| Fuzzy ORESTE [76] | Out-ranking (ordinal-distance ranking) | Ordinal ranks / preference intensities (possibly fuzzy); distance/aggregation rule | Ranking (often robust under weak data) | Uses (fuzzy) ordinal information and distance-based aggregation when precise cardinal evaluations are unavailable. |
| **G. Rule induction / learning / evidence / sequential decisions** | | | | |
| Fuzzy DRSA | Dominance-based rules | Ordered criteria; dominance relations; fuzzy boundaries | Decision rules; class approximations | Produces if–then rules and reduct-like structures. |
| Fuzzy Decision Tree [77] | Predictive model | Training data; fuzzy splits / fuzzy partitions | Tree model; predictions | Learning-oriented; can support decision recommendation. |
| Fuzzy Evidential Reasoning | Evidence aggregation | Belief degrees / evidences (often DS-like) + weights | Aggregated belief / utility; ranking | Combines multiple uncertain evidences systematically. |
| Fuzzy Markov Decision Process | Sequential control | States, actions, (fuzzy) transitions/rewards | Optimal policy; value function | Multi-stage decision under uncertainty (dynamic programming). |

# Chapter 2

# Preliminaries

This chapter collects the basic notation and background used throughout the book.

## 2.1 Fuzzy Set

Fuzzy set theory extends the classical notion of a subset by allowing graded membership, quantified by a value in $[0, 1]$ [1, 78, 79]. The core definition is recalled next.

**Definition 2.1.1** (Fuzzy set). [1] Let $X$ be a nonempty set. A *fuzzy set $A$* on $X$ is specified by a membership function

$$\mu_A : X \to [0, 1].$$

Equivalently, one may write

$$A = \{(x, \mu_A(x)) \mid x \in X\},$$

where $\mu_A(x)$ indicates the degree to which $x$ belongs to $A$.

A particularly common numeric object in fuzzy decision analysis is the *triangular fuzzy number* (TFN) [80, 81]. It models an imprecise quantity through a simple piecewise-linear membership profile.

**Definition 2.1.2** (Triangular fuzzy number (TFN)). [82, 83] Let $X$ be a universe. A *triangular fuzzy number* (TFN) $\tilde{x}$ is a fuzzy set on $\mathbb{R}$

whose membership function $\mu_{\tilde{x}} : \mathbb{R} \to [0, 1]$ is determined by three real parameters

$$\tilde{x} = (l_x, m_x, u_x) \in \mathbb{R}^3, \qquad l_x \leq m_x \leq u_x,$$

via

$$\mu_{\tilde{x}}(t) = \begin{cases} 0, & t < l_x, \\ \dfrac{t - l_x}{m_x - l_x}, & l_x \leq t \leq m_x, \quad (m_x > l_x), \\ \dfrac{u_x - t}{u_x - m_x}, & m_x \leq t \leq u_x, \quad (u_x > m_x), \\ 0, & t > u_x. \end{cases}$$

If $l_x = m_x$ (resp. $m_x = u_x$), the corresponding middle expression is interpreted in the limiting sense, so that $\mu_{\tilde{x}}(m_x) = 1$ and the membership curve remains triangular. If, in addition, $0 < l_x \leq m_x \leq u_x$, then $\tilde{x}$ is called a *positive* TFN.

Here $m_x$ represents the modal (most plausible) value, while $l_x$ and $u_x$ serve as lower and upper bounds.

## 2.2 Intuitionistic fuzzy set

Intuitionistic fuzzy sets enrich fuzzy sets by recording both membership and non-membership information, leaving room for an explicit hesitation component [2]. A standard formulation is as follows.

**Definition 2.2.1** (Intuitionistic fuzzy set). [84] Let $E$ be a nonempty set. An *intuitionistic fuzzy set* (IFS) $A$ on $E$ is given by

$$A = \big\{ \langle x, \mu_A(x), \nu_A(x) \rangle : x \in E \big\},$$

where

$$\mu_A, \nu_A : E \longrightarrow [0, 1]$$

are, respectively, the membership and non-membership functions, and for each $x \in E$,

$$0 \leq \mu_A(x) + \nu_A(x) \leq 1.$$

The remaining part,

$$\pi_A(x) := 1 - \mu_A(x) - \nu_A(x),$$

is called the *hesitation degree* at $x$.

The classical fuzzy-set situation is recovered when $\nu_A(x) = 1 - \mu_A(x)$ for all $x \in E$, equivalently $\pi_A(x) = 0$ for every $x$.

As extensions of *intuitionistic fuzzy sets*, several related concepts are known, such as the following:

- Generalized intuitionistic fuzzy sets [85]: Generalized intuitionistic fuzzy sets assign each element membership and nonmembership degrees in $[0, 1]$, constrained by $\min(\mu, \nu) \leq 0.5$, extending Atanassov's $\mu + \nu \leq 1$ condition for richer, flexible uncertainty modelling.

- Cubic intuitionistic fuzzy sets [86, 87]: Cubic intuitionistic fuzzy sets model each element using an interval valued intuitionistic pair and a crisp intuitionistic pair, preserving membership and nonmembership information simultaneously fully.

- Bipolar intuitionistic fuzzy sets [88, 89]: Bipolar intuitionistic fuzzy sets assign each element positive membership and nonmembership in $[0, 1]$ and negative membership and nonmembership in $[-1, 0]$, satisfying corresponding sum bounds simultaneously.

- Complex intuitionistic fuzzy sets [90, 91]: Complex intuitionistic fuzzy sets assign complex valued membership and nonmembership grades; membership, nonmembership, and their sum remain within the unit circle or unit square always.

- Interval-valued intuitionistic fuzzy sets [92, 93]: Interval-valued intuitionistic fuzzy sets represent each element by membership and nonmembership intervals in $[0, 1]$, with upper bounds satisfying $\overline{\mu} + \overline{\nu} \leq 1$, capturing imprecision explicitly.

## 2.3 Neutrosophic Set

Neutrosophic sets represent uncertainty by assigning three (in general independent) degrees to each element: truth $T$, indeterminacy $I$, and falsity $F$, typically taken in $[0, 1]$ [4, 14, 15, 94]. This explicit indeterminacy component provides a flexible generalization of both fuzzy sets and intuitionistic fuzzy sets.

**Definition 2.3.1** (Neutrosophic set). [95, 96] Let $X$ be a nonempty set. A *neutrosophic set* (NS) $A$ on $X$ is described by three functions

$$T_A : X \to [0, 1], \qquad I_A : X \to [0, 1], \qquad F_A : X \to [0, 1],$$

where, for each $x \in X$, the values $T_A(x)$, $I_A(x)$, and $F_A(x)$ quantify the degrees of truth, indeterminacy, and falsity of the statement "$x \in A$", respectively. They satisfy

$$0 \le T_A(x) + I_A(x) + F_A(x) \le 3.$$

For reference, an illustrative diagram of a Neutrosophic Set is presented in Figure 2.1.

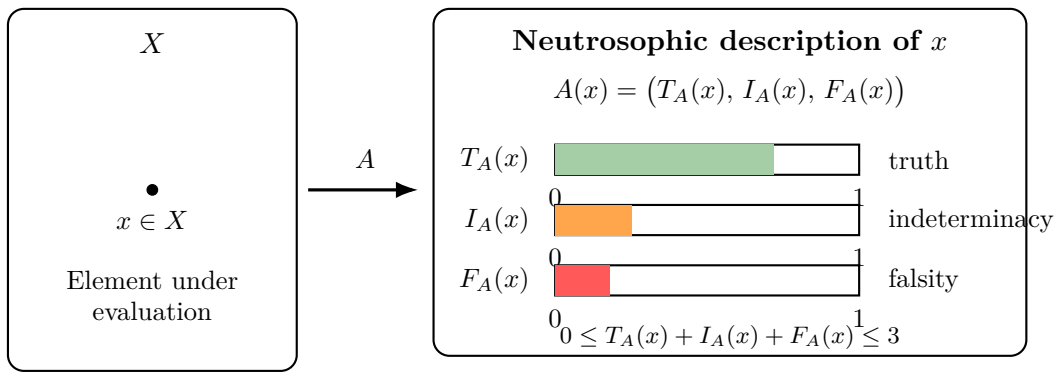

Figure 2.1: Conceptual illustration of a neutrosophic set. For each $x \in X$, the membership status is described by the triple $\big(T_A(x), I_A(x), F_A(x)\big)$ representing truth, indeterminacy, and falsity degrees, respectively.

As extensions of *neutrosophic sets*, several related concepts are known, such as the following:

- **Cubic Neutrosophic Sets [97, 98]:** Represent each element by both an interval neutrosophic triple $(T, I, F)$ and a single-valued triple, capturing range uncertainty plus a representative point simultaneously for decision tasks.

- **Bipolar Neutrosophic Sets [99, 100]:** Assign two neutrosophic triples per element: positive (supporting) and negative (opposing) truth–indeterminacy–falsity, modelling pros and cons with uncertainty on both sides in evaluations.

- **$m$-polar Neutrosophic Sets [101]:** Generalize bipolar to $m$ poles; each element has $m$ neutrosophic triples, one per perspective/agent/state, enabling multi-attitude assessments with indeterminacy and inconsistency across criteria sets too.

- **Interval-valued Neutrosophic Sets [102, 103]:** Replace truth, indeterminacy, and falsity degrees with intervals in $[0, 1]$, allowing bounded uncertainty about each component instead of single numbers during aggregation and ranking steps.

- **SuperHyperNeutrosophic Sets [104–106]:** Extends neutrosophic sets to superhyperstructures, assigning truth, indeterminacy, and falsity degrees to higher-order, nested entities and relations.

- **Complex Neutrosophic Sets [107, 108]:** Use complex-valued truth, indeterminacy, and falsity grades (typically magnitude and phase), representing oscillatory or periodic information while retaining neutrosophic separation of components for signal contexts.

- **Neutrosophic Offsets [109–111]:** Neutrosophic offsets extend neutrosophic sets by allowing truth, indeterminacy, or falsity values to fall below 0 or exceed 1 independently. As subclasses of neutrosophic offsets, concepts such as neutrosophic oversets [112, 113] and neutrosophic undersets [3, 114] are also known.

- **Spherical Neutrosophic Sets [115,116]:** Constrain single-valued degrees so that $T$, $I$, and $F$ lie in $[0, 1]$ and satisfy a spherical norm bound, e.g., $T^2 + I^2 + F^2 \leq 1$, ensuring feasible triples for decision models. As a further extension, n-HyperSpherical Neutrosophic Sets are also known [117–119].

## 2.4 Plithogenic Set

Plithogenic set assigns multi-criteria membership vectors to elements, modulated by contradiction degrees between attribute values and dominance levels interactions globally [17, 120–122].

**Definition 2.4.1** (Plithogenic Set). [17, 120] Let $P$ be a nonempty universe of discourse, and let $v$ be a (fixed) attribute whose possible values form a nonempty set $Pv$. Fix dimensions $s, t \in \mathbb{N}$.

A *plithogenic set* on $(P, v, Pv)$ is a quintuple

$$PS = (P, v, Pv, pdf, pCF),$$

where

- $pdf : P \times Pv \longrightarrow [0,1]^s$ is the *degree of appurtenance function* (DAF); for $x \in P$ and $a \in Pv$, $pdf(x, a)$ is the (possibly vector–valued) membership degree of $x$ corresponding to the attribute value $a$;

- $pCF : Pv \times Pv \longrightarrow [0,1]^t$ is the *degree of contradiction function* (DCF), satisfying

$$pCF(a, a) = 0, \qquad pCF(a, b) = pCF(b, a) \quad \text{for all } a, b \in Pv.$$

In plithogenic theory, a (typically fixed) *dominant attribute value* $a^* \in Pv$ is chosen, and set–theoretic operations (such as union and intersection) are defined by combining the appurtenance degrees $pdf$ with the contradiction degrees $pCF(\cdot, a^*)$ in order to model interaction and opposition between different attribute values.

Plithogenic sets are known for their ability to generalize a wide variety of concepts. If needed, see [20] for further details.

## 2.5 Rough Set

Rough set theory models imprecision by approximating a target subset through a *certain* part (lower approximation) and a *possible* part (upper approximation), constructed from an indiscernibility relation [123–126]. The classical Pawlak approximations are recalled below.

**Definition 2.5.1** (Rough set approximations). [127] Let $X$ be a nonempty universe, and let $R \subseteq X \times X$ be an equivalence (indiscernibility) relation. For $x \in X$, write the equivalence class of $x$ as

$$[x]_R := \{ y \in X \mid (x, y) \in R \}.$$

Given any subset $U \subseteq X$, define:

1. *Lower approximation*:
$$\underline{U} := \{ x \in X \mid [x]_R \subseteq U \}.$$

   Elements of $\underline{U}$ are those that belong to $U$ with certainty (their entire class is contained in $U$).

2. *Upper approximation*:
$$\overline{U} := \{ x \in X \mid [x]_R \cap U \neq \varnothing \}.$$

   Elements of $\overline{U}$ are those that may belong to $U$ (their class meets $U$).

The pair $(\underline{U}, \overline{U})$ is the rough-set representation of $U$, and it always satisfies

$$\underline{U} \subseteq U \subseteq \overline{U}.$$

As extensions of rough set theory, various concepts have been studied, such as probabilistic rough sets [128, 129] and granular rough sets [130, 131]. If needed, see [132] for further details.

## 2.6 Soft set

Soft sets provide a parameter-based description of uncertainty: each parameter selects a subset of the universe, and the family of all such selections forms the model. This framework was introduced by Molodtsov (1999) and has been widely used in decision problems [133, 134].

**Definition 2.6.1** (Soft set). [134] Let $U$ be a universe set and let $E$ be a set of parameters. Take $A \subseteq E$ and denote by $\mathcal{P}(U)$ the power set of $U$. A pair $(F, A)$ is called a *soft set* over $U$ if

$$F : A \to \mathcal{P}(U).$$

For each parameter $\epsilon \in A$, the subset $F(\epsilon) \subseteq U$ is called the $\epsilon$-*approximation* of $(F, A)$. Thus, a soft set is a parameterized family of subsets of $U$.

As extensions of soft set theory, a wide variety of concepts have been proposed, such as HyperSoft Sets [135, 136], SuperHyperSoft Sets [137, 138], and TreeSoft Sets [139, 140]. If needed, see, for example, [141] for further details.

## 2.7 Uncertain set

An *uncertain set* associates with each element a degree taken from a chosen uncertainty model, thereby providing a unifying umbrella for fuzzy, intuitionistic fuzzy, neutrosophic, plithogenic, and related frameworks [20, 142].

**Definition 2.7.1** (Uncertain model). [142] Let $U$ denote the class of all *uncertain models*. Each $M \in U$ is determined by:

- a nonempty set $\mathrm{Dom}(M) \subseteq [0,1]^k$ of *admissible degree tuples* for some fixed integer $k \geq 1$; and

- model-specific algebraic or geometric constraints imposed on elements of $\mathrm{Dom}(M)$ (for example, $\mu + \nu \leq 1$ in the intuitionistic fuzzy setting, or $0 \leq T + I + F \leq 3$ in the neutrosophic setting).

Typical instances include:

- **Fuzzy model:** $\mathrm{Dom}(M) = [0,1]$;

- **Intuitionistic fuzzy model:** $\mathrm{Dom}(M) = \{(\mu, \nu) \in [0,1]^2 : \mu + \nu \leq 1\}$;

- **Neutrosophic model:** $\mathrm{Dom}(M) = \{(T, I, F) \in [0,1]^3 : 0 \leq T + I + F \leq 3\}$;

- **Plithogenic model,** and many further extensions.

**Definition 2.7.2** (Uncertain set (U-set)). [142] Let $X$ be a nonempty universe, and fix an uncertain model $M$ with degree-domain $\mathrm{Dom}(M) \subseteq [0,1]^k$. An *uncertain set of type $M$* (briefly, a *U-set*) on $X$ is a pair

$$\mathcal{U} = (X, \mu_M),$$

where

$$\mu_M : X \longrightarrow \mathrm{Dom}(M)$$

is the *uncertainty-degree function* (membership map) of $\mathcal{U}$. For $x \in X$, the value $\mu_M(x) \in \mathrm{Dom}(M)$ encodes the degree(s) to which $x$ belongs to $\mathcal{U}$, as prescribed by the model $M$.

As noted in the remark, various generalizations are possible. For reference, Table 2.1 presents a catalogue of uncertainty-set families (U-Sets) organized by the dimension $k$ of the degree-domain $\mathrm{Dom}(M) \subseteq [0,1]^k$ (cf. [20]).

## 2.8 Fuzzy Graph

A fuzzy graph assigns each vertex and edge a membership degree in $[0,1]$, representing uncertain connectivity and relationship strength as graded, rather than binary, links [143]. As related concepts, Fuzzy Digraph [214], Fuzzy HyperGraph [143], and Fuzzy SuperHyperGraph [215] are also known.

**Definition 2.8.1** (Fuzzy graph). [79] A *fuzzy graph* on a vertex set $V$ is a pair $G = (\sigma, \mu)$ consisting of:

Table 2.1:  A catalogue of uncertainty-set families (U-Sets) by the dimension $k$ of the degree-domain $\mathrm{Dom}(M) \subseteq [0,1]^k$ [20].

| $k$ | note | Representative U-Set model(s) whose degree-domain is a subset of $[0,1]^k$ |
|---|---|---|
| 1 | | Fuzzy Set [1,143]; N-Fuzzy Set [144–146] Shadowed Set [147–149] |
| 2 | | Intuitionistic Fuzzy Set [2,150]; Vague Set [5,151]; Bipolar Fuzzy Set (two-component description) [152,153]; Pythagorean Fuzzy Set [154,155]; Fermatean fuzzy Set [156,157]; Variable Fuzzy Set [158–160]; Paraconsistent Fuzzy Set [161,162]; Bifuzzy Set [163,164] |
| 3 | | Single-Valued Neutrosophic Set [94,96]; Picture Fuzzy Set [7,165]; Ternary Fuzzy Set [166]; Hesitant Fuzzy Set [6,167]; Spherical Fuzzy Set [168,169]; Tripolar Fuzzy Set (three-component formalisms) [170–172]; Neutrosophic Vague Set [173,174] |
| 4 | | Quadripartitioned Neutrosophic Set [8,175]; Double-Valued Neutrosophic Set [176,177]; Dual Hesitant Fuzzy Set [178,179]; Ambiguous Set [180–182]; Turiyam Neutrosophic Set [183–186] |
| 5 | | Pentapartitioned Neutrosophic Set [187–189]; Triple-Valued Neutrosophic Set [190–193] |
| 6 | | Hexapartitioned Neutrosophic Set [194]; Bipolar Neutrosophic Set [99,195]; Bipolar Picture Fuzzy Set [196,197]; Quadruple-Valued Neutrosophic Set [192,198] |
| 7 | | Heptapartitioned Neutrosophic Set [199–201]; Quintuple-Valued Neutrosophic Set [192,202,203] |
| 8 | | Octapartitioned Neutrosophic Set [194]; Bipolar Quadripartitioned Neutrosophic Set [204,205]; Bipolar Double-valued Neutrosophic Set |
| 9 | | Nonapartitioned Neutrosophic Set [194] |
| $n$ | $(n \geq 1)$ | Multi-valued (Fuzzy) Sets [206]; MultiFuzzy Set [207]; $n$-Refined Fuzzy Set [208,209] |
| $2n$ | $(n \geq 1)$ | $n$-Refined Intuitionistic Fuzzy Set [209]; Multi-Intuitionistic Fuzzy Set [207] |
| $3n$ | $(n \geq 1)$ | $n$-Refined Neutrosophic Set [209,210]; Multi-Neutrosophic Set [207,211,212] |

**Reading guide.** In the U-Set scheme [142], each model $M$ is specified by a degree-domain $\mathrm{Dom}(M) \subseteq [0,1]^k$ and a membership map $\mu_M : X \to \mathrm{Dom}(M)$. The table groups representative families by the ambient dimension $k$ (i.e., how many numerical components are stored per element).
(a) A widely cited viewpoint is that neutrosophic sets provide a unifying umbrella covering several earlier multi-component fuzzy models (and their generalizations); see [16].
(b) Ambiguous sets are commonly presented as subclasses of certain four-component neutrosophic families; see [8,175,182].
(c) Turiyam neutrosophic sets are reported as subclasses of quadripartitioned neutrosophic sets; see [213].

- A vertex membership function $\sigma : V \to [0,1]$, where $\sigma(x)$ gives the degree to which $x \in V$ belongs to the graph.

- An edge membership function $\mu : V \times V \to [0,1]$, which is a fuzzy relation on $\sigma$, satisfying

$$\mu(x,y) \ \leq \ \sigma(x) \wedge \sigma(y), \quad \forall\, x,y \in V,$$

where $\wedge$ denotes the minimum operator.

The associated *crisp graph* $G^* = (\sigma^*, \mu^*)$ is determined by

$$\sigma^* = \{\, x \in V \mid \sigma(x) > 0 \,\}, \qquad \mu^* = \{\, (x,y) \in V \times V \mid \mu(x,y) > 0 \,\}.$$

A *fuzzy subgraph* $H = (\sigma', \mu')$ of $G$ is obtained by choosing a subset $X \subseteq V$ and defining

- a restricted vertex membership $\sigma' : X \to [0,1]$,

- an edge membership $\mu' : X \times X \to [0,1]$ such that

$$\mu'(x,y) \ \leq \ \sigma'(x) \wedge \sigma'(y), \quad \forall\, x,y \in X.$$

A wide variety of extensions of fuzzy graphs have been studied [4, 216, 217]. These can be represented within the framework of uncertain graphs, which extend uncertain sets to graph structures. We now state the uncertain graph-theoretic notions.

**Definition 2.8.2** (Uncertain graph). [218] Let $G = (V, E)$ be a finite, undirected, loopless graph, and let $M$ be an uncertain model with degree-domain $\mathrm{Dom}(M)$. An *uncertain graph of type $M$* is a triple

$$\mathcal{G}_M = (V, E, \mu_M),$$

where

$$\mu_M : V \cup E \longrightarrow \mathrm{Dom}(M)$$

assigns an uncertainty degree in $\mathrm{Dom}(M)$ to each vertex $v \in V$ and each edge $e \in E$. Optionally, one may impose model-dependent consistency relations between vertex- and edge-degrees (e.g., bounding $\mu_M(e)$ in terms of $\mu_M(u)$ and $\mu_M(v)$ for $e = \{u, v\}$ in fuzzy or intuitionistic fuzzy settings), but such constraints are dictated by the chosen model $M$ and are not fixed at the level of this general definition.

For convenience, Table 2.2 lists representative uncertainty-graph families, organized by the dimension $k$ of the degree-domain $\mathrm{Dom}(M) \subseteq [0, 1]^k$ (cf. [20, 218]).

Table 2.2: A catalogue of uncertainty-graph families (uncertain graphs) by the dimension $k$ of the degree-domain $\mathrm{Dom}(M) \subseteq [0, 1]^k$ (cf. [218]).

| $k$ | Representative uncertainty-graph type(s) $\mathcal{G}_M = (V, E, \mu_M)$ with $\mu_M : V \cup E \to \mathrm{Dom}(M) \subseteq [0, 1]^k$ |
|---|---|
| 1 | Fuzzy graph; $N$-graph [219]; shadowed-graph variants [220] |
| 2 | Intuitionistic fuzzy graph [221, 222]; vague graph [223]; bipolar fuzzy graph [224, 225]; intuitionistic evidence graph; variable fuzzy graph; paraconsistent fuzzy graph; bifuzzy graph [226, 227] |
| 3 | Neutrosophic graph [4](a); hesitant fuzzy graph [228, 229]; tripolar fuzzy graph; three-way fuzzy graph; picture fuzzy graph [230, 231]; spherical fuzzy graph [168, 232]; inconsistent intuitionistic fuzzy graph; ternary fuzzy / neutrosophic-fuzzy graph; neutrosophic vague graph |
| 4 | Quadripartitioned neutrosophic graph [233, 234]; double-valued neutrosophic graph [176]; dual hesitant fuzzy graph [235]; ambiguous graph(b); local-neutrosophic graph; support-neutrosophic graph; turiyam neutrosophic graph [185, 236](c) |
| 5 | Pentapartitioned neutrosophic graph [237]; triple-valued neutrosophic graph [191] |
| 6 | Hexapartitioned neutrosophic graph; quadruple-valued neutrosophic graph [191] |
| 7 | Heptapartitioned neutrosophic graph [238]; quintuple-valued neutrosophic graph [191] |
| 8 | Octapartitioned neutrosophic graph |
| 9 | Nonapartitioned neutrosophic graph |
| $n$ | $n$-refined fuzzy graph; multi-valued (fuzzy) graphs; multi-fuzzy graphs [239] |
| $2n$ | $n$-refined intuitionistic fuzzy graph; multi-intuitionistic fuzzy graphs |
| $3n$ | $n$-refined neutrosophic graph [240]; multi-neutrosophic graphs |

(a) Neutrosophic graph models are often treated as broad frameworks that can specialize to many degree-based graph formalisms under suitable constraints.
(b) Ambiguous-graph models are commonly presented as subclasses of certain quadripartitioned and also double-valued neutrosophic graph models.
(c) Turiyam neutrosophic graphs are reported as subclasses of certain quadripartitioned neutrosophic graph models.

## 2.9 Uncertain decision-making (UDM)

Using uncertain sets, one can define *uncertain decision-making* (UDM). The definition is given below.

**Definition 2.9.1** (Uncertain decision-making (UDM) of type $M$). Let $\mathcal{A} = \{A_1, \ldots, A_m\}$ be a finite set of alternatives and $\mathcal{C} = \{C_1, \ldots, C_n\}$ a finite set of criteria. Fix an *uncertain model $M$* with degree-domain $\mathrm{Dom}(M) \subseteq [0, 1]^k$ (cf. [20, 142]).

An *uncertain decision-making instance of type $M$* is the data

$$\mathfrak{D}_M = \big( \mathcal{A}, \mathcal{C}, X_M, w_M, \mathrm{Agg}_M, \mathrm{Score}_M \big),$$

where:

(i) $X_M = (\mu_{ij}) \in \mathrm{Dom}(M)^{m \times n}$ is an *uncertain evaluation matrix*, where $\mu_{ij} \in \mathrm{Dom}(M)$ encodes the uncertain assessment of $A_i$ under $C_j$;

(ii) $w_M = (\omega_1, \ldots, \omega_n) \in \mathrm{Dom}(M)^n$ (or, optionally, $w \in [0,1]^n$ with $\sum_j w_j = 1$) is an *uncertain criterion-importance profile*;

(iii) $\mathrm{Agg}_M$ is a model-dependent aggregation operator which assigns to each alternative $A_i$ an overall uncertain score

$$s_i := \mathrm{Agg}_M\big((\mu_{i1}, \ldots, \mu_{in}), (\omega_1, \ldots, \omega_n)\big) \in \mathrm{Dom}(M);$$

(iv) $\mathrm{Score}_M : \mathrm{Dom}(M) \to \mathbb{R}$ is a *ranking functional* (model-dependent score/defuzzification) inducing a preorder on $\mathcal{A}$ by

$$A_i \succeq_M A_j \iff \mathrm{Score}_M(s_i) \geq \mathrm{Score}_M(s_j).$$

A *solution* of $\mathfrak{D}_M$ is any alternative

$$A^\star \in \arg\max_{A_i \in \mathcal{A}} \mathrm{Score}_M(s_i),$$

together with the ranking induced by $\mathrm{Score}_M$.

**Remark 2.9.2** (Specializations). If $M$ is the fuzzy model $\mathrm{Dom}(M) = [0,1]$, then Definition 2.9.1 recovers a standard fuzzy decision-making setting. If $M$ is intuitionistic fuzzy, neutrosophic, plithogenic, etc., then one obtains the corresponding multi-component uncertain decision-making setting by changing $\mathrm{Dom}(M)$ and the operators $(\mathrm{Agg}_M, \mathrm{Score}_M)$.

**Theorem 2.9.3** (Well-definedness of uncertain decision-making of type $M$). *Let $\mathcal{A} = \{A_1, \ldots, A_m\}$ and $\mathcal{C} = \{C_1, \ldots, C_n\}$ be finite. Fix an uncertain model $M$ with degree-domain $\mathrm{Dom}(M) \subseteq [0,1]^k$. Assume:*

(A1) *the aggregation operator is a total map*

$$\mathrm{Agg}_M : \mathrm{Dom}(M)^n \times \mathrm{Dom}(M)^n \longrightarrow \mathrm{Dom}(M);$$

(A2) *the ranking functional is a total map*

$$\mathrm{Score}_M : \mathrm{Dom}(M) \longrightarrow \mathbb{R}.$$

*Then, for every UDM instance*

$$\mathfrak{D}_M = \big( \mathcal{A}, \mathcal{C}, X_M, w_M, \mathrm{Agg}_M, \mathrm{Score}_M \big)$$

*with* $\quad X_M = (\mu_{ij}) \in \mathrm{Dom}(M)^{m \times n}, \quad w_M = (\omega_1, \ldots, \omega_n) \in \mathrm{Dom}(M)^n,$

*the following objects are well-defined:*

(i) *the aggregated uncertain scores* $s_i \in \mathrm{Dom}(M)$, $i = 1, \ldots, m$;

(ii) *the induced preference relation* $\succeq_M$ *on* $\mathcal{A}$ *given by*

$$A_i \succeq_M A_j \iff \mathrm{Score}_M(s_i) \geq \mathrm{Score}_M(s_j);$$

(iii) *the solution set* $\arg\max_{A_i \in \mathcal{A}} \mathrm{Score}_M(s_i)$, *which is nonempty*.

*Moreover, $\succeq_M$ is a total preorder on $\mathcal{A}$ (reflexive, transitive, and total).*

*Proof.* **(i) Aggregated uncertain scores exist.** Fix $i \in \{1, \ldots, m\}$. Because $X_M \in \mathrm{Dom}(M)^{m \times n}$, the row $(\mu_{i1}, \ldots, \mu_{in})$ belongs to $\mathrm{Dom}(M)^n$. Also $w_M = (\omega_1, \ldots, \omega_n) \in \mathrm{Dom}(M)^n$. By assumption (A1), the value

$$s_i := \mathrm{Agg}_M\big((\mu_{i1}, \ldots, \mu_{in}), (\omega_1, \ldots, \omega_n)\big)$$

is defined and lies in $\mathrm{Dom}(M)$.

**(ii) Scores are real numbers.** By (A2), $\mathrm{Score}_M(s_i) \in \mathbb{R}$ is defined for each $i$.

**(iii) The relation $\succeq_M$ is well-defined and a total preorder.** Define $A_i \succeq_M A_j$ iff $\mathrm{Score}_M(s_i) \geq \mathrm{Score}_M(s_j)$. Since $\geq$ is a well-defined total preorder on $\mathbb{R}$, it follows immediately that:

- *Reflexive:* $\mathrm{Score}_M(s_i) \geq \mathrm{Score}_M(s_i)$ for all $i$, hence $A_i \succeq_M A_i$.

- *Transitive:* if $A_i \succeq_M A_j$ and $A_j \succeq_M A_\ell$, then $\mathrm{Score}_M(s_i) \geq \mathrm{Score}_M(s_j) \geq \mathrm{Score}_M(s_\ell)$, hence $A_i \succeq_M A_\ell$.

- *Total:* for any $i, j$, either $\mathrm{Score}_M(s_i) \geq \mathrm{Score}_M(s_j)$ or $\mathrm{Score}_M(s_j) \geq \mathrm{Score}_M(s_i)$, hence $A_i \succeq_M A_j$ or $A_j \succeq_M A_i$.

Therefore $\succeq_M$ is a total preorder on $\mathcal{A}$.

**(iv) A maximizer exists.** The set $\{\mathrm{Score}_M(s_i) : i = 1, \ldots, m\} \subset \mathbb{R}$ is finite, hence attains its maximum. Therefore,

$$\arg\max_{A_i \in \mathcal{A}} \mathrm{Score}_M(s_i) \neq \varnothing,$$

so at least one solution alternative $A^\star$ exists, and the solution set is well-defined. $\qquad \square$

For reference, Table 2.3 presents a catalogue of uncertainty-decision families (UDM).

Table 2.3: A catalogue of uncertainty-decision families (UDM) by the dimension $k$ of the degree-domain $\mathrm{Dom}(M) \subseteq [0,1]^k$ (cf. [20]).

| $k$ | note | Representative UDM family/families whose assessments take values in $\mathrm{Dom}(M) \subseteq [0,1]^k$ |
|---|---|---|
| 1 | | Fuzzy decision-making [241] / fuzzy MCDM and MADM (single membership degree); $N$-fuzzy decision-making (multi-membership collapsed to $k$=1 via aggregation); shadowed decision variants |
| 2 | | Intuitionistic fuzzy decision-making (membership/non-membership with constraint) [242, 243]; vague decision-making [244, 245]; bipolar/bi-component fuzzy decision-making [246, 247]; paraconsistent/bifuzzy decision-making |
| 3 | | Single-valued neutrosophic decision-making (truth/indeterminacy/falsity) [248, 249]; hesitant fuzzy decision-making [250, 251]; picture-fuzzy decision-making [252]; spherical-fuzzy decision-making [253,254]; tripolar fuzzy decision-making [255]; neutrosophic-vague decision-making [256] |
| 4 | | Quadripartitioned neutrosophic decision-making [257]; double-valued neutrosophic decision-making [258]; dual-hesitant decision-making [259,260]; ambiguous-decision-making; turiyam-neutrosophic decision-making [261] |
| 5 | | Pentapartitioned neutrosophic decision-making [262, 263]; triple-valued neutrosophic decision-making |
| 6 | | Hexapartitioned neutrosophic decision-making; quadruple-valued neutrosophic decision-making |
| 7 | | Heptapartitioned neutrosophic decision-making [199]; quintuple-valued neutrosophic decision-making |
| 8 | | Octapartitioned neutrosophic decision-making |
| 9 | | Nonapartitioned neutrosophic decision-making |
| $n$ | ($n \geq 1$) | $n$-refined fuzzy decision-making; multi-valued (fuzzy) decision-making; multi-fuzzy decision-making [264] |
| $2n$ | ($n \geq 1$) | $n$-refined intuitionistic fuzzy decision-making; multi-intuitionistic fuzzy decision-making |
| $3n$ | ($n \geq 1$) | $n$-refined neutrosophic decision-making [265–267]; multi-neutrosophic decision-making |

**Reading guide.** This table mirrors the "U-Set catalogue" viewpoint: an uncertain decision family is characterized, at the type level, by the degree-domain $\mathrm{Dom}(M) \subseteq [0,1]^k$ used to encode assessments and/or weights, together with model-dependent aggregation and ranking operators ($\mathrm{Agg}_M$, $\mathrm{Score}_M$). Thus, changing $k$ changes how many numerical components are stored per assessment.

# Chapter 3

# Problem-level frameworks

In this section, we present fundamental decision-making frameworks.

## 3.1 Fuzzy Multiple-Criteria Decision Making (Fuzzy MCDM)

Multiple-Criteria Decision Making evaluates alternatives across several criteria, assigns weights, aggregates scores or outranking relations, and ranks choices systematically overall [268, 269]. Fuzzy MCDM models ratings and weights as fuzzy numbers or linguistic terms, propagates uncertainty through aggregation, then defuzzifies rankings robustly [270, 271].

**Definition 3.1.1** (Fuzzy MCDM problem (evaluation-and-ranking form)). [270,271] Let $\mathcal{A} = \{A_1, \ldots, A_m\}$ be a finite set of alternatives and $\mathcal{C} = \{C_1, \ldots, C_k\}$ a finite set of criteria. Let $\mathsf{FN}(\mathbb{R})$ denote a chosen class of fuzzy numbers on $\mathbb{R}$ (e.g. triangular or trapezoidal), and let $\mathsf{FN}([0,1])$ denote fuzzy numbers supported in $[0,1]$.

A *fuzzy multiple-criteria decision-making (fuzzy MCDM) instance* is the data

$$\mathfrak{D} = \big(\mathcal{A}, \mathcal{C}, \widetilde{X}, \widetilde{w}, \mathrm{Agg}, \mathrm{Score}\big),$$

where:

(i) $\widetilde{X} = (\tilde{x}_{it}) \in \mathsf{FN}(\mathbb{R})^{m \times k}$ is the *fuzzy decision matrix*, where $\tilde{x}_{it}$ represents the (possibly linguistic) performance of $A_i$ under criterion $C_t$;

(ii) $\widetilde{w} = (\tilde{w}_1, \ldots, \tilde{w}_k) \in \mathsf{FN}([0,1])^k$ is the *fuzzy weight vector* with $\tilde{w}_t$ describing the uncertain importance of criterion $C_t$;

(iii) Agg is an *aggregation operator* producing, for each alternative $A_i$, an overall fuzzy score

$$\tilde{s}_i := \mathrm{Agg}\big((\tilde{x}_{i1}, \ldots, \tilde{x}_{ik}), (\tilde{w}_1, \ldots, \tilde{w}_k)\big) \in \mathsf{FN}(\mathbb{R});$$

a common choice is the fuzzy weighted average

$$\tilde{s}_i = \frac{\sum_{t=1}^k \tilde{w}_t \otimes \tilde{x}_{it}}{\sum_{t=1}^k \tilde{w}_t},$$

where $\otimes$ and $/$ are fuzzy arithmetic operations in $\mathsf{FN}(\mathbb{R})$;

(iv) Score : $\mathsf{FN}(\mathbb{R}) \to \mathbb{R}$ is a *ranking functional* (defuzzification/score), and the induced preorder on $\mathcal{A}$ is

$$A_i \succeq A_j \quad \Longleftrightarrow \quad \mathrm{Score}(\tilde{s}_i) \geq \mathrm{Score}(\tilde{s}_j).$$

A *solution* of the fuzzy MCDM instance is any alternative $A^\star \in \mathcal{A}$ satisfying

$$A^\star \in \arg\max_{A_i \in \mathcal{A}} \mathrm{Score}(\tilde{s}_i),$$

together with the (total preorder) ranking of $\mathcal{A}$ induced by $\mathrm{Score}(\tilde{s}_i)$.

**Remark 3.1.2** (Ideal/anti-ideal (distance-to-reference) fuzzy MCDM)**.** Many fuzzy MCDM methods replace Score by a distance-to-reference ranking. Given fuzzy positive and negative ideal points (or solutions) $I^+, I^-$ in the criterion space, define for each $A_i$ two distances $d_i^+ = d(\tilde{\mathbf{x}}_i, I^+)$ and $d_i^- = d(\tilde{\mathbf{x}}_i, I^-)$ (using a chosen fuzzy distance), and rank by a closeness/approximation index, e.g.

$$\Gamma_i := \frac{d_i^-}{d_i^+ + d_i^-} \in [0,1], \qquad A_i \succeq A_j \iff \Gamma_i \geq \Gamma_j.$$

This produces a compromise ranking relative to the ideal and anti-ideal references.

**Definition 3.1.3** (Uncertain Multiple-Criteria Decision Making (UMCDM))**.** Let $\mathcal{A} = \{A_1, \dots, A_m\}$ be a finite nonempty set of alternatives and $\mathcal{C} = \{C_1, \dots, C_n\}$ a finite nonempty set of criteria. Fix an uncertain model $M$.

An *UMCDM instance of type $M$* is the data

$$\mathfrak{D}_M = \big(\mathcal{A}, \mathcal{C}, X_M, w, \mathrm{Agg}_M, \mathrm{Score}_M\big),$$

where:

(i) $X_M = (\mu_{ij}) \in \mathrm{Dom}(M)^{m \times n}$ is the *uncertain evaluation matrix*, where $\mu_{ij} \in \mathrm{Dom}(M)$ encodes the (uncertain) evaluation of $A_i$ w.r.t. criterion $C_j$;

(ii) $w = (w_1, \dots, w_n) \in [0,1]^n$ is a *criterion weight vector* with $\sum_{j=1}^n w_j = 1$ (allowing uncertain weights is possible by taking $w \in \mathrm{Dom}(M)^n$ instead);

(iii) $\mathrm{Agg}_M$ is an *aggregation operator* producing an overall $M$-degree for each alternative:

$$s_i := \mathrm{Agg}_M\big((\mu_{i1}, \dots, \mu_{in}), w\big) \in \mathrm{Dom}(M), \qquad i = 1, \dots, m;$$

(iv) $\mathrm{Score}_M : \mathrm{Dom}(M) \to \mathbb{R}$ is a *ranking functional* (score/defuzzification), and it induces a preference relation on $\mathcal{A}$ by

$$A_i \succeq_M A_k \quad \Longleftrightarrow \quad \mathrm{Score}_M(s_i) \geq \mathrm{Score}_M(s_k).$$

A *best alternative* (solution) is any

$$A^\star \in \arg\max_{1 \leq i \leq m} \mathrm{Score}_M(s_i).$$

**Theorem 3.1.4** (Uncertain-set structure and well-definedness of UMCDM)**.** *Let*

$$\mathfrak{D}_M = \big(\mathcal{A}, \mathcal{C}, X_M, w, \mathrm{Agg}_M, \mathrm{Score}_M\big)$$

*be an UMCDM instance as in Definition 3.1.3, where $\mathcal{A}, \mathcal{C}$ are finite nonempty and $M$ is an uncertain model.*

*Assume:*

(A1) $\mathrm{Agg}_M : \mathrm{Dom}(M)^n \times \Delta_n \to \mathrm{Dom}(M)$ *is a total mapping, where* $\Delta_n := \{w \in [0,1]^n : \sum_{j=1}^n w_j = 1\}$*;*

(A2) $\mathrm{Score}_M : \mathrm{Dom}(M) \to \mathbb{R}$ *is a total mapping.*

*Then:*

(i) (***Uncertain-set structure***) *The evaluation matrix $X_M = (\mu_{ij})$ defines an uncertain set of type $M$ on the product universe $\mathcal{A} \times \mathcal{C}$ via*

$$\mu : \mathcal{A} \times \mathcal{C} \to \mathrm{Dom}(M), \qquad \mu(A_i, C_j) := \mu_{ij}.$$

*Hence $\big(\mathcal{A} \times \mathcal{C}, \mu\big)$ is an uncertain set of type $M$. Moreover, the aggregated-score map $s : \mathcal{A} \to \mathrm{Dom}(M)$ defined by $s(A_i) := s_i$ also yields an uncertain set $\big(\mathcal{A}, s\big)$ of type $M$.*

(ii) (***Well-defined ranking***) *The relation $\succeq_M$ on $\mathcal{A}$ induced by $\mathrm{Score}_M \circ s$ is a total preorder (reflexive, transitive, and total).*

(iii) (***Existence of a best alternative***) *The solution set $\arg\max_{1 \le i \le m} \mathrm{Score}_M(s_i)$ is nonempty.*

*Proof.* **(i) Uncertain-set structure.** Because $X_M \in \mathrm{Dom}(M)^{m \times n}$, each entry $\mu_{ij}$ lies in $\mathrm{Dom}(M)$. Define $\mu(A_i, C_j) := \mu_{ij}$ on $\mathcal{A} \times \mathcal{C}$. This is a well-defined function $\mu : \mathcal{A} \times \mathcal{C} \to \mathrm{Dom}(M)$, so $\big(\mathcal{A} \times \mathcal{C}, \mu\big)$ is an uncertain set of type $M$.

Next, fix $i \in \{1, \ldots, m\}$. Since $(\mu_{i1}, \ldots, \mu_{in}) \in \mathrm{Dom}(M)^n$ and $w \in \Delta_n$, assumption (A1) implies that

$$s_i = \mathrm{Agg}_M\big((\mu_{i1}, \ldots, \mu_{in}), w\big) \in \mathrm{Dom}(M)$$

exists. Hence the mapping $s : \mathcal{A} \to \mathrm{Dom}(M)$, $s(A_i) := s_i$, is well-defined and $\big(\mathcal{A}, s\big)$ is an uncertain set of type $M$.

**(ii) Well-defined ranking and total preorder.** By (A2), each $\mathrm{Score}_M(s_i) \in \mathbb{R}$ is defined. Define $A_i \succeq_M A_k$ iff $\mathrm{Score}_M(s_i) \ge \mathrm{Score}_M(s_k)$. Since $\ge$ on $\mathbb{R}$ is reflexive, transitive, and total, the induced relation $\succeq_M$ inherits these properties:

- Reflexive: $\mathrm{Score}_M(s_i) \ge \mathrm{Score}_M(s_i)$, hence $A_i \succeq_M A_i$.

- Transitive: if $A_i \succeq_M A_j$ and $A_j \succeq_M A_\ell$, then $\mathrm{Score}_M(s_i) \geq \mathrm{Score}_M(s_j) \geq \mathrm{Score}_M(s_\ell)$, hence $A_i \succeq_M A_\ell$.

- Total: for any $i, k$, either $\mathrm{Score}_M(s_i) \geq \mathrm{Score}_M(s_k)$ or $\mathrm{Score}_M(s_k) \geq \mathrm{Score}_M(s_i)$, hence $A_i \succeq_M A_k$ or $A_k \succeq_M A_i$.

Thus $\succeq_M$ is a total preorder.

**(iii) Existence of a best alternative.** The set $\{\mathrm{Score}_M(s_i) : i = 1, \ldots, m\} \subset \mathbb{R}$ is finite, hence it attains a maximum. Therefore $\arg\max_{1 \leq i \leq m} \mathrm{Score}_M(s_i) \neq \varnothing$, proving existence of at least one best alternative. □

As a reference, a catalogue of uncertainty-aware MCDM families organized by the dimension $k$ of the degree-domain is presented in Table 3.1.

Table 3.1: A catalogue of uncertainty-aware MCDM families by the dimension $k$ of the degree-domain $\mathrm{Dom}(M) \subseteq [0,1]^k$.

| $k$ | note | Representative uncertainty-aware MCDM model(s) whose degree-domain is a subset of $[0,1]^k$ |
|---|---|---|
| 1 | | Fuzzy MCDM. |
| 2 | | Intuitionistic Fuzzy MCDM [272, 273]; Bipolar Fuzzy MCDM [274, 275]; Pythagorean Fuzzy MCDM [276, 277]; Fermatean Fuzzy MCDM [278, 279] |
| 3 | | Hesitant Fuzzy MCDM [280, 281]; Picture Fuzzy MCDM [282, 283]; Spherical Fuzzy MCDM [284, 285]; (Single-valued) Neutrosophic MCDM [286, 287]. |
| 4 | | Quadripartitioned Neutrosophic MCDM [288, 289]. |
| $n$ | $(n \geq 1)$ | Plithogenic MCDM [290, 291] (vector-valued degrees over attribute values: for $\mathrm{Valset}(v) = \{a_1, \ldots, a_n\}$, $\mu(x) = (\mu(x, a_1), \ldots, \mu(x, a_n)) \in [0,1]^n$; augmented with a contradiction function $c : \mathrm{Valset}(v) \times \mathrm{Valset}(v) \to [0,1]$ on values). |

**Reading guide.** Here, $k$ counts the number of numerical components stored per evaluation (i.e., the ambient dimension of the degree-domain). For hesitant fuzzy information, the classification $k = 3$ can be realized by a standard vectorization (e.g., an envelope/summary map of a hesitant set of grades into three numbers in $[0,1]^3$). Plithogenic models naturally induce variable dimension $k = n$ because degrees are assigned per attribute–value and $|\mathrm{Valset}(v)|$ depends on the chosen value set; the contradiction map $c(\cdot, \cdot)$ is an additional structure on values.

In addition to uncertainty-related extensions, approaches such as Linguistic MCDM [292, 293], Behavioral MCDM [294], Rough MCDM [295, 296], Hybrid MCDM [297, 298], Soft MCDM [299, 300], Grey MCDM [301, 302], MCGDM [303, 304], Stratified Multi-Criteria Decision-Making (SMCDM) [305, 306], Z-Number Based MCDM [307, 308], Potentially All Pairwise RanKings of all possible Alternatives (PAPRIKA) [309, 310], Aggregated Indices Randomization Method (AIRM) [311, 312], Multi-Attribute Global Inference of Quality (MAGIQ) [313], UTA methods [314], Stochastic Multicriteria Acceptability Analysis (SMAA) [315], Conjoint Value Hierarchy (CVA) [316], and Large-scale GDM [317, 318] have also been studied, and further developments and extensions can be expected in the future.

## 3.2 Fuzzy MADM (Fuzzy Multi-attribute decision-making)

MADM evaluates a finite set of alternatives across multiple attributes, weights attributes, aggregates performances, and selects the best option [319, 320]. Fuzzy MADM expresses attribute ratings and weights as fuzzy numbers or linguistic terms, aggregates fuzzily, then defuzzifies to rank alternatives [321, 322].

**Definition 3.2.1** (Fuzzy MADM problem (discrete multi-attribute decision model)). [322, 323] Let

$$\mathcal{A} = \{A_1, \dots, A_m\}$$

be a finite set of alternatives and

$$\mathcal{C} = \{C_1, \dots, C_n\}$$

a finite set of attributes (criteria). A *fuzzy MADM instance* is specified by:

(1) a *fuzzy decision matrix*

$$\widetilde{R} = (\tilde{r}_{ij}) \in \mathsf{FN}(\mathbb{R})^{m \times n}, \qquad \tilde{r}_{ij} \text{ is the fuzzy performance rating of } A_i \text{ w.r.t. } C_j;$$

(2) a *fuzzy weight vector*

$$\widetilde{W} = (\tilde{w}_1, \dots, \tilde{w}_n) \in \mathsf{FN}(\mathbb{R}_{\geq 0})^n, \qquad \tilde{w}_j \text{ is the fuzzy importance of } C_j.$$

Here $\mathsf{FN}(\cdot)$ denotes a chosen class of fuzzy numbers (e.g. triangular/trapezoidal).

A *decision procedure* consists of two phases:

(P1) **(Aggregation)** For each alternative $A_i$, compute an aggregated fuzzy utility

$$\tilde{u}_i := \mathrm{Agg}(\tilde{r}_{i1}, \dots, \tilde{r}_{in}; \tilde{w}_1, \dots, \tilde{w}_n) \in \mathsf{FN}(\mathbb{R}),$$

where Agg is a specified fuzzy aggregation operator. A common choice is the fuzzy weighted average (FWA)

$$\tilde{u}_i = \frac{\bigoplus_{j=1}^{n} (\tilde{w}_j \otimes \tilde{r}_{ij})}{\bigoplus_{j=1}^{n} \tilde{w}_j},$$

with $\oplus, \otimes$ and $/$ interpreted in the adopted fuzzy arithmetic.

(P2) **(Ranking)** Rank alternatives by a fuzzy ranking rule

$$A_i \succeq A_k \iff \tilde{u}_i \preceq_F \tilde{u}_k,$$

where $\preceq_F$ is a fixed preorder on fuzzy numbers (e.g. induced by a defuzzification/score map Score : $\mathsf{FN}(\mathbb{R}) \to \mathbb{R}$ via $\tilde{u}_i \preceq_F \tilde{u}_k \iff \mathrm{Score}(\tilde{u}_i) \leq \mathrm{Score}(\tilde{u}_k)$).

A *(best) solution* is any $A^\star \in \mathcal{A}$ such that $A^\star \succeq A_i$ for all $A_i \in \mathcal{A}$.

We next consider Uncertain MADM, obtained by generalizing the above framework using Uncertain Sets.

**Definition 3.2.2** (Uncertain MADM (UMADM)). Let $\mathcal{A} = \{A_1, \dots, A_m\}$ be a finite nonempty set of alternatives and $\mathcal{C} = \{C_1, \dots, C_n\}$ a finite nonempty set of attributes (criteria). Fix an uncertain model $M$ with degree-domain $\mathrm{Dom}(M) \subseteq [0,1]^k$.

An *uncertain multi-attribute decision-making instance of type $M$* (UMADM instance) is the data

$$\mathfrak{D}_M = \big(\mathcal{A}, \mathcal{C}, R_M, w, \mathrm{Agg}_M, \mathrm{Score}_M\big),$$

where:

(i) $R_M = (\mu_{ij}) \in \mathrm{Dom}(M)^{m \times n}$ is the *uncertain decision matrix*, where $\mu_{ij} \in \mathrm{Dom}(M)$ encodes the uncertain evaluation of alternative $A_i$ under attribute $C_j$;

(ii) $w = (w_1, \ldots, w_n) \in [0,1]^n$ is a (crisp) attribute-weight vector with $\sum_{j=1}^n w_j = 1$ (optionally, one may allow uncertain weights $w \in \mathrm{Dom}(M)^n$);

(iii) $\mathrm{Agg}_M$ is an aggregation operator producing an overall $M$-degree for each alternative:

$$s_i := \mathrm{Agg}_M\big((\mu_{i1}, \ldots, \mu_{in}), w\big) \in \mathrm{Dom}(M), \qquad i = 1, \ldots, m;$$

(iv) $\mathrm{Score}_M : \mathrm{Dom}(M) \to \mathbb{R}$ is a ranking functional inducing the preference relation

$$A_i \succeq_M A_\ell \quad \Longleftrightarrow \quad \mathrm{Score}_M(s_i) \geq \mathrm{Score}_M(s_\ell).$$

A *best alternative* (solution) is any

$$A^\star \in \arg \max_{1 \leq i \leq m} \ \mathrm{Score}_M(s_i).$$

**Theorem 3.2.3** (Uncertain-set structure and well-definedness of UMADM)**.** *Let*

$$\mathfrak{D}_M = (\mathcal{A}, \mathcal{C}, R_M, w, \mathrm{Agg}_M, \mathrm{Score}_M)$$

*be a UMADM instance in the sense of Definition 3.2.2, with $\mathcal{A}, \mathcal{C}$ finite nonempty and $M$ an uncertain model.*

*Assume:*

(A1) $\mathrm{Agg}_M : \mathrm{Dom}(M)^n \times \Delta_n \to \mathrm{Dom}(M)$ *is a total mapping, where* $\Delta_n := \{w \in [0,1]^n : \sum_{j=1}^n w_j = 1\}$;

(A2) $\mathrm{Score}_M : \mathrm{Dom}(M) \to \mathbb{R}$ *is a total mapping.*

*Then:*

(i) (***Uncertain-set structure on the evaluation domain***) *The decision matrix $R_M = (\mu_{ij})$ induces an uncertain set of type $M$ on $\mathcal{A} \times \mathcal{C}$ via*

$$\mu : \mathcal{A} \times \mathcal{C} \to \mathrm{Dom}(M), \qquad \mu(A_i, C_j) := \mu_{ij}.$$

*Hence $(\mathcal{A} \times \mathcal{C}, \mu)$ is an uncertain set of type $M$.*

(ii) (***Uncertain-set structure on alternatives***) *The aggregated-score assignment $s : \mathcal{A} \to \mathrm{Dom}(M)$ defined by $s(A_i) := s_i$ is well-defined and yields an uncertain set $(\mathcal{A}, s)$ of type $M$.*

(iii) (***Well-defined preference and existence of a best alternative***) *The induced relation $\succeq_M$ on $\mathcal{A}$ is a total preorder (reflexive, transitive, total), and the solution set $\arg \max_{1 \leq i \leq m} \mathrm{Score}_M(s_i)$ is nonempty.*

*Proof.* **(i) Uncertain-set structure on $\mathcal{A} \times \mathcal{C}$.** Because $R_M \in \mathrm{Dom}(M)^{m \times n}$, each entry $\mu_{ij}$ belongs to $\mathrm{Dom}(M)$. Define $\mu(A_i, C_j) := \mu_{ij}$. This is a well-defined mapping $\mu : \mathcal{A} \times \mathcal{C} \to \mathrm{Dom}(M)$, hence $(\mathcal{A} \times \mathcal{C}, \mu)$ is an uncertain set of type $M$.

**(ii) Aggregated uncertain scores exist and form an uncertain set on $\mathcal{A}$.** Fix $i \in \{1, \ldots, m\}$. Since $(\mu_{i1}, \ldots, \mu_{in}) \in \mathrm{Dom}(M)^n$ and $w \in \Delta_n$, assumption (A1) implies that

$$s_i = \mathrm{Agg}_M\big((\mu_{i1}, \ldots, \mu_{in}), w\big) \in \mathrm{Dom}(M)$$

is defined. Therefore the mapping $s : \mathcal{A} \to \mathrm{Dom}(M)$, $s(A_i) := s_i$, is well-defined, and $(\mathcal{A}, s)$ is an uncertain set of type $M$.

**(iii) The induced preference $\succeq_M$ is well-defined and a total preorder.** By (A2), each value $\mathrm{Score}_M(s_i) \in \mathbb{R}$ exists. Define $A_i \succeq_M A_\ell$ iff $\mathrm{Score}_M(s_i) \geq \mathrm{Score}_M(s_\ell)$. Because $\geq$ on $\mathbb{R}$ is reflexive, transitive, and total, the induced relation $\succeq_M$ inherits these properties:

- Reflexive: $\mathrm{Score}_M(s_i) \geq \mathrm{Score}_M(s_i)$, hence $A_i \succeq_M A_i$.

- Transitive: if $A_i \succeq_M A_j$ and $A_j \succeq_M A_\ell$, then $\mathrm{Score}_M(s_i) \geq \mathrm{Score}_M(s_j) \geq \mathrm{Score}_M(s_\ell)$, hence $A_i \succeq_M A_\ell$.

- Total: for any $i, \ell$, either $\mathrm{Score}_M(s_i) \geq \mathrm{Score}_M(s_\ell)$ or $\mathrm{Score}_M(s_\ell) \geq \mathrm{Score}_M(s_i)$, hence $A_i \succeq_M A_\ell$ or $A_\ell \succeq_M A_i$.

Thus $\succeq_M$ is a total preorder.

**(iv) Existence of a best alternative.** The set $\{\mathrm{Score}_M(s_i) : i = 1, \ldots, m\} \subset \mathbb{R}$ is finite, hence attains a maximum. Therefore,

$$\arg \max_{1 \leq i \leq m} \mathrm{Score}_M(s_i) \neq \varnothing,$$

so at least one best alternative exists and the solution set is well-defined. $\qquad\square$

Examples of uncertainty-aware MADM families, grouped by the degree-domain dimension $k$ of $\mathrm{Dom}(M) \subseteq [0,1]^k$, are listed in Table 3.2.

As related concepts beyond uncertain MADM, Rough MADM [335], Soft MADM [336, 337], Multiple attribute group decision-making (MAGDM) [338, 339], Grey MADM [340], Z-number madm [341, 342], and Linguistic MADM [343, 344] are also known.

Table 3.2: Related uncertainty-aware MADM families (examples) grouped by the degree-domain dimension $k$ of $\mathrm{Dom}(M) \subseteq [0,1]^k$.

| $k$ | note | Representative uncertainty-aware MADM model(s) |
|---|---|---|
| 1 | | Fuzzy multi-attribute decision making. |
| 2 | | Intuitionistic Fuzzy multi-attribute decision making [324, 325]. |
| 3 | | Hesitant Fuzzy multi-attribute decision making [326, 327]; Spherical fuzzy multi-attribute decision making [328, 329]; Picture Fuzzy multi-attribute decision making [330, 331]; Neutrosophic multi-attribute decision making [332, 333]. |
| $n$ | $(n \geq 1)$ | Plithogenic multi-attribute decision making [334] (vector-valued degrees over attribute values, coupled with a contradiction function on values). |

**Note.** The $k = 3$ placement for hesitant fuzzy MADM follows the same convention as in Table 3.1, namely, representing hesitant information via a 3-component embedding/summary in $[0,1]^3$. Plithogenic MADM yields $k = n$ where $n = |\mathrm{Valset}(v)|$ depends on the chosen attribute value set.

## 3.3 Fuzzy Group-Decision Making

Group Decision-Making combines the preferences of multiple decision makers, negotiates trade-offs, measures consensus, and selects an alternative that reflects collective judgments [345, 346]. As a concept that may be viewed as roughly opposite to Group Decision-Making, personal decision-making [347, 348] is also known. Fuzzy Group-Decision Making represents opinions by fuzzy numbers or linguistic terms, aggregates them using fuzzy operators, evaluates consensus, and then ranks the alternatives [349, 350].

**Definition 3.3.1** (Fuzzy group decision making (consensus + selection framework)). [351] A *fuzzy group decision making* (FGDM) problem consists of:

$$X = \{x_1, \ldots, x_n\} \text{ (alternatives)}, \qquad E = \{e_1, \ldots, e_m\} \text{ (experts/decision makers)}.$$

Experts express their opinions in *fuzzy* form (often by linguistic terms encoded as fuzzy numbers), and their assessments are aggregated to obtain a collective decision; a consensus stage is typically applied before the final selection stage.

**(A) Multi-attribute evaluation data (typical MADM-type FGDM).** Let $C = \{c_1, \ldots, c_p\}$ be criteria. Fix a class $\mathsf{FN}(\mathbb{R})$ of fuzzy numbers (e.g. TFNs). Each expert $e_h$ provides:

$$\widetilde{R}^{(h)} = \big(\tilde{r}_{ij}^{(h)}\big) \in \mathsf{FN}(\mathbb{R})^{n \times p} \quad \text{(fuzzy ratings)},$$

$$\tilde{w}^{(h)} = (\tilde{w}_1^{(h)}, \ldots, \tilde{w}_p^{(h)}) \in \mathsf{FN}(\mathbb{R}_{\geq 0})^p \quad \text{(fuzzy criterion weights)}.$$

**(B) Group aggregation.** Choose aggregation operators

$$\mathrm{Agg}_R : \mathsf{FN}(\mathbb{R})^m \to \mathsf{FN}(\mathbb{R}),$$

$$\mathrm{Agg}_w : \mathsf{FN}(\mathbb{R}_{\geq 0})^m \to \mathsf{FN}(\mathbb{R}_{\geq 0}),$$

and define the collective fuzzy ratings and weights by

$$\tilde{r}_{ij} := \mathrm{Agg}_R\big(\tilde{r}_{ij}^{(1)}, \ldots, \tilde{r}_{ij}^{(m)}\big), \qquad \tilde{w}_j := \mathrm{Agg}_w\big(\tilde{w}_j^{(1)}, \ldots, \tilde{w}_j^{(m)}\big).$$

(For TFNs, a common choice is componentwise arithmetic mean.)

**(C) Selection (collective scoring + ranking).** Fix a fuzzy aggregation/scoring operator $\mathrm{Agg}_C$ across criteria. A standard choice is the fuzzy weighted average (FWA):

$$\tilde{s}_i := \mathrm{Agg}_C\big((\tilde{r}_{i1}, \ldots, \tilde{r}_{ip}), (\tilde{w}_1, \ldots, \tilde{w}_p)\big) = \frac{\bigoplus_{j=1}^p (\tilde{w}_j \otimes \tilde{r}_{ij})}{\bigoplus_{j=1}^p \tilde{w}_j} \in \mathsf{FN}(\mathbb{R}),$$

where $\oplus, \otimes$ and $/$ are the adopted fuzzy arithmetic operations.

Let $\mathrm{Score} : \mathsf{FN}(\mathbb{R}) \to \mathbb{R}$ be a ranking functional (defuzzification/score). Define the collective preorder on $X$ by

$$x_i \succeq x_k \iff \mathrm{Score}(\tilde{s}_i) \geq \mathrm{Score}(\tilde{s}_k),$$

and select any

$$x^\star \in \arg\max_{x_i \in X} \mathrm{Score}(\tilde{s}_i).$$

**(D) Consensus (optional but standard in FGDM).** Fix a consensus measure Cons (e.g. built from pairwise similarity of experts' fuzzy preferences) and a threshold $\gamma \in [0, 1]$. A *consensus state* is reached when $\mathrm{Cons} \geq \gamma$; otherwise, a moderator-guided iterative revision of experts' opinions may be performed until $\mathrm{Cons} \geq \gamma$, after which the selection step (C) is applied.

As an extension of the above framework, we define Uncertain Group-Decision Making. The definition is given below.

**Definition 3.3.2** (Uncertain Group-Decision Making (UGDM) of type $M$). Let $\mathcal{A} = \{A_1, \ldots, A_m\}$ be a finite nonempty set of alternatives, $\mathcal{C} = \{C_1, \ldots, C_n\}$ a finite nonempty set of criteria, and $\mathcal{E} = \{E_1, \ldots, E_r\}$ a finite nonempty set of experts (decision makers). Fix an uncertain model $M$ with degree-domain $\mathrm{Dom}(M) \subseteq [0, 1]^k$.

An *uncertain group decision-making instance of type $M$* (UGDM instance) is the data

$$\mathfrak{G}_M = \big(\mathcal{A}, \mathcal{C}, \mathcal{E}, X_M, w, \lambda, \mathrm{Agg}_E, \mathrm{Agg}_C, \mathrm{Score}_M, \mathrm{Cons}\big),$$

where:

(i) $X_M = \big(\mu_{ij}^{(t)}\big) \in \mathrm{Dom}(M)^{m \times n \times r}$ is the *expert-indexed uncertain decision array*, where $\mu_{ij}^{(t)} \in \mathrm{Dom}(M)$ encodes expert $E_t$'s uncertain evaluation of $A_i$ under $C_j$;

(ii) $w = (w_1, \ldots, w_n) \in [0, 1]^n$ is a criterion-weight vector with $\sum_{j=1}^n w_j = 1$;

(iii) $\lambda = (\lambda_1, \ldots, \lambda_r) \in [0, 1]^r$ is an expert-weight vector with $\sum_{t=1}^r \lambda_t = 1$ (optional; if omitted set $\lambda_t = 1/r$);

(iv) $\mathrm{Agg}_E : \mathrm{Dom}(M)^r \to \mathrm{Dom}(M)$ is an *expert-aggregation operator* used entrywise to obtain a collective matrix

$$\bar{\mu}_{ij} := \mathrm{Agg}_E\big(\mu_{ij}^{(1)}, \ldots, \mu_{ij}^{(r)}; \lambda\big) \in \mathrm{Dom}(M), \qquad i = 1, \ldots, m, \; j = 1, \ldots, n;$$

(we allow $\mathrm{Agg}_E$ to depend parametrically on $\lambda$);

(v) $\mathrm{Agg}_C : \mathrm{Dom}(M)^n \times \Delta_n \to \mathrm{Dom}(M)$ is a *criterion-aggregation operator* (with $\Delta_n = \{w \in [0,1]^n : \sum_j w_j = 1\}$), producing an overall $M$-degree

$$s_i := \mathrm{Agg}_C\big((\bar{\mu}_{i1}, \dots, \bar{\mu}_{in}), w\big) \in \mathrm{Dom}(M), \qquad i = 1, \dots, m;$$

(vi) $\mathrm{Score}_M : \mathrm{Dom}(M) \to \mathbb{R}$ is a *ranking functional* inducing a preorder on $\mathcal{A}$ by

$$A_i \succeq_M A_\ell \quad \Longleftrightarrow \quad \mathrm{Score}_M(s_i) \geq \mathrm{Score}_M(s_\ell);$$

(vii) Cons : $[0,1]$ is a (possibly optional) *consensus index* computed from the experts' assessments (e.g. via distances/similarities between $\{\mu_{ij}^{(t)}\}$ and the collective $\{\bar{\mu}_{ij}\}$), together with a chosen threshold $\gamma \in [0,1]$ used operationally to accept the group aggregation.

A *UGDM best alternative* (solution) is any

$$A^\star \in \arg \max_{1 \leq i \leq m} \mathrm{Score}_M(s_i).$$

**Theorem 3.3.3** (Uncertain-set structure and well-definedness of UGDM)**.** *Let $\mathfrak{G}_M$ be a UGDM instance, with $\mathcal{A}, \mathcal{C}, \mathcal{E}$ finite nonempty and $M$ an uncertain model.*

*Assume:*

(A1) $\mathrm{Agg}_E : \mathrm{Dom}(M)^r \times \Delta_r \to \mathrm{Dom}(M)$ *is a total mapping (where $\Delta_r = \{\lambda \in [0,1]^r : \sum_t \lambda_t = 1\}$);*

(A2) $\mathrm{Agg}_C : \mathrm{Dom}(M)^n \times \Delta_n \to \mathrm{Dom}(M)$ *is a total mapping;*

(A3) $\mathrm{Score}_M : \mathrm{Dom}(M) \to \mathbb{R}$ *is a total mapping.*

*Then:*

(i) *(**Uncertain-set structure on the evaluation universe**) The expert-indexed array $X_M$ induces an uncertain set of type $M$ on*

$$U := \mathcal{A} \times \mathcal{C} \times \mathcal{E}$$

*via*

$$\mu : U \to \mathrm{Dom}(M), \qquad \mu(A_i, C_j, E_t) := \mu_{ij}^{(t)}.$$

*Hence $(U, \mu)$ is an uncertain set of type $M$.*

(ii) *(**Collective matrix is well-defined and yields an uncertain set**) The entrywise aggregated mapping*

$$\bar{\mu} : \mathcal{A} \times \mathcal{C} \to \mathrm{Dom}(M), \qquad \bar{\mu}(A_i, C_j) := \bar{\mu}_{ij},$$

*is well-defined, so $(\mathcal{A} \times \mathcal{C}, \bar{\mu})$ is an uncertain set of type $M$.*

(iii) *(**Aggregated alternative scores form an uncertain set on $\mathcal{A}$**) The mapping*

$$s : \mathcal{A} \to \mathrm{Dom}(M), \qquad s(A_i) := s_i,$$

*is well-defined, so $(\mathcal{A}, s)$ is an uncertain set of type $M$.*

(iv) (***Preference relation and solution set are well-defined***) *The induced relation $\succeq_M$ on $\mathcal{A}$ is a total preorder, and the solution set $\arg\max_{1 \leq i \leq m} \mathrm{Score}_M(s_i)$ is nonempty.*

*Proof.* **(i) Uncertain-set structure on $U = \mathcal{A} \times \mathcal{C} \times \mathcal{E}$.** Because $X_M \in \mathrm{Dom}(M)^{m \times n \times r}$, every entry $\mu_{ij}^{(t)}$ belongs to $\mathrm{Dom}(M)$. Define $\mu(A_i, C_j, E_t) := \mu_{ij}^{(t)}$. This is a well-defined mapping $U \to \mathrm{Dom}(M)$, hence $(U, \mu)$ is an uncertain set of type $M$.

**(ii) Collective matrix exists and defines an uncertain set on $\mathcal{A} \times \mathcal{C}$.** Fix $(i, j)$. The expert list $(\mu_{ij}^{(1)}, \ldots, \mu_{ij}^{(r)})$ lies in $\mathrm{Dom}(M)^r$ and $\lambda \in \Delta_r$. By (A1), the aggregated value

$$\bar{\mu}_{ij} := \mathrm{Agg}_E\big(\mu_{ij}^{(1)}, \ldots, \mu_{ij}^{(r)}; \lambda\big) \in \mathrm{Dom}(M)$$

exists. Therefore $\bar{\mu} : \mathcal{A} \times \mathcal{C} \to \mathrm{Dom}(M)$, $\bar{\mu}(A_i, C_j) = \bar{\mu}_{ij}$ is well-defined, and $(\mathcal{A} \times \mathcal{C}, \bar{\mu})$ is an uncertain set of type $M$.

**(iii) Alternative-level aggregation exists and defines an uncertain set on $\mathcal{A}$.** Fix $i$. The criterion tuple $(\bar{\mu}_{i1}, \ldots, \bar{\mu}_{in})$ belongs to $\mathrm{Dom}(M)^n$ and $w \in \Delta_n$. By (A2), the overall score

$$s_i := \mathrm{Agg}_C\big((\bar{\mu}_{i1}, \ldots, \bar{\mu}_{in}), w\big) \in \mathrm{Dom}(M)$$

exists. Hence $s : \mathcal{A} \to \mathrm{Dom}(M)$, $s(A_i) = s_i$ is well-defined and $(\mathcal{A}, s)$ is an uncertain set of type $M$.

**(iv) Well-defined preference relation and existence of a best alternative.** By (A3), each $\mathrm{Score}_M(s_i) \in \mathbb{R}$ exists. Define $A_i \succeq_M A_\ell$ iff $\mathrm{Score}_M(s_i) \geq \mathrm{Score}_M(s_\ell)$. Since $\geq$ on $\mathbb{R}$ is reflexive, transitive, and total, $\succeq_M$ is a total preorder.

Finally, the finite set $\{\mathrm{Score}_M(s_i) : i = 1, \ldots, m\} \subset \mathbb{R}$ attains a maximum, hence

$$\arg\max_{1 \leq i \leq m} \mathrm{Score}_M(s_i) \neq \varnothing,$$

so at least one best alternative exists and the solution set is well-defined. $\qquad\square$

A catalogue of uncertainty-aware *group decision-making* (GDM) families is presented in Table 3.3.

Uncertain Group Decision-Making is not the only perspective; related notions such as rough group decision making [360, 361], soft group decision making [362], grey group decision making [363, 364], multigroup decision making [365, 366], Z-Number group decision making [367, 368], consensus group decision making [369, 370], multi-stage group decision making [371, 372], multi-criteria group decision-making (MCGDM) [373, 374], and linguistic group decision making [375, 376] are also known.

Table 3.3: A compact catalogue of uncertainty-aware *group decision-making* (GDM) families by the dimension $k$ of the degree-domain $\mathrm{Dom}(M) \subseteq [0,1]^k$.

| $k$ | note | Representative GDM family/families whose per-evaluation degree-domain is a subset of $[0,1]^k$ |
|---|---|---|
| 1 | | Fuzzy group decision making: group aggregation/consensus over scalar fuzzy evaluations in $[0,1]$ (membership grades), typically via expert weights and fuzzy aggregation operators. |
| 2 | | Intuitionistic fuzzy group decision making [352, 353]: group aggregation/consensus over evaluations in $[0,1]^2$ (membership, non-membership). |
| 3 | | Hesitant fuzzy group decision making [354, 355]: group aggregation/consensus where each evaluation is represented in a $[0,1]^3$-type degree-domain (per the adopted hesitant encoding). |
| 3 | | Neutrosophic group decision making [356, 357]: group aggregation/consensus over evaluations in $[0,1]^3$ (truth, indeterminacy, falsity). |
| 3 | | Spherical fuzzy group decision making [358, 359]: group aggregation/consensus over evaluations in $[0,1]^3$ satisfying $\mu^2 + \nu^2 + \pi^2 \leq 1$ (membership, non-membership, hesitancy). |
| $n^\dagger$ | non-vector add-on | Plithogenic group decision making: scalar degrees on attribute–value pairs ($n = |\mathrm{Valset}(v)|$) coupled with a contradiction function $c(\cdot, \cdot)$ on values. |

**Reading guide.** Here $k$ denotes the dimension of the degree-domain used for each expert's evaluation (or preference statement). In GDM, evaluations are collected per expert and then aggregated (optionally with a consensus-reaching stage) into a collective assessment. $^{(\dagger)}$ For plithogenic models, the effective dimension depends on the number $n$ of admissible values of the chosen attribute, because degrees are attached to attribute–value pairs rather than stored as a fixed-length vector.

## 3.4 Fuzzy dynamic decision-making

Dynamic decision-making chooses actions over time, updating information and preferences, accounting for state transitions, and optimizing cumulative outcomes [377, 378]. Fuzzy dynamic decision-making models evolving states, rewards, or criteria with fuzzy sets, propagates uncertainty through updates, and selects adaptive policies [379, 380].

**Definition 3.4.1** (Fuzzy dynamic decision-making (multi-stage fuzzy MADM)). [379, 380] Let

$$\mathcal{A} = \{A_1, \ldots, A_m\} \quad \text{and} \quad \mathcal{C} = \{C_1, \ldots, C_n\}$$

be finite sets of alternatives and criteria. Let

$$\mathcal{T} = \{t_1, \ldots, t_p\}$$

be a finite set of decision stages (time periods).

Fix a chosen class of fuzzy numbers $\mathsf{FN}(\mathbb{R})$ (e.g. triangular/trapezoidal), equipped with fuzzy arithmetic $(\oplus, \otimes, \oslash)$ consistent with $\mathsf{FN}(\mathbb{R})$.

A *fuzzy dynamic decision-making instance* is specified by:

(1) **Stage-wise fuzzy decision matrices:** for each stage $t_k \in \mathcal{T}$, a fuzzy decision matrix

$$\widetilde{R}^{(k)} = \left(\widetilde{r}_{ij}^{(k)}\right) \in \mathsf{FN}(\mathbb{R})^{m \times n}, \qquad \widetilde{r}_{ij}^{(k)} \text{ rates } A_i \text{ w.r.t. } C_j \text{ at time } t_k.$$

(2) **(Optional) Stage-wise fuzzy criterion weights:** for each stage $t_k$, a fuzzy weight vector

$$\widetilde{W}^{(k)} = (\tilde{w}_1^{(k)}, \ldots, \tilde{w}_n^{(k)}) \in \mathsf{FN}(\mathbb{R}_{\geq 0})^n.$$

(If weights are time-invariant, write $\widetilde{W}^{(k)} = \widetilde{W}$.)

(3) **Stage weights (dynamic importance of time):** a weight vector

$$\omega = (\omega_1, \ldots, \omega_p) \in [0,1]^p, \qquad \sum_{k=1}^{p} \omega_k = 1,$$

where $\omega_k$ quantifies the importance of stage $t_k$ (e.g. giving larger weight to recent stages).

(4) **Aggregation operators:**

- a within-stage (across-criteria) fuzzy aggregation map

$$\mathrm{Agg}_{\mathcal{C}} : \mathsf{FN}(\mathbb{R})^n \times \mathsf{FN}(\mathbb{R}_{\geq 0})^n \to \mathsf{FN}(\mathbb{R}),$$

- a cross-stage (dynamic) fuzzy aggregation map

$$\mathrm{Agg}_{\mathcal{T}} : \mathsf{FN}(\mathbb{R})^p \times [0,1]^p \to \mathsf{FN}(\mathbb{R}).$$

(5) **A ranking functional:** a score/defuzzification map

$$\mathrm{Score} : \mathsf{FN}(\mathbb{R}) \to \mathbb{R}$$

that induces a preorder on fuzzy numbers by $\tilde{x} \preceq_F \tilde{y} \iff \mathrm{Score}(\tilde{x}) \leq \mathrm{Score}(\tilde{y})$.

A *dynamic fuzzy decision procedure* computes, for each alternative $A_i$:

(P1) **Within-stage fuzzy utility.** For each stage $t_k$, define a stage utility

$$\tilde{s}_i^{(k)} := \mathrm{Agg}_{\mathcal{C}}\Big(\tilde{r}_{i1}^{(k)}, \ldots, \tilde{r}_{in}^{(k)}; \ \tilde{w}_1^{(k)}, \ldots, \tilde{w}_n^{(k)}\Big) \in \mathsf{FN}(\mathbb{R}).$$

A standard choice is the fuzzy weighted average (FWA):

$$\tilde{s}_i^{(k)} = \left(\bigoplus_{j=1}^{n} (\tilde{w}_j^{(k)} \otimes \tilde{r}_{ij}^{(k)})\right) \oslash \left(\bigoplus_{j=1}^{n} \tilde{w}_j^{(k)}\right),$$

assuming the denominator is nonzero in the adopted fuzzy arithmetic.

(P2) **Cross-stage (dynamic) aggregation.** Aggregate the stage utilities into an overall dynamic utility:

$$\tilde{u}_i := \mathrm{Agg}_{\mathcal{T}}\Big(\tilde{s}_i^{(1)}, \ldots, \tilde{s}_i^{(p)}; \ \omega_1, \ldots, \omega_p\Big) \in \mathsf{FN}(\mathbb{R}).$$

A standard choice is the dynamic fuzzy weighted average (DFWA):

$$\tilde{u}_i = \bigoplus_{k=1}^{p} (\omega_k \otimes \tilde{s}_i^{(k)}),$$

interpreting $\omega_k \otimes (\cdot)$ as scalar multiplication in the fuzzy-number model.

(P3) **Ranking and selection.** Define the crisp overall score $U_i := \mathrm{Score}(\tilde{u}_i) \in \mathbb{R}$ and rank by

$$A_i \succeq A_\ell \iff U_i \geq U_\ell.$$

A *(best) solution* is any

$$A^\star \in \arg\max_{A_i \in \mathcal{A}} \mathrm{Score}(\tilde{u}_i).$$

In the *uncertain-set* viewpoint, the core data at every stage are degrees in a fixed degree-domain $\mathrm{Dom}(M) \subseteq [0,1]^k$.

**Definition 3.4.2** (Uncertain dynamic decision-making (UDDM) of type $M$)**.** Let

$$\mathcal{A} = \{A_1, \ldots, A_m\} \quad \text{(alternatives)}, \qquad \mathcal{C} = \{C_1, \ldots, C_n\} \quad \text{(criteria)},$$

and let

$$\mathcal{T} = \{t_1, \ldots, t_p\} \quad \text{(decision stages / time periods)}$$

be finite nonempty sets. Fix an uncertain model $M$ with degree-domain $\mathrm{Dom}(M) \subseteq [0,1]^k$.

An *uncertain dynamic decision-making instance of type $M$* is the tuple

$$\mathfrak{D}_M^{\mathrm{dyn}} = \Big( \mathcal{A}, \mathcal{C}, \mathcal{T}, (X_M^{(1)}, \ldots, X_M^{(p)}), (w_M^{(1)}, \ldots, w_M^{(p)}), \omega, \mathrm{Agg}_{\mathcal{C},M}, \mathrm{Agg}_{\mathcal{T},M}, \mathrm{Score}_M \Big),$$

where:

(i) **Stage-wise uncertain evaluation matrices:** for each $k \in \{1, \ldots, p\}$,

$$X_M^{(k)} = (\mu_{ij}^{(k)}) \in \mathrm{Dom}(M)^{m \times n},$$

where $\mu_{ij}^{(k)} \in \mathrm{Dom}(M)$ encodes the uncertain assessment of $A_i$ under criterion $C_j$ at stage $t_k$.

(ii) **Stage-wise uncertain criterion weights (optional but standard):** for each $k$,

$$w_M^{(k)} = (\omega_1^{(k)}, \ldots, \omega_n^{(k)}) \in \mathrm{Dom}(M)^n \quad \text{(or, alternatively, } w^{(k)} \in [0,1]^n \text{ with } \sum_j w_j^{(k)} = 1\text{)}.$$

(iii) **Stage-importance weights (time aggregation weights):** a vector

$$\omega = (\omega_1, \ldots, \omega_p) \in [0,1]^p, \qquad \sum_{k=1}^{p} \omega_k = 1.$$

(Thus $\omega_k$ models the relative importance of stage $t_k$, e.g. recency weighting.)

(iv) **Within-stage aggregation (across criteria):** a model-dependent total map

$$\mathrm{Agg}_{\mathcal{C},M} : \mathrm{Dom}(M)^n \times \mathrm{Dom}(M)^n \longrightarrow \mathrm{Dom}(M),$$

which assigns to each alternative $A_i$ at stage $t_k$ an aggregated uncertain stage-utility

$$s_i^{(k)} := \mathrm{Agg}_{\mathcal{C},M}\big((\mu_{i1}^{(k)}, \ldots, \mu_{in}^{(k)}), w_M^{(k)}\big) \in \mathrm{Dom}(M).$$

(v) **Cross-stage aggregation (dynamic synthesis):** a model-dependent total map

$$\mathrm{Agg}_{\mathcal{T},M} : \mathrm{Dom}(M)^p \times [0,1]^p \longrightarrow \mathrm{Dom}(M),$$

which aggregates stage-utilities into an overall uncertain dynamic utility

$$u_i := \mathrm{Agg}_{\mathcal{T},M}((s_i^{(1)}, \ldots, s_i^{(p)}), \omega) \in \mathrm{Dom}(M).$$

(vi) **Ranking functional:** a total map

$$\mathrm{Score}_M : \mathrm{Dom}(M) \longrightarrow \mathbb{R}.$$

It induces a preference relation on $\mathcal{A}$ by

$$A_i \succeq_M^{\mathrm{dyn}} A_j \quad \Longleftrightarrow \quad \mathrm{Score}_M(u_i) \geq \mathrm{Score}_M(u_j).$$

The *(dynamic) uncertain best-alternative set* is

$$\arg\max_{A_i \in \mathcal{A}} \mathrm{Score}_M(u_i).$$

**Theorem 3.4.3** (Uncertain-set structure and well-definedness of UDDM). *Let $\mathfrak{D}_M^{\mathrm{dyn}}$ be an uncertain dynamic decision-making instance as in Definition 3.4.2. Assume:*

(A1) $\mathrm{Agg}_{\mathcal{C},M} : \mathrm{Dom}(M)^n \times \mathrm{Dom}(M)^n \to \mathrm{Dom}(M)$ *is a total map;*

(A2) $\mathrm{Agg}_{\mathcal{T},M} : \mathrm{Dom}(M)^p \times [0,1]^p \to \mathrm{Dom}(M)$ *is a total map;*

(A3) $\mathrm{Score}_M : \mathrm{Dom}(M) \to \mathbb{R}$ *is a total map;*

(A4) $\omega \in [0,1]^p$ *satisfies $\sum_{k=1}^p \omega_k = 1$.*

*Then:*

(i) *The overall dynamic utilities $u_i \in \mathrm{Dom}(M)$ are well-defined for all $i = 1, \ldots, m$.*

(ii) *The mapping*

$$\mu_M^{\mathrm{dyn}} : \mathcal{A} \to \mathrm{Dom}(M), \qquad \mu_M^{\mathrm{dyn}}(A_i) := u_i,$$

*is well-defined; hence*

$$\mathcal{U}_M^{\mathrm{dyn}} := (\mathcal{A}, \mu_M^{\mathrm{dyn}})$$

*is an* uncertain set (U-set) *of type $M$ on the universe $\mathcal{A}$.*

(iii) *The induced relation $\succeq_M^{\mathrm{dyn}}$ on $\mathcal{A}$ is a* total preorder *(reflexive, transitive, and total).*

(iv) *The solution set $\arg\max_{A_i \in \mathcal{A}} \mathrm{Score}_M(u_i)$ is nonempty (so a best alternative exists).*

*Proof.* **(i) Stage-utilities exist in** $\mathrm{Dom}(M)$. Fix $i \in \{1, \ldots, m\}$ and $k \in \{1, \ldots, p\}$. Because $X_M^{(k)} \in \mathrm{Dom}(M)^{m \times n}$, the $i$th row

$$(\mu_{i1}^{(k)}, \ldots, \mu_{in}^{(k)}) \in \mathrm{Dom}(M)^n.$$

Also, by definition, $w_M^{(k)} \in \mathrm{Dom}(M)^n$ (or it is a crisp admissible surrogate; the present theorem covers the $\mathrm{Dom}(M)^n$ case explicitly). By assumption (A1), the value

$$s_i^{(k)} = \mathrm{Agg}_{\mathcal{C},M}\big((\mu_{i1}^{(k)}, \ldots, \mu_{in}^{(k)}), w_M^{(k)}\big)$$

is well-defined and lies in $\mathrm{Dom}(M)$.

**(ii) Dynamic utilities exist in** $\mathrm{Dom}(M)$. For fixed $i$, the stage-utility vector $(s_i^{(1)}, \ldots, s_i^{(p)})$ lies in $\mathrm{Dom}(M)^p$ by (i), and the stage-weight vector $\omega$ lies in $[0,1]^p$ by (A4). By (A2), the overall dynamic utility

$$u_i = \mathrm{Agg}_{\mathcal{T},M}\big((s_i^{(1)}, \ldots, s_i^{(p)}), \omega\big)$$

is well-defined and lies in $\mathrm{Dom}(M)$, proving (i) of the theorem statement.

**(iii) U-set structure.** Define $\mu_M^{\mathrm{dyn}} : \mathcal{A} \to \mathrm{Dom}(M)$ by $\mu_M^{\mathrm{dyn}}(A_i) = u_i$. By (ii) this is a well-defined map into $\mathrm{Dom}(M)$, hence $\mathcal{U}_M^{\mathrm{dyn}} = (\mathcal{A}, \mu_M^{\mathrm{dyn}})$ is a U-set of type $M$ on $\mathcal{A}$, proving (ii).

**(iv) The preference relation is a total preorder.** By (A3), each $\mathrm{Score}_M(u_i) \in \mathbb{R}$ is well-defined. Define $A_i \succeq_M^{\mathrm{dyn}} A_j$ iff $\mathrm{Score}_M(u_i) \geq \mathrm{Score}_M(u_j)$. Since $\geq$ on $\mathbb{R}$ is reflexive, transitive, and total, the induced relation on $\mathcal{A}$ inherits these properties; thus $\succeq_M^{\mathrm{dyn}}$ is a total preorder, proving (iii).

**(v) Existence of a maximizer.** The set $\{\mathrm{Score}_M(u_i) : i = 1, \ldots, m\} \subset \mathbb{R}$ is finite, hence attains its maximum. Therefore $\arg\max_{A_i \in \mathcal{A}} \mathrm{Score}_M(u_i) \neq \varnothing$, proving (iv). $\qquad\square$

Table 3.4 presents related uncertainty-model variants of Dynamic Fuzzy Decision-Making.

Table 3.4: Related uncertainty-model variants of Dynamic Fuzzy Decision-Making.

| $k$ | Related Dynamic Fuzzy Decision-Making variant(s) |
|---|---|
| 1 | Dynamic Fuzzy Decision-Making |
| 2 | Dynamic Intuitionistic Fuzzy Decision-Making |
| 3 | Dynamic Hesitant Fuzzy Decision-Making |
| 3 | Dynamic Spherical Fuzzy Decision-Making |
| 3 | Dynamic Neutrosophic Decision-Making |
| $n$ | Dynamic Plithogenic Decision-Making |

## 3.5 Fuzzy Multiple Objective Decision-making

Multiple Objective Decision-making optimizes several conflicting objectives simultaneously, generating Pareto-efficient solutions and selecting a compromise using preferences or weights [381, 382]. Fuzzy Multiple Objective Decision-making represents objectives, goals, or constraints fuzzily, computes satisfaction degrees, aggregates them, and selects a compromise solution [383, 384].

**Definition 3.5.1** (FMODM as fuzzy multi-objective programming with fuzzy goals). [383,384] Let $X \subseteq \mathbb{R}^n$ be a nonempty feasible set and let $k \geq 2$ be the number of objectives. In *fuzzy multi-objective decision-making* (FMODM), the objectives remain explicit and (typically) carry weights reflecting their relative significance.

Fix a continuous $t$-norm $T : [0,1]^2 \to [0,1]$ and its $k$-ary extension (still denoted $T$). Let $F(\mathbb{R})$ be a chosen class of fuzzy numbers on $\mathbb{R}$, and suppose the $i$th objective is a fuzzy-number-valued mapping

$$\tilde{f}_i : X \to F(\mathbb{R}), \qquad i = 1, \dots, k.$$

Assume for each objective $i$ two fuzzy reference levels are given:

$$\tilde{m}_i \in F(\mathbb{R}) \ \ \text{(undesired level)}, \qquad \tilde{M}_i \in F(\mathbb{R}) \ \ \text{(desired level)},$$

together with a metric (distance) $D : F(\mathbb{R}) \times F(\mathbb{R}) \to \mathbb{R}_{\geq 0}$.

**(1) Objective-wise satisfaction (application) functions.** Define, for each $i$, a satisfaction degree $H_i : X \to [0,1]$ by

$$H_i(x) := \min \left\{ 1 - \frac{1}{1 + D\big(\tilde{m}_i, \tilde{f}_i(x)\big)}, \ \ \frac{1}{1 + D\big(\tilde{M}_i, \tilde{f}_i(x)\big)} \right\},$$

or more generally by the $t$-norm aggregation

$$H_i(x) := T \left( 1 - \frac{1}{1 + D\big(\tilde{m}_i, \tilde{f}_i(x)\big)}, \ \ \frac{1}{1 + D\big(\tilde{M}_i, \tilde{f}_i(x)\big)} \right).$$

Intuitively, larger $H_i(x)$ means "far from undesired" and "close to desired" for objective $i$.

**(2) Fuzzy decision (intersection of fuzzy goals).** Let $w_1, \dots, w_k \in [0,1]$ be (crisp) importance weights with $\sum_{i=1}^{k} w_i = 1$. Define the overall fuzzy decision membership (the intersection of fuzzy goals) by

$$\mu_{\mathcal{D}}(x) := T\big(H_1(x)^{w_1}, \dots, H_k(x)^{w_k}\big), \qquad x \in X.$$

(Any other weighted $t$-norm construction may be used; the key point is that a $t$-norm models the conjunction of goals.)

**(3) FMODM solution.** An *FMODM (compromise) solution* is any

$$x^\star \in \arg\max_{x \in X} \ \mu_{\mathcal{D}}(x).$$

Equivalently, the fuzzy multi-objective problem is reduced to the single-objective scalarization

$$\max_{x \in X} \ T\big(H_1(x), \dots, H_k(x)\big),$$

with $H_i$ constructed from desired/undesired levels and a metric on fuzzy numbers.

In an *uncertain* setting, objective values and/or goal achievements are represented in a fixed degree-domain $\mathrm{Dom}(M) \subseteq [0,1]^d$ (fuzzy, intuitionistic fuzzy, neutrosophic, plithogenic, etc.), and the multiobjective problem is reduced to a *U-set* on the feasible set via a model-dependent aggregation and a real-valued score.

**Definition 3.5.2** (Uncertain multiple objective decision-making (UMODM) of type $M$). Let $X \subseteq \mathbb{R}^n$ be a nonempty feasible set, and let $k \geq 2$. Fix an uncertain model $M$ with degree-domain $\mathrm{Dom}(M) \subseteq [0,1]^d$.

An *uncertain multiple objective decision-making instance of type $M$* is the tuple

$$\mathfrak{P}_M = \big(X,\ (f_1, \ldots, f_k),\ (G_1, \ldots, G_k),\ w,\ \mathrm{Agg}_M,\ \mathrm{Score}_M\big),$$

where:

(i) **Objectives:** $f_i : X \to \mathbb{R}$ (crisp objectives) for $i = 1, \ldots, k$. (If objective evaluations are uncertain, replace $f_i$ by an uncertain evaluation map $\tilde{f}_i : X \to \mathrm{Dom}(M)$; the construction below is unchanged by composing with $G_i$.)

(ii) **Objective-to-degree (goal/satisfaction) maps:** for each $i$, a map

$$G_i : \mathbb{R} \to \mathrm{Dom}(M) \quad \text{(or more generally } G_i : X \to \mathrm{Dom}(M)\text{)}$$

that converts the objective level $f_i(x)$ into an uncertain degree $\mu_i(x) := G_i(f_i(x)) \in \mathrm{Dom}(M)$.

(iii) **Importance weights:** a vector $w = (w_1, \ldots, w_k) \in [0,1]^k$ with $\sum_{i=1}^{k} w_i = 1$.

(iv) **Uncertain goal aggregation:** a total map

$$\mathrm{Agg}_M : \mathrm{Dom}(M)^k \times [0,1]^k \longrightarrow \mathrm{Dom}(M),$$

interpreted as the conjunction/compromise operator over the $k$ uncertain goal degrees.

(v) **Ranking functional:** a total map

$$\mathrm{Score}_M : \mathrm{Dom}(M) \to \mathbb{R}.$$

For each feasible $x \in X$, define the *objective-wise uncertain degree vector*

$$\mu(x) := \big(\mu_1(x), \ldots, \mu_k(x)\big) \in \mathrm{Dom}(M)^k, \qquad \mu_i(x) := G_i(f_i(x)),$$

and define the *overall uncertain decision degree*

$$\mu_{\mathcal{D},M}(x) := \mathrm{Agg}_M\big(\mu(x), w\big) = \mathrm{Agg}_M\big((\mu_1(x), \ldots, \mu_k(x)), (w_1, \ldots, w_k)\big) \in \mathrm{Dom}(M).$$

The induced preference on $X$ is

$$x \succeq_M y \iff \mathrm{Score}_M\big(\mu_{\mathcal{D},M}(x)\big) \geq \mathrm{Score}_M\big(\mu_{\mathcal{D},M}(y)\big).$$

A *UMODM (compromise) solution* is any

$$x^\star \in \arg\max_{x \in X} \ \mathrm{Score}_M\big(\mu_{\mathcal{D},M}(x)\big).$$

**Theorem 3.5.3** (Uncertain-set structure and well-definedness of UMODM). *Let $\mathfrak{P}_M$ be a UMODM instance as in Definition 3.5.2. Assume:*

(A1) $X \neq \varnothing$.

(A2) *For each $i$, $G_i : \mathbb{R} \to \mathrm{Dom}(M)$ is a total map and $f_i : X \to \mathbb{R}$ is a total map.*

(A3) $\mathrm{Agg}_M : \mathrm{Dom}(M)^k \times [0,1]^k \to \mathrm{Dom}(M)$ *is a total map.*

(A4) $\mathrm{Score}_M : \mathrm{Dom}(M) \to \mathbb{R}$ *is a total map.*

(A5) $w \in [0,1]^k$ *satisfies* $\sum_{i=1}^k w_i = 1$.

*Then:*

(i) *The mapping $\mu_{\mathcal{D},M} : X \to \mathrm{Dom}(M)$ is well-defined. Consequently,*

$$\mathcal{U}_M^{\mathrm{modm}} := (X, \mu_{\mathcal{D},M})$$

*is an* uncertain set (U-set) *of type $M$ on the universe $X$.*

(ii) *The induced relation $\succeq_M$ on $X$ is a* total preorder.

(iii) *If $X$ is finite, then the UMODM solution set $\arg\max_{x \in X} \mathrm{Score}_M(\mu_{\mathcal{D},M}(x))$ is nonempty.*

*Proof.* **(i) Well-definedness and U-set structure.** Fix $x \in X$. For each $i$, since $f_i$ is total on $X$, $f_i(x) \in \mathbb{R}$ is well-defined. By (A2), $G_i$ is total, hence $\mu_i(x) = G_i(f_i(x)) \in \mathrm{Dom}(M)$ is well-defined. Therefore $\mu(x) = (\mu_1(x), \ldots, \mu_k(x)) \in \mathrm{Dom}(M)^k$ is well-defined. By (A3) and (A5), $\mathrm{Agg}_M(\mu(x), w) \in \mathrm{Dom}(M)$ is well-defined, hence $\mu_{\mathcal{D},M}(x) \in \mathrm{Dom}(M)$ exists for every $x \in X$. Thus $\mu_{\mathcal{D},M} : X \to \mathrm{Dom}(M)$ is a well-defined map, and $\mathcal{U}_M^{\mathrm{modm}} = (X, \mu_{\mathcal{D},M})$ is a U-set of type $M$.

**(ii) $\succeq_M$ is a total preorder.** By (A4), $\mathrm{Score}_M(\mu_{\mathcal{D},M}(x)) \in \mathbb{R}$ is well-defined for each $x \in X$. Define $x \succeq_M y$ iff $\mathrm{Score}_M(\mu_{\mathcal{D},M}(x)) \geq \mathrm{Score}_M(\mu_{\mathcal{D},M}(y))$. Since $\geq$ on $\mathbb{R}$ is reflexive, transitive, and total, the induced relation $\succeq_M$ inherits these properties and is therefore a total preorder.

**(iii) Existence of a maximizer when $X$ is finite.** If $X$ is finite, then the real set

$$S := \{\mathrm{Score}_M(\mu_{\mathcal{D},M}(x)) : x \in X\}$$

is finite and thus attains a maximum. Hence $\arg\max_{x \in X} \mathrm{Score}_M(\mu_{\mathcal{D},M}(x)) \neq \varnothing$. $\qquad\square$

Table 3.5 presents related uncertainty-model variants of Multiple Objective Decision-Making (MODM).

Table 3.5: Related uncertainty-model variants of Multiple Objective Decision-Making (MODM).

| $k$ | Related MODM variant(s) |
|---|---|
| 1 | Fuzzy Multiple Objective Decision-Making [385] |
| 2 | Intuitionistic Fuzzy Multiple Objective Decision-Making |
| 3 | Hesitant Fuzzy Multiple Objective Decision-Making |
| 3 | Spherical Fuzzy Multiple Objective Decision-Making |
| 3 | Neutrosophic Multiple Objective Decision-Making |
| $n$ | Plithogenic Multiple Objective Decision-Making |

## 3.6 Fuzzy ethical decision making

Ethical decision making selects actions by comparing moral principles and consequences, balancing rights, duties, and stakeholders to justify choices [386,387]. Fuzzy ethical decision making assigns graded compliance to ethical dimensions, applies fuzzy rules and inference, defuzzifies scores, and selects actions.

**Definition 3.6.1** (Fuzzy ethical decision making (rule-based fuzzy selection))**.** Let $\mathcal{S} \neq \varnothing$ be a set of *situations* (contexts, states) and let $\mathcal{A} \neq \varnothing$ be a set of feasible *actions*. Let $\mathcal{D} = \{d_1, \ldots, d_k\}$ be a finite set of ethical dimensions (e.g. privacy, fairness, transparency), each evaluated on a graded scale in $[0, 1]$.

For each dimension $d_j$, assume an *ethical compliance degree*

$$\mu_j : \mathcal{S} \times \mathcal{A} \longrightarrow [0, 1], \qquad (s, a) \longmapsto \mu_j(s, a),$$

where $\mu_j(s, a)$ is the degree to which action $a$ satisfies dimension $d_j$ in situation $s$.

Fix a continuous $t$-norm $T$ and an $s$-norm $S$ on $[0, 1]$ (for conjunction/disjunction in fuzzy inference), and fix a finite fuzzy rule base $\mathcal{R}$ whose rules have the form

$$R : \text{IF } (\mu_1 \text{ is } L_1) \text{ AND } \cdots \text{ AND } (\mu_k \text{ is } L_k) \text{ THEN } (E \text{ is } L_E),$$

where each $L_j$ and $L_E$ is a linguistic label (e.g. *low/medium/high*) represented by membership functions on $[0, 1]$. Such conditional rules score ethical behaviour by combining (for example) privacy, fairness, and transparency into an *ethical decision score*.

For each $(s, a) \in \mathcal{S} \times \mathcal{A}$, fuzzy inference (Mamdani-type) produces a fuzzy output $\widetilde{E}_{s,a}$ on $[0, 1]$ by

$$\mu_{\widetilde{E}_{s,a}}(e) := S_{R \in \mathcal{R}} \Big( T \big( \alpha_R(s, a), \ \mu_{L_E^R}(e) \big) \Big),$$

where $\alpha_R(s, a) \in [0, 1]$ is the firing strength of rule $R$ computed from the antecedent labels via $T$, and $\mu_{L_E^R}$ is the membership function of the consequent label of $R$.

Let Defuzz be a defuzzification functional (e.g. centroid), and define the *crisp ethical score*

$$E(s, a) := \text{Defuzz}\big(\widetilde{E}_{s,a}\big) \in [0, 1].$$

(Defuzzification is required when non-boolean linguistic values must be converted to actionable numeric outputs.)

A *fuzzy ethical decision* in situation $s$ is any action

$$a^\star \in \arg\max_{a \in \mathcal{A}} E(s, a).$$

In an *uncertain* formulation, each ethical dimension is evaluated in a fixed uncertain degree-domain $\mathrm{Dom}(M) \subseteq [0,1]^d$ (chosen uncertain model $M$), and the overall ethical acceptability is represented as an *uncertain set* on the action space; actions are then ranked via a real-valued score.

**Definition 3.6.2** (Uncertain ethical decision making (UEDM) of type $M$). Let $\mathcal{S} \neq \varnothing$ be a set of situations (contexts) and $\mathcal{A} \neq \varnothing$ a set of feasible actions. Let $\mathcal{D} = \{d_1, \ldots, d_k\}$ be a finite set of ethical dimensions ($k \geq 1$). Fix an uncertain model $M$ with degree-domain $\mathrm{Dom}(M) \subseteq [0,1]^d$.

An *uncertain ethical decision making instance of type $M$* is a tuple

$$\mathfrak{E}_M = \big(\mathcal{S}, \mathcal{A}, \mathcal{D}, (\mu_1, \ldots, \mu_k), w, \mathrm{Agg}_M, \mathrm{Score}_M\big),$$

where:

(i) **Dimension-wise uncertain compliance degrees:** for each $j = 1, \ldots, k$, a total map

$$\mu_j : \mathcal{S} \times \mathcal{A} \longrightarrow \mathrm{Dom}(M),$$

where $\mu_j(s, a)$ encodes the uncertain degree to which action $a$ satisfies ethical dimension $d_j$ in situation $s$.

(ii) **Importance weights:** a vector $w = (w_1, \ldots, w_k) \in [0,1]^k$ with $\sum_{j=1}^{k} w_j = 1$.

(iii) **Uncertain ethical aggregation operator:** a total map

$$\mathrm{Agg}_M : \mathrm{Dom}(M)^k \times [0,1]^k \longrightarrow \mathrm{Dom}(M),$$

interpreted as combining the $k$ ethical-dimension degrees into a single overall ethical acceptability degree.

(iv) **Ranking functional:** a total map

$$\mathrm{Score}_M : \mathrm{Dom}(M) \to \mathbb{R}.$$

For each $(s, a) \in \mathcal{S} \times \mathcal{A}$ define the *overall uncertain ethical degree*

$$\mu_{\mathrm{eth},M}(s, a) := \mathrm{Agg}_M\big((\mu_1(s, a), \ldots, \mu_k(s, a)), \ (w_1, \ldots, w_k)\big) \in \mathrm{Dom}(M).$$

Thus, for each fixed situation $s$, the mapping

$$\mu_{\mathrm{eth},M}(s, \cdot) : \mathcal{A} \to \mathrm{Dom}(M), \qquad a \mapsto \mu_{\mathrm{eth},M}(s, a),$$

defines an *uncertain set (U-set) of type $M$* on the universe $\mathcal{A}$.

A *UEDM preference* at situation $s$ is the relation $\succeq_{M,s}$ on $\mathcal{A}$ given by

$$a \succeq_{M,s} b \quad \Longleftrightarrow \quad \mathrm{Score}_M\big(\mu_{\mathrm{eth},M}(s, a)\big) \geq \mathrm{Score}_M\big(\mu_{\mathrm{eth},M}(s, b)\big).$$

A *UEDM ethical decision* at situation $s$ is any action

$$a^{\star}(s) \in \arg\max_{a \in \mathcal{A}} \ \mathrm{Score}_M\big(\mu_{\mathrm{eth},M}(s, a)\big).$$

**Theorem 3.6.3** (Uncertain-set structure and well-definedness of UEDM)**.** *Let $\mathfrak{E}_M$ be a UEDM instance as in Definition 3.6.2. Assume:*

(A1)  $\mathcal{S} \neq \varnothing$ *and* $\mathcal{A} \neq \varnothing$ *and* $\mathcal{D}$ *is finite.*

(A2)  *Each* $\mu_j : \mathcal{S} \times \mathcal{A} \to \mathrm{Dom}(M)$ *is a total map.*

(A3)  $\mathrm{Agg}_M : \mathrm{Dom}(M)^k \times [0,1]^k \to \mathrm{Dom}(M)$ *is a total map.*

(A4)  $\mathrm{Score}_M : \mathrm{Dom}(M) \to \mathbb{R}$ *is a total map.*

(A5)  $w \in [0,1]^k$ *satisfies* $\sum_{j=1}^{k} w_j = 1.$

*Then:*

(i)   *The mapping* $\mu_{\mathrm{eth},M} : \mathcal{S} \times \mathcal{A} \to \mathrm{Dom}(M)$ *is well-defined. For every* $s \in \mathcal{S}$, $(\mathcal{A}, \mu_{\mathrm{eth},M}(s,\cdot))$ *is a U-set of type $M$ on $\mathcal{A}$.*

(ii)  *For each* $s \in \mathcal{S}$, *the relation* $\succeq_{M,s}$ *is a total preorder on* $\mathcal{A}$.

(iii) *If $\mathcal{A}$ is finite, then for every $s \in \mathcal{S}$ the argmax set* $\arg\max_{a \in \mathcal{A}} \mathrm{Score}_M(\mu_{\mathrm{eth},M}(s,a))$ *is nonempty.*

*Proof.* **(i) Well-definedness and U-set structure.** Fix $(s,a) \in \mathcal{S} \times \mathcal{A}$. By (A2), each value $\mu_j(s,a) \in \mathrm{Dom}(M)$ exists and is uniquely determined, hence the $k$-tuple $(\mu_1(s,a), \ldots, \mu_k(s,a)) \in \mathrm{Dom}(M)^k$ is well-defined. By (A3) and (A5), the aggregated degree

$$\mu_{\mathrm{eth},M}(s,a) = \mathrm{Agg}_M\big((\mu_1(s,a), \ldots, \mu_k(s,a)), w\big)$$

is a well-defined element of $\mathrm{Dom}(M)$. Therefore $\mu_{\mathrm{eth},M}$ is a well-defined map on $\mathcal{S} \times \mathcal{A}$. For fixed $s$, the function $\mu_{\mathrm{eth},M}(s,\cdot) : \mathcal{A} \to \mathrm{Dom}(M)$ is a membership assignment in the model $M$, hence $(\mathcal{A}, \mu_{\mathrm{eth},M}(s,\cdot))$ is a U-set of type $M$.

**(ii) $\succeq_{M,s}$ is a total preorder.** Fix $s \in \mathcal{S}$. By (A4), the real score $\mathrm{Score}_M(\mu_{\mathrm{eth},M}(s,a)) \in \mathbb{R}$ is well-defined for each $a \in \mathcal{A}$. Define $a \succeq_{M,s} b$ iff $\mathrm{Score}_M(\mu_{\mathrm{eth},M}(s,a)) \geq \mathrm{Score}_M(\mu_{\mathrm{eth},M}(s,b))$. Since $\geq$ on $\mathbb{R}$ is reflexive, transitive, and total, the induced relation $\succeq_{M,s}$ is a total preorder.

**(iii) Existence of a maximizer for finite $\mathcal{A}$.** If $\mathcal{A}$ is finite, then the finite set

$$S_s := \{\mathrm{Score}_M(\mu_{\mathrm{eth},M}(s,a)) : \ a \in \mathcal{A}\} \subseteq \mathbb{R}$$

attains a maximum. Hence the argmax set is nonempty. $\qquad\square$

Table 3.6 presents related uncertainty-model variants of Fuzzy Ethical Decision-Making.

Table 3.6: Related uncertainty-model variants of Fuzzy Ethical Decision-Making.

| $k$ | Related Fuzzy Ethical Decision-Making variant(s) |
|---|---|
| 1 | Fuzzy Ethical Decision-Making |
| 2 | Intuitionistic Fuzzy Ethical Decision-Making |
| 3 | Hesitant Fuzzy Ethical Decision-Making |
| 3 | Spherical Fuzzy Ethical Decision-Making |
| 3 | Neutrosophic Ethical Decision-Making |
| $n$ | Plithogenic Ethical Decision-Making |

## 3.7 Fuzzy Consensus decision-making

Consensus decision-making is a collaborative process in which participants discuss alternatives, reconcile differing views, and seek a broadly acceptable decision supported by the whole group [388, 389]. Fuzzy consensus decision-making is a group process where experts' fuzzy evaluations are iteratively adjusted toward a collective fuzzy matrix until consensus threshold holds, then alternatives ranked [390, 391].

**Definition 3.7.1** (Fuzzy consensus decision-making (consensus-reaching + selection)). [390, 391] Let

$$\mathcal{A} = \{A_1, \ldots, A_m\} \quad \text{(alternatives)},$$

$$\mathcal{C} = \{C_1, \ldots, C_n\} \quad \text{(criteria)},$$

$$\mathcal{E} = \{e_1, \ldots, e_p\} \quad \text{(experts)}.$$

Fix a class $\mathsf{FN}$ of fuzzy numbers (e.g. TFNs), and let each expert $e_k$ provide a fuzzy decision matrix

$$\widetilde{R}^{(k)} = \big(\tilde{r}_{ij}^{(k)}\big) \in \mathsf{FN}^{m \times n}, \qquad \tilde{r}_{ij}^{(k)} \text{ is the fuzzy evaluation of } A_i \text{ under } C_j.$$

Let $w = (w_1, \ldots, w_p)$ be expert weights with $w_k \geq 0$ and $\sum_{k=1}^{p} w_k = 1$.

**(1) Distance, similarity, and consensus index.** Assume a normalized distance on fuzzy numbers

$$d : \mathsf{FN} \times \mathsf{FN} \to [0, 1].$$

Extend it to decision matrices by

$$D\big(\widetilde{R}^{(k)}, \widetilde{R}^{(\ell)}\big) := \frac{1}{mn} \sum_{i=1}^{m} \sum_{j=1}^{n} d\big(\tilde{r}_{ij}^{(k)}, \tilde{r}_{ij}^{(\ell)}\big) \in [0, 1],$$

and define the corresponding similarity

$$S\big(\widetilde{R}^{(k)}, \widetilde{R}^{(\ell)}\big) := 1 - D\big(\widetilde{R}^{(k)}, \widetilde{R}^{(\ell)}\big) \in [0, 1].$$

A (weighted) *group consensus index* is

$$\mathrm{GCI}\big(\widetilde{R}^{(1)}, \ldots, \widetilde{R}^{(p)}\big) := \sum_{k=1}^{p} w_k \, S\big(\widetilde{R}^{(k)}, \widetilde{R}^c\big) \in [0, 1],$$

where $\widetilde{R}^c$ is a collective matrix obtained by a fixed fuzzy aggregation operator $\mathrm{Agg}_R$ (e.g. componentwise weighted average):

$$\widetilde{R}^c := \mathrm{Agg}_R\big(\widetilde{R}^{(1)}, \ldots, \widetilde{R}^{(p)}; w\big) \in \mathsf{FN}^{m \times n}.$$

**(2) Consensus-reaching condition.** Fix a consensus threshold $\beta \in [0,1]$. The experts are said to *reach consensus* if

$$\text{GCI}(\widetilde{R}^{(1)}, \ldots, \widetilde{R}^{(p)}) \geq \beta.$$

If this fails, a consensus-reaching process iteratively revises experts' opinions (and possibly their weights) until the threshold is met. Such iterative revision/feedback is the standard viewpoint of consensus reaching in group decision making.

**(3) A generic fuzzy feedback (opinion-adjustment) step.** At iteration $t$, given current matrices $\widetilde{R}^{(k)}(t)$, compute the collective matrix $\widetilde{R}^c(t) = \text{Agg}_R(\widetilde{R}^{(1)}(t), \ldots, \widetilde{R}^{(p)}(t); w(t))$. For each expert $e_k$ that is requested to adjust, choose an *acceptance degree* $\theta_k(t) \in [0,1]$ and update entrywise by the convex-combination rule

$$\tilde{r}_{ij}^{(k)}(t+1) := (1 - \theta_k(t)) \otimes \tilde{r}_{ij}^{(k)}(t) \ \oplus \ \theta_k(t) \otimes \tilde{r}_{ij}^c(t), \qquad i = 1, \ldots, m, \ j = 1, \ldots, n,$$

where $\oplus, \otimes$ denote the adopted fuzzy arithmetic (e.g. extension-principle addition and scalar multiplication). This expresses "move (to degree $\theta_k$) toward a trusted/collective opinion", a common mechanism in consensus-reaching models.

Repeat (1)–(3) until $\text{GCI} \geq \beta$.

**(4) Selection after consensus.** Let $\widetilde{R}^c$ be the final collective matrix. Fix criterion weights $\tilde{v} = (\tilde{v}_1, \ldots, \tilde{v}_n)$ (crisp or fuzzy) and a final scoring/defuzzification map $\text{Score} : \mathsf{FN} \to \mathbb{R}$. Define the overall fuzzy score of alternative $A_i$ by a chosen operator $\text{Agg}_C$, e.g. the fuzzy weighted average:

$$\tilde{s}_i := \text{Agg}_C(\tilde{r}_{i1}^c, \ldots, \tilde{r}_{in}^c; \tilde{v}_1, \ldots, \tilde{v}_n) \in \mathsf{FN},$$

$$A^\star \in \arg \max_{1 \leq i \leq m} \text{Score}(\tilde{s}_i).$$

Any such $A^\star$ is called a *fuzzy consensus decision*.

Uncertain consensus decision-making generalizes consensus-reaching group decision processes by allowing each assessment to take values in a fixed *uncertain degree-domain* $\text{Dom}(M) \subseteq [0,1]^d$ (an uncertain model $M$). The collective outcome after consensus is represented as an *uncertain set* on the alternative space, and a real-valued scoring functional induces ranking and selection.

**Definition 3.7.2** (Uncertain consensus decision-making (UCDM) of type $M$). Let

$$\mathcal{A} = \{A_1, \ldots, A_m\} \text{ (alternatives)}, \qquad \mathcal{C} = \{C_1, \ldots, C_n\} \text{ (criteria)}, \qquad \mathcal{E} = \{e_1, \ldots, e_p\} \text{ (experts)},$$

with $m, n, p \geq 1$. Fix an uncertain model $M$ with degree-domain $\text{Dom}(M) \subseteq [0,1]^d$.

An *uncertain consensus decision-making instance of type $M$* is a tuple

$$\mathfrak{C}_M = \big(\mathcal{A}, \mathcal{C}, \mathcal{E}, (R^{(k)})_{k=1}^p, w, d_M, \text{Agg}_R, \beta, \text{Up}_M, \text{Agg}_C, \text{Score}_M\big),$$

where:

(i) **Uncertain evaluation matrices:** each expert $e_k$ provides an uncertain decision matrix

$$R^{(k)} = \left(r_{ij}^{(k)}\right) \in \mathrm{Dom}(M)^{m \times n}, \qquad r_{ij}^{(k)} \in \mathrm{Dom}(M) \text{ is the uncertain evaluation of } A_i \text{ under } C_j.$$

(ii) **Expert weights:** $w = (w_1, \ldots, w_p) \in [0,1]^p$ with $\sum_{k=1}^p w_k = 1$.

(iii) **Distance and similarity on** $\mathrm{Dom}(M)$**:** a normalized distance

$$d_M : \mathrm{Dom}(M) \times \mathrm{Dom}(M) \to [0,1],$$

and its entrywise extension to matrices

$$D_M(R^{(k)}, R^{(\ell)}) := \frac{1}{mn} \sum_{i=1}^m \sum_{j=1}^n d_M(r_{ij}^{(k)}, r_{ij}^{(\ell)}) \in [0,1],$$

with similarity $S_M(R^{(k)}, R^{(\ell)}) := 1 - D_M(R^{(k)}, R^{(\ell)})$.

(iv) **Collective aggregation across experts:** a total aggregation operator

$$\mathrm{Agg}_R : \mathrm{Dom}(M)^p \times [0,1]^p \to \mathrm{Dom}(M)$$

extended entrywise to matrices by

$$R^c := \mathrm{Agg}_R(R^{(1)}, \ldots, R^{(p)}; w) \in \mathrm{Dom}(M)^{m \times n}, \qquad r_{ij}^c := \mathrm{Agg}_R(r_{ij}^{(1)}, \ldots, r_{ij}^{(p)}; w).$$

(v) **Consensus index and threshold:** a group consensus index

$$\mathrm{GCI}_M(R^{(1)}, \ldots, R^{(p)}) := \sum_{k=1}^p w_k \, S_M(R^{(k)}, R^c) \in [0,1],$$

and a threshold $\beta \in [0,1]$.

(vi) **Opinion-update (feedback) operator:** a total update map

$$\mathrm{Up}_M : \mathrm{Dom}(M) \times \mathrm{Dom}(M) \times [0,1] \to \mathrm{Dom}(M),$$

which updates an expert's entry toward the collective entry with acceptance degree $\theta \in [0,1]$. At iteration $t$, for each requested expert $k$ and each $(i,j)$:

$$r_{ij}^{(k)}(t+1) := \mathrm{Up}_M(r_{ij}^{(k)}(t), r_{ij}^c(t), \theta_k(t)), \qquad \theta_k(t) \in [0,1].$$

(vii) **Post-consensus selection:** a criterion-aggregation operator

$$\mathrm{Agg}_C : \mathrm{Dom}(M)^n \times [0,1]^n \to \mathrm{Dom}(M),$$

(using criterion weights $v = (v_1, \ldots, v_n) \in [0,1]^n$, $\sum_j v_j = 1$), and a scoring functional

$$\mathrm{Score}_M : \mathrm{Dom}(M) \to \mathbb{R}.$$

**Consensus-reaching phase.** Starting from $(R^{(k)}(0))_{k=1}^p$, define for $t = 0, 1, 2, \ldots$ the collective matrix $R^c(t) = \mathrm{Agg}_R(R^{(1)}(t), \ldots, R^{(p)}(t); w)$ and compute $\mathrm{GCI}_M(t) := \mathrm{GCI}_M(R^{(1)}(t), \ldots, R^{(p)}(t))$. Consensus is *reached* at the first time $T$ such that $\mathrm{GCI}_M(T) \geq \beta$.

**Decision (uncertain-set output) phase.** Let $R^c := R^c(T)$ be the final collective matrix. For each alternative $A_i$, define the overall uncertain utility degree

$$u_i := \mathrm{Agg}_C(r_{i1}^c, \dots, r_{in}^c; v_1, \dots, v_n) \in \mathrm{Dom}(M).$$

Then the mapping

$$\mu_{\mathcal{U}} : \mathcal{A} \to \mathrm{Dom}(M), \qquad \mu_{\mathcal{U}}(A_i) := u_i,$$

is the *collective uncertain utility set* (a U-set of type $M$) on $\mathcal{A}$. A final ranking is induced by

$$A_i \succeq_M A_\ell \quad \Longleftrightarrow \quad \mathrm{Score}_M(u_i) \geq \mathrm{Score}_M(u_\ell),$$

and any

$$A^\star \in \arg\max_{A_i \in \mathcal{A}} \ \mathrm{Score}_M\big(\mu_{\mathcal{U}}(A_i)\big)$$

is called an *uncertain consensus decision*.

**Theorem 3.7.3** (Uncertain-set structure and well-definedness of UCDM)**.** *Consider a UCDM instance $\mathfrak{C}_M$ as in Definition 3.7.2. Assume:*

(A1) *$m, n, p \geq 1$ and $\mathcal{A}, \mathcal{C}, \mathcal{E}$ are finite nonempty.*

(A2) *$R^{(k)} \in \mathrm{Dom}(M)^{m \times n}$ for all $k$ (hence each entry $r_{ij}^{(k)} \in \mathrm{Dom}(M)$ exists).*

(A3) *$d_M$ is a total map $\mathrm{Dom}(M) \times \mathrm{Dom}(M) \to [0,1]$.*

(A4) *$\mathrm{Agg}_R$ and $\mathrm{Agg}_C$ are total maps with codomain $\mathrm{Dom}(M)$.*

(A5) *$\mathrm{Up}_M$ is a total map with codomain $\mathrm{Dom}(M)$, and $\theta_k(t) \in [0,1]$ for all $k, t$.*

(A6) *$\mathrm{Score}_M : \mathrm{Dom}(M) \to \mathbb{R}$ is a total map; and $w, v$ are normalized weight vectors.*

*Then:*

(i) *For every $t \geq 0$, the matrices $R^{(k)}(t)$ produced by the update rule are well-defined elements of $\mathrm{Dom}(M)^{m \times n}$, and the collective matrix $R^c(t)$ is well-defined. Consequently $\mathrm{GCI}_M(t) \in [0,1]$ is well-defined.*

(ii) *If consensus is reached at some finite time $T$, then the post-consensus utility mapping $\mu_{\mathcal{U}} : \mathcal{A} \to \mathrm{Dom}(M)$ is a well-defined U-set of type $M$ on $\mathcal{A}$.*

(iii) *The induced relation $\succeq_M$ on $\mathcal{A}$ is a total preorder.*

(iv) *If consensus is reached at some finite time $T$ and $\mathcal{A}$ is finite, then the argmax set defining an uncertain consensus decision is nonempty.*

*Proof.* **(i) Well-definedness along iterations.** Fix $t \geq 0$ and suppose inductively that $R^{(k)}(t) \in \mathrm{Dom}(M)^{m \times n}$ for all $k$ (true at $t = 0$ by (A2)). By (A4), for each $(i,j)$ the value

$$r_{ij}^c(t) = \mathrm{Agg}_R(r_{ij}^{(1)}(t), \dots, r_{ij}^{(p)}(t); w) \in \mathrm{Dom}(M)$$

is well-defined, hence $R^c(t) \in \mathrm{Dom}(M)^{m \times n}$ is well-defined. By (A5), the update

$$r_{ij}^{(k)}(t+1) = \mathrm{Up}_M\big(r_{ij}^{(k)}(t), r_{ij}^c(t), \theta_k(t)\big) \in \mathrm{Dom}(M)$$

is well-defined for each $k, i, j$, so $R^{(k)}(t+1) \in \mathrm{Dom}(M)^{m \times n}$ for all $k$. Thus, by induction, all $R^{(k)}(t)$ and $R^c(t)$ are well-defined for every $t$.

Next, by (A3) each entrywise distance $d_M(r_{ij}^{(k)}(t), r_{ij}^c(t)) \in [0, 1]$ is well-defined, so the averages $D_M(R^{(k)}(t), R^c(t)) \in [0, 1]$ and similarities $S_M(R^{(k)}(t), R^c(t)) \in [0, 1]$ are well-defined. Finally, $\mathrm{GCI}_M(t)$ is a convex combination of values in $[0, 1]$ using weights $w$, hence $\mathrm{GCI}_M(t) \in [0, 1]$ is well-defined.

**(ii) U-set structure after consensus.** Assume consensus is reached at time $T$. Then $R^c = R^c(T) \in \mathrm{Dom}(M)^{m \times n}$ is well-defined by (i). For each alternative $A_i$, by (A4) the aggregation

$$u_i = \mathrm{Agg}_C(r_{i1}^c, \ldots, r_{in}^c; v) \in \mathrm{Dom}(M)$$

is well-defined. Therefore $\mu_{\mathcal{U}}(A_i) := u_i$ defines a well-defined map $\mu_{\mathcal{U}} : \mathcal{A} \to \mathrm{Dom}(M)$, i.e. a U-set of type $M$ on $\mathcal{A}$.

**(iii) $\succeq_M$ is a total preorder.** By (A6), each score $\mathrm{Score}_M(u_i) \in \mathbb{R}$ is well-defined. Since $\geq$ on $\mathbb{R}$ is reflexive, transitive, and total, the induced relation $A_i \succeq_M A_\ell \iff \mathrm{Score}_M(u_i) \geq \mathrm{Score}_M(u_\ell)$ is a total preorder.

**(iv) Existence of a maximizer for finite $\mathcal{A}$.** If $\mathcal{A}$ is finite and consensus is reached, then the finite set of real scores $\{\mathrm{Score}_M(\mu_{\mathcal{U}}(A_i)) : A_i \in \mathcal{A}\}$ attains a maximum. Hence the argmax set is nonempty. □

Table 3.7 presents related uncertainty-model variants of Fuzzy Consensus Decision-Making.

Table 3.7: Related uncertainty-model variants of Fuzzy Consensus Decision-Making.

| $k$ | Related Fuzzy Consensus Decision-Making variant(s) |
|---|---|
| 1 | Fuzzy Consensus Decision-Making |
| 2 | Intuitionistic Fuzzy Consensus Decision-Making |
| 3 | Hesitant Fuzzy Consensus Decision-Making |
| 3 | Spherical Fuzzy Consensus Decision-Making |
| 3 | Neutrosophic Consensus Decision-Making |
| $n$ | Plithogenic Consensus Decision-Making |

## 3.8 Fuzzy Strategic decision making

Strategic decision making selects long-term initiatives under resource constraints, aligning actions with organizational goals, risks, and competitive priorities [392, 393]. Fuzzy strategic decision making scores strategic plans by fuzzy goal achievement and cost satisfaction, aggregates via t-norms, selects best [394, 395].

**Definition 3.8.1** (Fuzzy strategic decision-making (strategy-map based)). Let $\mathcal{M} = \{M_1, \ldots, M_p\}$ be a finite set of strategic measures (projects). A *strategic plan* (decision) is a vector

$$x = (x_1, \ldots, x_p) \in \mathcal{D} \subseteq \{0, 1\}^p,$$

where $x_\ell = 1$ means that measure $M_\ell$ is selected, and $\mathcal{D}$ encodes feasibility constraints (e.g. resource budgets, precedence constraints).

Let $\mathcal{G} = \{g_1, \ldots, g_s\}$ be strategic goals (from a formalized strategy map). For each goal $g_i$, fix a set $\mathcal{I}_i = \{I_{i1}, \ldots, I_{im_i}\}$ of its (resultant) indicators and initial/target values $(v_{ij}^0, v_{ij}^\star)$ for each indicator $I_{ij}$.

**(1) Strategic criteria (goal-achievement degrees).** For each $i$, assume the strategy-map formalization provides a mapping that assigns to every plan $x \in \mathcal{D}$ a *degree of achievement*

$$S_i(x) \in [0, 1],$$

interpreted as the (graded) level at which goal $g_i$ is achieved under $x$.

**(2) Economic criteria (resource costs and efficiency).** Let $\mathcal{R} = \{1, \ldots, r\}$ be resource types. Define cost/consumption criteria

$$C_u : \mathcal{D} \to \mathbb{R}_{\geq 0} \qquad (u \in \mathcal{R}),$$

and (optionally) additional economic-efficiency criteria

$$E_v : \mathcal{D} \to \mathbb{R} \qquad (v = 1, \ldots, q).$$

Thus, the criteria are partitioned into a *strategic group* $\{S_i\}_{i=1}^s$ and an *economic group* $\{C_u\}_{u=1}^r$ (and possibly $\{E_v\}_{v=1}^q$).

**(3) Fuzzy evaluation and fuzzy selection.** Assume each criterion is assessed fuzzily by a satisfaction (membership) degree:

$$\mu_i^S : \mathcal{D} \to [0, 1] \quad (i = 1, \ldots, s),$$
$$\mu_u^C : \mathcal{D} \to [0, 1] \quad (u = 1, \ldots, r),$$
$$\mu_v^E : \mathcal{D} \to [0, 1] \quad (v = 1, \ldots, q).$$

Let $T$ be a $k$-ary $t$-norm (conjunction operator), and let $w = (w_1, \ldots, w_{s+r+q})$ be nonnegative importance weights with $\sum w_j = 1$. Define the overall *fuzzy decision membership* of a plan $x$ by

$$\mu_{\mathcal{D}}(x) := T\Big((\mu_1^S(x))^{w_1}, \ldots, (\mu_s^S(x))^{w_s}, (\mu_1^C(x))^{w_{s+1}}, \ldots, (\mu_r^C(x))^{w_{s+r}},$$

$$(\mu_1^E(x))^{w_{s+r+1}}, \ldots, (\mu_q^E(x))^{w_{s+r+q}}\Big) \in [0, 1].$$

A *fuzzy strategic decision* (optimal strategic plan) is any

$$x^\star \in \arg\max_{x \in \mathcal{D}} \ \mu_{\mathcal{D}}(x).$$

We now define Uncertain Strategic Decision-Making (USDM), obtained by extending this framework using Uncertain Sets.

**Definition 3.8.2** (Uncertain strategic decision-making (USDM) of type $M$)**.** Let $\mathcal{M} = \{M_1, \ldots, M_p\}$ be a finite set of strategic measures (projects), and let

$$\mathcal{D} \subseteq \{0, 1\}^p$$

be a nonempty feasible set of strategic plans, where $x = (x_1, \ldots, x_p) \in \mathcal{D}$ indicates the selected measures.

Fix an *uncertain model* $M$ with degree-domain $\mathrm{Dom}(M) \subseteq [0, 1]^d$.

Let $\mathcal{G} = \{g_1, \ldots, g_s\}$ be strategic goals and let $\mathcal{R} = \{1, \ldots, r\}$ be resource types. Assume that each plan $x \in \mathcal{D}$ induces uncertain evaluations (degrees) for:

$$\varphi_i : \mathcal{D} \to \mathrm{Dom}(M) \qquad (i = 1, \ldots, s) \quad \text{(goal-achievement degrees)},$$

$$\psi_u : \mathcal{D} \to \mathrm{Dom}(M) \qquad (u = 1, \ldots, r) \quad \text{(resource-feasibility/cost satisfaction degrees)},$$

and optionally additional uncertain criteria

$$\eta_v : \mathcal{D} \to \mathrm{Dom}(M) \qquad (v = 1, \ldots, q).$$

Let $k := s + r + q$ and enumerate these criteria as a single family

$$\zeta_1, \ldots, \zeta_k : \mathcal{D} \to \mathrm{Dom}(M),$$

where $(\zeta_1, \ldots, \zeta_s) = (\varphi_1, \ldots, \varphi_s)$, $(\zeta_{s+1}, \ldots, \zeta_{s+r}) = (\psi_1, \ldots, \psi_r)$, and $(\zeta_{s+r+1}, \ldots, \zeta_k) = (\eta_1, \ldots, \eta_q)$ (if $q = 0$, omit the last block).

**(1) Aggregation in the uncertain degree-domain.** Choose a total aggregation operator

$$\mathrm{Agg}_M : \mathrm{Dom}(M)^k \times \Delta_k \longrightarrow \mathrm{Dom}(M), \qquad \Delta_k := \Big\{ w \in [0, 1]^k : \sum_{j=1}^{k} w_j = 1 \Big\},$$

and fix a weight vector $w \in \Delta_k$.

**(2) Scoring and selection.** Fix a total scoring functional

$$\mathrm{Score}_M : \mathrm{Dom}(M) \to \mathbb{R}.$$

Define the overall uncertain strategic utility of a plan $x \in \mathcal{D}$ by

$$U_M(x) := \mathrm{Agg}_M\big(\zeta_1(x), \ldots, \zeta_k(x); w\big) \in \mathrm{Dom}(M),$$

and the induced real score by $\mathrm{Score}(x) := \mathrm{Score}_M(U_M(x)) \in \mathbb{R}$. An *uncertain strategic decision* is any plan

$$x^\star \in \arg\max_{x \in \mathcal{D}} \ \mathrm{Score}_M\big(U_M(x)\big).$$

**(3) Uncertain-set output (strategic utility set).** Define the mapping

$$\mu_{\mathcal{U}} : \mathcal{D} \to \mathrm{Dom}(M), \qquad \mu_{\mathcal{U}}(x) := U_M(x).$$

Then $\mathcal{U} = (\mathcal{D}, \mu_{\mathcal{U}})$ is called the *strategic uncertain utility set* (a U-set of type $M$) induced by the instance.

**Theorem 3.8.3** (Uncertain-set structure and well-definedness of USDM). *Let a USDM instance of type M be given as in Definition 3.8.2. Assume:*

(A1) $p \geq 1$ *and* $\mathcal{D} \subseteq \{0, 1\}^p$ *is finite and nonempty.*

(A2) $\mathrm{Dom}(M) \subseteq [0, 1]^d$ *is nonempty and each criterion map* $\zeta_j : \mathcal{D} \to \mathrm{Dom}(M)$ *is total.*

(A3) $\mathrm{Agg}_M : \mathrm{Dom}(M)^k \times \Delta_k \to \mathrm{Dom}(M)$ *is total and* $w \in \Delta_k$.

(A4) $\mathrm{Score}_M : \mathrm{Dom}(M) \to \mathbb{R}$ *is total.*

*Then:*

(i) *The mapping* $\mu_{\mathcal{U}} : \mathcal{D} \to \mathrm{Dom}(M)$ *defined by* $\mu_{\mathcal{U}}(x) = \mathrm{Agg}_M(\zeta_1(x), \ldots, \zeta_k(x); w)$ *is well-defined; hence* $\mathcal{U} = (\mathcal{D}, \mu_{\mathcal{U}})$ *is a well-defined uncertain set (U-set) of type M on* $\mathcal{D}$.

(ii) *The binary relation* $\succeq_M$ *on* $\mathcal{D}$ *defined by*

$$x \succeq_M y \iff \mathrm{Score}_M(\mu_{\mathcal{U}}(x)) \geq \mathrm{Score}_M(\mu_{\mathcal{U}}(y))$$

*is a total preorder.*

(iii) *The set* $\arg\max_{x \in \mathcal{D}} \mathrm{Score}_M(\mu_{\mathcal{U}}(x))$ *is nonempty; hence an uncertain strategic decision exists.*

*Proof.* **(i) Uncertain-set structure.** Fix $x \in \mathcal{D}$. By (A2), each $\zeta_j(x) \in \mathrm{Dom}(M)$ is well-defined. By (A3), applying the total map $\mathrm{Agg}_M(\cdot; w)$ to the $k$-tuple $(\zeta_1(x), \ldots, \zeta_k(x))$ yields a unique value

$$\mu_{\mathcal{U}}(x) = \mathrm{Agg}_M(\zeta_1(x), \ldots, \zeta_k(x); w) \in \mathrm{Dom}(M).$$

Thus $\mu_{\mathcal{U}} : \mathcal{D} \to \mathrm{Dom}(M)$ is a well-defined mapping, and therefore $\mathcal{U} = (\mathcal{D}, \mu_{\mathcal{U}})$ is a U-set of type $M$.

**(ii) Total preorder.** By (A4), $\mathrm{Score}_M(\mu_{\mathcal{U}}(x)) \in \mathbb{R}$ is well-defined for all $x \in \mathcal{D}$. Since $\geq$ on $\mathbb{R}$ is reflexive, transitive, and total, the induced relation $\succeq_M$ inherits these properties, hence is a total preorder.

**(iii) Existence of an optimizer.** Because $\mathcal{D}$ is finite and nonempty by (A1), the finite set of real numbers $\{\mathrm{Score}_M(\mu_{\mathcal{U}}(x)) : x \in \mathcal{D}\}$ attains a maximum. Therefore $\arg\max_{x \in \mathcal{D}} \mathrm{Score}_M(\mu_{\mathcal{U}}(x)) \neq \varnothing$. $\square$

Table 3.8 presents related uncertainty-model variants of Fuzzy Strategic Decision-Making.

Table 3.8: Related uncertainty-model variants of Fuzzy Strategic Decision-Making.

| $k$ | Related Fuzzy Strategic Decision-Making variant(s) |
|---|---|
| 1 | Fuzzy Strategic Decision-Making |
| 2 | Intuitionistic Fuzzy Strategic Decision-Making |
| 3 | Hesitant Fuzzy Strategic Decision-Making |
| 3 | Spherical Fuzzy Strategic Decision-Making |
| 3 | Neutrosophic Strategic Decision-Making |
| $n$ | Plithogenic Strategic Decision-Making |

## 3.9 Fuzzy Multi-Expert Decision-Making

Multi-expert decision-making aggregates several experts' evaluations or pairwise preferences, weights experts, measures consensus, and produces a collective ranking or choice [396, 397]. Fuzzy multi-expert decision-making represents experts' judgments as fuzzy numbers or linguistic terms, aggregates them via fuzzy operators, enforces consensus, and ranks alternatives robustly [398].

**Definition 3.9.1** (Fuzzy Multi-Expert Decision-Making (FMEDM)). (cf. [399, 400]) Let $\mathcal{A} = \{A_1, \ldots, A_m\}$ be a finite set of alternatives, $\mathcal{K} = \{K_1, \ldots, K_p\}$ a finite set of criteria, and $\mathcal{E} = \{E_1, \ldots, E_n\}$ a finite set of experts.

**(1) Expert contribution factors.** Assign each expert $E_i$ a contribution factor (weight) $c_i \in [0, 1]$ such that

$$\sum_{i=1}^{n} c_i = 1.$$

**(2) Fuzzy evaluations.** Let $\mathbb{F}$ denote a chosen class of fuzzy numbers on $\mathbb{R}$. Each expert $E_i$ provides a fuzzy evaluation

$$\tilde{x}_{rj}^{(i)} \in \mathbb{F} \qquad (r = 1, \ldots, m; \ j = 1, \ldots, p),$$

interpreted as the (uncertain/vague) performance of alternative $A_r$ under criterion $K_j$.

**(3) Multi-expert aggregation (group evaluation).** Fix fuzzy-number operations $\oplus$ (addition) and $\otimes$ (scalar multiplication by $c_i$). The aggregated (group) fuzzy evaluation is defined entrywise by the weighted fuzzy averaging operator

$$\tilde{x}_{rj} := \bigoplus_{i=1}^{n} \left( c_i \otimes \tilde{x}_{rj}^{(i)} \right), \qquad (r = 1, \ldots, m; \ j = 1, \ldots, p).$$

**(4) Aggregation across criteria and decision rule.** Let $w_1, \ldots, w_p \geq 0$ be criterion weights with $\sum_{j=1}^{p} w_j = 1$ (crisp weights; fuzzy weights can be used analogously). Define the fuzzy overall score of $A_r$ by

$$\tilde{s}_r := \bigoplus_{j=1}^{p} \left( w_j \otimes \tilde{x}_{rj} \right) \in \mathbb{F}.$$

Let $\Phi : \mathbb{F} \to \mathbb{R}$ be a ranking functional (e.g., a defuzzification map).  A (recommended) group decision is any

$$A^{\star} \in \arg \max_{1 \leq r \leq m} \Phi(\tilde{s}_r).$$

**(Common instantiation: trapezoidal fuzzy numbers).**  A standardized trapezoidal fuzzy number (STFN) is a quadruple $\tilde{x} = (a^l, a^m, a^n, a^u)$ with $a^l \leq a^m \leq a^n \leq a^u$ and membership function

$$\mu_{\tilde{x}}(t) = \begin{cases} \dfrac{t - a^l}{a^m - a^l}, & a^l \leq t \leq a^m, \\ 1, & a^m \leq t \leq a^n, \\ \dfrac{a^u - t}{a^u - a^n}, & a^n \leq t \leq a^u, \\ 0, & \text{otherwise.} \end{cases}$$

For $c \geq 0$, define $c \otimes (a^l, a^m, a^n, a^u) := (ca^l, ca^m, ca^n, ca^u)$ and

$$(a^l, a^m, a^n, a^u) \oplus (b^l, b^m, b^n, b^u) := (a^l + b^l, \ a^m + b^m, \ a^n + b^n, \ a^u + b^u),$$

so the multi-expert aggregation in (3) becomes componentwise weighted averaging.

**Definition 3.9.2** (Uncertain multi-expert decision-making (UMEDM) of type $M$).  Let

$$\mathcal{A} = \{A_1, \ldots, A_m\} \neq \varnothing \quad \text{(alternatives)}, \qquad \mathcal{C} = \{C_1, \ldots, C_n\} \neq \varnothing \quad \text{(criteria)},$$

and

$$\mathcal{E} = \{e_1, \ldots, e_p\} \neq \varnothing \quad \text{(experts)}.$$

Fix an *uncertain model* $M$ with degree-domain $\mathrm{Dom}(M) \subseteq [0, 1]^d$ (for some $d \geq 1$).

**(1) Expert and criterion weights.**  Let

$$\alpha = (\alpha_1, \ldots, \alpha_p) \in \Delta_p, \qquad \Delta_p := \Big\{ \alpha \in [0, 1]^p : \sum_{h=1}^{p} \alpha_h = 1 \Big\},$$

be expert weights, and let

$$w = (w_1, \ldots, w_n) \in \Delta_n$$

be criterion weights.

**(2) Expert-wise uncertain evaluations.**  Each expert $e_h$ provides an uncertain decision matrix

$$R^{(h)} = \big(r_{ij}^{(h)}\big) \in \mathrm{Dom}(M)^{m \times n}, \qquad r_{ij}^{(h)} \in \mathrm{Dom}(M) \text{ evaluates } A_i \text{ w.r.t. } C_j \text{ by expert } e_h.$$

**(3) Aggregation operators in $\mathrm{Dom}(M)$.**  Choose total aggregation operators

$$\mathrm{Agg}_{\mathcal{E}} : \mathrm{Dom}(M)^p \times \Delta_p \to \mathrm{Dom}(M) \quad \text{(across experts)},$$

$$\mathrm{Agg}_{\mathcal{C}} : \mathrm{Dom}(M)^n \times \Delta_n \to \mathrm{Dom}(M) \quad \text{(across criteria)},$$

and a total scoring functional

$$\mathrm{Score}_M : \mathrm{Dom}(M) \to \mathbb{R}.$$

**(4) Collective evaluation, utility, and selection.** Define the *collective uncertain decision matrix* $R^c = (r_{ij}^c) \in \mathrm{Dom}(M)^{m \times n}$ entrywise by

$$r_{ij}^c := \mathrm{Agg}_{\mathcal{E}}\big(r_{ij}^{(1)}, \ldots, r_{ij}^{(p)}; {}_a lpha\big) \in \mathrm{Dom}(M), \qquad i = 1, \ldots, m, \ j = 1, \ldots, n.$$

For each alternative $A_i$, define its *overall uncertain utility degree* by

$$u_i := \mathrm{Agg}_{\mathcal{C}}\big(r_{i1}^c, \ldots, r_{in}^c; w\big) \in \mathrm{Dom}(M),$$

and its induced real score by

$$U_i := \mathrm{Score}_M(u_i) \in \mathbb{R}.$$

A *(recommended) uncertain multi-expert decision* is any

$$A^{\star} \in \arg\max_{1 \le i \le m} \ U_i \ = \ \arg\max_{A_i \in \mathcal{A}} \ \mathrm{Score}_M(u_i).$$

**(5) Uncertain-set output (multi-expert utility set).** Define

$$\mu_{\mathcal{U}} : \mathcal{A} \to \mathrm{Dom}(M), \qquad \mu_{\mathcal{U}}(A_i) := u_i.$$

Then $\mathcal{U} := (\mathcal{A}, \mu_{\mathcal{U}})$ is called the *multi-expert uncertain utility set* (U-set of type $M$) induced by the instance.

**Theorem 3.9.3** (Uncertain-set structure and well-definedness of UMEDM). *Let a UMEDM instance of type $M$ be given as in Definition 3.9.2. Assume:*

(A1) *$\mathcal{A}, \mathcal{C}, \mathcal{E}$ are finite and nonempty.*

(A2) *$\mathrm{Dom}(M) \subseteq [0,1]^d$ is nonempty.*

(A3) *For every expert $e_h$ and every $(i,j)$, the evaluation entry $r_{ij}^{(h)} \in \mathrm{Dom}(M)$ is specified (i.e. each $R^{(h)}$ is a total $m \times n$ matrix over $\mathrm{Dom}(M)$).*

(A4) *$\alpha \in \Delta_p$ and $w \in \Delta_n$.*

(A5) *$\mathrm{Agg}_{\mathcal{E}}$ and $\mathrm{Agg}_{\mathcal{C}}$ are total functions with codomain $\mathrm{Dom}(M)$, and $\mathrm{Score}_M$ is a total function with codomain $\mathbb{R}$.*

*Then:*

(i) *$\mu_{\mathcal{U}} : \mathcal{A} \to \mathrm{Dom}(M)$ is well-defined; hence $\mathcal{U} = (\mathcal{A}, \mu_{\mathcal{U}})$ is a well-defined uncertain set (U-set) of type $M$ on $\mathcal{A}$.*

(ii) *The binary relation $\succeq_M$ on $\mathcal{A}$ given by*

$$A_i \succeq_M A_{\ell} \iff \mathrm{Score}_M\big(\mu_{\mathcal{U}}(A_i)\big) \ge \mathrm{Score}_M\big(\mu_{\mathcal{U}}(A_{\ell})\big)$$

*is a total preorder.*

(iii) *The maximizer set* $\arg\max_{A_i \in \mathcal{A}} \mathrm{Score}_M(\mu_{\mathcal{U}}(A_i))$ *is nonempty; in particular, a UMEDM solution* $A^\star$ *exists.*

*Proof.* **(i)** Fix $A_i \in \mathcal{A}$ and $C_j \in \mathcal{C}$. By (A3), the $p$-tuple $(r_{ij}^{(1)}, \ldots, r_{ij}^{(p)}) \in \mathrm{Dom}(M)^p$ is well-defined. By (A4)–(A5), applying the total map $\mathrm{Agg}_{\mathcal{E}}(\cdot\,;\alpha)$ yields a unique value

$$r_{ij}^c = \mathrm{Agg}_{\mathcal{E}}(r_{ij}^{(1)}, \ldots, r_{ij}^{(p)}; \alpha) \in \mathrm{Dom}(M).$$

Thus $R^c \in \mathrm{Dom}(M)^{m \times n}$ is well-defined. Again by (A4)–(A5), the $n$-tuple $(r_{i1}^c, \ldots, r_{in}^c) \in \mathrm{Dom}(M)^n$ is well-defined and

$$u_i = \mathrm{Agg}_{\mathcal{C}}(r_{i1}^c, \ldots, r_{in}^c; w) \in \mathrm{Dom}(M)$$

is uniquely determined. Therefore $\mu_{\mathcal{U}}(A_i) := u_i$ is well-defined for each $A_i$, hence $\mu_{\mathcal{U}} : \mathcal{A} \to \mathrm{Dom}(M)$ is a well-defined mapping and $\mathcal{U} = (\mathcal{A}, \mu_{\mathcal{U}})$ is a U-set of type $M$.

**(ii)** By (A5), each score $\mathrm{Score}_M(\mu_{\mathcal{U}}(A_i)) \in \mathbb{R}$ is well-defined. Since $\geq$ on $\mathbb{R}$ is reflexive, transitive, and total, the induced relation $\succeq_M$ is reflexive, transitive, and total; hence it is a total preorder.

**(iii)** Because $\mathcal{A}$ is finite and nonempty by (A1), the finite set of real numbers $\{\mathrm{Score}_M(\mu_{\mathcal{U}}(A_i)) : A_i \in \mathcal{A}\}$ attains a maximum. Therefore the argmax set is nonempty, so at least one maximizer $A^\star$ exists. $\qquad\square$

Table 3.9 presents related uncertainty-model variants of Fuzzy Multi-Expert Decision-Making.

Table 3.9: Related uncertainty-model variants of Fuzzy Multi-Expert Decision-Making.

| $k$ | **Related Fuzzy Multi-Expert Decision-Making variant(s)** |
|---|---|
| 1 | Fuzzy Multi-Expert Decision-Making |
| 2 | Intuitionistic Fuzzy Multi-Expert Decision-Making |
| 3 | Hesitant Fuzzy Multi-Expert Decision-Making |
| 3 | Spherical Fuzzy Multi-Expert Decision-Making |
| 3 | Neutrosophic Multi-Expert Decision-Making [401, 402] |
| $n$ | Plithogenic Multi-Expert Decision-Making |

## 3.10 Fuzzy Multi-Stage Decision-Making

Multi-stage decision-making selects actions across sequential stages, where earlier choices affect later feasible sets, costs, and outcomes, optimizing total performance [403, 404]. Fuzzy multi-stage decision-making models stage-dependent evaluations, constraints, or transitions with fuzzy sets, propagating uncertainty through stages to choose an adaptive plan [405, 406].

**Definition 3.10.1** (Fuzzy Multi-Stage Decision-Making (FMSDM)). [405, 406] Fix a finite horizon $m \in \mathbb{N}$. For each stage $i \in \{1, \ldots, m\}$, let $S_i$ be a (finite) state set and, for each $s \in S_i$, let $D_i(s)$ be a nonempty (finite) set of admissible decisions. Let the (crisp) transition map be

$$\tau_i : S_i \times D_i(s) \to S_{i+1} \qquad (i = 1, \ldots, m-1).$$

Assume that each stage-wise cost is given by a fuzzy number:

$$\beta_i : S_i \times D_i(s) \to \mathcal{FN}(\mathbb{R}),$$

where $\mathcal{FN}(\mathbb{R})$ denotes the class of fuzzy numbers on $\mathbb{R}$.

A (deterministic) policy is a sequence $\pi = (\pi_1, \ldots, \pi_m)$ with $\pi_i(s) \in D_i(s)$. Given an initial state $s_1 \in S_1$, the induced trajectory is

$$s_{i+1} = \tau_i\big(s_i, \pi_i(s_i)\big) \quad (i = 1, \ldots, m-1),$$

and the induced *composite fuzzy cost* is the fuzzy number

$$\boldsymbol{\beta}^\pi(s_1) := \beta_1\big(s_1, \pi_1(s_1)\big) \ \oplus \ \beta_2\big(s_2, \pi_2(s_2)\big) \ \oplus \ \cdots \ \oplus \ \beta_m\big(s_m, \pi_m(s_m)\big),$$

where $\oplus$ is the *extended (fuzzy) sum* defined by $\alpha$-cuts: for fuzzy numbers $X, Y$ and $\alpha \in (0, 1]$,

$$[X \oplus Y]_\alpha = [X]_\alpha + [Y]_\alpha \quad \text{(Minkowski sum of intervals)}.$$

To compare fuzzy costs under a minimization objective, fix the *$\alpha$-cut dominance preorder* $\preceq$ on $\mathcal{FN}(\mathbb{R})$: for fuzzy numbers $X, Y$,

$$X \preceq Y \iff \forall \alpha \in (0, 1] : \ \inf[X]_\alpha \leq \inf[Y]_\alpha \ \text{ and } \ \sup[X]_\alpha \leq \sup[Y]_\alpha.$$

Write $X \prec Y$ if $X \preceq Y$ and not $Y \preceq X$.

A policy $\pi^\star$ is called *(dominance-)optimal* at $s_1$ if there is no policy $\pi$ such that $\boldsymbol{\beta}^\pi(s_1) \prec \boldsymbol{\beta}^{\pi^\star}(s_1)$. Equivalently, the (possibly set-valued) solution set is the set of $\preceq$-minimal composite costs:

$$\Pi^\star(s_1) := \Big\{ \pi : \ \nexists \pi' \text{ with } \boldsymbol{\beta}^{\pi'}(s_1) \prec \boldsymbol{\beta}^\pi(s_1) \Big\}.$$

**Definition 3.10.2** (Extended minimum cost and fuzzy set of best policies). Assume $\Pi$ is a finite set of admissible policies (e.g., all deterministic policies over finite $S_i$ and $D_i(\cdot)$). Define the *extended minimum* of fuzzy numbers $(X_k)_{k=1}^N$ by Zadeh's extension principle:

$$\mu_{\widetilde{\min}(X_1, \ldots, X_N)}(u) := \sup_{u = \min(u_1, \ldots, u_N)} \ \min_{1 \leq k \leq N} \ \mu_{X_k}(u_k), \qquad u \in \mathbb{R}.$$

Then the *minimum (extended) fuzzy cost* at $s_1$ is

$$\boldsymbol{\beta}^\star(s_1) := \widetilde{\min}\big\{ \boldsymbol{\beta}^\pi(s_1) : \ \pi \in \Pi \big\}.$$

To associate a *fuzzy set of best policies* with $\boldsymbol{\beta}^\star(s_1)$, fix any well-defined matching (similarity) index Match : $\mathcal{FN}(\mathbb{R}) \times \mathcal{FN}(\mathbb{R}) \to [0, 1]$. A common choice is the *intersection height*:

$$\text{Match}(X, Y) := \sup_{u \in \mathbb{R}} \min\{\mu_X(u), \mu_Y(u)\}.$$

Define the fuzzy set $\widetilde{\Pi}^\star(s_1)$ on $\Pi$ by

$$\mu_{\widetilde{\Pi}^\star(s_1)}(\pi) := \text{Match}\big(\boldsymbol{\beta}^\pi(s_1), \boldsymbol{\beta}^\star(s_1)\big).$$

In particular, if a unique policy $\pi^\star$ strictly dominates all others (under $\prec$), then $\widetilde{\Pi}^\star(s_1)$ collapses to the crisp singleton $\{\pi^\star\}$.

We now define *Uncertain Multi-Stage Decision-Making* (UMSDM) by extending stage-wise evaluations from fuzzy quantities to a general uncertain model $M$ with degree-domain $\mathrm{Dom}(M) \subseteq [0,1]^k$.

**Definition 3.10.3** (Uncertain Multi-Stage Decision-Making (UMSDM) of type $M$). Let $M$ be an uncertain model with degree-domain

$$\mathrm{Dom}(M) \subseteq [0,1]^k$$

for some integer $k \geq 1$. Fix a finite horizon $m \in \mathbb{N}$.

For each stage $i \in \{1, \ldots, m\}$, let $S_i$ be a finite nonempty state set, and for each $s \in S_i$, let $D_i(s)$ be a finite nonempty set of admissible decisions at stage $i$. Define the feasible state-decision set

$$\mathcal{D}_i := \{(s,d) : s \in S_i, \ d \in D_i(s)\}.$$

For each $i = 1, \ldots, m-1$, let

$$\tau_i : \mathcal{D}_i \longrightarrow S_{i+1}$$

be a (crisp) transition map, and for each $i = 1, \ldots, m$, let

$$\beta_i : \mathcal{D}_i \longrightarrow \mathrm{Dom}(M)$$

be the *uncertain stage-evaluation map*, where $\beta_i(s,d)$ represents the uncertain cost, utility, reward, or performance degree associated with taking decision $d$ in state $s$ at stage $i$.

Assume further that a model-dependent stage-composition operator

$$\mathrm{Comp}_M : \mathrm{Dom}(M)^m \longrightarrow \mathrm{Dom}(M)$$

and a model-dependent ranking functional

$$\mathrm{Score}_M : \mathrm{Dom}(M) \longrightarrow \mathbb{R}$$

are fixed.

A *deterministic policy* is a sequence

$$\pi = (\pi_1, \ldots, \pi_m),$$

where each

$$\pi_i : S_i \longrightarrow \bigcup_{s \in S_i} D_i(s)$$

satisfies $\pi_i(s) \in D_i(s)$ for all $s \in S_i$. Let $\Pi_i$ denote the set of all such stage-$i$ decision rules, and let

$$\Pi := \Pi_1 \times \cdots \times \Pi_m$$

be the set of all deterministic policies.

Given an initial state $s_1 \in S_1$ and a policy $\pi \in \Pi$, the induced state trajectory

$$s_1^\pi, s_2^\pi, \ldots, s_m^\pi$$

is defined recursively by

$$s_1^\pi := s_1, \qquad s_{i+1}^\pi := \tau_i\big(s_i^\pi, \pi_i(s_i^\pi)\big) \quad (i = 1, \ldots, m-1).$$

The induced stage-wise uncertain evaluations are

$$b_i^\pi(s_1) := \beta_i\big(s_i^\pi, \pi_i(s_i^\pi)\big) \in \mathrm{Dom}(M) \qquad (i = 1, \dots, m).$$

The *composite uncertain performance* of $\pi$ at $s_1$ is

$$B_M^\pi(s_1) := \mathrm{Comp}_M\big(b_1^\pi(s_1), \dots, b_m^\pi(s_1)\big) \in \mathrm{Dom}(M),$$

and the associated scalarized performance value is

$$J_M^\pi(s_1) := \mathrm{Score}_M\big(B_M^\pi(s_1)\big) \in \mathbb{R}.$$

Under a minimization objective, define the induced policy preorder by

$$\pi \preceq_M \rho \quad \Longleftrightarrow \quad J_M^\pi(s_1) \le J_M^\rho(s_1).$$

A policy $\pi^\star \in \Pi$ is called *optimal* at $s_1$ if

$$J_M^{\pi^\star}(s_1) = \min_{\pi \in \Pi} J_M^\pi(s_1).$$

Equivalently, the optimal-policy set is

$$\Pi_M^\star(s_1) := \arg\min_{\pi \in \Pi} J_M^\pi(s_1).$$

If a maximization objective is intended instead, one replaces $\arg\min$ with $\arg\max$, or equivalently replaces $\le$ with $\ge$ in the induced preorder.

**Theorem 3.10.4** (Well-definedness of UMSDM of type $M$)**.** *Let $M$ be an uncertain model with degree-domain $\mathrm{Dom}(M) \subseteq [0,1]^k$, and let*

$$\mathfrak{U}_M = \big((S_i)_{i=1}^m, (D_i)_{i=1}^m, (\tau_i)_{i=1}^{m-1}, (\beta_i)_{i=1}^m, \mathrm{Comp}_M, \mathrm{Score}_M\big)$$

*be a UMSDM instance as in Definition 3.10.3. Assume:*

*(A1) each $S_i$ is finite and nonempty;*

*(A2) each $D_i(s)$ is finite and nonempty for every $i \in \{1, \dots, m\}$ and $s \in S_i$;*

*(A3) each transition map*

$$\tau_i : \mathcal{D}_i \to S_{i+1} \quad (i = 1, \dots, m-1)$$

*is a total function;*

*(A4) each stage-evaluation map*

$$\beta_i : \mathcal{D}_i \to \mathrm{Dom}(M) \quad (i = 1, \dots, m)$$

*is a total function;*

*(A5) the composition operator*

$$\mathrm{Comp}_M : \mathrm{Dom}(M)^m \to \mathrm{Dom}(M)$$

*is a total function;*

*(A6) the ranking functional*

$$\text{Score}_M : \text{Dom}(M) \to \mathbb{R}$$

*is a total function.*

*Then, for every initial state $s_1 \in S_1$, the following statements hold:*

*(i) the policy set $\Pi$ is finite and nonempty;*

*(ii) for every policy $\pi \in \Pi$, the induced trajectory*

$$s_1^\pi, s_2^\pi, \ldots, s_m^\pi$$

*is uniquely determined;*

*(iii) for every policy $\pi \in \Pi$, the stage-wise uncertain evaluations*

$$b_i^\pi(s_1) \in \text{Dom}(M) \qquad (i = 1, \ldots, m)$$

*and the composite uncertain performance*

$$B_M^\pi(s_1) \in \text{Dom}(M)$$

*are well-defined;*

*(iv) for every policy $\pi \in \Pi$, the scalar value*

$$J_M^\pi(s_1) = \text{Score}_M\big(B_M^\pi(s_1)\big) \in \mathbb{R}$$

*is well-defined;*

*(v) the relation $\preceq_M$ on $\Pi$, defined by*

$$\pi \preceq_M \rho \iff J_M^\pi(s_1) \le J_M^\rho(s_1),$$

*is a total preorder;*

*(vi) the optimal-policy set*

$$\Pi_M^\star(s_1) = \arg\min_{\pi \in \Pi} J_M^\pi(s_1)$$

*is nonempty.*

*Hence Uncertain Multi-Stage Decision-Making of type $M$ is well-defined.*

*Proof.* First, for each stage $i$, define

$$\Pi_i := \{\pi_i : S_i \to \bigcup_{s \in S_i} D_i(s) \ : \ \pi_i(s) \in D_i(s) \text{ for all } s \in S_i\}.$$

Since $S_i$ is finite and each $D_i(s)$ is finite and nonempty, the number of such functions is

$$|\Pi_i| = \prod_{s \in S_i} |D_i(s)|,$$

which is a positive finite integer. Therefore each $\Pi_i$ is finite and nonempty, and hence

$$\Pi = \Pi_1 \times \cdots \times \Pi_m$$

is also finite and nonempty. This proves $(i)$.

Next, fix an initial state $s_1 \in S_1$ and a policy $\pi \in \Pi$. We show that the induced trajectory is uniquely determined. Set $s_1^\pi := s_1$. Assume inductively that $s_i^\pi \in S_i$ has been uniquely defined for some $i \in \{1, \ldots, m-1\}$. Because $\pi_i(s_i^\pi) \in D_i(s_i^\pi)$, the pair

$$\big(s_i^\pi, \pi_i(s_i^\pi)\big) \in \mathcal{D}_i.$$

By assumption $(A3)$, the transition map $\tau_i$ is total, so

$$s_{i+1}^\pi := \tau_i\big(s_i^\pi, \pi_i(s_i^\pi)\big) \in S_{i+1}$$

exists and is unique. By induction, the whole trajectory

$$s_1^\pi, s_2^\pi, \ldots, s_m^\pi$$

exists and is unique. Thus $(ii)$ holds.

For each stage $i$, since

$$\big(s_i^\pi, \pi_i(s_i^\pi)\big) \in \mathcal{D}_i$$

and $\beta_i : \mathcal{D}_i \to \mathrm{Dom}(M)$ is total by $(A4)$, the value

$$b_i^\pi(s_1) := \beta_i\big(s_i^\pi, \pi_i(s_i^\pi)\big)$$

is well-defined and belongs to $\mathrm{Dom}(M)$. Hence the $m$-tuple

$$\big(b_1^\pi(s_1), \ldots, b_m^\pi(s_1)\big) \in \mathrm{Dom}(M)^m$$

is well-defined. By $(A5)$, the composition operator $\mathrm{Comp}_M$ is total, so

$$B_M^\pi(s_1) = \mathrm{Comp}_M\big(b_1^\pi(s_1), \ldots, b_m^\pi(s_1)\big) \in \mathrm{Dom}(M)$$

is well-defined. This proves $(iii)$.

By $(A6)$, $\mathrm{Score}_M$ is a total map from $\mathrm{Dom}(M)$ into $\mathbb{R}$. Therefore

$$J_M^\pi(s_1) := \mathrm{Score}_M\big(B_M^\pi(s_1)\big) \in \mathbb{R}$$

is well-defined for every $\pi \in \Pi$. Thus $(iv)$ holds.

Now define, for $\pi, \rho \in \Pi$,

$$\pi \preceq_M \rho \iff J_M^\pi(s_1) \leq J_M^\rho(s_1).$$

Since $\leq$ on $\mathbb{R}$ is reflexive, transitive, and total, the induced relation $\preceq_M$ on $\Pi$ is also reflexive, transitive, and total. Hence $\preceq_M$ is a total preorder. This proves $(v)$.

Finally, because $\Pi$ is finite and nonempty by $(i)$, the set

$$\{J_M^\pi(s_1) : \pi \in \Pi\} \subseteq \mathbb{R}$$

is a finite nonempty set of real numbers. Every finite nonempty subset of $\mathbb{R}$ attains its minimum. Therefore there exists at least one policy $\pi^\star \in \Pi$ such that

$$J_M^{\pi^\star}(s_1) = \min_{\pi \in \Pi} J_M^\pi(s_1).$$

Equivalently,

$$\Pi_M^\star(s_1) = \arg\min_{\pi \in \Pi} J_M^\pi(s_1) \neq \varnothing.$$

Thus $(vi)$ holds.

All required objects therefore exist and are unambiguously defined. Hence UMSDM of type $M$ is well-defined. □

Table 3.10 presents related uncertainty-model variants of Fuzzy Multi-Stage Decision-Making.

Table 3.10: Related uncertainty-model variants of Fuzzy Multi-Stage Decision-Making.

| $k$ | Related Fuzzy Multi-Stage Decision-Making variant(s) |
|---|---|
| 1 | Fuzzy Multi-Stage Decision-Making |
| 2 | Intuitionistic Fuzzy Multi-Stage Decision-Making |
| 3 | Hesitant Fuzzy Multi-Stage Decision-Making |
| 3 | Spherical Fuzzy Multi-Stage Decision-Making |
| 3 | Neutrosophic Multi-Stage Decision-Making |
| $n$ | Plithogenic Multi-Stage Decision-Making |

## 3.11 Fuzzy Multi-Level Decision-Making

Multi-level decision-making models hierarchical leaders and followers, where upper-level decisions constrain lower-level responses, solved as bilevel or hierarchical optimization problems [407–409]. Fuzzy multi-level decision-making introduces fuzzy goals, constraints, or preferences at different hierarchy levels, yielding compromise solutions under imprecise leader–follower information.

**Definition 3.11.1** (Fuzzy $L$-Level Decision-Making). Let $L \geq 2$ be the number of decision levels. For each level $\ell \in \{1, \ldots, L\}$, let $x_\ell \in X_\ell \subseteq \mathbb{R}^{n_\ell}$ be the decision vector controlled by the $\ell$-th level, and set

$$x := (x_1, \ldots, x_L) \in X := \prod_{\ell=1}^{L} X_\ell.$$

Assume that uncertainty/imprecision is modeled by fuzzy numbers. Let $\tilde{\mathbb{R}}$ denote a chosen class of fuzzy numbers on $\mathbb{R}$. For each level $\ell$, let the (fuzzy-valued) objective be

$$\tilde{f}_\ell : X \to \tilde{\mathbb{R}}.$$

Let the system constraints be given by fuzzy-valued mappings

$$\tilde{g}_r : X \to \tilde{\mathbb{R}}, \qquad r = 1, \ldots, p.$$

Fix:

- an *order-inducing ranking functional* $\mathcal{R} : \tilde{\mathbb{R}} \to \mathbb{R}$ (e.g., a defuzzification/ranking operator) that will be used to compare fuzzy objectives;

- a *feasibility satisfaction level* $\eta \in (0, 1]$.

For a fuzzy number $\tilde{z} \in \tilde{\mathbb{R}}$ with membership function $\mu_{\tilde{z}} : \mathbb{R} \to [0, 1]$, define the *possibility* that $\tilde{z} \leq 0$ by

$$\mathrm{Pos}(\tilde{z} \leq 0) := \sup_{t \leq 0} \mu_{\tilde{z}}(t) \in [0, 1].$$

Define the $\eta$-feasible set

$$F_\eta := \Big\{ x \in X : \ \mathrm{Pos}(\tilde{g}_r(x) \leq 0) \geq \eta \text{ for all } r = 1, \dots, p \Big\}.$$

Define recursively the *rational-response correspondences* $\mathcal{S}_\ell$ (backward induction):

$$\mathcal{S}_L(x_{1:L-1}) := \arg \min_{x_L \in X_L} \Big\{ \mathcal{R}\big(\tilde{f}_L(x_{1:L-1}, x_L)\big) : (x_{1:L-1}, x_L) \in F_\eta \Big\},$$

$$\mathcal{S}_\ell(x_{1:\ell-1}) := \arg \min_{x_\ell \in X_\ell} \Big\{ \mathcal{R}\big(\tilde{f}_\ell(x_{1:\ell-1}, x_\ell, x_{\ell+1:L})\big) : \exists x_{\ell+1:L} \in \mathcal{S}_{\ell+1}(x_{1:\ell}), \ (x_{1:L}) \in F_\eta \Big\},$$

for $\ell = L-1, L-2, \dots, 1$, where $x_{i:j} := (x_i, \dots, x_j)$.

A vector $x^\star = (x_1^\star, \dots, x_L^\star) \in F_\eta$ is called a *fuzzy L-level (Stackelberg) solution* if

$$x_{2:L}^\star \in \mathcal{S}_2(x_1^\star) \quad \text{and} \quad x_1^\star \in \mathcal{S}_1(\varnothing),$$

equivalently, if $x^\star$ is generated by the above backward-induction optimal reactions.

**Remark 3.11.2** (Recovery of classical multilevel programming). If all data are crisp (so $\tilde{f}_\ell$ and $\tilde{g}_r$ take values in $\mathbb{R}$), $\mathrm{Pos}(\tilde{g}_r(x) \leq 0) \in \{0, 1\}$, and one takes $\eta = 1$ with $\mathcal{R}$ equal to the identity, then Definition 3.11.1 reduces to the standard (crisp) $L$-level Stackelberg decision/optimization model.

**Theorem 3.11.3** (Well-definedness / existence under compactness). *Assume:*

1. *each $X_\ell$ is nonempty and compact;*

2. *$F_\eta$ is nonempty and closed in $X$ (hence compact);*

3. *for each $\ell$, the scalarized objective $x \mapsto \mathcal{R}(\tilde{f}_\ell(x))$ is continuous on $X$.*

*Then for every $\ell$, the correspondence $\mathcal{S}_\ell$ is nonempty-valued, and there exists at least one fuzzy L-level solution $x^\star \in F_\eta$.*

*Proof.* We argue by backward induction.

*Level $L$.* Fix $x_{1:L-1}$ such that the feasible slice $\{x_L \in X_L : (x_{1:L-1}, x_L) \in F_\eta\}$ is nonempty. Because $F_\eta$ is closed and $X_L$ is compact, this slice is compact. Continuity of $x_L \mapsto \mathcal{R}(\tilde{f}_L(x_{1:L-1}, x_L))$ implies the minimum is attained (Weierstrass theorem), hence $\mathcal{S}_L(x_{1:L-1}) \neq \varnothing$.

*Induction step.* Assume $\mathcal{S}_{\ell+1}$ is nonempty-valued. Fix $x_{1:\ell-1}$ and consider the set of admissible $x_\ell \in X_\ell$ for which there exists $x_{\ell+1:L} \in \mathcal{S}_{\ell+1}(x_{1:\ell})$ with $(x_{1:L}) \in F_\eta$. Nonemptiness follows from the induction hypothesis together with $F_\eta \neq \varnothing$. Compactness follows from compactness of $X$ and closedness of $F_\eta$ (after projecting onto the relevant coordinates). Since $x \mapsto \mathcal{R}(\tilde{f}_\ell(x))$ is continuous on $X$, the induced minimization over this compact nonempty set attains its minimum, so $\mathcal{S}_\ell(x_{1:\ell-1}) \neq \varnothing$.

Applying this to $\ell = 1$ yields a nonempty $\mathcal{S}_1(\varnothing)$ and thus a backward-induction chain producing at least one $x^\star \in F_\eta$. $\qquad\square$

We now define *Uncertain Multi-Level Decision-Making* (UMLDM) by replacing the level-wise crisp/fuzzy objectives in multi-level decision-making with *uncertain-set*-valued objectives, in the sense of [20, 142].

**Definition 3.11.4** (Uncertain $L$-Level Decision-Making (UMLDM) of type $M$)**.** Let $L \geq 2$ be the number of hierarchical decision levels, and let $M$ be an uncertain model with degree-domain

$$\mathrm{Dom}(M) \subseteq [0,1]^k$$

for some integer $k \geq 1$.

For each level $\ell \in \{1, \ldots, L\}$, let $X_\ell$ be a nonempty decision set for the $\ell$-th decision maker, and put

$$X := \prod_{\ell=1}^{L} X_\ell.$$

An element of $X$ is written as

$$x = (x_1, \ldots, x_L), \qquad x_\ell \in X_\ell.$$

We write

$$x_{i:j} := (x_i, \ldots, x_j) \qquad (1 \leq i \leq j \leq L).$$

Assume that the set of globally feasible decisions is a nonempty subset

$$F_M \subseteq X.$$

For each level $\ell \in \{1, \ldots, L\}$, let

$$f_{\ell,M} : X \longrightarrow \mathrm{Dom}(M)$$

be the *uncertain objective map* of level $\ell$, and let

$$\mathrm{Score}_{\ell,M} : \mathrm{Dom}(M) \longrightarrow \mathbb{R}$$

be a total *ranking / scalarization map* used to compare uncertain objective values at level $\ell$. Thus each pair $(X, f_{\ell,M})$ is an uncertain set (U-set) of type $M$ on the common decision universe $X$.

For each $\ell \in \{1, \dots, L\}$ and each feasible prefix $x_{1:\ell-1}$, define the *feasible continuation set*

$$F_M^{(\ell)}(x_{1:\ell-1}) := \big\{ x_{\ell:L} \in X_\ell \times \cdots \times X_L : (x_{1:\ell-1}, x_{\ell:L}) \in F_M \big\}.$$

For convenience, when $\ell = 1$, we interpret $x_{1:0}$ as the empty prefix $\varnothing$, so that

$$F_M^{(1)}(\varnothing) = F_M.$$

The *rational-response correspondences* are defined recursively from the lowest level upward.

**Level $L$:** for each feasible prefix $x_{1:L-1}$, set

$$\mathcal{R}_{L,M}(x_{1:L-1}) := \arg \min_{x_L \in F_M^{(L)}(x_{1:L-1})} \mathrm{Score}_{L,M}\big(f_{L,M}(x_{1:L-1}, x_L)\big).$$

**Levels $\ell = L-1, L-2, \dots, 1$:** for each feasible prefix $x_{1:\ell-1}$, define the *admissible rational-continuation set*

$$\mathcal{A}_{\ell,M}(x_{1:\ell-1}) := \big\{ x_{\ell:L} \in F_M^{(\ell)}(x_{1:\ell-1}) : x_{\ell+1:L} \in \mathcal{R}_{\ell+1,M}(x_{1:\ell}) \big\},$$

and then define

$$\mathcal{R}_{\ell,M}(x_{1:\ell-1}) = \arg \min_{x_{\ell:L} \in \mathcal{A}_{\ell,M}(x_{1:\ell-1})} \mathrm{Score}_{\ell,M}\big(f_{\ell,M}(x_{1:\ell-1}, x_{\ell:L})\big).$$

An element

$$x^\star = (x_1^\star, \dots, x_L^\star) \in F_M$$

is called an *uncertain $L$-level solution* (or *uncertain Stackelberg solution*) of type $M$ if

$$x^\star \in \mathcal{R}_{1,M}(\varnothing).$$

Equivalently, $x^\star$ is obtained by backward induction, where each level $\ell$ chooses a continuation minimizing its scalarized uncertain objective under the assumption that all lower levels $\ell+1, \dots, L$ respond rationally.

**Remark 3.11.5** (Specializations). If $M$ is the fuzzy model with $\mathrm{Dom}(M) = [0,1]$, then Definition 3.11.4 reduces to a fuzzy multi-level decision-making model. If $M$ is chosen as an intuitionistic fuzzy, neutrosophic, plithogenic, or other uncertainty model, then one obtains the corresponding uncertainty-aware multi-level decision-making framework by changing the degree-domain $\mathrm{Dom}(M)$ and the model-dependent ranking maps $\mathrm{Score}_{\ell,M}$.

**Theorem 3.11.6** (Well-definedness of UMLDM). *Let*

$$\mathfrak{M} = \big( L, \, M, \, (X_\ell)_{\ell=1}^L, \, F_M, \, (f_{\ell,M})_{\ell=1}^L, \, (\mathrm{Score}_{\ell,M})_{\ell=1}^L \big)$$

*be a UMLDM instance as in Definition 3.11.4. Assume:*

1. *each $X_\ell$ is finite and nonempty;*

2. *$F_M \subseteq X = \prod_{\ell=1}^L X_\ell$ is nonempty;*

3. *for each $\ell \in \{1, \ldots, L\}$, the map*
$$f_{\ell,M} : X \to \mathrm{Dom}(M)$$
   *is total;*

4. *for each $\ell \in \{1, \ldots, L\}$, the map*
$$\mathrm{Score}_{\ell,M} : \mathrm{Dom}(M) \to \mathbb{R}$$
   *is total.*

*Then the following hold.*

1. *For every $\ell \in \{1, \ldots, L\}$ and every feasible prefix $x_{1:\ell-1}$ arising from some feasible vector in $F_M$, the feasible continuation set $F_M^{(\ell)}(x_{1:\ell-1})$ is a finite nonempty set.*

2. *For every $\ell \in \{1, \ldots, L\}$ and every such feasible prefix $x_{1:\ell-1}$, the response set $\mathcal{R}_{\ell,M}(x_{1:\ell-1})$ is well-defined and nonempty.*

3. *In particular, $\mathcal{R}_{1,M}(\varnothing) \neq \varnothing$, so the set of uncertain $L$-level solutions is nonempty.*
   *Hence Uncertain Multi-Level Decision-Making of type $M$ is well-defined.*

*Proof.* For each $\ell \in \{1, \ldots, L\}$, define the set of feasible prefixes of length $\ell - 1$ by
$$P_{\ell-1} := \Big\{ x_{1:\ell-1} \in X_1 \times \cdots \times X_{\ell-1} : \exists\, x_{\ell:L} \text{ with } (x_{1:\ell-1}, x_{\ell:L}) \in F_M \Big\}.$$
By convention,
$$P_0 = \{\varnothing\}.$$
Since $F_M \neq \varnothing$, we have $P_0 \neq \varnothing$.

We prove by backward induction on $\ell$ that for every $x_{1:\ell-1} \in P_{\ell-1}$,
$$F_M^{(\ell)}(x_{1:\ell-1})$$
is finite and nonempty, and
$$\mathcal{R}_{\ell,M}(x_{1:\ell-1})$$
is well-defined and nonempty.

**Step 1: the last level $\ell = L$.**

Fix $x_{1:L-1} \in P_{L-1}$. By definition of $P_{L-1}$, there exists at least one $x_L \in X_L$ such that
$$(x_{1:L-1}, x_L) \in F_M.$$
Hence
$$F_M^{(L)}(x_{1:L-1}) = \{ x_L \in X_L : (x_{1:L-1}, x_L) \in F_M \}$$
is nonempty. Since $X_L$ is finite, $F_M^{(L)}(x_{1:L-1})$ is also finite.

Because $f_{L,M}$ is total, for each $x_L \in F_M^{(L)}(x_{1:L-1})$,

$$f_{L,M}(x_{1:L-1}, x_L) \in \mathrm{Dom}(M)$$

is defined. Because $\mathrm{Score}_{L,M}$ is total,

$$\mathrm{Score}_{L,M}\big(f_{L,M}(x_{1:L-1}, x_L)\big) \in \mathbb{R}$$

is defined for every $x_L \in F_M^{(L)}(x_{1:L-1})$.

Therefore we are minimizing a real-valued function over a finite nonempty set. Hence the minimum is attained, so

$$\mathcal{R}_{L,M}(x_{1:L-1}) = \arg \min_{x_L \in F_M^{(L)}(x_{1:L-1})} \mathrm{Score}_{L,M}\big(f_{L,M}(x_{1:L-1}, x_L)\big)$$

is well-defined and nonempty.

**Step 2: induction step.**

Assume that for some $\ell \in \{1, \dots, L-1\}$, the statement has already been proved for level $\ell + 1$; namely, for every feasible prefix $x_{1:\ell} \in P_\ell$,

$$F_M^{(\ell+1)}(x_{1:\ell})$$

is finite and nonempty, and

$$\mathcal{R}_{\ell+1,M}(x_{1:\ell})$$

is well-defined and nonempty.

Now fix $x_{1:\ell-1} \in P_{\ell-1}$. By definition of $P_{\ell-1}$, there exists some

$$\bar{x}_{\ell:L} \in X_\ell \times \cdots \times X_L$$

such that

$$(x_{1:\ell-1}, \bar{x}_{\ell:L}) \in F_M.$$

Hence

$$F_M^{(\ell)}(x_{1:\ell-1})$$

is nonempty; since $X_\ell \times \cdots \times X_L$ is finite, it is also finite.

Let $\bar{x}_\ell$ denote the $\ell$-th component of $\bar{x}_{\ell:L}$. Then

$$(x_{1:\ell-1}, \bar{x}_\ell, \bar{x}_{\ell+1:L}) \in F_M,$$

so the prefix

$$(x_{1:\ell-1}, \bar{x}_\ell) \in P_\ell.$$

By the induction hypothesis,

$$\mathcal{R}_{\ell+1,M}(x_{1:\ell-1}, \bar{x}_\ell) \neq \varnothing.$$

Choose any

$$y_{\ell+1:L} \in \mathcal{R}_{\ell+1,M}(x_{1:\ell-1}, \bar{x}_\ell).$$

Again by the induction hypothesis,

$$y_{\ell+1:L} \in F_M^{(\ell+1)}(x_{1:\ell-1}, \bar{x}_\ell),$$

which means

$$(x_{1:\ell-1}, \bar{x}_\ell, y_{\ell+1:L}) \in F_M.$$

Therefore

$$(\bar{x}_\ell, y_{\ell+1:L}) \in \mathcal{A}_{\ell,M}(x_{1:\ell-1}).$$

So the admissible rational-continuation set

$$\mathcal{A}_{\ell,M}(x_{1:\ell-1})$$

is nonempty. Since it is a subset of the finite set $F_M^{(\ell)}(x_{1:\ell-1})$, it is finite.

Now, for every

$$x_{\ell:L} \in \mathcal{A}_{\ell,M}(x_{1:\ell-1}),$$

the point $(x_{1:\ell-1}, x_{\ell:L}) \in X$ is defined, hence by totality of $f_{\ell,M}$,

$$f_{\ell,M}(x_{1:\ell-1}, x_{\ell:L}) \in \mathrm{Dom}(M)$$

is defined; and by totality of $\mathrm{Score}_{\ell,M}$,

$$\mathrm{Score}_{\ell,M}\big(f_{\ell,M}(x_{1:\ell-1}, x_{\ell:L})\big) \in \mathbb{R}$$

is defined.

Thus $\mathcal{R}_{\ell,M}(x_{1:\ell-1})$ is the argmin of a real-valued function over a finite nonempty set:

$$\mathcal{R}_{\ell,M}(x_{1:\ell-1}) = \arg\min_{x_{\ell:L} \in \mathcal{A}_{\ell,M}(x_{1:\ell-1})} \mathrm{Score}_{\ell,M}\big(f_{\ell,M}(x_{1:\ell-1}, x_{\ell:L})\big).$$

Hence $\mathcal{R}_{\ell,M}(x_{1:\ell-1})$ is well-defined and nonempty.

This completes the backward induction.

Applying the result to $\ell = 1$, we obtain that

$$\mathcal{R}_{1,M}(\varnothing) \neq \varnothing.$$

Every element of $\mathcal{R}_{1,M}(\varnothing)$ is an uncertain $L$-level solution. Therefore the solution set is nonempty, and UMLDM is well-defined. $\qquad\square$

Table 3.11 presents related uncertainty-model variants of Fuzzy Multi-Level Decision-Making.

Table 3.11: Related uncertainty-model variants of Fuzzy Multi-Level Decision-Making.

| $k$ | Related Fuzzy Multi-Level Decision-Making variant(s) |
|---|---|
| 1 | Fuzzy Multi-Level Decision-Making |
| 2 | Intuitionistic Fuzzy Multi-Level Decision-Making |
| 3 | Hesitant Fuzzy Multi-Level Decision-Making |
| 3 | Spherical Fuzzy Multi-Level Decision-Making |
| 3 | Neutrosophic Multi-Level Decision-Making |
| $n$ | Plithogenic Multi-Level Decision-Making |

## 3.12 Fuzzy Multi-Agent Decision-Making

Multi-agent decision-making coordinates or analyzes multiple autonomous agents with possibly conflicting utilities, seeking joint policies, equilibria, or negotiated agreements under interaction rules [410–412]. Fuzzy multi-agent decision-making represents agents' utilities, beliefs, or strategies by fuzzy sets, enabling negotiation and coordination under vague preferences and uncertain perceptions.

**Definition 3.12.1** (Fuzzy-number domain). Let $\mathbb{F}$ denote the class of (real) fuzzy numbers, i.e., fuzzy sets $\tilde{a}$ on $\mathbb{R}$ with membership function $\mu_{\tilde{a}} : \mathbb{R} \to [0,1]$ that are (i) normal, (ii) convex, (iii) upper semicontinuous, and (iv) have compact support. For $\alpha \in (0,1]$, the $\alpha$-cut is the (nonempty) compact interval

$$\tilde{a}^{\alpha} := \{x \in \mathbb{R} : \mu_{\tilde{a}}(x) \geq \alpha\} = [\underline{a}(\alpha), \overline{a}(\alpha)].$$

Define the (Zadeh-extension) sum $\tilde{a} \oplus \tilde{b} \in \mathbb{F}$ and positive scalar product $\gamma \odot \tilde{a} \in \mathbb{F}$ ($\gamma \geq 0$) by $\alpha$-cuts:

$$(\tilde{a} \oplus \tilde{b})^{\alpha} = \tilde{a}^{\alpha} + \tilde{b}^{\alpha} = [\underline{a}(\alpha) + \underline{b}(\alpha), \overline{a}(\alpha) + \overline{b}(\alpha)], \qquad (\gamma \odot \tilde{a})^{\alpha} = \gamma \, \tilde{a}^{\alpha}.$$

**Definition 3.12.2** (Fuzzy multi-agent decision-making). Let

$$\mathcal{A} = \{A_1, \ldots, A_m\} \quad \text{(alternatives)},$$

$$\mathcal{C} = \{C_1, \ldots, C_n\} \quad \text{(criteria)},$$

$$\mathcal{D} = \{1, \ldots, K\} \quad \text{(agents)}.$$

A *Fuzzy Multi-Agent Decision-Making* (FMADM) instance is a tuple

$$\mathsf{FMADM} = \Big(\mathcal{A}, \mathcal{C}, \mathcal{D}, (\tilde{X}^{(k)})_{k \in \mathcal{D}}, (w^{(k)})_{k \in \mathcal{D}}, \lambda, \mathrm{Agg}, \mathrm{Score}, \Pi\Big),$$

where:

1. For each agent $k \in \mathcal{D}$,

$$\tilde{X}^{(k)} = (\tilde{x}_{ij}^{(k)})_{m \times n} \in \mathbb{F}^{m \times n}$$

   is the fuzzy evaluation matrix, with $\tilde{x}_{ij}^{(k)} \in \mathbb{F}$ describing the (perceived) performance of $A_i$ on $C_j$ by agent $k$.

2. For each agent $k \in \mathcal{D}$,

$$w^{(k)} = (w_1^{(k)}, \ldots, w_n^{(k)}) \in [0,1]^n, \qquad \sum_{j=1}^{n} w_j^{(k)} = 1,$$

   is the criterion-weight vector of agent $k$ (crisp weights; fuzzy weights can be handled analogously by replacing $w_j^{(k)}$ with $\tilde{w}_j^{(k)} \in \mathbb{F}$).

3. $\lambda = (\lambda_1, \ldots, \lambda_K) \in [0,1]^K$ is the agent-importance vector with $\sum_{k=1}^{K} \lambda_k = 1$.

4. $\mathrm{Agg} : \mathbb{F}^K \to \mathbb{F}$ is an agent-aggregation operator. A canonical choice is the weighted fuzzy mean

$$\mathrm{Agg}(\tilde{a}_1, \ldots, \tilde{a}_K) := \bigoplus_{k=1}^{K} \lambda_k \odot \tilde{a}_k.$$

5. $\mathrm{Score} : \mathbb{F} \to \mathbb{R}$ is a scoring (defuzzification) functional. A standard example is the centroid score

$$\mathrm{Score}(\tilde{a}) = \frac{\int_{\mathbb{R}} x \, \mu_{\tilde{a}}(x) \, dx}{\int_{\mathbb{R}} \mu_{\tilde{a}}(x) \, dx}, \quad \text{assuming} \ \int_{\mathbb{R}} \mu_{\tilde{a}}(x) \, dx > 0.$$

6. $\Pi$ is a decision rule producing either a ranking $\preceq$ on $\mathcal{A}$ or a choice set $\mathcal{S} \subseteq \mathcal{A}$.

The *collective* fuzzy evaluation matrix is defined entrywise by

$$\tilde{x}_{ij} := \mathrm{Agg}(\tilde{x}_{ij}^{(1)}, \ldots, \tilde{x}_{ij}^{(K)}) \in \mathbb{F}, \qquad \tilde{X} := (\tilde{x}_{ij})_{m \times n} \in \mathbb{F}^{m \times n}.$$

A common induced (collective) fuzzy utility of $A_i$ is

$$\tilde{u}_i := \bigoplus_{j=1}^{n} w_j \odot \tilde{x}_{ij} \in \mathbb{F}, \qquad \text{with some } w \in [0,1]^n, \ \sum_{j=1}^{n} w_j = 1 \ \left(\text{e.g., } w = \sum_{k=1}^{K} \lambda_k \, w^{(k)}\right).$$

Then $\Pi$ may return the score-based preorder

$$A_i \preceq A_\ell \iff \mathrm{Score}(\tilde{u}_i) \leq \mathrm{Score}(\tilde{u}_\ell),$$

or the maximal (choice) set $\mathcal{S} = \{A_i \in \mathcal{A} : \ \mathrm{Score}(\tilde{u}_i) = \max_r \mathrm{Score}(\tilde{u}_r)\}$.

**Theorem 3.12.3** (Well-definedness of the canonical FMADM aggregation). *Assume $\mathbb{F}$ is as in Definition 3.12.1. Let $\mathrm{Agg}(\tilde{a}_1, \ldots, \tilde{a}_K) = \bigoplus_{k=1}^{K} \lambda_k \odot \tilde{a}_k$ with $\lambda_k \geq 0$ and $\sum_{k=1}^{K} \lambda_k = 1$. Then, for every $(\tilde{a}_1, \ldots, \tilde{a}_K) \in \mathbb{F}^K$, one has $\mathrm{Agg}(\tilde{a}_1, \ldots, \tilde{a}_K) \in \mathbb{F}$. Consequently, the collective matrix $\tilde{X}$ in Definition 3.12.2 is well-defined in $\mathbb{F}^{m \times n}$.*

*Proof.* Fix $\alpha \in (0,1]$. Each $\tilde{a}_k^\alpha$ is a compact interval in $\mathbb{R}$. By the $\alpha$-cut definitions,

$$\left(\bigoplus_{k=1}^{K} \lambda_k \odot \tilde{a}_k\right)^\alpha = \sum_{k=1}^{K} \lambda_k \, \tilde{a}_k^\alpha,$$

which is a (nonempty) compact interval since Minkowski sums and nonnegative scalar multiples preserve compact intervals. Standard closure properties for fuzzy numbers imply that the resulting fuzzy set is again normal, convex, upper semicontinuous, and has compact support; hence it lies in $\mathbb{F}$. Applying this entrywise yields $\tilde{X} \in \mathbb{F}^{m \times n}$. $\qquad \square$

We now define *Uncertain Multi-Agent Decision-Making* (UMADM) by extending fuzzy multi-agent decision-making from fuzzy-number-valued evaluations to a general uncertain model $M$ with degree-domain $\mathrm{Dom}(M) \subseteq [0,1]^k$.

**Definition 3.12.4** (Uncertain Multi-Agent Decision-Making (UMADM) of type $M$)**.** Let

$$\mathcal{A} = \{A_1, \ldots, A_m\} \quad \text{(alternatives)}, \qquad \mathcal{C} = \{C_1, \ldots, C_n\} \quad \text{(criteria)}, \qquad \mathcal{D} = \{1, \ldots, K\} \quad \text{(agents)},$$

where $m, n, K \in \mathbb{N}$ and $m, n, K \geq 1$.

Fix an uncertain model $M$ with degree-domain

$$\mathrm{Dom}(M) \subseteq [0, 1]^k$$

for some integer $k \geq 1$.

An *Uncertain Multi-Agent Decision-Making instance of type $M$* is a tuple

$$\mathsf{UMADM}_M = \Big( \mathcal{A}, \mathcal{C}, \mathcal{D}, (X_M^{(k)})_{k \in \mathcal{D}}, (w_M^{(k)})_{k \in \mathcal{D}}, \lambda, \mathrm{Agg}_M, \mathrm{Util}_M, \mathrm{Score}_M \Big),$$

where:

1. For each agent $k \in \mathcal{D}$,
$$X_M^{(k)} = (\mu_{ij}^{(k)})_{m \times n} \in \mathrm{Dom}(M)^{m \times n}$$
   is the *uncertain evaluation matrix* of agent $k$, where $\mu_{ij}^{(k)} \in \mathrm{Dom}(M)$ denotes the uncertain assessment of alternative $A_i$ under criterion $C_j$ given by agent $k$.

2. For each agent $k \in \mathcal{D}$,
$$w_M^{(k)} = (\omega_1^{(k)}, \ldots, \omega_n^{(k)}) \in \mathrm{Dom}(M)^n$$
   is the *uncertain criterion-importance profile* of agent $k$. (If desired, one may instead use crisp weights $w^{(k)} \in [0, 1]^n$, but here we keep the general uncertain form.)

3. 
$$\lambda = (\lambda_1, \ldots, \lambda_K) \in [0, 1]^K, \qquad \sum_{k=1}^{K} \lambda_k = 1,$$
   is the *agent-importance vector*.

4. 
$$\mathrm{Agg}_M : \mathrm{Dom}(M)^K \longrightarrow \mathrm{Dom}(M)$$
   is a *model-dependent agent-aggregation operator*, used to combine the $K$ agent-specific assessments of the same entry into a collective uncertain assessment.

5. 
$$\mathrm{Util}_M : \mathrm{Dom}(M)^n \times \mathrm{Dom}(M)^n \longrightarrow \mathrm{Dom}(M)$$
   is a *model-dependent utility aggregation operator*, used to combine the collective criterion-wise assessments of an alternative with a collective criterion-weight profile.

6. 
$$\mathrm{Score}_M : \mathrm{Dom}(M) \longrightarrow \mathbb{R}$$
   is a *ranking functional* (score / scalarization map).

The *collective uncertain evaluation matrix* is defined entrywise by

$$\mu_{ij} := \mathrm{Agg}_M\big(\mu_{ij}^{(1)}, \ldots, \mu_{ij}^{(K)}\big) \in \mathrm{Dom}(M), \qquad X_M := (\mu_{ij})_{m \times n} \in \mathrm{Dom}(M)^{m \times n}.$$

Similarly, the *collective uncertain criterion-weight profile* is defined componentwise by

$$\omega_j := \mathrm{Agg}_M\big(\omega_j^{(1)}, \ldots, \omega_j^{(K)}\big) \in \mathrm{Dom}(M), \qquad w_M := (\omega_1, \ldots, \omega_n) \in \mathrm{Dom}(M)^n.$$

For each alternative $A_i \in \mathcal{A}$, define its *collective uncertain utility* by

$$u_i := \mathrm{Util}_M\big((\mu_{i1}, \ldots, \mu_{in}), (\omega_1, \ldots, \omega_n)\big) \in \mathrm{Dom}(M).$$

The induced score of $A_i$ is

$$s_i := \mathrm{Score}_M(u_i) \in \mathbb{R}.$$

The *collective preference relation* $\succeq_M$ on $\mathcal{A}$ is defined by

$$A_i \succeq_M A_j \quad \Longleftrightarrow \quad s_i \geq s_j.$$

A *solution* of $\mathsf{UMADM}_M$ is any alternative

$$A^\star \in \arg\max_{A_i \in \mathcal{A}} s_i.$$

The corresponding set of optimal alternatives is denoted by

$$\mathcal{A}_M^\star := \arg\max_{A_i \in \mathcal{A}} s_i.$$

**Remark 3.12.5** (Interpretation). Definition 3.12.4 separates the multi-agent decision process into three layers:

1. agent-wise uncertain assessments $X_M^{(k)}$ and weight profiles $w_M^{(k)}$;

2. inter-agent aggregation via $\mathrm{Agg}_M$;

3. alternative-wise utility construction via $\mathrm{Util}_M$, followed by ranking through $\mathrm{Score}_M$.

Thus UMADM is a model-independent framework: once $M$, $\mathrm{Agg}_M$, $\mathrm{Util}_M$, and $\mathrm{Score}_M$ are specified, one obtains a concrete uncertainty-aware multi-agent decision method.

**Remark 3.12.6** (Specializations). If $M$ is the fuzzy model with $\mathrm{Dom}(M) = [0, 1]$, then Definition 3.12.4 reduces to a fuzzy multi-agent decision-making framework. If $M$ is chosen as an intuitionistic fuzzy, neutrosophic, hesitant fuzzy, spherical fuzzy, plithogenic, or other uncertainty model, then one obtains the corresponding uncertainty-aware multi-agent decision-making setting by replacing $\mathrm{Dom}(M)$, $\mathrm{Agg}_M$, $\mathrm{Util}_M$, and $\mathrm{Score}_M$ accordingly.

**Theorem 3.12.7** (Well-definedness of UMADM of type $M$)**.** *Let*

$$\mathsf{UMADM}_M = \left( \mathcal{A}, \mathcal{C}, \mathcal{D}, (X_M^{(k)})_{k \in \mathcal{D}}, (w_M^{(k)})_{k \in \mathcal{D}}, \lambda, \mathrm{Agg}_M, \mathrm{Util}_M, \mathrm{Score}_M \right)$$

*be a UMADM instance as in Definition 3.12.4. Assume:*

*(A1) $\mathcal{A}$, $\mathcal{C}$, and $\mathcal{D}$ are finite nonempty sets;*

*(A2) for each $k \in \mathcal{D}$,*

$$X_M^{(k)} = (\mu_{ij}^{(k)}) \in \mathrm{Dom}(M)^{m \times n} \qquad and \qquad w_M^{(k)} = (\omega_1^{(k)}, \ldots, \omega_n^{(k)}) \in \mathrm{Dom}(M)^n;$$

*(A3)*

$$\mathrm{Agg}_M : \mathrm{Dom}(M)^K \to \mathrm{Dom}(M)$$

*is a total map;*

*(A4)*

$$\mathrm{Util}_M : \mathrm{Dom}(M)^n \times \mathrm{Dom}(M)^n \to \mathrm{Dom}(M)$$

*is a total map;*

*(A5)*

$$\mathrm{Score}_M : \mathrm{Dom}(M) \to \mathbb{R}$$

*is a total map.*

*Then the following objects are well-defined:*

*(i) the collective uncertain evaluation matrix*

$$X_M = (\mu_{ij})_{m \times n} \in \mathrm{Dom}(M)^{m \times n};$$

*(ii) the collective uncertain criterion-weight profile*

$$w_M = (\omega_1, \ldots, \omega_n) \in \mathrm{Dom}(M)^n;$$

*(iii) the collective uncertain utilities*

$$u_i \in \mathrm{Dom}(M) \qquad (i = 1, \ldots, m);$$

*(iv) the real-valued scores*

$$s_i = \mathrm{Score}_M(u_i) \in \mathbb{R} \qquad (i = 1, \ldots, m);$$

*(v) the preference relation $\succeq_M$ on $\mathcal{A}$, defined by*

$$A_i \succeq_M A_j \iff s_i \geq s_j,$$

*which is a total preorder;*

*(vi) the optimal set*

$$\mathcal{A}_M^\star = \arg\max_{A_i \in \mathcal{A}} s_i,$$

*which is nonempty.*

*Hence Uncertain Multi-Agent Decision-Making of type M is well-defined.*

*Proof.* By (A1), the sets $\mathcal{A} = \{A_1, \ldots, A_m\}$, $\mathcal{C} = \{C_1, \ldots, C_n\}$, and $\mathcal{D} = \{1, \ldots, K\}$ are finite and nonempty.

**Step 1: Well-definedness of the collective matrix $X_M$.**

Fix $i \in \{1, \ldots, m\}$ and $j \in \{1, \ldots, n\}$. By (A2), for every $k \in \mathcal{D}$,

$$\mu_{ij}^{(k)} \in \mathrm{Dom}(M).$$

Hence the $K$-tuple

$$(\mu_{ij}^{(1)}, \ldots, \mu_{ij}^{(K)}) \in \mathrm{Dom}(M)^K.$$

Since $\mathrm{Agg}_M$ is total by (A3), the value

$$\mu_{ij} := \mathrm{Agg}_M(\mu_{ij}^{(1)}, \ldots, \mu_{ij}^{(K)})$$

is defined and belongs to $\mathrm{Dom}(M)$. Because this holds for every pair $(i, j)$, the matrix

$$X_M = (\mu_{ij})_{m \times n}$$

belongs to $\mathrm{Dom}(M)^{m \times n}$. Thus (i) is proved.

**Step 2: Well-definedness of the collective weight profile $w_M$.**

Fix $j \in \{1, \ldots, n\}$. By (A2), for every $k \in \mathcal{D}$,

$$\omega_j^{(k)} \in \mathrm{Dom}(M).$$

Hence

$$(\omega_j^{(1)}, \ldots, \omega_j^{(K)}) \in \mathrm{Dom}(M)^K.$$

Again by totality of $\mathrm{Agg}_M$,

$$\omega_j := \mathrm{Agg}_M(\omega_j^{(1)}, \ldots, \omega_j^{(K)})$$

is defined and belongs to $\mathrm{Dom}(M)$. Therefore

$$w_M = (\omega_1, \ldots, \omega_n) \in \mathrm{Dom}(M)^n.$$

So (ii) holds.

**Step 3: Well-definedness of the collective uncertain utilities $u_i$.**

Fix $i \in \{1, \ldots, m\}$. From Step 1,

$$(\mu_{i1}, \ldots, \mu_{in}) \in \mathrm{Dom}(M)^n,$$

and from Step 2,

$$(\omega_1, \ldots, \omega_n) \in \mathrm{Dom}(M)^n.$$

Hence

$$\big((\mu_{i1}, \ldots, \mu_{in}), (\omega_1, \ldots, \omega_n)\big) \in \mathrm{Dom}(M)^n \times \mathrm{Dom}(M)^n.$$

By totality of $\mathrm{Util}_M$ in (A4), the value

$$u_i = \mathrm{Util}_M\big((\mu_{i1}, \ldots, \mu_{in}), (\omega_1, \ldots, \omega_n)\big)$$

is defined and belongs to $\mathrm{Dom}(M)$. Thus (iii) holds.

**Step 4: Well-definedness of the scores $s_i$.**

By Step 3, $u_i \in \mathrm{Dom}(M)$ for each $i$. Since $\mathrm{Score}_M$ is total by (A5),

$$s_i := \mathrm{Score}_M(u_i) \in \mathbb{R}$$

is well-defined for every $i = 1, \ldots, m$. Hence (iv) holds.

**Step 5: The induced preference relation is a total preorder.**

Define

$$A_i \succeq_M A_j \iff s_i \geq s_j.$$

Since $\geq$ on $\mathbb{R}$ is reflexive, transitive, and total, the induced relation $\succeq_M$ on $\mathcal{A}$ is also reflexive, transitive, and total. Therefore $\succeq_M$ is a total preorder on $\mathcal{A}$. This proves (v).

**Step 6: Existence of an optimal alternative.**

The set of scores

$$\{s_1, \ldots, s_m\} \subset \mathbb{R}$$

is finite and nonempty because $\mathcal{A}$ is finite and nonempty. Every finite nonempty subset of $\mathbb{R}$ has a maximum. Hence there exists at least one index $i^\star \in \{1, \ldots, m\}$ such that

$$s_{i^\star} = \max_{1 \leq i \leq m} s_i.$$

Therefore

$$\mathcal{A}_M^\star = \arg\max_{A_i \in \mathcal{A}} s_i \neq \varnothing.$$

So (vi) holds.

All required objects are thus well-defined, and the optimal set is nonempty. Hence UMADM of type $M$ is well-defined. $\qquad\square$

Table 3.12 presents related uncertainty-model variants of Fuzzy Multi-Agent Decision-Making.

Table 3.12: Related uncertainty-model variants of Fuzzy Multi-Agent Decision-Making.

| $k$ | **Related Fuzzy Multi-Agent Decision-Making variant(s)** |
| --- | --- |
| 1 | Fuzzy Multi-Agent Decision-Making |
| 2 | Intuitionistic Fuzzy Multi-Agent Decision-Making |
| 3 | Hesitant Fuzzy Multi-Agent Decision-Making |
| 3 | Spherical Fuzzy Multi-Agent Decision-Making |
| 3 | Neutrosophic Multi-Agent Decision-Making |
| $n$ | Plithogenic Multi-Agent Decision-Making |

## 3.13 Fuzzy Multi-Scenario Decision-Making

Multi-scenario decision-making evaluates alternatives under several scenarios, aggregates scenario outcomes using probabilities or weights, and selects robust or expected-optimal alternatives [413, 414]. Fuzzy multi-scenario decision-making treats scenario performances or scenario weights as fuzzy, aggregating with fuzzy expectation, worst-case, or regret criteria to select robustly.

**Definition 3.13.1** (Fuzzy multi-scenario decision model). A *fuzzy multi-scenario decision-making problem* is a tuple

$$\mathfrak{D} = \big(X,\ K,\ (K_j)_{j=1}^{m},\ (Y_{jt})_{j,t},\ \Sigma,\ \mu,\ w,\ \pi,\ \mathrm{Agg}_C,\ \mathrm{Agg}_S,\ \preceq\big),$$

where:

1. $X = \{x_1, \ldots, x_n\}$ is the finite set of alternatives.

2. $K = \{K_1, \ldots, K_m\}$ is the finite set of (top-level) criteria.

3. For each $j \in \{1, \ldots, m\}$, $K_j = \{k_{j1}, \ldots, k_{jT_j}\}$ is the finite set of indicators (subcriteria) defining criterion $K_j$.

4. For each indicator $k_{jt}$, $Y_{jt}$ is its value domain (numerical, ordinal, linguistic, etc.).

5. $\Sigma = \{\sigma_1, \ldots, \sigma_s\}$ is a finite set of *scenarios* (distinct requirement profiles / evaluation policies).

6. $\mu$ is a family of scenario-dependent membership mappings (defined precisely in Definition 3.13.2).

7. $w = (w_1, \ldots, w_m) \in [0,1]^m$ is a criterion-weight vector with $\sum_{j=1}^{m} w_j = 1$.

8. $\pi = (\pi_1, \ldots, \pi_s) \in [0,1]^s$ is a scenario-weight vector with $\sum_{r=1}^{s} \pi_r = 1$.

9. $\mathrm{Agg}_C$ is a criterion-aggregation operator (within a scenario).

10. $\mathrm{Agg}_S$ is a scenario-aggregation operator (across scenarios).

11. $\preceq$ is a total preorder on the chosen score space (e.g., on $[0,1]$ or on a class of fuzzy numbers) used for ranking.

**Definition 3.13.2** (Scenario-dependent membership structure). Fix $x \in X$, $j \in \{1, \ldots, m\}$, and a scenario $\sigma \in \Sigma$. Let the indicator observations for $x$ be

$$y_{jt}(x) \in Y_{jt} \qquad (t = 1, \ldots, T_j).$$

A *scenario-dependent indicator membership* is a mapping

$$\mu_{jt}^{\sigma} : Y_{jt} \longrightarrow [0,1], \qquad y \longmapsto \mu_{jt}^{\sigma}(y),$$

and the induced *scenario-dependent criterion satisfaction* is defined by an indicator-aggregator $\Phi_j^\sigma$ as

$$\mu_j^\sigma(x) := \Phi_j^\sigma\Big(\mu_{j1}^\sigma(y_{j1}(x)), \ldots, \mu_{jT_j}^\sigma(y_{jT_j}(x))\Big) \in [0,1].$$

Typical choices include:

$$\Phi_j^\sigma = \min \quad \text{(conjunctive / "all indicators must be good"),}$$

$$\Phi_j^\sigma = \max \quad \text{(disjunctive / "any indicator may suffice"),}$$

or a weighted mean on $[0,1]$, or a t-norm/t-conorm composition. Scenario semantics (e.g., *obligatory*, *desirable*, *not required*) are encoded by how $\mu_{jt}^\sigma$ and $\Phi_j^\sigma$ are chosen.

**Definition 3.13.3** (Within-scenario overall score). Assume the score space is $[0,1]$. For each scenario $\sigma \in \Sigma$, define the *within-scenario overall score*

$$U^\sigma(x) := \mathrm{Agg}_C\big((\mu_j^\sigma(x))_{j=1}^m;\ w\big) \in [0,1].$$

A canonical (compensatory) choice is the convex combination

$$U^\sigma(x) = \sum_{j=1}^m w_j\, \mu_j^\sigma(x).$$

A canonical non-compensatory alternative is the weighted minimum

$$U^\sigma(x) = \min_{1 \le j \le m} \left(\mu_j^\sigma(x)\right)^{\lambda_j}, \qquad \lambda_j > 0,$$

or any monotone aggregation on $[0,1]^m$ compatible with the intended decision logic.

**Definition 3.13.4** (Across-scenario aggregation and ranking). Define the *multi-scenario overall score* of $x \in X$ by

$$U(x) := \mathrm{Agg}_S\big((U^{\sigma_r}(x))_{r=1}^s;\ \pi\big).$$

A standard choice is again the convex combination

$$U(x) = \sum_{r=1}^s \pi_r\, U^{\sigma_r}(x) \in [0,1].$$

The induced ranking is

$$x_a \preceq x_b \quad \Longleftrightarrow \quad U(x_a) \le U(x_b),$$

and the set of optimal alternatives is

$$\arg\max_{x \in X} U(x) = \{x \in X :\ U(x) \ge U(z)\ \forall z \in X\}.$$

**Proposition 3.13.5** (Well-definedness). *Assume:*

1. *For every $\sigma \in \Sigma$, $j$, and $t$, $\mu_{jt}^\sigma : Y_{jt} \to [0,1]$ is well-defined.*

2. *For every $\sigma$ and $j$, $\Phi_j^\sigma : [0,1]^{T_j} \to [0,1]$ is well-defined.*

3. *$w \in [0,1]^m$ and $\sum_{j=1}^m w_j = 1$, and $\pi \in [0,1]^s$ and $\sum_{r=1}^s \pi_r = 1$.*

4. *$\mathrm{Agg}_C : [0,1]^m \times \Delta_m \to [0,1]$ and $\mathrm{Agg}_S : [0,1]^s \times \Delta_s \to [0,1]$ (where $\Delta_m, \Delta_s$ are simplices of weights) are aggregation operators that return values in $[0,1]$.*

*Then for every $x \in X$, the quantities $\mu_j^\sigma(x)$, $U^\sigma(x)$, and $U(x)$ are well-defined and belong to $[0,1]$. Hence, Definition 3.13.4 yields a well-defined ranking.*

*Proof.* Fix $x \in X$. By (1), each $\mu_{jt}^\sigma(y_{jt}(x)) \in [0,1]$ is defined. By (2), $\mu_j^\sigma(x) = \Phi_j^\sigma(\cdots) \in [0,1]$ is defined for each $j$. Thus $(\mu_j^\sigma(x))_{j=1}^m \in [0,1]^m$, and by (4) we obtain $U^\sigma(x) = \mathrm{Agg}_C((\mu_j^\sigma(x))_j; w) \in [0,1]$. Likewise, $(U^{\sigma_r}(x))_{r=1}^s \in [0,1]^s$, and by (4) again $U(x) = \mathrm{Agg}_S((U^{\sigma_r}(x))_r; \pi) \in [0,1]$. Therefore $U : X \to [0,1]$ is well-defined, and the preorder induced by $\leq$ on $[0,1]$ defines a well-defined ranking on $X$. $\qquad\square$

We now define *Uncertain Multi-Scenario Decision-Making* (UMScDM) by extending fuzzy multi-scenario decision-making from fuzzy assessments to a general uncertain model $M$ with degree-domain $\mathrm{Dom}(M) \subseteq [0,1]^k$.

**Definition 3.13.6** (Uncertain Multi-Scenario Decision-Making (UMScDM) of type $M$)**.** Let

$$\mathcal{A} = \{A_1, \ldots, A_m\} \quad \text{(alternatives)}, \qquad \mathcal{C} = \{C_1, \ldots, C_n\} \quad \text{(criteria)}, \qquad \Sigma = \{\sigma_1, \ldots, \sigma_s\} \quad \text{(scenarios)},$$

where $m, n, s \in \mathbb{N}$ and $m, n, s \geq 1$.

Fix an uncertain model $M$ with degree-domain

$$\mathrm{Dom}(M) \subseteq [0,1]^k$$

for some integer $k \geq 1$.

An *Uncertain Multi-Scenario Decision-Making instance of type $M$* is a tuple

$$\mathsf{UMScDM}_M = \Big(\mathcal{A}, \mathcal{C}, \Sigma, (X_M^{(r)})_{r=1}^s, w_M, \pi, \mathrm{Agg}_{C,M}, \mathrm{Agg}_{S,M}, \mathrm{Score}_M\Big),$$

where:

1. For each scenario $r \in \{1, \ldots, s\}$,

$$X_M^{(r)} = (\mu_{ij}^{(r)})_{m \times n} \in \mathrm{Dom}(M)^{m \times n}$$

   is the *scenario-dependent uncertain evaluation matrix*, where $\mu_{ij}^{(r)} \in \mathrm{Dom}(M)$ denotes the uncertain assessment of alternative $A_i$ under criterion $C_j$ in scenario $\sigma_r$.

2. 

$$w_M = (\omega_1, \ldots, \omega_n) \in \mathrm{Dom}(M)^n$$

   is the *uncertain criterion-importance profile*.

3. 

$$\pi = (\pi_1, \ldots, \pi_s) \in [0,1]^s, \qquad \sum_{r=1}^s \pi_r = 1,$$

   is the *scenario-weight vector*.

4.
$$\mathrm{Agg}_{C,M} : \mathrm{Dom}(M)^n \times \mathrm{Dom}(M)^n \longrightarrow \mathrm{Dom}(M)$$

is a *within-scenario criterion-aggregation operator*. For each scenario $r$ and each alternative $A_i$, it produces an uncertain scenario-specific utility

$$u_i^{(r)} := \mathrm{Agg}_{C,M}\big((\mu_{i1}^{(r)}, \ldots, \mu_{in}^{(r)}), (\omega_1, \ldots, \omega_n)\big) \in \mathrm{Dom}(M).$$

5.
$$\mathrm{Agg}_{S,M} : \mathrm{Dom}(M)^s \times [0,1]^s \longrightarrow \mathrm{Dom}(M)$$

is an *across-scenario aggregation operator*. It combines the scenario-specific utilities of an alternative into its overall uncertain utility

$$u_i := \mathrm{Agg}_{S,M}\big((u_i^{(1)}, \ldots, u_i^{(s)}), \pi\big) \in \mathrm{Dom}(M).$$

6.
$$\mathrm{Score}_M : \mathrm{Dom}(M) \longrightarrow \mathbb{R}$$

is a *ranking functional* (score / scalarization map), and the corresponding real-valued score of $A_i$ is

$$s_i := \mathrm{Score}_M(u_i) \in \mathbb{R}.$$

The induced preference relation $\succeq_M$ on $\mathcal{A}$ is defined by

$$A_i \succeq_M A_j \quad \Longleftrightarrow \quad s_i \geq s_j.$$

A *solution* of $\mathsf{UMScDM}_M$ is any alternative

$$A^\star \in \arg\max_{A_i \in \mathcal{A}} s_i,$$

and the set of all optimal alternatives is denoted by

$$\mathcal{A}_M^\star := \arg\max_{A_i \in \mathcal{A}} s_i.$$

**Remark 3.13.7** (Interpretation). Definition 3.13.6 separates the decision process into two aggregation layers:

1. *within-scenario aggregation*, where criterion-wise uncertain assessments are combined into a scenario-specific uncertain utility $u_i^{(r)}$;

2. *across-scenario aggregation*, where the family $(u_i^{(1)}, \ldots, u_i^{(s)})$ is combined into the overall uncertain utility $u_i$.

Thus UMScDM models decisions in which the performance of each alternative depends on several possible scenarios, while all evaluations remain encoded in a general uncertain model $M$.

**Remark 3.13.8** (Specializations). If $M$ is the fuzzy model with $\mathrm{Dom}(M) = [0,1]$, then Definition 3.13.6 reduces to a fuzzy multi-scenario decision-making framework. If $M$ is chosen as an intuitionistic fuzzy, neutrosophic, hesitant fuzzy, spherical fuzzy, plithogenic, or another uncertainty model, then one obtains the corresponding uncertainty-aware multi-scenario decision-making framework by replacing $\mathrm{Dom}(M)$, $\mathrm{Agg}_{C,M}$, $\mathrm{Agg}_{S,M}$, and $\mathrm{Score}_M$ accordingly.

**Theorem 3.13.9** (Well-definedness of UMScDM of type $M$)**.** *Let*

$$\mathrm{UMScDM}_M = \Big( \mathcal{A}, \mathcal{C}, \Sigma, (X_M^{(r)})_{r=1}^s, w_M, \pi, \mathrm{Agg}_{C,M}, \mathrm{Agg}_{S,M}, \mathrm{Score}_M \Big)$$

*be a UMScDM instance as in Definition 3.13.6. Assume:*

*(A1)* $\mathcal{A}$, $\mathcal{C}$, *and* $\Sigma$ *are finite nonempty sets;*

*(A2) for every* $r \in \{1, \ldots, s\}$,

$$X_M^{(r)} = (\mu_{ij}^{(r)}) \in \mathrm{Dom}(M)^{m \times n};$$

*(A3)*

$$w_M = (\omega_1, \ldots, \omega_n) \in \mathrm{Dom}(M)^n;$$

*(A4)*

$$\pi = (\pi_1, \ldots, \pi_s) \in [0,1]^s \qquad with \qquad \sum_{r=1}^s \pi_r = 1;$$

*(A5)*

$$\mathrm{Agg}_{C,M} : \mathrm{Dom}(M)^n \times \mathrm{Dom}(M)^n \to \mathrm{Dom}(M)$$

> *is a total map;*

*(A6)*

$$\mathrm{Agg}_{S,M} : \mathrm{Dom}(M)^s \times [0,1]^s \to \mathrm{Dom}(M)$$

> *is a total map;*

*(A7)*

$$\mathrm{Score}_M : \mathrm{Dom}(M) \to \mathbb{R}$$

> *is a total map.*

*Then the following objects are well-defined:*

*(i) for every alternative* $A_i$ *and every scenario* $\sigma_r$, *the scenario-specific uncertain utility*

$$u_i^{(r)} = \mathrm{Agg}_{C,M}\big((\mu_{i1}^{(r)}, \ldots, \mu_{in}^{(r)}), (\omega_1, \ldots, \omega_n)\big)$$

> *belongs to* $\mathrm{Dom}(M)$;

*(ii) for every alternative* $A_i$, *the overall uncertain utility*

$$u_i = \mathrm{Agg}_{S,M}\big((u_i^{(1)}, \ldots, u_i^{(s)}), \pi\big)$$

> *belongs to* $\mathrm{Dom}(M)$;

*(iii) for every alternative* $A_i$, *the real-valued score*

$$s_i = \mathrm{Score}_M(u_i)$$

> *is well-defined;*

*(iv) the induced preference relation $\succeq_M$ on $\mathcal{A}$, defined by*

$$A_i \succeq_M A_j \iff s_i \geq s_j,$$

*is a total preorder;*

*(v) the optimal set*

$$\mathcal{A}_M^\star = \arg\max_{A_i \in \mathcal{A}} s_i$$

*is nonempty.*

*Hence Uncertain Multi-Scenario Decision-Making of type $M$ is well-defined.*

*Proof.* By (A1), the sets

$$\mathcal{A} = \{A_1, \ldots, A_m\}, \qquad \mathcal{C} = \{C_1, \ldots, C_n\}, \qquad \Sigma = \{\sigma_1, \ldots, \sigma_s\}$$

are finite and nonempty.

### Step 1: Well-definedness of the scenario-specific uncertain utilities $u_i^{(r)}$.

Fix $i \in \{1, \ldots, m\}$ and $r \in \{1, \ldots, s\}$. By (A2), the $i$-th row of the scenario-dependent matrix $X_M^{(r)}$ is

$$(\mu_{i1}^{(r)}, \ldots, \mu_{in}^{(r)}) \in \mathrm{Dom}(M)^n.$$

By (A3),

$$(\omega_1, \ldots, \omega_n) \in \mathrm{Dom}(M)^n.$$

Hence

$$\big((\mu_{i1}^{(r)}, \ldots, \mu_{in}^{(r)}), (\omega_1, \ldots, \omega_n)\big) \in \mathrm{Dom}(M)^n \times \mathrm{Dom}(M)^n.$$

Since $\mathrm{Agg}_{C,M}$ is total by (A5), the value

$$u_i^{(r)} = \mathrm{Agg}_{C,M}\big((\mu_{i1}^{(r)}, \ldots, \mu_{in}^{(r)}), (\omega_1, \ldots, \omega_n)\big)$$

is defined and belongs to $\mathrm{Dom}(M)$. This proves (i).

### Step 2: Well-definedness of the overall uncertain utilities $u_i$.

Fix $i \in \{1, \ldots, m\}$. By Step 1, for each $r = 1, \ldots, s$,

$$u_i^{(r)} \in \mathrm{Dom}(M).$$

Therefore

$$(u_i^{(1)}, \ldots, u_i^{(s)}) \in \mathrm{Dom}(M)^s.$$

By (A4),

$$\pi = (\pi_1, \ldots, \pi_s) \in [0,1]^s.$$

Hence

$$\big((u_i^{(1)}, \ldots, u_i^{(s)}), \pi\big) \in \text{Dom}(M)^s \times [0,1]^s.$$

Since $\text{Agg}_{S,M}$ is total by (A6), the value

$$u_i = \text{Agg}_{S,M}\big((u_i^{(1)}, \ldots, u_i^{(s)}), \pi\big)$$

is defined and belongs to $\text{Dom}(M)$. Thus (ii) holds.

## Step 3: Well-definedness of the scores $s_i$.

By Step 2, $u_i \in \text{Dom}(M)$ for each $i$. Since $\text{Score}_M$ is total by (A7),

$$s_i := \text{Score}_M(u_i) \in \mathbb{R}$$

is well-defined for each $i = 1, \ldots, m$. Hence (iii) holds.

## Step 4: The induced preference relation is a total preorder.

Define

$$A_i \succeq_M A_j \iff s_i \geq s_j.$$

Since $\geq$ on $\mathbb{R}$ is reflexive, transitive, and total, the induced relation $\succeq_M$ on $\mathcal{A}$ is also reflexive, transitive, and total. Therefore $\succeq_M$ is a total preorder on $\mathcal{A}$. This proves (iv).

## Step 5: Existence of an optimal alternative.

The set of scores

$$\{s_1, \ldots, s_m\} \subset \mathbb{R}$$

is finite and nonempty because $\mathcal{A}$ is finite and nonempty. Every finite nonempty subset of $\mathbb{R}$ has a maximum. Hence there exists at least one index $i^\star \in \{1, \ldots, m\}$ such that

$$s_{i^\star} = \max_{1 \leq i \leq m} s_i.$$

Therefore

$$\mathcal{A}_M^\star = \arg\max_{A_i \in \mathcal{A}} s_i \neq \varnothing.$$

Thus (v) holds.

All required objects are therefore defined unambiguously, and the solution set is nonempty. Hence UMScDM of type $M$ is well-defined. $\qquad\square$

Table 3.13 presents related uncertainty-model variants of Fuzzy Multi-Scenario Decision-Making.

Table 3.13: Related uncertainty-model variants of Fuzzy Multi-Scenario Decision-Making.

| $k$ | Related Fuzzy Multi-Scenario Decision-Making variant(s) |
|---|---|
| 1 | Fuzzy Multi-Scenario Decision-Making |
| 2 | Intuitionistic Fuzzy Multi-Scenario Decision-Making |
| 3 | Hesitant Fuzzy Multi-Scenario Decision-Making |
| 3 | Spherical Fuzzy Multi-Scenario Decision-Making |
| 3 | Neutrosophic Multi-Scenario Decision-Making |
| $n$ | Plithogenic Multi-Scenario Decision-Making |

# Chapter 4

# Weight elicitation Decision-Methods

Weight elicitation estimates criterion importance in decision-making by surveys, pairwise comparisons, trade-off queries, entropy/optimization, or learning, producing normalized weights for aggregation and ranking. For reference, a comparison table is presented in Table 4.1.

Table 4.1: A concise comparison of fuzzy weight-elicitation methods.

| Method | Type | Typical input | Main idea / strength | Main limitation / note |
|---|---|---|---|---|
| Fuzzy AHP | Subjective | Fuzzy pairwise comparison matrix | Classical hierarchical weighting with consistency checking; intuitive, transparent, and widely used in practice. | Requires many pairwise comparisons when the number of criteria becomes large. |
| Fuzzy LOP-COW | Objective | Fuzzy decision matrix | Data-driven weighting based on criterion variability or discrimination power. | Does not directly incorporate the decision-maker's subjective preferences. |
| F-SIWEC | Subjective | Expert importance assessments | A simple expert-oriented weighting procedure with relatively low elicitation burden. | The precise mathematical formulation should be stated explicitly, since variants may differ across studies. |
| Fuzzy judgment matrix | Foundational tool | Fuzzy pairwise judgments | A basic structure for representing fuzzy preference information used by many weighting approaches. | Not a standalone weighting method unless coupled with a specific derivation rule. |
| Fuzzy ANP | Subjective | Fuzzy pairwise comparisons on a network | Extends AHP to settings with interdependent criteria and feedback relations. | Model construction and computation are generally more involved than in AHP. |

*Continued on the next page.*

*Table 4.1 (continued).*

| Method | Type | Typical input | Main idea / strength | Main limitation / note |
|---|---|---|---|---|
| Fuzzy OPA | Subjective | Ordinal rankings or priority orders | Requires less information than full pairwise comparison and is efficient in elicitation. | Uses less detailed preference information than cardinal comparison-based methods. |
| Fuzzy PI-PRECIA | Subjective | Sequential relative importance judgments | Stepwise weighting method with lower burden than full pairwise comparison. | Final weights may depend on the ordering of criteria. |
| Fuzzy SWARA | Subjective | Ordered criteria and comparative importance coefficients | Expert-friendly and easy-to-implement stepwise weighting approach. | Sensitive to criterion ordering and the reliability of expert judgments. |
| Fuzzy CILOS | Objective | Fuzzy decision matrix | Determines weights through the criterion impact-loss concept without requiring direct preference input. | Although preference-free, it may be less intuitive for practitioners. |
| Fuzzy IDOCRIW | Integrated objective | Fuzzy decision matrix | Integrates multiple objective weighting ideas, often combining CILOS-type and entropy-type information. | Computationally more complex than a single objective weighting scheme. |
| Fuzzy BWM | Subjective | Best-to-others and others-to-worst comparisons | Requires fewer comparisons than AHP and provides a consistency-based framework. | Requires clear and reliable identification of the best and worst criteria. |
| Fuzzy CRITIC | Objective | Fuzzy decision matrix | Uses both contrast intensity and inter-criterion conflict to derive weights. | Does not reflect direct expert preferences unless combined with a subjective approach. |
| Fuzzy MEREC | Objective | Fuzzy decision matrix | Evaluates the effect of removing each criterion on the overall performance structure. | Can be sensitive to the selected normalization procedure and performance aggregation model. |
| Fuzzy FU-COM | Subjective | Ranked criteria and comparative priorities | Achieves weighting with very few comparisons while emphasizing consistency. | Requires a reliable prior ranking of criteria before weight derivation. |

## 4.1 Fuzzy Analytic hierarchy process (Fuzzy AHP)

AHP structures decisions into a hierarchy, uses pairwise comparisons to derive weights, and synthesizes global priorities for ranking [415–417]. Fuzzy AHP replaces crisp comparisons with fuzzy numbers, computes fuzzy weights and defuzzified priorities to rank alternatives under uncertainty [418–420].

**Definition 4.1.1** (TFNs and FAHP). (cf. [421,422]) **(0) Positive TFN, arithmetic, and defuzzification.** A *positive triangular fuzzy number* (TFN) is $\tilde{x} = (l, m, u) \in \mathbb{R}_{>0}^3$ with $0 < l \le m \le u$. For positive TFNs $\tilde{x} = (l_x, m_x, u_x)$ and $\tilde{y} = (l_y, m_y, u_y)$, define componentwise

$$\tilde{x} \oplus \tilde{y} := (l_x + l_y,\ m_x + m_y,\ u_x + u_y), \quad \tilde{x} \otimes \tilde{y} := (l_x l_y,\ m_x m_y,\ u_x u_y),$$

$$\tilde{x}^\alpha := (l_x^\alpha,\ m_x^\alpha,\ u_x^\alpha)\ (\alpha > 0), \quad \tilde{x}^{-1} := (1/u_x,\ 1/m_x,\ 1/l_x), \quad \tilde{x} \oslash \tilde{y} := \tilde{x} \otimes \tilde{y}^{-1}.$$

A standard crisp representative (centroid/COA) is

$$\mathrm{COA}(\tilde{x}) := \frac{l + m + u}{3}.$$

**(1) Decision hierarchy.** A *decision hierarchy* is a rooted tree $\mathcal{H} = (V, \mathrm{root}, \mathrm{ch})$, where each internal node $p \in V$ has children $\mathrm{ch}(p) = \{e_1, \ldots, e_n\}$ representing criteria/subcriteria, and leaves represent alternatives.

**(2) Fuzzy pairwise comparisons at a node.** Fix an internal node $p$ with children $\mathrm{ch}(p) = \{e_1, \ldots, e_n\}$. A *fuzzy reciprocal pairwise comparison matrix* at $p$ is

$$\tilde{A}^{(p)} = (\tilde{a}_{ij}^{(p)}) \in (\mathsf{TFN}_{>0})^{n \times n}, \qquad \tilde{a}_{ii}^{(p)} = (1,1,1), \quad \tilde{a}_{ji}^{(p)} = (\tilde{a}_{ij}^{(p)})^{-1}.$$

**(3) Local weights by fuzzy geometric mean.** For each row $i$, define the fuzzy geometric mean

$$\tilde{g}_i^{(p)} := \left( \bigotimes_{j=1}^n \tilde{a}_{ij}^{(p)} \right)^{1/n},$$

and the normalized local fuzzy weight

$$\tilde{w}_i^{(p)} := \tilde{g}_i^{(p)} \oslash \left( \bigoplus_{k=1}^n \tilde{g}_k^{(p)} \right).$$

Optionally obtain crisp local weights by $w_i^{(p)} := \mathrm{COA}(\tilde{w}_i^{(p)})$ and normalize $(w_1^{(p)}, \ldots, w_n^{(p)})$ to sum to 1.

**(4) Hierarchical synthesis (global priorities).** Set $W(\mathrm{root}) = 1$ and propagate priorities down the tree:

$$W(e_i) := W(p)\, w_i^{(p)} \qquad \text{for each edge } p \to e_i.$$

For a leaf (alternative) $a$, $W(a)$ is its final priority and alternatives are ranked by decreasing $W(a)$. (A fully fuzzy-end variant replaces products by $\otimes$ and defuzzifies only at the end.)

**Remark 4.1.2** (Reduction to classical AHP). If $\tilde{a}_{ij}^{(p)} = (a_{ij}^{(p)}, a_{ij}^{(p)}, a_{ij}^{(p)})$ for all $i, j, p$, then the above reduces to the corresponding crisp AHP geometric-mean weighting and synthesis.

By extending AHP using Uncertain Sets, we obtain the following formulation.

**Definition 4.1.3** (Uncertain AHP (UAHP) of type $M$). Let $\mathcal{H} = (V, \text{root}, \text{ch})$ be a finite rooted tree (decision hierarchy), where each internal node $p \in V$ has a finite nonempty set of children

$$\text{ch}(p) = \{e_1, \ldots, e_{n_p}\},$$

representing criteria/subcriteria, and the leaves $\mathsf{Alt} \subseteq V$ represent alternatives. Fix an *uncertainty model* $M$ with degree-domain $\text{Dom}(M) \subseteq [0,1]^d$ ($d \geq 1$).

**(1) Local uncertain pairwise comparisons.** For each internal node $p \in V$, let $n_p := |\text{ch}(p)|$ and assume a local *uncertain pairwise comparison matrix*

$$A^{(p)} = \big(a_{ij}^{(p)}\big) \in \text{Dom}(M)^{n_p \times n_p},$$

subject to the basic AHP normalization constraints in $\text{Dom}(M)$:

$$a_{ii}^{(p)} = \mathbf{1}_M \quad (i = 1, \ldots, n_p), \qquad a_{ji}^{(p)} = \text{Inv}_M\big(a_{ij}^{(p)}\big) \quad (i \neq j),$$

where $\mathbf{1}_M \in \text{Dom}(M)$ is the model's "unit-comparison" element and $\text{Inv}_M : \text{Dom}(M) \to \text{Dom}(M)$ is a (total) reciprocal/inversion operator in $\text{Dom}(M)$.

**(2) Local priority extraction.** Fix a (total) priority-extraction operator

$$\text{Pri}_M : \text{Dom}(M)^{n \times n} \longrightarrow \Delta_n, \qquad \Delta_n := \Big\{ x \in \mathbb{R}_{\geq 0}^n : \sum_{i=1}^{n} x_i = 1 \Big\},$$

which, given a local matrix $A^{(p)}$, returns a *crisp* local priority vector

$$w^{(p)} = \text{Pri}_M\big(A^{(p)}\big) \in \Delta_{n_p}.$$

(Thus $w_i^{(p)}$ is the relative importance of child $e_i \in \text{ch}(p)$ w.r.t. node $p$.)

**(3) Hierarchical synthesis (global priorities).** Define global priorities $W : V \to [0,1]$ recursively by

$$W(\text{root}) = 1, \qquad W(e_i) = W(p)\, w_i^{(p)} \quad \text{for each edge } p \to e_i.$$

For each alternative $a \in \mathsf{Alt}$, its final priority is $W(a)$, and alternatives are ranked by decreasing $W(a)$.

**(4) UAHP output as an uncertain set over alternatives.** Let $\mathsf{Alt} = \{A_1, \ldots, A_m\}$ be the leaf set. Define a membership map

$$\mu_{\mathcal{U}} : \mathsf{Alt} \to [0,1], \qquad \mu_{\mathcal{U}}(A_i) := W(A_i).$$

Then $\mathcal{U} := (\mathsf{Alt}, \mu_{\mathcal{U}})$ is called the *UAHP priority uncertain set* induced by the hierarchy and uncertain comparisons.

**Theorem 4.1.4** (Uncertain-set structure and well-definedness of UAHP). *Consider a UAHP instance of type $M$ as in Definition 4.1.3. Assume:*

(A1) *The hierarchy $\mathcal{H}$ is a finite rooted tree and every internal node has finitely many children.*

(A2) $\mathrm{Dom}(M) \subseteq [0,1]^d$ *is nonempty,* $\mathbf{1}_M \in \mathrm{Dom}(M)$ *is fixed, and* $\mathrm{Inv}_M : \mathrm{Dom}(M) \to \mathrm{Dom}(M)$ *is total.*

(A3) *For each internal node* $p$, *the matrix* $A^{(p)} \in \mathrm{Dom}(M)^{n_p \times n_p}$ *is fully specified and satisfies* $a_{ii}^{(p)} = \mathbf{1}_M$ *and* $a_{ji}^{(p)} = \mathrm{Inv}_M(a_{ij}^{(p)})$.

(A4) $\mathrm{Pri}_M$ *is total and satisfies* $\mathrm{Pri}_M(A) \in \Delta_n$ *for every* $A \in \mathrm{Dom}(M)^{n \times n}$.

*Then:*

(i) *The global priority map* $W : V \to [0,1]$ *is well-defined and satisfies* $\sum_{a \in \mathsf{Alt}} W(a) = 1$.

(ii) *The induced mapping* $\mu_{\mathcal{U}} : \mathsf{Alt} \to [0,1]$ *is well-defined; hence* $\mathcal{U} = (\mathsf{Alt}, \mu_{\mathcal{U}})$ *is a well-defined uncertain set on* $\mathsf{Alt}$.

(iii) *The UAHP solution set*
$$\arg\max_{A \in \mathsf{Alt}} \; \mu_{\mathcal{U}}(A)$$
*is nonempty; therefore at least one optimal alternative exists.*

*Proof.* **(i) Well-definedness of local weights.** Fix an internal node $p$ with $n_p$ children. By (A3), $A^{(p)}$ is a well-defined element of $\mathrm{Dom}(M)^{n_p \times n_p}$. By (A4), applying the total map $\mathrm{Pri}_M$ yields a uniquely determined vector
$$w^{(p)} = \mathrm{Pri}_M(A^{(p)}) \in \Delta_{n_p}.$$
In particular, $w_i^{(p)} \geq 0$ and $\sum_{i=1}^{n_p} w_i^{(p)} = 1$.

**(ii) Well-definedness of $W$ and normalization on leaves.** Define $W(\mathrm{root}) = 1$. For any node $v \neq \mathrm{root}$, there is a unique directed path $\mathrm{root} = v_0 \to v_1 \to \cdots \to v_\ell = v$ in a rooted tree. Define recursively $W(v_t) = W(v_{t-1}) \, w_{\iota_t}^{(v_{t-1})}$, where $v_t$ is the $\iota_t$-th child of $v_{t-1}$. This gives a unique real value $W(v) \geq 0$, hence $W$ is well-defined on all nodes.

To show $\sum_{a \in \mathsf{Alt}} W(a) = 1$, proceed by induction on the tree. For each internal node $p$, the total weight assigned to its children is
$$\sum_{e \in \mathrm{ch}(p)} W(e) = \sum_{i=1}^{n_p} W(p) \, w_i^{(p)} = W(p) \sum_{i=1}^{n_p} w_i^{(p)} = W(p).$$

Thus, the mass $W(p)$ is conserved when propagated from $p$ to its children. Starting from $W(\mathrm{root}) = 1$, iterating this conservation down to the leaves implies the sum of leaf weights equals 1.

**(iii) Uncertain-set output and existence of an optimum.** By (ii), each $W(a) \in [0,1]$ is well-defined for every alternative $a \in \mathsf{Alt}$, so $\mu_{\mathcal{U}}(a) = W(a)$ defines a well-defined mapping $\mu_{\mathcal{U}} : \mathsf{Alt} \to [0,1]$ and hence $\mathcal{U} = (\mathsf{Alt}, \mu_{\mathcal{U}})$ is an uncertain set.

Because $\mathsf{Alt}$ is finite and nonempty, the real-valued function $\mu_{\mathcal{U}}$ attains its maximum on $\mathsf{Alt}$, so $\arg\max_{A \in \mathsf{Alt}} \mu_{\mathcal{U}}(A) \neq \varnothing$. $\qquad\square$

Table 4.2: Related uncertainty-model variants of AHP (classified by the degree-domain dimension $k$).

| $k$ | Related AHP variant(s) |
|---|---|
| 1 | Fuzzy AHP [423, 424] |
| 2 | Intuitionistic Fuzzy AHP [425, 426] |
| 2 | Pythagorean Fuzzy AHP [427, 428] |
| 3 | Hesitant Fuzzy AHP [429, 430] |
| 3 | Spherical Fuzzy AHP [431, 432] |
| 3 | Neutrosophic AHP [433, 434] |
| $n$ | Plithogenic AHP [435, 436] |

Related uncertainty-model variants of AHP, classified by the degree-domain dimension $k$, are listed in Table 4.2.

In addition to Uncertain AHP, related concepts such as Rough AHP [437,438], Soft AHP [439,440], Modified analytic hierarchy process [441,442], TOPSIS-AHP [443,444], Interval AHP [445,446], Grey AHP [447,448], AHP–BOCR [449, 450], Linking Pin AHP [451], AHPSort [452, 453], Stochastic AHP [454, 455], AHP-K [456, 457], Incomplete AHP [458, 459], Monte Carlo analytic hierarchical process (MCAHP) [460, 461], Entropy-weight AHP [462, 463], and Linguistic AHP [464, 465] have also been studied.

## 4.2 Fuzzy LOPCOW (Fuzzy Linear Optimization for Comprehensive Weight)

LOPCOW computes objective criterion weights from normalized data using logarithmic percentage change of dispersion, emphasizing informative criteria [466,467]. Fuzzy LOPCOW defuzzifies fuzzy ratings, normalizes performances, computes RMS-to-variance log coefficients, and yields objective weights under uncertainty [468,469].

**Definition 4.2.1** (Fuzzy LOPCOW objective weighting (score-based, well-defined form)). [468, 469] Let $\mathcal{A} = \{A_1, \ldots, A_m\}$ be a finite set of alternatives and $\mathcal{C} = \{C_1, \ldots, C_n\}$ a finite set of criteria. Let $\mathsf{FN}(\mathbb{R})$ be a chosen class of fuzzy numbers (e.g., TFNs). Assume a fuzzy decision matrix

$$\widetilde{X} = (\tilde{x}_{ij}) \in \mathsf{FN}(\mathbb{R})^{m \times n}.$$

Let $\mathcal{C} = \mathcal{C}^+ \dot\cup \mathcal{C}^-$ be a partition into benefit ($\uparrow$) and cost ($\downarrow$) criteria.

**(0) Score/defuzzification map.** Fix a map (ranking/defuzzification)

$$S : \mathsf{FN}(\mathbb{R}) \to \mathbb{R},$$

and define the induced crisp matrix $X = (x_{ij}) \in \mathbb{R}^{m \times n}$ by

$$x_{ij} := S(\tilde{x}_{ij}).$$

(For TFNs $\tilde{x} = (l, m, u)$, a common choice is $S(\tilde{x}) = (l + m + u)/3$.)

**(1) Linear normalization.** For each criterion $j$, set

$$x_j^{\max} := \max_{1 \le i \le m} x_{ij}, \qquad x_j^{\min} := \min_{1 \le i \le m} x_{ij}.$$

Define the normalized matrix $R = (r_{ij}) \in [0,1]^{m \times n}$ by

$$r_{ij} := \begin{cases} \dfrac{x_j^{\max} - x_{ij}}{x_j^{\max} - x_j^{\min}}, & C_j \in \mathcal{C}^- \quad (\text{cost}), \\[2ex] \dfrac{x_{ij} - x_j^{\min}}{x_j^{\max} - x_j^{\min}}, & C_j \in \mathcal{C}^+ \quad (\text{benefit}), \end{cases}$$

with the convention that if $x_j^{\max} = x_j^{\min}$ then $r_{ij} := 0$ for all $i$ (zero-variation criterion).

**(2) Dispersion statistics per criterion.** For each $j$, define the root-mean-square (RMS) and the standard deviation:

$$\text{RMS}_j := \sqrt{\frac{1}{m} \sum_{i=1}^{m} r_{ij}^2}, \qquad \bar{r}_j := \frac{1}{m} \sum_{i=1}^{m} r_{ij}, \qquad \sigma_j := \sqrt{\frac{1}{m} \sum_{i=1}^{m} (r_{ij} - \bar{r}_j)^2}.$$

(So $\sigma_j \geq 0$ and $\text{RMS}_j \geq 0$.)

**(3) Logarithmic percentage-change coefficient.** Define the LOPCOW percentage value of criterion $j$ by

$$PV_j := \begin{cases} \left| \ln\left( \dfrac{\text{RMS}_j}{\sigma_j} \right) \right| \cdot 100, & \sigma_j > 0, \\[2ex] 0, & \sigma_j = 0, \end{cases}$$

where ln denotes the natural logarithm.

**(4) Objective criterion weights.** If $\sum_{\ell=1}^{n} PV_\ell > 0$, define

$$w_j := \frac{PV_j}{\sum_{\ell=1}^{n} PV_\ell} \in [0,1], \qquad j = 1, \dots, n,$$

so that $\sum_{j=1}^{n} w_j = 1$. If $\sum_{\ell=1}^{n} PV_\ell = 0$ (all criteria have zero variation after normalization), set $w_j := 1/n$.

The resulting vector $w = (w_1, \dots, w_n)$ is called the *Fuzzy LOPCOW (objective) weight vector* associated with $(\widetilde{X}, S, \mathcal{C}^+, \mathcal{C}^-)$.

The definition of Uncertain LOPCOW (ULOPCOW) is given below.

**Definition 4.2.2** (Uncertain LOPCOW (ULOPCOW) objective weighting)**.** Let $\mathcal{A} = \{A_1, \dots, A_m\}$ be a finite set of alternatives and $\mathcal{C} = \{C_1, \dots, C_n\}$ a finite set of criteria. Fix an uncertainty model $M$ with degree-domain $\text{Dom}(M) \subseteq [0,1]^d$ $(d \geq 1)$. Assume an *uncertain decision matrix*

$$\widetilde{X} = (\tilde{x}_{ij}) \in \text{Dom}(M)^{m \times n},$$

where $\tilde{x}_{ij}$ is the uncertain evaluation of $A_i$ under $C_j$. Let $\mathcal{C} = \mathcal{C}^+ \,\dot{\cup}\, \mathcal{C}^-$ be a partition into benefit and cost criteria.

**(0) Score (crisp representative) map.** Fix a total mapping

$$S_M : \text{Dom}(M) \to \mathbb{R},$$

and define the induced crisp matrix $X = (x_{ij}) \in \mathbb{R}^{m \times n}$ by

$$x_{ij} := S_M(\tilde{x}_{ij}).$$

(Examples: for TFNs $\tilde{x} = (l, m, u)$, one may take $S_M(\tilde{x}) = (l + m + u)/3$; for interval values $\tilde{x} = [\ell, u]$, one may take $S_M(\tilde{x}) = (\ell + u)/2$.)

**(1) Linear normalization (benefit/cost).** For each criterion $j$, set

$$x_j^{\max} := \max_{1 \le i \le m} x_{ij}, \qquad x_j^{\min} := \min_{1 \le i \le m} x_{ij}.$$

Define the normalized matrix $R = (r_{ij}) \in [0,1]^{m \times n}$ by

$$r_{ij} := \begin{cases} \dfrac{x_j^{\max} - x_{ij}}{x_j^{\max} - x_j^{\min}}, & C_j \in \mathcal{C}^- \quad \text{(cost)}, \\[2mm] \dfrac{x_{ij} - x_j^{\min}}{x_j^{\max} - x_j^{\min}}, & C_j \in \mathcal{C}^+ \quad \text{(benefit)}, \end{cases}$$

with the convention that if $x_j^{\max} = x_j^{\min}$ then $r_{ij} := 0$ for all $i$.

**(2) Dispersion statistics per criterion.** For each $j$, define

$$\text{RMS}_j := \sqrt{\frac{1}{m} \sum_{i=1}^{m} r_{ij}^2}, \qquad \bar{r}_j := \frac{1}{m} \sum_{i=1}^{m} r_{ij}, \qquad \sigma_j := \sqrt{\frac{1}{m} \sum_{i=1}^{m} (r_{ij} - \bar{r}_j)^2}.$$

**(3) LOPCOW percentage value.** Define

$$PV_j := \begin{cases} \left| \ln\left( \dfrac{\text{RMS}_j}{\sigma_j} \right) \right| \cdot 100, & \sigma_j > 0, \\[3mm] 0, & \sigma_j = 0, \end{cases} \qquad j = 1, \dots, n,$$

where ln denotes the natural logarithm.

**(4) Objective weight vector.** If $\sum_{\ell=1}^{n} PV_\ell > 0$, define

$$w_j := \frac{PV_j}{\sum_{\ell=1}^{n} PV_\ell} \in [0,1], \qquad j = 1, \dots, n,$$

so that $\sum_{j=1}^{n} w_j = 1$. If $\sum_{\ell=1}^{n} PV_\ell = 0$, set $w_j := 1/n$.

The vector $w = (w_1, \dots, w_n)$ is called the *Uncertain LOPCOW (ULOPCOW) objective weight vector* associated with $(\widetilde{X}, S_M, \mathcal{C}^+, \mathcal{C}^-)$.

**Theorem 4.2.3** (Uncertain-set output and well-definedness of ULOPCOW). *Let $\widetilde{X} \in \text{Dom}(M)^{m \times n}$ be given as in Definition 4.2.2, with $m, n \ge 1$. Assume:*

(A1) *The score map $S_M : \mathrm{Dom}(M) \to \mathbb{R}$ is total (defined for every element of $\mathrm{Dom}(M)$).*

(A2) *The benefit/cost partition $\mathcal{C} = \mathcal{C}^+ \dot\cup \mathcal{C}^-$ is fixed.*

*Then:*

(i) *The normalized matrix $R = (r_{ij})$ defined in (4.2.2) is well-defined and satisfies $0 \le r_{ij} \le 1$ for all $i, j$.*

(ii) *For each $j$, $\mathrm{RMS}_j$ and $\sigma_j$ are well-defined real numbers and satisfy $\mathrm{RMS}_j \ge 0$, $\sigma_j \ge 0$.*

(iii) *Each $PV_j$ is well-defined and satisfies $PV_j \ge 0$.*

(iv) *The weight vector $w = (w_1, \dots, w_n)$ is well-defined and belongs to the probability simplex*

$$\Delta_n = \Big\{ w \in \mathbb{R}^n_{\ge 0} : \sum_{j=1}^n w_j = 1 \Big\}.$$

(v) *The mapping*

$$\mu_w : \mathcal{C} \to [0, 1], \qquad \mu_w(C_j) := w_j,$$

*defines an* uncertain set $\mathcal{W} := (\mathcal{C}, \mu_w)$ *(the* ULOPCOW *weight uncertain set).*

*Proof.* **(i) Normalization is well-defined and bounded.** Fix $j \in \{1, \dots, n\}$. By (A1), each $x_{ij} = S_M(\tilde{x}_{ij})$ is a real number, hence $x_j^{\max}, x_j^{\min} \in \mathbb{R}$ are well-defined. If $x_j^{\max} = x_j^{\min}$, we set $r_{ij} = 0$ by convention, so $r_{ij}$ is defined.

Assume $x_j^{\max} > x_j^{\min}$. For benefit criteria,

$$r_{ij} = \frac{x_{ij} - x_j^{\min}}{x_j^{\max} - x_j^{\min}},$$

and since $x_j^{\min} \le x_{ij} \le x_j^{\max}$, it follows that $0 \le r_{ij} \le 1$. For cost criteria,

$$r_{ij} = \frac{x_j^{\max} - x_{ij}}{x_j^{\max} - x_j^{\min}},$$

and again $0 \le r_{ij} \le 1$ by the same bounding.

**(ii) Dispersion statistics exist and are nonnegative.** For each fixed $j$, the values $\{r_{ij}\}_{i=1}^m$ are real and bounded by (i). Therefore $\mathrm{RMS}_j$ is the square root of an average of squares of real numbers and is well-defined with $\mathrm{RMS}_j \ge 0$. Similarly, $\bar{r}_j$ is a finite average and $\sigma_j$ is the square root of an average of squared deviations, hence is well-defined with $\sigma_j \ge 0$.

**(iii) Log-percentage values are well-defined.** If $\sigma_j = 0$, then $PV_j$ is defined to be 0. If $\sigma_j > 0$, then $\mathrm{RMS}_j \ge 0$. Moreover, if $\sigma_j > 0$ then the column $\{r_{ij}\}$ is not constant, so at least one $r_{ij} \ne 0$ and hence $\mathrm{RMS}_j > 0$. Thus $\mathrm{RMS}_j / \sigma_j > 0$ and $\ln(\mathrm{RMS}_j / \sigma_j)$ is defined. Taking absolute value and multiplying by 100 yields $PV_j \ge 0$.

**(iv) Weights form a simplex vector.** If $\sum_{\ell=1}^{n} PV_\ell > 0$, then each $w_j$ is a nonnegative ratio, so $w_j \geq 0$ and

$$\sum_{j=1}^{n} w_j = \sum_{j=1}^{n} \frac{PV_j}{\sum_{\ell=1}^{n} PV_\ell} = \frac{\sum_{j=1}^{n} PV_j}{\sum_{\ell=1}^{n} PV_\ell} = 1.$$

If $\sum_{\ell=1}^{n} PV_\ell = 0$, then $w_j = 1/n \geq 0$ and $\sum_{j=1}^{n} w_j = 1$.

**(v) Uncertain-set structure on criteria.** By (iv), $w_j \in [0,1]$ for all $j$. Hence $\mu_w(C_j) := w_j$ defines a well-defined membership map $\mu_w : \mathcal{C} \to [0,1]$, so $\mathcal{W} = (\mathcal{C}, \mu_w)$ is an uncertain set. □

Related concepts of LOPCOW under uncertainty-aware models are listed in Table 4.3.

Table 4.3: Related concepts of LOPCOW under uncertainty-aware models.

| $k$ | **Related LOPCOW concept(s)** |
|---|---|
| 2 | Intuitionistic Fuzzy LOPCOW |
| 2 | $q$-Rung Fuzzy LOPCOW [470, 471] |
| 3 | Hesitant Fuzzy LOPCOW [472] |
| 3 | Spherical Fuzzy LOPCOW [473] |
| 3 | Neutrosophic LOPCOW [474] |

## 4.3 Fuzzy SIWEC (Fuzzy simple weight calculation for criteria)

SIWEC derives criterion weights from experts' direct ratings, normalizes scores, aggregates across experts, and outputs a weight vector [475, 476]. Fuzzy SIWEC maps linguistic ratings to fuzzy numbers, normalizes, scales by dispersion, aggregates, then defuzzifies or ranks fuzzy weights [47, 477].

**Definition 4.3.1** (Fuzzy SIWEC (F-SIWEC): fuzzy simple weight calculation for criteria). [477, 478] Let $E = \{1, \ldots, e\}$ be the set of expert decision makers (EDMs) and $C = \{C_1, \ldots, C_n\}$ the set of criteria.

**Step 1 (linguistic evaluation → triangular fuzzy numbers).** Each EDM $i \in E$ evaluates each criterion $C_j$ using a linguistic term that is mapped to a triangular fuzzy number (TFN)

$$\tilde{x}_{ij} = (x_{ij}^l, x_{ij}^m, x_{ij}^u) \in \mathbb{R}_+^3, \qquad x_{ij}^l \leq x_{ij}^m \leq x_{ij}^u.$$

Collect these ratings into the fuzzy decision matrix

$$\tilde{X} = [\tilde{x}_{ij}]_{e \times n}.$$

**Step 2 (normalization).** For each criterion $j$, let

$$M_j := \max_{i \in E} x_{ij}^u,$$

and define the normalized TFN

$$\tilde{n}_{ij} = \left( \frac{x_{ij}^l}{M_j}, \ \frac{x_{ij}^m}{M_j}, \ \frac{x_{ij}^u}{M_j} \right).$$

**Step 3 (standard-deviation scaling).** Defuzzify each normalized TFN by

$$n_{ij}^{\mathrm{def}} := \frac{n_{ij}^l + 4n_{ij}^m + n_{ij}^u}{6},$$

and compute the (sample) standard deviation for each criterion $j$ across EDMs:

$$\mathrm{stdev}_j := \sqrt{\frac{1}{e-1}\sum_{i=1}^{e}\left(n_{ij}^{\mathrm{def}} - \bar{n}_j^{\mathrm{def}}\right)^2}, \qquad \bar{n}_j^{\mathrm{def}} := \frac{1}{e}\sum_{i=1}^{e} n_{ij}^{\mathrm{def}}.$$

Scale the normalized TFNs by this dispersion:

$$\tilde{v}_{ij} := \mathrm{stdev}_j \odot \tilde{n}_{ij} = (\mathrm{stdev}_j n_{ij}^l,\ \mathrm{stdev}_j n_{ij}^m,\ \mathrm{stdev}_j n_{ij}^u).$$

**Step 4 (aggregation over EDMs).** For each criterion $j$, aggregate scaled TFNs across EDMs:

$$\tilde{s}_j := \bigoplus_{i=1}^{e} \tilde{v}_{ij} = (s_j^l, s_j^m, s_j^u),$$

where $\oplus$ denotes componentwise addition of TFNs.

**Step 5 (fuzzy weight vector).** Define the fuzzy weight of criterion $C_j$ as the TFN

$$\tilde{w}_j := (w_j^l, w_j^m, w_j^u) = \left(\frac{s_j^l}{\sum_{k=1}^{n} s_k^u},\ \frac{s_j^m}{\sum_{k=1}^{n} s_k^m},\ \frac{s_j^u}{\sum_{k=1}^{n} s_k^l}\right), \qquad j = 1, \dots, n.$$

**Step 6 (optional defuzzification and normalization).** A crisp weight can be obtained by

$$w_j^{\mathrm{def}} := \frac{w_j^l + 4w_j^m + w_j^u}{6} \in \mathbb{R}_+.$$

If required, renormalize:

$$w_j^* := \frac{w_j^{\mathrm{def}}}{\sum_{k=1}^{n} w_k^{\mathrm{def}}}.$$

The output $\tilde{w} = (\tilde{w}_1, \dots, \tilde{w}_n)$ (or $w^*$) is called the *Fuzzy SIWEC* criteria-weight vector.

Fuzzy SIWEC is extended by using Uncertain Sets.

**Definition 4.3.2** (Uncertain sets and uncertain numbers)**.** Let $U$ be a nonempty universe. An *uncertain set* on $U$ is a mapping

$$\mu : U \to [0, 1],$$

where $\mu(u)$ is interpreted as the degree of uncertainty-membership of $u$.

An *uncertain number* is an uncertain set $\tilde{x}$ on $\mathbb{R}$ whose $\alpha$-cuts

$$[\tilde{x}]_\alpha := \{t \in \mathbb{R} : \mu_{\tilde{x}}(t) \geq \alpha\} \qquad (\alpha \in (0,1])$$

are nonempty compact intervals.  Write

$$[\tilde{x}]_\alpha = [x_\alpha^-, x_\alpha^+] \quad (\alpha \in (0,1]).$$

Denote by $\mathsf{UN}$ a chosen class of uncertain numbers closed under the operations below, and

$$\mathsf{UN}_{>0} := \{\tilde{x} \in \mathsf{UN} : \operatorname{supp}(\tilde{x}) \subseteq (0,\infty)\}, \quad \operatorname{supp}(\tilde{x}) := \overline{\{t : \mu_{\tilde{x}}(t) > 0\}}.$$

Define the *upper support endpoint* (a positive scalar representative) by

$$\operatorname{up}(\tilde{x}) := \sup \operatorname{supp}(\tilde{x}) \in (0,\infty) \qquad (\tilde{x} \in \mathsf{UN}_{>0}).$$

**Definition 4.3.3** ((Recall) Uncertain-number arithmetic via $\alpha$-cuts)**.** Let $\tilde{x}, \tilde{y} \in \mathsf{UN}$ and $c \geq 0$.  Define:

$$[\tilde{x} \oplus \tilde{y}]_\alpha := [x_\alpha^- + y_\alpha^-, \; x_\alpha^+ + y_\alpha^+], \qquad [c \odot \tilde{x}]_\alpha := [c\,x_\alpha^-, \; c\,x_\alpha^+],$$

for all $\alpha \in (0,1]$.  (These are Minkowski sum and scalar multiplication of interval $\alpha$-cuts.)

Let $\operatorname{Score} : \mathsf{UN} \to \mathbb{R}_{\geq 0}$ be a fixed *score* (crisp representative) map satisfying:

(S1)  (Positivity) $\operatorname{Score}(\tilde{x}) \geq 0$ for all $\tilde{x} \in \mathsf{UN}$, and $\operatorname{Score}(\tilde{x}) > 0$ for all $\tilde{x} \in \mathsf{UN}_{>0}$;

(S2)  (Positive homogeneity) $\operatorname{Score}(c \odot \tilde{x}) = c \operatorname{Score}(\tilde{x})$ for all $c \geq 0$;

(S3)  (Additivity) $\operatorname{Score}(\tilde{x} \oplus \tilde{y}) = \operatorname{Score}(\tilde{x}) + \operatorname{Score}(\tilde{y})$.

(For instance, one may take $\operatorname{Score}(\tilde{x}) = \int_0^1 \frac{x_\alpha^- + x_\alpha^+}{2}\,d\alpha$ when it is finite.)

Here, we define Uncertain SIWEC (U-SIWEC), obtained by extending SIWEC using Uncertain Sets.

**Definition 4.3.4** (Uncertain SIWEC (U-SIWEC): uncertain simple integrated weight estimation for criteria)**.** Let $E = \{1, \ldots, e\}$ be a finite set of experts and $C = \{C_1, \ldots, C_n\}$ a finite set of criteria.  An *uncertain SIWEC instance* is an uncertain evaluation matrix

$$\widetilde{X} = (\tilde{x}_{ij}) \in (\mathsf{UN}_{>0})^{e \times n}, \qquad \tilde{x}_{ij} \text{ is expert } i\text{'s uncertain rating of criterion } C_j.$$

Fix up and Score as in Definitions 4.3.2–4.3.3.  The *U-SIWEC procedure* constructs criterion weights as follows.

**Step 1 (normalization by upper support).**  For each criterion $j$, set

$$M_j := \max_{i \in E} \operatorname{up}(\tilde{x}_{ij}) \in (0,\infty),$$

and define normalized uncertain ratings

$$\tilde{n}_{ij} := \frac{1}{M_j} \odot \tilde{x}_{ij} \in \mathsf{UN}_{>0}.$$

**Step 2 (crisp representatives).** Define

$$n_{ij}^{\mathrm{cr}} := \mathrm{Score}(\tilde{n}_{ij}) \in \mathbb{R}_{>0}.$$

**Step 3 (dispersion across experts).** For each criterion $j$, compute the mean and (population) standard deviation

$$\bar{n}_j^{\mathrm{cr}} := \frac{1}{e} \sum_{i=1}^{e} n_{ij}^{\mathrm{cr}}, \qquad \sigma_j := \sqrt{\frac{1}{e} \sum_{i=1}^{e} \left( n_{ij}^{\mathrm{cr}} - \bar{n}_j^{\mathrm{cr}} \right)^2} \in \mathbb{R}_{\geq 0}.$$

**Step 4 (dispersion scaling in the uncertain domain).** Scale each normalized uncertain rating by $\sigma_j$:

$$\tilde{v}_{ij} := \sigma_j \odot \tilde{n}_{ij} \in \mathsf{UN}.$$

**Step 5 (aggregation over experts).** Aggregate expert information for each criterion by

$$\tilde{s}_j := \bigoplus_{i=1}^{e} \tilde{v}_{ij} \in \mathsf{UN}.$$

**Step 6 (objective weights).** Let

$$S_j := \mathrm{Score}(\tilde{s}_j) \in \mathbb{R}_{\geq 0}, \qquad T := \sum_{k=1}^{n} S_k \in \mathbb{R}_{\geq 0}.$$

Define the *crisp weight vector* $w = (w_1, \dots, w_n)$ by

$$w_j := \begin{cases} \dfrac{S_j}{T}, & T > 0, \\ \dfrac{1}{n}, & T = 0. \end{cases}$$

Optionally, define an *uncertain weight vector* $\tilde{w} = (\tilde{w}_1, \dots, \tilde{w}_n)$ by

$$\tilde{w}_j := \begin{cases} \dfrac{1}{T} \odot \tilde{s}_j, & T > 0, \\ \dfrac{1}{n} \odot \tilde{u}, & T = 0, \end{cases}$$

where $\tilde{u} \in \mathsf{UN}_{>0}$ is any fixed reference uncertain number (e.g. $\tilde{u}$ with $\mathrm{Score}(\tilde{u}) = 1$).

**Theorem 4.3.5** (Uncertain-set structure and well-definedness of U-SIWEC). *Assume $\widetilde{X} \in (\mathsf{UN}_{>0})^{e \times n}$ and that $\mathsf{UN}$ is closed under $\oplus$ and $c \odot (\cdot)$ for $c \geq 0$. Assume* Score *satisfies* (S1)–(S3) *in Definition 4.3.3. Then the U-SIWEC procedure in Definition 4.3.4 is well-defined and satisfies:*

(i) *Each $\tilde{n}_{ij}$, $\tilde{v}_{ij}$, and $\tilde{s}_j$ is an uncertain number; hence each is an uncertain set on $\mathbb{R}$.*

(ii) *The crisp output $w = (w_1, \ldots, w_n)$ is a valid criterion-weight vector:*

$$w_j \geq 0 \text{ for all } j, \qquad \sum_{j=1}^n w_j = 1.$$

(iii) *If $T > 0$, then the uncertain output satisfies $\tilde{w}_j \in \mathsf{UN}$ and*

$$\mathrm{Score}(\tilde{w}_j) = w_j, \qquad \sum_{j=1}^n \mathrm{Score}(\tilde{w}_j) = 1.$$

*Proof.* **(i) Closure / uncertain-set structure.** Fix $j$. Since each $\tilde{x}_{ij} \in \mathsf{UN}_{>0}$, the upper endpoint $\mathrm{up}(\tilde{x}_{ij})$ is finite and positive, hence $M_j = \max_i \mathrm{up}(\tilde{x}_{ij}) \in (0, \infty)$ exists. By closure of $\mathsf{UN}$ under scalar multiplication, $\tilde{n}_{ij} = (1/M_j) \odot \tilde{x}_{ij} \in \mathsf{UN}_{>0}$ is defined. Next $\sigma_j \in \mathbb{R}_{\geq 0}$ is defined by a finite sum of real numbers $n_{ij}^{\mathrm{cr}} = \mathrm{Score}(\tilde{n}_{ij})$. Again by closure, $\tilde{v}_{ij} = \sigma_j \odot \tilde{n}_{ij} \in \mathsf{UN}$, and then $\tilde{s}_j = \bigoplus_{i=1}^e \tilde{v}_{ij} \in \mathsf{UN}$ by closure under $\oplus$. Each of these objects is, by definition, an uncertain number and therefore an uncertain set on $\mathbb{R}$.

**(ii) Nonnegativity of scores and weights.** By (S1), $S_j = \mathrm{Score}(\tilde{s}_j) \geq 0$ for all $j$, hence $T = \sum_k S_k \geq 0$ and $w_j \geq 0$ in both branches ($T > 0$ and $T = 0$).

**(iii) Normalization $\sum_j w_j = 1$.** If $T > 0$, then

$$\sum_{j=1}^n w_j = \sum_{j=1}^n \frac{S_j}{T} = \frac{1}{T} \sum_{j=1}^n S_j = \frac{T}{T} = 1.$$

If $T = 0$, then $w_j = 1/n$ gives $\sum_j w_j = 1$.

**(iv) Consistency of uncertain weights when $T > 0$.** Assume $T > 0$ and define $\tilde{w}_j = (1/T) \odot \tilde{s}_j$. Then $\tilde{w}_j \in \mathsf{UN}$ by closure. Using (S2),

$$\mathrm{Score}(\tilde{w}_j) = \mathrm{Score}\left( \frac{1}{T} \odot \tilde{s}_j \right) = \frac{1}{T} \mathrm{Score}(\tilde{s}_j) = \frac{S_j}{T} = w_j.$$

Summing yields $\sum_j \mathrm{Score}(\tilde{w}_j) = \sum_j w_j = 1$. $\qquad \square$

For reference, related concepts of SIWEC under uncertainty-aware models are listed in Table 4.4.

Table 4.4: Related concepts of SIWEC under uncertainty-aware models.

| $k$ | Related SIWEC concept(s) |
|---|---|
| 1 | Fuzzy SIWEC |
| 2 | Intuitionistic Fuzzy SIWEC |
| 3 | Neutrosophic SIWEC |
| $n$ | Plithogenic SIWEC |

## 4.4 Fuzzy judgment matrix

Judgment matrix is a pairwise-comparison matrix in which each entry $a_{ij}$ represents the relative preference (or importance) of alternative $X_i$ over alternative $X_j$; it is often assumed reciprocal and is used to derive priority weights and to assess consistency [479–481]. A fuzzy judgment matrix is a pairwise-comparison matrix whose entries lie in $[0, 1]$ and quantify graded preference degrees, so that $a_{ij} > 0.5$ favors $X_i$ over $X_j$, $a_{ij} = 0.5$ indicates indifference, and (in the complementary case) $a_{ij} + a_{ji} = 1$ [482, 483].

**Definition 4.4.1** (Fuzzy judgment matrix). [482, 483] Let $X = \{X_1, \ldots, X_n\}$ be a finite set of alternatives. A matrix

$$A = (a_{ij}) \in \mathbb{R}^{n \times n}$$

is called a *fuzzy judgment matrix* if

$$0 \leq a_{ij} \leq 1 \qquad (\forall\, i, j \in \{1, \ldots, n\}),$$

where $a_{ij}$ is interpreted as the (graded) preference degree of $X_i$ over $X_j$ in a pairwise comparison.

A common semantic convention is:

$$a_{ij} > 0.5 \Rightarrow X_i \text{ is preferred to } X_j,$$

$$a_{ij} = 0.5 \Rightarrow X_i \text{ and } X_j \text{ are indifferent,}$$

$$a_{ij} < 0.5 \Rightarrow X_j \text{ is preferred to } X_i.$$

**Definition 4.4.2** (Fuzzy complementary judgment matrix). A fuzzy judgment matrix $A = (a_{ij})$ is called *fuzzy complementary* if it satisfies

$$a_{ij} + a_{ji} = 1 \qquad (\forall\, i, j \in \{1, \ldots, n\}),$$

(which implies $a_{ii} = 0.5$ for all $i$).

**Definition 4.4.3** (Uncertain preference numbers and arithmetic). Let

$$\mathsf{UN}_{[0,1]} := \{\tilde{x} \in \mathsf{UN} : \mathrm{supp}(\tilde{x}) \subseteq [0, 1]\}.$$

For $\tilde{x}, \tilde{y} \in \mathsf{UN}_{[0,1]}$ define the *(truncated) sum* and *complement* by $\alpha$-cuts:

$$[\tilde{x} \oplus_T \tilde{y}]_\alpha := \big[\min\{1, x_\alpha^- + y_\alpha^-\},\ \min\{1, x_\alpha^+ + y_\alpha^+\}\big],$$

$$[\mathbf{1} \ominus \tilde{x}]_\alpha := \big[1 - x_\alpha^+,\ 1 - x_\alpha^-\big], \qquad \alpha \in (0, 1],$$

where $[\tilde{x}]_\alpha = [x_\alpha^-, x_\alpha^+]$ and $[\tilde{y}]_\alpha = [y_\alpha^-, y_\alpha^+]$. (Thus $\mathbf{1} \ominus \tilde{x}$ is the pointwise complement of $\tilde{x}$ on $[0, 1]$.)

Next, we extend the judgment matrix using Uncertain Sets. The definition is given below.

**Definition 4.4.4** (Uncertain judgment matrix). Let $X = \{X_1, \ldots, X_n\}$ be a finite set of alternatives. An *uncertain judgment matrix* (UJM) on $X$ is a matrix

$$\widetilde{A} = (\tilde{a}_{ij}) \in \left(\mathsf{UN}_{[0,1]}\right)^{n \times n},$$

where $\tilde{a}_{ij}$ is interpreted as the uncertain (graded) preference of $X_i$ over $X_j$. A semantic convention is induced by Score:

$$\text{Score}(\tilde{a}_{ij}) > 0.5 \Rightarrow X_i \text{ is preferred to } X_j, \qquad \text{Score}(\tilde{a}_{ij}) = 0.5 \Rightarrow X_i \text{ and } X_j \text{ are indifferent.}$$

The UJM is called:

(i) *uncertain complementary* if

$$\tilde{a}_{ij} = \mathbf{1} \ominus \tilde{a}_{ji} \qquad (\forall\, i, j),$$

(which implies $\tilde{a}_{ii}$ is the "uncertain indifference" around 0.5);

(ii) *uncertain reciprocal* if all entries are supported in $(0, 1]$ and a fixed uncertain-number product $\otimes$ is available such that

$$\tilde{a}_{ij} \otimes \tilde{a}_{ji} = \tilde{\mathbf{1}} \qquad (\forall\, i, j),$$

for a unit uncertain number $\tilde{\mathbf{1}}$ concentrated at 1.

**Theorem 4.4.5** (Uncertain-set structure and well-definedness). *Let $\widetilde{A} = (\tilde{a}_{ij}) \in (\mathsf{UN}_{[0,1]})^{n \times n}$. Then:*

(i) *Each entry $\tilde{a}_{ij}$ is an uncertain set on $[0, 1]$ and, a fortiori, an uncertain set on $\mathbb{R}$. Hence an uncertain judgment matrix is a matrix whose entries carry uncertain-set structure.*

(ii) *The complement operator $\mathbf{1} \ominus (\cdot)$ in Definition 4.4.3 is well-defined on $\mathsf{UN}_{[0,1]}$ and maps $\mathsf{UN}_{[0,1]}$ into itself.*

(iii) *If $\widetilde{A}$ is uncertain complementary, then the complement relation is coherent: for all $i, j$ and all $\alpha \in (0, 1]$,*

$$[\tilde{a}_{ij}]_\alpha = [1 - a_{ji,\alpha}^+, \, 1 - a_{ji,\alpha}^-] \quad \Longrightarrow \quad \text{Score}(\tilde{a}_{ij}) = 1 - \text{Score}(\tilde{a}_{ji})$$

*whenever* Score *is affine with respect to complements, i.e.* $\text{Score}(\mathbf{1} \ominus \tilde{x}) = 1 - \text{Score}(\tilde{x})$.

*Proof.* **(i)** By definition, $\tilde{a}_{ij} \in \mathsf{UN}_{[0,1]} \subseteq \mathsf{UN}$ is an uncertain number, hence an uncertain set on $\mathbb{R}$. Therefore each matrix entry carries uncertain-set structure.

**(ii)** Fix $\tilde{x} \in \mathsf{UN}_{[0,1]}$ and write $[\tilde{x}]_\alpha = [x_\alpha^-, x_\alpha^+]$ with $0 \leq x_\alpha^- \leq x_\alpha^+ \leq 1$. Then Definition 4.4.3 gives

$$[\mathbf{1} \ominus \tilde{x}]_\alpha = [1 - x_\alpha^+, \, 1 - x_\alpha^-].$$

Since $0 \leq 1 - x_\alpha^+ \leq 1 - x_\alpha^- \leq 1$, each $\alpha$-cut is a nonempty compact interval in $[0, 1]$. Hence $\mathbf{1} \ominus \tilde{x} \in \mathsf{UN}_{[0,1]}$, so the operation is well-defined and closed.

**(iii)** If $\widetilde{A}$ is uncertain complementary, then $\tilde{a}_{ij} = \mathbf{1} \ominus \tilde{a}_{ji}$ by definition. Assuming $\text{Score}(\mathbf{1} \ominus \tilde{x}) = 1 - \text{Score}(\tilde{x})$, we obtain

$$\text{Score}(\tilde{a}_{ij}) = \text{Score}(\mathbf{1} \ominus \tilde{a}_{ji}) = 1 - \text{Score}(\tilde{a}_{ji}),$$

which is exactly the coherence of complementary semantics at the score level. $\qquad \square$

Related concepts of judgment matrices under uncertainty-aware models are listed in Table 4.5.

Table 4.5: Related concepts of judgment matrices under uncertainty-aware models.

| $k$ | Related judgment matrix concept(s) |
|---|---|
| 1 | Fuzzy Judgment Matrix |
| 2 | Intuitionistic Fuzzy Judgment Matrix [426, 484] |
| 3 | Hesitant Fuzzy Judgment Matrix [485–487] |
| 3 | Spherical Fuzzy Judgment Matrix [488, 489] |
| 3 | Neutrosophic Judgment Matrix [490, 491] |

## 4.5 Fuzzy Analytic network process (ANP)

ANP evaluates interdependent criteria and alternatives using pairwise comparisons, builds a supermatrix, and derives global priorities via limit supermatrix [492–494]. Fuzzy ANP extends ANP with fuzzy pairwise judgments, forming fuzzy supermatrices and defuzzified limit priorities to handle uncertainty [39, 495, 496].

**Definition 4.5.1** (Fuzzy Analytic Network Process (FANP): supermatrix formulation)**.** [497, 498] Let $\mathcal{C} = \{C_1, \ldots, C_m\}$ be clusters, where $C_i = \{e_{i1}, \ldots, e_{in_i}\}$ with $n_i \geq 1$, and let

$$E := \bigsqcup_{i=1}^{m} C_i$$

be the set of all elements. Let $\mathcal{D} \subseteq \mathcal{C} \times \mathcal{C}$ be a directed dependence relation, where $(C_i, C_j) \in \mathcal{D}$ means that "$C_i$ influences $C_j$" (allowing inner dependence $i = j$ and outer dependence $i \neq j$).

Fix a class $\mathcal{FN}_{>0}$ of positive fuzzy numbers, closed under the operations used below, and fix a ranking (defuzzification) map

$$\text{Score} : \mathcal{FN}_{>0} \longrightarrow \mathbb{R}_{>0}.$$

**(0) Fuzzy pairwise comparisons and local priorities.** A *positive reciprocal fuzzy pairwise comparison matrix* is

$$\widetilde{A} = (\tilde{a}_{rs}) \in \mathcal{FN}_{>0}^{n \times n},$$

satisfying

$$\tilde{a}_{rr} = 1, \qquad \tilde{a}_{sr} = \tilde{a}_{rs}^{-1} \qquad (1 \leq r, s \leq n).$$

Its *local fuzzy priority vector*

$$\widetilde{w} = (\tilde{w}_1, \ldots, \tilde{w}_n) \in \mathcal{FN}_{>0}^n$$

is obtained, for example, by the fuzzy geometric mean method:

$$\tilde{g}_r := \Big(\prod_{s=1}^{n} \tilde{a}_{rs}\Big)^{1/n}, \qquad \tilde{w}_r := \tilde{g}_r \Big/ \Big(\sum_{t=1}^{n} \tilde{g}_t\Big), \qquad r = 1, \ldots, n,$$

where the product, power, sum, and division are understood in the fuzzy-number sense (e.g. via the extension principle or an equivalent admissible fuzzy arithmetic). The associated *crisp normalized priority vector* is defined by

$$w_r^{\sharp} := \frac{\text{Score}(\tilde{w}_r)}{\sum_{t=1}^{n} \text{Score}(\tilde{w}_t)}, \qquad r = 1, \ldots, n.$$

Then $w^{\sharp} = (w_1^{\sharp}, \ldots, w_n^{\sharp}) \in \mathbb{R}_{\geq 0}^n$ and

$$\sum_{r=1}^{n} w_r^{\sharp} = 1.$$

**(1) Element-to-element influence vectors and unweighted supermatrix.** Fix $(C_i, C_j) \in \mathcal{D}$ and a target element $e_{jk} \in C_j$. Compare the elements of $C_i$ pairwise with respect to their influence on $e_{jk}$ by a positive reciprocal fuzzy comparison matrix

$$\widetilde{A}^{(i \to j \,|\, k)} \in \mathcal{FN}_{>0}^{n_i \times n_i}.$$

Let

$$w^{\sharp(i \to j \,|\, k)} \in \mathbb{R}_{\geq 0}^{n_i}$$

be the corresponding crisp normalized local priority vector obtained from the above procedure. Define the block $W_{ij} \in \mathbb{R}^{n_i \times n_j}$ by

$$(W_{ij})_{\bullet k} := w^{\sharp(i \to j \,|\, k)} \qquad (k = 1, \ldots, n_j),$$

and set $W_{ij} := 0$ if $(C_i, C_j) \notin \mathcal{D}$. The resulting block matrix

$$W := \big(W_{ij}\big)_{1 \leq i,j \leq m} \in \mathbb{R}^{(\sum_i n_i) \times (\sum_j n_j)}$$

is called the *unweighted supermatrix*.

**(2) Cluster weights and weighted (column-stochastic) supermatrix.** For each cluster $C_j$, let

$$\Gamma(j) := \{\, i : (C_i, C_j) \in \mathcal{D} \,\}.$$

Obtain fuzzy cluster-comparison judgments among the influencing clusters $\{C_i : i \in \Gamma(j)\}$, compute the corresponding fuzzy cluster-priority vector, and then defuzzify/normalize it to get

$$\alpha^{(j),\sharp} = (\alpha_i^{(j),\sharp})_{i \in \Gamma(j)} \in \mathbb{R}_{\geq 0}^{|\Gamma(j)|}, \qquad \sum_{i \in \Gamma(j)} \alpha_i^{(j),\sharp} = 1.$$

Define the weighted blocks

$$\bar{W}_{ij} := \begin{cases} \alpha_i^{(j),\sharp} \, W_{ij}, & i \in \Gamma(j), \\ 0, & i \notin \Gamma(j), \end{cases} \qquad \bar{W} := (\bar{W}_{ij})_{1 \leq i,j \leq m}.$$

Then $\bar{W}$ is column-stochastic, i.e. each column sums to 1.

**(3) Limit supermatrix and global priorities.** If the limit exists, define the *limit supermatrix* by

$$W^\infty := \lim_{t \to \infty} \bar{W}^t.$$

If $\bar{W}^t$ is cyclic with period $N$, use the cycle-average

$$W^\infty := \frac{1}{N} \sum_{t=0}^{N-1} \bar{W}^t.$$

Let $A \subseteq E$ be the designated set of alternatives (usually a subset of the elements in one cluster). The *global priority* of an alternative is read from the corresponding row of $W^\infty$ (equivalently, from stabilized columns), and the alternatives are ranked in decreasing order of global priority.

We now extend Fuzzy ANP using Uncertain Sets. The related definitions are given below.

**Definition 4.5.2** (Uncertain set and uncertain number). Let $U$ be a nonempty universe. An *uncertain set* on $U$ is a mapping $\mu : U \to [0, 1]$.

An *uncertain number* is an uncertain set $\tilde{x}$ on $\mathbb{R}$ whose $\alpha$-cuts

$$[\tilde{x}]_\alpha := \{t \in \mathbb{R} : \mu_{\tilde{x}}(t) \geq \alpha\} \qquad (\alpha \in (0, 1])$$

are nonempty compact intervals. Let $\mathsf{UN}$ be a fixed class of uncertain numbers and set

$$\mathsf{UN}_{>0} := \{\tilde{x} \in \mathsf{UN} : \mathrm{supp}(\tilde{x}) \subseteq (0, \infty)\}.$$

Assume $\mathsf{UN}_{>0}$ is closed under the *uncertain inverse*: if $[\tilde{x}]_\alpha = [x_\alpha^-, x_\alpha^+]$ with $0 < x_\alpha^- \leq x_\alpha^+$, define

$$[\tilde{x}^{-1}]_\alpha := \left[\frac{1}{x_\alpha^+}, \frac{1}{x_\alpha^-}\right] \qquad (\alpha \in (0, 1]).$$

Fix a *score* (crisp representative) map

$$\mathrm{Score} : \mathsf{UN}_{>0} \to (0, \infty),$$

and assume it is *reciprocity-compatible*:

$$\mathrm{Score}(\tilde{x}^{-1}) = \frac{1}{\mathrm{Score}(\tilde{x})} \quad (\forall \tilde{x} \in \mathsf{UN}_{>0}), \qquad \mathrm{Score}(\tilde{\mathbf{1}}) = 1,$$

where $\tilde{\mathbf{1}}$ is the uncertain number concentrated at 1.

**Definition 4.5.3** (Uncertain reciprocal judgment matrix). Let $n \geq 2$. An *uncertain reciprocal judgment matrix* is

$$\widetilde{A} = (\tilde{a}_{rs}) \in (\mathsf{UN}_{>0})^{n \times n}$$

such that

$$\tilde{a}_{rr} = \tilde{\mathbf{1}} \quad (\forall r), \qquad \tilde{a}_{sr} = \tilde{a}_{rs}^{-1} \quad (\forall r \neq s).$$

Its induced crisp reciprocal matrix is

$$A := (a_{rs}) \in (0, \infty)^{n \times n}, \qquad a_{rs} := \mathrm{Score}(\tilde{a}_{rs}).$$

**Definition 4.5.4** (Uncertain ANP (UANP): score-induced supermatrix formulation). Let $\mathcal{C} = \{C_1, \ldots, C_m\}$ be *clusters*, where

$$C_i = \{e_{i1}, \ldots, e_{in_i}\} \quad (n_i \geq 1), \qquad E := \bigsqcup_{i=1}^m C_i$$

is the set of all elements. Let $\mathcal{D} \subseteq \mathcal{C} \times \mathcal{C}$ be a directed dependence relation; $(C_i, C_j) \in \mathcal{D}$ means "$C_i$ influences $C_j$" (allowing $i = j$).

**(0) Local priorities from uncertain pairwise judgments.** Fix $(C_i, C_j) \in \mathcal{D}$ and a target element $e_{jk} \in C_j$. Decision makers provide an uncertain reciprocal judgment matrix

$$\widetilde{A}^{(i \to j \,|\, k)} \in (\mathsf{UN}_{>0})^{n_i \times n_i}$$

comparing the elements of $C_i$ with respect to their influence on $e_{jk}$. Let

$$A^{(i \to j \,|\, k)} := \left(\mathrm{Score}(\tilde{a}_{rs}^{(i \to j \,|\, k)})\right) \in (0, \infty)^{n_i \times n_i}$$

be the induced crisp reciprocal matrix. Define the *local priority vector* $w^{(i \to j \,|\, k)} \in \mathbb{R}^{n_i}_{>0}$ as the normalized Perron vector of $A^{(i \to j \,|\, k)}$:

$$A^{(i \to j \,|\, k)} \, w^{(i \to j \,|\, k)} = \lambda_{\max} \, w^{(i \to j \,|\, k)}, \qquad \sum_{r=1}^{n_i} w_r^{(i \to j \,|\, k)} = 1.$$

**(1) Unweighted supermatrix.** For each $(i, j)$, define the block $W_{ij} \in \mathbb{R}^{n_i \times n_j}_{\geq 0}$ by setting its $k$th column as

$$(W_{ij})_{\bullet k} := w^{(i \to j \,|\, k)} \quad (k = 1, \dots, n_j),$$

and set $W_{ij} := 0$ if $(C_i, C_j) \notin \mathcal{D}$. The *unweighted supermatrix* is the block matrix

$$W := \big(W_{ij}\big)_{1 \leq i,j \leq m} \in \mathbb{R}^{N \times N}_{\geq 0}, \qquad N := \sum_{i=1}^{m} n_i.$$

**(2) Cluster weights and weighted supermatrix.** For each target cluster $C_j$, let $\Gamma(j) := \{i : (C_i, C_j) \in \mathcal{D}\}$. Cluster weights are obtained by uncertain pairwise comparisons among clusters in $\Gamma(j)$, yielding (after applying Score and Perron normalization) a vector

$$\alpha^{(j)} = (\alpha_i^{(j)})_{i \in \Gamma(j)} \in \mathbb{R}^{|\Gamma(j)|}_{\geq 0}, \qquad \sum_{i \in \Gamma(j)} \alpha_i^{(j)} = 1.$$

Define the weighted blocks

$$\bar{W}_{ij} := \begin{cases} \alpha_i^{(j)} \, W_{ij}, & i \in \Gamma(j), \\ 0, & i \notin \Gamma(j), \end{cases} \qquad \bar{W} := \big(\bar{W}_{ij}\big)_{1 \leq i,j \leq m} \in \mathbb{R}^{N \times N}_{\geq 0}.$$

**(3) Limit supermatrix and global priorities.** If $\bar{W}$ is *primitive* (some power has strictly positive entries), define

$$W^{\infty} := \lim_{t \to \infty} \bar{W}^{\,t}.$$

In general (even if periodic), define the Cesàro limit (always used in practice when cycling occurs):

$$W^{\infty} := \lim_{T \to \infty} \frac{1}{T} \sum_{t=0}^{T-1} \bar{W}^{\,t}, \quad \text{whenever the limit exists.}$$

Let $A \subseteq E$ be the designated set of alternatives (elements representing alternatives). The *global priority* of $a \in A$ is read from the corresponding row of $W^{\infty}$ (equivalently from stabilized columns), and alternatives are ranked by decreasing global priority.

**Theorem 4.5.5** (Uncertain-set structure and well-definedness of UANP). *Under the assumptions of Definitions 4.5.2–4.5.4, the UANP construction is well-defined in the following sense.*

(i) *(Uncertain-set structure) Every entry $\tilde{a}_{rs}^{(i \to j \,|\, k)} \in \mathsf{UN}_{>0}$ is an uncertain set on $(0, \infty)$ (hence on $\mathbb{R}$). Thus each local comparison matrix $\widetilde{A}^{(i \to j \,|\, k)}$ is a matrix of uncertain sets.*

(ii) *(Local priorities are well-defined)* *For each local uncertain reciprocal judgment matrix $\widetilde{A}^{(i \to j \,|\, k)}$, the induced crisp matrix $A^{(i \to j \,|\, k)} = (\mathrm{Score}(\tilde{a}_{rs}^{(i \to j \,|\, k)}))$ is a positive reciprocal matrix. Hence its Perron eigenvector exists, is unique up to scaling, and the normalized vector $w^{(i \to j \,|\, k)}$ is uniquely determined in the simplex $\Delta_{n_i} := \{x \in \mathbb{R}_{\geq 0}^{n_i} : \sum_r x_r = 1\}$.*

(iii) *(Weighted supermatrix is column-stochastic)* *Each column of $\bar{W}$ sums to 1, i.e.*

$$\mathbf{1}^\top \bar{W} = \mathbf{1}^\top,$$

*where $\mathbf{1}$ is the all-ones vector in $\mathbb{R}^N$.*

(iv) *(Limit priorities are well-defined)* *If $\bar{W}$ is primitive, then the limit $W^\infty = \lim_{t \to \infty} \bar{W}^t$ exists. Consequently, global priorities read from $W^\infty$ are well-defined (independent of the iteration index for sufficiently large $t$). More generally, whenever the Cesàro limit in Definition 4.5.4 exists, it yields a well-defined steady-state priority extraction even in cyclic cases.*

*Proof.* **(i)** Each $\tilde{a}_{rs} \in \mathsf{UN}_{>0}$ is an uncertain set on $\mathbb{R}$. Hence every local matrix $\widetilde{A}^{(i \to j \,|\, k)}$ is entrywise an uncertain set.

**(ii)** Let $\widetilde{A} = (\tilde{a}_{rs})$ be an uncertain reciprocal judgment matrix. Then $\tilde{a}_{rr} = \tilde{\mathbf{1}}$ implies $a_{rr} = \mathrm{Score}(\tilde{\mathbf{1}}) = 1$. Also $\tilde{a}_{sr} = \tilde{a}_{rs}^{-1}$ and reciprocity-compatibility give

$$a_{sr} = \mathrm{Score}(\tilde{a}_{sr}) = \mathrm{Score}(\tilde{a}_{rs}^{-1}) = \frac{1}{\mathrm{Score}(\tilde{a}_{rs})} = \frac{1}{a_{rs}}.$$

Thus $A = (a_{rs})$ is a positive reciprocal matrix. By the Perron–Frobenius theorem, $A$ has a (strictly) positive principal eigenvector, unique up to a positive scalar; normalizing it to sum to 1 yields a unique $w \in \Delta_n$.

**(iii)** Fix a target cluster $C_j$ and an element $e_{jk} \in C_j$. The $k$th column of the block $W_{ij}$ equals $w^{(i \to j \,|\, k)}$ for each $i \in \Gamma(j)$, so

$$\sum_{r=1}^{n_i} (W_{ij})_{rk} = \sum_{r=1}^{n_i} w_r^{(i \to j \,|\, k)} = 1.$$

In the weighted supermatrix, the same column is scaled by $\alpha_i^{(j)}$ and summed over $i \in \Gamma(j)$, giving

$$\sum_{i \in \Gamma(j)} \sum_{r=1}^{n_i} (\bar{W}_{ij})_{rk} = \sum_{i \in \Gamma(j)} \alpha_i^{(j)} \sum_{r=1}^{n_i} (W_{ij})_{rk} = \sum_{i \in \Gamma(j)} \alpha_i^{(j)} = 1.$$

Hence every column of $\bar{W}$ sums to 1, i.e. $\mathbf{1}^\top \bar{W} = \mathbf{1}^\top$.

**(iv)** If $\bar{W}$ is primitive and column-stochastic, then $\bar{W}^\top$ is a primitive row-stochastic matrix. By standard Markov-chain (Perron–Frobenius) convergence, $(\bar{W}^\top)^t$ converges as $t \to \infty$, hence $\bar{W}^t$ converges as well, and $W^\infty := \lim_{t \to \infty} \bar{W}^t$ exists. Therefore the priorities extracted from $W^\infty$ are well-defined. If $\bar{W}$ is not primitive, the Cesàro averaging is the standard remedy; whenever the Cesàro limit exists, it yields a unique steady-state projection for priority extraction. $\square$

Related concepts of ANP under uncertainty-aware models are listed in Table 4.6.

As concepts other than Uncertain ANP, DEMATEL-ANP [507, 508], BOCR-based ANP [509], Group ANP [510, 511], ANP-TOPSIS [510, 512], and Rough ANP [513, 514] are also known.

Table 4.6: Related concepts of ANP under uncertainty-aware models.

| $k$ | Related ANP concept(s) |
|---|---|
| 2 | Intuitionistic Fuzzy ANP [499, 500] |
| 2 | Pythagorean Fuzzy ANP [501, 502] |
| 3 | Hesitant Fuzzy ANP [503] |
| 3 | Spherical Fuzzy ANP [504] |
| 3 | Neutrosophic ANP [505, 506] |

## 4.6 Fuzzy Ordinal Priority Approach (OPA)

Ordinal Priority Approach (OPA) is a max–min optimization method that converts experts' ordinal rankings into consistent criterion weights, without pairwise comparisons, yielding a prioritized alternative ordering [515, 516]. Fuzzy Ordinal Priority Approach (OPA) extends OPA by representing linguistic importance as triangular fuzzy numbers, solving a fuzzy max–min model to obtain fuzzy weights and rankings [517–519].

**Definition 4.6.1** (Fuzzy Ordinal Priority Approach (OPA-F): TFN-based formulation). [517–519] Let $I$ be a finite set of experts, $J$ a finite set of criteria, and $K$ a finite set of alternatives. Put $n := |J|$. Let

$$\mathsf{TFN} := \{(l, m, u) \in \mathbb{R}^3 : \ l \leq m \leq u\}$$

be the set of triangular fuzzy numbers (TFNs). We use the componentwise partial order on $\mathsf{TFN}$:

$$(l_1, m_1, u_1) \succeq (l_2, m_2, u_2) \iff l_1 \geq l_2, \ m_1 \geq m_2, \ u_1 \geq u_2.$$

(Any fixed TFN ranking/defuzzification rule can be used later for producing a crisp ranking.)

**Fuzzy linguistic inputs.** Let $\mathcal{L}$ be a linguistic scale and let $\varphi : \mathcal{L} \to \mathsf{TFN}$ map each term to a TFN. Each expert $i \in I$ provides an *ordinal ranking* of criteria, represented by a permutation

$$\pi_i : \{1, \ldots, n\} \to J,$$

where $\pi_i(1)$ is the most important criterion under expert $i$ and $\pi_i(n)$ is the least important. In addition, expert $i$ provides a linguistic importance assessment for each ranked criterion, encoded as TFNs

$$\tilde{a}_{i,r} := \varphi(\ell_{i,r}) \in \mathsf{TFN}, \qquad r = 1, \ldots, n,$$

where $\ell_{i,r} \in \mathcal{L}$ is the linguistic term attached to the $r$-th ranked criterion $\pi_i(r)$.

**Decision variables (fuzzy weights).** Introduce TFN decision variables

$$\tilde{W}_{i,r} \in \mathsf{TFN}_{\geq 0} \quad (i \in I, \ r = 1, \ldots, n), \qquad \tilde{Z} \in \mathsf{TFN},$$

where $\tilde{W}_{i,r}$ is the (fuzzy) weight assigned by expert $i$ to the $r$-th ranked criterion $\pi_i(r)$, and $\tilde{Z}$ is the max–min (bottleneck) objective variable.

**TFN arithmetic convention.** For $\tilde{x} = (l_x, m_x, u_x)$ and $\tilde{y} = (l_y, m_y, u_y)$ and $\lambda \geq 0$, use

$$\tilde{x} \oplus \tilde{y} := (l_x + l_y, \ m_x + m_y, \ u_x + u_y), \qquad \lambda \odot \tilde{x} := (\lambda l_x, \ \lambda m_x, \ \lambda u_x),$$

and the common triangular subtraction approximation

$$\tilde{x} \ominus \tilde{y} := (l_x - u_y, \ m_x - m_y, \ u_x - l_y).$$

(Any other consistent TFN arithmetic can be substituted if preferred.)

**OPA-F core model (fuzzy max–min program).** Fix a (possibly fuzzy) normalization constant $\tilde{1} \in \mathsf{TFN}$, typically $\tilde{1} = (1,1,1)$ (or a tolerance form such as $(1-\varepsilon, 1, 1+\varepsilon)$). The *Fuzzy Ordinal Priority Approach (OPA-F)* determines fuzzy criterion weights by solving:

$$\max \ \tilde{Z} \tag{4.1}$$

$$\text{s.t. } \tilde{a}_{i,r} \odot \left( \tilde{W}_{i,r} \ominus \tilde{W}_{i,r+1} \right) \ \succeq \ \tilde{Z}, \qquad \forall i \in I, \ \forall r = 1, \ldots, n-1, \tag{4.2}$$

$$\tilde{a}_{i,n} \odot \tilde{W}_{i,n} \ \succeq \ \tilde{Z}, \qquad \forall i \in I, \tag{4.3}$$

$$\bigoplus_{i \in I} \bigoplus_{r=1}^{n} \tilde{W}_{i,r} \ = \ \tilde{1}, \tag{4.4}$$

$$\tilde{W}_{i,r} = (l_{i,r}^w, m_{i,r}^w, u_{i,r}^w), \quad 0 \le l_{i,r}^w \le m_{i,r}^w \le u_{i,r}^w, \qquad \forall i \in I, \ \forall r. \tag{4.5}$$

**Aggregated criterion weights and ranking.** Define the aggregated fuzzy weight of criterion $j \in J$ by collecting the ranks where $j$ appears:

$$\tilde{W}_j := \bigoplus_{i \in I} \tilde{W}_{i,r(i,j)}, \quad \text{where } r(i,j) \text{ is the unique rank with } \pi_i(r(i,j)) = j.$$

A crisp weight vector can be obtained via any defuzzification/ranking map $\mathrm{Defuzz} : \mathsf{TFN} \to \mathbb{R}_{\ge 0}$, e.g. $\mathrm{Defuzz}(l,m,u) = (l+m+u)/3$, followed by normalization:

$$w_j := \frac{\mathrm{Defuzz}(\tilde{W}_j)}{\sum_{t \in J} \mathrm{Defuzz}(\tilde{W}_t)}.$$

**Reduction to crisp OPA.** If all inputs are degenerate TFNs $(l,m,u) = (p,p,p)$ and one takes $\tilde{1} = (1,1,1)$, then (4.1)–(4.5) reduces to a crisp OPA-type max–min linear program.

By extending it using Uncertain Sets, we obtain the following formulation.

**Definition 4.6.2** (Uncertainty space, uncertain set, membership, expected value)**.** An *uncertainty space* is a triple $(\Gamma, \mathcal{L}, \mathcal{M})$ where $\Gamma$ is a nonempty set, $\mathcal{L}$ is a $\sigma$-algebra on $\Gamma$, and $\mathcal{M} : \mathcal{L} \to [0,1]$ is an *uncertain measure* (satisfying, e.g., normality, duality, and subadditivity).

An *uncertain set* is a measurable mapping

$$\xi : \Gamma \to \mathcal{P}(\mathbb{R}),$$

such that for every Borel set $B \subseteq \mathbb{R}$, the events

$$\{\gamma \in \Gamma : \ \xi(\gamma) \subseteq B\} \in \mathcal{L}, \qquad \{\gamma \in \Gamma : \ B \subseteq \xi(\gamma)\} \in \mathcal{L}.$$

An uncertain set $\xi$ is said to have a *membership function* $\mu_\xi : \mathbb{R} \to [0, 1]$ if it represents the uncertain-set membership semantics in the standard way (equivalently, $\sup_{x \in \mathbb{R}} \mu_\xi(x) = 1$).

If $\xi$ is a nonempty uncertain set, its *expected value* is defined by

$$\mathbb{E}[\xi] := \int_0^\infty \mathcal{M}\{\xi \geq r\}\, dr - \int_{-\infty}^0 \mathcal{M}\{\xi \leq r\}\, dr,$$

provided at least one of the two integrals is finite.

**Definition 4.6.3** (Uncertain Ordinal Priority Approach (UOPA)). Let $I$ be a finite set of experts and $J$ a finite set of criteria, with $n := |J| \geq 2$. Each expert $i \in I$ provides:

1. an *ordinal ranking* of criteria, represented as a permutation $\pi_i : \{1, \ldots, n\} \to J$ (rank 1 = most important);

2. an *uncertain importance profile* along the ranks: for each rank $r \in \{1, \ldots, n\}$, an uncertain set (uncertain number)

$$\xi_{i,r} : \Gamma \to \mathcal{P}(\mathbb{R})$$

with finite expected value $a_{i,r} := \mathbb{E}[\xi_{i,r}] \in \mathbb{R}$.

**(1) Reduction to a deterministic OPA core via expected values.** Define deterministic coefficients

$$a_{i,r} := \mathbb{E}[\xi_{i,r}], \qquad i \in I, \ r = 1, \ldots, n.$$

Introduce decision variables $w_{i,r} \in \mathbb{R}_{\geq 0}$ and $z \in \mathbb{R}$, and solve the linear program

$$\max z \tag{4.6}$$
$$\text{s.t. } a_{i,r}\big(w_{i,r} - w_{i,r+1}\big) \ \geq \ z, \qquad \forall i \in I, \ \forall r = 1, \ldots, n-1, \tag{4.7}$$
$$a_{i,n} w_{i,n} \ \geq \ z, \qquad \forall i \in I, \tag{4.8}$$
$$\sum_{i \in I} \sum_{r=1}^n w_{i,r} = 1, \tag{4.9}$$
$$w_{i,r} \geq 0, \qquad \forall i \in I, \ \forall r. \tag{4.10}$$

**(2) Aggregated criterion weights.** For each criterion $j \in J$, define its aggregated (crisp) weight by

$$W_j := \sum_{i \in I} w_{i,r(i,j)}, \quad \text{where } r(i,j) \text{ is the unique rank with } \pi_i(r(i,j)) = j.$$

Then $W_j \geq 0$ and $\sum_{j \in J} W_j = 1$.

**(3) Output as uncertain sets (degenerate uncertain numbers).** Define, for each $j \in J$, a degenerate uncertain set (singleton-valued mapping)

$$\Xi_j : \Gamma \to \mathcal{P}(\mathbb{R}), \qquad \Xi_j(\gamma) := \{W_j\}.$$

The family $\{\Xi_j\}_{j \in J}$ is called the *UOPA uncertain weight vector*.

**Theorem 4.6.4** (Uncertain-set structure and well-definedness of UOPA). *Assume in Definition 4.6.3 that, for all $i \in I$ and $r \in \{1, \ldots, n\}$, the uncertain set $\xi_{i,r}$ has a finite expected value $a_{i,r} = \mathbb{E}[\xi_{i,r}]$ and that $a_{i,r} \geq 0$.*

*Then:*

1. *the coefficients $a_{i,r}$ in (4.6)–(4.10) are well-defined real numbers, so the optimization model is well-posed as a finite-dimensional linear program;*

2. *the feasible region of (4.6)–(4.10) is nonempty and compact, and the objective is bounded above; hence an optimal solution exists;*

3. *the UOPA output $\{\Xi_j\}_{j \in J}$ defined in Definition 4.6.3(3) forms a family of uncertain sets (on the same uncertainty space), i.e., the method yields an* uncertain-set structured *weight vector.*

*Proof.* **(1) Well-defined coefficients.** By assumption, each $\xi_{i,r}$ has finite expected value $\mathbb{E}[\xi_{i,r}]$ in the sense of Definition 4.6.2. Therefore $a_{i,r} \in \mathbb{R}$ is uniquely determined, and (4.6)–(4.10) is a deterministic linear program with finitely many variables and linear constraints.

**(2) Nonemptiness (feasibility).** Let $N := |I| \, n$. Set $w_{i,r} := 1/N$ for all $i, r$, and set $z := 0$. Then (4.9) and (4.10) hold. Moreover, since $a_{i,r} \geq 0$ and $w_{i,r} - w_{i,r+1} = 0$, the left-hand side of (4.7) equals 0, and the left-hand side of (4.8) equals $a_{i,n}/N \geq 0$. Hence all constraints hold with $z = 0$. So the feasible set is nonempty.

**(3) Compactness and boundedness.** From (4.9) and (4.10), the vector $(w_{i,r})$ lies in a standard simplex, hence the feasible set projected to $w$ is closed and bounded. The full feasible set in $(w, z)$ is also closed: it is an intersection of closed halfspaces.

Also, for every feasible $(w, z)$ and every $i \in I$, (4.8) implies $z \leq a_{i,n} w_{i,n} \leq a_{i,n} \cdot 1 = a_{i,n}$ because $0 \leq w_{i,n} \leq 1$ under (4.9)–(4.10). Hence $z$ is bounded above by $\max_{i \in I} a_{i,n} < \infty$. Therefore the linear objective $\max z$ is bounded above on the feasible region.

**(4) Existence of an optimizer.** A bounded linear functional attains its maximum on a nonempty compact polyhedron. Thus (4.6)–(4.10) has at least one optimal solution.

**(5) Uncertain-set structured output.** For each $j \in J$, the mapping $\Xi_j(\gamma) = \{W_j\}$ is a constant (singleton-valued) set-valued function, hence measurable with respect to $\mathcal{L}$, and therefore is an uncertain set in the sense of Definition 4.6.2. Consequently, $\{\Xi_j\}_{j \in J}$ is a family of uncertain sets representing the UOPA weight vector. $\square$

Related concepts of Ordinal Priority Approach (OPA) under uncertainty-aware models are listed in Table 4.7.

Related concepts other than Uncertain OPA include the interval ordinal priority approach [516, 522], the grey ordinal priority approach [523, 524], and the rough ordinal priority approach [525, 526].

Table 4.7: Related concepts of Ordinal Priority Approach (OPA) under uncertainty-aware models.

| $k$ | Related OPA concept(s) |
|---|---|
| 1 | Fuzzy Ordinal Priority Approach (OPA) |
| 2 | Intuitionistic Fuzzy Ordinal Priority Approach (OPA) |
| 3 | Neutrosophic Ordinal Priority Approach (OPA) [520, 521] |

## 4.7 Fuzzy PIPRECIA (Pivot Pairwise Relative Criteria Importance Assessment)

PIPRECIA is an MCDM weighting method using a pivot criterion and sequential pairwise comparisons to derive criteria importance coefficients efficiently [527, 528]. Fuzzy PIPRECIA extends PIPRECIA by expressing judgments with fuzzy numbers, handling vagueness and uncertainty while computing robust criteria weights consistently [44, 529].

**Definition 4.7.1** (TFN-based fuzzy PIPRECIA (ordinary + inverse)). [44, 529] Let $C_1, \ldots, C_n$ be criteria written in a fixed order (the order is *not* required to be sorted by importance). Let $R \geq 1$ decision makers (DMs) provide linguistic judgments encoded as *positive triangular fuzzy numbers* (TFNs), using two TFN-valued scales: a "> 1" scale (often called the 1–2 scale) and a "< 1" scale (often called the 0–1 scale). (Concrete TFN tables for these two scales are commonly given in the literature.)

Assume TFN arithmetic ($\otimes, \oslash$) and TFN powers are available (e.g., componentwise in the standard fuzzy-AHP convention).

**Step 1 (ordinary consecutive comparisons).** For each $j = 2, \ldots, n$ and each DM $r \in \{1, \ldots, R\}$, let $\bar{s}_j^{(r)} \in \mathsf{TFN}_{>0}$ encode the relative importance of $C_j$ versus $C_{j-1}$:

$$\bar{s}_j^{(r)} \begin{cases} > \tilde{1} & \text{if } C_j \text{ is judged more important than } C_{j-1}, \\ = \tilde{1} & \text{if } C_j \text{ is judged equally important as } C_{j-1}, \qquad \tilde{1} := (1,1,1), \\ < \tilde{1} & \text{if } C_j \text{ is judged less important than } C_{j-1}, \end{cases}$$

where $>/<$ are interpreted at the linguistic-scale level.

**Step 2 (aggregation by geometric mean).** Aggregate the DMs' TFNs by the (TFN) geometric mean:

$$\tilde{s}_j := \mathrm{GM}\big(\tilde{s}_j^{(1)}, \ldots, \tilde{s}_j^{(R)}\big) := \left( \bigotimes_{r=1}^{R} \tilde{s}_j^{(r)} \right)^{1/R}, \qquad j = 2, \ldots, n,$$

as commonly required in fuzzy PIPRECIA implementations.

**Step 3 (coefficients).** Define TFN coefficients $\tilde{k}_j$ by

$$\tilde{k}_1 := \tilde{1}, \qquad \tilde{k}_j := \tilde{2} \ominus \tilde{s}_j \ \ (j = 2, \ldots, n), \qquad \tilde{2} := (2,2,2),$$

i.e., if $\tilde{s}_j = (l_j, m_j, u_j)$ then $\tilde{k}_j := (2 - u_j, \ 2 - m_j, \ 2 - l_j)$ (to preserve the TFN order). This corresponds to the "subtract from 2" rule in fuzzy PIPRECIA.

**Step 4 (recursive fuzzy weights).** Define $\tilde{q}_j$ by

$$\tilde{q}_1 := \tilde{1}, \qquad \tilde{q}_j := \tilde{q}_{j-1} \oslash \tilde{k}_j \quad (j = 2, \ldots, n).$$

**Step 5 (relative weights from the ordinary pass).** Define the (ordinary-pass) fuzzy relative weights by

$$\tilde{w}_j := \tilde{q}_j \oslash \left( \bigoplus_{t=1}^{n} \tilde{q}_t \right), \qquad j = 1, \ldots, n.$$

**Step 6–9 (inverse pass).** Repeat Steps 1–5 *in reverse direction* by comparing $C_j$ with $C_{j+1}$ (starting from the penultimate criterion) to obtain inverse-pass TFNs $\tilde{w}'_1, \ldots, \tilde{w}'_n$.

**Step 10 (defuzzification and final criterion weights).** Let Defuzz : $\mathsf{TFN}_{>0} \to \mathbb{R}_{>0}$ be a fixed defuzzification map. A widely used choice (often reported as DFV) is the weighted mean

$$\mathrm{Defuzz}(l, m, u) := \frac{l + 4m + u}{6}.$$

Defuzzify and average the ordinary and inverse weights:

$$w_j := \mathrm{Defuzz}(\tilde{w}_j), \qquad w'_j := \mathrm{Defuzz}(\tilde{w}'_j), \qquad w_j^{\mathrm{final}} := \frac{w_j + w'_j}{2}, \qquad j = 1, \ldots, n.$$

This "defuzzify then average" rule is the standard finalization step in fuzzy PIPRECIA.

Finally, normalize

$$w_j^{\mathrm{final}} \leftarrow \frac{w_j^{\mathrm{final}}}{\sum_{t=1}^{n} w_t^{\mathrm{final}}} \quad \text{so that} \quad \sum_{j=1}^{n} w_j^{\mathrm{final}} = 1,$$

and use $(w_1^{\mathrm{final}}, \ldots, w_n^{\mathrm{final}})$ as the criterion-weight vector.

As an extension based on Uncertain Sets, the definition of Uncertain PIPRECIA (U-PIPRECIA) is given below.

**Definition 4.7.2** (Uncertain PIPRECIA (U-PIPRECIA): expected-value realization). Let $C_1, \ldots, C_n$ be criteria listed in a fixed order ($n \geq 2$), and let $\mathcal{D} = \{1, \ldots, R\}$ be a finite set of decision makers (DMs). Work on a fixed uncertainty space $(\Gamma, \mathcal{L}, \mathcal{M})$.

For each $j = 2, \ldots, n$ and each DM $r \in \mathcal{D}$, assume the DM provides an *uncertain relative-importance assessment*

$$\Xi_j^{(r)} : \Gamma \to \mathcal{P}(\mathbb{R}),$$

interpreted as the (uncertain) importance ratio of $C_j$ versus $C_{j-1}$. Assume:

$$\Xi_j^{(r)}(\gamma) \subseteq \mathbb{R}_{>0} \quad (\forall \gamma \in \Gamma), \qquad \mathbb{E}[\Xi_j^{(r)}] < \infty.$$

Define the corresponding deterministic coefficients by expected values:

$$s_j^{(r)} := \mathbb{E}[\Xi_j^{(r)}] \in \mathbb{R}_{>0}, \qquad j = 2, \dots, n, \ r \in \mathcal{D}.$$

**Step 1 (DM aggregation).** Aggregate DMs by the geometric mean (positive scalar aggregation):

$$s_j := \Big(\prod_{r=1}^{R} s_j^{(r)}\Big)^{1/R} \in \mathbb{R}_{>0}, \qquad j = 2, \dots, n,$$

and set $s_1 := 1$.

**Step 2 (PIPRECIA coefficients).** Define

$$k_1 := 1, \qquad k_j := 2 - s_j, \quad j = 2, \dots, n.$$

(These are the crisp counterparts of the standard "subtract from 2" rule.)

**Step 3 (recursive importance sequence).** Define

$$q_1 := 1, \qquad q_j := \frac{q_{j-1}}{k_j} \quad (j = 2, \dots, n),$$

whenever $k_j > 0$.

**Step 4 (normalized criterion weights).** Let $Q := \sum_{t=1}^{n} q_t$ (assume $Q > 0$). Define

$$w_j := \frac{q_j}{Q}, \qquad j = 1, \dots, n.$$

The vector $w = (w_1, \dots, w_n)$ is called the *U-PIPRECIA (ordinary-pass) weight vector*.

**Step 5 (optional inverse pass and finalization).** Repeat Steps 1–4 in reverse order by comparing $C_j$ with $C_{j+1}$ to obtain $w_1', \dots, w_n'$, and set

$$w_j^{\text{final}} := \frac{w_j + w_j'}{2}, \qquad w_j^{\text{final}} \leftarrow \frac{w_j^{\text{final}}}{\sum_{t=1}^{n} w_t^{\text{final}}}.$$

**Definition 4.7.3** (Uncertain-set output of U-PIPRECIA). Let $w = (w_1, \dots, w_n)$ be the (ordinary-pass or final) weight vector obtained from Definition 4.7.2. Define, for each $j = 1, \dots, n$, a degenerate uncertain set (singleton-valued mapping)

$$W_j : \Gamma \to \mathcal{P}(\mathbb{R}), \qquad W_j(\gamma) := \{w_j\}.$$

Then $W = (W_1, \dots, W_n)$ is called the *U-PIPRECIA uncertain weight vector*.

**Theorem 4.7.4** (Uncertain-set structure and well-definedness of U-PIPRECIA). *Assume the setting of Definition 4.7.2 and suppose:*

(A1) *(Finite positive expectations) For all $j = 2, \dots, n$ and $r \in \{1, \dots, R\}$, $s_j^{(r)} = \mathbb{E}[\Xi_j^{(r)}] \in (0, 2)$.*

*Then:*

(i) $s_j \in (0, 2)$ *and hence* $k_j = 2 - s_j \in (0, 2)$ *for all* $j \geq 2$;

(ii) *the recursion* $q_1 = 1$, $q_j = q_{j-1}/k_j$ *is well-defined and yields* $q_j > 0$ *for all* $j$;

(iii) $Q = \sum_{t=1}^{n} q_t$ *satisfies* $Q \in (0, \infty)$, *and therefore* $w_j = q_j/Q$ *is well-defined with* $w_j > 0$ *and* $\sum_{j=1}^{n} w_j = 1$;

(iv) *the output family* $(W_j)_{j=1}^{n}$ *in Definition 4.7.3 is a family of uncertain sets on* $(\Gamma, \mathcal{L}, \mathcal{M})(=$ *an uncertain-set structure). The same conclusions hold for the inverse-pass and the final averaged weights whenever (A1) holds for the inverse comparisons as well.*

*Proof.* **(i)** Fix $j \in \{2, \ldots, n\}$. By (A1), each $s_j^{(r)} \in (0, 2)$, hence $\prod_{r=1}^{R} s_j^{(r)} \in (0, 2^R)$ and therefore

$$s_j = \Big(\prod_{r=1}^{R} s_j^{(r)}\Big)^{1/R} \in (0, 2).$$

Consequently $k_j = 2 - s_j \in (0, 2)$. Also $k_1 = 1$ by definition.

**(ii)** Since $k_j > 0$ for all $j \geq 2$, division by $k_j$ is valid. Inductively, $q_1 = 1 > 0$ and if $q_{j-1} > 0$ then $q_j = q_{j-1}/k_j > 0$. Thus all $q_j$ are well-defined and positive.

**(iii)** Because each $q_t > 0$ and $n < \infty$, the sum $Q = \sum_{t=1}^{n} q_t$ is finite and strictly positive. Hence each $w_j = q_j/Q$ is well-defined and positive. Moreover,

$$\sum_{j=1}^{n} w_j = \sum_{j=1}^{n} \frac{q_j}{Q} = \frac{1}{Q} \sum_{j=1}^{n} q_j = \frac{Q}{Q} = 1.$$

**(iv)** For each fixed $j$, the mapping $W_j(\gamma) = \{w_j\}$ is constant (singleton-valued), hence measurable with respect to $\mathcal{L}$ and therefore qualifies as an uncertain set. Thus $(W_j)_{j=1}^{n}$ is an uncertain-set structured output. The inverse pass is the same construction applied to the reverse ordering, so the same argument applies, and the final averaging/renormalization is an ordinary algebraic post-processing that preserves well-definedness whenever both passes are well-defined. $\square$

For reference, related concepts of PIPRECIA under uncertainty-aware models are listed in Table 4.8.

As related PIPRECIA-based methods other than Uncertain PIPRECIA, PIPRECIA-S [536, 537], Grey PIPRECIA [538, 539], and Rough PIPRECIA [540, 541] are also known.

Table 4.8: Related concepts of PIPRECIA under uncertainty-aware models.

| $k$ | Related PIPRECIA concept(s) |
|---|---|
| 2 | Intuitionistic Fuzzy PIPRECIA [530] |
| 2 | Fermatean Fuzzy PIPRECIA [531] |
| 3 | Picture Fuzzy PIPRECIA [532] |
| 3 | Hesitant Fuzzy PIPRECIA [533] |
| 3 | Neutrosophic PIPRECIA [534, 535] |

## 4.8 Fuzzy SWARA (Fuzzy Stepwise Weight Assessment Ratio Analysis)

SWARA ranks criteria by importance, elicits stepwise comparative coefficients between successive criteria, computes recalculated weights sequentially, and normalizes them for use in scoring models later [542, 543]. Fuzzy SWARA uses linguistic terms for stepwise importance ratios, converts them to fuzzy numbers, propagates fuzziness through weight calculation, and defuzzifies final weights for ranking [544–546].

**Definition 4.8.1** (Fuzzy SWARA (TFN-based step-wise weighting)). [544–546] Let $\mathcal{C} = \{C_1, \ldots, C_n\}$ be a finite set of criteria. Assume that the decision makers (DMs) provide a *consensus ordering*

$$C_1 \succ C_2 \succ \cdots \succ C_n$$

(from the most important to the least important criterion).

**(0) Triangular fuzzy numbers and arithmetic.** Let

$$\mathsf{TFN}_{>0} := \{(l, m, u) \in \mathbb{R}^3 : \ 0 < l \le m \le u\}$$

be the set of positive triangular fuzzy numbers (TFNs). For $\tilde{x} = (l_x, m_x, u_x)$ and $\tilde{y} = (l_y, m_y, u_y)$ in $\mathsf{TFN}_{>0}$, define

$$\tilde{x} \oplus \tilde{y} := (l_x + l_y, \ m_x + m_y, \ u_x + u_y), \qquad \tilde{y}^{-1} := \left(\frac{1}{u_y}, \frac{1}{m_y}, \frac{1}{l_y}\right),$$

$$\tilde{x} \oslash \tilde{y} := \tilde{x} \otimes \tilde{y}^{-1} = \left(\frac{l_x}{u_y}, \ \frac{m_x}{m_y}, \ \frac{u_x}{l_y}\right), \qquad \tilde{1} := (1, 1, 1).$$

(Thus $\oslash$ is well-defined on $\mathsf{TFN}_{>0}$.)

**(1) Fuzzy stepwise comparative importance.** For each $j = 2, \ldots, n$, let

$$\tilde{b}_j = (l_j, m_j, u_j) \in \mathsf{TFN}_{>0}$$

be the *fuzzy comparative importance* of $C_{j-1}$ relative to $C_j$ (elicited from linguistic judgments encoded as TFNs). No $\tilde{b}_1$ is needed.

**(2) Fuzzy coefficients.** Define TFN coefficients $\tilde{e}_j$ by

$$\tilde{e}_1 := \tilde{1}, \qquad \tilde{e}_j := \tilde{b}_j \oplus \tilde{1} \quad (j = 2, \ldots, n).$$

**(3) Recalculated fuzzy weights (unnormalized).** Define $\tilde{f}_j \in \mathsf{TFN}_{>0}$ recursively by

$$\tilde{f}_1 := \tilde{1}, \qquad \tilde{f}_j := \tilde{f}_{j-1} \oslash \tilde{e}_j \quad (j = 2, \ldots, n).$$

**(4) Normalized fuzzy weights.** Let $\tilde{F} := \bigoplus_{t=1}^{n} \tilde{f}_t$. The *fuzzy SWARA weight* of $C_j$ is

$$\tilde{w}_j := \tilde{f}_j \oslash \tilde{F} = (w_j^\ell, w_j^m, w_j^u) \in \mathsf{TFN}_{>0}, \qquad j = 1, \ldots, n.$$

**(5) Defuzzification (crisp weights).** A commonly used crisp weight is obtained by centroid (center-of-area) defuzzification:

$$w_j := \frac{w_j^\ell + w_j^m + w_j^u}{3}, \qquad j = 1, \ldots, n.$$

Optionally, normalize $w$ again by $w_j \leftarrow w_j / \sum_{t=1}^{n} w_t$ so that $\sum_{j=1}^{n} w_j = 1$.

The resulting vector $(w_1, \ldots, w_n)$ is called the *Fuzzy SWARA (defuzzified) criterion-weight vector*.

Using Uncertain Sets, we define Uncertain SWARA (U-SWARA) as follows.

**Definition 4.8.2** (Uncertain SWARA (U-SWARA): expected-value realization). Let $\mathcal{C} = \{C_1, \ldots, C_n\}$ be a finite set of criteria, $n \geq 2$. Assume a fixed *consensus importance order*

$$C_1 \succ C_2 \succ \cdots \succ C_n$$

(from most to least important). Work on a fixed uncertainty space $(\Gamma, \mathcal{L}, \mathcal{M})$.

For each step $j = 2, \ldots, n$, decision makers provide an *uncertain stepwise comparative-importance coefficient* (the "relative importance of $C_{j-1}$ over $C_j$")

$$\Xi_j : \Gamma \to \mathcal{P}(\mathbb{R}),$$

assumed to satisfy

$$\Xi_j(\gamma) \subseteq \mathbb{R}_{\geq 0} \quad (\forall \gamma \in \Gamma), \qquad s_j := \mathbb{E}[\Xi_j] < \infty.$$

(Thus $s_j$ is the crisp stepwise coefficient obtained by expected-value reduction of uncertainty.)

Define the *SWARA recalculation factors* by

$$k_1 := 1, \qquad k_j := 1 + s_j \quad (j = 2, \ldots, n).$$

Define the *unnormalized sequential weights* recursively by

$$q_1 := 1, \qquad q_j := \frac{q_{j-1}}{k_j} \quad (j = 2, \ldots, n),$$

whenever $k_j > 0$.

Let $Q := \sum_{t=1}^{n} q_t$ (assume $Q > 0$), and define the *normalized SWARA weights* by

$$w_j := \frac{q_j}{Q}, \qquad j = 1, \ldots, n.$$

The vector $w = (w_1, \ldots, w_n)$ is called the *U-SWARA criterion-weight vector*.

**Definition 4.8.3** (Uncertain-set output of U-SWARA). Let $w = (w_1, \ldots, w_n)$ be obtained from Definition 4.8.2. Define degenerate uncertain sets (singleton-valued mappings)

$$W_j : \Gamma \to \mathcal{P}(\mathbb{R}), \qquad W_j(\gamma) := \{w_j\}, \qquad j = 1, \ldots, n.$$

Then $W = (W_1, \ldots, W_n)$ is called the *U-SWARA uncertain weight vector*.

**Theorem 4.8.4** (Uncertain-set structure and well-definedness of U-SWARA). *Assume the setting of Definition 4.8.2 and suppose:*

(A1)  *(Nonnegative finite expectations)* $s_j = \mathbb{E}[\Xi_j] \in [0, \infty)$ *for all* $j = 2, \ldots, n$.

*Then:*

(i)  $k_j = 1 + s_j \in [1, \infty)$ *for all* $j = 2, \ldots, n$, *hence* $k_j > 0$;

(ii)  *the recursion* $q_1 = 1$, $q_j = q_{j-1}/k_j$ *is well-defined and yields* $q_j > 0$ *for all* $j$;

(iii)  $Q = \sum_{t=1}^{n} q_t$ *satisfies* $Q \in (0, \infty)$, *and thus* $w_j = q_j/Q$ *is well-defined with* $w_j > 0$ *and* $\sum_{j=1}^{n} w_j = 1$;

(iv)  *the family* $(W_j)_{j=1}^{n}$ *in Definition 4.8.3 forms an uncertain-set structured output on* $(\Gamma, \mathcal{L}, \mathcal{M})$.

*Proof.* **(i)** By (A1), for each $j \geq 2$ one has $s_j \geq 0$, hence $k_j = 1 + s_j \geq 1 > 0$.

**(ii)** Since $k_j > 0$, division by $k_j$ is valid. Inductively $q_1 = 1 > 0$; if $q_{j-1} > 0$ then $q_j = q_{j-1}/k_j > 0$. Thus all $q_j$ exist and are positive.

**(iii)** Because $n < \infty$ and each $q_t > 0$, the sum $Q = \sum_{t=1}^{n} q_t$ is finite and strictly positive. Hence $w_j = q_j/Q$ is well-defined and positive. Moreover,

$$\sum_{j=1}^{n} w_j = \sum_{j=1}^{n} \frac{q_j}{Q} = \frac{1}{Q} \sum_{j=1}^{n} q_j = \frac{Q}{Q} = 1.$$

**(iv)** For each $j$, the map $W_j(\gamma) = \{w_j\}$ is constant (singleton-valued), hence measurable; therefore it is an uncertain set. Consequently $(W_j)_{j=1}^{n}$ is an uncertain-set structured output.   $\square$

Related uncertainty-model variants of SWARA, classified by the degree-domain dimension $k$, are listed in Table 4.9.

As related concepts other than Uncertain SWARA, Rough SWARA [561], Soft SWARA [562], SWARA-TOPSIS [563, 564], SWARA-AHP [565, 566], and Grey SWARA [567, 568] are also known.

Table 4.9: Related uncertainty-model variants of SWARA (classified by the degree-domain dimension $k$).

| $k$ | Related SWARA variant(s) |
|---|---|
| 1 | Fuzzy SWARA |
| 2 | Intuitionistic Fuzzy SWARA [547,548] |
| 2 | Pythagorean Fuzzy SWARA [549,550] |
| 2 | Fermatean Fuzzy SWARA [551,552] |
| 3 | Hesitant Fuzzy SWARA [553,554] |
| 3 | Spherical Fuzzy SWARA [555,556] |
| 3 | Neutrosophic SWARA [557,558] |
| n | Plithogenic SWARA [559,560] |

## 4.9 Fuzzy CILOS (Fuzzy Criterion Impact Loss)

CILOS derives objective criterion weights by quantifying information or impact loss when each criterion deteriorates, emphasizing criteria that most affect overall performance [569,570]. Fuzzy CILOS extends CILOS using fuzzy numbers for performances, computes fuzzy impact-loss measures per criterion, and produces objective weights robust under imprecision [571].

**Definition 4.9.1** (Fuzzy CILOS (Criterion Impact Loss) weighting). [571] Let

$$\mathcal{A} = \{A_1, \ldots, A_m\} \quad \text{and} \quad \mathcal{C} = \{C_1, \ldots, C_n\}$$

be the sets of alternatives and criteria, respectively. Assume that each criterion $C_j$ is either of *benefit type* or *cost type*.

Let $\mathcal{FN}_{>0}$ be a fixed class of positive fuzzy numbers, and let

$$\widetilde{X} = (\tilde{x}_{ij})_{m \times n} \in \mathcal{FN}_{>0}^{m \times n}$$

be the fuzzy decision matrix, where $\tilde{x}_{ij}$ is the fuzzy performance of alternative $A_i$ under criterion $C_j$.

Fix a total ranking / defuzzification map

$$\text{Score} : \mathcal{FN}_{>0} \to \mathbb{R}_{>0}.$$

Define the associated positive crisp score matrix

$$S = (s_{ij})_{m \times n}, \qquad s_{ij} := \text{Score}(\tilde{x}_{ij}) > 0.$$

**Step 1: Benefit-type transformation.** Define the transformed matrix $X = (x_{ij})_{m \times n} \in \mathbb{R}_{>0}^{m \times n}$ by

$$x_{ij} := \begin{cases} s_{ij}, & \text{if } C_j \text{ is a benefit criterion,} \\ \dfrac{\min_{1 \le r \le m} s_{rj}}{s_{ij}}, & \text{if } C_j \text{ is a cost criterion.} \end{cases}$$

Thus, after transformation, every criterion is treated as a benefit criterion.

**Step 2: Criterion-wise optima.** For each criterion $j \in \{1, \ldots, n\}$, define

$$x_j^\star := \max_{1 \le i \le m} x_{ij}.$$

Choose an index

$$\iota_j \in \arg \max_{1 \le i \le m} x_{ij}.$$

**Step 3: Square matrix of criterion-optimal alternatives.** Define the matrix $A = (a_{ij})_{n \times n}$ by

$$a_{ij} := x_{\iota_i j}, \qquad i, j = 1, \dots, n.$$

Hence, the $i$-th row of $A$ is the transformed performance vector of an alternative that is optimal for criterion $C_i$. In particular,

$$a_{ii} = x_i^\star.$$

**Step 4: Relative-loss matrix.** Define the matrix $P = (p_{ij})_{n \times n}$ by

$$p_{ij} := \begin{cases} \dfrac{x_j^\star - a_{ij}}{x_j^\star}, & i \ne j, \\ 0, & i = j. \end{cases}$$

Thus $p_{ij} \in [0, 1]$ represents the relative loss of criterion $C_j$ when the alternative that is best for criterion $C_i$ is selected.

**Step 5: CILOS balance matrix.** Define $F = (f_{ij})_{n \times n}$ by

$$f_{ij} := \begin{cases} p_{ij}, & i \ne j, \\ -\displaystyle\sum_{\substack{r=1 \\ r \ne j}}^{n} p_{rj}, & i = j. \end{cases}$$

Equivalently,

$$F = \begin{pmatrix} -\sum_{r \ne 1} p_{r1} & p_{12} & \cdots & p_{1n} \\ p_{21} & -\sum_{r \ne 2} p_{r2} & \cdots & p_{2n} \\ \vdots & \vdots & \ddots & \vdots \\ p_{n1} & p_{n2} & \cdots & -\sum_{r \ne n} p_{rn} \end{pmatrix}.$$

**Step 6: Criterion-significance vector and weights.** A nonzero vector

$$q = (q_1, \dots, q_n)^\top \in \mathbb{R}_{\ge 0}^n \setminus \{0\}$$

satisfying

$$Fq = 0$$

is called a *criterion-significance vector* of the fuzzy CILOS model. The corresponding *Fuzzy CILOS weight vector* is

$$w = (w_1, \dots, w_n)^\top, \qquad w_j := \frac{q_j}{\sum_{\ell=1}^{n} q_\ell}, \quad j = 1, \dots, n.$$

Then

$$w_j \ge 0 \qquad \text{and} \qquad \sum_{j=1}^{n} w_j = 1.$$

We now define *Uncertain CILOS* (UCILOS) by extending the criterion-impact-loss idea to a general uncertain model $M$.

**Definition 4.9.2** (Uncertain CILOS (UCILOS) of type $M$)**.** Let

$$\mathcal{A} = \{A_1, \ldots, A_m\} \quad \text{and} \quad \mathcal{C} = \{C_1, \ldots, C_n\}$$

be the sets of alternatives and criteria, respectively, where $m, n \in \mathbb{N}$ and $m, n \geq 1$.

Fix an uncertain model $M$ with degree-domain

$$\mathrm{Dom}(M) \subseteq [0,1]^k$$

for some integer $k \geq 1$.

Assume that each criterion $C_j$ is designated either as a *benefit criterion* or as a *cost criterion*. Let

$$X_M = (\mu_{ij})_{m \times n} \in \mathrm{Dom}(M)^{m \times n}$$

be an uncertain decision matrix, where $\mu_{ij} \in \mathrm{Dom}(M)$ is the uncertain performance of alternative $A_i$ under criterion $C_j$.

Assume further that a total score map

$$\mathrm{Score}_M : \mathrm{Dom}(M) \longrightarrow \mathbb{R}_{>0}$$

is fixed. Define the associated positive score matrix

$$S = (s_{ij})_{m \times n} \in \mathbb{R}_{>0}^{m \times n}, \qquad s_{ij} := \mathrm{Score}_M(\mu_{ij}).$$

**Step 1: Benefit-type transformation.** Define $B = (b_{ij})_{m \times n} \in \mathbb{R}_{>0}^{m \times n}$ by

$$b_{ij} := \begin{cases} s_{ij}, & \text{if } C_j \text{ is a benefit criterion,} \\[2mm] \dfrac{\min_{1 \leq r \leq m} s_{rj}}{s_{ij}}, & \text{if } C_j \text{ is a cost criterion.} \end{cases}$$

Thus every criterion is transformed into benefit form.

**Step 2: Column normalization.** For each criterion $j$, let

$$\beta_j := \sum_{r=1}^{m} b_{rj}.$$

Define the normalized matrix

$$Z = (z_{ij})_{m \times n} \in \mathbb{R}_{>0}^{m \times n}$$

by

$$z_{ij} := \frac{b_{ij}}{\beta_j}.$$

**Step 3: Criterion-wise optima.** For each $j \in \{1, \ldots, n\}$, define

$$z_j^\star := \max_{1 \leq i \leq m} z_{ij},$$

and choose any index

$$\iota_j \in \arg \max_{1 \leq i \leq m} z_{ij}.$$

**Step 4: Square matrix of criterion-optimal alternatives.** Define the square matrix

$$A = (a_{ij})_{n \times n}$$

by

$$a_{ij} := z_{\iota_i j}, \qquad i, j = 1, \ldots, n.$$

Hence the $i$-th row of $A$ is the normalized criterion profile of an alternative that is optimal for criterion $C_i$. In particular,

$$a_{ii} = z_i^\star.$$

**Step 5: Relative-loss matrix.** Define $P = (p_{ij})_{n \times n}$ by

$$p_{ij} := \begin{cases} 0, & i = j, \\ \dfrac{z_j^\star - a_{ij}}{z_j^\star}, & i \neq j. \end{cases}$$

Thus $p_{ij}$ is the relative loss of criterion $C_j$ when the alternative that is optimal for criterion $C_i$ is selected.

**Step 6: Total loss and criterion significance.** For each criterion $j$, define its total impact loss by

$$L_j := \sum_{\substack{i=1 \\ i \neq j}}^{n} p_{ij}.$$

Define the corresponding criterion-significance score by

$$q_j := \frac{1}{1 + L_j}, \qquad j = 1, \ldots, n.$$

**Step 7: UCILOS weight vector.** The *Uncertain CILOS weight vector* is

$$w = (w_1, \ldots, w_n)^\top \in [0,1]^n, \qquad w_j := \frac{q_j}{\sum_{\ell=1}^{n} q_\ell}, \quad j = 1, \ldots, n.$$

**Remark 4.9.3.** Definition 4.10.2 is an objective weighting construction. It generalizes the criterion-impact-loss idea by first converting each uncertain entry $\mu_{ij} \in \mathrm{Dom}(M)$ into a positive representative score and then applying a loss-based criterion-weighting mechanism to the induced score matrix.

**Theorem 4.9.4** (Well-definedness of UCILOS). *Let*

$$\mathfrak{U}_{\mathrm{CILOS}} = \big(\mathcal{A}, \mathcal{C}, M, X_M, \mathrm{Score}_M\big)$$

*be a UCILOS instance as in Definition 4.10.2. Assume:*

*(A1) $\mathcal{A}$ and $\mathcal{C}$ are finite nonempty sets;*

*(A2)*

$$X_M = (\mu_{ij})_{m \times n} \in \mathrm{Dom}(M)^{m \times n};$$

*(A3) each criterion $C_j$ is designated either as benefit type or as cost type;*

*(A4)*

$$\mathrm{Score}_M : \mathrm{Dom}(M) \to \mathbb{R}_{>0}$$

*is a total map.*

*Then all quantities appearing in Definition 4.10.2 are well-defined. More precisely:*

*(i) the score matrix $S = (s_{ij}) \in \mathbb{R}_{>0}^{m \times n}$ is well-defined;*

*(ii) the benefit-type matrix $B = (b_{ij}) \in \mathbb{R}_{>0}^{m \times n}$ is well-defined;*

*(iii) the normalized matrix $Z = (z_{ij}) \in \mathbb{R}_{>0}^{m \times n}$ is well-defined, and for each $j$,*

$$\sum_{i=1}^{m} z_{ij} = 1;$$

*(iv) for each $j$, the optimal value $z_j^\star$ and at least one maximizer $\iota_j \in \arg\max_i z_{ij}$ exist;*

*(v) the square matrix $A = (a_{ij})$ and the relative-loss matrix $P = (p_{ij})$ are well-defined, with*

$$0 \le p_{ij} \le 1 \qquad (1 \le i, j \le n);$$

*(vi) the total losses $L_j$ and significance scores $q_j$ are well-defined and satisfy*

$$L_j \in [0, n-1], \qquad q_j \in \left[\frac{1}{n}, 1\right];$$

*(vii) the weight vector $w = (w_1, \ldots, w_n)^\top$ is well-defined and satisfies*

$$w_j > 0 \qquad \text{for all } j, \qquad \text{and} \qquad \sum_{j=1}^{n} w_j = 1.$$

*Hence Uncertain CILOS of type $M$ is well-defined.*

*Proof.* By (A2), for every $i \in \{1, \ldots, m\}$ and $j \in \{1, \ldots, n\}$,

$$\mu_{ij} \in \mathrm{Dom}(M).$$

Since $\mathrm{Score}_M$ is total by (A4), each value

$$s_{ij} := \mathrm{Score}_M(\mu_{ij})$$

is well-defined and belongs to $\mathbb{R}_{>0}$. Therefore the score matrix

$$S = (s_{ij})_{m \times n} \in \mathbb{R}_{>0}^{m \times n}$$

is well-defined. This proves (i).

Next, define $B = (b_{ij})$ as in Definition 4.10.2. If $C_j$ is a benefit criterion, then

$$b_{ij} = s_{ij} > 0.$$

If $C_j$ is a cost criterion, then

$$b_{ij} = \frac{\min_{1 \le r \le m} s_{rj}}{s_{ij}}.$$

Because the finite set $\{s_{1j}, \ldots, s_{mj}\} \subset \mathbb{R}_{>0}$ has a positive minimum and $s_{ij} > 0$, this quotient is well-defined and strictly positive. Hence every $b_{ij} \in \mathbb{R}_{>0}$, so $B \in \mathbb{R}_{>0}^{m \times n}$. This proves (ii).

For each $j$, define

$$\beta_j := \sum_{r=1}^{m} b_{rj}.$$

Since each $b_{rj} > 0$ and $m \ge 1$, we have $\beta_j > 0$. Therefore

$$z_{ij} := \frac{b_{ij}}{\beta_j}$$

is well-defined and positive. Thus $Z = (z_{ij}) \in \mathbb{R}_{>0}^{m \times n}$ is well-defined. Moreover,

$$\sum_{i=1}^{m} z_{ij} = \sum_{i=1}^{m} \frac{b_{ij}}{\beta_j} = \frac{1}{\beta_j} \sum_{i=1}^{m} b_{ij} = 1.$$

This proves (iii).

Fix $j \in \{1, \ldots, n\}$. Since the finite nonempty set $\{z_{1j}, \ldots, z_{mj}\} \subset \mathbb{R}_{>0}$ has a maximum, the number

$$z_j^\star := \max_{1 \le i \le m} z_{ij}$$

exists, and the argmax set

$$\arg \max_{1 \le i \le m} z_{ij}$$

is nonempty. Hence at least one index $\iota_j$ can be chosen. This proves (iv).

Now define

$$a_{ij} := z_{\iota_i j}.$$

Since each $\iota_i \in \{1, \ldots, m\}$, every $a_{ij}$ is well-defined and belongs to $\mathbb{R}_{>0}$. Therefore $A = (a_{ij})$ is well-defined.

For the relative-loss matrix, first note that $z_j^\star > 0$. Also, since $z_j^\star$ is the maximum in column $j$,

$$0 < a_{ij} \le z_j^\star.$$

Hence, for $i \neq j$,

$$p_{ij} = \frac{z_j^\star - a_{ij}}{z_j^\star}$$

is well-defined and satisfies $0 \leq p_{ij} \leq 1$. By definition $p_{jj} = 0$. Therefore $P = (p_{ij})$ is well-defined and all its entries lie in $[0, 1]$. This proves (v).

For each $j$,

$$L_j := \sum_{\substack{i=1 \\ i \neq j}}^{n} p_{ij}$$

is a finite sum of $n - 1$ numbers from $[0, 1]$. Therefore

$$0 \leq L_j \leq n - 1.$$

Consequently,

$$q_j := \frac{1}{1 + L_j}$$

is well-defined and satisfies

$$\frac{1}{n} \leq q_j \leq 1.$$

This proves (vi).

Finally, because each $q_j > 0$, the denominator

$$\sum_{\ell=1}^{n} q_\ell$$

is strictly positive. Hence

$$w_j := \frac{q_j}{\sum_{\ell=1}^{n} q_\ell}$$

is well-defined for every $j$, and $w_j > 0$. Moreover,

$$\sum_{j=1}^{n} w_j = \sum_{j=1}^{n} \frac{q_j}{\sum_{\ell=1}^{n} q_\ell} = \frac{\sum_{j=1}^{n} q_j}{\sum_{\ell=1}^{n} q_\ell} = 1.$$

Thus $w$ is a well-defined normalized weight vector. This proves (vii).

Therefore every step of the UCILOS construction is mathematically well-defined. □

Related concepts of CILOS under uncertainty-aware models are listed in Table 4.10.

Table 4.10: Related concepts of CILOS under uncertainty-aware models.

| $k$ | Related CILOS concept(s) |
|---|---|
| 1 | Fuzzy CILOS |
| 2 | Intuitionistic Fuzzy CILOS |
| 3 | Neutrosophic CILOS |

## 4.10  Fuzzy IDOCRIW (Fuzzy Entropy-CILOS integrated objective weighting)

IDOCRIW objectively derives criterion weights by integrating entropy dispersion and CILOS impact-loss indices from a normalized decision matrix data alone [572,573]. Fuzzy IDOCRIW extends IDOCRIW using fuzzy numbers for performances, computing entropy and loss measures fuzzily, then defuzzifying weights robustly overall [48,574].

**Definition 4.10.1** (Fuzzy IDOCRIW weights (Entropy–CILOS integrated objective weighting)).  [48,574] Let $\mathcal{A} = \{A_1, \ldots, A_m\}$ be a finite set of alternatives and $\mathcal{C} = \{C_1, \ldots, C_n\}$ a finite set of criteria. Let $\widetilde{X} = (\widetilde{x}_{ij}) \in \mathbb{F}^{m \times n}$ be a fuzzy decision matrix, where $\mathbb{F}$ denotes the chosen class of fuzzy evaluations (e.g., TFNs, IVIFNs, picture-fuzzy numbers, etc.).

Assume a *score/defuzzification* map

$$s : \mathbb{F} \to \mathbb{R}_{>0}$$

that induces comparisons and yields strictly positive scalars

$$d_{ij} := s(\widetilde{x}_{ij}) \quad (i = 1, \ldots, m, \ j = 1, \ldots, n).$$

Partition the criteria into benefit and cost types: $\mathcal{C} = B \sqcup C$, where $B$ are beneficial and $C$ are non-beneficial criteria. Form a benefit-oriented (positive) matrix $\widehat{D} = (\widehat{d}_{ij})$ by

$$\widehat{d}_{ij} := \begin{cases} d_{ij}, & j \in B, \\ \dfrac{\min_{1 \le i \le m} d_{ij}}{d_{ij}}, & j \in C, \end{cases} \quad \text{for all } i, j.$$

(Thus all criteria are converted to the "larger-is-better" direction.)

**(1) Entropy objective weights.** Define the column-normalized proportions

$$p_{ij} := \frac{\widehat{d}_{ij}}{\sum_{r=1}^{m} \widehat{d}_{rj}}, \qquad i = 1, \ldots, m, \ j = 1, \ldots, n,$$

and adopt the convention $0 \ln 0 := 0$. Let $k := 1/\ln(m)$. The entropy of criterion $j$ is

$$E_j := -k \sum_{i=1}^{m} p_{ij} \ln(p_{ij}),$$

its divergence is $\Delta_j := 1 - E_j$, and the entropy weight vector is

$$w_j^{(E)} := \frac{\Delta_j}{\sum_{\ell=1}^{n} \Delta_\ell}, \qquad j = 1, \ldots, n.$$

**(2) CILOS objective weights.** For each criterion $j$, pick an index

$$s_j \in \arg \max_{1 \le i \le m} \widehat{d}_{ij}$$

(one of the best-performing alternatives under criterion $j$). Construct the square matrix $B = (b_{ij}) \in \mathbb{R}_{>0}^{n \times n}$ by

$$b_{ij} := \widehat{d}_{s_i j} \qquad (i, j = 1, \ldots, n),$$

so that $b_{ii} = \max_{1 \leq r \leq m} \widehat{d}_{ri}$. Define the relative impact-loss matrix $R = (r_{ij}) \in [0,1]^{n \times n}$ by

$$r_{ij} := \frac{b_{jj} - b_{ij}}{b_{jj}}, \qquad r_{jj} := 0 \quad (i, j = 1, \ldots, n).$$

Form the (column-Laplacian) weight-system matrix $F = (f_{ij}) \in \mathbb{R}^{n \times n}$:

$$f_{ij} := \begin{cases} r_{ij}, & i \neq j, \\ -\displaystyle\sum_{r=1}^{n} r_{rj}, & i = j. \end{cases}$$

A *CILOS weight vector* is any vector $w^{(C)} \in \mathbb{R}_{\geq 0}^n \setminus \{0\}$ satisfying

$$F \, w^{(C)} = 0, \qquad \sum_{j=1}^{n} w_j^{(C)} = 1.$$

(Under standard irreducibility/nondegeneracy conditions on $R$, such a normalized nonnegative null-vector is unique.)

**(3) IDOCRIW aggregation (integrated objective weights).** The *fuzzy IDOCRIW* weight vector is defined by the multiplicative fusion

$$w_j^{(\mathrm{IDOCRIW})} := \frac{w_j^{(E)} w_j^{(C)}}{\sum_{\ell=1}^{n} w_\ell^{(E)} w_\ell^{(C)}}, \qquad j = 1, \ldots, n.$$

Then $w^{(\mathrm{IDOCRIW})} \in \mathbb{R}_{\geq 0}^n$ and $\sum_{j=1}^{n} w_j^{(\mathrm{IDOCRIW})} = 1$.

We now define *Uncertain CILOS* (UCILOS) by extending the criterion-impact-loss idea to a general uncertain model $M$.

**Definition 4.10.2** (Uncertain CILOS (UCILOS) of type $M$)**.** Let

$$\mathcal{A} = \{A_1, \ldots, A_m\} \quad \text{and} \quad \mathcal{C} = \{C_1, \ldots, C_n\}$$

be the sets of alternatives and criteria, respectively, where $m, n \in \mathbb{N}$ and $m, n \geq 1$.

Fix an uncertain model $M$ with degree-domain

$$\mathrm{Dom}(M) \subseteq [0,1]^k$$

for some integer $k \geq 1$.

Assume that each criterion $C_j$ is designated either as a *benefit criterion* or as a *cost criterion*. Let

$$X_M = (\mu_{ij})_{m \times n} \in \mathrm{Dom}(M)^{m \times n}$$

be an uncertain decision matrix, where $\mu_{ij} \in \mathrm{Dom}(M)$ is the uncertain performance of alternative $A_i$ under criterion $C_j$.

Assume further that a total score map

$$\text{Score}_M : \text{Dom}(M) \longrightarrow \mathbb{R}_{>0}$$

is fixed. Define the associated positive score matrix

$$S = (s_{ij})_{m \times n} \in \mathbb{R}_{>0}^{m \times n}, \qquad s_{ij} := \text{Score}_M(\mu_{ij}).$$

**Step 1: Benefit-type transformation.** Define $B = (b_{ij})_{m \times n} \in \mathbb{R}_{>0}^{m \times n}$ by

$$b_{ij} := \begin{cases} s_{ij}, & \text{if } C_j \text{ is a benefit criterion,} \\ \dfrac{\min_{1 \le r \le m} s_{rj}}{s_{ij}}, & \text{if } C_j \text{ is a cost criterion.} \end{cases}$$

Thus every criterion is transformed into benefit form.

**Step 2: Column normalization.** For each criterion $j$, let

$$\beta_j := \sum_{r=1}^{m} b_{rj}.$$

Define the normalized matrix

$$Z = (z_{ij})_{m \times n} \in \mathbb{R}_{>0}^{m \times n}$$

by

$$z_{ij} := \frac{b_{ij}}{\beta_j}.$$

**Step 3: Criterion-wise optima.** For each $j \in \{1, \dots, n\}$, define

$$z_j^\star := \max_{1 \le i \le m} z_{ij},$$

and choose any index

$$\iota_j \in \arg \max_{1 \le i \le m} z_{ij}.$$

**Step 4: Square matrix of criterion-optimal alternatives.** Define the square matrix

$$A = (a_{ij})_{n \times n}$$

by

$$a_{ij} := z_{\iota_i, j}, \qquad i, j = 1, \dots, n.$$

Hence the $i$-th row of $A$ is the normalized criterion profile of an alternative that is optimal for criterion $C_i$. In particular,

$$a_{ii} = z_i^\star.$$

**Step 5: Relative-loss matrix.** Define $P = (p_{ij})_{n \times n}$ by

$$p_{ij} := \begin{cases} 0, & i = j, \\ \dfrac{z_j^\star - a_{ij}}{z_j^\star}, & i \neq j. \end{cases}$$

Thus $p_{ij}$ is the relative loss of criterion $C_j$ when the alternative that is optimal for criterion $C_i$ is selected.

**Step 6: Total loss and criterion significance.** For each criterion $j$, define its total impact loss by

$$L_j := \sum_{\substack{i=1 \\ i \neq j}}^{n} p_{ij}.$$

Define the corresponding criterion-significance score by

$$q_j := \frac{1}{1 + L_j}, \qquad j = 1, \ldots, n.$$

**Step 7: UCILOS weight vector.** The *Uncertain CILOS weight vector* is

$$w = (w_1, \ldots, w_n)^\top \in [0,1]^n, \qquad w_j := \frac{q_j}{\sum_{\ell=1}^{n} q_\ell}, \quad j = 1, \ldots, n.$$

**Remark 4.10.3.** Definition 4.10.2 is an objective weighting construction. It generalizes the criterion-impact-loss idea by first converting each uncertain entry $\mu_{ij} \in \mathrm{Dom}(M)$ into a positive representative score and then applying a loss-based criterion-weighting mechanism to the induced score matrix.

**Theorem 4.10.4** (Well-definedness of UCILOS). *Let*

$$\mathfrak{U}_{\mathrm{CILOS}} = \big( \mathcal{A}, \mathcal{C}, M, X_M, \mathrm{Score}_M \big)$$

*be a UCILOS instance as in Definition 4.10.2. Assume:*

*(A1) $\mathcal{A}$ and $\mathcal{C}$ are finite nonempty sets;*

*(A2)*

$$X_M = (\mu_{ij})_{m \times n} \in \mathrm{Dom}(M)^{m \times n};$$

*(A3) each criterion $C_j$ is designated either as benefit type or as cost type;*

*(A4)*

$$\mathrm{Score}_M : \mathrm{Dom}(M) \to \mathbb{R}_{>0}$$

*is a total map.*

*Then all quantities appearing in Definition 4.10.2 are well-defined. More precisely:*

*(i) the score matrix $S = (s_{ij}) \in \mathbb{R}_{>0}^{m \times n}$ is well-defined;*

*(ii) the benefit-type matrix $B = (b_{ij}) \in \mathbb{R}_{>0}^{m \times n}$ is well-defined;*

*(iii) the normalized matrix $Z = (z_{ij}) \in \mathbb{R}_{>0}^{m \times n}$ is well-defined, and for each $j$,*

$$\sum_{i=1}^{m} z_{ij} = 1;$$

*(iv) for each $j$, the optimal value $z_j^\star$ and at least one maximizer $\iota_j \in \arg\max_i z_{ij}$ exist;*

*(v) the square matrix $A = (a_{ij})$ and the relative-loss matrix $P = (p_{ij})$ are well-defined, with*

$$0 \le p_{ij} \le 1 \qquad (1 \le i, j \le n);$$

*(vi) the total losses $L_j$ and significance scores $q_j$ are well-defined and satisfy*

$$L_j \in [0, n-1], \qquad q_j \in \left[\frac{1}{n}, 1\right];$$

*(vii) the weight vector $w = (w_1, \ldots, w_n)^\top$ is well-defined and satisfies*

$$w_j > 0 \qquad \text{for all } j, \qquad \text{and} \qquad \sum_{j=1}^{n} w_j = 1.$$

*Hence Uncertain CILOS of type M is well-defined.*

*Proof.* By (A2), for every $i \in \{1, \ldots, m\}$ and $j \in \{1, \ldots, n\}$,

$$\mu_{ij} \in \mathrm{Dom}(M).$$

Since $\mathrm{Score}_M$ is total by (A4), each value

$$s_{ij} := \mathrm{Score}_M(\mu_{ij})$$

is well-defined and belongs to $\mathbb{R}_{>0}$. Therefore the score matrix

$$S = (s_{ij})_{m \times n} \in \mathbb{R}_{>0}^{m \times n}$$

is well-defined. This proves (i).

Next, define $B = (b_{ij})$ as in Definition 4.10.2. If $C_j$ is a benefit criterion, then

$$b_{ij} = s_{ij} > 0.$$

If $C_j$ is a cost criterion, then

$$b_{ij} = \frac{\min_{1 \le r \le m} s_{rj}}{s_{ij}}.$$

Because the finite set $\{s_{1j}, \ldots, s_{mj}\} \subset \mathbb{R}_{>0}$ has a positive minimum and $s_{ij} > 0$, this quotient is well-defined and strictly positive. Hence every $b_{ij} \in \mathbb{R}_{>0}$, so $B \in \mathbb{R}_{>0}^{m \times n}$. This proves (ii).

For each $j$, define

$$\beta_j := \sum_{r=1}^{m} b_{rj}.$$

Since each $b_{rj} > 0$ and $m \geq 1$, we have $\beta_j > 0$. Therefore

$$z_{ij} := \frac{b_{ij}}{\beta_j}$$

is well-defined and positive. Thus $Z = (z_{ij}) \in \mathbb{R}_{>0}^{m \times n}$ is well-defined. Moreover,

$$\sum_{i=1}^{m} z_{ij} = \sum_{i=1}^{m} \frac{b_{ij}}{\beta_j} = \frac{1}{\beta_j} \sum_{i=1}^{m} b_{ij} = 1.$$

This proves (iii).

Fix $j \in \{1, \ldots, n\}$. Since the finite nonempty set $\{z_{1j}, \ldots, z_{mj}\} \subset \mathbb{R}_{>0}$ has a maximum, the number

$$z_j^\star := \max_{1 \leq i \leq m} z_{ij}$$

exists, and the argmax set

$$\arg \max_{1 \leq i \leq m} z_{ij}$$

is nonempty. Hence at least one index $\iota_j$ can be chosen. This proves (iv).

Now define

$$a_{ij} := z_{\iota_i j}.$$

Since each $\iota_i \in \{1, \ldots, m\}$, every $a_{ij}$ is well-defined and belongs to $\mathbb{R}_{>0}$. Therefore $A = (a_{ij})$ is well-defined.

For the relative-loss matrix, first note that $z_j^\star > 0$. Also, since $z_j^\star$ is the maximum in column $j$,

$$0 < a_{ij} \leq z_j^\star.$$

Hence, for $i \neq j$,

$$p_{ij} = \frac{z_j^\star - a_{ij}}{z_j^\star}$$

is well-defined and satisfies $0 \leq p_{ij} \leq 1$. By definition $p_{jj} = 0$. Therefore $P = (p_{ij})$ is well-defined and all its entries lie in $[0, 1]$. This proves (v).

For each $j$,

$$L_j := \sum_{\substack{i=1 \\ i \neq j}}^{n} p_{ij}$$

is a finite sum of $n - 1$ numbers from $[0, 1]$. Therefore

$$0 \leq L_j \leq n - 1.$$

Consequently,

$$q_j := \frac{1}{1 + L_j}$$

is well-defined and satisfies

$$\frac{1}{n} \leq q_j \leq 1.$$

This proves (vi).

Finally, because each $q_j > 0$, the denominator

$$\sum_{\ell=1}^{n} q_\ell$$

is strictly positive. Hence

$$w_j := \frac{q_j}{\sum_{\ell=1}^{n} q_\ell}$$

is well-defined for every $j$, and $w_j > 0$. Moreover,

$$\sum_{j=1}^{n} w_j = \sum_{j=1}^{n} \frac{q_j}{\sum_{\ell=1}^{n} q_\ell} = \frac{\sum_{j=1}^{n} q_j}{\sum_{\ell=1}^{n} q_\ell} = 1.$$

Thus $w$ is a well-defined normalized weight vector. This proves (vii).

Therefore every step of the UCILOS construction is mathematically well-defined. $\square$

## 4.11 Fuzzy BWM (Fuzzy Best-Worst Method)

BWM selects the best and worst criteria, compares best-to-others and others-to-worst, then solves a constrained optimization to obtain consistent criterion weights for MCDA ranking tasks [575, 576]. Fuzzy BWM replaces pairwise comparison ratios with fuzzy numbers, solves for fuzzy weights under consistency constraints, and defuzzifies or ranks by possibility to prioritize criteria [577–579].

**Definition 4.11.1** (TFN-based Fuzzy Best–Worst Method (FBWM)). [577–579] Let $\mathcal{C} = \{c_1, \ldots, c_n\}$ be a set of decision criteria ($n \geq 2$). A triangular fuzzy number (TFN) is denoted by $\tilde{a} = (a^l, a^m, a^u)$ with $0 < a^l \leq a^m \leq a^u$.

**Step 1 (Best and worst criteria).** A decision-maker selects:

$$c_B \in \mathcal{C} \text{ (best/most important)}, \qquad c_W \in \mathcal{C} \text{ (worst/least important)}.$$

**Step 2 (Fuzzy pairwise-comparison vectors).** Using a linguistic scale mapped to TFNs, the decision-maker provides:

- the *Fuzzy Best-to-Others* vector

$$\tilde{\mathbf{A}}_B = (\tilde{a}_{B1}, \tilde{a}_{B2}, \ldots, \tilde{a}_{Bn}), \qquad \tilde{a}_{Bj} = (a_{Bj}^l, a_{Bj}^m, a_{Bj}^u),$$

where $\tilde{a}_{Bj}$ encodes the degree to which $c_B$ is preferred over $c_j$;

- the *Fuzzy Others-to-Worst* vector

$$\tilde{\mathbf{A}}_W = (\tilde{a}_{1W}, \tilde{a}_{2W}, \ldots, \tilde{a}_{nW}), \qquad \tilde{a}_{jW} = (a_{jW}^l, a_{jW}^m, a_{jW}^u),$$

where $\tilde{a}_{jW}$ encodes the degree to which $c_j$ is preferred over $c_W$.

As neutral self-comparisons, set $\tilde{a}_{BB} = \tilde{a}_{WW} = (1, 1, 1)$.

**Step 3 (Deriving fuzzy weights by linear programming).** The FBWM output is a TFN weight for each criterion

$$\tilde{w}_j = (w_j^l, w_j^m, w_j^u) \quad (j = 1, \ldots, n),$$

together with a satisfaction degree $\beta \in [0, 1]$. Let $t \in \{l, m, u\}$ denote the TFN component (lower/middle/upper). Introduce tolerance parameters $d_j^t > 0$ and $q_j^t > 0$.

Define the following linear program:

$$\max \ \beta$$

$$\text{s.t.} \ \ 0 \le w_B^t - a_{Bj}^t w_j^t \le d_j^t, \qquad\qquad \beta \le 1 - \frac{w_B^t - a_{Bj}^t w_j^t}{d_j^t}, \quad \forall j, \ \forall t \in \{l, m, u\},$$

$$- q_j^t \le w_j^t - a_{jW}^t w_W^t \le 0, \qquad\qquad \beta \le 1 + \frac{w_j^t - a_{jW}^t w_W^t}{q_j^t}, \quad \forall j, \ \forall t \in \{l, m, u\},$$

$$\sum_{i=1}^{n} w_i^m = 1,$$

$$w_j^u + \sum_{\substack{i=1 \\ i \ne j}}^{n} w_i^l \le 1, \qquad w_j^l + \sum_{\substack{i=1 \\ i \ne j}}^{n} w_i^u \ge 1, \qquad \forall j,$$

$$0 \le w_j^l \le w_j^m \le w_j^u, \qquad \forall j,$$

$$0 \le \beta \le 1.$$

Any optimal solution $\{\tilde{w}_j\}_{j=1}^n$ is called the *FBWM fuzzy weight vector* associated with $(\tilde{\mathbf{A}}_B, \tilde{\mathbf{A}}_W)$.

Uncertain BWM of type $(M)$ is defined as follows.

**Definition 4.11.2** (Admissible comparison scale for an uncertain model). Let $M$ be an uncertain model with degree-domain $\mathrm{Dom}(M) \subseteq [0, 1]^k$ (nonempty). An *admissible (ratio) comparison scale* for $M$ is a map

$$\kappa_M : \mathrm{Dom}(M) \longrightarrow [1, \Lambda]$$

for some fixed constant $\Lambda \ge 1$. The value $\kappa_M(d)$ is interpreted as a *crisp preference intensity* induced by the uncertainty-degree tuple $d \in \mathrm{Dom}(M)$.

**Remark.** The choice of $\kappa_M$ is model-dependent (e.g., a score/defuzzification plus rescaling) and should be specified by the decision analyst.

**Definition 4.11.3** (Uncertain BWM of type $M$ (U-BWM)). Let $\mathcal{C} = \{c_1, \ldots, c_n\}$ be a set of criteria with $n \geq 2$. Fix an uncertain model $M$ with $\mathrm{Dom}(M) \neq \emptyset$ and an admissible scale $\kappa_M$ as in Definition 4.11.2.

**Step 1 (Best and worst criteria).** Select

$$c_B \in \mathcal{C} \quad \text{(best / most important)}, \qquad c_W \in \mathcal{C} \quad \text{(worst / least important)}.$$

**Step 2 (Uncertain comparison vectors).** Provide the following two U-sets (uncertain vectors) of type $M$:

- *Best-to-Others* uncertain preferences

$$\mathbf{A}_B^{(M)} : \mathcal{C} \to \mathrm{Dom}(M), \qquad c_j \mapsto a_{Bj}^{(M)},$$

   where $a_{Bj}^{(M)}$ encodes how strongly $c_B$ is preferred over $c_j$;

- *Others-to-Worst* uncertain preferences

$$\mathbf{A}_W^{(M)} : \mathcal{C} \to \mathrm{Dom}(M), \qquad c_j \mapsto a_{jW}^{(M)},$$

   where $a_{jW}^{(M)}$ encodes how strongly $c_j$ is preferred over $c_W$.

Set neutral self-comparisons $a_{BB}^{(M)} = a_{WW}^{(M)}$ so that

$$\kappa_M\big(a_{BB}^{(M)}\big) = \kappa_M\big(a_{WW}^{(M)}\big) = 1.$$

**Step 3 (Crisp intensities induced by $M$).** Define induced (crisp) ratio intensities

$$\hat{a}_{Bj} := \kappa_M\big(a_{Bj}^{(M)}\big) \in [1, \Lambda], \qquad \hat{a}_{jW} := \kappa_M\big(a_{jW}^{(M)}\big) \in [1, \Lambda] \quad (j = 1, \ldots, n).$$

**Step 4 (Weight derivation via a minimax linear program).** The *U-BWM weight vector* is any optimal solution $w = (w_1, \ldots, w_n)$ of the linear program

$$\begin{aligned}
\min \ & \xi \\
\text{s.t.} \ & -\xi \leq w_B - \hat{a}_{Bj} w_j \leq \xi, \qquad j = 1, \ldots, n, \\
& -\xi \leq w_j - \hat{a}_{jW} w_W \leq \xi, \qquad j = 1, \ldots, n, \\
& \sum_{j=1}^{n} w_j = 1, \qquad w_j \geq 0 \ (j = 1, \ldots, n), \qquad \xi \geq 0.
\end{aligned}$$

The optimal value $\xi^\star$ is interpreted as a (model-induced) *consistency deviation*.

**Theorem 4.11.4** (Well-definedness of U-BWM). *Under the assumptions of Definition 4.11.3 (in particular, $\mathrm{Dom}(M) \neq \emptyset$ and $\kappa_M : \mathrm{Dom}(M) \to [1, \Lambda]$ exists), the U-BWM optimization problem is well-defined:*

*(i) the feasible region is nonempty;*

*(ii) an optimal solution $(w^\star, \xi^\star)$ exists;*

*(iii) every optimal weight vector satisfies $w^\star \in \Delta^{n-1} := \{w \in \mathbb{R}_{\geq 0}^n : \sum_{j=1}^n w_j = 1\}$.*

*Proof.* (i) Since $\mathrm{Dom}(M) \neq \emptyset$, the uncertain comparison vectors $\mathbf{A}_B^{(M)}, \mathbf{A}_W^{(M)}$ can be chosen (and are given) with values in $\mathrm{Dom}(M)$, hence the induced intensities $\hat{a}_{Bj}, \hat{a}_{jW}$ are well-defined real numbers in $[1, \Lambda]$. Consider the uniform weights $w_j := 1/n$ for all $j$; then $\sum_{j=1}^n w_j = 1$ and $w_j \geq 0$ hold.

Define
$$\xi_0 := \max\left\{ \max_{1 \leq j \leq n} |w_B - \hat{a}_{Bj} w_j|, \ \max_{1 \leq j \leq n} |w_j - \hat{a}_{jW} w_W| \right\} \geq 0.$$

With $(w, \xi) = (\frac{1}{n}\mathbf{1}, \xi_0)$ all linear inequalities in Definition 4.11.3 are satisfied, so the feasible region is nonempty.

(ii) The feasible set is described by finitely many linear equalities/inequalities, hence it is a closed polyhedron. Moreover, the constraints $\sum_{j=1}^n w_j = 1$ and $w_j \geq 0$ imply $0 \leq w_j \leq 1$ for all $j$. For any feasible $w$, choosing
$$\xi = \max\left\{ \max_j |w_B - \hat{a}_{Bj} w_j|, \ \max_j |w_j - \hat{a}_{jW} w_W| \right\}$$
always yields a feasible pair $(w, \xi)$, so the objective $\xi$ is bounded below by 0. Because the simplex constraint makes the $w$-coordinates bounded and the inequalities force $\xi$ to be at least the maximum of finitely many continuous expressions, the minimum of the linear objective over this nonempty closed feasible set is attained. Equivalently (and standard in linear programming), a feasible LP with objective bounded below admits an optimal solution.

(iii) By construction, every feasible (hence every optimal) solution satisfies the simplex constraints $\sum_{j=1}^n w_j = 1$ and $w_j \geq 0$, i.e., $w \in \Delta^{n-1}$. $\qquad\square$

Related concepts of BWM under uncertainty-aware models are listed in Table 4.11.

Table 4.11: Related concepts of BWM under uncertainty-aware models.

| $k$ | Related BWM concept(s) |
|---|---|
| 2 | Intuitionistic Fuzzy BWM [580, 581] |
| 2 | Bipolar Fuzzy BWM [582] |
| 2 | Pythagorean Fuzzy BWM [40] |
| 3 | Spherical Fuzzy BWM [583, 584] |
| 3 | Hesitant Fuzzy BWM [585, 586] |
| 3 | Picture fuzzy BWM (cf. [587]) |
| 3 | Neutrosophic BWM [588–590] |
| $n$ | Plithogenic BWM |

As extension- or related concepts other than Uncertain BWM, BWAHP [581], Bayesian BWM [591], Rough BWM [592, 593], BWM-TOPSIS [594, 595], Z-number BWM [596], Interval BWM [597], Robust BWM [598], Belief-based BWM [599], and Grey BWM [600, 601] are also known.

## 4.12 Fuzzy CRITIC

CRITIC is an objective weighting method that assigns criterion weights using data dispersion and inter-criterion conflict, emphasizing informative, discriminative, and nonredundant criteria [602, 603]. Fuzzy CRITIC computes objective criterion weights from fuzzy ratings using dispersion and intercriteria correlation, emphasizing informative, nonredundant criteria in scoring [604–606].

**Definition 4.12.1** (Fuzzy CRITIC weighting). [604–606] Let $\mathcal{A} = \{A_1, \ldots, A_m\}$ be a set of alternatives and $\mathcal{C} = \{C_1, \ldots, C_n\}$ a set of criteria. Assume a fuzzy decision matrix

$$\widetilde{X} = (\widetilde{x}_{ij})_{m \times n}, \qquad \widetilde{x}_{ij} \in \mathbb{F},$$

where $\mathbb{F}$ is a chosen family of fuzzy evaluations (e.g., TFNs, trapezoidal fuzzy numbers, interval-valued fuzzy numbers, etc.). Partition criteria into benefit and cost sets:

$$\mathcal{C} = \mathcal{C}^{\mathrm{ben}} \,\dot{\cup}\, \mathcal{C}^{\mathrm{cost}}.$$

Fix a *defuzzification/score* map

$$S : \mathbb{F} \to \mathbb{R},$$

which converts fuzzy evaluations into real numbers (e.g., for a TFN $\widetilde{x} = (\ell, m, u)$ one may take $S(\widetilde{x}) = (\ell + m + u)/3$). Define the crisp matrix $X = (x_{ij})$ by

$$x_{ij} := S(\widetilde{x}_{ij}) \qquad (i = 1, \ldots, m; \; j = 1, \ldots, n).$$

**(1) Min–max normalization.** For each criterion $C_j$, define

$$x_j^{\mathrm{min}} := \min_{1 \leq i \leq m} x_{ij}, \qquad x_j^{\mathrm{max}} := \max_{1 \leq i \leq m} x_{ij},$$

and assume $x_j^{\mathrm{max}} \neq x_j^{\mathrm{min}}$ for at least one $j$. The normalized matrix $R = (r_{ij})$ is defined by

$$r_{ij} := \begin{cases} \dfrac{x_{ij} - x_j^{\mathrm{min}}}{x_j^{\mathrm{max}} - x_j^{\mathrm{min}}}, & C_j \in \mathcal{C}^{\mathrm{ben}}, \\[2mm] \dfrac{x_j^{\mathrm{max}} - x_{ij}}{x_j^{\mathrm{max}} - x_j^{\mathrm{min}}}, & C_j \in \mathcal{C}^{\mathrm{cost}}. \end{cases}$$

Then $r_{ij} \in [0, 1]$.

**(2) Dispersion of each criterion.** Let

$$\bar{r}_j := \frac{1}{m} \sum_{i=1}^{m} r_{ij}, \qquad \sigma_j := \sqrt{\frac{1}{m-1} \sum_{i=1}^{m} (r_{ij} - \bar{r}_j)^2}.$$

**(3) Inter-criteria correlation.** For $j, k \in \{1, \ldots, n\}$ define Pearson correlation

$$\rho_{jk} := \frac{\sum_{i=1}^{m} (r_{ij} - \bar{r}_j)(r_{ik} - \bar{r}_k)}{\sqrt{\sum_{i=1}^{m} (r_{ij} - \bar{r}_j)^2} \sqrt{\sum_{i=1}^{m} (r_{ik} - \bar{r}_k)^2}},$$

whenever the denominator is nonzero (otherwise set $\rho_{jk} := 0$).

**(4) Information content and weights.** Define the CRITIC information content of criterion $C_j$ as

$$\Gamma_j := \sigma_j \sum_{k=1}^{n} (1 - \rho_{jk}),$$

and (when $\sum_{j=1}^{n} \Gamma_j > 0$) define fuzzy-CRITIC objective weights by

$$w_j := \frac{\Gamma_j}{\sum_{\ell=1}^{n} \Gamma_\ell}, \qquad j = 1, \ldots, n.$$

The resulting vector $w = (w_1, \ldots, w_n)^\top$ is called the *Fuzzy CRITIC* weight vector associated with $(\widetilde{X}, S)$.

**Proposition 4.12.2** (Well-definedness and simplex property)**.** *In Definition 4.12.1, assume $\sum_{j=1}^{n} \Gamma_j > 0$. Then $w_j \geq 0$ for all $j$ and $\sum_{j=1}^{n} w_j = 1$.*

*Proof.* For each $j$, $\sigma_j \geq 0$ by definition. Also, $\rho_{jk} \leq 1$, hence $1 - \rho_{jk} \geq 0$, so $\sum_{k=1}^{n} (1 - \rho_{jk}) \geq 0$ and therefore $\Gamma_j \geq 0$. With $\sum_{j=1}^{n} \Gamma_j > 0$, the formula $w_j = \Gamma_j / \sum_{\ell=1}^{n} \Gamma_\ell$ yields $w_j \geq 0$ and $\sum_{j=1}^{n} w_j = 1$ immediately. $\square$

Using Uncertain Sets, we define Uncertain CRITIC weighting of type $M$ as follows.

**Definition 4.12.3** (Admissible scoring map for an uncertain model)**.** Let $M$ be an uncertain model with nonempty degree-domain $\mathrm{Dom}(M) \subseteq [0,1]^k$. An *admissible score* for $M$ is a map

$$S_M : \mathrm{Dom}(M) \longrightarrow \mathbb{R}$$

such that $S_M(d)$ is finite for every $d \in \mathrm{Dom}(M)$. (Examples include defuzzification/score functions for fuzzy degrees, or scalarization maps for multi-component degrees such as intuitionistic/neutrosophic/plithogenic tuples.)

**Definition 4.12.4** (Uncertain CRITIC weighting of type $M$)**.** Let $\mathcal{A} = \{A_1, \ldots, A_m\}$ be alternatives and $\mathcal{C} = \{C_1, \ldots, C_n\}$ criteria with $m \geq 2$ and $n \geq 1$. Partition criteria into benefit and cost sets:

$$\mathcal{C} = \mathcal{C}^{\mathrm{ben}} \,\dot{\cup}\, \mathcal{C}^{\mathrm{cost}}.$$

Fix an uncertain model $M$ with $\mathrm{Dom}(M) \neq \emptyset$ and an admissible score $S_M$ as in Definition 4.12.3.

Assume an *uncertain decision matrix*

$$X^{(M)} = \big(x_{ij}^{(M)}\big)_{m \times n}, \qquad x_{ij}^{(M)} \in \mathrm{Dom}(M),$$

where $x_{ij}^{(M)}$ encodes the uncertainty-degree evaluation of alternative $A_i$ under criterion $C_j$.

**Step 0 (Crisp projection via scoring).** Define the real-valued matrix $X = (x_{ij})$ by

$$x_{ij} := S_M\big(x_{ij}^{(M)}\big) \in \mathbb{R}.$$

**Step 1 (Min–max normalization).** For each criterion $C_j$, set

$$x_j^{\min} := \min_{1 \leq i \leq m} x_{ij}, \qquad x_j^{\max} := \max_{1 \leq i \leq m} x_{ij}.$$

Define the normalized matrix $R = (r_{ij})$ by

$$r_{ij} := \begin{cases} \dfrac{x_{ij} - x_j^{\min}}{x_j^{\max} - x_j^{\min}}, & \text{if } C_j \in \mathcal{C}^{\mathrm{ben}} \text{ and } x_j^{\max} > x_j^{\min}, \\[2ex] \dfrac{x_j^{\max} - x_{ij}}{x_j^{\max} - x_j^{\min}}, & \text{if } C_j \in \mathcal{C}^{\mathrm{cost}} \text{ and } x_j^{\max} > x_j^{\min}, \\[2ex] 0, & \text{if } x_j^{\max} = x_j^{\min}. \end{cases}$$

(Thus degenerate criteria with zero range are handled by $r_{ij} = 0$.)

**Step 2 (Dispersion).** Let

$$\bar{r}_j := \frac{1}{m} \sum_{i=1}^{m} r_{ij}, \qquad \sigma_j := \sqrt{\frac{1}{m-1} \sum_{i=1}^{m} (r_{ij} - \bar{r}_j)^2} \quad (j = 1, \dots, n).$$

**Step 3 (Inter-criteria correlation).** For $j, k \in \{1, \dots, n\}$ define

$$\rho_{jk} := \begin{cases} \dfrac{\sum_{i=1}^{m} (r_{ij} - \bar{r}_j)(r_{ik} - \bar{r}_k)}{\sqrt{\sum_{i=1}^{m} (r_{ij} - \bar{r}_j)^2} \sqrt{\sum_{i=1}^{m} (r_{ik} - \bar{r}_k)^2}}, & \text{if } \sigma_j > 0 \text{ and } \sigma_k > 0, \\[2ex] 0, & \text{otherwise.} \end{cases}$$

**Step 4 (CRITIC information and objective weights).** Define the information content of $C_j$ by

$$\Gamma_j := \sigma_j \sum_{k=1}^{n} \big(1 - \rho_{jk}\big), \qquad j = 1, \dots, n.$$

If $\sum_{j=1}^{n} \Gamma_j > 0$, define the *Uncertain CRITIC* weights by

$$w_j := \frac{\Gamma_j}{\sum_{\ell=1}^{n} \Gamma_\ell}, \qquad j = 1, \dots, n.$$

The vector $w = (w_1, \dots, w_n)^{\top}$ is called the *Uncertain CRITIC weight vector of type M* associated with $(X^{(M)}, S_M)$.

**Theorem 4.12.5** (Well-definedness of Uncertain CRITIC). *In Definition 4.12.4, assume $m \geq 2$, $\mathrm{Dom}(M) \neq \emptyset$, and $S_M$ is admissible. Then:*

- *the matrices $X$ and $R$ are well-defined (all entries are finite real numbers);*

- *for every $j, k$, $\rho_{jk}$ is well-defined and satisfies $-1 \leq \rho_{jk} \leq 1$;*

- $\Gamma_j \geq 0$ *for all $j$;*

- *if $\sum_{j=1}^{n} \Gamma_j > 0$, then the resulting weight vector satisfies*

$$w_j \geq 0 \quad (j = 1, \ldots, n), \qquad \sum_{j=1}^{n} w_j = 1,$$

*i.e., $w \in \Delta^{n-1} := \{w \in \mathbb{R}_{\geq 0}^{n} : \sum_{j=1}^{n} w_j = 1\}$.*

*Proof.* Since $S_M : \mathrm{Dom}(M) \to \mathbb{R}$ is admissible, each $x_{ij} = S_M(x_{ij}^{(M)})$ is a finite real number, hence $X$ is well-defined. For each fixed $j$, $x_j^{\min}$ and $x_j^{\max}$ exist because the index set $\{1, \ldots, m\}$ is finite. The normalization rule defines $r_{ij}$ either by a ratio with denominator $x_j^{\max} - x_j^{\min} > 0$ or by the constant value $0$ when $x_j^{\max} = x_j^{\min}$. Therefore $R$ is well-defined and has finite real entries.

Next, $\sigma_j \geq 0$ holds by definition of a square root of a nonnegative quantity. If $\sigma_j = 0$ or $\sigma_k = 0$, Definition 4.12.4 sets $\rho_{jk} = 0$, hence it is well-defined. Assume $\sigma_j > 0$ and $\sigma_k > 0$. Let

$$u_i := r_{ij} - \bar{r}_j, \qquad v_i := r_{ik} - \bar{r}_k \qquad (i = 1, \ldots, m).$$

Then the numerator of $\rho_{jk}$ equals $\sum_{i=1}^{m} u_i v_i$, and the denominator equals $\sqrt{\sum_{i=1}^{m} u_i^2} \sqrt{\sum_{i=1}^{m} v_i^2}$, which is positive under $\sigma_j > 0$ and $\sigma_k > 0$. By the Cauchy–Schwarz inequality,

$$\left| \sum_{i=1}^{m} u_i v_i \right| \leq \sqrt{\sum_{i=1}^{m} u_i^2} \sqrt{\sum_{i=1}^{m} v_i^2},$$

hence $|\rho_{jk}| \leq 1$, so $-1 \leq \rho_{jk} \leq 1$.

For each $j$, $\sigma_j \geq 0$ and $1 - \rho_{jk} \geq 0$ because $\rho_{jk} \leq 1$. Therefore $\sum_{k=1}^{n}(1 - \rho_{jk}) \geq 0$ and thus $\Gamma_j = \sigma_j \sum_{k=1}^{n}(1 - \rho_{jk}) \geq 0$.

Finally, if $\sum_{j=1}^{n} \Gamma_j > 0$, then $w_j = \Gamma_j / \sum_{\ell=1}^{n} \Gamma_\ell$ is well-defined, satisfies $w_j \geq 0$, and

$$\sum_{j=1}^{n} w_j = \frac{\sum_{j=1}^{n} \Gamma_j}{\sum_{\ell=1}^{n} \Gamma_\ell} = 1.$$

Hence $w \in \Delta^{n-1}$. $\qquad\qquad \square$

Related concepts of CRITIC under uncertainty-aware models are listed in Table 4.12.

As extensions of CRITIC other than Uncertain CRITIC, several related concepts are also known, including Rough CRITIC [623], Modified CRITIC [624,625], and CRITID [626].

Table 4.12: Related concepts of CRITIC under uncertainty-aware models.

| $k$ | Related CRITIC concept(s) |
|---|---|
| 2 | Intuitionistic Fuzzy CRITIC [607,608] |
| 2 | Pythagorean Fuzzy CRITIC [609,610] |
| 2 | Fermatean Fuzzy CRITIC [611,612] |
| 3 | Picture Fuzzy CRITIC [613,614] |
| 3 | Spherical Fuzzy CRITIC [615–617] |
| 3 | Neutrosophic CRITIC [618–620] |
| $n$ | Plithogenic CRITIC [621,622] |

## 4.13   Fuzzy MEREC (Fuzzy MEthod based on the Removal Effects of Criteria)

The Method based on the Removal Effects of Criteria (MEREC) is an objective weighting method that determines criterion importance by measuring how the overall performance changes when each criterion is removed [627]. As related methods, approaches such as ITARA are also known [628,629]. Fuzzy MEREC derives objective weights by evaluating the change in fuzzy overall performance caused by removing each criterion, assigning larger weights to criteria with greater impact [630,631].

**Definition 4.13.1** (TFN-based Fuzzy MEREC (MEREC-F)). [630,631] Let $\mathcal{A} = \{A_1, \ldots, A_m\}$ be a finite set of alternatives and $\mathcal{C} = \{C_1, \ldots, C_n\}$ a finite set of criteria. Let $\mathcal{B} \subseteq \mathcal{C}$ and $\mathcal{K} \subseteq \mathcal{C}$ denote the sets of benefit-type and cost-type criteria, respectively, with $\mathcal{B} \cup \mathcal{K} = \mathcal{C}$ and $\mathcal{B} \cap \mathcal{K} = \varnothing$.

Assume the (positive) triangular fuzzy decision matrix

$$\tilde{X} = (\tilde{x}_{ij})_{m \times n}, \qquad \tilde{x}_{ij} = (l_{ij}, m_{ij}, u_{ij}), \quad 0 < l_{ij} \leq m_{ij} \leq u_{ij} \qquad (i = 1, \ldots, m; \ j = 1, \ldots, n).$$

**Step 1 (Fuzzy normalization).** Define the normalized TFNs $\tilde{r}_{ij} = (r_{ij}^l, r_{ij}^m, r_{ij}^u)$ by

$$\tilde{r}_{ij} = \begin{cases} \left(l_{ij}/u_j^\star, \ m_{ij}/u_j^\star, \ u_{ij}/u_j^\star\right), & C_j \in \mathcal{B}, \\ \left(l_j^-/u_{ij}, \ l_j^-/m_{ij}, \ l_j^-/l_{ij}\right), & C_j \in \mathcal{K}, \end{cases}$$

where

$$u_j^\star := \max_{1 \leq i \leq m} u_{ij} > 0, \qquad l_j^- := \min_{1 \leq i \leq m} l_{ij} > 0.$$

Then $\tilde{r}_{ij}$ is a positive TFN and (componentwise) lies in $(0, 1]$.

**Step 2 (Scalarization / defuzzification).** Convert $\tilde{r}_{ij}$ to a scalar $\eta_{ij} \in (0, 1]$ by a fixed defuzzification map $\delta : \mathrm{TFN}_{>0} \to (0, \infty)$. A standard choice (used widely for TFNs) is

$$\eta_{ij} := \delta(\tilde{r}_{ij}) = \frac{r_{ij}^l + 4r_{ij}^m + r_{ij}^u}{6}.$$

(If numerical zeros may occur in practice, one may replace $\eta_{ij}$ by $\max\{\eta_{ij}, \varepsilon\}$ with a fixed tiny $\varepsilon > 0$ so that $\ln(\eta_{ij})$ is well-defined.)

**Step 3 (Overall performance with all criteria).** Define, for each alternative $A_i$,

$$S_i := \ln\left(1 + \frac{1}{n}\sum_{j=1}^{n} |\ln(\eta_{ij})|\right).$$

**Step 4 (Leave-one-criterion-out performance).** For each criterion $C_j$ and each alternative $A_i$, define

$$S_i^{(-j)} := \ln\left(1 + \frac{1}{n}\sum_{\substack{k=1 \\ k\neq j}}^{n}|\ln(\eta_{ik})|\right).$$

**Step 5 (Removal effect).** For each criterion $C_j$, define its removal effect (impact) by

$$V_j := \sum_{i=1}^{m}|S_i^{(-j)} - S_i|.$$

**Step 6 (Objective criterion weights).** If $\sum_{j=1}^{n}V_j > 0$, define the (crisp) MEREC-F weights by

$$w_j := \frac{V_j}{\sum_{k=1}^{n}V_k}, \qquad j = 1,\ldots,n.$$

**Theorem 4.13.2** (Well-definedness of MEREC-F weights)**.** *Under the assumptions of Definition (MEREC-F), if $\sum_{j=1}^{n}V_j > 0$, then*

$$w_j \geq 0 \quad (j = 1,\ldots,n), \qquad and \qquad \sum_{j=1}^{n}w_j = 1.$$

*Moreover, $\sum_{j=1}^{n}V_j = 0$ holds if and only if $S_i^{(-j)} = S_i$ for all $i,j$, i.e., removing any single criterion never changes the overall performance values.*

*Proof.* By construction, each $V_j$ is a finite sum of absolute values, hence $V_j \geq 0$. If $\sum_{j=1}^{n}V_j > 0$, then $w_j$ is well-defined and nonnegative, and

$$\sum_{j=1}^{n}w_j = \sum_{j=1}^{n}\frac{V_j}{\sum_{k=1}^{n}V_k} = \frac{\sum_{j=1}^{n}V_j}{\sum_{k=1}^{n}V_k} = 1.$$

Finally, $\sum_{j=1}^{n}V_j = 0$ holds exactly when every $V_j = 0$, equivalently $|S_i^{(-j)} - S_i| = 0$ for all $i,j$, i.e., $S_i^{(-j)} = S_i$ for all $i,j$. $\qquad\square$

Uncertain MEREC is defined as follows.

**Definition 4.13.3** (Uncertain MEREC weighting of type $M$)**.** Let $\mathcal{A} = \{A_1,\ldots,A_m\}$ be a finite set of alternatives and $\mathcal{C} = \{C_1,\ldots,C_n\}$ a finite set of criteria, with $m \geq 1$ and $n \geq 2$. Let $\mathcal{B} \subseteq \mathcal{C}$ and $\mathcal{K} \subseteq \mathcal{C}$ denote the sets of benefit-type and cost-type criteria, respectively, with $\mathcal{B} \cup \mathcal{K} = \mathcal{C}$ and $\mathcal{B} \cap \mathcal{K} = \varnothing$.

Fix an uncertain model $M$ with $\mathrm{Dom}(M) \neq \emptyset$ and an admissible positive score $S_M$. Assume an *uncertain decision matrix*

$$X^{(M)} = \left(x_{ij}^{(M)}\right)_{m\times n}, \qquad x_{ij}^{(M)} \in \mathrm{Dom}(M) \quad (i = 1,\ldots,m;\ j = 1,\ldots,n).$$

**Step 0 (Crisp projection).** Define the positive real matrix $X = (x_{ij})$ by

$$x_{ij} := S_M\big(x_{ij}^{(M)}\big) \in (0, \infty).$$

**Step 1 (Positive normalization).** For each criterion $C_j$, define

$$x_j^\star := \max_{1 \le i \le m} x_{ij} > 0, \qquad x_j^- := \min_{1 \le i \le m} x_{ij} > 0.$$

Define the normalized scalars $\eta_{ij} \in (0, 1]$ by

$$\eta_{ij} := \begin{cases} \dfrac{x_{ij}}{x_j^\star}, & C_j \in \mathcal{B}, \\[2ex] \dfrac{x_j^-}{x_{ij}}, & C_j \in \mathcal{K}. \end{cases}$$

**(Optional numerical stabilization).** Fix a tiny constant $\varepsilon \in (0, 1)$ and replace $\eta_{ij}$ by

$$\eta_{ij}^\varepsilon := \max\{\eta_{ij}, \varepsilon\} \in [\varepsilon, 1].$$

For notational simplicity, write $\eta_{ij}$ for the stabilized values in what follows.

**Step 2 (Overall performance with all criteria).** For each alternative $A_i$, define

$$S_i := \ln\left(1 + \frac{1}{n} \sum_{j=1}^n \big|\ln(\eta_{ij})\big|\right).$$

**Step 3 (Leave-one-criterion-out performance).** For each $j \in \{1, \ldots, n\}$ and each $i \in \{1, \ldots, m\}$, define

$$S_i^{(-j)} := \ln\left(1 + \frac{1}{n} \sum_{\substack{k=1 \\ k \ne j}}^n \big|\ln(\eta_{ik})\big|\right).$$

**Step 4 (Removal effect).** For each criterion $C_j$, define its removal effect by

$$V_j := \sum_{i=1}^m \big|S_i^{(-j)} - S_i\big|.$$

**Step 5 (Objective weights).** If $\sum_{j=1}^n V_j > 0$, define the *Uncertain MEREC weights* by

$$w_j := \frac{V_j}{\sum_{k=1}^n V_k}, \qquad j = 1, \ldots, n.$$

The vector $w = (w_1, \ldots, w_n)^\top$ is called the *Uncertain MEREC* weight vector of type $M$ associated with $(X^{(M)}, S_M)$.

**Theorem 4.13.4** (Well-definedness of Uncertain MEREC). *Under the assumptions of Definition 4.13.3:*

1. *All quantities $\eta_{ij}$, $S_i$, $S_i^{(-j)}$, and $V_j$ are well-defined and finite real numbers, with $\eta_{ij} \in (0,1]$ and $V_j \geq 0$.*

2. *If $\sum_{j=1}^n V_j > 0$, then the weights are well-defined and satisfy*

$$w_j \geq 0 \quad (j = 1, \ldots, n), \qquad \sum_{j=1}^n w_j = 1,$$

   *i.e., $w \in \Delta^{n-1} := \{w \in \mathbb{R}_{\geq 0}^n : \sum_{j=1}^n w_j = 1\}$.*

3. *Moreover, $\sum_{j=1}^n V_j = 0$ holds if and only if $S_i^{(-j)} = S_i$ for all $i$ and $j$, i.e., removing any single criterion never changes the performance values.*

*Proof.* (1) By admissibility of $S_M$, each $x_{ij} = S_M(x_{ij}^{(M)})$ is strictly positive and finite. Hence for each $j$, the extrema $x_j^\star$ and $x_j^-$ exist (finite index set) and are positive. If $C_j \in \mathcal{B}$ then $\eta_{ij} = x_{ij}/x_j^\star \in (0,1]$; if $C_j \in \mathcal{K}$ then $\eta_{ij} = x_j^-/x_{ij} \in (0,1]$. After optional stabilization, $\eta_{ij} \in [\varepsilon, 1]$. Therefore $\ln(\eta_{ij})$ is defined and finite, so $|\ln(\eta_{ij})|$ is finite. Consequently, the sums in $S_i$ and $S_i^{(-j)}$ are finite, their arguments are $> 1$, and thus $S_i$ and $S_i^{(-j)}$ are finite real numbers. Finally, each $V_j$ is a finite sum of absolute values, hence $V_j \geq 0$ and finite.

(2) If $\sum_{j=1}^n V_j > 0$, then each $w_j = V_j / \sum_{k=1}^n V_k$ is well-defined and nonnegative. Moreover,

$$\sum_{j=1}^n w_j = \sum_{j=1}^n \frac{V_j}{\sum_{k=1}^n V_k} = \frac{\sum_{j=1}^n V_j}{\sum_{k=1}^n V_k} = 1,$$

so $w \in \Delta^{n-1}$.

(3) Since $V_j \geq 0$ for all $j$, one has $\sum_{j=1}^n V_j = 0$ if and only if $V_j = 0$ for all $j$. But $V_j = \sum_{i=1}^m |S_i^{(-j)} - S_i| = 0$ holds if and only if $|S_i^{(-j)} - S_i| = 0$ for all $i$, equivalently $S_i^{(-j)} = S_i$ for all $i$. $\qquad \square$

Related concepts of MEREC under uncertainty-aware models are listed in Table 4.13.

Table 4.13: Related concepts of MEREC under uncertainty-aware models.

| $k$ | **Related MEREC concept(s)** |
|---|---|
| 2 | Intuitionistic Fuzzy MEREC [632] |
| 2 | Pythagorean Fuzzy MEREC [633,634] |
| 3 | Spherical Fuzzy MEREC [630,635] |
| 3 | Neutrosophic MEREC [636,637] |

## 4.14 Fuzzy FUCOM

FUCOM orders criteria, elicits comparative priorities between consecutive criteria, and finds weights by minimizing deviation from these ratios while enforcing full consistency conditions in practice [638–640]. Fuzzy FUCOM expresses comparative priorities as fuzzy ratios, solves for fuzzy weights under consistency constraints, and applies defuzzification/possibility measures to obtain a stable ranking result [641,642].

**Definition 4.14.1** (Fuzzy FUCOM (FUCOM-F) with triangular fuzzy numbers). [641,642] Let $\mathcal{C} = \{C_1, \ldots, C_n\}$ be a finite set of criteria. Assume that decision makers (DMs) provide an *importance ranking*

$$C_{j(1)} \succcurlyeq C_{j(2)} \succcurlyeq \cdots \succcurlyeq C_{j(n)},$$

where $j(\cdot)$ is a permutation of $\{1, \ldots, n\}$ and ties are allowed.

**(0) Triangular fuzzy numbers (TFNs) and basic operations.** Let

$$\mathsf{TFN}_{>0} := \{(l, m, u) \in \mathbb{R}^3 : \ 0 < l \le m \le u\}.$$

For $\tilde{x} = (l_x, m_x, u_x)$ and $\tilde{y} = (l_y, m_y, u_y)$ in $\mathsf{TFN}_{>0}$ define

$$\tilde{x} \oplus \tilde{y} := (l_x + l_y, \ m_x + m_y, \ u_x + u_y), \qquad \tilde{x} \otimes \tilde{y} := (l_x l_y, \ m_x m_y, \ u_x u_y),$$

$$\tilde{y}^{-1} := \left(\frac{1}{u_y}, \frac{1}{m_y}, \frac{1}{l_y}\right), \qquad \tilde{x} \oslash \tilde{y} := \tilde{x} \otimes \tilde{y}^{-1} = \left(\frac{l_x}{u_y}, \frac{m_x}{m_y}, \frac{u_x}{l_y}\right), \qquad \tilde{1} := (1, 1, 1).$$

We use the componentwise preorder on TFNs:

$$\tilde{a} \preceq \tilde{b} \iff a^\ell \le b^\ell, \ a^m \le b^m, \ a^u \le b^u \quad \text{for } \tilde{a} = (a^\ell, a^m, a^u), \ \tilde{b} = (b^\ell, b^m, b^u),$$

and componentwise absolute value $|\tilde{a}| := (|a^\ell|, |a^m|, |a^u|)$.

**(1) Fuzzy comparative priorities between consecutive ranks.** For each rank $k = 1, \ldots, n-1$, DMs elicit a TFN

$$\tilde{\varphi}_{k/(k+1)} \in \mathsf{TFN}_{>0},$$

interpreted as the comparative priority of $C_{j(k)}$ over $C_{j(k+1)}$.

**(2) Unknown fuzzy weights.** The FUCOM-F goal is to determine fuzzy criterion weights

$$\tilde{w}_{j(k)} = (w_{j(k)}^\ell, w_{j(k)}^m, w_{j(k)}^u) \in \mathsf{TFN}_{\ge 0} \qquad (k = 1, \ldots, n),$$

with the usual TFN feasibility $0 \le w_{j(k)}^\ell \le w_{j(k)}^m \le w_{j(k)}^u$.

**(3) FUCOM-F consistency conditions (ideal equalities).** The two FUCOM consistency requirements are:

(C1) **(Ratio consistency)** For each $k = 1, \ldots, n-1$,

$$\frac{\tilde{w}_{j(k)}}{\tilde{w}_{j(k+1)}} = \tilde{\varphi}_{k/(k+1)} \qquad \text{equivalently} \qquad \tilde{w}_{j(k)} = \tilde{w}_{j(k+1)} \otimes \tilde{\varphi}_{k/(k+1)}.$$

(C2) **(Transitivity)** For each $k = 1, \ldots, n-2$,

$$\frac{\tilde{w}_{j(k)}}{\tilde{w}_{j(k+2)}} = \tilde{\varphi}_{k/(k+1)} \otimes \tilde{\varphi}_{(k+1)/(k+2)} \qquad \text{equivalently} \qquad \tilde{w}_{j(k)} = \tilde{w}_{j(k+2)} \otimes \tilde{\varphi}_{k/(k+1)} \otimes \tilde{\varphi}_{(k+1)/(k+2)}.$$

In practice, exact equality may not hold; FUCOM-F therefore minimizes the deviation from full consistency.

**(4) FUCOM-F optimization model (minimizing deviation from maximum consistency).** Let $\xi \geq 0$ be a (crisp) deviation variable (DMC). FUCOM-F determines the fuzzy weights by solving:

$$\min \xi \tag{4.11}$$
$$\text{s.t.} \ \left| \tilde{w}_{j(k)} - \tilde{w}_{j(k+1)} \otimes \tilde{\varphi}_{k/(k+1)} \right| \ \preceq \ (\xi, \xi, \xi), \qquad\qquad k = 1, \ldots, n-1, \tag{4.12}$$
$$\left| \tilde{w}_{j(k)} - \tilde{w}_{j(k+2)} \otimes \tilde{\varphi}_{k/(k+1)} \otimes \tilde{\varphi}_{(k+1)/(k+2)} \right| \ \preceq \ (\xi, \xi, \xi), \qquad k = 1, \ldots, n-2, \tag{4.13}$$
$$\bigoplus_{k=1}^{n} \tilde{w}_{j(k)} = \tilde{1}, \tag{4.14}$$
$$0 \leq w_{j(k)}^{\ell} \leq w_{j(k)}^{m} \leq w_{j(k)}^{u}, \qquad\qquad k = 1, \ldots, n. \tag{4.15}$$

If $\xi = 0$ is attainable, the obtained weights satisfy full (fuzzy) consistency.

**(5) Defuzzification (optional; crisp weights).** A common defuzzification for $\tilde{w} = (w^{\ell}, w^m, w^u)$ is the graded mean integration representation (GMIR):

$$\text{GMIR}(\tilde{w}) := \frac{w^{\ell} + 4w^m + w^u}{6}.$$

Define crisp weights and (optionally) renormalize:

$$w_i := \text{GMIR}(\tilde{w}_i), \qquad w_i \leftarrow \frac{w_i}{\sum_{t=1}^{n} w_t} \quad \text{so that} \quad \sum_{i=1}^{n} w_i = 1.$$

The resulting $(w_1, \ldots, w_n)$ is called the *FUCOM-F (defuzzified) criterion-weight vector*.

Using Uncertain Sets, we define Uncertain FUCOM (U-FUCOM) as follows.

**Definition 4.14.2** (Uncertain FUCOM (U-FUCOM): expected-value realization). Let $\mathcal{C} = \{C_1, \ldots, C_n\}$ be a finite set of criteria, $n \geq 2$. Assume decision makers provide an *importance order*

$$C_{j(1)} \succcurlyeq C_{j(2)} \succcurlyeq \cdots \succcurlyeq C_{j(n)},$$

where $j(\cdot)$ is a permutation of $\{1, \ldots, n\}$ (ties allowed). Work on a fixed uncertainty space $(\Gamma, \mathcal{L}, \mathcal{M})$.

**(1) Uncertain comparative priorities.** For each consecutive rank $k = 1, \ldots, n-1$, let

$$\Phi_k : \Gamma \to \mathcal{P}(\mathbb{R})$$

be an *uncertain comparative-priority coefficient* representing the relative importance of $C_{j(k)}$ over $C_{j(k+1)}$. Assume:

$$\Phi_k(\gamma) \subseteq \mathbb{R}_{>0} \quad (\forall \gamma \in \Gamma), \qquad \varphi_k := \mathbb{E}[\Phi_k] \in (0, \infty).$$

Define also $\varphi_{k,k+2} := \varphi_k \varphi_{k+1}$ for $k = 1, \ldots, n-2$.

**(2) Decision variables (crisp weights) and deviation.** Let $w_{j(1)}, \ldots, w_{j(n)} \in \mathbb{R}_{\geq 0}$ be unknown criterion weights and let $\xi \in \mathbb{R}_{\geq 0}$ be the deviation-from-full-consistency variable.

**(3) U-FUCOM optimization model.** U-FUCOM determines $(w, \xi)$ by solving the linear program:

$$\min \xi \tag{4.16}$$

$$\text{s.t.} \ -\xi \leq w_{j(k)} - \varphi_k \, w_{j(k+1)} \leq \xi, \qquad k = 1, \ldots, n-1, \tag{4.17}$$

$$-\xi \leq w_{j(k)} - \varphi_{k,k+2} \, w_{j(k+2)} \leq \xi, \qquad k = 1, \ldots, n-2, \tag{4.18}$$

$$\sum_{k=1}^{n} w_{j(k)} = 1, \tag{4.19}$$

$$w_{j(k)} \geq 0 \ (k = 1, \ldots, n), \qquad \xi \geq 0. \tag{4.20}$$

Any optimal solution $w^\star = (w_1^\star, \ldots, w_n^\star)$ (with indices permuted back from $j(\cdot)$) is called a *U-FUCOM weight vector*. If the optimal value satisfies $\xi^\star = 0$, the obtained weights are (fully) consistent with the expected ratios $\varphi_k$ and transitivity products $\varphi_k \varphi_{k+1}$.

**Definition 4.14.3** (Uncertain-set output of U-FUCOM). Let $w^\star$ be any U-FUCOM optimal weight vector from Definition 4.14.2. Define singleton-valued uncertain sets

$$W_i : \Gamma \to \mathcal{P}(\mathbb{R}), \qquad W_i(\gamma) := \{w_i^\star\}, \qquad i = 1, \ldots, n.$$

Then $W = (W_1, \ldots, W_n)$ is called the *U-FUCOM uncertain weight vector*.

**Theorem 4.14.4** (Uncertain-set structure and well-definedness of U-FUCOM). *Assume the setting of Definition 4.14.2 and suppose:*

(A1) *(Positive finite expected ratios)* $\varphi_k = \mathbb{E}[\Phi_k] \in (0, \infty)$ *for all* $k = 1, \ldots, n-1$.

*Then:*

(i) *the feasible set of* (4.16)–(4.20) *is nonempty;*

(ii) *an optimal solution* $(w^\star, \xi^\star)$ *exists;*

(iii) *every optimal weight vector satisfies* $w_i^\star \geq 0$ *and* $\sum_{i=1}^{n} w_i^\star = 1$;

(iv) *the output* $W = (W_1, \ldots, W_n)$ *in Definition 4.14.3 is an uncertain-set structured output on* $(\Gamma, \mathcal{L}, \mathcal{M})$.

*Proof.* **(i) Feasibility.** Consider the uniform weight vector $\bar{w}$ defined by $\bar{w}_i := 1/n$ for all $i$. Let

$$B := \max\left\{ 1, \ \max_{1 \leq k \leq n-1} \varphi_k, \ \max_{1 \leq k \leq n-2} \varphi_k \varphi_{k+1} \right\} \in (0, \infty).$$

For any $k$,

$$\left| \bar{w}_{j(k)} - \varphi_k \, \bar{w}_{j(k+1)} \right| \leq \bar{w}_{j(k)} + \varphi_k \, \bar{w}_{j(k+1)} \leq \frac{1}{n} + \frac{\varphi_k}{n} \leq \frac{1+B}{n},$$

and similarly

$$\left| \bar{w}_{j(k)} - (\varphi_k \varphi_{k+1}) \, \bar{w}_{j(k+2)} \right| \leq \frac{1}{n} + \frac{\varphi_k \varphi_{k+1}}{n} \leq \frac{1+B}{n}.$$

Hence $(\bar{w}, \bar{\xi})$ with $\bar{\xi} := (1+B)/n$ satisfies all constraints, so the feasible set is nonempty.

**(ii) Existence of an optimizer.** All feasible $w$ lie in the standard simplex

$$\Delta := \{ w \in \mathbb{R}^n : \ w_i \geq 0, \ \sum_{i=1}^{n} w_i = 1 \},$$

which is compact. Moreover, part (i) shows there exists a feasible point with $\xi = \bar{\xi}$, hence the optimal value satisfies $0 \leq \xi^\star \leq \bar{\xi}$. Therefore it suffices to minimize the continuous function $(w, \xi) \mapsto \xi$ over the compact set

$$\mathcal{F} \cap \left( \Delta \times [0, \bar{\xi}] \right),$$

where $\mathcal{F}$ is the feasible set; compactness follows because all constraints in (4.17)–(4.20) are closed linear inequalities/equalities. By the Weierstrass theorem, an optimizer $(w^\star, \xi^\star)$ exists.

**(iii) Normalization and nonnegativity.** These are enforced directly by (4.19)–(4.20).

**(iv) Uncertain-set structure.** For each $i$, the map $W_i(\gamma) = \{w_i^\star\}$ is constant (singleton-valued), hence measurable, and thus an uncertain set. Therefore $W = (W_1, \ldots, W_n)$ is an uncertain-set structured output on $(\Gamma, \mathcal{L}, \mathcal{M})$. $\qquad \square$

# Chapter 5

# Structure / causality decision-modelling (inter-criteria influence)

Structure/causality modelling methods (e.g., DEMATEL, ISM, MICMAC, FCM) elicit inter-criteria influence matrices, derive causal graphs, identify driving/dependent factors, and prioritize criteria accordingly.

For convenience, a concise comparison of the four structure/causality decision-modeling frameworks considered in this chapter is presented in Table 5.1.

Table 5.1: A concise comparison of four structure/causality decision-modeling frameworks.

| Method | Primary purpose | Typical input / core mechanism | Typical output / role in analysis |
|---|---|---|---|
| DEMATEL | To analyze the strength and direction of inter-criteria causal influence. | Direct-influence matrix; normalization and total-relation matrix computation; cause–effect analysis via prominence and relation indices. | Cause/effect grouping, influence intensity, and overall centrality of criteria. |
| ISM | To derive a hierarchical structural model from pairwise influence or reachability judgments. | Reachability-based relation matrix; transitive closure; level partitioning of factors. | A layered directed hierarchy showing which factors are foundational, intermediate, or top-level. |
| MICMAC | To classify factors according to driving power and dependence power. | Reachability matrix (often obtained from ISM); row/column aggregation and quadrant-based classification. | Partition of factors into autonomous, dependent, linkage, and driving classes. |
| FCM | To model feedback-rich causal systems and simulate their dynamic behavior over time. | Signed weighted causal graph (or weight matrix) together with an activation/update rule. | Dynamic trajectories, equilibrium states, cycles, and scenario-based system behavior under feedback. |

## 5.1 Fuzzy DEMATEL (Fuzzy Decision Making Trial and Evaluation Laboratory)

DEMATEL analyzes causal relationships among criteria using direct influence matrices, computes total influence, and separates factors into cause and effect groups [643–645]. Fuzzy DEMATEL models uncertain causal influences among criteria using fuzzy direct-relation matrices, derives total relations, and identifies cause–effect groups via prominence (D+R) and relation (D−R) [154, 646, 647].

**Definition 5.1.1** (TFN-based fuzzy DEMATEL data and outputs). [154, 646, 647] Let $C = \{C_1, \ldots, C_n\}$ be the set of criteria. Let $\mathsf{TFN} := \{(l, m, u) \in \mathbb{R}^3 : l \leq m \leq u\}$.

**(1) Linguistic-to-fuzzy encoding and expert aggregation.** Let $\mathcal{L}$ be a linguistic scale and let $\varphi : \mathcal{L} \to [0,1]^3$ map each linguistic term $\ell$ to a TFN $\varphi(\ell) = (l_\ell, m_\ell, u_\ell)$ (e.g. a 5-level scale). Assume $P \geq 1$ experts provide judgments $\ell_{ij}^{(p)} \in \mathcal{L}$ for the influence $C_i \to C_j$. Define the individual fuzzy direct-relation matrices

$$\tilde{X}^{(p)} = (\tilde{x}_{ij}^{(p)})_{n \times n}, \qquad \tilde{x}_{ij}^{(p)} := \varphi(\ell_{ij}^{(p)}), \qquad \tilde{x}_{ii}^{(p)} = (0, 0, 0),$$

and their componentwise mean (aggregation)

$$\tilde{X} = (\tilde{x}_{ij})_{n \times n}, \qquad \tilde{x}_{ij} := \frac{1}{P} \sum_{p=1}^{P} \tilde{x}_{ij}^{(p)}.$$

**(2) Defuzzification, normalization, and total relation.** For a TFN $\tilde{x} = (l, m, u)$, define BNP defuzzification by

$$\mathrm{BNP}(\tilde{x}) := \frac{l + m + u}{3}.$$

Defuzzify $\tilde{X}$ entrywise to obtain the crisp direct-relation matrix

$$X = (x_{ij}), \qquad x_{ij} := \mathrm{BNP}(\tilde{x}_{ij}).$$

Let

$$s := \max_{1 \leq i \leq n} \sum_{j=1}^{n} x_{ij}, \qquad N := \frac{1}{s} X.$$

Assume $\rho(N) < 1$. The total-relation matrix is

$$T := N(I - N)^{-1} = \sum_{k=1}^{\infty} N^k.$$

**(3) Dispatch/receive and cause–effect indices.** Let $\mathbf{1} \in \mathbb{R}^n$ be the all-ones vector and define

$$D := T\mathbf{1}, \qquad R := T^{\mathsf{T}}\mathbf{1}.$$

For each criterion $C_i$, define

$$\mathrm{Prominence}_i := D_i + R_i, \qquad \mathrm{Relation}_i := D_i - R_i.$$

Then $\mathrm{Prominence}_i$ measures overall centrality, while $\mathrm{Relation}_i > 0$ (resp. $< 0$) indicates a net cause (resp. net effect).

**Definition 5.1.2** (Uncertain DEMATEL (U-DEMATEL): expected-value realization). Let $C = \{C_1, \ldots, C_n\}$ be a finite set of criteria with $n \geq 2$, and fix an uncertainty space $(\Gamma, \mathcal{L}, \mathcal{M})$.

**(1) Uncertain direct-relation matrix.** An *uncertain direct influence assessment* is an $n \times n$ matrix

$$\tilde{X} = (\tilde{X}_{ij}) \in \mathsf{US}(\mathbb{R}_{\geq 0})^{n \times n}, \qquad \tilde{X}_{ii} \equiv 0,$$

where each entry $\widetilde{X}_{ij} : \Gamma \to \mathcal{P}(\mathbb{R}_{\geq 0})$ is an uncertain number encoding the (nonnegative) direct influence $C_i \to C_j$. Assume the expected values exist and define the *crisp expected direct-relation matrix*

$$X = (x_{ij}) \in \mathbb{R}_{\geq 0}^{n \times n}, \qquad x_{ij} := \mathbb{E}[\widetilde{X}_{ij}] < \infty.$$

**(2) Normalization.** Let

$$s := \max_{1 \leq i \leq n} \sum_{j=1}^{n} x_{ij}.$$

Assume $s > 0$ and define the normalized matrix

$$N := \frac{1}{s} X.$$

**(3) Total-relation matrix.** Define the *total-relation matrix* by the Neumann series

$$T := \sum_{k=1}^{\infty} N^k,$$

whenever the series converges. (Equivalently, when it converges one may write $T = N(I - N)^{-1}$.)

**(4) Prominence and relation indices.** Let $\mathbf{1} \in \mathbb{R}^n$ be the all-ones vector and set

$$D := T\mathbf{1}, \qquad R := T^{\mathsf{T}}\mathbf{1}.$$

For each criterion $C_i$, define

$$\text{Prominence}_i := D_i + R_i, \qquad \text{Relation}_i := D_i - R_i.$$

Then $\text{Relation}_i > 0$ indicates a net cause and $\text{Relation}_i < 0$ indicates a net effect (with respect to the expected influence network).

Using Uncertain Sets, we define U-DEMATEL as follows.

**Definition 5.1.3** (Uncertain-set output of U-DEMATEL). Let $(\text{Prominence}_i, \text{Relation}_i)_{i=1}^{n}$ be the indices produced by Definition 5.1.2. Define singleton-valued uncertain sets

$$\Pi_i : \Gamma \to \mathcal{P}(\mathbb{R}), \quad \Pi_i(\gamma) := \{\text{Prominence}_i\}, \qquad \mathcal{R}_i : \Gamma \to \mathcal{P}(\mathbb{R}), \quad \mathcal{R}_i(\gamma) := \{\text{Relation}_i\}.$$

The collection $\big((\Pi_i, \mathcal{R}_i)\big)_{i=1}^{n}$ is called the *U-DEMATEL uncertain index family*.

**Theorem 5.1.4** (Uncertain-set structure and well-definedness of U-DEMATEL). *In the setting of Definition 5.1.2, assume:*

(A1) *(Finite expectations)* $x_{ij} = \mathbb{E}[\widetilde{X}_{ij}] < \infty$ *for all* $i, j$, *and* $X \neq 0$ *(so $s > 0$).*

(A2) *(Spectral-radius condition)* $\rho(N) < 1$, *where $\rho(\cdot)$ denotes the spectral radius.*

*Then:*

  (i)  *the series* $T = \sum_{k=1}^{\infty} N^k$ *converges (entrywise and in any matrix norm);*

 (ii)  $(I - N)$ *is invertible and* $T = N(I - N)^{-1}$;

(iii)  *the vectors* $D = T\mathbf{1}$ *and* $R = T^{\mathsf{T}}\mathbf{1}$ *are well-defined and finite;*

 (iv)  *the indices* Prominence$_i$ *and* Relation$_i$ *are well-defined real numbers;*

  (v)  *the outputs in Definition 5.1.3 form an uncertain-set structured output on* $(\Gamma, \mathcal{L}, \mathcal{M})$.

*Proof.* **(i)** Under (A2), the matrix $N$ satisfies $\rho(N) < 1$. A standard result in linear algebra implies that for any induced matrix norm $\|\cdot\|$ there exists $k_0$ such that $\|N^{k_0}\| < 1$; hence $\sum_{k \geq 1} N^k$ converges absolutely in that norm. Therefore the Neumann series

$$\sum_{k=0}^{\infty} N^k$$

converges, and so does $T = \sum_{k=1}^{\infty} N^k$.

**(ii)** Since $\sum_{k=0}^{\infty} N^k$ converges, the Neumann-series identity holds:

$$(I - N)\left(\sum_{k=0}^{\infty} N^k\right) = I = \left(\sum_{k=0}^{\infty} N^k\right)(I - N),$$

so $(I - N)$ is invertible and

$$(I - N)^{-1} = \sum_{k=0}^{\infty} N^k.$$

Multiplying by $N$ gives

$$T = \sum_{k=1}^{\infty} N^k = N \sum_{k=0}^{\infty} N^k = N(I - N)^{-1}.$$

**(iii)** By (i), $T$ has finite real entries, hence $D = T\mathbf{1}$ and $R = T^{\mathsf{T}}\mathbf{1}$ are finite vectors in $\mathbb{R}^n$.

**(iv)** Each Prominence$_i = D_i + R_i$ and Relation$_i = D_i - R_i$ is therefore a finite real number.

**(v)** For each $i$, $\Pi_i(\gamma) = \{$Prominence$_i\}$ and $\mathcal{R}_i(\gamma) = \{$Relation$_i\}$ are constant singleton-valued maps, hence measurable and therefore uncertain sets. This yields an uncertain-set structured output on $(\Gamma, \mathcal{L}, \mathcal{M})$.   □

Related concepts of DEMATEL under uncertainty-aware models are listed in Table 5.2.

Uncertain DEMATEL has also been extended in other directions; for example, Rough DEMATEL [666, 667], AHP-DEMATEL [668, 669], TOPSIS-DEMATEL [670, 671], Interval DEMATEL [672, 673], Linguistic DEMATEL [674, 675], and Grey DEMATEL [676, 677] are also known as related variants.

Table 5.2: Related concepts of DEMATEL under uncertainty-aware models.

| $k$ | Related DEMATEL concept(s) | Representative references |
|---|---|---|
| 1 | Fuzzy DEMATEL | [648, 649] |
| 2 | Intuitionistic Fuzzy DEMATEL | [650–652] |
| 2 | Bipolar Fuzzy DEMATEL | [653, 654] |
| 2 | Pythagorean fuzzy DEMATEL | [655, 656] |
| 2 | Fermatean fuzzy DEMATEL | [657–659] |
| 3 | Picture Fuzzy DEMATEL | [660] |
| 3 | Hesitant Fuzzy DEMATEL | [661] |
| 3 | Spherical Fuzzy DEMATEL | [662, 663] |
| 3 | Neutrosophic DEMATEL | [664, 665] |

## 5.2  Fuzzy ISM (Fuzzy Interpretive Structural Modeling)

Interpretive Structural Modeling structures complex factors by expert judgments on pairwise reachability, builds a transitive reachability matrix, and derives a hierarchical directed graph [678–680]. Fuzzy ISM replaces binary reachability with graded membership degrees from linguistic judgments, propagates fuzziness through transitive closure, and yields a hierarchical influence graph [681, 682]. As an extension, concepts such as Fuzzy Total Interpretive Structural Modeling have also been studied [683, 684].

**Definition 5.2.1** (Fuzzy Interpretive Structural Modeling (Fuzzy ISM)).  [681, 682] Let $\mathcal{F} = \{f_1, \ldots, f_n\}$ be a finite set of factors (system variables), with $n \geq 2$.

**Step 1 (Directional judgments: SSIM symbols).** For each unordered pair $\{i, j\}$ with $i \neq j$, an expert specifies the *direction* of influence using

$$s_{ij} \in \{V, A, X, O\},$$

interpreted as:

$$
\begin{aligned}
V : & \quad f_i \text{ influences } f_j, \\
A : & \quad f_j \text{ influences } f_i, \\
X : & \quad f_i \text{ and } f_j \text{ influence each other}, \\
O : & \quad f_i \text{ and } f_j \text{ are unrelated}.
\end{aligned}
$$

(Equivalently, one may store these symbols in a self-structural interaction matrix (SSIM).)

**Step 2 (Fuzzy strength of influence).** Fix a family $\mathbb{F}$ of fuzzy degrees representing influence strengths, together with a mapping from linguistic terms to $\mathbb{F}$. Typical choices include:

- $\mathbb{F} = [0, 1]$ (direct fuzzy degrees), or

- $\mathbb{F} = \text{TFN}_{[0,1]}$ (triangular fuzzy numbers within $[0, 1]$).

For each pair $\{i, j\}$ with $i \neq j$, assign a fuzzy influence strength

$$\tilde{a}_{ij} \in \mathbb{F}$$

corresponding to the expert's linguistic assessment of the influence magnitude.

**Step 3 (Fuzzy adjacency / AFSSIM).** Define the (possibly aggregated) fuzzy adjacency matrix $\widetilde{A} = (\tilde{a}^{\rightarrow}_{ij})_{n \times n} \in \mathbb{F}^{n \times n}$ by setting $\tilde{a}^{\rightarrow}_{ii} := 1_{\mathbb{F}}$ (full self-reachability) and, for $i \neq j$,

$$(\tilde{a}^{\rightarrow}_{ij}, \tilde{a}^{\rightarrow}_{ji}) := \begin{cases} (\tilde{a}_{ij}, 0_{\mathbb{F}}), & s_{ij} = V, \\ (0_{\mathbb{F}}, \tilde{a}_{ij}), & s_{ij} = A, \\ (\tilde{a}_{ij}, \tilde{a}_{ij}), & s_{ij} = X, \\ (0_{\mathbb{F}}, 0_{\mathbb{F}}), & s_{ij} = O, \end{cases}$$

where $0_{\mathbb{F}}$ and $1_{\mathbb{F}}$ denote the null and unit elements in $\mathbb{F}$. If multiple experts provide matrices $\widetilde{A}^{(1)}, \ldots, \widetilde{A}^{(p)}$, their aggregation (AFSSIM) is obtained by a fixed aggregation operator $\mathrm{Agg} : \mathbb{F}^p \to \mathbb{F}$ applied entrywise:

$$\widetilde{A}_{ij} := \mathrm{Agg}\big(\widetilde{A}^{(1)}_{ij}, \ldots, \widetilde{A}^{(p)}_{ij}\big).$$

**Step 4 (Fuzzy reachability matrix via fuzzy transitive closure).** Assume $\mathbb{F}$ supports a fuzzy conjunction $\wedge$ and disjunction $\vee$ (e.g., $\min/\max$ on $[0,1]$). For matrices over $\mathbb{F}$, define the max–min product $\circ$ by

$$(\widetilde{P} \circ \widetilde{Q})_{ik} := \bigvee_{j=1}^{n} \big(\widetilde{P}_{ij} \wedge \widetilde{Q}_{jk}\big).$$

Let $\widetilde{A}^{(1)} := \widetilde{A}$ and $\widetilde{A}^{(t+1)} := \widetilde{A}^{(t)} \circ \widetilde{A}$. The *fuzzy reachability matrix* (FRM) is defined as the fuzzy transitive closure

$$\widetilde{R} := \widetilde{A}^{(1)} \ \vee \ \widetilde{A}^{(2)} \ \vee \ \cdots \ \vee \ \widetilde{A}^{(n)}.$$

**Step 5 (Defuzzification and thresholding: DFRM).** Fix a crisp score/defuzzification map $\mathrm{Score} : \mathbb{F} \to [0,1]$ and a threshold $\theta \in (0,1]$. Define the *defuzzified (binary) reachability matrix* $D = (d_{ij}) \in \{0,1\}^{n \times n}$ by

$$d_{ij} := \begin{cases} 1, & \mathrm{Score}(\widetilde{R}_{ij}) \geq \theta, \\ 0, & \mathrm{Score}(\widetilde{R}_{ij}) < \theta. \end{cases}$$

**Step 6 (Level partitioning and ISM hierarchy).** For each $i$, define the reachability set, antecedent set, and intersection set:

$$\mathrm{Reach}(i) := \{\, f_j \in \mathcal{F} \mid d_{ij} = 1 \,\}, \qquad \mathrm{Ante}(i) := \{\, f_j \in \mathcal{F} \mid d_{ji} = 1 \,\}, \qquad \mathrm{Inter}(i) := \mathrm{Reach}(i) \cap \mathrm{Ante}(i).$$

The *top level* consists of all $f_i$ satisfying $\mathrm{Reach}(i) = \mathrm{Inter}(i)$. Remove these factors and repeat the same construction on the remaining set until all factors are assigned to levels. The resulting layered digraph (with edges $f_i \to f_j$ when $d_{ij} = 1$) is called the *Fuzzy ISM model* induced by the expert judgments and the chosen fuzzy/defuzzification settings.

**Proposition 5.2.2** (Basic well-posedness and termination). *Assume $\mathcal{F}$ is finite and $\mathbb{F}$ admits $\wedge, \vee$ so that max–min composition is defined. Then:*

- *the fuzzy transitive closure $\widetilde{R}$ in Definition 5.2.1 is well-defined;*

- *for any score map* Score *and threshold* $\theta \in (0, 1]$, *the binary matrix $D$ is well-defined;*

- *the level partitioning procedure terminates after at most $n$ iterations and assigns every factor to exactly one level.*

*Proof.* All entries of $\widetilde{A}$ lie in $\mathbb{F}$, and $\widetilde{A}^{(t)}$ is obtained from finitely many $\wedge, \vee$ operations, hence is well-defined; so is $\widetilde{R}$ as a finite join. Applying Score and thresholding yields a binary entry for each pair $(i, j)$, so $D$ is well-defined. At each level-partition iteration, at least one factor is selected for the current top level (a standard property of reachability/antecedent-based partitioning on finite sets), hence the remaining set strictly decreases. Therefore the procedure terminates in at most $n$ steps, and disjointness of removed sets implies each factor is assigned to exactly one level. $\square$

We now generalize this framework by allowing pairwise influences to take values in an arbitrary uncertain model $M$.

**Definition 5.2.3** (Uncertain ISM of type $M$). Let

$$\mathcal{F} = \{f_1, \ldots, f_n\}$$

be a finite nonempty set of factors, where $n \geq 1$. Fix an uncertain model $M$ with degree-domain

$$\mathrm{Dom}(M) \subseteq [0, 1]^k$$

for some integer $k \geq 1$.

Assume that the direct influence structure is represented by an *uncertain influence relation*

$$R_M : \mathcal{F} \times \mathcal{F} \longrightarrow \mathrm{Dom}(M),$$

where $R_M(f_i, f_j)$ expresses the uncertain degree to which factor $f_i$ directly influences factor $f_j$.

Fix further:

- a total score map

$$\mathrm{Score}_M : \mathrm{Dom}(M) \longrightarrow [0, 1],$$

which converts uncertain influence values into scalar influence strengths;

- a threshold

$$\theta \in (0, 1].$$

**Step 1 (Scored adjacency matrix).** Define the scored adjacency matrix

$$A = (a_{ij})_{n \times n} \in [0, 1]^{n \times n}$$

by

$$a_{ij} := \begin{cases} 1, & i = j, \\ \mathrm{Score}_M\big(R_M(f_i, f_j)\big), & i \neq j. \end{cases}$$

Thus self-reachability is fixed at 1, while off-diagonal entries are induced from the uncertain relation.

**Step 2 (Max–min reachability closure).** For matrices $P, Q \in [0,1]^{n \times n}$, define their max–min product $P \circ Q$ by

$$(P \circ Q)_{ik} := \max_{1 \leq j \leq n} \min\{P_{ij}, Q_{jk}\}.$$

Define recursively

$$A^{[1]} := A, \qquad A^{[t+1]} := A^{[t]} \circ A \qquad (t \geq 1).$$

The *uncertain reachability matrix* is defined by

$$\widetilde{R} := A^{[1]} \vee A^{[2]} \vee \cdots \vee A^{[n]},$$

where $\vee$ denotes the entrywise maximum. Writing

$$\widetilde{R} = (r_{ij})_{n \times n},$$

each $r_{ij} \in [0,1]$ represents the aggregated uncertain reachability strength from $f_i$ to $f_j$.

**Step 3 (Binary reachability matrix).** Define the binary reachability matrix

$$D = (d_{ij})_{n \times n} \in \{0,1\}^{n \times n}$$

by thresholding:

$$d_{ij} := \begin{cases} 1, & r_{ij} \geq \theta, \\ 0, & r_{ij} < \theta. \end{cases}$$

**Step 4 (Level partitioning).** Set

$$U_1 := \mathcal{F}.$$

For each iteration $t \geq 1$, provided $U_t \neq \varnothing$, define for every $f_i \in U_t$:

$$\text{Reach}_t(i) := \{ f_j \in U_t : d_{ij} = 1 \},$$

$$\text{Ante}_t(i) := \{ f_j \in U_t : d_{ji} = 1 \},$$

$$\text{Inter}_t(i) := \text{Reach}_t(i) \cap \text{Ante}_t(i).$$

The $t$-th level is

$$L_t := \{ f_i \in U_t : \text{Reach}_t(i) = \text{Inter}_t(i) \}.$$

If $U_t \neq \varnothing$, remove the identified level and define

$$U_{t+1} := U_t \setminus L_t.$$

Repeat until the remaining set becomes empty.

The ordered family of levels

$$(L_1, L_2, \ldots, L_T)$$

together with the directed graph on $\mathcal{F}$ whose edges are given by

$$f_i \rightarrow f_j \quad \Longleftrightarrow \quad d_{ij} = 1$$

is called the *Uncertain ISM model of type $M$* induced by

$$(\mathcal{F}, R_M, \text{Score}_M, \theta).$$

**Theorem 5.2.4** (Well-definedness of Uncertain ISM). *Assume the data in Definition 5.2.3 satisfy:*

*(A1)* $\mathcal{F} = \{f_1, \ldots, f_n\}$ *is finite and nonempty;*

*(A2)* $R_M : \mathcal{F} \times \mathcal{F} \to \mathrm{Dom}(M)$ *is a total map;*

*(A3)* $\mathrm{Score}_M : \mathrm{Dom}(M) \to [0, 1]$ *is a total map;*

*(A4)* $\theta \in (0, 1]$.

*Then:*

*(i) the scored adjacency matrix $A$, all powers $A^{[t]}$ ($t = 1, \ldots, n$), and the uncertain reachability matrix $\widetilde{R}$ are well-defined;*

*(ii) the binary reachability matrix $D$ is well-defined, reflexive, and transitive;*

*(iii) for every iteration $t$ with $U_t \neq \varnothing$, the level $L_t$ is well-defined and nonempty;*

*(iv) the level-partitioning procedure terminates after at most $n$ iterations;*

*(v) the obtained levels $L_1, \ldots, L_T$ are pairwise disjoint and satisfy*

$$\mathcal{F} = L_1 \sqcup L_2 \sqcup \cdots \sqcup L_T.$$

*Hence Uncertain ISM of type $M$ is well-defined.*

*Proof.* **(i) Well-definedness of $A$, $A^{[t]}$, and $\widetilde{R}$.** For each pair $(i, j)$, the value $R_M(f_i, f_j) \in \mathrm{Dom}(M)$ is defined by (A2), and

$$\mathrm{Score}_M\big(R_M(f_i, f_j)\big) \in [0, 1]$$

is defined by (A3). Therefore every entry $a_{ij}$ of $A$ is well-defined, so

$$A \in [0, 1]^{n \times n}.$$

Now let $P, Q \in [0, 1]^{n \times n}$. Since $\min\{P_{ij}, Q_{jk}\} \in [0, 1]$ for every $j$, and the maximum of finitely many numbers in $[0, 1]$ again lies in $[0, 1]$, the entry

$$(P \circ Q)_{ik} = \max_{1 \leq j \leq n} \min\{P_{ij}, Q_{jk}\}$$

is well-defined and belongs to $[0, 1]$. Hence $P \circ Q \in [0, 1]^{n \times n}$.

By induction, each $A^{[t]} \in [0, 1]^{n \times n}$ is well-defined for $t = 1, \ldots, n$. Since $\widetilde{R}$ is the entrywise maximum of finitely many matrices $A^{[1]}, \ldots, A^{[n]}$, it is also well-defined and belongs to $[0, 1]^{n \times n}$.

**(ii) Well-definedness, reflexivity, and transitivity of $D$.** Because every $r_{ij} \in [0,1]$ is well-defined and $\theta \in (0,1]$, each entry $d_{ij} \in \{0,1\}$ is well-defined by thresholding. Thus $D \in \{0,1\}^{n \times n}$.

Since $a_{ii} = 1$ for every $i$, we also have $r_{ii} \geq 1$, hence $r_{ii} = 1$. Because $\theta \leq 1$, it follows that $d_{ii} = 1$ for all $i$. Therefore $D$ is reflexive.

We next show transitivity. For a directed path

$$\pi = (i = i_0, i_1, \ldots, i_m = j)$$

in $\{1, \ldots, n\}$, define its strength by

$$\text{str}(\pi) := \min_{0 \leq q < m} a_{i_q i_{q+1}}.$$

By construction of $\widetilde{R} = A^{[1]} \vee \cdots \vee A^{[n]}$, the entry $r_{ij}$ is the maximum strength of all simple directed paths from $i$ to $j$; it suffices to consider simple paths because removing a cycle does not decrease the minimum edge value along the path, and any simple path has length at most $n - 1$.

Now suppose $d_{ij} = 1$ and $d_{jk} = 1$. Then

$$r_{ij} \geq \theta, \qquad r_{jk} \geq \theta.$$

Choose simple paths $\pi_{ij}$ from $i$ to $j$ and $\pi_{jk}$ from $j$ to $k$ whose strengths are $r_{ij}$ and $r_{jk}$, respectively. Concatenating them gives a path from $i$ to $k$; after removing repeated cycles, one obtains a simple path from $i$ to $k$ whose strength is at least

$$\min\{r_{ij}, r_{jk}\} \geq \theta.$$

Hence

$$r_{ik} \geq \min\{r_{ij}, r_{jk}\} \geq \theta,$$

so $d_{ik} = 1$. Therefore $D$ is transitive.

**(iii) Nonemptiness of each level $L_t$.** Fix an iteration $t$ with $U_t \neq \varnothing$. Restrict the relation represented by $D$ to $U_t$. Since $D$ is reflexive and transitive, the restricted relation is also reflexive and transitive.

Define an equivalence relation $\sim_t$ on $U_t$ by

$$f_i \sim_t f_j \quad \Longleftrightarrow \quad d_{ij} = 1 \text{ and } d_{ji} = 1.$$

Let $U_t / \sim_t$ be the set of equivalence classes, and define a relation $\preceq_t$ on these classes by

$$[f_i] \preceq_t [f_j] \quad \Longleftrightarrow \quad d_{ij} = 1.$$

Because $D$ is reflexive and transitive, $\preceq_t$ is a partial order on the finite set $U_t / \sim_t$. Hence it has at least one maximal class; choose one and denote it by $[f_{i^\star}]$.

Take any $f_j \in \text{Reach}_t(i^\star)$. Then $d_{i^\star j} = 1$, so

$$[f_{i^\star}] \preceq_t [f_j].$$

By maximality of $[f_{i^\star}]$, this implies $[f_j] = [f_{i^\star}]$. Therefore $d_{ji^\star} = 1$, i.e.,

$$f_j \in \text{Ante}_t(i^\star).$$

Thus

$$\mathrm{Reach}_t(i^\star) \subseteq \mathrm{Ante}_t(i^\star).$$

Consequently,

$$\mathrm{Reach}_t(i^\star) = \mathrm{Reach}_t(i^\star) \cap \mathrm{Ante}_t(i^\star) = \mathrm{Inter}_t(i^\star).$$

Hence $f_{i^\star} \in L_t$, so $L_t \neq \varnothing$.

**(iv) Termination.** Whenever $U_t \neq \varnothing$, part (iii) shows that $L_t \neq \varnothing$. Therefore

$$U_{t+1} = U_t \setminus L_t$$

is a strict subset of $U_t$. Since $|U_1| = n$, after at most $n$ iterations the remaining set becomes empty. Hence the procedure terminates.

**(v) Disjointness and covering property.** By construction,

$$L_t \subseteq U_t \quad \text{and} \quad U_{t+1} = U_t \setminus L_t.$$

Therefore the sets $L_1, L_2, \ldots, L_T$ are pairwise disjoint. Since the procedure stops exactly when no factors remain, every factor is removed at some step, whence

$$\mathcal{F} = L_1 \sqcup L_2 \sqcup \cdots \sqcup L_T.$$

All required objects are therefore well-defined, and the hierarchical decomposition exists and terminates.   $\square$

For reference, related concepts of Interpretive Structural Modeling under uncertainty-aware models are listed in Table 5.3.

Table 5.3: Related concepts of Interpretive Structural Modeling under uncertainty-aware models.

| $k$ | **Related ISM concept(s)** |
|---|---|
| 1 | Fuzzy Interpretive Structural Modeling |
| 2 | Intuitionistic Fuzzy Interpretive Structural Modeling |
| 3 | Neutrosophic Interpretive Structural Modeling [685–687] |

## 5.3   Fuzzy MICMAC (cross-impact / driving–dependence analysis)

MICMAC analyzes a binary reachability matrix to compute driving power and dependence, classifying factors into autonomous, dependent, linkage, and driving clusters [688, 689]. Fuzzy MICMAC computes driving and dependence strengths from fuzzy reachability values, classifies criteria using graded impacts, and supports robust clustering under ambiguous judgments [690–692].

**Definition 5.3.1** (Fuzzy MICMAC (driving–dependence analysis)). [693–695] Let $\mathcal{F} = \{f_1, \ldots, f_n\}$ be a finite set of system factors ($n \geq 2$). Assume that a *fuzzy reachability matrix* (FRM) has been obtained (typically after fuzzy transitive closure in Fuzzy ISM),

$$\widetilde{R} = (\tilde{r}_{ij})_{n \times n}, \qquad \tilde{r}_{ij} \in \mathbb{F},$$

where $\mathbb{F}$ is a chosen fuzzy evaluation family (e.g., TFNs in $[0,1]$), introduced to handle uncertainty in expert judgments.

**Step 1 (Defuzzification / scoring).** Fix a score map $\mathrm{Score} : \mathbb{F} \to [0,1]$ and define the crisp reachability matrix

$$R = (r_{ij})_{n \times n}, \qquad r_{ij} := \mathrm{Score}(\tilde{r}_{ij}) \in [0,1].$$

**Step 2 (Thresholding: DFRM).** Fix a threshold $\theta \in (0,1]$ and define the *defuzzified (binary) reachability matrix* (DFRM)

$$D = (d_{ij})_{n \times n} \in \{0,1\}^{n \times n}, \qquad d_{ij} := \mathbf{1}[\, r_{ij} \geq \theta\,].$$

(Thresholding assigns 1 when the value is at least the cutoff and 0 otherwise.)

**Step 3 (Driving power and dependence power).** Define the *driving power* and *dependence power* of each factor $f_i$ by

$$\mathrm{DRP}(i) := \sum_{j=1}^{n} d_{ij}, \qquad \mathrm{DEP}(i) := \sum_{j=1}^{n} d_{ji}.$$

Row sums measure how many factors are reachable (directly or indirectly), while column sums measure how strongly a factor is influenced by others.

**Step 4 (Quadrant-based MICMAC classification).** Choose cutoffs $(\tau_D, \tau_P)$ for "high/low" driving and dependence; a common choice is

$$\tau_D := \frac{1}{n} \sum_{i=1}^{n} \mathrm{DRP}(i), \qquad \tau_P := \frac{1}{n} \sum_{i=1}^{n} \mathrm{DEP}(i).$$

Classify each factor $f_i$ into one of the four MICMAC clusters (autonomous, dependent, linkage, independent) according to its $(\mathrm{DRP}(i), \mathrm{DEP}(i))$ location. Formally,

$$\mathcal{F}_{\mathrm{aut}} := \{f_i : \ \mathrm{DRP}(i) < \tau_D \ \text{and} \ \mathrm{DEP}(i) < \tau_P\},$$
$$\mathcal{F}_{\mathrm{dep}} := \{f_i : \ \mathrm{DRP}(i) < \tau_D \ \text{and} \ \mathrm{DEP}(i) \geq \tau_P\},$$
$$\mathcal{F}_{\mathrm{lin}} := \{f_i : \ \mathrm{DRP}(i) \geq \tau_D \ \text{and} \ \mathrm{DEP}(i) \geq \tau_P\},$$
$$\mathcal{F}_{\mathrm{ind}} := \{f_i : \ \mathrm{DRP}(i) \geq \tau_D \ \text{and} \ \mathrm{DEP}(i) < \tau_P\}.$$

The resulting partition is called the *Fuzzy MICMAC classification* induced by $(\widetilde{R}, \mathrm{Score}, \theta, \tau_D, \tau_P)$. (Driving/dependence-based grouping is the standard purpose of fuzzy MICMAC in ISM-based studies.)

**Theorem 5.3.2** (Well-definedness of Fuzzy MICMAC). *Under Definition 5.3.1, assume:*

- *$\mathcal{F}$ is finite with $n \geq 2$;*

- Score : $\mathbb{F} \to [0,1]$ *is well-defined on $\mathbb{F}$;*

- $\theta \in (0,1]$ *and* $(\tau_D, \tau_P) \in \mathbb{R}^2$ *are fixed.*

*Then:*

1. *The matrices $R$ and $D$ are well-defined, with $R \in [0,1]^{n \times n}$ and $D \in \{0,1\}^{n \times n}$.*

2. *For each $i$, $\mathrm{DRP}(i)$ and $\mathrm{DEP}(i)$ are well-defined integers satisfying*

$$0 \leq \mathrm{DRP}(i) \leq n, \qquad 0 \leq \mathrm{DEP}(i) \leq n.$$

3. *The four sets $\mathcal{F}_{\mathrm{aut}}, \mathcal{F}_{\mathrm{dep}}, \mathcal{F}_{\mathrm{lin}}, \mathcal{F}_{\mathrm{ind}}$ form a partition of $\mathcal{F}$, i.e., they are pairwise disjoint and their union equals $\mathcal{F}$.*

*Proof.* (1) Since Score maps every $\tilde{r}_{ij} \in \mathbb{F}$ into $[0,1]$, each entry $r_{ij}$ is a well-defined real number in $[0,1]$. With $\theta \in (0,1]$, the indicator $d_{ij} = \mathbf{1}[r_{ij} \geq \theta]$ is well-defined and lies in $\{0,1\}$ for every $(i,j)$. Hence $R$ and $D$ are well-defined matrices of the stated types.

(2) Because each $d_{ij} \in \{0,1\}$ and there are exactly $n$ terms in each sum, $\mathrm{DRP}(i) = \sum_{j=1}^{n} d_{ij}$ and $\mathrm{DEP}(i) = \sum_{j=1}^{n} d_{ji}$ are well-defined integers with bounds $0 \leq \mathrm{DRP}(i) \leq n$ and $0 \leq \mathrm{DEP}(i) \leq n$.

(3) Fix any $f_i \in \mathcal{F}$. Exactly one of the two inequalities $\mathrm{DRP}(i) < \tau_D$ or $\mathrm{DRP}(i) \geq \tau_D$ holds, and exactly one of $\mathrm{DEP}(i) < \tau_P$ or $\mathrm{DEP}(i) \geq \tau_P$ holds. Therefore, $f_i$ belongs to exactly one of the four sets defined in Step 4. This shows $\mathcal{F}_{\mathrm{aut}} \cup \mathcal{F}_{\mathrm{dep}} \cup \mathcal{F}_{\mathrm{lin}} \cup \mathcal{F}_{\mathrm{ind}} = \mathcal{F}$. Moreover, no $f_i$ can satisfy two incompatible pairs of inequalities simultaneously, so the four sets are pairwise disjoint. Thus they form a partition of $\mathcal{F}$. $\qquad \square$

We now extend this idea to a general uncertain model $M$.

**Definition 5.3.3** (Uncertain MICMAC of type $M$)**.** Let

$$\mathcal{F} = \{f_1, \ldots, f_n\}$$

be a finite nonempty set of factors, where $n \geq 1$. Fix an uncertain model $M$ with degree-domain

$$\mathrm{Dom}(M) \subseteq [0,1]^k$$

for some integer $k \geq 1$.

Assume that an *uncertain reachability matrix* of type $M$ is given:

$$\widetilde{R}_M = (\rho_{ij})_{n \times n} \in \mathrm{Dom}(M)^{n \times n},$$

where $\rho_{ij} \in \mathrm{Dom}(M)$ represents the uncertain reachability degree from factor $f_i$ to factor $f_j$. Typically, $\widetilde{R}_M$ is obtained from an Uncertain ISM procedure.

Fix further:

- a total score map
$$\mathrm{Score}_M : \mathrm{Dom}(M) \longrightarrow [0,1],$$

- a threshold
$$\theta \in (0,1].$$

**Step 1 (Scored reachability matrix).** Define the scored reachability matrix
$$R = (r_{ij})_{n \times n} \in [0,1]^{n \times n}$$
by
$$r_{ij} := \mathrm{Score}_M(\rho_{ij}), \qquad i,j = 1, \ldots, n.$$

**Step 2 (Binary reachability matrix).** Define the binary reachability matrix
$$D = (d_{ij})_{n \times n} \in \{0,1\}^{n \times n}$$
by thresholding:
$$d_{ij} := \begin{cases} 1, & r_{ij} \geq \theta, \\ 0, & r_{ij} < \theta. \end{cases}$$

**Step 3 (Driving power and dependence power).** For each factor $f_i \in \mathcal{F}$, define its *driving power* and *dependence power* by
$$\mathrm{Drv}_M(i) := \sum_{j=1}^n d_{ij}, \qquad \mathrm{Dep}_M(i) := \sum_{j=1}^n d_{ji}.$$
Thus $\mathrm{Drv}_M(i)$ is the row sum of $D$, and $\mathrm{Dep}_M(i)$ is the column sum of $D$.

**Step 4 (MICMAC cutoffs).** Define the average driving-power and dependence-power cutoffs by
$$\tau_{\mathrm{drv}} := \frac{1}{n} \sum_{i=1}^n \mathrm{Drv}_M(i), \qquad \tau_{\mathrm{dep}} := \frac{1}{n} \sum_{i=1}^n \mathrm{Dep}_M(i).$$

**Step 5 (Quadrant-based classification).** Define the four MICMAC classes by
$$\mathcal{F}_{\mathrm{aut}} := \big\{ f_i \in \mathcal{F} : \mathrm{Drv}_M(i) < \tau_{\mathrm{drv}} \text{ and } \mathrm{Dep}_M(i) < \tau_{\mathrm{dep}} \big\},$$
$$\mathcal{F}_{\mathrm{dep}} := \big\{ f_i \in \mathcal{F} : \mathrm{Drv}_M(i) < \tau_{\mathrm{drv}} \text{ and } \mathrm{Dep}_M(i) \geq \tau_{\mathrm{dep}} \big\},$$
$$\mathcal{F}_{\mathrm{lin}} := \big\{ f_i \in \mathcal{F} : \mathrm{Drv}_M(i) \geq \tau_{\mathrm{drv}} \text{ and } \mathrm{Dep}_M(i) \geq \tau_{\mathrm{dep}} \big\},$$
$$\mathcal{F}_{\mathrm{drv}} := \big\{ f_i \in \mathcal{F} : \mathrm{Drv}_M(i) \geq \tau_{\mathrm{drv}} \text{ and } \mathrm{Dep}_M(i) < \tau_{\mathrm{dep}} \big\}.$$
These are called, respectively, the *autonomous*, *dependent*, *linkage*, and *driving* classes.

The quadruple
$$\big( \mathcal{F}_{\mathrm{aut}}, \mathcal{F}_{\mathrm{dep}}, \mathcal{F}_{\mathrm{lin}}, \mathcal{F}_{\mathrm{drv}} \big)$$
is called the *Uncertain MICMAC classification of type $M$* induced by
$$\big( \widetilde{R}_M, \mathrm{Score}_M, \theta \big).$$

**Remark 5.3.4.** Definition 5.3.3 is model-independent. The uncertain model $M$ only enters through the degree-domain $\mathrm{Dom}(M)$ and the score map $\mathrm{Score}_M$, which converts uncertain reachability values into scalar strengths in $[0,1]$.

**Theorem 5.3.5** (Well-definedness of Uncertain MICMAC). *Under Definition 5.3.3, assume:*

*(A1)* $\mathcal{F} = \{f_1, \ldots, f_n\}$ *is finite and nonempty;*

*(A2)* $\widetilde{R}_M = (\rho_{ij})_{n \times n} \in \mathrm{Dom}(M)^{n \times n}$;

*(A3)* $\mathrm{Score}_M : \mathrm{Dom}(M) \to [0,1]$ *is a total map;*

*(A4)* $\theta \in (0,1]$.

*Then:*

*(i) the scored matrix $R = (r_{ij}) \in [0,1]^{n \times n}$ and the binary matrix $D = (d_{ij}) \in \{0,1\}^{n \times n}$ are well-defined;*

*(ii) for every $i \in \{1, \ldots, n\}$, the quantities $\mathrm{Drv}_M(i)$ and $\mathrm{Dep}_M(i)$ are well-defined integers satisfying*

$$0 \leq \mathrm{Drv}_M(i) \leq n, \qquad 0 \leq \mathrm{Dep}_M(i) \leq n;$$

*(iii) the cutoffs $\tau_{\mathrm{drv}}$ and $\tau_{\mathrm{dep}}$ are well-defined real numbers;*

*(iv) the four classes*

$$\mathcal{F}_{\mathrm{aut}}, \quad \mathcal{F}_{\mathrm{dep}}, \quad \mathcal{F}_{\mathrm{lin}}, \quad \mathcal{F}_{\mathrm{drv}}$$

*are well-defined and form a partition of $\mathcal{F}$, that is, they are pairwise disjoint and their union is $\mathcal{F}$.*

*Hence Uncertain MICMAC of type $M$ is well-defined.*

*Proof.* **(i) Well-definedness of $R$ and $D$.** By (A2), each entry $\rho_{ij}$ belongs to $\mathrm{Dom}(M)$. Since $\mathrm{Score}_M$ is total by (A3), the value

$$r_{ij} := \mathrm{Score}_M(\rho_{ij})$$

is well-defined and belongs to $[0,1]$ for every $i, j$. Therefore

$$R = (r_{ij})_{n \times n} \in [0,1]^{n \times n}$$

is well-defined.

Because $\theta \in (0,1]$, for each $r_{ij} \in [0,1]$ exactly one of the inequalities

$$r_{ij} \geq \theta \qquad \text{or} \qquad r_{ij} < \theta$$

holds. Hence each $d_{ij}$ is well-defined and belongs to $\{0,1\}$. Therefore

$$D = (d_{ij})_{n \times n} \in \{0,1\}^{n \times n}$$

is well-defined.

**(ii) Well-definedness of driving and dependence powers.** For each fixed $i$, the sums

$$\mathrm{Drv}_M(i) = \sum_{j=1}^{n} d_{ij}, \qquad \mathrm{Dep}_M(i) = \sum_{j=1}^{n} d_{ji}$$

contain exactly $n$ terms, each equal to 0 or 1. Therefore both are well-defined integers, and clearly

$$0 \le \mathrm{Drv}_M(i) \le n, \qquad 0 \le \mathrm{Dep}_M(i) \le n.$$

**(iii) Well-definedness of the cutoffs.** Since each $\mathrm{Drv}_M(i)$ and $\mathrm{Dep}_M(i)$ is a real number, the averages

$$\tau_{\mathrm{drv}} = \frac{1}{n} \sum_{i=1}^{n} \mathrm{Drv}_M(i), \qquad \tau_{\mathrm{dep}} = \frac{1}{n} \sum_{i=1}^{n} \mathrm{Dep}_M(i)$$

are well-defined real numbers.

**(iv) Partition property of the four classes.** Fix any factor $f_i \in \mathcal{F}$. Exactly one of the two relations

$$\mathrm{Drv}_M(i) < \tau_{\mathrm{drv}} \qquad \text{or} \qquad \mathrm{Drv}_M(i) \ge \tau_{\mathrm{drv}}$$

holds, and exactly one of the two relations

$$\mathrm{Dep}_M(i) < \tau_{\mathrm{dep}} \qquad \text{or} \qquad \mathrm{Dep}_M(i) \ge \tau_{\mathrm{dep}}$$

holds. Hence exactly one of the following four combinations is true:

$$\mathrm{Drv}_M(i) < \tau_{\mathrm{drv}} \text{ and } \mathrm{Dep}_M(i) < \tau_{\mathrm{dep}},$$
$$\mathrm{Drv}_M(i) < \tau_{\mathrm{drv}} \text{ and } \mathrm{Dep}_M(i) \ge \tau_{\mathrm{dep}},$$
$$\mathrm{Drv}_M(i) \ge \tau_{\mathrm{drv}} \text{ and } \mathrm{Dep}_M(i) \ge \tau_{\mathrm{dep}},$$
$$\mathrm{Drv}_M(i) \ge \tau_{\mathrm{drv}} \text{ and } \mathrm{Dep}_M(i) < \tau_{\mathrm{dep}}.$$

Therefore $f_i$ belongs to exactly one of

$$\mathcal{F}_{\mathrm{aut}}, \ \mathcal{F}_{\mathrm{dep}}, \ \mathcal{F}_{\mathrm{lin}}, \ \mathcal{F}_{\mathrm{drv}}.$$

This shows both that the union of the four classes is $\mathcal{F}$ and that they are pairwise disjoint.

Hence the four classes form a partition of $\mathcal{F}$, and Uncertain MICMAC is well-defined. $\quad\square$

Related concepts of MICMAC under uncertainty-aware models are listed in Table 5.4.

Table 5.4: Related concepts of MICMAC under uncertainty-aware models.

| $k$ | Related MICMAC concept(s) |
|---|---|
| 1 | Fuzzy MICMAC |
| 2 | Intuitionistic Fuzzy MICMAC |
| 3 | Neutrosophic MICMAC [696, 697] |

## 5.4  Fuzzy Cognitive Map (FCM)

A cognitive map is a signed directed graph of causal concepts; updating node states via weighted sums simulates system behavior and feedback loops [698, 699]. Fuzzy Cognitive Maps use fuzzy causal weights and fuzzy concept activations, apply nonlinear update functions, and model uncertain feedback dynamics for scenario evaluation [700–702].

**Definition 5.4.1** (Fuzzy Cognitive Map (FCM)). [700, 701] Let $N \geq 1$ and let $\mathcal{C} = \{C_1, \ldots, C_N\}$ be a finite set of *concepts* (states/variables/features of the modeled system). A *Fuzzy Cognitive Map (FCM)* is a weighted directed graph (equivalently, a weight matrix)

$$\mathsf{FCM} = (\mathcal{C}, W, f),$$

where:

(i) $W = (w_{ji}) \in [-1, 1]^{N \times N}$ is the *causal connection (adjacency) matrix* with $w_{ii} = 0$. The entry $w_{ji}$ represents the signed strength of the causal influence $C_j \to C_i$:

$$w_{ji} > 0 \text{ (positive causality)}, \qquad w_{ji} < 0 \text{ (negative causality)}, \qquad w_{ji} = 0 \text{ (no direct relation)}.$$

(ii) $f : \mathbb{R} \to I$ is an *activation (transfer) function* mapping to a bounded interval $I = [0, 1]$ (unipolar) or $I = [-1, 1]$ (bipolar), typically monotone and continuous (e.g. sigmoid).

**FCM state and reasoning dynamics.** The *state* of the FCM at discrete time $k \in \mathbb{N}_0$ is a vector

$$A(k) = (A_1(k), \ldots, A_N(k)) \in I^N,$$

where $A_i(k)$ is the activation level of concept $C_i$. Given an initial state $A(0) \in I^N$, the standard synchronous FCM update rule is

$$A_i(k+1) = f\left( A_i(k) + \sum_{\substack{j=1 \\ j \neq i}}^{N} A_j(k)\, w_{ji} \right), \qquad i = 1, \ldots, N. \tag{5.1}$$

Equivalently, with componentwise application of $f$,

$$A(k+1) = f\big(A(k) + A(k)W\big).$$

(Variants omit the self-memory term $A_i(k)$ and use $A(k+1) = f(A(k)W)$.)

**Common choice of $f$ (unipolar sigmoid).** A widely used activation function is the logistic sigmoid

$$f_\lambda(x) = \frac{1}{1 + e^{-\lambda x}}, \qquad \lambda > 0,$$

which ensures $A_i(k) \in [0,1]$ for all $k$.

**Outcome.** Iterating (5.1) generates a trajectory in $I^N$ that may converge to a fixed point, enter a limit cycle, or exhibit more complex attractors depending on $A(0)$, $W$, and $f$.

Below, we present the Uncertain Cognitive Map obtained by extending the framework using Uncertain Sets.

**Definition 5.4.2** (Uncertain Cognitive Map (UCM): expected-weight dynamics). Let $N \geq 1$ and let $\mathcal{C} = \{C_1, \ldots, C_N\}$ be a finite set of concepts. Fix an uncertainty space $(\Gamma, \mathcal{L}, \mathcal{M})$ and a bounded interval $I = [0,1]$ (unipolar) or $I = [-1,1]$ (bipolar).

**(1) Uncertain causal weights.** An *uncertain cognitive map* is specified by a matrix of uncertain numbers

$$\widetilde{W} = (\widetilde{w}_{ji}) \in \mathsf{US}([-1,1])^{N \times N}, \qquad \widetilde{w}_{ii} \equiv 0,$$

where $\widetilde{w}_{ji}$ encodes the uncertain signed causal influence $C_j \to C_i$. Assume the expectations exist and define the *expected weight matrix*

$$W := (w_{ji}) \in [-1,1]^{N \times N}, \qquad w_{ji} := \mathbb{E}[\widetilde{w}_{ji}].$$

**(2) State space and activation function.** Let $f : \mathbb{R} \to I$ be an activation (transfer) function such that

$$f(\mathbb{R}) \subseteq I.$$

(Examples include the unipolar logistic sigmoid or a bipolar tanh-type map.)

**(3) Expected-value UCM dynamics.** A *UCM state* at discrete time $k \in \mathbb{N}_0$ is a vector

$$A(k) = (A_1(k), \ldots, A_N(k)) \in I^N.$$

Given $A(0) \in I^N$, define the synchronous update rule by

$$A_i(k+1) = f\left(A_i(k) + \sum_{\substack{j=1 \\ j \neq i}}^{N} A_j(k)\, w_{ji}\right), \qquad i = 1, \ldots, N. \tag{5.2}$$

Equivalently,

$$A(k+1) = f\big(A(k) + A(k)W\big),$$

where $f$ is applied componentwise. The triple $\mathsf{UCM} := (\mathcal{C}, \widetilde{W}, f)$ is called an *Uncertain Cognitive Map*.

**Definition 5.4.3** (Uncertain-set output induced by a UCM). Let $A(0) \in I^N$ be fixed and let $(A(k))_{k \geq 0}$ be generated by (5.2). For each $k \in \mathbb{N}_0$ and each concept $i$, define the singleton-valued uncertain set

$$\mathcal{A}_{i,k} : \Gamma \to \mathcal{P}(I), \qquad \mathcal{A}_{i,k}(\gamma) := \{A_i(k)\}.$$

Then $\big(\mathcal{A}_{i,k}\big)_{i=1,\ldots,N}^{k \in \mathbb{N}_0}$ is called the *uncertain-set state family* induced by the UCM (via expected-value realization).

**Theorem 5.4.4** (Uncertain-set structure and well-definedness of UCM dynamics). *In the setting of Definition 5.4.2, assume:*

(A1) *(Finite expectations)* $\mathbb{E}[\widetilde{w}_{ji}]$ *exists and is finite for all* $i, j$.

(A2) *(Bounded transfer)* $f(\mathbb{R}) \subseteq I$.

*Then, for every initial state* $A(0) \in I^N$:

(i) *the recursion (5.2) defines a unique sequence* $(A(k))_{k \geq 0}$ *in* $I^N$;

(ii) *each* $A_i(k) \in I$ *is well-defined for all* $i$ *and all* $k$;

(iii) *the family in Definition 5.4.3 is an uncertain-set structured output on* $(\Gamma, \mathcal{L}, \mathcal{M})$.

*Proof.* Fix $A(0) \in I^N$.

**Step 1 (existence of the expected weight matrix).** By (A1), each $w_{ji} := \mathbb{E}[\widetilde{w}_{ji}]$ exists as a real number, hence $W = (w_{ji})$ is a well-defined real matrix.

**Step 2 (one-step update is well-defined).** Assume $A(k) \in I^N$. For each $i$, define the real input

$$u_i(k) := A_i(k) + \sum_{j \neq i} A_j(k)\, w_{ji} \in \mathbb{R},$$

which is well-defined because it is a finite sum of real products. Then (A2) implies $A_i(k+1) := f(u_i(k)) \in I$. Therefore $A(k+1) \in I^N$.

**Step 3 (existence and uniqueness for all times).** Step 2 shows that the map $F : I^N \to I^N$ defined by $F(A) = f(A + AW)$ (componentwise $f$) is a well-defined function. Hence the recursion $A(k+1) = F(A(k))$ produces a unique sequence $(A(k))_{k \geq 0}$ in $I^N$ by deterministic iteration of a function.

**Step 4 (uncertain-set structured output).** For each $(i, k)$, $\mathcal{A}_{i,k}(\gamma) = \{A_i(k)\}$ is a constant singleton-valued set map, hence it is an uncertain set on $(\Gamma, \mathcal{L}, \mathcal{M})$. This proves (iii). $\square$

As related concepts beyond Uncertain Cognitive Maps, Grey Cognitive Maps [717, 718], Rough Cognitive Maps [719, 720], Evidential Cognitive Maps [721, 722], Dynamic Cognitive Maps [723, 724], Granular Cognitive Maps [725, 726], Cognitive HyperMaps [727–729], Bipolar Cognitive Maps [730, 731], Linguistic Cognitive Maps [732], and Probabilistic Cognitive Maps [733, 734] are also well known.

Table 5.5: A compact catalogue of Cognitive Map variants by the uncertainty encoding (indexed by a convenient degree-domain dimension $k$).

| $k$ | Cognitive Map variant | Remark (how uncertainty is represented) |
|---|---|---|
| 1 | Fuzzy Cognitive Maps (FCM) [703, 704] | Causal influences (edge weights) and/or concept activations are modeled by single graded degrees (typically normalized to $[0, 1]$ when needed) and updated by nonlinear dynamical rules. |
| 2 | Intuitionistic Fuzzy Cognitive Maps [705, 706] | Each influence is encoded by a two-component degree (e.g., membership and non-membership, with implicit hesitation), yielding a richer uncertainty description than scalar weights. |
| 2 | Pythagorean Fuzzy Cognitive Maps (PFCM) [707, 708] | Pythagorean Fuzzy Cognitive Maps (PFCM) (degrees $(\mu, \nu) \in [0, 1]^2$ with $\mu^2 + \nu^2 \leq 1$; hesitation $\pi = \sqrt{1 - \mu^2 - \nu^2}$ is derived). |
| 3 | Hesitant Fuzzy Cognitive Maps [709, 710] | Each influence is assessed by multiple plausible degrees (a finite hesitation set); for catalogue purposes this is often summarized into a fixed-length $k$=3 encoding (e.g., min/representative/max). |
| 3 | Neutrosophic Cognitive Maps [711–714] | Each influence carries truth/indeterminacy/falsity components in $[0, 1]^3$, enabling explicit representation of indeterminacy and inconsistency in causal strengths. |
| $3n$ ($n \geq$ 1) | Refined Neutrosophic Cognitive Maps [704, 714] | **Refined Neutrosophic Cognitive Maps (R-NCMs):** degrees in $[0, 1]^{3n}$, typically $(T_1, \ldots, T_n, \ I_1, \ldots, I_n, \ F_1, \ldots, F_n)$ for each concept/edge. |
| $n$ | Plithogenic Cognitive Maps [715, 716] | Influences are attribute–value based with scalar appurtenance degrees on attribute–value pairs, coupled with a contradiction function on values; aggregation uses plithogenic operators. |

# Chapter 6

# Compensatory scoring methods on a decision matrix ("weighted-sum" families)

Compensatory scoring aggregates weighted, normalized criterion performances into overall utility scores, allowing trade-offs where strong performance compensates weak criteria, producing rankings via sums/products.

For convenience, a concise comparison of the representative decision-matrix methods discussed in this chapter is presented in Table 6.1.

Table 6.1: A concise comparison of representative decision-matrix methods discussed in this chapter.

| Method | Primary role | Core mechanism | Typical final quantity | Short note |
|---|---|---|---|---|
| COPRAS | Utility ranking | Uses weighted normalized performances, separates benefit and cost sums, and computes relative significance. | Relative significance and utility degree | Explicit benefit–cost decomposition. |
| MAC-BETH | Value-scale construction and ranking | Converts qualitative pairwise attractiveness judgments into a numerical value scale, usually via linear programming. | Cardinal value scale / overall scores | Strongly judgment-driven and scale-oriented. |
| CoCoSo | Compromise ranking | Combines weighted-sum and weighted-product logics into a compromise aggregation. | Combined compromise score | Hybrid SAW–WPM style method. |
| SAW | Utility ranking | Aggregates normalized criterion performances by a simple weighted sum. | Weighted-sum utility | Canonical compensatory baseline. |
| RAFSI | Utility ranking | Maps criterion values to a common interval using ideal and anti-ideal anchors, then aggregates them. | RAFSI score | Emphasizes common-interval functional mapping. |

*Table 6.1 (continued).*

| Method | Primary role | Core mechanism | Typical final quantity | Short note |
|--------|-------------|----------------|------------------------|------------|
| RATMI | Utility ranking | Combines a trace-based performance term with a median-similarity term after weighting. | Trace-to-median index | Matrix-geometry flavored ranking rule. |
| RANCOM | Criteria weighting | Ranks criteria, builds a comparison matrix from rank relations, and normalizes row sums. | Criterion-weight vector | Mainly a weighting tool rather than a direct alternative-ranking method. |
| AROMAN | Utility ranking | Uses two-step normalization, separates benefit and cost effects, and applies an exponential preference score. | Ranking index | Designed to balance two normalization views. |
| MAUT | Utility ranking | Applies single-attribute utility functions and combines them into an overall multi-attribute utility. | Overall utility | Preference-theoretic and utility-based. |
| SMART | Utility ranking | Uses value functions and swing-type weights, then computes a simple weighted additive score. | SMART score | Transparent and easy to implement. |
| REGIME | Pairwise ranking | Compares alternatives pairwise on each criterion, aggregates weighted win–loss indicators, and forms net dominance. | Net guide index | Comparison-based rather than purely direct scoring. |
| TODIM | Dominance ranking | Uses prospect-theoretic gains and losses relative to a reference and aggregates dominance values. | Overall dominance value | Captures asymmetric treatment of losses. |
| GRA | Similarity ranking | Measures closeness to an ideal or reference sequence via grey relational coefficients and grades. | Grey relational grade | Reference-similarity based. |
| ARAS | Utility ratio ranking | Aggregates weighted normalized performances relative to an explicitly defined ideal alternative. | Utility degree | Ideal-alternative baseline is central. |
| WASPAS | Compromise ranking | Combines weighted-sum and weighted-product utilities through a mixing parameter. | Integrated WASPAS utility | Robust WSM–WPM hybrid. |
| MOORA | Ratio ranking | Uses vector normalization and computes a weighted benefit-minus-cost score. | MOORA score | Simple ratio-analysis family. |
| PSI | Objective weighting and ranking | Derives criterion weights from dispersion of normalized data and then computes a preference index. | Preference selection index | More data-driven than subjective weighting methods. |
| ROV | Interval utility ranking | Forms pessimistic and optimistic weighted utilities and ranks by an attitude-dependent score. | Value interval / attitude score | Naturally supports optimistic–pessimistic evaluation. |

*Table 6.1 (continued).*

| Method | Primary role | Core mechanism | Typical final quantity | Short note |
|---|---|---|---|---|
| MOOSRA | Ratio ranking | Computes a weighted benefit-to-cost ratio after normalization. | MOOSRA score | Direct benefit/cost ratio form. |

**Note.** Although all methods are discussed in the same general decision-matrix context, they are not identical in role. In particular, RANCOM is primarily a criterion-weighting method, while MACBETH is fundamentally a value-scale construction framework that can subsequently support ranking.

## 6.1 Fuzzy COPRAS (Fuzzy Complex Proportional Assessment)

COPRAS ranks alternatives using weighted normalized criteria, separately aggregating benefit and cost sums to compute relative significance and utility [618, 735, 736]. Fuzzy COPRAS models ratings and weights as fuzzy numbers, defuzzifies or compares them, then computes COPRAS significance and utility under uncertainty [737, 738].

**Definition 6.1.1** (Fuzzy COPRAS (COPRAS-F))**.** [737, 738] Let $\mathcal{A} = \{A_1, \ldots, A_m\}$ be alternatives and $\mathcal{C} = \{C_1, \ldots, C_n\}$ criteria. Partition criteria into benefit and cost types

$$\mathcal{C} = \mathcal{C}^+ \,\dot{\cup}\, \mathcal{C}^-, \qquad J^+ := \{j : C_j \in \mathcal{C}^+\}, \ \ J^- := \{j : C_j \in \mathcal{C}^-\}.$$

Assume a TFN decision matrix $\tilde{D} = (\tilde{x}_{ij}) \in (\mathsf{TFN})^{m \times n}$ and TFN criterion weights $\tilde{\mathbf{w}} = (\tilde{w}_1, \ldots, \tilde{w}_n) \in (\mathsf{TFN})^n$.

**(0) BNP defuzzification.** For a TFN $\tilde{x} = (l, m, u)$ define

$$\mathrm{BNP}(\tilde{x}) := \frac{l + m + u}{3}.$$

**(1) Defuzzify and (optionally) normalize weights.** Set

$$x_{ij} := \mathrm{BNP}(\tilde{x}_{ij}), \qquad q_j := \mathrm{BNP}(\tilde{w}_j),$$

and (optionally) normalize $q$ by $q_j \leftarrow q_j / \sum_{t=1}^n q_t$ so that $\sum_{j=1}^n q_j = 1$.

**(2) Column-sum normalization and weighting.** Define

$$n_{ij} := \frac{x_{ij}}{\sum_{p=1}^m x_{pj}}, \qquad \hat{x}_{ij} := q_j\, n_{ij}.$$

**(3) Benefit/cost sums.** For each alternative $A_i$, set

$$P_i := \sum_{j \in J^+} \hat{x}_{ij}, \qquad R_i := \sum_{j \in J^-} \hat{x}_{ij}, \qquad R_{\min} := \min_{1 \le i \le m} R_i.$$

**(4) Relative significance and utility degree.** Define

$$Q_i := P_i + \frac{R_{\min} \sum_{t=1}^{m} R_t}{R_i \sum_{t=1}^{m} \left(\frac{R_{\min}}{R_t}\right)}, \qquad Q_{\max} := \max_{1 \le i \le m} Q_i, \qquad N_i := \frac{Q_i}{Q_{\max}} \times 100\%.$$

The final ranking is obtained by sorting $Q_i$ (equivalently $N_i$) in descending order.

Using Uncertain Sets, we define Uncertain COPRAS (U-COPRAS) as follows.

**Definition 6.1.2** (Uncertain COPRAS (U-COPRAS): expected-score formulation). Let $\mathcal{A} = \{A_1, \ldots, A_m\}$ be a finite set of alternatives and $\mathcal{C} = \{C_1, \ldots, C_n\}$ a finite set of criteria. Partition criteria into benefit and cost types

$$\mathcal{C} = \mathcal{C}^+ \,\dot\cup\, \mathcal{C}^-, \qquad J^+ := \{j : C_j \in \mathcal{C}^+\}, \quad J^- := \{j : C_j \in \mathcal{C}^-\}.$$

Fix an uncertainty space $(\Gamma, \mathcal{L}, \mathcal{M})$.

**(1) Uncertain decision matrix and uncertain weights.** An *uncertain COPRAS instance* consists of:

- an *uncertain decision matrix*
$$\widetilde{X} = (\widetilde{x}_{ij}) \in \mathsf{US}(\mathbb{R}_{\ge 0})^{m \times n},$$
where $\widetilde{x}_{ij}$ is the uncertain performance of $A_i$ under criterion $C_j$;

- an *uncertain weight vector*
$$\widetilde{w} = (\widetilde{w}_1, \ldots, \widetilde{w}_n) \in \mathsf{US}(\mathbb{R}_{\ge 0})^n,$$
where $\widetilde{w}_j$ is the uncertain importance of criterion $C_j$.

Assume all expectations exist and define the *expected (crisp) scores*

$$x_{ij} := \mathbb{E}[\widetilde{x}_{ij}] \in \mathbb{R}_{\ge 0}, \qquad q_j := \mathbb{E}[\widetilde{w}_j] \in \mathbb{R}_{\ge 0}.$$

**(2) Normalized expected weights.** If $\sum_{t=1}^{n} q_t > 0$, set

$$\bar{q}_j := \frac{q_j}{\sum_{t=1}^{n} q_t} \in [0, 1] \quad (j = 1, \ldots, n),$$

so that $\sum_{j=1}^{n} \bar{q}_j = 1$. If $\sum_{t=1}^{n} q_t = 0$, set $\bar{q}_j := 1/n$ for all $j$.

**(3) Column-sum normalization and weighting (on expected data).** For each criterion $j$, let

$$s_j := \sum_{p=1}^{m} x_{pj}.$$

Assume $s_j > 0$ for all $j$ (non-degenerate criteria). Define

$$n_{ij} := \frac{x_{ij}}{s_j} \in [0, 1], \qquad \hat{x}_{ij} := \bar{q}_j \, n_{ij} \in [0, 1].$$

**(4) Benefit/cost aggregated sums.** For each alternative $A_i$, define

$$P_i := \sum_{j \in J^+} \hat{x}_{ij}, \qquad R_i := \sum_{j \in J^-} \hat{x}_{ij}.$$

If $J^- \neq \varnothing$, also define

$$R_{\min} := \min_{1 \leq i \leq m} R_i.$$

**(5) Relative significance and utility degree.** If $J^- = \varnothing$, set $Q_i := P_i$. If $J^- \neq \varnothing$ and $R_i > 0$ for all $i$, define

$$Q_i := P_i + \frac{R_{\min} \sum_{t=1}^m R_t}{R_i \sum_{t=1}^m \left( \frac{R_{\min}}{R_t} \right)}.$$

Let $Q_{\max} := \max_{1 \leq i \leq m} Q_i$ and, if $Q_{\max} > 0$, define the utility percentage

$$N_i := \frac{Q_i}{Q_{\max}} \times 100\%.$$

The U-COPRAS ranking is obtained by sorting $Q_i$ (equivalently $N_i$) in descending order.

**Definition 6.1.3** (Uncertain-set output induced by U-COPRAS). Under Definition 6.1.2, define the singleton-valued uncertain sets

$$\mathcal{Q}_i : \Gamma \to \mathcal{P}(\mathbb{R}), \qquad \mathcal{Q}_i(\gamma) := \{Q_i\},$$

and (when $Q_{\max} > 0$) the singleton-valued uncertain sets

$$\mathcal{N}_i : \Gamma \to \mathcal{P}([0, 100]), \qquad \mathcal{N}_i(\gamma) := \{N_i\}.$$

Then $(\mathcal{Q}_i)_{i=1}^m$ (and $(\mathcal{N}_i)_{i=1}^m$) is called the *uncertain-set structured COPRAS output* (via expected-score realization).

**Theorem 6.1.4** (Uncertain-set structure and well-definedness of U-COPRAS). *In the setting of Definition 6.1.2, assume:*

(A1) *(Finite expectations)* $\mathbb{E}[\tilde{x}_{ij}]$ *and* $\mathbb{E}[\tilde{w}_j]$ *exist and are finite for all* $i, j$.

(A2) *(Non-degenerate columns)* $s_j = \sum_{p=1}^m x_{pj} > 0$ *for all* $j$.

(A3) *(Cost-sum positivity when needed)* *Either* $J^- = \varnothing$, *or else* $R_i > 0$ *for all* $i$.

(A4) *(Nontrivial overall score)* $Q_{\max} > 0$ *whenever* $N_i$ *is formed.*

*Then:*

(i) *all quantities in Steps* (2)–(5) *of Definition 6.1.2 are well-defined real numbers;*

(ii) *the induced ranking preorder* $A_i \succeq A_k \iff Q_i \geq Q_k$ *is well-defined on* $\mathcal{A}$;

(iii) *the outputs* $(\mathcal{Q}_i)$ *and* $(\mathcal{N}_i)$ *in Definition 6.1.3 constitute uncertain sets on* $(\Gamma, \mathcal{L}, \mathcal{M})$.

*Proof.* Assumption (A1) guarantees that $x_{ij} = \mathbb{E}[\widetilde{x}_{ij}]$ and $q_j = \mathbb{E}[\widetilde{w}_j]$ exist as finite real numbers, hence Steps (1)–(2) are well-defined. If $\sum_t q_t > 0$, the normalization $\bar{q}_j = q_j / \sum_t q_t$ is a real number in $[0,1]$; if $\sum_t q_t = 0$, the fallback $\bar{q}_j = 1/n$ is a real number and still satisfies $\sum_j \bar{q}_j = 1$.

By (A2), each $s_j > 0$, so $n_{ij} = x_{ij}/s_j$ is well-defined and lies in $[0,1]$ because $x_{ij} \geq 0$ and $x_{ij} \leq s_j$. Then $\hat{x}_{ij} = \bar{q}_j n_{ij}$ is well-defined and nonnegative, so $P_i = \sum_{j \in J^+} \hat{x}_{ij}$ and $R_i = \sum_{j \in J^-} \hat{x}_{ij}$ are well-defined finite sums.

If $J^- = \varnothing$, then $Q_i = P_i$ is well-defined. Otherwise, by (A3) we have $R_i > 0$ and also $R_t > 0$ for every $t$, hence each fraction $R_{\min}/R_t$ is well-defined and positive, so $\sum_{t=1}^m (R_{\min}/R_t) > 0$. Therefore the COPRAS expression for $Q_i$ is well-defined for all $i$.

The ranking preorder $A_i \succeq A_k \iff Q_i \geq Q_k$ is well-defined because $(Q_i)$ are real numbers. Finally, when $Q_{\max} > 0$ (assumption (A4)), each $N_i = (Q_i/Q_{\max}) \cdot 100$ is a well-defined real number in $[0,100]$. The set maps $\mathcal{Q}_i(\gamma) = \{Q_i\}$ and $\mathcal{N}_i(\gamma) = \{N_i\}$ are singleton-valued uncertain sets on $(\Gamma, \mathcal{L}, \mathcal{M})$, proving (iii). $\qquad\square$

For reference, COPRAS and representative uncertainty-aware variants are listed in Table 6.2.

Table 6.2: COPRAS and representative uncertainty-aware variants (organized by the degree-domain dimension $k$).

| $k$ | Representative COPRAS-type model(s) whose evaluation degrees live in a subset of $[0,1]^k$ |
| --- | --- |
| 1 | Fuzzy COPRAS. |
| 2 | Intuitionistic Fuzzy COPRAS [57, 739]; Pythagorean fuzzy COPRAS [740, 741]. |
| 3 | Hesitant Fuzzy COPRAS [742, 743]; Spherical fuzzy COPRAS [744, 745]; Picture Fuzzy COPRAS [746, 747]; Neutrosophic COPRAS [748, 749]. |

**Note.** Here $k$ denotes the ambient dimension of the degree-domain used to encode each criterion evaluation (and, if applicable, each weight).

## 6.2 Fuzzy MACBETH (Fuzzy Measuring Attractiveness by a Categorical Based Evaluation Technique)

MACBETH elicits qualitative pairwise judgments of attractiveness differences, converts them into numerical value scales via linear programming, and aggregates weighted scores to rank alternatives consistently [750, 751]. Fuzzy MACBETH replaces crisp categories with fuzzy linguistic terms or fuzzy numbers, propagating imprecision through the scale-construction model to obtain fuzzy scores and robust rankings [752–754].

**Definition 6.2.1** (Fuzzy MACBETH (triangular fuzzy semantic scale + F-LP-MACBETH)). [752–754] Let $A = \{a_1, \ldots, a_n\}$ be a finite set of *evaluation elements* (actions, alternatives, or performance levels) that are totally preordered by a decision maker ($P$ = strict preference, $I$ = indifference). Assume $a^+ \in A$ is the most attractive element and $a^- \in A$ is the least attractive element.

**(1) Semantic categories and their triangular fuzzy encoding.** Let the semantic categories be $C_0, C_1, \ldots, C_6$, where $C_0$ denotes *indifference* and $C_1, \ldots, C_6$ denote increasing *differences of attractiveness*. Associate each category $C_k$ with a triangular fuzzy number (TFN)

$$\tilde{A}_k = (\ell_k, m_k, u_k) \in \mathbb{R}^3_{\geq 0}, \qquad \ell_k \leq m_k \leq u_k,$$

given by the standard Fuzzy-MACBETH scale

$$\tilde{A}_0 = (0,0,0), \ \tilde{A}_1 = (1,1,2), \ \tilde{A}_2 = (1,2,3), \ \tilde{A}_3 = (2,3,4),$$

$$\tilde{A}_4 = (3,4,5), \ \tilde{A}_5 = (4,5,6), \ \tilde{A}_6 = (5,6,6).$$

(Equivalently: for $k = 2,3,4,5$, take $\ell_k = k-1$, $m_k = k$, $u_k = k+1$; for $k = 1$, $\ell_1 = m_1 = 1, u_1 = 2$; for $k = 6$, $\ell_6 = 5, m_6 = u_6 = 6$.)

**(2) Fuzzified judgment matrix.** For each ordered pair $(x,y) \in A \times A$ with $xPy$, the decision maker assigns a category $c(x,y) \in \{1,\ldots,6\}$ (or 0 for indifference). Define the fuzzified judgment as $\tilde{d}(x,y) := \tilde{A}_{c(x,y)}$.

**(3) TFN arithmetic and TFN order.** For TFNs $\tilde{p} = (p_1, p_2, p_3)$ and $\tilde{q} = (q_1, q_2, q_3)$ define

$$\tilde{p} \oplus \tilde{q} = (p_1 + q_1, \ p_2 + q_2, \ p_3 + q_3), \qquad \tilde{p} \ominus \tilde{q} = (p_1 - q_1, \ p_2 - q_2, \ p_3 - q_3).$$

Use the componentwise preorder $\succeq$ on TFNs:

$$(\ell_1, m_1, u_1) \succeq (\ell_2, m_2, u_2) \iff \ell_1 \geq \ell_2, \ m_1 \geq m_2, \ u_1 \geq u_2.$$

Let $\mathrm{COA}(\ell, m, u) := (\ell + m + u)/3$ be the centroid defuzzification map.

**(4) F-LP-MACBETH (fuzzy linear program) and fuzzy basic scale.** Decision variables are TFNs $\tilde{v}(x) = (v_x^L, v_x^M, v_x^U)$ for all $x \in A$ (the *pre-cardinal / basic fuzzy scale*). Compute $\tilde{v}$ by solving the following optimization problem:

$$\min \ \mathrm{COA}\big(\tilde{v}(a^+) \ominus \tilde{v}(a^-)\big)$$

subject to the constraints

| | |
|---|---|
| (Origin) | $\tilde{v}(a^-) = (0,0,0),$ |
| (Indifference) | $\tilde{v}(x) \ominus \tilde{v}(y) = (0,0,0) \quad \forall (x,y) \in C_0,$ |
| (Positive separation) | $\tilde{v}(x) \ominus \tilde{v}(y) \succeq (1,1,2) \quad \forall (x,y) \in C_k, \ k \in \{1,\ldots,6\},$ |
| (Cardinal consistency across categories) | $\tilde{v}(x) \ominus \tilde{v}(y) \ \succeq \ \tilde{v}(w) \ominus \tilde{v}(z)$ |
| | $\forall (x,y) \in C_k, \ \forall (w,z) \in C_{k'}, \ k > k', \ k, k' \in \{1,\ldots,6\}.$ |

Any feasible optimizer $\tilde{v}$ is called a *Fuzzy-MACBETH basic (pre-cardinal) scale.*

**(5) Defuzzification and cardinalization.** Define the basic crisp scale $v_x := \mathrm{COA}(\tilde{v}(x))$ for each $x \in A$. To obtain a *cardinal* scale $E_x$, anchor two reference levels: choose a "neutral" level $x^{\mathrm{neu}}$ and a "good" level $x^{\mathrm{good}}$ and impose

$$E_{x^{\mathrm{neu}}} = 0, \qquad E_{x^{\mathrm{good}}} = 100.$$

Then define the affine transformation

$$E_x = \alpha \, v_x + \beta,$$

with $(\alpha, \beta)$ determined by the two anchoring equations above. The resulting mapping $x \mapsto E_x$ is called the *(fuzzy) MACBETH cardinal value function.*

Using Uncertain Sets, we present Uncertain MACBETH (U-MACBETH) as an extension of the original framework below.

**Definition 6.2.2** (Uncertain MACBETH (U-MACBETH): expected-interval LP scale construction)**.** Let $A = \{a_1, \ldots, a_N\}$ be a finite set of evaluation elements (actions/alternatives/levels). Assume a total preorder $\succeq$ on $A$ with strict part $P$ and indifference part $I$:

$$xPy \iff (x \succeq y \text{ and } y \not\succeq x), \qquad xIy \iff (x \succeq y \text{ and } y \succeq x).$$

Let $a^+ \in A$ be a most attractive element and $a^- \in A$ a least attractive element (with $a^+Pa^-$).

**(1) Semantic categories and uncertain numerical bounds.** Let the MACBETH semantic categories be $C_0, C_1, \ldots, C_6$, where $C_0$ denotes *indifference* and $C_1, \ldots, C_6$ denote increasing *differences of attractiveness*. Fix an uncertainty space $(\Gamma, \mathcal{L}, \mathcal{M})$. For each category $k \in \{0, \ldots, 6\}$, assign an *uncertain interval bound pair*

$$(\widetilde{L}_k, \widetilde{U}_k) \in \mathsf{US}(\mathbb{R}_{\geq 0}) \times \mathsf{US}(\mathbb{R}_{\geq 0}),$$

interpreted as (uncertain) lower/upper bounds on a value difference belonging to category $C_k$. Assume the expected bounds exist and put

$$L_k := \mathbb{E}[\widetilde{L}_k], \qquad U_k := \mathbb{E}[\widetilde{U}_k] \qquad (k = 0, \ldots, 6),$$

with

$$0 = L_0 = U_0, \qquad 0 \leq L_k \leq U_k < \infty \ (k \geq 1), \qquad U_k \leq L_{k+1} \ (k = 0, \ldots, 5)$$

(category separation on expected bounds).

**(2) Uncertain MACBETH judgment matrix.** For each ordered pair $(x, y) \in A \times A$ with $xPy$, the decision maker provides a category label

$$c(x, y) \in \{1, \ldots, 6\}.$$

For indifference pairs $(x, y) \in I$, set $c(x, y) := 0$. This yields a (category-valued) judgment matrix $c : A \times A \to \{0, \ldots, 6\}$.

**(3) Expected-interval linear program (basic value scale).** Decision variables are real values $v(x) \in \mathbb{R}$ for all $x \in A$. Define the feasible set by the expected-interval constraints:

| | | |
|---|---|---|
| (Anchor) | $v(a^-) = 0,$ | |
| (Indifference) | $v(x) - v(y) = 0$ | $\forall (x, y) \in I,$ |
| (Category bounds) | $L_{c(x,y)} \ \leq \ v(x) - v(y) \ \leq \ U_{c(x,y)}$ | $\forall (x, y) \in A \times A$ with $xPy,$ |
| (Monotonicity) | $v(x) \geq v(y)$ | $\forall x, y \in A$ with $x \succeq y.$ |

Among all feasible $v$, define a *basic MACBETH scale* by solving

$$\min \ v(a^+) \quad \text{subject to the above constraints.} \tag{6.1}$$

Any optimizer $v^\star$ of (6.1) is called an *U-MACBETH basic value scale*.

**(4) Cardinalization (0–100 scale) and ranking.** If $v^\star(a^+) > 0$, define the cardinal value function $E : A \to [0, 100]$ by

$$E(x) := 100 \cdot \frac{v^\star(x) - v^\star(a^-)}{v^\star(a^+) - v^\star(a^-)} = 100 \cdot \frac{v^\star(x)}{v^\star(a^+)} \qquad (x \in A).$$

Rank elements by the induced preorder

$$x \succeq_{\text{U-MACBETH}} y \iff E(x) \geq E(y).$$

**Definition 6.2.3** (Uncertain-set structured U-MACBETH output). Under Definition 6.2.2, fix an optimizer $v^\star$ of (6.1) (and the induced $E$ when $v^\star(a^+) > 0$). Define singleton-valued uncertain sets

$$\mathcal{V}_x : \Gamma \to \mathcal{P}(\mathbb{R}), \qquad \mathcal{V}_x(\gamma) := \{v^\star(x)\},$$

and, when $v^\star(a^+) > 0$,

$$\mathcal{E}_x : \Gamma \to \mathcal{P}([0, 100]), \qquad \mathcal{E}_x(\gamma) := \{E(x)\}.$$

Then $(\mathcal{V}_x)_{x \in A}$ (and $(\mathcal{E}_x)_{x \in A}$) is called the *uncertain-set structured U-MACBETH output* (via expected-interval realization).

**Theorem 6.2.4** (Uncertain-set structure and well-definedness of U-MACBETH). *In the setting of Definition 6.2.2, assume:*

(A1) *(Finite expectations)* $\mathbb{E}[\widetilde{L}_k]$ *and* $\mathbb{E}[\widetilde{U}_k]$ *exist and are finite for all* $k = 0, \ldots, 6$.

(A2) *(Separated expected category bounds)* $0 = L_0 = U_0$, $0 \leq L_k \leq U_k$, *and* $U_k \leq L_{k+1}$ *for* $k = 0, \ldots, 5$.

(A3) *(Feasibility)* *The constraint set of* (6.1) *is nonempty.*

(A4) *(Nontrivial top anchor)* *Every feasible* $v$ *satisfies* $v(a^+) > 0$ *(equivalently, the constraints force a strictly positive separation between* $a^+$ *and* $a^-$).

*Then:*

(i) *the LP* (6.1) *is well-defined and attains an optimal solution* $v^\star$;

(ii) *the cardinal values* $E(x)$ *are well-defined real numbers in* $[0, 100]$ *for all* $x \in A$;

(iii) *the ranking preorder* $\succeq_{\text{U-MACBETH}}$ *is well-defined on* $A$;

(iv) *the maps* $\mathcal{V}_x$ *and* $\mathcal{E}_x$ *in Definition 6.2.3 are uncertain sets on* $(\Gamma, \mathcal{L}, \mathcal{M})$.

*Proof.* By (A1) the expected bounds $L_k, U_k$ are finite real numbers. Hence all constraints in Definition 6.2.2 are ordinary linear equalities/inequalities, so (6.1) is a standard linear program.

By (A3) the feasible region is nonempty. The anchor constraint $v(a^-) = 0$ and the monotonicity constraints ensure $v(a^+) \geq 0$ for every feasible $v$ because $a^+ \succeq a^-$. Therefore the objective $\min v(a^+)$ is bounded below by 0. Since the feasible region is a closed polyhedron and the objective is continuous linear, the minimum is attained at some optimizer $v^\star$, proving (i).

By (A4) one has $v^\star(a^+) > 0$, so the normalization $E(x) = 100\, v^\star(x)/v^\star(a^+)$ is well-defined for every $x \in A$. Moreover, $v^\star(a^-) = 0$ and monotonicity implies $0 \leq v^\star(x) \leq v^\star(a^+)$ for all $x \in A$ (because $a^+ \succeq x \succeq a^-$ holds in a total preorder). Hence $E(x) \in [0, 100]$, proving (ii).

The preorder $\succeq_{\text{U-MACBETH}}$ is well-defined because it compares real numbers $E(x)$, proving (iii).

Finally, $\mathcal{V}_x(\gamma) = \{v^\star(x)\}$ and $\mathcal{E}_x(\gamma) = \{E(x)\}$ are singleton-valued set maps from $\Gamma$ into $\mathcal{P}(\mathbb{R})$ (resp. $\mathcal{P}([0, 100])$), hence they are uncertain sets. This proves (iv). $\qquad \square$

Related concepts of MACBETH under uncertainty-aware models are listed in Table 6.3.

Table 6.3: Related concepts of MACBETH under uncertainty-aware models.

| $k$ | Related MACBETH concept(s) |
|---|---|
| 1 | Fuzzy MACBETH |
| 2 | Intuitionistic Fuzzy MACBETH |
| 3 | Neutrosophic MACBETH |

## 6.3 Fuzzy CoCoSo (Fuzzy Combined Compromise Solution)

CoCoSo combines SAW and WEP aggregation to compute compromise scores, producing stable, robust rankings even after alternative changes small perturbations [755, 756]. Fuzzy CoCoSo expresses performances and preferences by triangular fuzzy numbers, normalizes them, aggregates fuzzy sums/powers, then defuzzifies final scores consistently [641, 757].

**Definition 6.3.1** (Fuzzy CoCoSo (TFN-based Combined Compromise Solution)). [641, 757] Let $\mathcal{A} = \{A_1, \ldots, A_m\}$ be alternatives and $\mathcal{C} = \{C_1, \ldots, C_n\}$ criteria. Partition criteria into benefit and cost sets

$$\mathcal{C} = \mathcal{C}^+ \dot\cup \mathcal{C}^-.$$

Let $\mathsf{TFN}_{\geq 0} := \{(l, m, u) \in \mathbb{R}_{\geq 0}^3 : l \leq m \leq u\}$.

**(0) TFN arithmetic (nonnegative convention).** For $\tilde{x} = (l_x, m_x, u_x), \tilde{y} = (l_y, m_y, u_y) \in \mathsf{TFN}_{\geq 0}$ and $c > 0$ define

$$\tilde{x} \oplus \tilde{y} := (l_x + l_y, \ m_x + m_y, \ u_x + u_y), \qquad \tilde{x} \otimes \tilde{y} := (l_x l_y, \ m_x m_y, \ u_x u_y),$$

$$c \odot \tilde{x} := (c l_x, \ c m_x, \ c u_x), \qquad \tilde{x} \oslash c := \left(\frac{l_x}{c}, \frac{m_x}{c}, \frac{u_x}{c}\right).$$

For $w \geq 0$, define the (scalar) TFN power by

$$\tilde{x}^w := (l_x^w, \ m_x^w, \ u_x^w),$$

(with the standard convention $0^0 := 1$ if needed). Fix a defuzzification/score map (centroid)

$$\mathrm{Defuzz}(l, m, u) := \frac{l + m + u}{3}.$$

**(1) Input data (fuzzy decision matrix + weights).** Assume a TFN decision matrix

$$\widetilde{Y} = (\tilde{y}_{ij}) \in (\mathsf{TFN}_{\geq 0})^{m \times n}, \qquad \tilde{y}_{ij} = (l_{ij}, m_{ij}, u_{ij}),$$

and a (crisp) weight vector $w = (w_1, \ldots, w_n)$ with $w_j \geq 0$ and $\sum_{j=1}^n w_j = 1$. (If TFN weights are given, one may first defuzzify and renormalize them to obtain such $w$.)

**(2) Normalization (fuzzy lift of CoCoSo (7)–(8)).** For each criterion $j$, set the global lower/upper scalars

$$\underline{y}_j := \min_{1 \leq i \leq m} l_{ij}, \qquad \overline{y}_j := \max_{1 \leq i \leq m} u_{ij}, \qquad \Delta_j := \overline{y}_j - \underline{y}_j > 0.$$

Define the normalized TFNs $\tilde{t}_{ij} \in \mathsf{TFN}_{\geq 0}$ by

$$\tilde{t}_{ij} := \begin{cases} \left( \dfrac{l_{ij} - \underline{y}_j}{\Delta_j}, \ \dfrac{m_{ij} - \underline{y}_j}{\Delta_j}, \ \dfrac{u_{ij} - \underline{y}_j}{\Delta_j} \right), & C_j \in \mathcal{C}^+ \text{ (benefit)}, \\[3mm] \left( \dfrac{\overline{y}_j - u_{ij}}{\Delta_j}, \ \dfrac{\overline{y}_j - m_{ij}}{\Delta_j}, \ \dfrac{\overline{y}_j - l_{ij}}{\Delta_j} \right), & C_j \in \mathcal{C}^- \text{ (cost)}. \end{cases}$$

Then $0 \leq \tilde{t}_{ij} \leq (1,1,1)$ componentwise and the construction reduces to the crisp normalization when $\tilde{y}_{ij} = (y_{ij}, y_{ij}, y_{ij})$.

### (3) Fuzzy comparability sequences (lift of CoCoSo (9)–(10)). Define

$$\tilde{G}_i := \bigotimes_{j=1}^{n} (\tilde{t}_{ij})^{w_j} \in \mathsf{TFN}_{\geq 0}, \qquad \tilde{H}_i := \bigoplus_{j=1}^{n} (w_j \odot \tilde{t}_{ij}) \in \mathsf{TFN}_{\geq 0}.$$

### (4) Three fuzzy appraisal scores (lift of CoCoSo (11)–(13)). Let

$$S_{GH} := \sum_{p=1}^{m} \mathrm{Defuzz}(\tilde{G}_p \oplus \tilde{H}_p) \ > \ 0, \qquad G_{\min} := \min_p \mathrm{Defuzz}(\tilde{G}_p), \ \ H_{\min} := \min_p \mathrm{Defuzz}(\tilde{H}_p),$$

$$G_{\max} := \max_p \mathrm{Defuzz}(\tilde{G}_p), \ \ H_{\max} := \max_p \mathrm{Defuzz}(\tilde{H}_p).$$

Define

$$\tilde{as}_i^{(\alpha)} := \left( \tilde{G}_i \oplus \tilde{H}_i \right) \oslash S_{GH},$$

$$\tilde{as}_i^{(\beta)} := \left( \tilde{H}_i \oslash H_{\min} \right) \ \oplus \ \left( \tilde{G}_i \oslash G_{\min} \right),$$

and for a parameter $\lambda \in [0,1]$ (often $\lambda = 0.5$),

$$\tilde{as}_i^{(\gamma)} := \left( \lambda \odot \tilde{H}_i \ \oplus \ (1-\lambda) \odot \tilde{G}_i \right) \oslash \left( \lambda H_{\max} + (1-\lambda) G_{\max} \right).$$

### (5) Final fuzzy CoCoSo score (lift of CoCoSo (14)) and ranking. Define the final TFN score

$$\tilde{as}_i := \left( \tilde{as}_i^{(\alpha)} \otimes \tilde{as}_i^{(\beta)} \otimes \tilde{as}_i^{(\gamma)} \right)^{1/3} \ \oplus \ \frac{1}{3} \odot \left( \tilde{as}_i^{(\alpha)} \oplus \tilde{as}_i^{(\beta)} \oplus \tilde{as}_i^{(\gamma)} \right).$$

Defuzzify $AS_i := \mathrm{Defuzz}(\tilde{as}_i) \in \mathbb{R}_{\geq 0}$ and rank alternatives by

$$A_p \succeq_{\mathrm{FCoCoSo}} A_q \quad \Longleftrightarrow \quad AS_p \geq AS_q.$$

Any $A^\star \in \arg\max_{1 \leq i \leq m} AS_i$ is called a *Fuzzy CoCoSo* (compromise) solution.

Using Uncertain Sets, we define Uncertain CoCoSo as follows.

**Definition 6.3.2** (Uncertainty space, uncertain variable, and induced uncertain set). An *uncertainty space* is a triple $(\Gamma, \mathcal{L}, \mathcal{M})$ where $\Gamma \neq \varnothing$, $\mathcal{L}$ is a $\sigma$-algebra on $\Gamma$, and $\mathcal{M} : \mathcal{L} \to [0,1]$ is an uncertain measure. A (real-valued) *uncertain variable* is an $\mathcal{L}$-measurable map $\xi : \Gamma \to \mathbb{R}$. Any uncertain variable $\xi$ induces a singleton-valued *uncertain set*

$$\mathcal{X}_\xi : \Gamma \to \mathcal{P}(\mathbb{R}), \qquad \mathcal{X}_\xi(\gamma) := \{\xi(\gamma)\}.$$

When $\mathbb{E}[\xi] \in \mathbb{R}$ exists (finite), it is used as a crisp score.

**Definition 6.3.3** (Uncertain CoCoSo (U-CoCoSo): scenario-wise CoCoSo with expected-score decision).
Let $\mathcal{A} = \{A_1, \ldots, A_m\}$ be alternatives and $\mathcal{C} = \{C_1, \ldots, C_n\}$ criteria, partitioned as

$$\mathcal{C} = \mathcal{C}^+ \,\dot\cup\, \mathcal{C}^- \quad \text{(benefit/cost)}, \qquad J^+ := \{j : C_j \in \mathcal{C}^+\}, \quad J^- := \{j : C_j \in \mathcal{C}^-\}.$$

Fix an uncertainty space $(\Gamma, \mathcal{L}, \mathcal{M})$.

**(1) Uncertain input data.** An *uncertain CoCoSo instance* consists of

- an *uncertain decision matrix* $\Xi = (\xi_{ij}) \in \mathsf{UV}_{\geq 0}^{m \times n}$, where each $\xi_{ij} : \Gamma \to \mathbb{R}_{\geq 0}$ is an uncertain variable representing the performance of $A_i$ under $C_j$;

- an *uncertain weight vector* $\Omega = (\omega_1, \ldots, \omega_n) \in \mathsf{UV}_{\geq 0}^n$, where $\omega_j : \Gamma \to \mathbb{R}_{\geq 0}$ is an uncertain variable representing the importance of $C_j$;

- a parameter $\lambda \in [0, 1]$ (often $\lambda = \frac{1}{2}$).

Here $\mathsf{UV}_{\geq 0}$ denotes the class of nonnegative uncertain variables on $(\Gamma, \mathcal{L}, \mathcal{M})$.

**(2) Scenario-wise normalized weights.** For each $\gamma \in \Gamma$, define

$$W(\gamma) := \sum_{t=1}^{n} \omega_t(\gamma), \qquad w_j(\gamma) := \frac{\omega_j(\gamma)}{W(\gamma)} \quad (j = 1, \ldots, n),$$

so that $\sum_{j=1}^{n} w_j(\gamma) = 1$ whenever $W(\gamma) > 0$.

**(3) Scenario-wise normalization of performances.** For each criterion $j$ and scenario $\gamma$, set

$$\underline{\xi}_j(\gamma) := \min_{1 \leq i \leq m} \xi_{ij}(\gamma), \qquad \overline{\xi}_j(\gamma) := \max_{1 \leq i \leq m} \xi_{ij}(\gamma), \qquad \Delta_j(\gamma) := \overline{\xi}_j(\gamma) - \underline{\xi}_j(\gamma).$$

Assuming $\Delta_j(\gamma) > 0$, define the normalized scores $t_{ij}(\gamma) \in [0, 1]$ by

$$t_{ij}(\gamma) := \begin{cases} \dfrac{\xi_{ij}(\gamma) - \underline{\xi}_j(\gamma)}{\Delta_j(\gamma)}, & j \in J^+ \text{ (benefit)}, \\[2ex] \dfrac{\overline{\xi}_j(\gamma) - \xi_{ij}(\gamma)}{\Delta_j(\gamma)}, & j \in J^- \text{ (cost)}. \end{cases}$$

**(4) CoCoSo comparability sequences (SAW-like and WEP-like).** Define, for each alternative $A_i$ and scenario $\gamma$,

$$H_i(\gamma) := \sum_{j=1}^{n} w_j(\gamma)\, t_{ij}(\gamma), \qquad G_i(\gamma) := \prod_{j=1}^{n} \big(t_{ij}(\gamma)\big)^{w_j(\gamma)},$$

with the convention $0^0 := 1$ when it occurs.

**(5) Three appraisal scores.** Let

$$S_{GH}(\gamma) := \sum_{p=1}^{m} \big(G_p(\gamma) + H_p(\gamma)\big), \quad H_{\min}(\gamma) := \min_p H_p(\gamma), \quad G_{\min}(\gamma) := \min_p G_p(\gamma),$$

$$H_{\max}(\gamma) := \max_p H_p(\gamma), \quad G_{\max}(\gamma) := \max_p G_p(\gamma).$$

Assuming the denominators below are positive, define

$$as_i^{(\alpha)}(\gamma) := \frac{G_i(\gamma) + H_i(\gamma)}{S_{GH}(\gamma)},$$

$$as_i^{(\beta)}(\gamma) := \frac{H_i(\gamma)}{H_{\min}(\gamma)} + \frac{G_i(\gamma)}{G_{\min}(\gamma)},$$

$$as_i^{(\gamma)}(\gamma) := \frac{\lambda H_i(\gamma) + (1-\lambda)G_i(\gamma)}{\lambda H_{\max}(\gamma) + (1-\lambda)G_{\max}(\gamma)}.$$

**(6) Final U-CoCoSo score and decision rule.** Define the final scenario-wise score

$$AS_i(\gamma) := \Big(as_i^{(\alpha)}(\gamma)\, as_i^{(\beta)}(\gamma)\, as_i^{(\gamma)}(\gamma)\Big)^{1/3} + \frac{as_i^{(\alpha)}(\gamma) + as_i^{(\beta)}(\gamma) + as_i^{(\gamma)}(\gamma)}{3}.$$

Then $AS_i : \Gamma \to \mathbb{R}_{\geq 0}$ is an uncertain variable (under the well-definedness conditions below), and it induces an uncertain set

$$\mathcal{AS}_i(\gamma) := \{AS_i(\gamma)\} \subseteq \mathbb{R}_{\geq 0}.$$

If the expectation exists, define the crisp utility

$$U_i := \mathbb{E}[AS_i] \in \mathbb{R},$$

rank alternatives by $A_p \succeq_{\text{U-CoCoSo}} A_q \iff U_p \geq U_q$, and select any

$$A^{\star} \in \arg\max_{1 \leq i \leq m} U_i.$$

**Theorem 6.3.4** (Uncertain-set structure and well-definedness of U-CoCoSo). *In Definition 6.3.3, assume:*

(A1) *Each $\xi_{ij}$ and $\omega_j$ is an $\mathcal{L}$-measurable map $\Gamma \to \mathbb{R}_{\geq 0}$.*

(A2) *Weight normalization is valid: $W(\gamma) = \sum_{t=1}^{n} \omega_t(\gamma) > 0$ for all $\gamma \in \Gamma$.*

(A3) *Nondegenerate criteria ranges: $\Delta_j(\gamma) > 0$ for all $j = 1, \ldots, n$ and all $\gamma \in \Gamma$.*

(A4) *Positive denominators for appraisal scores:*

$$S_{GH}(\gamma) > 0, \quad H_{\min}(\gamma) > 0, \quad G_{\min}(\gamma) > 0, \quad \lambda H_{\max}(\gamma) + (1-\lambda)G_{\max}(\gamma) > 0 \quad (\forall \gamma \in \Gamma).$$

(A5) *Finite expected scores: $\mathbb{E}[AS_i] \in \mathbb{R}$ exists for all $i = 1, \ldots, m$.*

*Then:*

(i) *all quantities in Steps (2)–(6) of Definition 6.3.3 are well-defined for every $\gamma \in \Gamma$;*

(ii) *each $AS_i : \Gamma \to \mathbb{R}_{\geq 0}$ is an uncertain variable, hence $\mathcal{AS}_i(\gamma) = \{AS_i(\gamma)\}$ is an uncertain set on $(\Gamma, \mathcal{L}, \mathcal{M})$;*

(iii) *the ranking preorder $\succeq_{\text{U-CoCoSo}}$ induced by $U_i = \mathbb{E}[AS_i]$ is well-defined on $\mathcal{A}$, and $\arg\max_i U_i$ is nonempty.*

*Proof.* Fix any $\gamma \in \Gamma$.

By (A2), $W(\gamma) > 0$, so each normalized weight $w_j(\gamma) = \omega_j(\gamma)/W(\gamma)$ is a well-defined real number and $\sum_j w_j(\gamma) = 1$.

By (A3), $\Delta_j(\gamma) > 0$, so each normalized performance $t_{ij}(\gamma)$ is well-defined. Moreover, by construction, $t_{ij}(\gamma) \in [0, 1]$.

Hence the weighted sum $H_i(\gamma) = \sum_j w_j(\gamma) t_{ij}(\gamma)$ is well-defined and belongs to $[0, 1]$. Also $G_i(\gamma) = \prod_j (t_{ij}(\gamma))^{w_j(\gamma)}$ is well-defined as a finite product of nonnegative reals (using $0^0 := 1$ when needed), so $G_i(\gamma) \in [0, 1]$.

Assumption (A4) guarantees that every denominator appearing in $as_i^{(\alpha)}(\gamma), as_i^{(\beta)}(\gamma), as_i^{(\gamma)}(\gamma)$ is strictly positive, so these appraisal scores are well-defined real numbers. Consequently $AS_i(\gamma)$ is well-defined.

To show measurability, note that sums, products, scalar division by a strictly positive measurable function, and pointwise min / max preserve $\mathcal{L}$-measurability. Under (A1)–(A4), each mapping $\gamma \mapsto AS_i(\gamma)$ is obtained from $(\xi_{ij})$ and $(\omega_j)$ by finitely many such operations, therefore $AS_i$ is an $\mathcal{L}$-measurable map $\Gamma \to \mathbb{R}_{\geq 0}$, i.e., an uncertain variable. This proves (i) and (ii).

Finally, by (A5) each $U_i = \mathbb{E}[AS_i]$ is a finite real number, so the preorder $A_p \succeq_{\text{U-CoCoSo}} A_q \iff U_p \geq U_q$ is well-defined. Since $\mathcal{A}$ is finite, $\max_i U_i$ exists and $\arg\max_i U_i$ is nonempty, proving (iii). □

CoCoSo and representative uncertainty-aware variants are listed in Table 6.4.

Table 6.4: CoCoSo and representative uncertainty-aware variants (organized by the degree-domain dimension $k$).

| $k$ | Representative CoCoSo-type model(s) whose evaluation degrees live in a subset of $[0,1]^k$ |
|---|---|
| 1 | Fuzzy CoCoSo. |
| 2 | Intuitionistic Fuzzy CoCoSo [758, 759]; Pythagorean fuzzy Cocoso [760, 761]; Fermatean fuzzy CoCoSo [762]. |
| 3 | Picture Fuzzy CoCoSo [763]; Spherical fuzzy Cocoso [764, 765]; Neutrosophic CoCoSo [766, 767]. |
| $n$ | Plithogenic CoCoSo [768] |

**Note.** Here $k$ denotes the ambient dimension of the degree-domain used to encode each criterion evaluation (and, if applicable, each weight). In this convention, picture-fuzzy and (single-valued) neutrosophic evaluations are treated as three-component degrees, hence $k = 3$.

As concepts other than Uncertain CoCoSo, Rough CoCoSo [769], Grey CoCoSo [770, 771], SWARA-CoCoSo [772, 773], and Extended CoCoSo [774, 775] are also known.

## 6.4 Fuzzy SAW (Fuzzy Simple Additive Weighting)

Classical SAW normalizes performance scores, multiplies each criterion by its weight, sums weighted values, and ranks alternatives by maximum total score under given criteria [776, 777]. Fuzzy SAW normalizes fuzzy performance ratings, multiplies by criterion weights, sums weighted fuzzy scores, defuzzifies them, and ranks alternatives by highest overall utility final index [778–780].

**Definition 6.4.1** (Fuzzy SAW (FSAW) with triangular fuzzy numbers). [778–780] Let $\mathcal{A} = \{A_1, \ldots, A_m\}$ be a finite set of alternatives and $\mathcal{C} = \{C_1, \ldots, C_n\}$ a finite set of criteria. Assume each criterion $C_j$ is either a *benefit* criterion ($\uparrow$; larger is better) or a *cost* criterion ($\downarrow$; smaller is better).

**(0) Triangular fuzzy numbers (TFNs) and arithmetic.** Let

$$\mathsf{TFN}_{\geq 0} := \{(l, m, u) \in \mathbb{R}^3 : 0 \leq l \leq m \leq u\}.$$

For $\tilde{x} = (l_x, m_x, u_x), \tilde{y} = (l_y, m_y, u_y) \in \mathsf{TFN}_{\geq 0}$ and $\alpha \geq 0$, define

$$\tilde{x} \oplus \tilde{y} := (l_x + l_y, \ m_x + m_y, \ u_x + u_y), \qquad \alpha \odot \tilde{x} := (\alpha l_x, \ \alpha m_x, \ \alpha u_x).$$

**(1) Fuzzy decision matrix.** A *TFN-based fuzzy decision matrix* is

$$\tilde{X} = (\tilde{x}_{ij}) \in (\mathsf{TFN}_{\geq 0})^{m \times n}, \qquad \tilde{x}_{ij} = (a_{ij}, b_{ij}, c_{ij}),$$

where $\tilde{x}_{ij}$ represents the (possibly linguistic) performance of $A_i$ under $C_j$. If $P \geq 1$ decision makers provide TFNs $\tilde{x}_{ij}^{(p)} = (a_{ij}^{(p)}, b_{ij}^{(p)}, c_{ij}^{(p)})$, a standard aggregation is the TFN geometric mean

$$\tilde{x}_{ij} := \Big( \prod_{p=1}^{P} a_{ij}^{(p)} \Big)^{1/P}, \quad \Big( \prod_{p=1}^{P} b_{ij}^{(p)} \Big)^{1/P}, \quad \Big( \prod_{p=1}^{P} c_{ij}^{(p)} \Big)^{1/P}.$$

(Geometric-mean aggregation for the fuzzy MCDM matrix is commonly used.)

**(2) Criterion weights.** Let $w = (w_1, \ldots, w_n)$ be a (crisp) weight vector with

$$w_j \geq 0, \qquad \sum_{j=1}^{n} w_j = 1.$$

(Weights can be obtained, for example, by defuzzifying and normalizing linguistic judgments.)

**(3) Normalization of TFNs.** To map heterogeneous criteria to comparable scales and keep normalized TFNs in $[0, 1]$, a standard linear scale transformation for benefit criteria is:

$$\tilde{r}_{ij} := \Big( \frac{a_{ij}}{c_j^*}, \ \frac{b_{ij}}{c_j^*}, \ \frac{c_{ij}}{c_j^*} \Big), \qquad c_j^* := \max_{1 \leq i \leq m} c_{ij}, \qquad (C_j : \uparrow).$$

This normalization is widely used in fuzzy SAW implementations.

For cost criteria, one common dual form (after ensuring positivity) is:

$$\tilde{r}_{ij} := \left( \frac{a_j^-}{c_{ij}}, \ \frac{a_j^-}{b_{ij}}, \ \frac{a_j^-}{a_{ij}} \right), \qquad a_j^- := \min_{1 \le i \le m} a_{ij}, \qquad (C_j : \downarrow),$$

which is the TFN analogue of reciprocal cost normalization.

**(4) Weighted normalized matrix and fuzzy SAW score.** Define the weighted normalized TFN entries by

$$\tilde{v}_{ij} := w_j \odot \tilde{r}_{ij} = (w_j r_{ij}^{\ell}, \ w_j r_{ij}^{m}, \ w_j r_{ij}^{u}),$$

and the total fuzzy score of alternative $A_i$ by the TFN sum

$$\tilde{s}_i := \bigoplus_{j=1}^{n} \tilde{v}_{ij} \in \mathsf{TFN}_{\ge 0}.$$

Equivalently, in matrix form this is the fuzzy weighted summation $\tilde{s} = \tilde{R} w$ (the SAW aggregation of normalized ratings with weights).

**(5) Defuzzification and ranking.** Convert $\tilde{s}_i = (s_i^{\ell}, s_i^{m}, s_i^{u})$ to a crisp score via (signed-distance / centroid) defuzzification:

$$\mathrm{Score}(A_i) := \mathrm{Defuzz}(\tilde{s}_i) := \frac{s_i^{\ell} + s_i^{m} + s_i^{u}}{3}.$$

Then rank alternatives by descending score:

$$A_p \succeq_{\mathrm{FSAW}} A_q \quad \Longleftrightarrow \quad \mathrm{Score}(A_p) \ge \mathrm{Score}(A_q).$$

(Using $(a + b + c)/3$ as a defuzzification rule for TFNs is standard in fuzzy SAW practice.)

**(6) Reduction to crisp SAW.** If every TFN is degenerate $\tilde{x}_{ij} = (x_{ij}, x_{ij}, x_{ij})$, then $\tilde{r}_{ij}$ and $\tilde{s}_i$ are degenerate as well, and the above procedure reduces to the classical (crisp) SAW method.

Using Uncertain Sets, we define Uncertain SAW (U-SAW) as follows.

**Definition 6.4.2** (Uncertain SAW (U-SAW): scenario-wise SAW with expected-score decision)**.** Let $\mathcal{A} = \{A_1, \dots, A_m\}$ be a finite set of alternatives and $\mathcal{C} = \{C_1, \dots, C_n\}$ a finite set of criteria, partitioned as

$$\mathcal{C} = \mathcal{C}^+ \,\dot{\cup}\, \mathcal{C}^- \quad \text{(benefit/cost)}, \qquad J^+ := \{j : C_j \in \mathcal{C}^+\}, \ \ J^- := \{j : C_j \in \mathcal{C}^-\}.$$

Fix an uncertainty space $(\Gamma, \mathcal{L}, \mathcal{M})$.

**(1) Uncertain input data.** An *uncertain SAW instance* consists of:

- an *uncertain decision matrix*

$$\Xi = (\xi_{ij}) \in \mathsf{UV}^{m \times n}, \qquad \xi_{ij} : \Gamma \to \mathbb{R}_{\ge 0},$$

  where $\xi_{ij}(\gamma)$ is the (nonnegative) performance of $A_i$ under $C_j$ in scenario $\gamma$;

- an *uncertain criterion-weight vector*

$$\Omega = (\omega_1, \ldots, \omega_n) \in (\mathsf{UV}_{\geq 0})^n, \qquad \omega_j : \Gamma \to \mathbb{R}_{\geq 0}.$$

**(2) Scenario-wise weight normalization.** For each scenario $\gamma \in \Gamma$, define

$$W(\gamma) := \sum_{t=1}^{n} \omega_t(\gamma), \qquad w_j(\gamma) := \frac{\omega_j(\gamma)}{W(\gamma)} \quad (j = 1, \ldots, n),$$

so that $\sum_{j=1}^{n} w_j(\gamma) = 1$ whenever $W(\gamma) > 0$.

**(3) Scenario-wise performance normalization.** For each criterion $j$ and scenario $\gamma$, define

$$\underline{\xi}_j(\gamma) := \min_{1 \leq i \leq m} \xi_{ij}(\gamma), \qquad \overline{\xi}_j(\gamma) := \max_{1 \leq i \leq m} \xi_{ij}(\gamma).$$

Assuming $\overline{\xi}_j(\gamma) > 0$ for benefit criteria and $\underline{\xi}_j(\gamma) > 0$ for cost criteria, define normalized scores $r_{ij}(\gamma) \in [0,1]$ by

$$r_{ij}(\gamma) := \begin{cases} \dfrac{\xi_{ij}(\gamma)}{\overline{\xi}_j(\gamma)}, & j \in J^+ \text{ (benefit)}, \\[2ex] \dfrac{\underline{\xi}_j(\gamma)}{\xi_{ij}(\gamma)}, & j \in J^- \text{ (cost)}. \end{cases}$$

**(4) Scenario-wise SAW aggregation.** Define, for each alternative $A_i$ and scenario $\gamma$,

$$S_i(\gamma) := \sum_{j=1}^{n} w_j(\gamma) \, r_{ij}(\gamma) \ \in \ [0,1].$$

Then $S_i : \Gamma \to [0,1]$ is the *uncertain SAW score* (a scenario-wise SAW utility).

**(5) Uncertain-set structure and decision rule.** Each $S_i$ induces an uncertain set

$$\mathcal{S}_i(\gamma) := \{S_i(\gamma)\} \subseteq [0,1].$$

If $\mathbb{E}[S_i]$ exists and is finite, define the crisp utility

$$U_i := \mathbb{E}[S_i] \in \mathbb{R},$$

rank alternatives by

$$A_p \succeq_{\text{U-SAW}} A_q \quad \Longleftrightarrow \quad U_p \geq U_q,$$

and select any

$$A^\star \in \arg\max_{1 \leq i \leq m} U_i.$$

**Theorem 6.4.3** (Uncertain-set structure and well-definedness of U-SAW). *In Definition 6.4.2, assume:*

**(A1)** *(Measurability) Each $\xi_{ij}$ and $\omega_j$ is $\mathcal{L}$-measurable.*

(A2) *(Positive total weight)* $W(\gamma) = \sum_{t=1}^{n} \omega_t(\gamma) > 0$ *for all* $\gamma \in \Gamma$.

(A3) *(Normalization denominators)* *For all* $\gamma \in \Gamma$:

$$\overline{\xi}_j(\gamma) > 0 \ \ \forall j \in J^+, \qquad \underline{\xi}_j(\gamma) > 0 \ \ \forall j \in J^-, \qquad \xi_{ij}(\gamma) > 0 \ \ \forall (i,j) \in \{1,\ldots,m\} \times J^-.$$

(A4) *(Finite expectations)* $\mathbb{E}[S_i]$ *exists and is finite for all* $i = 1,\ldots,m$.

*Then:*

(i) *the normalized weights* $w_j(\gamma)$ *and normalized scores* $r_{ij}(\gamma)$ *are well-defined for every* $\gamma \in \Gamma$, *and* $r_{ij}(\gamma) \in [0,1]$;

(ii) *each* $S_i : \Gamma \to [0,1]$ *is an uncertain variable, hence* $\mathcal{S}_i(\gamma) = \{S_i(\gamma)\}$ *is an uncertain set on* $(\Gamma, \mathcal{L}, \mathcal{M})$;

(iii) *the preorder* $\succeq_{\text{U-SAW}}$ *induced by* $U_i = \mathbb{E}[S_i]$ *is well-defined, and since* $\mathcal{A}$ *is finite,* $\arg\max_i U_i$ *is nonempty.*

*Proof.* Fix $\gamma \in \Gamma$.

By (A2), $W(\gamma) > 0$, hence each normalized weight $w_j(\gamma) = \omega_j(\gamma)/W(\gamma)$ is a well-defined real number and $\sum_{j=1}^{n} w_j(\gamma) = 1$.

For $j \in J^+$, (A3) gives $\overline{\xi}_j(\gamma) > 0$, so $r_{ij}(\gamma) = \xi_{ij}(\gamma)/\overline{\xi}_j(\gamma)$ is well-defined and lies in $[0,1]$ because $0 \leq \xi_{ij}(\gamma) \leq \overline{\xi}_j(\gamma)$ by definition of the maximum. For $j \in J^-$, (A3) ensures $\xi_{ij}(\gamma) > 0$ and $\underline{\xi}_j(\gamma) > 0$, so $r_{ij}(\gamma) = \underline{\xi}_j(\gamma)/\xi_{ij}(\gamma)$ is well-defined and belongs to $[0,1]$ since $\underline{\xi}_j(\gamma) \leq \xi_{ij}(\gamma)$.

Therefore, for each $i$ the SAW score $S_i(\gamma) = \sum_{j=1}^{n} w_j(\gamma) r_{ij}(\gamma)$ is well-defined and lies in $[0,1]$, as a convex combination of numbers in $[0,1]$.

To prove that $S_i$ is an uncertain variable, observe that under (A1) the maps $\xi_{ij}$ and $\omega_j$ are measurable; pointwise min / max of finitely many measurable functions is measurable; sums/products and division by a strictly positive measurable function preserve measurability. Hence $\gamma \mapsto w_j(\gamma)$, $\gamma \mapsto r_{ij}(\gamma)$, and consequently $\gamma \mapsto S_i(\gamma)$ are all $\mathcal{L}$-measurable. Thus $S_i$ is an uncertain variable $\Gamma \to [0,1]$, and the induced singleton mapping $\mathcal{S}_i(\gamma) = \{S_i(\gamma)\}$ is an uncertain set. This proves (i) and (ii).

Finally, (A4) yields finite reals $U_i = \mathbb{E}[S_i]$. Hence $A_p \succeq_{\text{U-SAW}} A_q \iff U_p \geq U_q$ defines a well-defined preorder on $\mathcal{A}$. Since $\mathcal{A}$ is finite, $\max_i U_i$ exists and $\arg\max_i U_i$ is nonempty. This proves (iii). $\square$

Related concepts of SAW under uncertainty-aware models are listed in Table 6.5.

Table 6.5: Related concepts of SAW under uncertainty-aware models.

| $k$ | Related SAW concept(s) |
|---|---|
| 1 | Fuzzy SAW |
| 2 | Intuitionistic Fuzzy SAW [781, 782] |
| 3 | Neutrosophic SAW [783, 784] |

## 6.5 Fuzzy RAFSI (Fuzzy Ranking of Alternatives through Functional mapping of criterion Sub-Intervals into a single Interval)

RAFSI maps criterion values to a common interval using ideal and anti-ideal points, then aggregates weighted normalized scores [785, 786]. Fuzzy RAFSI uses fuzzy numbers for evaluations, maps them to a common interval, normalizes benefit/cost, aggregates fuzzy weighted scores [787, 788].

**Definition 6.5.1** (TFN-based Fuzzy RAFSI). [787, 788] Let $\mathcal{A} = \{A_1, \ldots, A_m\}$ be a finite set of alternatives and $\mathcal{C} = \{C_1, \ldots, C_n\}$ a finite set of criteria. Partition criteria into benefit (max) and cost (min) types:

$$\mathcal{C} = \mathcal{C}^+ \,\dot\cup\, \mathcal{C}^-.$$

Assume evaluations are given by triangular fuzzy numbers (TFNs)

$$\tilde{x}_{ij} = (l_{ij}, m_{ij}, u_{ij}), \qquad 0 \le l_{ij} \le m_{ij} \le u_{ij},$$

forming the fuzzy decision matrix $\widetilde{X} = (\tilde{x}_{ij}) \in (\mathsf{TFN}_{\ge 0})^{m \times n}$. Let $w = (w_1, \ldots, w_n) \in [0,1]^n$ be criterion weights with $\sum_{j=1}^n w_j = 1$.

**TFN arithmetic (standard positive convention).** For TFNs $\tilde{a} = (a_1, a_2, a_3)$, $\tilde{b} = (b_1, b_2, b_3)$ and $\alpha \ge 0$, define

$$\tilde{a} \oplus \tilde{b} := (a_1 + b_1, \ a_2 + b_2, \ a_3 + b_3), \qquad \alpha \odot \tilde{a} := (\alpha a_1, \ \alpha a_2, \ \alpha a_3),$$

and (for strictly positive TFNs) the reciprocal

$$\tilde{a}^{-1} := \left( \frac{1}{a_3}, \frac{1}{a_2}, \frac{1}{a_1} \right).$$

Fix the centroid score (defuzzification) map

$$\mathrm{COA}(l, m, u) := \frac{l + m + u}{3}.$$

**Step 1 (reference points: ideal/anti-ideal).** For each criterion $C_j$, define crisp reference points (via COA-ordering):

$$a_j^I := \begin{cases} \max_{1 \le i \le m} \mathrm{COA}(\tilde{x}_{ij}), & C_j \in \mathcal{C}^+, \\ \min_{1 \le i \le m} \mathrm{COA}(\tilde{x}_{ij}), & C_j \in \mathcal{C}^-, \end{cases} \qquad a_j^N := \begin{cases} \min_{1 \le i \le m} \mathrm{COA}(\tilde{x}_{ij}), & C_j \in \mathcal{C}^+, \\ \max_{1 \le i \le m} \mathrm{COA}(\tilde{x}_{ij}), & C_j \in \mathcal{C}^-. \end{cases}$$

(Equivalently, a decision-maker may specify $a_j^I, a_j^N$ externally.)

**Step 2 (functional mapping to a common interval).** Fix numbers $n_1 > 0$ and $n_2 > n_1$ (often $n_1 = 1$, $n_2 = 6$). Define the affine map

$$f_j(x) := \frac{n_2 - n_1}{a_j^I - a_j^N}\, x + \frac{a_j^I n_1 - a_j^N n_2}{a_j^I - a_j^N} \qquad (a_j^I \neq a_j^N),$$

and set the standardized TFN

$$\tilde{s}_{ij} := f_j(\tilde{x}_{ij}) := \big(f_j(l_{ij}),\, f_j(m_{ij}),\, f_j(u_{ij})\big) \in \mathsf{TFN}_{>0}.$$

(If $a_j^I = a_j^N$, set $\tilde{s}_{ij} := (n_1, n_1, n_1)$ for all $i$.)

**Step 3 (arithmetic and harmonic means of the common interval).** Define

$$A := \frac{n_1 + n_2}{2}, \qquad H := \frac{2}{\frac{1}{n_1} + \frac{1}{n_2}}.$$

**Step 4 (RAFSI normalization).** Define the normalized TFNs $\widehat{s}_{ij}$ by

$$\widehat{s}_{ij} := \begin{cases} \dfrac{1}{2A} \odot \tilde{s}_{ij}, & C_j \in \mathcal{C}^+, \\[2ex] \dfrac{H}{2} \odot \tilde{s}_{ij}^{-1}, & C_j \in \mathcal{C}^-. \end{cases}$$

**Step 5 (criteria function / overall score and ranking).** For each alternative $A_i$, define its overall fuzzy RAFSI score

$$\tilde{V}(A_i) := \bigoplus_{j=1}^{n} \big(w_j \odot \widehat{s}_{ij}\big) \in \mathsf{TFN}_{>0}.$$

Defuzzify $V_i := \mathrm{COA}(\tilde{V}(A_i))$ and rank alternatives by

$$A_p \succeq_{\text{F-RAFSI}} A_q \quad \Longleftrightarrow \quad V_p \geq V_q.$$

Any maximizer $A^\star \in \arg\max_i V_i$ is called a *Fuzzy RAFSI best alternative*.

Using Uncertain Sets, we define Uncertain RAFSI (U-RAFSI) as follows.

**Definition 6.5.2** (Uncertainty space, uncertain variable, and induced uncertain set). An *uncertainty space* is a triple $(\Gamma, \mathcal{L}, \mathcal{M})$ where $\Gamma \neq \varnothing$, $\mathcal{L}$ is a $\sigma$-algebra on $\Gamma$, and $\mathcal{M} : \mathcal{L} \to [0, 1]$ is an uncertain measure. A (real-valued) *uncertain variable* is an $\mathcal{L}$-measurable map $\xi : \Gamma \to \mathbb{R}$. Every uncertain variable $\xi$ induces a singleton-valued *uncertain set*

$$\mathcal{X}_\xi : \Gamma \to \mathcal{P}(\mathbb{R}), \qquad \mathcal{X}_\xi(\gamma) := \{\xi(\gamma)\}.$$

Whenever it exists and is finite, $\mathbb{E}[\xi] \in \mathbb{R}$ denotes the expected value of $\xi$.

**Definition 6.5.3** (Uncertain RAFSI (U-RAFSI): scenario-wise RAFSI with expected-score decision)**.** Let $\mathcal{A} = \{A_1, \ldots, A_m\}$ be alternatives and $\mathcal{C} = \{C_1, \ldots, C_n\}$ criteria, partitioned into benefit/cost sets

$$\mathcal{C} = \mathcal{C}^+ \,\dot{\cup}\, \mathcal{C}^-, \qquad J^+ := \{j : C_j \in \mathcal{C}^+\}, \ \ J^- := \{j : C_j \in \mathcal{C}^-\}.$$

Fix an uncertainty space $(\Gamma, \mathcal{L}, \mathcal{M})$ and fixed constants $0 < n_1 < n_2$. Set the arithmetic and harmonic anchors

$$A := \frac{n_1 + n_2}{2}, \qquad H := \frac{2}{\frac{1}{n_1} + \frac{1}{n_2}}.$$

**(1) Uncertain evaluations and weights.** An *uncertain RAFSI instance* is specified by:

- an *uncertain decision matrix*

$$\Xi = (\xi_{ij}) \in \mathsf{UV}^{m \times n}, \qquad \xi_{ij} : \Gamma \to \mathbb{R},$$

- an *uncertain criterion-weight vector*

$$\Omega = (\omega_1, \ldots, \omega_n) \in (\mathsf{UV}_{\geq 0})^n, \qquad \omega_j : \Gamma \to \mathbb{R}_{\geq 0}.$$

**(2) Scenario-wise weight normalization.** For each scenario $\gamma \in \Gamma$, define

$$W(\gamma) := \sum_{t=1}^{n} \omega_t(\gamma), \qquad w_j(\gamma) := \frac{\omega_j(\gamma)}{W(\gamma)} \quad (j = 1, \ldots, n).$$

**(3) Scenario-wise ideal/anti-ideal reference points.** For each criterion $j$ and scenario $\gamma$, define the reference scalars

$$a_j^I(\gamma) := \begin{cases} \max\limits_{1 \leq i \leq m} \xi_{ij}(\gamma), & j \in J^+, \\ \min\limits_{1 \leq i \leq m} \xi_{ij}(\gamma), & j \in J^-, \end{cases} \qquad a_j^N(\gamma) := \begin{cases} \min\limits_{1 \leq i \leq m} \xi_{ij}(\gamma), & j \in J^+, \\ \max\limits_{1 \leq i \leq m} \xi_{ij}(\gamma), & j \in J^-. \end{cases}$$

(Thus $a_j^I(\gamma)$ is the scenario-wise ideal and $a_j^N(\gamma)$ is the scenario-wise anti-ideal.)

**(4) Scenario-wise functional mapping to the common interval** $[n_1, n_2]$**.** Whenever $a_j^I(\gamma) \neq a_j^N(\gamma)$, define the affine map

$$f_{j,\gamma}(x) := \frac{n_2 - n_1}{a_j^I(\gamma) - a_j^N(\gamma)} \, x + \frac{a_j^I(\gamma) n_1 - a_j^N(\gamma) n_2}{a_j^I(\gamma) - a_j^N(\gamma)}.$$

Define the standardized (mapped) performance

$$s_{ij}(\gamma) := \begin{cases} f_{j,\gamma}(\xi_{ij}(\gamma)), & a_j^I(\gamma) \neq a_j^N(\gamma), \\ n_1, & a_j^I(\gamma) = a_j^N(\gamma), \end{cases} \qquad (i = 1, \ldots, m; \ j = 1, \ldots, n).$$

(So $s_{ij}(\gamma) \in [n_1, n_2]$ whenever $a_j^I(\gamma) \neq a_j^N(\gamma)$.)

**(5) RAFSI normalization (benefit/cost).** Define normalized values $\widehat{s}_{ij}(\gamma)$ by

$$\widehat{s}_{ij}(\gamma) := \begin{cases} \dfrac{s_{ij}(\gamma)}{2A}, & j \in J^+, \\[2ex] \dfrac{H}{2} \cdot \dfrac{1}{s_{ij}(\gamma)}, & j \in J^-. \end{cases}$$

**(6) Scenario-wise RAFSI score and uncertain-set output.** For each alternative $A_i$, define its (scenario-wise) RAFSI score

$$V_i(\gamma) := \sum_{j=1}^{n} w_j(\gamma)\,\widehat{s}_{ij}(\gamma) \in \mathbb{R}.$$

Then $V_i : \Gamma \to \mathbb{R}$ is an uncertain variable, and it induces an uncertain set

$$\mathcal{V}_i(\gamma) := \{V_i(\gamma)\}.$$

**(7) Expected-score ranking and selection.** If $\mathbb{E}[V_i]$ exists and is finite, define

$$U_i := \mathbb{E}[V_i] \in \mathbb{R}, \qquad A_p \succeq_{\text{U-RAFSI}} A_q \iff U_p \geq U_q,$$

and select any

$$A^\star \in \arg \max_{1 \leq i \leq m} U_i.$$

**Theorem 6.5.4** (Uncertain-set structure and well-definedness of U-RAFSI). *In Definition 6.5.3, assume:*

(A1) *(Measurability) Each $\xi_{ij}$ and $\omega_j$ is $\mathcal{L}$-measurable.*

(A2) *(Positive total weight) $W(\gamma) = \sum_{t=1}^{n} \omega_t(\gamma) > 0$ for all $\gamma \in \Gamma$.*

(A3) *(Common-interval anchors) $0 < n_1 < n_2$.*

(A4) *(Cost positivity after mapping) For every $\gamma \in \Gamma$ and $j \in J^-$, the mapped values satisfy $s_{ij}(\gamma) > 0$ for all $i$ (in particular this holds if $n_1 > 0$).*

(A5) *(Finite expectations) $\mathbb{E}[V_i]$ exists and is finite for all $i = 1, \ldots, m$.*

*Then:*

(i) *for every $\gamma \in \Gamma$, the quantities $w_j(\gamma)$, $a_j^I(\gamma)$, $a_j^N(\gamma)$, $s_{ij}(\gamma)$, $\widehat{s}_{ij}(\gamma)$, and $V_i(\gamma)$ are well-defined real numbers;*

(ii) *each $V_i$ is an uncertain variable, hence $\mathcal{V}_i(\gamma) = \{V_i(\gamma)\}$ is an uncertain set;*

(iii) *the preorder $\succeq_{\text{U-RAFSI}}$ is well-defined and $\arg \max_i \mathbb{E}[V_i] \neq \varnothing$.*

*Proof.* Fix $\gamma \in \Gamma$.

(i) By (A2), $W(\gamma) > 0$, hence $w_j(\gamma) = \omega_j(\gamma)/W(\gamma)$ is well-defined and $\sum_{j=1}^n w_j(\gamma) = 1$. Since $m < \infty$, the pointwise maxima/minima of $\{\xi_{ij}(\gamma)\}_{i=1}^m$ exist in $\mathbb{R}$, so $a_j^I(\gamma)$ and $a_j^N(\gamma)$ are well-defined reals.

If $a_j^I(\gamma) \neq a_j^N(\gamma)$, then $f_{j,\gamma}$ is a well-defined affine map and $s_{ij}(\gamma) = f_{j,\gamma}(\xi_{ij}(\gamma))$ is a real number. If $a_j^I(\gamma) = a_j^N(\gamma)$, we set $s_{ij}(\gamma) = n_1$, which is well-defined by (A3). Thus $s_{ij}(\gamma)$ is well-defined in all cases. Moreover, for $a_j^I(\gamma) \neq a_j^N(\gamma)$ one has $f_{j,\gamma}(a_j^N(\gamma)) = n_2$ and $f_{j,\gamma}(a_j^I(\gamma)) = n_1$, so $s_{ij}(\gamma) \in [n_1, n_2]$ whenever $\xi_{ij}(\gamma) \in [a_j^N(\gamma), a_j^I(\gamma)]$, which holds by definition of max/min. Hence $s_{ij}(\gamma) \geq n_1 > 0$. In particular, for $j \in J^-$ we have $s_{ij}(\gamma) > 0$, so the reciprocal $1/s_{ij}(\gamma)$ in the cost normalization is well-defined. Therefore $\widehat{s}_{ij}(\gamma)$ is well-defined for all $i, j$, and so is $V_i(\gamma) = \sum_{j=1}^n w_j(\gamma)\widehat{s}_{ij}(\gamma)$.

(ii) Under (A1), each $\xi_{ij}$ and $\omega_j$ is measurable. Finite sums preserve measurability, and pointwise min / max of finitely many measurable functions is measurable. On the set $\{\gamma : a_j^I(\gamma) \neq a_j^N(\gamma)\}$, $s_{ij}(\gamma)$ is obtained from measurable functions by arithmetic operations and division by a nonzero measurable function, hence is measurable there; on the complementary set it is the constant $n_1$, hence measurable. Thus $s_{ij}$ is measurable on all of $\Gamma$. Since $1/s_{ij}$ is measurable wherever $s_{ij} > 0$ (and (A4) guarantees this for cost criteria), it follows that $\widehat{s}_{ij}$ is measurable. Consequently each $V_i$ is $\mathcal{L}$-measurable, i.e., an uncertain variable, and $\mathcal{V}_i(\gamma) = \{V_i(\gamma)\}$ is an uncertain set.

(iii) By (A5) each $U_i = \mathbb{E}[V_i]$ is a finite real number, so $A_p \succeq_{\text{U-RAFSI}} A_q \iff U_p \geq U_q$ defines a well-defined preorder. Since $\mathcal{A}$ is finite, $\max_i U_i$ exists and $\arg\max_i U_i$ is nonempty. $\square$

Related concepts of RAFSI under uncertainty-aware models are listed in Table 6.6.

Table 6.6: Related concepts of RAFSI under uncertainty-aware models.

| $k$ | Related RAFSI concept(s) |
|---|---|
| 1 | Fuzzy RAFSI |
| 2 | Intuitionistic Fuzzy RAFSI |
| 2 | Fermatean Fuzzy RAFSI [789] |
| 3 | Neutrosophic RAFSI [790] |

## 6.6 Fuzzy RATMI (Fuzzy Ranking of Alternatives by Trace-to-Median Index)

RATMI is a matrix-based MCDM ranking method that normalizes criterion values, applies criterion weights, and produces a single trace-to-median index by combining a trace-based performance term with a median-similarity term [791, 792]. Related concepts, such as RAMS-RATMI [793, 794], are also known. Fuzzy RATMI extends RATMI to settings where evaluations are fuzzy or linguistic; the fuzzy assessments are first converted to comparable scalar scores (e.g., by defuzzification), and then the same trace-to-median aggregation is used to rank alternatives under uncertainty [795].

**Definition 6.6.1** (Fuzzy RATMI (FRATMI): Fuzzy Ranking of Alternatives by Trace-to-Median Index). [795] Let $\mathcal{A} = \{A_1, \ldots, A_m\}$ be alternatives and $\mathcal{C} = \{C_1, \ldots, C_n\}$ criteria $(m, n \geq 2)$. Partition criteria into benefit and cost sets:

$$\mathcal{C} = \mathcal{C}^{\text{ben}} \dot{\cup} \mathcal{C}^{\text{cost}}.$$

Assume a *positive* fuzzy decision matrix

$$\widetilde{X} = (\tilde{x}_{ij})_{m \times n}, \qquad \tilde{x}_{ij} \in \mathbb{F}_{>0},$$

where $\mathbb{F}_{>0}$ is a chosen family of positive fuzzy evaluations (e.g., TFNs, trapezoidal fuzzy numbers, interval-valued fuzzy numbers).

**Step 0 (Score/defuzzification).** Fix a score map $S : \mathbb{F}_{>0} \to (0, \infty)$ and define a crisp matrix

$$x_{ij} := S(\tilde{x}_{ij}) > 0 \qquad (i = 1, \dots, m; \ j = 1, \dots, n).$$

(Example for TFN $\tilde{x} = (\ell, m, u)$: $S(\tilde{x}) = (\ell + m + u)/3$.)

**Step 1 (Normalization).** For each criterion $C_j$, define

$$r_{ij} := \begin{cases} \dfrac{x_{ij}}{\max_{1 \leq i \leq m} x_{ij}}, & C_j \in \mathcal{C}^{\text{ben}}, \\[2mm] \dfrac{\min_{1 \leq i \leq m} x_{ij}}{x_{ij}}, & C_j \in \mathcal{C}^{\text{cost}}. \end{cases}$$

Then $r_{ij} \in (0, 1]$.

**Step 2 (Weights and weighted normalization).** Let $w = (w_1, \dots, w_n)^\top$ be criterion weights with

$$w_j \geq 0, \qquad \sum_{j=1}^{n} w_j = 1.$$

Define the weighted normalized matrix $U = (u_{ij})$ by

$$u_{ij} := w_j \, r_{ij} \qquad (i = 1, \dots, m; \ j = 1, \dots, n).$$

**Step 3 (Optimal alternative components).** Define the componentwise optimal vector $Q = (q_1, \dots, q_n)$ by

$$q_j := \max_{1 \leq i \leq m} u_{ij} \qquad (j = 1, \dots, n).$$

Let $k := |\mathcal{C}^{\text{ben}}|$ and $h := |\mathcal{C}^{\text{cost}}|$ ($k + h = n$), and decompose $Q = Q_{\max} \cup Q_{\min}$ according to $\mathcal{C}^{\text{ben}}$ and $\mathcal{C}^{\text{cost}}$.

**Step 4 (Magnitudes).** Define the magnitudes of the benefit- and cost-components:

$$Q_k := \sqrt{\sum_{C_j \in \mathcal{C}^{\text{ben}}} q_j^2}, \qquad Q_h := \sqrt{\sum_{C_j \in \mathcal{C}^{\text{cost}}} q_j^2}.$$

For each alternative $A_i$, define

$$U_{ik} := \sqrt{\sum_{C_j \in \mathcal{C}^{\text{ben}}} u_{ij}^2}, \qquad U_{ih} := \sqrt{\sum_{C_j \in \mathcal{C}^{\text{cost}}} u_{ij}^2}.$$

**Step 5 (Trace index: MCRAT component).** Define diagonal matrices

$$F := \begin{pmatrix} Q_k & 0 \\ 0 & Q_h \end{pmatrix}, \qquad G_i := \begin{pmatrix} U_{ik} & 0 \\ 0 & U_{ih} \end{pmatrix}, \qquad T_i := F\,G_i.$$

Define the trace score

$$\mathrm{tr}_i := \mathrm{tr}(T_i) = Q_k U_{ik} + Q_h U_{ih}.$$

**Step 6 (Median similarity: RAMS component).** Define the median lengths

$$M := \frac{\sqrt{Q_k^2 + Q_h^2}}{2}, \qquad M_i := \frac{\sqrt{U_{ik}^2 + U_{ih}^2}}{2},$$

and the median similarity

$$\mathrm{MS}_i := \frac{M_i}{M}.$$

**Step 7 (Trace-to-median index: RATMI).** Let $v \in [0,1]$ (typically $v = 0.5$). Define

$$\mathrm{tr}^* := \min_{1 \le i \le m} \mathrm{tr}_i, \quad \mathrm{tr}^- := \max_{1 \le i \le m} \mathrm{tr}_i, \quad \mathrm{MS}^* := \min_{1 \le i \le m} \mathrm{MS}_i, \quad \mathrm{MS}^- := \max_{1 \le i \le m} \mathrm{MS}_i.$$

Assume $\mathrm{tr}^- > \mathrm{tr}^*$ and $\mathrm{MS}^- > \mathrm{MS}^*$. The *(fuzzy) RATMI index* of $A_i$ is

$$E_i := v\,\frac{\mathrm{tr}_i - \mathrm{tr}^*}{\mathrm{tr}^- - \mathrm{tr}^*} + (1 - v)\,\frac{\mathrm{MS}_i - \mathrm{MS}^*}{\mathrm{MS}^- - \mathrm{MS}^*}.$$

**Ranking rule:** rank alternatives in descending order of $E_i$.

We now present Uncertain RATMI of type $M$, obtained by extending the framework using Uncertain Sets.

**Definition 6.6.2** (Uncertain RATMI of type $M$ (U-RATMI)). Let $\mathcal{A} = \{A_1, \ldots, A_m\}$ be alternatives and $\mathcal{C} = \{C_1, \ldots, C_n\}$ criteria, with $m, n \ge 2$. Partition criteria into benefit and cost sets:

$$\mathcal{C} = \mathcal{C}^{\mathrm{ben}} \,\dot{\cup}\, \mathcal{C}^{\mathrm{cost}}.$$

Fix an uncertain model $M$ with $\mathrm{Dom}(M) \ne \emptyset$ and an admissible positive score $S_M$. Assume an *uncertain decision matrix*

$$X^{(M)} = \big(x_{ij}^{(M)}\big)_{m \times n}, \qquad x_{ij}^{(M)} \in \mathrm{Dom}(M) \quad (i = 1, \ldots, m;\ j = 1, \ldots, n).$$

**Step 0 (Crisp projection).** Define the positive real matrix $X = (x_{ij})$ by

$$x_{ij} := S_M\big(x_{ij}^{(M)}\big) \in (0, \infty).$$

**Step 1 (Normalization).** For each criterion $C_j$, define

$$x_j^{\max} := \max_{1 \le i \le m} x_{ij}, \qquad x_j^{\min} := \min_{1 \le i \le m} x_{ij}.$$

Then $x_j^{\max} > 0$ and $x_j^{\min} > 0$. Define

$$r_{ij} := \begin{cases} \dfrac{x_{ij}}{x_j^{\max}}, & C_j \in \mathcal{C}^{\mathrm{ben}}, \\[2ex] \dfrac{x_j^{\min}}{x_{ij}}, & C_j \in \mathcal{C}^{\mathrm{cost}}. \end{cases} \qquad (i = 1, \ldots, m; \ j = 1, \ldots, n).$$

Hence $r_{ij} \in (0, 1]$.

**Step 2 (Weights and weighted normalization).** Let $w = (w_1, \ldots, w_n)^\top$ be criterion weights satisfying

$$w_j \geq 0, \qquad \sum_{j=1}^{n} w_j = 1.$$

Define the weighted normalized matrix $U = (u_{ij})$ by

$$u_{ij} := w_j\, r_{ij} \qquad (i = 1, \ldots, m; \ j = 1, \ldots, n).$$

**Step 3 (Componentwise optimal vector).** Define

$$q_j := \max_{1 \leq i \leq m} u_{ij} \qquad (j = 1, \ldots, n),$$

and set $Q = (q_1, \ldots, q_n)$.

**Step 4 (Benefit/cost magnitudes).** Let $k := |\mathcal{C}^{\mathrm{ben}}|$ and $h := |\mathcal{C}^{\mathrm{cost}}|$ ($k + h = n$). Define

$$Q_k := \sqrt{\sum_{C_j \in \mathcal{C}^{\mathrm{ben}}} q_j^2}, \qquad Q_h := \sqrt{\sum_{C_j \in \mathcal{C}^{\mathrm{cost}}} q_j^2},$$

and for each alternative $A_i$,

$$U_{ik} := \sqrt{\sum_{C_j \in \mathcal{C}^{\mathrm{ben}}} u_{ij}^2}, \qquad U_{ih} := \sqrt{\sum_{C_j \in \mathcal{C}^{\mathrm{cost}}} u_{ij}^2}.$$

(Empty sums are understood as 0.)

**Step 5 (Trace index component).** Define diagonal matrices

$$F := \begin{pmatrix} Q_k & 0 \\ 0 & Q_h \end{pmatrix}, \qquad G_i := \begin{pmatrix} U_{ik} & 0 \\ 0 & U_{ih} \end{pmatrix}, \qquad T_i := F\, G_i,$$

and the trace score

$$\mathrm{tr}_i := \mathrm{tr}(T_i) = Q_k U_{ik} + Q_h U_{ih}.$$

**Step 6 (Median similarity component).** Define

$$M := \frac{\sqrt{Q_k^2 + Q_h^2}}{2}, \qquad M_i := \frac{\sqrt{U_{ik}^2 + U_{ih}^2}}{2}, \qquad \mathrm{MS}_i := \frac{M_i}{M}.$$

**Step 7 (Trace-to-median index).** Let $v \in [0,1]$. Define

$$\mathrm{tr}^* := \min_{1 \le i \le m} \mathrm{tr}_i, \quad \mathrm{tr}^- := \max_{1 \le i \le m} \mathrm{tr}_i, \quad \mathrm{MS}^* := \min_{1 \le i \le m} \mathrm{MS}_i, \quad \mathrm{MS}^- := \max_{1 \le i \le m} \mathrm{MS}_i,$$

and the ranges $\Delta_{\mathrm{tr}} := \mathrm{tr}^- - \mathrm{tr}^*$, $\Delta_{\mathrm{MS}} := \mathrm{MS}^- - \mathrm{MS}^*$. Define normalized components

$$T_i := \begin{cases} \dfrac{\mathrm{tr}_i - \mathrm{tr}^*}{\Delta_{\mathrm{tr}}}, & \Delta_{\mathrm{tr}} > 0, \\[2mm] 0, & \Delta_{\mathrm{tr}} = 0, \end{cases} \qquad S_i := \begin{cases} \dfrac{\mathrm{MS}_i - \mathrm{MS}^*}{\Delta_{\mathrm{MS}}}, & \Delta_{\mathrm{MS}} > 0, \\[2mm] 0, & \Delta_{\mathrm{MS}} = 0. \end{cases}$$

The *Uncertain RATMI index* of $A_i$ is

$$E_i := v\,T_i + (1-v)\,S_i.$$

**Ranking rule:** rank alternatives in descending order of $E_i$.

**Theorem 6.6.3** (Well-definedness and boundedness of U-RATMI)**.** *Under the assumptions of Definition 6.6.2 (in particular, $m, n \ge 2$, $\mathrm{Dom}(M) \ne \emptyset$, and $S_M : \mathrm{Dom}(M) \to (0, \infty)$), all quantities in the U-RATMI procedure are well-defined and finite. Moreover, for each alternative $A_i$,*

$$0 \le T_i \le 1, \qquad 0 \le S_i \le 1, \qquad 0 \le E_i \le 1.$$

*Hence the ranking rule in Definition 6.6.2 is always well-defined.*

*Proof.* Since $S_M$ maps $\mathrm{Dom}(M)$ into $(0, \infty)$, each $x_{ij} > 0$ is finite. Because $\{1, \dots, m\}$ is finite, $x_j^{\max}$ and $x_j^{\min}$ exist and satisfy $x_j^{\max} \ge x_{ij} \ge x_j^{\min} > 0$. Thus, for benefit criteria, $r_{ij} = x_{ij}/x_j^{\max} \in (0,1]$, and for cost criteria, $r_{ij} = x_j^{\min}/x_{ij} \in (0,1]$. With $w_j \ge 0$ and $\sum_j w_j = 1$, each $u_{ij} = w_j r_{ij}$ is finite and lies in $[0,1]$. Therefore each $q_j = \max_i u_{ij}$ exists and is finite.

The quantities $Q_k, Q_h, U_{ik}, U_{ih}$ are square roots of finite sums of squares, hence finite and nonnegative. At least one weight is positive, so at least one $q_j > 0$, which implies $Q_k^2 + Q_h^2 = \sum_{j=1}^n q_j^2 > 0$, hence $M > 0$ and $\mathrm{MS}_i = M_i/M$ is well-defined and finite.

The extrema $\mathrm{tr}^*, \mathrm{tr}^-, \mathrm{MS}^*, \mathrm{MS}^-$ exist because $m$ is finite. If $\Delta_{\mathrm{tr}} > 0$, then $T_i$ is the standard min–max normalization of $\mathrm{tr}_i$, hence $0 \le T_i \le 1$; if $\Delta_{\mathrm{tr}} = 0$, the definition sets $T_i = 0$, so still $0 \le T_i \le 1$. The same argument gives $0 \le S_i \le 1$.

Finally, with $v \in [0,1]$, $E_i = vT_i + (1-v)S_i$ is a convex combination of two numbers in $[0,1]$, so $0 \le E_i \le 1$. Therefore the index $E_i$ exists for every $i$, and sorting by $E_i$ defines a valid ranking. $\qquad \square$

Related concepts of RATMI under uncertainty-aware models are listed in Table 6.7.

Table 6.7: Related concepts of RATMI under uncertainty-aware models.

| $k$ | Related RATMI concept(s) |
|---|---|
| 1 | Fuzzy RATMI |
| 2 | Intuitionistic Fuzzy RATMI |
| 2 | Fermatean Fuzzy RATMI |
| 3 | Neutrosophic RATMI |

## 6.7 Fuzzy RANCOM (Fuzzy RANking COMparison)

RANCOM ranks criteria by importance judgments, builds a comparison matrix from ranks, sums rows, and normalizes to weights [796, 797]. Fuzzy RANCOM encodes linguistic importance as fuzzy numbers, aggregates experts, scores and ranks criteria, then derives normalized weights [65, 798].

**Definition 6.7.1** (Fuzzy RANCOM (PTF-RANCOM) for subjective criteria weighting). [65, 798] Let $\mathcal{C} = \{C_1, \ldots, C_n\}$ be a finite set of criteria and $\mathcal{E} = \{E_1, \ldots, E_r\}$ a finite set of experts. Fix an integer $q \geq 1$.

**(A) Polytopic fuzzy numbers (PTFNs).** A *polytopic fuzzy number (PTFN)* is a triple

$$z = \langle \alpha, \eta, \zeta \rangle \in [0,1]^3 \quad \text{satisfying} \quad \alpha^q + \eta^q + \zeta^q \leq 1,$$

where $\alpha$ is the positive-membership degree, $\eta$ is the neutral-membership degree, and $\zeta$ is the negative-membership degree.

**(B) Score functional.** For $z = \langle \alpha, \eta, \zeta \rangle$, define the (crisp) score

$$\text{Score}(z) := \frac{1 + \alpha^q + \eta^q - \zeta^q}{3} \in \mathbb{R}.$$

**(C) PTF weighted aggregation (PTFWA).** Given PTFNs $z_1 = \langle \alpha_1, \eta_1, \zeta_1 \rangle, \ldots, z_r = \langle \alpha_r, \eta_r, \zeta_r \rangle$ and expert weights $\lambda = (\lambda_1, \ldots, \lambda_r)$ with $\lambda_k \geq 0$ and $\sum_{k=1}^r \lambda_k = 1$, define

$$\text{PTFWA}_\lambda(z_1, \ldots, z_r) := \left\langle \left(1 - \prod_{k=1}^r (1 - \alpha_k^q)^{\lambda_k}\right)^{1/q}, \prod_{k=1}^r \eta_k^{\lambda_k}, \prod_{k=1}^r \zeta_k^{\lambda_k} \right\rangle.$$

**(D) Input data of Fuzzy RANCOM.** Each expert $E_k$ provides, for every criterion $C_j$, a PTFN importance assessment

$$\iota_{jk} = \langle \alpha_{jk}, \eta_{jk}, \zeta_{jk} \rangle \qquad (j = 1, \ldots, n; \; k = 1, \ldots, r),$$

typically obtained by encoding linguistic terms into PTFNs.

**(E) PTF-RANCOM procedure and output weights.**

Step 1. (**Aggregate experts**) For each criterion $C_j$, compute its integrated PTF importance

$$\iota_j := \mathrm{PTFWA}_\lambda(\iota_{j1}, \ldots, \iota_{jr}) \in [0,1]^3.$$

Step 2. (**Score and rank**) Compute $s_j := \mathrm{Score}(\iota_j)$ and define the rank position

$$\varsigma_j := 1 + |\{t \in \{1, \ldots, n\} : s_t > s_j\}| \in \{1, \ldots, n\},$$

so that smaller $\varsigma_j$ means higher importance (ties allowed).

Step 3. (**Ranking comparison matrix**) Define $B = (b_{gj}) \in [0,1]^{n \times n}$ by

$$b_{gj} := \begin{cases} 1, & \varsigma_g < \varsigma_j, \\ 0.5, & \varsigma_g = \varsigma_j, \qquad (g, j = 1, \ldots, n). \\ 0, & \varsigma_g > \varsigma_j, \end{cases}$$

Step 4. (**Row sums**) Define the row-sum vector $h \in \mathbb{R}_{\geq 0}^n$ by

$$h_j := \sum_{g=1}^{n} b_{jg} \qquad (j = 1, \ldots, n).$$

Step 5. (**Normalized criterion weights**) Define the criterion weights

$$w_j := \frac{h_j}{\sum_{t=1}^{n} h_t} \in [0,1], \qquad j = 1, \ldots, n,$$

so that $\sum_{j=1}^{n} w_j = 1$.

The vector $w = (w_1, \ldots, w_n)$ is called the *(PTF-)fuzzy RANCOM weight vector* of $\mathcal{C}$.

**Definition 6.7.2** (Uncertain RANCOM of type $M$). Let $\mathcal{C} = \{C_1, \ldots, C_n\}$ be a finite set of criteria ($n \geq 2$) and $\mathcal{E} = \{E_1, \ldots, E_r\}$ a finite set of experts ($r \geq 1$). Fix an uncertain model $M$ with $\mathrm{Dom}(M) \neq \emptyset$ and an admissible score $S_M$.

**Input (expert uncertain importance assessments).** Each expert $E_k$ provides, for every criterion $C_j$, an uncertain importance degree

$$\iota_{jk} \in \mathrm{Dom}(M) \qquad (j = 1, \ldots, n; \; k = 1, \ldots, r),$$

interpreted as the importance of $C_j$ under the uncertainty model $M$.

**Step 1 (Expert aggregation).** Let $\lambda = (\lambda_1, \ldots, \lambda_r)$ be expert weights with $\lambda_k \geq 0$ and $\sum_{k=1}^{r} \lambda_k = 1$. Fix an aggregation operator

$$\mathrm{Agg}_M : \mathrm{Dom}(M)^r \to \mathrm{Dom}(M),$$

and compute the aggregated importance for each criterion:

$$\iota_j := \mathrm{Agg}_M(\iota_{j1}, \ldots, \iota_{jr}) \in \mathrm{Dom}(M), \qquad j = 1, \ldots, n.$$

(When $r = 1$, take $\mathrm{Agg}_M$ as the identity.)

**Step 2 (Score and induced ranking).** Compute crisp scores

$$s_j := S_M(\iota_j) \in \mathbb{R}, \qquad j = 1, \dots, n,$$

and define the rank position (ties allowed) by

$$\varsigma_j := 1 + \big| \{t \in \{1, \dots, n\} : s_t > s_j\} \big| \in \{1, \dots, n\}.$$

Thus smaller $\varsigma_j$ means higher importance.

**Step 3 (Ranking comparison matrix).** Define $B = (b_{gj}) \in [0,1]^{n \times n}$ by

$$b_{gj} := \begin{cases} 1, & \varsigma_g < \varsigma_j, \\ 0.5, & \varsigma_g = \varsigma_j, \qquad (g, j = 1, \dots, n). \\ 0, & \varsigma_g > \varsigma_j, \end{cases}$$

**Step 4 (Row sums).** Define

$$h_j := \sum_{g=1}^{n} b_{jg} \qquad (j = 1, \dots, n).$$

**Step 5 (Normalized criterion weights).** Define the Uncertain RANCOM weights by

$$w_j := \frac{h_j}{\sum_{t=1}^{n} h_t} \qquad (j = 1, \dots, n),$$

provided that $\sum_{t=1}^{n} h_t > 0$. The vector $w = (w_1, \dots, w_n)^\top$ is called the *Uncertain RANCOM weight vector of type $M$*.

**Theorem 6.7.3** (Well-definedness of Uncertain RANCOM). *Under Definition 6.7.2, assume:*

- $n \geq 2$, $r \geq 1$, *and* $\mathrm{Dom}(M) \neq \emptyset$;
- $\mathrm{Agg}_M : \mathrm{Dom}(M)^r \to \mathrm{Dom}(M)$ *is well-defined on* $\mathrm{Dom}(M)^r$;
- $S_M : \mathrm{Dom}(M) \to \mathbb{R}$ *is admissible.*

*Then:*

(i) *The aggregated importances $\iota_j$ and scores $s_j$ are well-defined for all $j$.*

(ii) *The rank positions $\varsigma_j$ are well-defined integers in $\{1, \dots, n\}$.*

(iii) *The matrix $B$ and the row-sums $h_j$ are well-defined, with*

$$0 \leq b_{gj} \leq 1, \qquad 0 \leq h_j \leq n \quad (g, j = 1, \dots, n).$$

*(iv)* $\sum_{t=1}^{n} h_t > 0$ *always holds, hence the weights $w_j$ are well-defined and satisfy*

$$w_j \geq 0 \quad (j = 1, \ldots, n), \qquad \sum_{j=1}^{n} w_j = 1,$$

*i.e., $w \in \Delta^{n-1} := \{w \in \mathbb{R}_{\geq 0}^n : \sum_{j=1}^n w_j = 1\}$.*

*Proof.* (i) Since each $\iota_{jk} \in \text{Dom}(M)$ and $\text{Agg}_M$ maps $\text{Dom}(M)^r$ into $\text{Dom}(M)$, each $\iota_j = \text{Agg}_M(\iota_{j1}, \ldots, \iota_{jr})$ is well-defined and belongs to $\text{Dom}(M)$. Admissibility of $S_M$ implies each $s_j = S_M(\iota_j)$ is a finite real number.

(ii) For each fixed $j$, the set $\{t : s_t > s_j\}$ is a subset of the finite set $\{1, \ldots, n\}$, so its cardinality is a well-defined integer between 0 and $n-1$. Hence $\varsigma_j = 1 + |\{t : s_t > s_j\}|$ lies in $\{1, \ldots, n\}$.

(iii) For any pair $(g, j)$, exactly one of the three relations $\varsigma_g < \varsigma_j$, $\varsigma_g = \varsigma_j$, or $\varsigma_g > \varsigma_j$ holds, so $b_{gj}$ is well-defined and belongs to $\{0, 0.5, 1\} \subset [0, 1]$. Therefore each row sum $h_j = \sum_{g=1}^{n} b_{jg}$ is well-defined and satisfies $0 \leq h_j \leq n$.

(iv) Consider the diagonal entries $b_{jj}$. Since $\varsigma_j = \varsigma_j$, we have $b_{jj} = 0.5$ for all $j$. Thus

$$\sum_{t=1}^{n} h_t = \sum_{t=1}^{n} \sum_{g=1}^{n} b_{tg} = \sum_{t=1}^{n} b_{tt} + \sum_{t=1}^{n} \sum_{\substack{g=1 \\ g \neq t}}^{n} b_{tg} \ \geq \ \sum_{t=1}^{n} b_{tt} = \frac{n}{2} > 0.$$

Hence the denominator in $w_j = h_j / \sum_{t=1}^{n} h_t$ is strictly positive, so $w_j$ is well-defined. Since $h_j \geq 0$, we have $w_j \geq 0$ for all $j$, and

$$\sum_{j=1}^{n} w_j = \frac{\sum_{j=1}^{n} h_j}{\sum_{t=1}^{n} h_t} = 1.$$

Therefore $w \in \Delta^{n-1}$. $\qquad \square$

Related concepts of RANCOM under uncertainty-aware models are listed in Table 6.8.

Table 6.8: Related concepts of RANCOM under uncertainty-aware models.

| $k$ | Related RANCOM concept(s) |
|---|---|
| 1 | Fuzzy RANCOM |
| 2 | Intuitionistic Fuzzy RANCOM |
| 2 | Fermatean Fuzzy RANCOM |
| 3 | Neutrosophic RANCOM |

## 6.8 Fuzzy AROMAN (Fuzzy Alternative Ranking Order Method accounting for two-step normalization)

AROMAN ranks alternatives using two-step normalization, separates benefit and cost effects, then computes an exponential preference score [799, 800]. Fuzzy AROMAN replaces crisp ratings with fuzzy numbers, performs two-step fuzzy normalization, aggregates weighted effects, and defuzzifies rankings [801].

**Definition 6.8.1** (Fuzzy AROMAN (Alternative Ranking Order Method accounting for two-step normalization)). [801] Let $\mathcal{A} = \{A_1, \ldots, A_m\}$ be a finite set of alternatives and $\mathcal{C} = \{C_1, \ldots, C_n\}$ a finite set of criteria. Partition the criteria into benefit- and cost-type indices

$$J_{\max} \,\dot{\cup}\, J_{\min} = \{1, \ldots, n\},$$

where $j \in J_{\max}$ means "larger is better" and $j \in J_{\min}$ means "smaller is better".

Assume evaluations are given as triangular fuzzy numbers (TFNs)

$$\tilde{x}_{ij} = (l_{ij}, m_{ij}, u_{ij}) \in \mathsf{TFN}_{\geq 0}, \qquad i = 1, \ldots, m, \; j = 1, \ldots, n.$$

If $K \geq 1$ experts provide TFNs $\tilde{x}_{ij}^{(k)}$, set the aggregated fuzzy decision matrix

$$\tilde{x}_{ij} := \frac{1}{K}\Big(\tilde{x}_{ij}^{(1)} \oplus \cdots \oplus \tilde{x}_{ij}^{(K)}\Big).$$

**TFN arithmetic convention (componentwise, positive case).** For TFNs $\tilde{a} = (a_1, a_2, a_3)$, $\tilde{b} = (b_1, b_2, b_3)$ and $\alpha \geq 0$, define

$$\tilde{a} \oplus \tilde{b} := (a_1 + b_1, a_2 + b_2, a_3 + b_3), \qquad \alpha \odot \tilde{a} := (\alpha a_1, \alpha a_2, \alpha a_3).$$

If all TFNs involved are nonnegative, define the (approximate) product and power by

$$\tilde{a} \otimes \tilde{b} := (a_1 b_1, a_2 b_2, a_3 b_3), \qquad \tilde{a}^p := (a_1^p, a_2^p, a_3^p) \; (p > 0),$$

and for a monotone function $f : \mathbb{R}_{\geq 0} \to \mathbb{R}_{\geq 0}$ (e.g. $f = \exp$),

$$f(\tilde{a}) := (f(a_1), f(a_2), f(a_3)).$$

**Step 1 (Normalization No. 1: min–max).** For each criterion $j$, define fuzzy extrema across alternatives by any fixed TFN preorder (e.g. by a score/defuzzification map; see below):

$$\tilde{x}_j^{\min} := \min_{1 \leq i \leq m} \tilde{x}_{ij}, \qquad \tilde{x}_j^{\max} := \max_{1 \leq i \leq m} \tilde{x}_{ij}.$$

Set the first normalized TFNs $\tilde{t}_{ij}$ by

$$\tilde{t}_{ij} := \begin{cases} (\tilde{x}_{ij} \ominus \tilde{x}_j^{\min}) \oslash (\tilde{x}_j^{\max} \ominus \tilde{x}_j^{\min}), & j \in J_{\max}, \\ (\tilde{x}_j^{\max} \ominus \tilde{x}_{ij}) \oslash (\tilde{x}_j^{\max} \ominus \tilde{x}_j^{\min}), & j \in J_{\min}, \end{cases}$$

where $\ominus, \oslash$ denote the chosen TFN subtraction/division rules (any consistent TFN arithmetic may be used).

**Step 2 (Normalization No. 2: vector-type).** Define the second normalized TFNs $\tilde{t}_{ij}^*$ by

$$\tilde{t}_{ij}^* := \begin{cases} \tilde{x}_{ij} \oslash \sqrt{\displaystyle\bigoplus_{p=1}^{m}(\tilde{x}_{pj} \otimes \tilde{x}_{pj})}, & j \in J_{\max}, \\[4mm] \tilde{x}_{ij}^{-1} \oslash \sqrt{\displaystyle\bigoplus_{p=1}^{m}(\tilde{x}_{pj}^{-1} \otimes \tilde{x}_{pj}^{-1})}, & j \in J_{\min}, \end{cases}$$

where $\tilde{x}^{-1}$ is the TFN reciprocal (defined whenever the TFN is strictly positive).

**Step 3 (Aggregated two-step normalization).** Fix $\beta \in [0,1]$ and set

$$\tilde{t}_{ij}^{\mathrm{norm}} := \beta\,\tilde{t}_{ij}\ \oplus\ (1-\beta)\,\tilde{t}_{ij}^{*}.$$

**Step 4 (Weighting).** Let $w = (w_1, \ldots, w_n)$ be nonnegative criterion weights with $\sum_{j=1}^{n} w_j = 1$. Define the weighted normalized TFNs

$$\widehat{\tilde{t}}_{ij} := w_j \odot \tilde{t}_{ij}^{\mathrm{norm}}.$$

**Step 5 (Separate aggregation by criterion type).** Define the fuzzy sums

$$\tilde{L}_i := \bigoplus_{j \in J_{\min}} \widehat{\tilde{t}}_{ij}, \qquad \tilde{A}_i := \bigoplus_{j \in J_{\max}} \widehat{\tilde{t}}_{ij}.$$

**Step 6 (Type-balance exponent and final fuzzy ranking index).** Set the type-balance coefficient

$$\lambda := \sum_{j \in J_{\min}} w_j \in [0,1],$$

and define

$$\widehat{\tilde{L}}_i := \tilde{L}_i^{\lambda}, \qquad \widehat{\tilde{A}}_i := \tilde{A}_i^{1-\lambda}, \qquad \tilde{R}_i := \exp\!\big(\widehat{\tilde{A}}_i \ominus \widehat{\tilde{L}}_i\big).$$

**Step 7 (Defuzzification and ordering).** Fix a TFN score/defuzzification map, e.g. the centroid score

$$\mathrm{Score}(l, m, u) := \frac{l + m + u}{3}.$$

The *Fuzzy AROMAN ranking* is the preorder on $\mathcal{A}$ given by

$$A_p \succeq_{\mathrm{FAROMAN}} A_q \quad \Longleftrightarrow \quad \mathrm{Score}(\tilde{R}_p) \geq \mathrm{Score}(\tilde{R}_q).$$

Any maximizer of $\mathrm{Score}(\tilde{R}_i)$ is called a *Fuzzy AROMAN best alternative*.

We now define *Uncertain AROMAN* (U-AROMAN) by extending AROMAN to a general uncertain model $M$, using a total score map to convert uncertain evaluations into positive real values and then performing the two-step normalization on the induced score matrix.

**Definition 6.8.2** (Uncertain AROMAN of type $M$)**.** Let

$$\mathcal{A} = \{A_1, \ldots, A_m\} \quad \text{and} \quad \mathcal{C} = \{C_1, \ldots, C_n\}$$

be the sets of alternatives and criteria, respectively, where $m, n \in \mathbb{N}$, $m \geq 1$, and $n \geq 2$.

Fix an uncertain model $M$ with degree-domain

$$\mathrm{Dom}(M) \subseteq [0,1]^k$$

for some integer $k \geq 1$. Let

$$X_M = (\mu_{ij})_{m \times n} \in \mathrm{Dom}(M)^{m \times n}$$

be an uncertain decision matrix, where $\mu_{ij} \in \mathrm{Dom}(M)$ is the uncertain evaluation of alternative $A_i$ under criterion $C_j$.

Partition the criterion index set as

$$J_{\max} \,\dot\cup\, J_{\min} = \{1, \dots, n\},$$

where $J_{\max}$ is the set of benefit criteria and $J_{\min}$ is the set of cost criteria.

Fix:

- a total score map

$$\mathrm{Score}_M : \mathrm{Dom}(M) \longrightarrow \mathbb{R}_{>0},$$

- a parameter

$$\beta \in [0,1],$$

- a weight vector

$$w = (w_1, \dots, w_n) \in (0,1)^n, \qquad \sum_{j=1}^n w_j = 1.$$

Define the positive score matrix

$$S = (s_{ij})_{m \times n} \in \mathbb{R}_{>0}^{m \times n}, \qquad s_{ij} := \mathrm{Score}_M(\mu_{ij}).$$

For each criterion $j$, define

$$s_j^{\min} := \min_{1 \leq i \leq m} s_{ij}, \qquad s_j^{\max} := \max_{1 \leq i \leq m} s_{ij}, \qquad \delta_j := s_j^{\max} - s_j^{\min}.$$

Assume $\delta_j > 0$ for all $j$.

**Step 1 (Normalization No. 1: min–max normalization).** Define

$$t_{ij}^{(1)} := \begin{cases} \dfrac{s_{ij} - s_j^{\min}}{\delta_j}, & j \in J_{\max}, \\[2mm] \dfrac{s_j^{\max} - s_{ij}}{\delta_j}, & j \in J_{\min}. \end{cases}$$

**Step 2 (Normalization No. 2: vector-type normalization).** For each criterion $j$, define

$$\nu_j := \sqrt{\sum_{r=1}^{m} s_{rj}^2}, \qquad \eta_j := \sqrt{\sum_{r=1}^{m} s_{rj}^{-2}}.$$

Then define

$$t_{ij}^{(2)} := \begin{cases} \dfrac{s_{ij}}{\nu_j}, & j \in J_{\max}, \\ \dfrac{s_{ij}^{-1}}{\eta_j}, & j \in J_{\min}. \end{cases}$$

**Step 3 (Aggregated two-step normalization).** Define

$$t_{ij} := \beta\, t_{ij}^{(1)} + (1-\beta)\, t_{ij}^{(2)}.$$

**Step 4 (Weighting).** Define

$$u_{ij} := w_j\, t_{ij}, \qquad i = 1, \ldots, m, \ \ j = 1, \ldots, n.$$

**Step 5 (Separate aggregation by criterion type).** Define

$$L_i := \sum_{j \in J_{\min}} u_{ij}, \qquad A_i := \sum_{j \in J_{\max}} u_{ij}, \qquad i = 1, \ldots, m.$$

**Step 6 (Type-balance coefficient and final ranking index).** Define

$$\lambda := \sum_{j \in J_{\min}} w_j.$$

Assume

$$0 < \lambda < 1.$$

The *Uncertain AROMAN ranking index* of alternative $A_i$ is

$$R_i := \exp\!\big(A_i^{1-\lambda} - L_i^{\lambda}\big), \qquad i = 1, \ldots, m.$$

**Step 7 (Ranking).** The U-AROMAN preference relation on $\mathcal{A}$ is defined by

$$A_p \succeq_{\text{UAROMAN}} A_q \quad \Longleftrightarrow \quad R_p \geq R_q.$$

Any alternative

$$A^\star \in \arg \max_{1 \leq i \leq m} R_i$$

is called a *U-AROMAN best alternative.*

**Theorem 6.8.3** (Well-definedness of Uncertain AROMAN). *Under Definition 6.8.2, assume:*

*(A1) $\mathcal{A}$ and $\mathcal{C}$ are finite nonempty sets;*

*(A2) $X_M = (\mu_{ij}) \in \mathrm{Dom}(M)^{m \times n}$;*

*(A3) $\mathrm{Score}_M : \mathrm{Dom}(M) \to \mathbb{R}_{>0}$ is a total map;*

*(A4) $J_{\max} \dot{\cup} J_{\min} = \{1, \ldots, n\}$, with both $J_{\max} \neq \varnothing$ and $J_{\min} \neq \varnothing$;*

*(A5) for every $j \in \{1, \ldots, n\}$,*
$$\delta_j = s_j^{\max} - s_j^{\min} > 0;$$

*(A6) $w = (w_1, \ldots, w_n) \in (0, 1)^n$ satisfies $\sum_{j=1}^n w_j = 1$;*

*(A7) $\beta \in [0, 1]$.*

*Then the following hold:*

*(i) the matrices $S = (s_{ij})$, $T^{(1)} = (t_{ij}^{(1)})$, $T^{(2)} = (t_{ij}^{(2)})$, $T = (t_{ij})$, and $U = (u_{ij})$ are well-defined;*

*(ii) for all $i, j$,*
$$0 \le t_{ij}^{(1)} \le 1, \qquad 0 < t_{ij}^{(2)} \le 1, \qquad 0 \le t_{ij} \le 1, \qquad 0 \le u_{ij} \le w_j;$$

*(iii) for every $i$,*
$$0 \le L_i \le \lambda, \qquad 0 \le A_i \le 1 - \lambda;$$

*(iv) the coefficient $\lambda$ satisfies $0 < \lambda < 1$, each ranking index $R_i$ is a well-defined positive real number, and the relation $\succeq_{\mathrm{UAROMAN}}$ is a total preorder on $\mathcal{A}$;*

*(v) the set*
$$\arg \max_{1 \le i \le m} R_i$$
*is nonempty.*

*Hence Uncertain AROMAN of type $M$ is well-defined.*

*Proof.* By (A2), each $\mu_{ij} \in \mathrm{Dom}(M)$. Since $\mathrm{Score}_M$ is total by (A3),
$$s_{ij} := \mathrm{Score}_M(\mu_{ij})$$
is well-defined and belongs to $\mathbb{R}_{>0}$. Therefore $S = (s_{ij}) \in \mathbb{R}_{>0}^{m \times n}$ is well-defined.

For each fixed criterion $j$, the set $\{s_{1j}, \ldots, s_{mj}\} \subset \mathbb{R}_{>0}$ is finite and nonempty, so $s_j^{\min}$ and $s_j^{\max}$ are well-defined. By (A5), $\delta_j > 0$, hence the min–max normalized values
$$t_{ij}^{(1)} = \begin{cases} \dfrac{s_{ij} - s_j^{\min}}{\delta_j}, & j \in J_{\max}, \\ \dfrac{s_j^{\max} - s_{ij}}{\delta_j}, & j \in J_{\min} \end{cases}$$

are well-defined. Since $s_j^{\min} \le s_{ij} \le s_j^{\max}$, it follows that

$$0 \le t_{ij}^{(1)} \le 1.$$

Next, because each $s_{rj} > 0$, the sums

$$\sum_{r=1}^{m} s_{rj}^2 \qquad \text{and} \qquad \sum_{r=1}^{m} s_{rj}^{-2}$$

are strictly positive, so $\nu_j > 0$ and $\eta_j > 0$. Hence

$$t_{ij}^{(2)} = \begin{cases} \dfrac{s_{ij}}{\nu_j}, & j \in J_{\max}, \\ \dfrac{s_{ij}^{-1}}{\eta_j}, & j \in J_{\min} \end{cases}$$

is well-defined for every $i, j$. Moreover,

$$\nu_j^2 = \sum_{r=1}^{m} s_{rj}^2 \ge s_{ij}^2 \implies 0 < \frac{s_{ij}}{\nu_j} \le 1,$$

and similarly

$$\eta_j^2 = \sum_{r=1}^{m} s_{rj}^{-2} \ge s_{ij}^{-2} \implies 0 < \frac{s_{ij}^{-1}}{\eta_j} \le 1.$$

Thus

$$0 < t_{ij}^{(2)} \le 1.$$

Since $\beta \in [0,1]$, the aggregated normalized value

$$t_{ij} = \beta t_{ij}^{(1)} + (1-\beta) t_{ij}^{(2)}$$

is a convex combination of numbers in $[0,1]$, hence

$$0 \le t_{ij} \le 1.$$

Because $w_j > 0$ by (A6), the weighted value

$$u_{ij} = w_j t_{ij}$$

is well-defined and satisfies

$$0 \le u_{ij} \le w_j.$$

This proves (i) and (ii).

Now define

$$L_i = \sum_{j \in J_{\min}} u_{ij}, \qquad A_i = \sum_{j \in J_{\max}} u_{ij}.$$

Since $0 \le u_{ij} \le w_j$, we obtain

$$0 \le L_i \le \sum_{j \in J_{\min}} w_j = \lambda, \qquad 0 \le A_i \le \sum_{j \in J_{\max}} w_j = 1 - \lambda.$$

This proves (iii).

Because $J_{\min} \neq \varnothing$, $J_{\max} \neq \varnothing$, and every $w_j > 0$, we have

$$0 < \lambda = \sum_{j \in J_{\min}} w_j < 1.$$

Hence the exponents $\lambda$ and $1 - \lambda$ are strictly positive. Since $L_i, A_i \geq 0$, the powers $L_i^\lambda$ and $A_i^{1-\lambda}$ are well-defined nonnegative real numbers. Therefore

$$R_i = \exp\!\big(A_i^{1-\lambda} - L_i^\lambda\big)$$

is well-defined and strictly positive for every $i$.

The relation

$$A_p \succeq_{\text{UAROMAN}} A_q \iff R_p \geq R_q$$

is induced by the usual order on $\mathbb{R}$, so it is reflexive, transitive, and total. Hence it is a total preorder on $\mathcal{A}$. This proves (iv).

Finally, since $\mathcal{A}$ is finite and nonempty, the finite set

$$\{R_1, \ldots, R_m\} \subset \mathbb{R}_{>0}$$

attains a maximum. Therefore

$$\arg \max_{1 \leq i \leq m} R_i \neq \varnothing.$$

This proves (v), and hence U-AROMAN is well-defined. $\qquad\square$

Related concepts of AROMAN under uncertainty-aware models are listed in Table 6.9.

Table 6.9: Related concepts of AROMAN under uncertainty-aware models.

| $k$ | Related AROMAN concept(s) |
|---|---|
| 1 | Fuzzy AROMAN |
| 2 | Intuitionistic Fuzzy AROMAN |
| 2 | Fermatean Fuzzy AROMAN |
| 3 | Neutrosophic AROMAN |

## 6.9 Fuzzy MAUT (Fuzzy multi-attribute utility theory)

MAUT (multi-attribute utility theory) models preferences with single-attribute utility functions and trade-off weights, combining them (often additively) into an overall utility used to choose the maximum-utility alternative option [802–804]. Fuzzy MAUT represents utilities, weights, or outcomes as fuzzy sets/numbers, computing fuzzy expected utilities and then defuzzifying or using dominance to select under vague assessments [805–807].

**Definition 6.9.1** (Fuzzy MAUT (fuzzy multi-attribute utility theory)). (cf. [808,809]) Let $\mathcal{A} = \{A_1, \ldots, A_m\}$ be a finite set of alternatives and $\mathcal{C} = \{C_1, \ldots, C_K\}$ a finite set of attributes (criteria).

**(0) Fuzzy numbers and defuzzification.** Fix a class $\mathsf{FN}(\mathbb{R})$ of fuzzy numbers on $\mathbb{R}$ (e.g. triangular or trapezoidal), and let $\mathsf{FN}([0,1])$ denote fuzzy numbers supported in $[0,1]$. Fix a defuzzification (score) functional

$$\mathrm{Defuzz} : \mathsf{FN}([0,1]) \to [0,1]$$

(e.g. centroid/COA), used only for final ranking.

**(1) Fuzzy performance ratings.** A *fuzzy MAUT instance* specifies, for each alternative $A_i$ and attribute $C_k$, a fuzzy rating

$$\tilde{x}_{ik} \in \mathsf{FN}(\mathbb{R}), \qquad i = 1, \ldots, m, \;\; k = 1, \ldots, K,$$

interpreted as the uncertain value of attribute $C_k$ for alternative $A_i$.

**(2) Attribute utility functions.** For each attribute $C_k$, fix a (normalized) utility function

$$u_k : \mathbb{R} \to [0,1].$$

Its *fuzzy extension* maps fuzzy ratings to fuzzy utilities:

$$\tilde{u}_{ik} := u_k(\tilde{x}_{ik}) \in \mathsf{FN}([0,1]),$$

defined by Zadeh's extension principle. Equivalently, in $\alpha$-cut form: if $(\tilde{x}_{ik})_\alpha = [x_{ik}^L(\alpha), x_{ik}^U(\alpha)]$, then

$$(\tilde{u}_{ik})_\alpha = \Big[ \min_{t \in [x_{ik}^L(\alpha), x_{ik}^U(\alpha)]} u_k(t), \; \max_{t \in [x_{ik}^L(\alpha), x_{ik}^U(\alpha)]} u_k(t) \Big] \quad (\alpha \in (0,1]).$$

(In particular, if $u_k$ is monotone increasing, then $(\tilde{u}_{ik})_\alpha = [u_k(x_{ik}^L(\alpha)), u_k(x_{ik}^U(\alpha))]$.)

**(3) Weights.** Either use crisp weights $w = (w_1, \ldots, w_K)$ with $w_k \geq 0$ and $\sum_{k=1}^K w_k = 1$, or use fuzzy weights $\tilde{w}_k \in \mathsf{FN}([0,1])$ (with a chosen normalization rule).

**(4) Overall fuzzy utility (additive MAUT aggregation).** Using fuzzy arithmetic on $\mathsf{FN}([0,1])$, define the overall fuzzy utility of $A_i$ by

$$\tilde{U}(A_i) := \begin{cases} \displaystyle\bigoplus_{k=1}^K (w_k \odot \tilde{u}_{ik}), & \text{(crisp weights)}, \\ \displaystyle\bigoplus_{k=1}^K (\tilde{w}_k \otimes \tilde{u}_{ik}), & \text{(fuzzy weights)}, \end{cases} \qquad i = 1, \ldots, m,$$

where $\oplus$ is fuzzy addition, $\odot$ is scalar multiplication, and $\otimes$ is fuzzy multiplication (or another chosen t-norm-like weighting operation).

**(5) Ranking and solution.** Defuzzify the overall utilities:

$$U_i := \mathrm{Defuzz}\big(\tilde{U}(A_i)\big) \in [0,1], \qquad i = 1, \ldots, m,$$

and define the induced preorder

$$A_i \succeq_{\text{FMAUT}} A_j \quad \Longleftrightarrow \quad U_i \geq U_j.$$

Any maximizer

$$A^\star \in \arg\max_{A_i \in \mathcal{A}} U_i$$

is called a *Fuzzy MAUT* (best-utility) solution.

**(Reduction to classical MAUT).** If each fuzzy rating is degenerate $\tilde{x}_{ik} = (x_{ik}, x_{ik}, x_{ik})$ (or the crisp singleton), and weights are crisp, then $\tilde{u}_{ik}$ becomes crisp $u_k(x_{ik})$ and the model reduces to

$$U(A_i) = \sum_{k=1}^{K} w_k \, u_k(x_{ik}).$$

Using Uncertain Sets, we define Uncertain MAUT (U-MAUT) as follows.

**Definition 6.9.2** (Uncertain MAUT (U-MAUT): multi-attribute utility with uncertain inputs). Let $\mathcal{A} = \{A_1, \ldots, A_m\}$ be a finite set of alternatives and $\mathcal{C} = \{C_1, \ldots, C_K\}$ a finite set of attributes (criteria). Fix an uncertainty space $(\Gamma, \mathcal{L}, \mathcal{M})$.

**(1) Uncertain performance ratings (attribute outcomes).** For each alternative $A_i$ and attribute $C_k$, the attribute outcome is modeled by an uncertain variable

$$x_{ik} : \Gamma \to \mathbb{R}, \qquad i = 1, \ldots, m, \ \ k = 1, \ldots, K.$$

Equivalently, each $x_{ik}$ induces an uncertain set $\mathcal{X}_{ik}(\gamma) := \{x_{ik}(\gamma)\}$.

**(2) Single-attribute utility functions and uncertain utilities.** For each attribute $C_k$, fix a utility function

$$u_k : \mathbb{R} \to [0, 1].$$

Define the *uncertain utility* of alternative $A_i$ under attribute $C_k$ by composition:

$$U_{ik}(\gamma) := (u_k \circ x_{ik})(\gamma) = u_k\big(x_{ik}(\gamma)\big) \in [0, 1].$$

Thus $U_{ik} : \Gamma \to [0, 1]$ is an uncertain variable and induces the uncertain set $\mathcal{U}_{ik}(\gamma) := \{U_{ik}(\gamma)\}$.

**(3) Attribute weights (crisp or uncertain).** Either:

(Wc) *crisp weights:* $w = (w_1, \ldots, w_K) \in [0, 1]^K$ with $\sum_{k=1}^{K} w_k = 1$; or

(Wu) *uncertain weights:* uncertain variables $w_k : \Gamma \to [0, 1]$ such that

$$\sum_{k=1}^{K} w_k(\gamma) = 1 \quad \text{for all } \gamma \in \Gamma.$$

**(4) Overall uncertain utility (additive MAUT aggregation).** Define the overall utility of $A_i$ by

$$U_i(\gamma) := \begin{cases} \displaystyle\sum_{k=1}^{K} w_k\, U_{ik}(\gamma), & \text{under (Wc),} \\ \displaystyle\sum_{k=1}^{K} w_k(\gamma)\, U_{ik}(\gamma), & \text{under (Wu),} \end{cases} \qquad \gamma \in \Gamma.$$

Each $U_i : \Gamma \to [0,1]$ induces an uncertain set

$$\mathcal{U}(A_i)(\gamma) := \{U_i(\gamma)\} \subseteq [0,1].$$

**(5) Decision rule (expected-utility ranking).** If $\mathbb{E}[U_i]$ exists for all $i$, define the (crisp) MAUT ranking by

$$A_p \succeq_{\text{U-MAUT}} A_q \quad \Longleftrightarrow \quad \mathbb{E}[U_p] \geq \mathbb{E}[U_q],$$

and select any

$$A^\star \in \arg\max_{A_i \in \mathcal{A}} \mathbb{E}[U_i].$$

**Theorem 6.9.3** (Uncertain-set structure and well-definedness of U-MAUT). *In Definition 6.9.2, assume:*

(A1) *(Measurability of outcomes) Each* $x_{ik} : \Gamma \to \mathbb{R}$ *is* $\mathcal{L}$*-measurable.*

(A2) *(Utility regularity) Each* $u_k : \mathbb{R} \to [0,1]$ *is Borel-measurable.*

(A3) *(Weights) Either* (Wc) *holds, or* (Wu) *holds with each* $w_k$ $\mathcal{L}$*-measurable.*

*Then:*

(i) *For all* $i, k$, $U_{ik} = u_k \circ x_{ik}$ *is an uncertain variable taking values in* $[0,1]$, *hence* $\mathcal{U}_{ik}(\gamma) = \{U_{ik}(\gamma)\}$ *is an uncertain set.*

(ii) *For all* $i$, *the aggregated utility* $U_i$ *is an uncertain variable taking values in* $[0,1]$, *hence* $\mathcal{U}(A_i)(\gamma) = \{U_i(\gamma)\}$ *is an uncertain set.*

(iii) *If, additionally,* $\mathbb{E}[U_i]$ *exists for all* $i$ *(for example, if the underlying expectation is defined for all bounded uncertain variables), then the expected-utility ranking in Definition 6.9.2 is well-defined.*

*Proof.* (i) Fix $i, k$. By (A1) $x_{ik}$ is $\mathcal{L}$-measurable, and by (A2) $u_k$ is Borel-measurable. Therefore the composition $U_{ik} = u_k \circ x_{ik}$ is $\mathcal{L}$-measurable, hence an uncertain variable. Since $u_k(\mathbb{R}) \subseteq [0,1]$, one has $U_{ik}(\gamma) \in [0,1]$ for all $\gamma$. Thus $\mathcal{U}_{ik}(\gamma) = \{U_{ik}(\gamma)\}$ defines a singleton-valued uncertain set.

(ii) Fix $i$. Under (Wc), $U_i(\gamma) = \sum_{k=1}^{K} w_k U_{ik}(\gamma)$ is a finite linear combination of $\mathcal{L}$-measurable functions, hence $\mathcal{L}$-measurable. Under (Wu), each product $w_k(\gamma)U_{ik}(\gamma)$ is $\mathcal{L}$-measurable (product of measurable functions), and the finite sum is measurable; hence $U_i$ is an uncertain variable.

Moreover, in both cases $w_k(\gamma) \geq 0$ and $\sum_{k=1}^{K} w_k(\gamma) = 1$ (with the crisp case interpreted as constant functions), and $0 \leq U_{ik}(\gamma) \leq 1$. Hence

$$0 \leq U_i(\gamma) = \sum_{k=1}^{K} w_k(\gamma) U_{ik}(\gamma) \leq \sum_{k=1}^{K} w_k(\gamma) \cdot 1 = 1,$$

so $U_i(\gamma) \in [0, 1]$ for all $\gamma$. Thus $\mathcal{U}(A_i)(\gamma) = \{U_i(\gamma)\}$ is a well-defined uncertain set.

(iii) If $\mathbb{E}[U_i]$ exists for all $i$, then each $\mathbb{E}[U_i]$ is a finite real number, so the preorder $A_p \succeq_{\text{U-MAUT}} A_q \iff \mathbb{E}[U_p] \geq \mathbb{E}[U_q]$ is well-defined and the argmax set is nonempty because $\mathcal{A}$ is finite. $\square$

Related concepts of MAUT under uncertainty-aware models are listed in Table 6.10.

Table 6.10: Related concepts of MAUT under uncertainty-aware models.

| $k$ | Related MAUT concept(s) |
|---|---|
| 1 | Fuzzy MAUT |
| 2 | Intuitionistic Fuzzy MAUT |
| 2 | Fermatean Fuzzy MAUT |
| 3 | Neutrosophic MAUT |

## 6.10 Fuzzy SMART (Fuzzy Simple Multi-Attribute Rating Technique)

SMART (Simple Multi-Attribute Rating Technique) assigns each criterion a value scale, elicits swing weights for importance, and computes a simple weighted-sum score for each alternative to rank them overall [810, 811]. Fuzzy SMART uses linguistic ratings and fuzzy weights (e.g., triangular numbers), aggregates via fuzzy arithmetic, and derives rankings by defuzzification or comparing fuzzy scores directly [812, 813].

**Definition 6.10.1** (Fuzzy SMART (fuzzy Simple Multi-Attribute Rating Technique)). [812, 813] Let $\mathcal{A} = \{A_1, \ldots, A_m\}$ be a finite set of alternatives and $\mathcal{C} = \{C_1, \ldots, C_n\}$ a finite set of criteria. Partition criteria into benefit and cost sets

$$\mathcal{C} = \mathcal{C}^+ \,\dot{\cup}\, \mathcal{C}^-.$$

Let $P = \{1, \ldots, p\}$ be the set of decision makers (DMs).

**(0) TFNs, fuzzy arithmetic, and defuzzification.** Let $\mathsf{TFN} := \{(l, m, u) \in \mathbb{R}^3 : l \leq m \leq u\}$ be the set of triangular fuzzy numbers (TFNs). For $\tilde{x} = (l_x, m_x, u_x), \tilde{y} = (l_y, m_y, u_y) \in \mathsf{TFN}$ and $\alpha \geq 0$, define

$$\tilde{x} \oplus \tilde{y} := (l_x + l_y, \ m_x + m_y, \ u_x + u_y), \qquad \alpha \odot \tilde{x} := (\alpha l_x, \ \alpha m_x, \ \alpha u_x).$$

Fix a defuzzification map (e.g. centroid/COA)

$$\text{Defuzz}(l, m, u) := \frac{l + m + u}{3}.$$

(When TFNs are used, centroid-type defuzzification is standard in fuzzy SMART implementations.)

**(1) Fuzzy criterion weights (group aggregation).** Each DM $t \in P$ assigns a TFN weight $\tilde{w}_j^{(t)} \in \mathsf{TFN}_{\geq 0}$ to criterion $C_j$. Define the aggregated fuzzy weight by the TFN mean

$$\tilde{w}_j := \frac{1}{p} \bigoplus_{t=1}^{p} \tilde{w}_j^{(t)} \in \mathsf{TFN}_{\geq 0}, \qquad j = 1, \dots, n.$$

Defuzzify and normalize to obtain crisp SMART weights

$$\hat{w}_j := \mathrm{Defuzz}(\tilde{w}_j), \qquad w_j := \frac{\hat{w}_j}{\sum_{r=1}^{n} \hat{w}_r}, \qquad j = 1, \dots, n,$$

so that $w_j \geq 0$ and $\sum_{j=1}^{n} w_j = 1$.

**(2) Fuzzy ratings and aggregation.** Each DM $t \in P$ provides a TFN rating $\tilde{x}_{ij}^{(t)} \in \mathsf{TFN}_{\geq 0}$ for $A_i$ under $C_j$. Aggregate by the TFN mean:

$$\tilde{x}_{ij} := \frac{1}{p} \bigoplus_{t=1}^{p} \tilde{x}_{ij}^{(t)} \in \mathsf{TFN}_{\geq 0}, \qquad i = 1, \dots, m, \ j = 1, \dots, n.$$

Thus $\tilde{X} = (\tilde{x}_{ij})$ is the aggregated fuzzy decision matrix.

**(3) Normalization (benefit/cost).** Choose a crisp ranking of TFNs via Defuzz and define, for each criterion $j$,

$$\underline{x}_j := \min_{1 \leq i \leq m} \mathrm{Defuzz}(\tilde{x}_{ij}), \qquad \overline{x}_j := \max_{1 \leq i \leq m} \mathrm{Defuzz}(\tilde{x}_{ij}), \qquad \Delta_j := \overline{x}_j - \underline{x}_j > 0.$$

Define the (crisp) normalized ratings $r_{ij} \in [0, 1]$ by

$$r_{ij} := \begin{cases} \dfrac{\mathrm{Defuzz}(\tilde{x}_{ij}) - \underline{x}_j}{\Delta_j}, & C_j \in \mathcal{C}^+, \\[2ex] \dfrac{\overline{x}_j - \mathrm{Defuzz}(\tilde{x}_{ij})}{\Delta_j}, & C_j \in \mathcal{C}^-. \end{cases}$$

(Equivalently, one may normalize directly at the TFN level and then defuzzify; both variants appear in fuzzy SMART practice.)

**(4) SMART aggregation (simple additive weighting).** The SMART score of alternative $A_i$ is

$$S_i := \sum_{j=1}^{n} w_j \, r_{ij}, \qquad i = 1, \dots, m,$$

which is the classical SAW-type synthesis used by SMART.

**(5) Ranking and solution.** Rank alternatives by descending $S_i$:

$$A_p \succeq_{\mathrm{FSMART}} A_q \iff S_p \geq S_q,$$

and any maximizer

$$A^\star \in \arg \max_{1 \leq i \leq m} S_i$$

is called a *Fuzzy SMART* solution.

**(Reduction to classical SMART).** If all TFNs are degenerate (crisp) and $p = 1$, then $\tilde{x}_{ij}$ and $\tilde{w}_j$ reduce to crisp values, and the above procedure reduces to the standard SMART/SAW scoring formula.

Using Uncertain Sets, we define Uncertain SMART (U-SMART) as follows.

**Definition 6.10.2** (Uncertain SMART (U-SMART): uncertain Simple Multi-Attribute Rating Technique).
Let $\mathcal{A} = \{A_1, \ldots, A_m\}$ be a finite set of alternatives and $\mathcal{C} = \{C_1, \ldots, C_n\}$ a finite set of criteria. Fix an uncertainty space $(\Gamma, \mathcal{L}, \mathcal{M})$.

**(1) Uncertain raw ratings (attribute performances).** For each alternative $A_i$ and criterion $C_j$, the (raw) performance is modeled by an uncertain variable

$$X_{ij} : \Gamma \to \mathbb{R}, \qquad i = 1, \ldots, m, \ \ j = 1, \ldots, n.$$

Thus each $X_{ij}$ induces the uncertain set $\mathcal{X}_{ij}(\gamma) := \{X_{ij}(\gamma)\}$.

**(2) Value functions and uncertain value scores.** For each criterion $C_j$, fix a Borel-measurable *value function*

$$v_j : \mathbb{R} \to [0, 1].$$

Define the uncertain value score

$$V_{ij}(\gamma) := (v_j \circ X_{ij})(\gamma) = v_j\big(X_{ij}(\gamma)\big) \in [0, 1].$$

Then $V_{ij} : \Gamma \to [0, 1]$ is an uncertain variable inducing $\mathcal{V}_{ij}(\gamma) := \{V_{ij}(\gamma)\}$.

**(3) Swing weights (crisp or uncertain).** Either:

(Wc) *crisp weights:* $w = (w_1, \ldots, w_n) \in [0, 1]^n$ with $\sum_{j=1}^n w_j = 1$; or

(Wu) *uncertain weights:* uncertain variables $W_j : \Gamma \to [0, 1]$ such that

$$\sum_{j=1}^n W_j(\gamma) = 1 \quad \text{for all } \gamma \in \Gamma.$$

**(4) Overall uncertain SMART score (simple weighted sum).** Define the overall SMART score of alternative $A_i$ by

$$S_i(\gamma) := \begin{cases} \displaystyle\sum_{j=1}^n w_j \, V_{ij}(\gamma), & \text{under (Wc)}, \\ \displaystyle\sum_{j=1}^n W_j(\gamma) \, V_{ij}(\gamma), & \text{under (Wu)}, \end{cases} \qquad \gamma \in \Gamma.$$

Each $S_i : \Gamma \to [0, 1]$ induces the uncertain set

$$\mathcal{S}(A_i)(\gamma) := \{S_i(\gamma)\} \subseteq [0, 1].$$

**(5) Decision rule (expected SMART score).** If $\mathbb{E}[S_i]$ exists for all $i$, define the ranking

$$A_p \succeq_{\text{U-SMART}} A_q \quad \Longleftrightarrow \quad \mathbb{E}[S_p] \geq \mathbb{E}[S_q],$$

and select any maximizer

$$A^\star \in \arg\max_{A_i \in \mathcal{A}} \mathbb{E}[S_i].$$

**Theorem 6.10.3** (Uncertain-set structure and well-definedness of U-SMART)**.** *In Definition 6.10.2, assume:*

(A1) *Each $X_{ij} : \Gamma \to \mathbb{R}$ is $\mathcal{L}$-measurable.*

(A2) *Each value function $v_j : \mathbb{R} \to [0,1]$ is Borel-measurable.*

(A3) *Either* (Wc) *holds, or* (Wu) *holds and each $W_j$ is $\mathcal{L}$-measurable.*

*Then:*

(i) *For all $i, j$, $V_{ij} = v_j \circ X_{ij}$ is an uncertain variable with values in $[0,1]$, so $\mathcal{V}_{ij}(\gamma) = \{V_{ij}(\gamma)\}$ is an uncertain set.*

(ii) *For all $i$, the aggregated score $S_i$ is an uncertain variable with values in $[0,1]$, so $\mathcal{S}(A_i)(\gamma) = \{S_i(\gamma)\}$ is an uncertain set.*

(iii) *If, additionally, $\mathbb{E}[S_i]$ exists for all $i$ (e.g., if expectation is defined for all bounded uncertain variables), then the expected-score ranking in Definition 6.10.2 is well-defined and $\arg\max_{A_i \in \mathcal{A}} \mathbb{E}[S_i] \neq \varnothing$.*

*Proof.* (i) Fix $i, j$. By (A1) $X_{ij}$ is $\mathcal{L}$-measurable, and by (A2) $v_j$ is Borel-measurable. Hence $V_{ij} = v_j \circ X_{ij}$ is $\mathcal{L}$-measurable, i.e., an uncertain variable. Since $v_j(\mathbb{R}) \subseteq [0,1]$, $V_{ij}(\gamma) \in [0,1]$ for all $\gamma$. Therefore $\mathcal{V}_{ij}(\gamma) = \{V_{ij}(\gamma)\}$ is a well-defined uncertain set.

(ii) Fix $i$. Under (Wc), $S_i(\gamma) = \sum_{j=1}^{n} w_j V_{ij}(\gamma)$ is a finite linear combination of measurable functions, hence measurable. Under (Wu), each product $W_j(\gamma) V_{ij}(\gamma)$ is measurable and the finite sum remains measurable; hence $S_i$ is an uncertain variable.

Moreover, in both cases the weights are nonnegative and sum to 1 (pointwise in the uncertain-weight case), and $0 \leq V_{ij}(\gamma) \leq 1$. Thus

$$0 \leq S_i(\gamma) = \sum_{j=1}^{n} \mathrm{weight}_j(\gamma)\, V_{ij}(\gamma) \leq \sum_{j=1}^{n} \mathrm{weight}_j(\gamma) \cdot 1 = 1,$$

so $S_i(\gamma) \in [0,1]$ for all $\gamma$. Hence $\mathcal{S}(A_i)(\gamma) = \{S_i(\gamma)\}$ is an uncertain set.

(iii) If each $\mathbb{E}[S_i]$ exists and is finite, then the relation $A_p \succeq_{\text{U-SMART}} A_q \iff \mathbb{E}[S_p] \geq \mathbb{E}[S_q]$ is a well-defined preorder. Since $\mathcal{A}$ is finite, $\max_{A_i \in \mathcal{A}} \mathbb{E}[S_i]$ is attained, so the argmax set is nonempty. $\qquad\square$

Related concepts of SMART under uncertainty-aware models are listed in Table 6.11.

Table 6.11: Related concepts of SMART under uncertainty-aware models.

| $k$ | Related SMART concept(s) |
|---|---|
| 1 | Fuzzy SMART |
| 2 | Intuitionistic Fuzzy SMART |
| 2 | Fermatean Fuzzy SMART |
| 3 | Neutrosophic SMART |

## 6.11  Fuzzy REGIME

REGIME compares alternatives pairwise on each criterion, records win/lose/tie signs, multiplies by weights, and sums to build a preference matrix for ranking across all criteria. Fuzzy REGIME replaces crisp performances and win/lose signs with fuzzy comparisons, yielding fuzzy preference intensities; aggregation and defuzzification produce rankings that reflect measurement uncertainty better [814].

**Definition 6.11.1** (Fuzzy REGIME (F-REGIME) for MADM). [814] Let $\mathcal{A} = \{A_1, \ldots, A_m\}$ be a finite set of alternatives and $\mathcal{C} = \{C_1, \ldots, C_n\}$ a finite set of criteria. For each criterion $C_j$, let $s_j \in \{+1, -1\}$ indicate its type ($s_j = +1$ for a benefit criterion; $s_j = -1$ for a cost criterion). Let $w = (w_1, \ldots, w_n)$ be criterion weights with

$$w_j \geq 0, \qquad \sum_{j=1}^{n} w_j = 1.$$

Assume a *fuzzy decision matrix* $\widetilde{X} = (\tilde{x}_{ij})_{m \times n}$ with $\tilde{x}_{ij} \in \mathrm{F}$, where F denotes a chosen class of fuzzy numbers on $\mathbb{R}$ (e.g., normal and convex fuzzy numbers).

**(1) Fuzzy pairwise comparator.** For each criterion $C_j$, fix a mapping

$$\kappa_j : \mathrm{F} \times \mathrm{F} \longrightarrow [-1, 1]$$

satisfying the following minimal axioms:

(K1)  (Anti-symmetry)   $\kappa_j(\tilde{a}, \tilde{b}) = -\kappa_j(\tilde{b}, \tilde{a})$ for all $\tilde{a}, \tilde{b} \in \mathrm{F}$.

(K2)  (Crisp consistency)   for crisp $a, b \in \mathbb{R}$ (embedded in F), $\kappa_j(a, b) = \mathrm{sgn}(a - b)$.

**(2) REGIME identifier (criterionwise).** For each ordered pair $(A_i, A_k)$ with $i \neq k$ and each criterion $C_j$, define the (possibly fuzzy/graded) REGIME identifier

$$R_{ik}^{(j)} := s_j \, \kappa_j(\tilde{x}_{ij}, \tilde{x}_{kj}) \in [-1, 1].$$

Collect these into the *REGIME vector*

$$\boldsymbol{R}(A_i, A_k) := \big(R_{ik}^{(1)}, \ldots, R_{ik}^{(n)}\big) \in [-1, 1]^n.$$

Stacking $\boldsymbol{R}(A_i, A_k)$ over all ordered pairs $(i, k)$, $i \neq k$, yields the *REGIME matrix*.

**(3) Guide index (pairwise aggregation).** Define the guide index of $A_i$ over $A_k$ by the weighted sum

$$\mathsf{GI}_{ik} := \sum_{j=1}^{n} w_j\, R_{ik}^{(j)} \in [-1, 1].$$

**(4) Net guide index and ranking.** Define the net guide index of each alternative by

$$\mathsf{NGI}(A_i) := \sum_{\substack{k=1 \\ k \neq i}}^{m} \mathsf{GI}_{ik}.$$

The F-REGIME ranking is obtained by sorting alternatives in descending order of $\mathsf{NGI}(A_i)$ (ties may be broken by a secondary rule, e.g., average rank under individual criteria or a chosen defuzzification tie-break).

**Proposition 6.11.2** (Basic well-definedness). *Under Theorem 6.11.1 and (K1), for all $i \neq k$,*

$$R_{ik}^{(j)} = -R_{ki}^{(j)} \quad (\forall j), \qquad \mathsf{GI}_{ik} = -\mathsf{GI}_{ki}.$$

*Hence $\mathsf{NGI}(A_i)$ is well-defined and measures the net pairwise advantage of $A_i$ over the remaining alternatives under the chosen fuzzy comparator.*

*Proof.* By definition, $R_{ik}^{(j)} = s_j \kappa_j(\tilde{x}_{ij}, \tilde{x}_{kj}) = -s_j \kappa_j(\tilde{x}_{kj}, \tilde{x}_{ij}) = -R_{ki}^{(j)}$ using (K1). Summing with weights yields $\mathsf{GI}_{ik} = -\mathsf{GI}_{ki}$. The formula for $\mathsf{NGI}(A_i)$ is a finite sum, hence well-defined. □

Using Uncertain Sets, we define Uncertain REGIME as follows.

**Definition 6.11.3** (Uncertain REGIME of type $M$ (U-REGIME)). Let $\mathcal{A} = \{A_1, \ldots, A_m\}$ be a finite set of alternatives and $\mathcal{C} = \{C_1, \ldots, C_n\}$ a finite set of criteria, with $m \geq 2$ and $n \geq 1$. For each criterion $C_j$, let $s_j \in \{+1, -1\}$ indicate its type ($s_j = +1$ for benefit; $s_j = -1$ for cost). Let $w = (w_1, \ldots, w_n)$ be weights with

$$w_j \geq 0, \qquad \sum_{j=1}^{n} w_j = 1.$$

Fix an uncertain model $M$ with nonempty degree-domain $\mathrm{Dom}(M) \subseteq [0, 1]^k$. Assume an *uncertain decision matrix*

$$X^{(M)} = \left(x_{ij}^{(M)}\right)_{m \times n}, \qquad x_{ij}^{(M)} \in \mathrm{Dom}(M),$$

where $x_{ij}^{(M)}$ encodes the (uncertain) performance of alternative $A_i$ on criterion $C_j$.

**(1) Criterionwise uncertain comparator.** For each criterion $C_j$, fix a mapping

$$\kappa_j^{(M)} : \mathrm{Dom}(M) \times \mathrm{Dom}(M) \longrightarrow [-1, 1]$$

satisfying the minimal axioms:

(U1) (Anti-symmetry) $\kappa_j^{(M)}(a, b) = -\kappa_j^{(M)}(b, a)$ for all $a, b \in \mathrm{Dom}(M)$.

(U2)  (Neutrality on equality)   $\kappa_j^{(M)}(a, a) = 0$ for all $a \in \text{Dom}(M)$.

(When $M$ is fuzzy and $\text{Dom}(M) = [0, 1]$, $\kappa_j^{(M)}$ may be a fuzzy/possibilistic comparison intensity; other $M$ admit analogous score-based or possibility-based comparators.)

**(2) REGIME identifiers (criterionwise).** For each ordered pair $(A_i, A_k)$ with $i \neq k$ and each criterion $C_j$, define

$$R_{ik}^{(j)} := s_j \, \kappa_j^{(M)}\big(x_{ij}^{(M)}, x_{kj}^{(M)}\big) \in [-1, 1].$$

Collect them into the *REGIME vector*

$$\boldsymbol{R}(A_i, A_k) := \big(R_{ik}^{(1)}, \ldots, R_{ik}^{(n)}\big) \in [-1, 1]^n.$$

**(3) Guide index (pairwise aggregation).** Define the pairwise guide index of $A_i$ over $A_k$ by

$$\mathsf{GI}_{ik} := \sum_{j=1}^{n} w_j \, R_{ik}^{(j)} \in [-1, 1].$$

**(4) Net guide index and ranking.** Define the net guide index of each alternative by

$$\mathsf{NGI}(A_i) := \sum_{\substack{k=1 \\ k \neq i}}^{m} \mathsf{GI}_{ik} \in \mathbb{R}.$$

The U-REGIME ranking is obtained by sorting alternatives in descending order of $\mathsf{NGI}(A_i)$ (ties may be handled by any fixed secondary rule).

**Theorem 6.11.4** (Well-definedness and anti-symmetry of U-REGIME). *Under Definition 6.11.3, the U-REGIME quantities are well-defined and satisfy:*

*(i) For all $i \neq k$ and all $j$, $R_{ik}^{(j)}$ and $\mathsf{GI}_{ik}$ exist and lie in $[-1, 1]$.*

*(ii) For all $i \neq k$ and all $j$,*
$$R_{ik}^{(j)} = -R_{ki}^{(j)}, \qquad \mathsf{GI}_{ik} = -\mathsf{GI}_{ki}.$$

*(iii) For each $i$, $\mathsf{NGI}(A_i)$ is a well-defined finite real number.*

*Proof.* (i) Since $x_{ij}^{(M)}, x_{kj}^{(M)} \in \text{Dom}(M)$ and $\kappa_j^{(M)}$ maps $\text{Dom}(M) \times \text{Dom}(M)$ into $[-1, 1]$, the value $\kappa_j^{(M)}(x_{ij}^{(M)}, x_{kj}^{(M)})$ is well-defined in $[-1, 1]$. Multiplying by $s_j \in \{\pm 1\}$ preserves the range, hence each $R_{ik}^{(j)} \in [-1, 1]$ exists. The weighted sum defining $\mathsf{GI}_{ik}$ is finite and is a convex combination of numbers in $[-1, 1]$, so $\mathsf{GI}_{ik} \in [-1, 1]$.

(ii) By anti-symmetry (U1),
$$R_{ki}^{(j)} = s_j \, \kappa_j^{(M)}(x_{kj}^{(M)}, x_{ij}^{(M)}) = -s_j \, \kappa_j^{(M)}(x_{ij}^{(M)}, x_{kj}^{(M)}) = -R_{ik}^{(j)}.$$
Multiplying by weights and summing over $j$ yields $\mathsf{GI}_{ki} = -\mathsf{GI}_{ik}$.

(iii) For fixed $i$, $\mathsf{NGI}(A_i) = \sum_{k \neq i} \mathsf{GI}_{ik}$ is a finite sum of finite real numbers (at most $m - 1$ terms), hence is well-defined and finite. $\qquad \square$

As related concepts, Spherical fuzzy REGIME [815] and Neutrosophic REGIME [816] are also known.

## 6.12 Fuzzy TODIM

TODIM ranks by prospect theory: it computes dominance from gains and losses relative to a reference, uses an attenuation parameter for losses, and aggregates dominance [817–819]. Fuzzy TODIM evaluates criteria values as fuzzy numbers, computes fuzzy dominance for gains and losses, and obtains a final ordering via defuzzification or fuzzy preference relations [820].

**Definition 6.12.1** (TFN-based Fuzzy TODIM). Let $\mathcal{A} = \{A_1, \ldots, A_m\}$ be a finite set of alternatives and $\mathcal{C} = \{C_1, \ldots, C_n\}$ a finite set of criteria. Partition $\mathcal{C} = \mathcal{C}^{\mathrm{ben}} \cup \mathcal{C}^{\mathrm{cost}}$ into benefit- and cost-type criteria.

**(1) Triangular fuzzy evaluations.** Assume a TFN decision matrix

$$\widetilde{X} = (\widetilde{x}_{ij})_{m \times n}, \qquad \widetilde{x}_{ij} = (\ell_{ij}, \mu_{ij}, u_{ij}), \qquad 0 < \ell_{ij} \le \mu_{ij} \le u_{ij},$$

where $\widetilde{x}_{ij}$ is the (linguistic-to-fuzzy) assessment of $A_i$ w.r.t. $C_j$.

If multiple decision-makers $e = 1, \ldots, k$ provide TFNs $\widetilde{x}_{ij}^{(e)}$, use the componentwise average aggregation

$$\widetilde{x}_{ij} := \frac{1}{k} \sum_{e=1}^{k} \widetilde{x}_{ij}^{(e)} = \Big( \frac{1}{k} \sum_{e=1}^{k} \ell_{ij}^{(e)}, \ \frac{1}{k} \sum_{e=1}^{k} \mu_{ij}^{(e)}, \ \frac{1}{k} \sum_{e=1}^{k} u_{ij}^{(e)} \Big).$$

**(2) TFN weights and reference criterion.** Let TFN criterion weights be

$$\widetilde{w}_j = (\ell_j^w, \mu_j^w, u_j^w), \qquad 0 < \ell_j^w \le \mu_j^w \le u_j^w, \qquad j = 1, \ldots, n,$$

(aggregated from multiple experts in the same way if needed).

Fix a score/defuzzification map $s : \mathrm{TFN}_{>0} \to \mathbb{R}_{>0}$, e.g.

$$s(\ell, \mu, u) := \frac{\ell + 2\mu + u}{4}.$$

Define the normalized crisp weights

$$w_j := \frac{s(\widetilde{w}_j)}{\sum_{t=1}^{n} s(\widetilde{w}_t)}, \qquad j = 1, \ldots, n.$$

Let the reference criterion be any maximizer

$$r \in \arg \max_{1 \le j \le n} w_j,$$

and define relative weights (projection to the reference criterion)

$$w_{jr} := \frac{w_j}{w_r}, \qquad j = 1, \ldots, n, \qquad W_r := \sum_{t=1}^{n} w_{tr}.$$

**(3) Normalized TFN performances.** Define normalized TFN performances $\widehat{x}_{ij}$ (larger is better) as follows.

For $C_j \in \mathcal{C}^{\mathrm{ben}}$, set

$$\widetilde{x}_j^{\max} := \big(\max_i \ell_{ij}, \ \max_i \mu_{ij}, \ \max_i u_{ij}\big), \qquad \widehat{x}_{ij} := \widetilde{x}_{ij} \oslash \widetilde{x}_j^{\max}.$$

For $C_j \in \mathcal{C}^{\mathrm{cost}}$, set

$$\widetilde{x}_j^{\min} := \big(\min_i \ell_{ij}, \ \min_i \mu_{ij}, \ \min_i u_{ij}\big), \qquad \widehat{x}_{ij} := \widetilde{x}_j^{\min} \oslash \widetilde{x}_{ij}.$$

Here TFN division is the standard positive-TFN operation

$$(\ell_1, \mu_1, u_1) \oslash (\ell_2, \mu_2, u_2) := \Big(\frac{\ell_1}{u_2}, \ \frac{\mu_1}{\mu_2}, \ \frac{u_1}{\ell_2}\Big).$$

**(4) TFN distance (vertex method).** For TFNs $\widetilde{a} = (\ell_a, \mu_a, u_a)$ and $\widetilde{b} = (\ell_b, \mu_b, u_b)$ define

$$d(\widetilde{a}, \widetilde{b}) := \sqrt{\frac{1}{3}\Big[(\ell_a - \ell_b)^2 + (\mu_a - \mu_b)^2 + (u_a - u_b)^2\Big]}.$$

**(5) Gain/Loss matrices (pairwise comparisons).** Use the score $s(\cdot)$ to compare TFNs: write $\widetilde{x} \succeq \widetilde{y}$ iff $s(\widetilde{x}) \geq s(\widetilde{y})$.

For each criterion $C_j$ and each pair $(i, k)$, define gain and loss:

$$G_{ik}^j, \ L_{ik}^j \in \mathbb{R}, \qquad i, k = 1, \dots, m, \qquad j = 1, \dots, n.$$

If $C_j \in \mathcal{C}^{\mathrm{ben}}$ (benefit), set

$$G_{ik}^j := \begin{cases} d(\widehat{x}_{ij}, \widehat{x}_{kj}), & \widetilde{x}_{ij} \succeq \widetilde{x}_{kj}, \\ 0, & \widetilde{x}_{ij} \prec \widetilde{x}_{kj}, \end{cases} \qquad L_{ik}^j := \begin{cases} 0, & \widetilde{x}_{ij} \succeq \widetilde{x}_{kj}, \\ -d(\widehat{x}_{ij}, \widehat{x}_{kj}), & \widetilde{x}_{ij} \prec \widetilde{x}_{kj}. \end{cases}$$

If $C_j \in \mathcal{C}^{\mathrm{cost}}$ (cost), set

$$G_{ik}^j := \begin{cases} 0, & \widetilde{x}_{ij} \succeq \widetilde{x}_{kj}, \\ d(\widehat{x}_{ij}, \widehat{x}_{kj}), & \widetilde{x}_{ij} \prec \widetilde{x}_{kj}, \end{cases} \qquad L_{ik}^j := \begin{cases} -d(\widehat{x}_{ij}, \widehat{x}_{kj}), & \widetilde{x}_{ij} \succeq \widetilde{x}_{kj}, \\ 0, & \widetilde{x}_{ij} \prec \widetilde{x}_{kj}. \end{cases}$$

(Thus $G_{ik}^j \geq 0$ and $L_{ik}^j \leq 0$ always, and $G_{ii}^j = L_{ii}^j = 0$.)

**(6) Dominance degree per criterion (prospect-type asymmetry).** Fix a loss attenuation parameter $\theta > 0$. Define (square-root form) dominance contributions:

$$\Phi_{ik}^{j(+)} := \sqrt{\frac{G_{ik}^j \, w_{jr}}{W_r}}, \qquad \Phi_{ik}^{j(-)} := -\frac{1}{\theta}\sqrt{\frac{(-L_{ik}^j) \, W_r}{w_{jr}}}, \qquad \Phi_{ik}^j := \Phi_{ik}^{j(+)} + \Phi_{ik}^{j(-)}.$$

Let $\Phi^j := (\Phi_{ik}^j)_{m \times m}$ be the dominance matrix for criterion $C_j$.

**(7) Overall dominance and overall value.** Define the overall dominance matrix $\Delta = (\delta_{ik})_{m \times m}$ by

$$\delta_{ik} := \sum_{j=1}^n \Phi_{ik}^j.$$

Define the overall value of alternative $A_i$ by normalizing row-sums:

$$S_i := \sum_{k=1}^m \delta_{ik}, \qquad \Xi(A_i) := \frac{S_i - \min_{1 \leq t \leq m} S_t}{\max_{1 \leq t \leq m} S_t - \min_{1 \leq t \leq m} S_t} \in [0, 1].$$

Rank alternatives by decreasing $\Xi(A_i)$ (ties allowed).

**Proposition 6.12.2** (Well-definedness). *Under the standing assumptions $\widetilde{x}_{ij} \in \mathrm{TFN}_{>0}$ and $\widetilde{w}_j \in \mathrm{TFN}_{>0}$, and with $\theta > 0$, the quantities $G_{ik}^j, L_{ik}^j, \Phi_{ik}^j, \delta_{ik}, \Xi(A_i)$ are well-defined real numbers and $0 \leq \Xi(A_i) \leq 1$ for all $i$.*

*Proof.* The vertex distance $d(\cdot, \cdot)$ is real and nonnegative by construction. Hence each $G_{ik}^j$ is either 0 or a nonnegative real, and each $L_{ik}^j$ is either 0 or a nonpositive real. Since $w_j > 0$ and $w_r = \max_j w_j > 0$, all ratios $w_{jr}$ and $W_r$ are positive, so $\Phi_{ik}^{j(+)}$ and $\Phi_{ik}^{j(-)}$ are real. Therefore $\delta_{ik}$ and $S_i$ are real. Finally, $\Xi(A_i)$ is the standard affine normalization of $\{S_i\}_{i=1}^m$, hence lies in $[0,1]$ whenever the denominator is nonzero; if all $S_i$ are equal, define $\Xi(A_i) := 0$ for all $i$ (or any constant in $[0,1]$). □

Using Uncertain Sets, we define Uncertain TODIM as follows.

**Definition 6.12.3** (Admissible distance on an uncertain model). Let $M$ be an uncertain model. An *admissible distance* is a map

$$d_M : \mathrm{Dom}(M) \times \mathrm{Dom}(M) \longrightarrow [0, \infty)$$

satisfying: (i) $d_M(a, b) = d_M(b, a)$, (ii) $d_M(a, a) = 0$, and (iii) $d_M(a, b) < \infty$ for all $a, b \in \mathrm{Dom}(M)$.

**Definition 6.12.4** (Uncertain TODIM of type $M$ (U-TODIM)). Let $\mathcal{A} = \{A_1, \ldots, A_m\}$ be alternatives and $\mathcal{C} = \{C_1, \ldots, C_n\}$ criteria, with $m \geq 2$ and $n \geq 1$. Partition $\mathcal{C} = \mathcal{C}^{\mathrm{ben}} \,\dot{\cup}\, \mathcal{C}^{\mathrm{cost}}$ into benefit and cost criteria.

Assume an *uncertain decision matrix*

$$X^{(M)} = \big(x_{ij}^{(M)}\big)_{m \times n}, \qquad x_{ij}^{(M)} \in \mathrm{Dom}(M),$$

and *uncertain criterion weights*

$$w_j^{(M)} \in \mathrm{Dom}(M) \qquad (j = 1, \ldots, n).$$

Fix an admissible score $S_M$ and an admissible distance $d_M$.

**Step 1 (Crisp weights and reference criterion).** Define positive crisp weights by

$$\omega_j := \frac{\exp(S_M(w_j^{(M)}))}{\sum_{t=1}^n \exp(S_M(w_t^{(M)}))} \in (0, 1), \qquad \sum_{j=1}^n \omega_j = 1.$$

Choose a reference criterion

$$r \in \arg\max_{1 \leq j \leq n} \omega_j,$$

and define relative weights

$$\omega_{jr} := \frac{\omega_j}{\omega_r} \quad (j = 1, \ldots, n), \qquad W_r := \sum_{t=1}^n \omega_{tr}.$$

**Step 2 (Normalization to a benefit orientation).** For each criterion $C_j$, define the criterionwise best and worst degrees in $\mathrm{Dom}(M)$ by

$$x_j^{\max} \in \arg\max_{1 \leq i \leq m} S_M(x_{ij}^{(M)}), \qquad x_j^{\min} \in \arg\min_{1 \leq i \leq m} S_M(x_{ij}^{(M)}).$$

Define the normalized distance-to-best performance

$$\Delta_i^{(j)} := \begin{cases} d_M(x_{ij}^{(M)}, x_j^{\max}), & C_j \in \mathcal{C}^{\text{ben}}, \\ d_M(x_{ij}^{(M)}, x_j^{\min}), & C_j \in \mathcal{C}^{\text{cost}}, \end{cases} \qquad (i = 1, \dots, m).$$

(Thus, for each criterion, smaller $\Delta_i^{(j)}$ means better.)

**Step 3 (Pairwise gains/losses).** For each criterion $C_j$ and pair $(i, k)$ define

$$G_{ik}^j := \max\{0,\ \Delta_k^{(j)} - \Delta_i^{(j)}\}\ \geq 0, \qquad L_{ik}^j := -\max\{0,\ \Delta_i^{(j)} - \Delta_k^{(j)}\}\ \leq 0.$$

Hence $G_{ik}^j$ measures gain of $A_i$ over $A_k$, while $L_{ik}^j$ is the (negative) loss.

**Step 4 (Prospect-type dominance per criterion).** Fix a loss attenuation parameter $\theta > 0$. Define dominance contributions

$$\Phi_{ik}^{j(+)} := \sqrt{\frac{G_{ik}^j\,\omega_{jr}}{W_r}}, \qquad \Phi_{ik}^{j(-)} := -\frac{1}{\theta}\sqrt{\frac{(-L_{ik}^j)\,W_r}{\omega_{jr}}}, \qquad \Phi_{ik}^j := \Phi_{ik}^{j(+)} + \Phi_{ik}^{j(-)}.$$

Let $\Phi^j := (\Phi_{ik}^j)_{m \times m}$ be the dominance matrix for criterion $C_j$.

**Step 5 (Overall dominance and final value).** Define the overall dominance matrix $\Delta = (\delta_{ik})$ by

$$\delta_{ik} := \sum_{j=1}^n \Phi_{ik}^j.$$

Let

$$S_i := \sum_{k=1}^m \delta_{ik}.$$

Define the final TODIM value by min–max normalization:

$$\Xi(A_i) := \begin{cases} \dfrac{S_i - \min_{1 \leq t \leq m} S_t}{\max_{1 \leq t \leq m} S_t - \min_{1 \leq t \leq m} S_t}, & \max S_t > \min S_t, \\ 0, & \max S_t = \min S_t, \end{cases}$$

and rank alternatives by decreasing $\Xi(A_i)$.

**Theorem 6.12.5** (Well-definedness of U-TODIM). *Under Definition 6.12.4, assume $\text{Dom}(M) \neq \emptyset$, $S_M$ is admissible, $d_M$ is an admissible distance, and $\theta > 0$. Then:*

(i) *All quantities in U-TODIM are well-defined finite real numbers.*

(ii) *The crisp weights satisfy $\omega_j \in (0, 1)$ and $\sum_{j=1}^n \omega_j = 1$, and thus $\omega_{jr} > 0$ and $W_r > 0$.*

(iii) *For each $i$, the final value satisfies $0 \leq \Xi(A_i) \leq 1$.*

*Proof.* (i) Admissibility of $S_M$ implies $S_M(w_j^{(M)}) \in \mathbb{R}$, hence $\exp(S_M(w_j^{(M)})) \in (0, \infty)$ is finite. Therefore the normalization defining $\omega_j$ yields well-defined positive reals with sum 1.

(ii) Since $r$ maximizes $\omega_j$, we have $\omega_r > 0$, hence each ratio $\omega_{jr} = \omega_j/\omega_r$ is well-defined and positive, and $W_r = \sum_t \omega_{tr} > 0$.

(iii) Because $m$ is finite and $S_M(x_{ij}^{(M)})$ are finite, the argmax/argmin sets used to choose $x_j^{\max}, x_j^{\min}$ are nonempty, so the selections exist. By admissibility of $d_M$, each $\Delta_i^{(j)}$ is a finite nonnegative real. Thus $G_{ik}^j$ and $L_{ik}^j$ are well-defined finite reals with $G_{ik}^j \geq 0$ and $L_{ik}^j \leq 0$. With $\theta > 0$, $\omega_{jr} > 0$, and $W_r > 0$, the square-root expressions in $\Phi_{ik}^{j(+)}$ and $\Phi_{ik}^{j(-)}$ are well-defined finite reals, so $\Phi_{ik}^j$ and hence $\delta_{ik}$ and $S_i$ are finite real numbers.

Finally, $\Xi(A_i)$ is obtained by standard min–max normalization. If $\max S_t > \min S_t$, then $0 \leq \Xi(A_i) \leq 1$ for all $i$; if $\max S_t = \min S_t$, the definition sets $\Xi(A_i) = 0$. $\qquad\square$

Related concepts of TODIM under uncertainty-aware models are listed in Table 6.12.

Table 6.12: Related concepts of TODIM under uncertainty-aware models.

| $k$ | Related TODIM concept(s) |
|---|---|
| 2 | Intuitionistic Fuzzy TODIM [821] |
| 2 | Pythagorean Fuzzy TODIM [822–824] |
| 2 | Fermatean Fuzzy TODIM [825, 826] |
| 3 | Hesitant Fuzzy TODIM [827, 828] |
| 3 | Picture Fuzzy TODIM [829] |
| 3 | Spherical Fuzzy TODIM [830, 831] |
| 3 | Neutrosophic TODIM [832–835] |
| $n$ | Plithogenic TODIM [836] |

As related concepts to Uncertain TODIM, several extensions are also known, including Rough TODIM [837, 838], Linguistic TODIM [839, 840], Grey TODIM [841], TODIM-VIKOR [826, 842], and Extended TODIM [843, 844].

## 6.13 Fuzzy GRA (Fuzzy Grey Relational Analysis)

Classical GRA (Grey Relational Analysis) sets a reference sequence, computes grey relational coefficients from absolute deviations, aggregates coefficients with weights, and ranks alternatives by highest relational grade overall [845, 846]. Fuzzy GRA normalizes fuzzy criterion ratings, defines an ideal reference sequence, measures fuzzy deviations, computes grey relational coefficients and weighted grades, then ranks alternatives by similarity [847, 848].

**Definition 6.13.1** (TFN-based Fuzzy Grey Relational Analysis (Fuzzy GRA)). [849, 850] Let $\mathcal{A} = \{A_1, \ldots, A_m\}$ be a finite set of alternatives and $\mathcal{C} = \{C_1, \ldots, C_n\}$ a finite set of criteria. Assume a partition into benefit and cost criteria

$$\mathcal{C} = \mathcal{C}^+ \dot{\cup} \mathcal{C}^-, \qquad J^+ := \{j : C_j \in \mathcal{C}^+\}, \quad J^- := \{j : C_j \in \mathcal{C}^-\}.$$

Let $\mathsf{TFN}_{>0} := \{(l, m, u) \in \mathbb{R}_{>0}^3 : l \leq m \leq u\}$ denote positive triangular fuzzy numbers (TFNs). Suppose the fuzzy decision matrix is

$$\widetilde{X} = (\tilde{x}_{ij}) \in (\mathsf{TFN}_{>0})^{m \times n}, \qquad \tilde{x}_{ij} = (l_{ij}, m_{ij}, u_{ij}),$$

and let $w = (w_1, \ldots, w_n) \in [0, 1]^n$ be criterion weights with $\sum_{j=1}^n w_j = 1$.

**Step 1 (Normalization to a comparable fuzzy scale).** For each criterion $j$, define the positive scalars

$$u_j^{\max} := \max_{1 \leq i \leq m} u_{ij}, \qquad l_j^{\min} := \min_{1 \leq i \leq m} l_{ij}.$$

Define the normalized TFNs $\tilde{r}_{ij} \in (\mathsf{TFN}_{>0})$ by

$$\tilde{r}_{ij} := \begin{cases} \left( \dfrac{l_{ij}}{u_j^{\max}}, \dfrac{m_{ij}}{u_j^{\max}}, \dfrac{u_{ij}}{u_j^{\max}} \right), & j \in J^+ \quad \text{(benefit)}, \\[2ex] \left( \dfrac{l_j^{\min}}{u_{ij}}, \dfrac{l_j^{\min}}{m_{ij}}, \dfrac{l_j^{\min}}{l_{ij}} \right), & j \in J^- \quad \text{(cost)}. \end{cases}$$

(Thus, larger performance yields larger normalized values for both benefit and cost criteria.)

**Step 2 (Reference/ideal sequence).** Define the fuzzy reference value for each criterion by the TFN

$$\tilde{r}_j^* := (1, 1, 1) \in \mathsf{TFN}_{>0}, \qquad j = 1, \ldots, n,$$

and the corresponding reference sequence $\widetilde{R}^* := (\tilde{r}_1^*, \ldots, \tilde{r}_n^*)$.

**Step 3 (Distance to the reference in the fuzzy domain).** Fix a metric (distance) $D : \mathsf{TFN}_{>0} \times \mathsf{TFN}_{>0} \to \mathbb{R}_{\geq 0}$. A standard choice is the vertex-distance:

$$D\big((l_1, m_1, u_1), (l_2, m_2, u_2)\big) := \sqrt{\frac{(l_1 - l_2)^2 + (m_1 - m_2)^2 + (u_1 - u_2)^2}{3}}.$$

Define the deviation (distance to the reference) by

$$\Delta_{ij} := D(\tilde{r}_{ij}, \tilde{r}_j^*) \in \mathbb{R}_{\geq 0}, \qquad i = 1, \ldots, m, \; j = 1, \ldots, n.$$

Let

$$\Delta_{\min} := \min_{1 \leq i \leq m} \min_{1 \leq j \leq n} \Delta_{ij}, \qquad \Delta_{\max} := \max_{1 \leq i \leq m} \max_{1 \leq j \leq n} \Delta_{ij}.$$

**Step 4 (Grey relational coefficients).** Fix the distinguishing coefficient $\rho \in (0, 1]$. If $\Delta_{\max} = 0$ (all alternatives coincide with the reference), define $\gamma_{ij} := 1$ for all $i, j$. Otherwise define the fuzzy grey relational coefficient by

$$\gamma_{ij} := \frac{\Delta_{\min} + \rho \, \Delta_{\max}}{\Delta_{ij} + \rho \, \Delta_{\max}} \in (0, 1], \qquad i = 1, \ldots, m, \; j = 1, \ldots, n.$$

**Step 5 (Grey relational grades and ranking).** Define the grey relational grade (overall similarity) of alternative $A_i$ by the weighted sum

$$\Gamma_i := \sum_{j=1}^n w_j \, \gamma_{ij} \in [0, 1], \qquad i = 1, \ldots, m.$$

Rank alternatives by descending $\Gamma_i$:

$$A_p \succeq_{\text{FGRA}} A_q \quad \Longleftrightarrow \quad \Gamma_p \geq \Gamma_q.$$

Any maximizer

$$A^\star \in \arg \max_{1 \leq i \leq m} \Gamma_i$$

is called a *Fuzzy GRA best alternative*.

The definition of Uncertain GRA of type $M$ is given below.

**Definition 6.13.2** (Uncertain GRA of type $M$ (U-GRA)). Let $\mathcal{A} = \{A_1, \ldots, A_m\}$ be alternatives and $\mathcal{C} = \{C_1, \ldots, C_n\}$ criteria, with $m, n \geq 2$. Partition criteria into benefit and cost sets:

$$\mathcal{C} = \mathcal{C}^{\text{ben}} \mathbin{\dot{\cup}} \mathcal{C}^{\text{cost}}.$$

Assume an *uncertain decision matrix*

$$X^{(M)} = \big(x_{ij}^{(M)}\big)_{m \times n}, \qquad x_{ij}^{(M)} \in \text{Dom}(M),$$

and criterion weights $w = (w_1, \ldots, w_n)$ satisfying $w_j \geq 0$ and $\sum_{j=1}^{n} w_j = 1$. Fix an uncertain model $M$ with $\text{Dom}(M) \neq \emptyset$, together with an admissible score $S_M$ and an admissible distance $d_M$. Fix a distinguishing coefficient $\rho \in (0, 1]$.

**Step 1 (Crisp projection).** Define $y_{ij} := S_M(x_{ij}^{(M)}) \in \mathbb{R}$ and form the real matrix $Y = (y_{ij})$.

**Step 2 (Min–max normalization to a benefit orientation).** For each criterion $C_j$, define

$$y_j^{\min} := \min_{1 \leq i \leq m} y_{ij}, \qquad y_j^{\max} := \max_{1 \leq i \leq m} y_{ij}.$$

Define a normalized scalar matrix $R = (r_{ij})$ by

$$r_{ij} := \begin{cases} \dfrac{y_{ij} - y_j^{\min}}{y_j^{\max} - y_j^{\min}}, & C_j \in \mathcal{C}^{\text{ben}} \text{ and } y_j^{\max} > y_j^{\min}, \\[2ex] \dfrac{y_j^{\max} - y_{ij}}{y_j^{\max} - y_j^{\min}}, & C_j \in \mathcal{C}^{\text{cost}} \text{ and } y_j^{\max} > y_j^{\min}, \\[2ex] 0, & y_j^{\max} = y_j^{\min}, \end{cases} \qquad (i = 1, \ldots, m; \ j = 1, \ldots, n),$$

so that $r_{ij} \in [0, 1]$ and larger values always indicate better performance.

**Step 3 (Reference sequence).** Define the reference (ideal) sequence by

$$r_j^* := 1 \qquad (j = 1, \ldots, n).$$

**Step 4 (Deviation to the reference).** Define

$$\Delta_{ij} := |r_j^* - r_{ij}| = |1 - r_{ij}| \in [0, 1], \qquad \Delta_{\min} := \min_{i,j} \Delta_{ij}, \qquad \Delta_{\max} := \max_{i,j} \Delta_{ij}.$$

**Step 5 (Grey relational coefficients).** If $\Delta_{\max} = 0$, set $\gamma_{ij} := 1$ for all $i, j$. Otherwise define

$$\gamma_{ij} := \frac{\Delta_{\min} + \rho\,\Delta_{\max}}{\Delta_{ij} + \rho\,\Delta_{\max}} \in (0, 1], \qquad i = 1, \dots, m, \; j = 1, \dots, n.$$

**Step 6 (Grey relational grades and ranking).** Define the grey relational grade of alternative $A_i$ by

$$\Gamma_i := \sum_{j=1}^{n} w_j\,\gamma_{ij} \in [0, 1], \qquad i = 1, \dots, m,$$

and rank by descending $\Gamma_i$.

**(Optional) Uncertainty-aware distance refinement.** Instead of $\Delta_{ij} = |1 - r_{ij}|$, one may define an uncertainty-aware deviation using $d_M$, e.g.,

$$\Delta_{ij}^{(M)} := d_M\big(x_{ij}^{(M)}, x_j^{(M)*}\big),$$

where $x_j^{(M)*}$ is a chosen reference degree in $\mathrm{Dom}(M)$ (e.g., an $S_M$-maximizer). The remaining steps are then applied with $\Delta_{ij}^{(M)}$ in place of $\Delta_{ij}$.

**Theorem 6.13.3** (Well-definedness of U-GRA)**.** *Under Definition 6.13.2, assume $m, n \geq 2$, $\mathrm{Dom}(M) \neq \emptyset$, $S_M$ is admissible, and $\rho \in (0, 1]$. Then:*

*(i) The normalized values $r_{ij}$ are well-defined and satisfy $0 \leq r_{ij} \leq 1$.*

*(ii) The deviations satisfy $0 \leq \Delta_{ij} \leq 1$, hence $\Delta_{\min}, \Delta_{\max}$ exist.*

*(iii) The grey relational coefficients $\gamma_{ij}$ are well-defined and satisfy $0 < \gamma_{ij} \leq 1$ when $\Delta_{\max} > 0$, and $\gamma_{ij} = 1$ when $\Delta_{\max} = 0$.*

*(iv) The grey relational grades satisfy $0 \leq \Gamma_i \leq 1$ for all $i$, and the ranking by $\Gamma_i$ is well-defined.*

*Proof.* (i) By admissibility of $S_M$, each $y_{ij} = S_M(x_{ij}^{(M)})$ is a finite real number, so for each $j$ the extrema $y_j^{\min}$ and $y_j^{\max}$ exist (finite index set). If $y_j^{\max} > y_j^{\min}$, the stated min–max formula defines $r_{ij}$. Because $y_{ij} \in [y_j^{\min}, y_j^{\max}]$, it follows that $r_{ij} \in [0, 1]$ for both benefit and cost cases. If $y_j^{\max} = y_j^{\min}$, the definition sets $r_{ij} = 0$, so $r_{ij} \in [0, 1]$ always.

(ii) Since $r_j^* = 1$ and $r_{ij} \in [0, 1]$, we have $\Delta_{ij} = |1 - r_{ij}| \in [0, 1]$. Thus $\Delta_{\min}$ and $\Delta_{\max}$ exist as minima/maxima over a finite set.

(iii) If $\Delta_{\max} = 0$, then all $\Delta_{ij} = 0$ and the definition sets $\gamma_{ij} = 1$. If $\Delta_{\max} > 0$, then $\rho\,\Delta_{\max} > 0$ and the denominator $\Delta_{ij} + \rho\Delta_{\max} \geq \rho\Delta_{\max} > 0$, so $\gamma_{ij}$ is well-defined. Moreover, $\Delta_{ij} \geq \Delta_{\min}$ implies

$$\gamma_{ij} = \frac{\Delta_{\min} + \rho\Delta_{\max}}{\Delta_{ij} + \rho\Delta_{\max}} \leq 1,$$

and positivity of the numerator yields $\gamma_{ij} > 0$.

(iv) Since $w_j \geq 0$, $\sum_j w_j = 1$, and $0 < \gamma_{ij} \leq 1$, the weighted sum satisfies $0 \leq \Gamma_i \leq 1$. Therefore sorting by $\Gamma_i$ defines a valid ranking (ties allowed). $\qquad\square$

Related concepts of GRA under uncertainty-aware models are listed in Table 6.13.

Table 6.13: Related concepts of GRA under uncertainty-aware models.

| $k$ | Related GRA concept(s) |
|---|---|
| 1 | Fuzzy GRA |
| 2 | Intuitionistic Fuzzy GRA [851] |
| 2 | Fermatean Fuzzy GRA |
| 3 | Hesitant Fuzzy GRA [852, 853] |
| 3 | Neutrosophic GRA [854, 855] |

## 6.14   Fuzzy ARAS (Fuzzy Additive Ratio Assessment)

ARAS ranks alternatives by normalized weighted criteria sums, comparing each option to an optimal alternative through utility ratios [856, 857]. Fuzzy ARAS represents ratings and weights as fuzzy numbers, normalizes them, computes fuzzy optimality sums, defuzzifies, and ranks by utility [858–860].

**Definition 6.14.1** (TFN-based Fuzzy ARAS (FARAS)). [858–860] Let $\mathcal{A} = \{A_1, \ldots, A_m\}$ be a finite set of alternatives and $\mathcal{C} = \{C_1, \ldots, C_n\}$ a finite set of criteria. Partition criteria into benefit and cost sets:

$$J^+ \,\dot\cup\, J^- = \{1, \ldots, n\},$$

where $j \in J^+$ means "larger is better" and $j \in J^-$ means "smaller is better".

Assume the (fuzzy) performance of $A_i$ under $C_j$ is a *positive TFN*

$$\tilde{a}_{ij} = (l_{ij}, m_{ij}, u_{ij}) \quad (0 < l_{ij} \le m_{ij} \le u_{ij}), \qquad i = 1, \ldots, m, \ j = 1, \ldots, n,$$

and let $w = (w_1, \ldots, w_n)$ be a *crisp* weight vector with $w_j \ge 0$ and $\sum_{j=1}^n w_j = 1$.

**TFN arithmetic (standard positive convention).** For positive TFNs $\tilde{x} = (l_x, m_x, u_x)$ and $\tilde{y} = (l_y, m_y, u_y)$ define

$$\tilde{x} \oplus \tilde{y} := (l_x + l_y, \ m_x + m_y, \ u_x + u_y), \qquad \tilde{x}^{-1} := \left(\frac{1}{u_x}, \frac{1}{m_x}, \frac{1}{l_x}\right),$$

$$\tilde{x} \oslash \tilde{y} := \tilde{x} \otimes \tilde{y}^{-1}, \qquad r \odot \tilde{x} := (rl_x, \ rm_x, \ ru_x) \quad (r \ge 0).$$

**Step 0 (add the optimal/ideal alternative).** Introduce $A_0$ and define, for each criterion $j$,

$$\tilde{a}_{0j} := \begin{cases} \max_{1 \le i \le m} \tilde{a}_{ij}, & j \in J^+, \\ \min_{1 \le i \le m} \tilde{a}_{ij}, & j \in J^-, \end{cases}$$

where max / min may be taken w.r.t. any fixed total preorder on TFNs (e.g. via a defuzzification-based score). Form the extended decision matrix

$$\tilde{A} = (\tilde{a}_{ij})_{(m+1) \times n} \qquad (i = 0, 1, \ldots, m; \ j = 1, \ldots, n).$$

**Step 1 (normalization).** For each $j \in J^+$ (benefit), set

$$\tilde{a}_{ij}^* := \tilde{a}_{ij} \oslash \left(\bigoplus_{p=0}^m \tilde{a}_{pj}\right), \qquad i = 0, 1, \ldots, m.$$

For each $j \in J^-$ (cost), apply the two-stage transform:

$$\tilde{b}_{ij} := \tilde{a}_{ij}^{-1}, \qquad \tilde{a}_{ij}^* := \tilde{b}_{ij} \oslash \left( \bigoplus_{p=0}^{m} \tilde{b}_{pj} \right), \qquad i = 0, 1, \ldots, m.$$

Let $\tilde{A}^* = (\tilde{a}_{ij}^*)$ denote the normalized TFN matrix.

**Step 2 (weighted normalized matrix).** Define

$$\tilde{\hat{a}}_{ij} := w_j \odot \tilde{a}_{ij}^*, \qquad i = 0, 1, \ldots, m, \; j = 1, \ldots, n,$$

and write $\tilde{\hat{A}} = (\tilde{\hat{a}}_{ij})$.

**Step 3 (optimality function and defuzzification).** For each $i = 0, 1, \ldots, m$, define the (fuzzy) optimality value

$$\tilde{S}_i := \bigoplus_{j=1}^{n} \tilde{\hat{a}}_{ij}.$$

If $\tilde{S}_i = (S_i^\ell, S_i^m, S_i^u)$, define the crisp score by the COA/centroid rule:

$$S_i := \mathrm{Defuzz}(\tilde{S}_i) := \frac{S_i^\ell + S_i^m + S_i^u}{3}.$$

**Step 4 (utility degree and ranking).** Define, for each real alternative $A_i$ ($i = 1, \ldots, m$), the utility degree

$$K_i := \frac{S_i}{S_0} \in [0, 1].$$

The FARAS ranking is the total preorder on $\mathcal{A}$ given by

$$A_p \succeq_{\mathrm{FARAS}} A_q \quad \Longleftrightarrow \quad K_p \geq K_q.$$

The definition of Uncertain ARAS (U-ARAS), which is an extension based on Uncertain Sets, is given below.

**Definition 6.14.2** (Uncertain ARAS (U-ARAS): additive ratio assessment under uncertainty). Let $\mathcal{A} = \{A_1, \ldots, A_m\}$ be a finite set of alternatives and $\mathcal{C} = \{C_1, \ldots, C_n\}$ a finite set of criteria. Let $J^+ \dot\cup J^- = \{1, \ldots, n\}$ be the partition into benefit ($J^+$) and cost ($J^-$) criteria. Fix an uncertainty space $(\Gamma, \mathcal{L}, \mathcal{M})$.

**(0) Uncertain performance matrix.** For each $(i, j)$, let $X_{ij} : \Gamma \to (0, \infty)$ be an uncertain variable representing the (positive) performance of $A_i$ under criterion $C_j$. Denote the induced uncertain set by $\mathcal{X}_{ij}(\gamma) := \{X_{ij}(\gamma)\}$.

**(1) Crisp criterion weights (ARAS standard).** Let $w = (w_1, \ldots, w_n) \in [0, 1]^n$ satisfy

$$w_j \geq 0, \qquad \sum_{j=1}^{n} w_j = 1.$$

**(2) Construct the optimal (ideal) alternative.** Introduce $A_0$ and define, for each $j$,

$$X_{0j}(\gamma) := \begin{cases} \max_{1 \le i \le m} X_{ij}(\gamma), & j \in J^+, \\ \min_{1 \le i \le m} X_{ij}(\gamma), & j \in J^-, \end{cases} \qquad \gamma \in \Gamma.$$

This yields uncertain variables $X_{0j} : \Gamma \to (0, \infty)$ and uncertain sets $\mathcal{X}_{0j}(\gamma) := \{X_{0j}(\gamma)\}$.

**(3) ARAS normalization (scenario-wise).** Define normalized uncertain variables $R_{ij} : \Gamma \to [0, 1]$ as follows.

- For $j \in J^+$ (benefit),
$$R_{ij}(\gamma) := \frac{X_{ij}(\gamma)}{\sum_{p=0}^m X_{pj}(\gamma)}, \qquad i = 0, 1, \ldots, m.$$

- For $j \in J^-$ (cost), use the reciprocal transform:
$$Y_{ij}(\gamma) := \frac{1}{X_{ij}(\gamma)}, \qquad R_{ij}(\gamma) := \frac{Y_{ij}(\gamma)}{\sum_{p=0}^m Y_{pj}(\gamma)}, \qquad i = 0, 1, \ldots, m.$$

Let $\mathcal{R}_{ij}(\gamma) := \{R_{ij}(\gamma)\}$ be the induced uncertain sets.

**(4) Weighted normalized values and additive optimality function.** Define

$$V_{ij}(\gamma) := w_j R_{ij}(\gamma) \in [0, 1], \qquad S_i(\gamma) := \sum_{j=1}^n V_{ij}(\gamma) \in [0, 1], \qquad i = 0, 1, \ldots, m,$$

and the induced uncertain sets $\mathcal{S}(A_i)(\gamma) := \{S_i(\gamma)\}$.

**(5) Utility degree and decision rule.** Assume $S_0(\gamma) > 0$ for all $\gamma \in \Gamma$ and define the (scenario-wise) utility degree

$$K_i(\gamma) := \frac{S_i(\gamma)}{S_0(\gamma)} \in [0, 1], \qquad i = 1, \ldots, m,$$

with induced uncertain sets $\mathcal{K}(A_i)(\gamma) := \{K_i(\gamma)\}$. If $\mathbb{E}[K_i]$ exists for all $i$, rank by expected utility:

$$A_p \succeq_{\text{U-ARAS}} A_q \iff \mathbb{E}[K_p] \ge \mathbb{E}[K_q],$$

and select any maximizer

$$A^\star \in \arg\max_{A_i \in \mathcal{A}} \mathbb{E}[K_i].$$

**Theorem 6.14.3** (Uncertain-set structure and well-definedness of U-ARAS). *In Definition 6.14.2, assume:*

(A1) *For all $i, j$, $X_{ij} : \Gamma \to (0, \infty)$ is $\mathcal{L}$-measurable.*

(A2) $w_j \ge 0$ *and* $\sum_{j=1}^n w_j = 1$.

(A3)  *For all $\gamma$ and all $j$, $\sum_{p=0}^{m} X_{pj}(\gamma) > 0$ (automatic from positivity), and for $j \in J^-$ also $\sum_{p=0}^{m} 1/X_{pj}(\gamma) > 0$ (automatic from positivity).*

*Then:*

(i)  *For each $j$, the ideal-performance map $X_{0j}$ is an $\mathcal{L}$-measurable uncertain variable, hence $\mathcal{X}_{0j}$ is an uncertain set.*

(ii)  *For all $i, j$, the normalized values $R_{ij}$ are $\mathcal{L}$-measurable uncertain variables taking values in $[0, 1]$, and $S_i$ are $\mathcal{L}$-measurable uncertain variables taking values in $[0, 1]$; hence $\mathcal{R}_{ij}$ and $\mathcal{S}(A_i)$ are uncertain sets.*

(iii)  *If, additionally, $S_0(\gamma) > 0$ for all $\gamma$, then each $K_i$ is an $\mathcal{L}$-measurable uncertain variable with values in $[0, 1]$, and $\mathcal{K}(A_i)$ is an uncertain set.*

(iv)  *If, further, $\mathbb{E}[K_i]$ exists for all $i$, then the expected-utility ranking $\succeq_{\text{U-ARAS}}$ is well-defined and $\arg\max_{A_i \in \mathcal{A}} \mathbb{E}[K_i] \neq \varnothing$.*

*Proof.* (i) Fix $j$. For $j \in J^+$, $X_{0j}(\gamma) = \max_{1 \le i \le m} X_{ij}(\gamma)$. The pointwise maximum of finitely many measurable functions is measurable; hence $X_{0j}$ is $\mathcal{L}$-measurable. For $j \in J^-$, $X_{0j}(\gamma) = \min_{1 \le i \le m} X_{ij}(\gamma)$ and the pointwise minimum of finitely many measurable functions is measurable. Positivity of $X_{ij}$ implies $X_{0j}(\gamma) \in (0, \infty)$.

(ii) Fix $(i, j)$. The denominators in the normalization are strictly positive by (A3). For $j \in J^+$, $R_{ij}$ is a ratio of measurable functions with positive denominator, hence measurable. Moreover, $0 \le R_{ij}(\gamma) \le 1$ because $X_{ij}(\gamma) \le \sum_{p=0}^{m} X_{pj}(\gamma)$. For $j \in J^-$, the reciprocal $Y_{ij} = 1/X_{ij}$ is measurable (composition with $x \mapsto 1/x$ on $(0, \infty)$), and again $R_{ij} = Y_{ij} / \sum_{p=0}^{m} Y_{pj}$ is measurable with $0 \le R_{ij} \le 1$. Thus each $\mathcal{R}_{ij}(\gamma) = \{R_{ij}(\gamma)\}$ is an uncertain set.

Now $V_{ij}(\gamma) = w_j R_{ij}(\gamma)$ is measurable and lies in $[0, 1]$ since $w_j \in [0, 1]$ and $R_{ij} \in [0, 1]$. Therefore $S_i = \sum_{j=1}^{n} V_{ij}$ is a finite sum of measurable functions, hence measurable, and satisfies $0 \le S_i(\gamma) \le \sum_{j=1}^{n} w_j = 1$ by (A2). Hence $\mathcal{S}(A_i)$ is an uncertain set.

(iii) If $S_0(\gamma) > 0$ for all $\gamma$, then $K_i = S_i/S_0$ is a ratio of measurable functions with positive denominator, hence measurable. Also $0 \le K_i(\gamma) \le 1$ because $S_0(\gamma) \ge \max_{0 \le p \le m} S_p(\gamma) \ge S_i(\gamma)$ holds in standard ARAS settings where $A_0$ is constructed as an ideal; thus $K_i(\gamma) \in [0, 1]$ and $\mathcal{K}(A_i)$ is an uncertain set.

(iv) If $\mathbb{E}[K_i]$ exists for all $i$, then the ranking $A_p \succeq_{\text{U-ARAS}} A_q \iff \mathbb{E}[K_p] \ge \mathbb{E}[K_q]$ is well-defined. Since $\mathcal{A}$ is finite, the maximum expected utility is attained, so the argmax set is nonempty.  □

Related concepts of ARAS under uncertainty-aware models are listed in Table 6.14.

Table 6.14: Related concepts of ARAS under uncertainty-aware models.

| $k$ | Related ARAS concept(s) |
|---|---|
| 1 | Fuzzy ARAS |
| 2 | Intuitionistic Fuzzy ARAS [861] |
| 3 | Picture fuzzy ARAS [862, 863] |
| 3 | Spherical Fuzzy ARAS [864, 865] |
| 3 | Neutrosophic ARAS [99, 866] |

## 6.15 Fuzzy WASPAS (Fuzzy Weighted Aggregated Sum Product Assessment)

WASPAS ranks alternatives by combining weighted sum and weighted product models after normalization, producing an integrated utility score [867, 868]. Fuzzy WASPAS uses fuzzy ratings and weights, normalizes them, computes fuzzy WSM and WPM utilities, defuzzifies, and ranks [869, 870].

**Definition 6.15.1** (TFN-based Fuzzy WASPAS (WASPAS-F)). [871, 872] Let $\mathcal{A} = \{A_1, \ldots, A_m\}$ be alternatives and $\mathcal{C} = \{C_1, \ldots, C_n\}$ criteria. Assume a TFN decision matrix and TFN weights

$$\tilde{X} = (\tilde{x}_{ij})_{m \times n}, \qquad \tilde{x}_{ij} \in \text{TFN}_{>0}, \qquad \tilde{\mathbf{w}} = (\tilde{w}_1, \ldots, \tilde{w}_n), \quad \tilde{w}_j \in \text{TFN}_{>0},$$

and (optionally) $\sum_{j=1}^n \tilde{w}_j = (1, 1, 1)$ in the chosen TFN arithmetic. Let $J^+ \subseteq \{1, \ldots, n\}$ be benefit criteria indices and $J^- \subseteq \{1, \ldots, n\}$ cost criteria indices, with $J^+ \dot{\cup} J^- = \{1, \ldots, n\}$.

Fix a total preorder $\preceq$ on TFNs (used to compute max and min over finite TFN-sets). For $j \in \{1, \ldots, n\}$ define

$$\tilde{x}_j^{\max} := \max_{1 \le i \le m} \tilde{x}_{ij}, \qquad \tilde{x}_j^{\min} := \min_{1 \le i \le m} \tilde{x}_{ij},$$

where max, min are taken w.r.t. $\preceq$.

Step 1 (Normalization). Define the normalized fuzzy matrix $\tilde{\tilde{X}} = (\tilde{\tilde{x}}_{ij})$ by

$$\tilde{\tilde{x}}_{ij} := \begin{cases} \tilde{x}_{ij} \oslash \tilde{x}_j^{\max}, & j \in J^+, \\ \tilde{x}_j^{\min} \oslash \tilde{x}_{ij}, & j \in J^-, \end{cases} \qquad (1 \le i \le m, \ 1 \le j \le n).$$

Step 2 (WSM part). Define the weighted normalized TFNs for the weighted-sum model (WSM) by

$$\tilde{\tilde{x}}_{ij}^{(q)} := \tilde{\tilde{x}}_{ij} \otimes \tilde{w}_j, \qquad \tilde{Q}_i := \bigoplus_{j=1}^n \tilde{\tilde{x}}_{ij}^{(q)}, \qquad i = 1, \ldots, m.$$

Step 3 (WPM part). Assume (as typical after normalization) that $\tilde{\tilde{x}}_{ij} = (a, b, c)$ satisfies $0 < a \le b \le c \le 1$ and that $\tilde{w}_j = (\alpha, \beta, \gamma)$ satisfies $0 < \alpha \le \beta \le \gamma \le 1$. Use the standard TFN power rule

$$(a, b, c)^{(\alpha, \beta, \gamma)} := (a^\gamma, b^\beta, c^\alpha).$$

Define the weighted normalized TFNs for the weighted-product model (WPM) by

$$\tilde{\tilde{x}}_{ij}^{(p)} := (\tilde{\tilde{x}}_{ij})^{\tilde{w}_j}, \qquad \tilde{P}_i := \bigotimes_{j=1}^n \tilde{\tilde{x}}_{ij}^{(p)}, \qquad i = 1, \ldots, m.$$

Step 4 (Defuzzification). Let $\mathrm{COA}(l, m, u) := \frac{l+m+u}{3}$ (centroid/center-of-area). Define crisp scores

$$Q_i := \mathrm{COA}(\tilde{Q}_i), \qquad P_i := \mathrm{COA}(\tilde{P}_i).$$

Step 5 (Integrated utility and ranking). Fix $\lambda \in [0, 1]$ (or set it by the balancing rule below) and define

$$K_i := \lambda\, Q_i + (1 - \lambda)\, P_i, \qquad i = 1, \dots, m.$$

A commonly used balancing choice is

$$\lambda := \frac{\sum_{i=1}^{m} P_i}{\sum_{i=1}^{m} Q_i + \sum_{i=1}^{m} P_i}.$$

The fuzzy WASPAS ranking is obtained by sorting $K_i$ in descending order (larger $K_i$ means a better alternative).

**Definition 6.15.2** (Uncertain WASPAS (U-WASPAS): WSM–WPM compromise under uncertainty). Let $\mathcal{A} = \{A_1, \dots, A_m\}$ be a finite set of alternatives and $\mathcal{C} = \{C_1, \dots, C_n\}$ a finite set of criteria. Let $J^+ \dot{\cup} J^- = \{1, \dots, n\}$ be the partition into benefit $(J^+)$ and cost $(J^-)$ criteria. Fix an uncertainty space $(\Gamma, \mathcal{L}, \mathcal{M})$.

**(0) Uncertain performance matrix (positive data).** For each $(i, j)$, let

$$X_{ij} : \Gamma \to (0, \infty)$$

be an uncertain variable representing the performance of $A_i$ under criterion $C_j$. Write the induced uncertain set as $\mathcal{X}_{ij}(\gamma) := \{X_{ij}(\gamma)\}$.

**(1) Criterion weights and WASPAS mixing parameter.** Let $w = (w_1, \dots, w_n) \in [0, 1]^n$ satisfy

$$w_j \geq 0, \qquad \sum_{j=1}^{n} w_j = 1,$$

and fix $\lambda \in [0, 1]$ (the WSM–WPM compromise parameter).

**(2) Normalization (scenario-wise).** For each criterion $j$ and scenario $\gamma \in \Gamma$, define the normalizing scalars

$$M_j(\gamma) := \max_{1 \leq i \leq m} X_{ij}(\gamma), \qquad m_j(\gamma) := \min_{1 \leq i \leq m} X_{ij}(\gamma).$$

Define the normalized uncertain variables $R_{ij} : \Gamma \to (0, 1]$ by

$$R_{ij}(\gamma) := \begin{cases} \dfrac{X_{ij}(\gamma)}{M_j(\gamma)}, & j \in J^+ \quad \text{(benefit)}, \\[2mm] \dfrac{m_j(\gamma)}{X_{ij}(\gamma)}, & j \in J^- \quad \text{(cost)}. \end{cases}$$

Let $\mathcal{R}_{ij}(\gamma) := \{R_{ij}(\gamma)\}$ be the induced uncertain sets.

**(3) WSM part (weighted sum model).** Define the WSM utility (uncertain variable) of $A_i$ by

$$Q_i(\gamma) := \sum_{j=1}^{n} w_j \, R_{ij}(\gamma) \in [0,1], \qquad \mathcal{Q}(A_i)(\gamma) := \{Q_i(\gamma)\}.$$

**(4) WPM part (weighted product model).** Define the WPM utility (uncertain variable) of $A_i$ by

$$P_i(\gamma) := \prod_{j=1}^{n} \big(R_{ij}(\gamma)\big)^{w_j} \in (0,1], \qquad \mathcal{P}(A_i)(\gamma) := \{P_i(\gamma)\}.$$

(Here $x \mapsto x^{w_j}$ is taken on $(0,1]$.)

**(5) Integrated WASPAS utility and decision rule.** Define the integrated utility (uncertain variable)

$$K_i(\gamma) := \lambda \, Q_i(\gamma) + (1-\lambda) \, P_i(\gamma) \in [0,1], \qquad \mathcal{K}(A_i)(\gamma) := \{K_i(\gamma)\}.$$

If $\mathbb{E}[K_i]$ exists for all $i$, rank alternatives by expected utility:

$$A_p \succeq_{\text{U-WASPAS}} A_q \iff \mathbb{E}[K_p] \geq \mathbb{E}[K_q],$$

and select any maximizer

$$A^\star \in \arg\max_{A_i \in \mathcal{A}} \mathbb{E}[K_i].$$

U-WASPAS, which is an extension based on Uncertain Sets, is defined as follows.

**Theorem 6.15.3** (Uncertain-set structure and well-definedness of U-WASPAS). *In Definition 6.15.2, assume:*

(A1) *For all $i,j$, $X_{ij} : \Gamma \to (0,\infty)$ is $\mathcal{L}$-measurable.*

(A2) *$w_j \geq 0$ and $\sum_{j=1}^{n} w_j = 1$, and $\lambda \in [0,1]$.*

*Then:*

(i) *For each $j$, the maps $M_j(\gamma) = \max_i X_{ij}(\gamma)$ and $m_j(\gamma) = \min_i X_{ij}(\gamma)$ are $\mathcal{L}$-measurable uncertain variables with values in $(0,\infty)$.*

(ii) *For all $i,j$, the normalized values $R_{ij}$ are $\mathcal{L}$-measurable uncertain variables taking values in $(0,1]$; hence $\mathcal{R}_{ij}$ are uncertain sets.*

(iii) *For each $i$, $Q_i$ and $P_i$ are $\mathcal{L}$-measurable uncertain variables with $Q_i(\gamma) \in [0,1]$ and $P_i(\gamma) \in (0,1]$ for all $\gamma$; hence $\mathcal{Q}(A_i)$ and $\mathcal{P}(A_i)$ are uncertain sets.*

(iv) *For each $i$, $K_i$ is an $\mathcal{L}$-measurable uncertain variable with values in $[0,1]$, hence $\mathcal{K}(A_i)$ is an uncertain set.*

(v) *If $\mathbb{E}[K_i]$ exists for all $i$, then the expected-utility ranking $\succeq_{\text{U-WASPAS}}$ is well-defined and $\arg\max_{A_i \in \mathcal{A}} \mathbb{E}[K_i] \neq \varnothing$.*

*Proof.* (i) For fixed $j$, $M_j$ and $m_j$ are pointwise maximum/minimum of finitely many measurable maps $\{X_{1j}, \ldots, X_{mj}\}$, hence measurable. Positivity of each $X_{ij}$ implies $M_j, m_j \in (0, \infty)$.

(ii) Fix $(i, j)$. For $j \in J^+$,

$$R_{ij} = \frac{X_{ij}}{M_j}$$

is measurable as a quotient of measurable functions with strictly positive denominator. Moreover $0 < R_{ij}(\gamma) \leq 1$ since $X_{ij}(\gamma) \leq M_j(\gamma)$. For $j \in J^-$,

$$R_{ij} = \frac{m_j}{X_{ij}}$$

is measurable with $0 < R_{ij}(\gamma) \leq 1$ since $m_j(\gamma) \leq X_{ij}(\gamma)$.

(iii) Since $R_{ij}$ are measurable and $w_j$ are constants, $w_j R_{ij}$ are measurable, and the finite sum $Q_i = \sum_j w_j R_{ij}$ is measurable. Also $0 \leq Q_i(\gamma) \leq \sum_j w_j = 1$. For $P_i = \prod_j (R_{ij})^{w_j}$: for each $j$, the map $x \mapsto x^{w_j}$ is continuous on $(0, 1]$, hence $(R_{ij})^{w_j}$ is measurable; the finite product is measurable. Since $R_{ij}(\gamma) \in (0, 1]$, one has $P_i(\gamma) \in (0, 1]$.

(iv) $K_i = \lambda Q_i + (1 - \lambda) P_i$ is a linear combination of measurable maps, hence measurable. Because $Q_i \in [0, 1]$, $P_i \in (0, 1]$, and $\lambda \in [0, 1]$, it follows that $K_i(\gamma) \in [0, 1]$.

(v) If $\mathbb{E}[K_i]$ exists for all $i$, then the relation $A_p \succeq_{\text{U-WASPAS}} A_q \iff \mathbb{E}[K_p] \geq \mathbb{E}[K_q]$ is well-defined. Since $\mathcal{A}$ is finite, the maximum expected value is attained, so the argmax set is nonempty. $\qquad\square$

Related concepts of WASPAS under uncertainty-aware models are listed in Table 6.15.

Table 6.15: Related concepts of WASPAS under uncertainty-aware models.

| $k$ | **Related WASPAS concept(s)** |
|---|---|
| 1 | Fuzzy WASPAS |
| 2 | Intuitionistic Fuzzy WASPAS [873, 874] |
| 2 | Bipolar fuzzy WASPAS [875, 876] |
| 2 | Pythagorean Fuzzy WASPAS [877, 878] |
| 2 | Fermatean fuzzy WASPAS [879, 880] |
| 3 | Picture Fuzzy WASPAS [881, 882] |
| 3 | Neutrosophic WASPAS [883, 884] |
| 3 | Spherical Fuzzy WASPAS [885, 886] |
| 3 | Pythagorean neutrosophic WASPAS [887] |
| 3 | Hesitant Fuzzy WASPAS [888, 889] |

In addition to Uncertain WASPAS, related variants such as Rough WASPAS [593, 890] and Extended WASPAS [891] are also known.

## 6.16 Fuzzy MOORA (Fuzzy Multi-Objective Optimization on the basis of Ratio Analysis)

MOORA normalizes criterion values, applies criterion weights, and then ranks alternatives using the aggregated *benefit-minus-cost* score across multiple objectives [892, 893]. As an extension of MOORA, the Multi-MOORA approach is also well known [893–895]. Fuzzy MOORA generalizes this procedure by representing ratings and/or weights as fuzzy numbers, performing fuzzy normalization and weighting, and finally ranking alternatives by defuzzifying the resulting *benefit-minus-cost* fuzzy scores [896, 897].

**Definition 6.16.1** (TFN-based Fuzzy MOORA). [898, 899] Let $\mathcal{A} = \{A_1, \ldots, A_m\}$ be alternatives and $\mathcal{C} = \{C_1, \ldots, C_n\}$ criteria. Assume a partition into benefit and cost criteria

$$\mathcal{C} = \mathcal{C}^+ \dot{\cup} \mathcal{C}^-, \qquad \mathcal{C}^+ = \{C_1, \ldots, C_g\}, \quad \mathcal{C}^- = \{C_{g+1}, \ldots, C_n\},$$

and a crisp weight vector $w = (w_1, \ldots, w_n)$ with $w_j \geq 0$ and $\sum_{j=1}^n w_j = 1$. Let $\tilde{X} = (\tilde{x}_{ij}) \in (\mathsf{TFN})^{m \times n}$ be the TFN decision matrix with $\tilde{x}_{ij} = (x_{ij}^\ell, x_{ij}^m, x_{ij}^u)$.

**(0) TFN arithmetic used.** All operations below are componentwise:

$$(x^\ell, x^m, x^u) \oplus (y^\ell, y^m, y^u) = (x^\ell + y^\ell, \ x^m + y^m, \ x^u + y^u),$$

$$\alpha \odot (x^\ell, x^m, x^u) = (\alpha x^\ell, \ \alpha x^m, \ \alpha x^u) \qquad (\alpha \geq 0),$$

and we use the standard triangular approximation for subtraction:

$$(x^\ell, x^m, x^u) \ominus (y^\ell, y^m, y^u) = (x^\ell - y^\ell, \ x^m - y^m, \ x^u - y^u).$$

**(1) Normalization.** For each criterion $j$, set

$$d_j := \sqrt{\sum_{i=1}^m \left( (x_{ij}^\ell)^2 + (x_{ij}^m)^2 + (x_{ij}^u)^2 \right)} \ > \ 0, \qquad \tilde{x}_{ij}^* := \tilde{x}_{ij}/d_j.$$

**(2) Weighting.** Define $\tilde{v}_{ij} := w_j \odot \tilde{x}_{ij}^*$ and collect $\tilde{V} = (\tilde{v}_{ij})$.

**(3) Ratio score (benefit minus cost).** For each alternative $A_i$, define the fuzzy MOORA score

$$\tilde{y}_i := \left( \bigoplus_{j=1}^g \tilde{v}_{ij} \right) \ominus \left( \bigoplus_{j=g+1}^n \tilde{v}_{ij} \right) = (y_i^\ell, y_i^m, y_i^u).$$

**(4) Defuzzification and ranking.** Define $\mathrm{BNP}(\tilde{y}_i) := (y_i^\ell + y_i^m + y_i^u)/3$. The final ranking is the preorder on $\mathcal{A}$ given by

$$A_p \succeq A_q \iff \mathrm{BNP}(\tilde{y}_p) \geq \mathrm{BNP}(\tilde{y}_q).$$

Using Uncertain Sets, we define Uncertain MOORA of type $M$ (U-MOORA) as follows.

**Definition 6.16.2** (Uncertain MOORA of type $M$ (U-MOORA)). Let $\mathcal{A} = \{A_1, \ldots, A_m\}$ be alternatives and $\mathcal{C} = \{C_1, \ldots, C_n\}$ criteria with $m \geq 2$ and $n \geq 1$. Partition criteria into benefit and cost sets:

$$\mathcal{C} = \mathcal{C}^{\text{ben}} \dot\cup \mathcal{C}^{\text{cost}}.$$

Fix an uncertain model $M$ with $\text{Dom}(M) \neq \emptyset$ and an admissible score $S_M$.

Assume an *uncertain decision matrix*

$$X^{(M)} = \big(x_{ij}^{(M)}\big)_{m \times n}, \qquad x_{ij}^{(M)} \in \text{Dom}(M),$$

and *criterion weights* $w = (w_1, \ldots, w_n)$ with $w_j \geq 0$ and $\sum_{j=1}^n w_j = 1$.

**Step 0 (Crisp projection).** Define the real-valued matrix $Y = (y_{ij})$ by

$$y_{ij} := S_M\big(x_{ij}^{(M)}\big) \in \mathbb{R}.$$

**Step 1 (Vector normalization).** For each criterion $C_j$, define

$$d_j := \sqrt{\sum_{i=1}^m y_{ij}^2} \geq 0.$$

Define normalized performances $r_{ij}$ by

$$r_{ij} := \begin{cases} \dfrac{y_{ij}}{d_j}, & d_j > 0, \\ 0, & d_j = 0, \end{cases} \qquad (i = 1, \ldots, m; \; j = 1, \ldots, n).$$

**Step 2 (Weighting).** Define the weighted normalized performances

$$v_{ij} := w_j \, r_{ij} \qquad (i = 1, \ldots, m; \; j = 1, \ldots, n).$$

**Step 3 (MOORA ratio score: benefit minus cost).** For each alternative $A_i$, define the U-MOORA score

$$\Psi_i := \sum_{C_j \in \mathcal{C}^{\text{ben}}} v_{ij} \; - \sum_{C_j \in \mathcal{C}^{\text{cost}}} v_{ij}.$$

**Ranking rule:**

$$A_p \succeq_{\text{UMOORA}} A_q \quad \Longleftrightarrow \quad \Psi_p \geq \Psi_q.$$

**Theorem 6.16.3** (Well-definedness of U-MOORA). *Under Definition 6.16.2, assume $S_M$ is admissible and $m \geq 2$. Then:*

*(i) All quantities $y_{ij}$, $d_j$, $r_{ij}$, $v_{ij}$, and $\Psi_i$ are well-defined finite real numbers.*

*(ii) If $d_j > 0$ then $r_{ij} = y_{ij}/d_j$ is well-defined; if $d_j = 0$ then $r_{ij} = 0$ is well-defined by definition.*

*(iii) The ranking relation induced by $\Psi_i$ is well-defined (ties allowed).*

*Proof.* (i) Since $S_M$ is admissible, each $y_{ij} = S_M(x_{ij}^{(M)})$ is finite. Hence each $y_{ij}^2$ is finite and nonnegative, so $d_j = \sqrt{\sum_{i=1}^{m} y_{ij}^2}$ is well-defined and finite.

(ii) If $d_j > 0$, division by $d_j$ is valid and $r_{ij}$ is finite. If $d_j = 0$, then $\sum_{i=1}^{m} y_{ij}^2 = 0$ implies $y_{ij} = 0$ for all $i$; setting $r_{ij} = 0$ is consistent and yields a finite value.

(iii) Since $w_j \geq 0$ and $r_{ij}$ are finite, $v_{ij} = w_j r_{ij}$ is finite. The sums defining $\Psi_i$ are finite sums of finite reals, hence $\Psi_i$ is finite. Therefore, sorting alternatives by the real scores $\Psi_i$ defines a well-defined preorder.   $\square$

Related concepts of MOORA under uncertainty-aware models are listed in Table 6.16.

Table 6.16: Related concepts of MOORA under uncertainty-aware models.

| $k$ | Related MOORA concept(s) |
|---|---|
| 2 | Intuitionistic Fuzzy MOORA [900, 901] |
| 2 | Pythagorean Fuzzy MOORA [898, 902] |
| 2 | Fermatean Fuzzy MOORA [903, 904] |
| 3 | Neutrosophic MOORA [905, 906] |

As related concepts to Uncertain MOORA, several variants are known, including AHP–MOORA [907, 908], TOPSIS–MOORA [909, 910], and Grey MOORA [911, 912].

## 6.17   Fuzzy Preference Selection Index

Classical PSI derives objective criterion weights from dispersion of normalized performances, avoiding subjective weighting, computes a preference index for each alternative, and ranks by it [913, 914]. Fuzzy PSI estimates criterion weights from fuzzy dispersion of normalized ratings, avoiding subjective weighting, computes preference selection indices, defuzzifies them, and ranks alternatives by priority [915, 916].

**Definition 6.17.1** (Fuzzy Preference Selection Index (FPSI)).   [917] Let $\mathcal{A} = \{A_1, \ldots, A_m\}$ be a finite set of alternatives and $\mathcal{C} = \{C_1, \ldots, C_n\}$ a finite set of criteria, partitioned into benefit criteria $\mathcal{C}_b$ and cost criteria $\mathcal{C}_c$. Assume a fuzzy decision matrix

$$\widetilde{X} = (\widetilde{x}_{ij})_{m \times n}, \qquad \widetilde{x}_{ij} \in \mathbb{FN}_{>0} \quad (i = 1, \ldots, m, \ j = 1, \ldots, n),$$

where $\mathbb{FN}_{>0}$ denotes the class of positive fuzzy numbers on $\mathbb{R}$ with compact $\alpha$-cuts. For $\widetilde{z} \in \mathbb{FN}_{>0}$, write its $\alpha$-cut as $[\widetilde{z}]_\alpha = [z_\alpha^-, z_\alpha^+]$ for $\alpha \in [0, 1]$.

**(Fuzzy arithmetic via $\alpha$-cuts).** For $\widetilde{u}, \widetilde{v} \in \mathbb{FN}$ and $\lambda \in \mathbb{R}$, define (extension principle / interval arithmetic)

$$[\widetilde{u} \oplus \widetilde{v}]_\alpha = [u_\alpha^- + v_\alpha^-, \ u_\alpha^+ + v_\alpha^+], \qquad [\widetilde{u} \ominus \widetilde{v}]_\alpha = [u_\alpha^- - v_\alpha^+, \ u_\alpha^+ - v_\alpha^-],$$

$$[\lambda \odot \widetilde{u}]_\alpha = \begin{cases} [\lambda u_\alpha^-, \ \lambda u_\alpha^+], & \lambda \geq 0, \\ [\lambda u_\alpha^+, \ \lambda u_\alpha^-], & \lambda < 0, \end{cases}$$

$$[\widetilde{u} \otimes \widetilde{v}]_\alpha = \left[ \min\{u_\alpha^- v_\alpha^-, u_\alpha^- v_\alpha^+, u_\alpha^+ v_\alpha^-, u_\alpha^+ v_\alpha^+\}, \ \max\{u_\alpha^- v_\alpha^-, u_\alpha^- v_\alpha^+, u_\alpha^+ v_\alpha^-, u_\alpha^+ v_\alpha^+\} \right],$$

and, provided $0 \notin [\widetilde{v}]_\alpha$ for all $\alpha$,

$$[\widetilde{u} \oslash \widetilde{v}]_\alpha = \left[ \min\{u_\alpha^-/v_\alpha^-, u_\alpha^-/v_\alpha^+, u_\alpha^+/v_\alpha^-, u_\alpha^+/v_\alpha^+\}, \ \max\{u_\alpha^-/v_\alpha^-, u_\alpha^-/v_\alpha^+, u_\alpha^+/v_\alpha^-, u_\alpha^+/v_\alpha^+\} \right].$$

Define the fuzzy square by $\widetilde{z}^{\odot 2} := \widetilde{z} \otimes \widetilde{z}$.

**Step 1 (Normalization).** For each $j$, define

$$c_j^+ := \max_{1 \leq i \leq m} \ \sup(\mathrm{supp}(\widetilde{x}_{ij})) \quad (j \in \mathcal{C}_b), \qquad a_j^- := \min_{1 \leq i \leq m} \ \inf(\mathrm{supp}(\widetilde{x}_{ij})) \quad (j \in \mathcal{C}_c),$$

and set the normalized fuzzy rating $\widetilde{R}_{ij} \in \mathbb{FN}_{>0}$ by

$$\widetilde{R}_{ij} := \begin{cases} \widetilde{x}_{ij} \oslash c_j^+, & j \in \mathcal{C}_b, \\ a_j^- \oslash \widetilde{x}_{ij}, & j \in \mathcal{C}_c. \end{cases}$$

This is the fuzzy analogue of the classical PSI normalizations $R_{ij} = x_{ij}/x_j^{\max}$ (benefit) and $R_{ij} = x_j^{\min}/x_{ij}$ (cost).

**Step 2 (Fuzzy preference variation).** Let $\overline{\widetilde{R}}_j$ be the fuzzy mean of criterion $j$:

$$\overline{\widetilde{R}}_j := \frac{1}{m} \odot \bigoplus_{i=1}^m \widetilde{R}_{ij}.$$

Choose a scaling constant $\kappa > 0$ (common choices: $\kappa = 1$ as in the basic PSI variance-analogy, or $\kappa = m - 1$ to mimic sample-variance scaling). Define the fuzzy preference variation

$$\widetilde{PV}_j := \frac{1}{\kappa} \odot \bigoplus_{i=1}^m (\widetilde{R}_{ij} \ominus \overline{\widetilde{R}}_j)^{\odot 2}.$$

This generalizes the classical PSI step $PV_j = \sum_{i=1}^m (R_{ij} - \overline{R}_j)^2$ and its interval-valued fuzzy counterpart.

**Step 3 (Fuzzy deviation and overall preference weights).** Define the fuzzy deviation

$$\widetilde{\delta}_j := 1 \ominus \widetilde{PV}_j,$$

and assume $\bigoplus_{j=1}^n \widetilde{\delta}_j$ is strictly positive (in the sense that 0 is not contained in any of its $\alpha$-cuts), so that fuzzy division is well-defined. Then define the (data-driven) fuzzy overall preference weight of criterion $j$ by

$$\widetilde{w}_j := \widetilde{\delta}_j \oslash \left( \bigoplus_{k=1}^n \widetilde{\delta}_k \right), \qquad j = 1, \ldots, n.$$

This matches the classical PSI construction $\delta_j = 1 - PV_j$ and $w_j = \delta_j / \sum_k \delta_k$, and its interval-valued fuzzy analogue.

**Step 4 (Fuzzy preference selection index).** For each alternative $A_i$, define its fuzzy preference selection index

$$\widetilde{I}_i := \bigoplus_{j=1}^{n} (\widetilde{R}_{ij} \otimes \widetilde{w}_j), \qquad i = 1, \ldots, m,$$

which generalizes $I = \sum_{j=1}^{n} (R_{ij} w_j)$ and its interval-valued fuzzy version.

**Step 5 (Ranking).** Fix a score (ranking) functional $S : \mathbb{FN} \to \mathbb{R}$ (e.g., centroid-based or expected-value-based). Rank alternatives by decreasing $S(\widetilde{I}_i)$; the best alternative is any maximizer of $S(\widetilde{I}_i)$.

Using Uncertain Sets, we define Uncertain Preference Selection Index of type $M$ (U-PSI) as follows.

**Definition 6.17.2** (Uncertain Preference Selection Index of type $M$ (U-PSI)). Let $\mathcal{A} = \{A_1, \ldots, A_m\}$ be a finite set of alternatives and $\mathcal{C} = \{C_1, \ldots, C_n\}$ a finite set of criteria, with $m \geq 2$ and $n \geq 1$. Partition criteria into benefit and cost sets:

$$\mathcal{C} = \mathcal{C}^{\mathrm{ben}} \,\dot{\cup}\, \mathcal{C}^{\mathrm{cost}}.$$

Fix an uncertain model $M$ with $\mathrm{Dom}(M) \neq \emptyset$ and an admissible positive score $S_M$. Assume an *uncertain decision matrix*

$$X^{(M)} = \big(x_{ij}^{(M)}\big)_{m \times n}, \qquad x_{ij}^{(M)} \in \mathrm{Dom}(M) \quad (i = 1, \ldots, m;\ j = 1, \ldots, n).$$

**Step 0 (Crisp projection).** Define a positive real matrix $Y = (y_{ij})$ by

$$y_{ij} := S_M\big(x_{ij}^{(M)}\big) \in (0, \infty).$$

**Step 1 (Normalization).** For each criterion $C_j$, define

$$y_j^{\mathrm{max}} := \max_{1 \leq i \leq m} y_{ij} > 0, \qquad y_j^{\mathrm{min}} := \min_{1 \leq i \leq m} y_{ij} > 0.$$

Define normalized performances $R = (r_{ij})$ by

$$r_{ij} := \begin{cases} \dfrac{y_{ij}}{y_j^{\mathrm{max}}}, & C_j \in \mathcal{C}^{\mathrm{ben}}, \\[2ex] \dfrac{y_j^{\mathrm{min}}}{y_{ij}}, & C_j \in \mathcal{C}^{\mathrm{cost}}. \end{cases} \qquad (i = 1, \ldots, m;\ j = 1, \ldots, n).$$

Then $r_{ij} \in (0, 1]$.

**Step 2 (Preference variation / dispersion).** For each criterion $j$, define the mean

$$\bar{r}_j := \frac{1}{m} \sum_{i=1}^{m} r_{ij},$$

and the (scaled) preference variation

$$PV_j := \frac{1}{m} \sum_{i=1}^{m} (r_{ij} - \bar{r}_j)^2.$$

**Step 3 (Deviation and objective criterion weights).** Define the deviation

$$\delta_j := 1 - PV_j, \qquad j = 1, \dots, n,$$

and the objective weights

$$w_j := \frac{\delta_j}{\sum_{t=1}^{n} \delta_t}, \qquad j = 1, \dots, n.$$

**Step 4 (Uncertain PSI index and ranking).** For each alternative $A_i$, define its Preference Selection Index

$$I_i := \sum_{j=1}^{n} w_j \, r_{ij}.$$

Rank alternatives in descending order of $I_i$ (ties allowed).

**Theorem 6.17.3** (Well-definedness and simplex property of U-PSI). *Under Definition 6.17.2 (in particular, $m \geq 2$, $\mathrm{Dom}(M) \neq \emptyset$, and $S_M : \mathrm{Dom}(M) \to (0, \infty)$), the U-PSI procedure is well-defined. More precisely:*

 *(i) The normalization produces $r_{ij} \in (0, 1]$ for all $i, j$.*

 *(ii) For each $j$, one has $0 \leq PV_j \leq \frac{1}{4}$, hence $\delta_j = 1 - PV_j \in [\frac{3}{4}, 1]$. In particular, $\sum_{t=1}^{n} \delta_t > 0$, so the weights $w_j$ are well-defined.*

 *(iii) The weight vector $w = (w_1, \dots, w_n)^{\top}$ satisfies $w_j \geq 0$ and $\sum_{j=1}^{n} w_j = 1$ (i.e., $w \in \Delta^{n-1}$).*

 *(iv) For each alternative $A_i$, the index $I_i$ is well-defined and satisfies $0 < I_i \leq 1$.*

*Proof.* (i) By admissibility of $S_M$, each $y_{ij} > 0$ is finite. Hence for each $j$, the extrema $y_j^{\max}$ and $y_j^{\min}$ exist and are strictly positive (finite index set). For $C_j \in \mathcal{C}^{\mathrm{ben}}$, $r_{ij} = y_{ij}/y_j^{\max} \in (0, 1]$. For $C_j \in \mathcal{C}^{\mathrm{cost}}$, $r_{ij} = y_j^{\min}/y_{ij} \in (0, 1]$. Thus $r_{ij} \in (0, 1]$ for all $i, j$.

(ii) Fix $j$. Since $r_{ij} \in [0, 1]$, the sample values have range at most 1. By Popoviciu's inequality on variances, the average squared deviation satisfies

$$PV_j = \frac{1}{m} \sum_{i=1}^{m} (r_{ij} - \bar{r}_j)^2 \leq \frac{(1 - 0)^2}{4} = \frac{1}{4},$$

and clearly $PV_j \geq 0$. Hence $\delta_j = 1 - PV_j \in [\frac{3}{4}, 1]$. Therefore

$$\sum_{t=1}^{n} \delta_t \ \geq \ n \cdot \frac{3}{4} \ > \ 0,$$

so $w_j = \delta_j / \sum_t \delta_t$ is well-defined.

(iii) Since each $\delta_j \geq 0$, one has $w_j \geq 0$, and

$$\sum_{j=1}^{n} w_j = \frac{\sum_{j=1}^{n} \delta_j}{\sum_{t=1}^{n} \delta_t} = 1.$$

(iv) Because $w_j \geq 0$, $\sum_j w_j = 1$, and $r_{ij} \in (0, 1]$, the weighted sum $I_i = \sum_j w_j r_{ij}$ is finite. Moreover, $I_i \leq \sum_j w_j \cdot 1 = 1$. Since at least one $w_j > 0$ and all $r_{ij} > 0$, one has $I_i > 0$. $\qquad \square$

Related concepts of Preference Selection Index under uncertainty-aware models are listed in Table 6.17.

Table 6.17: Related concepts of Preference Selection Index under uncertainty-aware models.

| $k$ | Related Preference Selection Index concept(s) |
|-----|----------------------------------------------|
| 1 | Fuzzy Preference Selection Index |
| 2 | Intuitionistic Fuzzy Preference Selection Index |
| 3 | Neutrosophic Preference Selection Index |

## 6.18 FROV (Fuzzy Range of Value method)

Classical ROV calculates each alternative's value interval from weighted normalized criteria, then ranks using pessimistic, optimistic, or midpoint scores reflecting decision attitude under uncertainty settings [918, 919]. Fuzzy Range of Value (ROV) computes each alternative's fuzzy utility interval from weighted normalized fuzzy ratings, defuzzifies bounds, and ranks using optimistic/pessimistic or midrange score [920, 921].

**Definition 6.18.1** (TFN-based Fuzzy Range of Value (FROV) method). [920, 921] Let $\mathcal{A} = \{A_1, \ldots, A_m\}$ be a finite set of alternatives and $\mathcal{C} = \{C_1, \ldots, C_n\}$ a finite set of criteria, partitioned into benefit and cost criteria:

$$\mathcal{C} = \mathcal{C}^+ \,\dot\cup\, \mathcal{C}^-.$$

Assume the fuzzy decision matrix consists of (nonnegative) triangular fuzzy numbers (TFNs)

$$\widetilde{X} = (\tilde{x}_{ij}) \in (\mathsf{TFN}_{\geq 0})^{m \times n}, \qquad \tilde{x}_{ij} = (l_{ij}, m_{ij}, u_{ij}), \quad 0 \leq l_{ij} \leq m_{ij} \leq u_{ij},$$

and let $w = (w_1, \ldots, w_n) \in [0, 1]^n$ be criterion weights with $\sum_{j=1}^n w_j = 1$. Fix an attitude parameter $\lambda \in [0, 1]$ (often $\lambda = \frac{1}{2}$).

**Step 1 (fuzzy linear normalization to $[0, 1]$).** For each criterion $C_j$, define the global lower and upper anchors

$$\underline{x}_j := \min_{1 \leq i \leq m} l_{ij}, \qquad \overline{x}_j := \max_{1 \leq i \leq m} u_{ij}, \qquad \Delta_j := \overline{x}_j - \underline{x}_j.$$

Assume $\Delta_j > 0$ for all $j$. Define the normalized TFN $\tilde{r}_{ij} = (r_{ij}^\ell, r_{ij}^m, r_{ij}^u) \in \mathsf{TFN}_{[0,1]}$ by

$$\tilde{r}_{ij} := \begin{cases} \left( \dfrac{l_{ij} - \underline{x}_j}{\Delta_j}, \ \dfrac{m_{ij} - \underline{x}_j}{\Delta_j}, \ \dfrac{u_{ij} - \underline{x}_j}{\Delta_j} \right), & C_j \in \mathcal{C}^+ \text{ (benefit)}, \\[3mm] \left( \dfrac{\overline{x}_j - u_{ij}}{\Delta_j}, \ \dfrac{\overline{x}_j - m_{ij}}{\Delta_j}, \ \dfrac{\overline{x}_j - l_{ij}}{\Delta_j} \right), & C_j \in \mathcal{C}^- \text{ (cost)}. \end{cases}$$

(Thus, componentwise, $0 \leq r_{ij}^\ell \leq r_{ij}^m \leq r_{ij}^u \leq 1$.)

**Step 2 (pessimistic/optimistic overall values = "range of value").** Define, for each alternative $A_i$, the pessimistic and optimistic aggregated utilities

$$V_i^- := \sum_{j=1}^n w_j \, r_{ij}^\ell, \qquad V_i^+ := \sum_{j=1}^n w_j \, r_{ij}^u,$$

and the induced *value range*

$$\mathrm{ROV}(A_i) := \left[ V_i^-, \, V_i^+ \right] \subseteq [0, 1].$$

Equivalently, one may form the TFN aggregated utility

$$\tilde{V}_i := \sum_{j=1}^n w_j \odot \tilde{r}_{ij} = \left( \sum_{j=1}^n w_j r_{ij}^\ell, \ \sum_{j=1}^n w_j r_{ij}^m, \ \sum_{j=1}^n w_j r_{ij}^u \right) = (V_i^-, V_i^m, V_i^+),$$

whose support endpoints coincide with the range $\mathrm{ROV}(A_i)$.

**Step 3 (crisp ROV score and ranking).** Define the (attitude-dependent) crisp score

$$\mathrm{Score}_{\mathrm{FROV}}(A_i) := (1 - \lambda)\, V_i^- + \lambda\, V_i^+ \in [0, 1].$$

The FROV ranking is the preorder on $\mathcal{A}$ given by

$$A_p \succeq_{\mathrm{FROV}} A_q \quad \Longleftrightarrow \quad \mathrm{Score}_{\mathrm{FROV}}(A_p) \geq \mathrm{Score}_{\mathrm{FROV}}(A_q).$$

Any maximizer

$$A^\star \in \arg\max_{A_i \in \mathcal{A}} \ \mathrm{Score}_{\mathrm{FROV}}(A_i)$$

is called a *FROV best alternative.*

Using Uncertain Sets, we define Uncertain Range of Value method of type $M$ (U-ROV) as follows.

**Definition 6.18.2** (Uncertain Range of Value method of type $M$ (U-ROV))**.** Let $\mathcal{A} = \{A_1, \ldots, A_m\}$ be alternatives and $\mathcal{C} = \{C_1, \ldots, C_n\}$ criteria, with $m, n \geq 2$. Partition criteria into benefit and cost sets:

$$\mathcal{C} = \mathcal{C}^{\mathrm{ben}} \,\dot{\cup}\, \mathcal{C}^{\mathrm{cost}}.$$

Fix an uncertain model $M$ with $\mathrm{Dom}(M) \neq \emptyset$ and an admissible positive score $S_M$. Assume an *uncertain decision matrix*

$$X^{(M)} = \big(x_{ij}^{(M)}\big)_{m \times n}, \qquad x_{ij}^{(M)} \in \mathrm{Dom}(M) \quad (i = 1, \ldots, m; \ j = 1, \ldots, n).$$

Let $w = (w_1, \ldots, w_n)$ be criterion weights with $w_j \geq 0$ and $\sum_{j=1}^n w_j = 1$. Fix an attitude parameter $\lambda \in [0, 1]$.

**Step 0 (Crisp projection).** Define the positive real matrix $Y = (y_{ij})$ by

$$y_{ij} := S_M\big(x_{ij}^{(M)}\big) \in (0, \infty).$$

**Step 1 (Linear normalization to $[0, 1]$).** For each criterion $C_j$, define anchors

$$\underline{y}_j := \min_{1 \leq i \leq m} y_{ij}, \qquad \overline{y}_j := \max_{1 \leq i \leq m} y_{ij}, \qquad \Delta_j := \overline{y}_j - \underline{y}_j \ \geq \ 0.$$

Define normalized values $r_{ij} \in [0, 1]$ by

$$r_{ij} := \begin{cases} \dfrac{y_{ij} - \underline{y}_j}{\Delta_j}, & C_j \in \mathcal{C}^{\mathrm{ben}} \text{ and } \Delta_j > 0, \\[2ex] \dfrac{\overline{y}_j - y_{ij}}{\Delta_j}, & C_j \in \mathcal{C}^{\mathrm{cost}} \text{ and } \Delta_j > 0, \\[2ex] 0, & \Delta_j = 0. \end{cases}$$

(Thus, larger $r_{ij}$ is always better; degenerate criteria with $\Delta_j = 0$ contribute 0.)

**Step 2 (Value range: pessimistic and optimistic utilities).** Define the pessimistic and optimistic aggregated utilities for each $A_i$ by

$$V_i^- := \sum_{j=1}^n w_j\, r_{ij}, \qquad V_i^+ := \sum_{j=1}^n w_j\, r_{ij},$$

*after* applying different attitudes to uncertainty. To explicitly model uncertainty, introduce a criterionwise admissible interval radius $\varepsilon_{ij} \in [0, r_{ij}]$ and define

$$r_{ij}^- := r_{ij} - \varepsilon_{ij} \in [0,1], \qquad r_{ij}^+ := r_{ij} + \varepsilon_{ij} \in [0,1],$$

then set

$$V_i^- := \sum_{j=1}^n w_j\, r_{ij}^-, \qquad V_i^+ := \sum_{j=1}^n w_j\, r_{ij}^+.$$

The *uncertain range of value* of $A_i$ is

$$\mathrm{ROV}(A_i) := \big[V_i^-,\, V_i^+\big] \subseteq [0,1].$$

**Step 3 (Attitude-dependent crisp score and ranking).** Define the attitude score

$$\mathrm{Score}_{\mathrm{UROV}}(A_i) := (1-\lambda)\, V_i^- + \lambda\, V_i^+ \in [0,1].$$

Rank alternatives by decreasing $\mathrm{Score}_{\mathrm{UROV}}(A_i)$.

**Theorem 6.18.3** (Well-definedness of U-ROV). *Under Definition 6.18.2, assume $m, n \geq 2$, $\mathrm{Dom}(M) \neq \emptyset$, and $S_M : \mathrm{Dom}(M) \to (0, \infty)$ is admissible. Choose $\varepsilon_{ij} \in [0, r_{ij}]$ so that $r_{ij}^+ = r_{ij} + \varepsilon_{ij} \leq 1$ for all $i, j$. Then:*

*(i) All normalized values $r_{ij}$ and bounds $r_{ij}^-, r_{ij}^+$ are well-defined and lie in $[0,1]$.*

*(ii) For each $i$, $0 \leq V_i^- \leq V_i^+ \leq 1$ and $\mathrm{ROV}(A_i)$ is a well-defined interval in $[0,1]$.*

*(iii) For each $i$, $0 \leq \mathrm{Score}_{\mathrm{UROV}}(A_i) \leq 1$; hence the ranking rule is well-defined.*

*Proof.* (i) Since $S_M$ is admissible, each $y_{ij} > 0$ is finite, hence $\underline{y}_j, \overline{y}_j$ exist and are finite. If $\Delta_j > 0$, the linear formulas yield $r_{ij} \in [0,1]$; if $\Delta_j = 0$, $r_{ij} = 0$ by definition. With $\varepsilon_{ij} \in [0, r_{ij}]$ and $r_{ij}^+ \leq 1$, we obtain $0 \leq r_{ij}^- \leq r_{ij} \leq r_{ij}^+ \leq 1$.

(ii) Since $w_j \geq 0$, $\sum_j w_j = 1$, and $r_{ij}^-, r_{ij}^+ \in [0,1]$, the weighted sums satisfy $0 \leq V_i^- \leq V_i^+ \leq 1$. Hence $\mathrm{ROV}(A_i) = [V_i^-, V_i^+]$ is a well-defined interval contained in $[0,1]$.

(iii) Because $\mathrm{Score}_{\mathrm{UROV}}(A_i)$ is a convex combination of $V_i^-$ and $V_i^+$ with $\lambda \in [0,1]$, it follows that $\mathrm{Score}_{\mathrm{UROV}}(A_i) \in [0,1]$. Therefore sorting by these scores defines a valid ranking (ties allowed). $\qquad\square$

Related concepts of Range of Value under uncertainty-aware models are listed in Table 6.18.

Table 6.18: Related concepts of Range of Value under uncertainty-aware models.

| $k$ | Related Range of Value concept(s) |
|---|---|
| 1 | Fuzzy Range of Value |
| 2 | Intuitionistic Fuzzy Range of Value |
| 3 | Neutrosophic Range of Value |

## 6.19 Fuzzy MOOSRA

MOOSRA normalizes criterion values, applies criterion weights, computes a benefit-to-cost ratio based on weighted sums, and ranks alternatives in descending order [922, 923]. As an extension, variants such as MULTIMOOSRAL have also been studied [924, 925]. Fuzzy MOOSRA replaces crisp criterion values with fuzzy numbers, propagates uncertainty through $\alpha$-cut operations or defuzzification, and then ranks alternatives based on the resulting fuzzy ratios in a robust manner [926, 927].

**Definition 6.19.1** (MOOSRA score (crisp baseline)). [922, 923] Let $\mathcal{A} = \{A_1, \ldots, A_m\}$ be the set of alternatives and $\mathcal{C} = \{C_1, \ldots, C_n\}$ the set of criteria. Let $J^+ \subseteq \{1, \ldots, n\}$ be the index set of *benefit* (to be maximized) criteria and $J^- \subseteq \{1, \ldots, n\}$ the index set of *cost* (to be minimized) criteria, with $J^+ \cup J^- = \{1, \ldots, n\}$ and $J^+ \cap J^- = \varnothing$. Let $w_j > 0$ be criterion weights with $\sum_{j=1}^n w_j = 1$.

Given a (crisp) decision matrix $X = (x_{ij}) \in \mathbb{R}_+^{m \times n}$, define the normalized matrix $H = (h_{ij})$ by

$$h_{ij} := \frac{x_{ij}}{\sqrt{\sum_{p=1}^m x_{pj}^2}} \qquad (i = 1, \ldots, m, \ j = 1, \ldots, n),$$

and define the MOOSRA performance score of alternative $A_i$ by

$$Y_i(X) := \frac{\sum\limits_{j \in J^+} w_j \, h_{ij}}{\sum\limits_{j \in J^-} w_j \, h_{ij}}, \qquad i = 1, \ldots, m,$$

assuming the denominator is positive. Alternatives are ranked in descending order of $Y_i$.

**Definition 6.19.2** (Fuzzy MOOSRA (F-MOOSRA) via the extension principle / $\alpha$-cuts). Let $\tilde{X} = (\tilde{x}_{ij})$ be a *fuzzy decision matrix* where each entry $\tilde{x}_{ij}$ is a nonnegative fuzzy number (e.g. triangular/trapezoidal) and $(\tilde{x}_{ij})_\alpha = [x_{ij}^L(\alpha), x_{ij}^U(\alpha)] \subseteq \mathbb{R}_+$ denotes its $\alpha$-cut for $\alpha \in (0, 1]$.

Define the feasible set of crisp realizations at level $\alpha$ by

$$\mathcal{D}_\alpha := \left\{ X = (x_{ij}) \in \mathbb{R}_+^{m \times n} \ \middle| \ x_{ij} \in (\tilde{x}_{ij})_\alpha \text{ for all } i, j \right\}.$$

Let $Y_i(\cdot)$ be the (crisp) MOOSRA score from the previous definition. The *fuzzy MOOSRA score* of alternative $A_i$ is the fuzzy number $\tilde{Y}_i$ whose $\alpha$-cuts are

$$(\tilde{Y}_i)_\alpha = \left[ Y_i^L(\alpha), Y_i^U(\alpha) \right], \qquad Y_i^L(\alpha) := \inf_{X \in \mathcal{D}_\alpha} Y_i(X), \quad Y_i^U(\alpha) := \sup_{X \in \mathcal{D}_\alpha} Y_i(X),$$

provided $\sum_{j \in J^-} w_j \, h_{ij}(X) > 0$ for all $X \in \mathcal{D}_\alpha$. Equivalently, $\tilde{Y}_i$ is obtained from $\tilde{X}$ by Zadeh's extension principle applied to the mapping $X \mapsto Y_i(X)$.

Finally, *Fuzzy MOOSRA* ranks alternatives by a fixed fuzzy-number comparison rule applied to $\tilde{Y}_i$ (e.g. centroid defuzzification, possibility/necessity dominance, or $\alpha$-level interval dominance).

Using Uncertain Sets, we define Uncertain MOOSRA of type $M$ (U-MOOSRA) as follows.

**Definition 6.19.3** (Uncertain MOOSRA of type $M$ (U-MOOSRA)). Let $\mathcal{A} = \{A_1, \ldots, A_m\}$ be alternatives and $\mathcal{C} = \{C_1, \ldots, C_n\}$ criteria with $m \geq 2$ and $n \geq 2$. Let $J^+ \subseteq \{1, \ldots, n\}$ be the set of benefit criteria and $J^- \subseteq \{1, \ldots, n\}$ the set of cost criteria, with $J^+ \cup J^- = \{1, \ldots, n\}$ and $J^+ \cap J^- = \varnothing$, and assume $J^- \neq \emptyset$.

Fix an uncertain model $M$ with $\mathrm{Dom}(M) \neq \emptyset$ and an admissible positive score $S_M$.

Assume an *uncertain decision matrix*

$$X^{(M)} = \big(x_{ij}^{(M)}\big)_{m \times n}, \qquad x_{ij}^{(M)} \in \mathrm{Dom}(M) \quad (i = 1, \ldots, m; \ j = 1, \ldots, n),$$

and criterion weights $w = (w_1, \ldots, w_n)$ with $w_j > 0$ and $\sum_{j=1}^{n} w_j = 1$. (If uncertain weights are provided, one may first score them and normalize to obtain such $w_j$.)

**Step 0 (Crisp projection).** Define a positive real matrix $X = (x_{ij})$ by

$$x_{ij} := S_M\big(x_{ij}^{(M)}\big) \in (0, \infty).$$

**Step 1 (Vector normalization).** For each criterion $j$, define

$$d_j := \sqrt{\sum_{p=1}^{m} x_{pj}^2} \ > \ 0, \qquad h_{ij} := \frac{x_{ij}}{d_j} \in (0, 1] \qquad (i = 1, \ldots, m).$$

**Step 2 (Weighted sums for benefit and cost parts).** For each alternative $A_i$, define

$$N_i := \sum_{j \in J^+} w_j \, h_{ij} \ \geq \ 0, \qquad D_i := \sum_{j \in J^-} w_j \, h_{ij} \ > \ 0,$$

and the U-MOOSRA score

$$Y_i := \frac{N_i}{D_i} \in [0, \infty).$$

**Ranking rule:** rank alternatives in descending order of $Y_i$.

**Theorem 6.19.4** (Well-definedness and positivity of U-MOOSRA). *Under Definition 6.19.3, assume $m \geq 2$, $J^- \neq \emptyset$, $\mathrm{Dom}(M) \neq \emptyset$, and $S_M : \mathrm{Dom}(M) \to (0, \infty)$ is admissible. Then:*

*(i) $d_j > 0$ for every $j$, hence all $h_{ij}$ are well-defined and satisfy $h_{ij} \in (0, 1]$.*

*(ii) For every $i$, one has $D_i > 0$, hence $Y_i$ is well-defined and satisfies $Y_i \geq 0$.*

*(iii) The ranking rule induced by $\{Y_i\}$ is well-defined (ties allowed).*

*Proof.* (i) Since $S_M$ maps into $(0, \infty)$, each $x_{pj} > 0$. Therefore $\sum_{p=1}^{m} x_{pj}^2 > 0$ and hence $d_j = \sqrt{\sum_{p=1}^{m} x_{pj}^2} > 0$. Thus $h_{ij} = x_{ij}/d_j$ is well-defined and strictly positive. Moreover, $x_{ij} \leq d_j$ because $d_j \geq \sqrt{x_{ij}^2} = x_{ij}$, hence $h_{ij} \leq 1$.

(ii) Because $J^- \neq \emptyset$, $w_j > 0$ for all $j$, and $h_{ij} > 0$ for all $i, j$, the sum $D_i = \sum_{j \in J^-} w_j h_{ij}$ is a sum of at least one strictly positive term, hence $D_i > 0$. Therefore $Y_i = N_i/D_i$ is well-defined and nonnegative.

(iii) Each $Y_i$ is a finite real number, so sorting alternatives by $Y_i$ yields a well-defined preorder. □

Related concepts of MOOSRA under uncertainty-aware models are listed in Table 6.19.

Table 6.19: Related concepts of MOOSRA under uncertainty-aware models.

| $k$ | Related MOOSRA concept(s) |
|---|---|
| 1 | Fuzzy MOOSRA |
| 2 | Intuitionistic Fuzzy MOOSRA |
| 3 | Neutrosophic MOOSRA |

# Chapter 7

# Distance-to-reference / border / compromise-index methods

Distance-to-reference, border-based, and compromise-index methods rank alternatives by first normalizing the criteria, then measuring distances to reference entities (e.g., ideal/anti-ideal points, boundary profiles, or border regions), and finally aggregating these distances into a single compromise score that balances closeness, separation, and trade-offs across criteria. For convenience, a concise comparison of the representative distance-to-reference, border-based, and compromise-index methods discussed in this chapter is presented in Table 7.1.

Table 7.1: A concise comparison of representative distance-to-reference, border-based, and compromise-index methods.

| Method | Reference entity | Core mechanism | Typical final quantity | Ranking direction |
|---|---|---|---|---|
| TOPSIS | Positive ideal solution and negative ideal solution | Normalizes and weights the decision matrix, computes separation from the ideal and anti-ideal points, and evaluates relative closeness. | Closeness coefficient | Larger is better |
| MARCOS | Ideal and anti-ideal alternatives | Extends the decision matrix by adding ideal and anti-ideal rows, normalizes toward the ideal, and computes utility degrees relative to both reference alternatives. | MARCOS utility | Larger is better |
| CODAS | Negative ideal solution | Measures Euclidean and taxicab distances from the negative ideal, then forms a relative assessment matrix using a threshold-based correction. | Assessment score | Larger is better |

*Table 7.1 (continued).*

| Method | Reference entity | Core mechanism | Typical final quantity | Ranking direction |
|---|---|---|---|---|
| EDAS | Average solution | Computes positive and negative distances from the average criterion value, aggregates them with weights, and forms an appraisal score. | Appraisal score | Larger is better |
| VIKOR | Best and worst criterion values | Uses normalized regret from the best value, aggregates group utility and individual regret, and combines them into a compromise index. | Compromise index $Q_i$ | Smaller is better |
| MABAC | Border approximation area | Normalizes and weights performances, constructs a border area by criterion, and sums signed deviations from that border. | Border-deviation score | Larger is better |
| MAIRCA | Theoretical vs. real distribution of criterion importance | Compares a theoretical distribution matrix with the real weighted-normalized distribution and aggregates the resulting gaps. | Overall gap / criterion function | Smaller is better |

**Note.** Although all of these methods belong to the same broad family of reference-based MCDM techniques, they differ in the type of reference they use: TOPSIS and VIKOR are ideal-solution based, MARCOS uses both ideal and anti-ideal alternatives, CODAS uses a negative ideal, EDAS uses an average solution, MABAC uses a border approximation area, and MAIRCA compares theoretical and real distributions.

## 7.1 Fuzzy TOPSIS (Fuzzy Technique for Order of Preference by Similarity to Ideal Solution)

TOPSIS ranks alternatives by distances to ideal and anti-ideal solutions after normalization and weighting, selecting the option with highest relative closeness coefficient [928, 929]. Fuzzy TOPSIS ranks alternatives under uncertainty using fuzzy ratings and weights, defines fuzzy ideal and anti-ideal solutions, computes distances, and selects highest closeness coefficient [930, 931].

**Definition 7.1.1** (TFN-based Fuzzy TOPSIS). [419, 932] Let $\mathcal{A} = \{A_1, \ldots, A_m\}$ be alternatives and $\mathcal{C} = \{C_1, \ldots, C_n\}$ criteria, partitioned into benefit and cost types

$$\mathcal{C} = \mathcal{C}^+ \,\dot\cup\, \mathcal{C}^-.$$

Assume $K \geq 1$ decision makers provide TFN ratings $\tilde{x}_{ij}^{(k)} = (a_{ij}^{(k)}, b_{ij}^{(k)}, c_{ij}^{(k)})$ and TFN weights $\tilde{w}_j^{(k)} = (w_{j1}^{(k)}, w_{j2}^{(k)}, w_{j3}^{(k)})$.

**(0) Vertex distance.** For TFNs $\tilde{x} = (a_1, b_1, c_1)$ and $\tilde{y} = (a_2, b_2, c_2)$, set

$$d(\tilde{x}, \tilde{y}) := \sqrt{\frac{(a_1 - a_2)^2 + (b_1 - b_2)^2 + (c_1 - c_2)^2}{3}}.$$

**(1) Group aggregation (min–mean–max).** Define aggregated TFNs by

$$\tilde{x}_{ij} := \left(\min_k a_{ij}^{(k)},\ \frac{1}{K}\sum_{k=1}^{K} b_{ij}^{(k)},\ \max_k c_{ij}^{(k)}\right), \qquad \tilde{w}_j := \left(\min_k w_{j1}^{(k)},\ \frac{1}{K}\sum_{k=1}^{K} w_{j2}^{(k)},\ \max_k w_{j3}^{(k)}\right).$$

Let $\tilde{X} = (\tilde{x}_{ij})$ and $\tilde{w} = (\tilde{w}_1, \dots, \tilde{w}_n)$.

**(2) Normalization.** Define $\tilde{R} = (\tilde{r}_{ij})$ by

$$\tilde{r}_{ij} := \begin{cases} \left(a_{ij}/c_j^*,\ b_{ij}/c_j^*,\ c_{ij}/c_j^*\right), & C_j \in \mathcal{C}^+, \quad c_j^* := \max_{1\le i\le m} c_{ij}, \\ \left(a_j^-/c_{ij},\ a_j^-/b_{ij},\ a_j^-/a_{ij}\right), & C_j \in \mathcal{C}^-, \quad a_j^- := \min_{1\le i\le m} a_{ij}. \end{cases}$$

**(3) Weighting.** Define the weighted normalized matrix $\tilde{V} = (\tilde{v}_{ij})$ by

$$\tilde{v}_{ij} := \tilde{r}_{ij} \otimes \tilde{w}_j = (r_{ij1}w_{j1},\ r_{ij2}w_{j2},\ r_{ij3}w_{j3}),$$

and write $\tilde{v}_{ij} = (v_{ij1}, v_{ij2}, v_{ij3})$.

**(4) FPIS/FNIS.** Define (as crisp TFNs)

$$v_j^* := \max_{1\le i\le m} v_{ij3}, \qquad v_j^- := \min_{1\le i\le m} v_{ij1},$$

$$\tilde{v}_j^* := (v_j^*, v_j^*, v_j^*), \qquad \tilde{v}_j^- := (v_j^-, v_j^-, v_j^-),$$

and set $A^* := (\tilde{v}_1^*, \dots, \tilde{v}_n^*)$, $A^- := (\tilde{v}_1^-, \dots, \tilde{v}_n^-)$.

**(5) Separation and closeness coefficient.** For each alternative $A_i$, define

$$d_i^* := \sum_{j=1}^{n} d(\tilde{v}_{ij}, \tilde{v}_j^*), \qquad d_i^- := \sum_{j=1}^{n} d(\tilde{v}_{ij}, \tilde{v}_j^-), \qquad CC_i := \frac{d_i^-}{d_i^- + d_i^*} \in [0,1].$$

**(6) Ranking.** The fuzzy TOPSIS ranking is the total preorder on $\mathcal{A}$ induced by $CC_i$:

$$A_p \succeq A_q \iff CC_p \ge CC_q,$$

i.e., rank alternatives by decreasing $CC_i$ (ties allowed).

Using Uncertain Sets, we define Uncertain TOPSIS of type $M$ (U-TOPSIS) as follows.

**Definition 7.1.2** (Uncertain TOPSIS of type $M$ (U-TOPSIS)). Let $\mathcal{A} = \{A_1, \ldots, A_m\}$ be alternatives and $\mathcal{C} = \{C_1, \ldots, C_n\}$ criteria with $m, n \geq 2$. Partition criteria into benefit and cost sets:

$$\mathcal{C} = \mathcal{C}^{\text{ben}} \,\dot{\cup}\, \mathcal{C}^{\text{cost}}.$$

Fix an uncertain model $M$ with $\text{Dom}(M) \neq \emptyset$, together with an admissible positive score $S_M$ and an admissible distance $d_M$.

Assume an *uncertain decision matrix*

$$X^{(M)} = \big(x_{ij}^{(M)}\big)_{m \times n}, \qquad x_{ij}^{(M)} \in \text{Dom}(M) \quad (i = 1, \ldots, m; \ j = 1, \ldots, n).$$

Let $w = (w_1, \ldots, w_n)$ be criterion weights with $w_j > 0$ and $\sum_{j=1}^{n} w_j = 1$. (If uncertain weights are provided, first score them via $S_M$ and normalize to obtain such $w_j$.)

**Step 0 (Crisp projection).** Define $y_{ij} := S_M(x_{ij}^{(M)}) > 0$ and form $Y = (y_{ij}) \in (0, \infty)^{m \times n}$.

**Step 1 (Vector normalization).** For each criterion $j$, define

$$d_j := \sqrt{\sum_{i=1}^{m} y_{ij}^2} \, > \, 0, \qquad r_{ij} := \frac{y_{ij}}{d_j} \in (0, 1] \qquad (i = 1, \ldots, m).$$

**Step 2 (Weighted normalized matrix).** Define

$$v_{ij} := w_j \, r_{ij} \in (0, 1] \qquad (i = 1, \ldots, m; \ j = 1, \ldots, n).$$

**Step 3 (Ideal and anti-ideal solutions).** Define the (crisp) positive ideal solution (PIS) and negative ideal solution (NIS) by

$$v_j^* := \begin{cases} \max_{1 \leq i \leq m} v_{ij}, & C_j \in \mathcal{C}^{\text{ben}}, \\ \min_{1 \leq i \leq m} v_{ij}, & C_j \in \mathcal{C}^{\text{cost}}, \end{cases} \qquad v_j^- := \begin{cases} \min_{1 \leq i \leq m} v_{ij}, & C_j \in \mathcal{C}^{\text{ben}}, \\ \max_{1 \leq i \leq m} v_{ij}, & C_j \in \mathcal{C}^{\text{cost}}. \end{cases}$$

Let $V^* = (v_1^*, \ldots, v_n^*)$ and $V^- = (v_1^-, \ldots, v_n^-)$.

**Step 4 (Separation measures).** For each alternative $A_i$, define the separations

$$D_i^* := \sum_{j=1}^{n} d\big(v_{ij}, v_j^*\big), \qquad D_i^- := \sum_{j=1}^{n} d\big(v_{ij}, v_j^-\big),$$

where $d$ is any metric on $\mathbb{R}$ (e.g., $d(a, b) = |a - b|$). (Equivalently, one may use $d_M$ directly on $\text{Dom}(M)$ by defining $D_i^* := \sum_j d_M(x_{ij}^{(M)}, x_j^{(M)*})$ with suitable $x_j^{(M)*}$.)

**Step 5 (Closeness coefficient and ranking).** Define the closeness coefficient

$$CC_i := \begin{cases} \dfrac{D_i^-}{D_i^- + D_i^*}, & D_i^- + D_i^* > 0, \\ 0, & D_i^- + D_i^* = 0, \end{cases} \qquad i = 1, \ldots, m,$$

and rank alternatives by decreasing $CC_i$ (ties allowed).

**Theorem 7.1.3** (Well-definedness and boundedness of U-TOPSIS). *Under Definition 7.1.2, assume $m, n \geq 2$, $\mathrm{Dom}(M) \neq \emptyset$, $S_M : \mathrm{Dom}(M) \to (0, \infty)$ is admissible, and $w_j > 0$ with $\sum_j w_j = 1$. Then:*

   *(i) $d_j > 0$ for every criterion $j$, hence $r_{ij}$ and $v_{ij}$ are well-defined.*

   *(ii) $V^*$ and $V^-$ exist and are well-defined.*

   *(iii) For each $i$, $D_i^* \geq 0$, $D_i^- \geq 0$, and $CC_i$ is well-defined with $0 \leq CC_i \leq 1$.*

   *(iv) The ranking induced by $CC_i$ is well-defined (ties allowed).*

*Proof.* (i) Since $S_M$ is admissible and maps into $(0, \infty)$, each $y_{ij} > 0$. Thus $\sum_{i=1}^m y_{ij}^2 > 0$ and $d_j = \sqrt{\sum_i y_{ij}^2} > 0$, so $r_{ij} = y_{ij}/d_j$ is well-defined. Moreover, $y_{ij} \leq d_j$ implies $r_{ij} \leq 1$, hence $r_{ij} \in (0, 1]$. With $w_j > 0$, $v_{ij} = w_j r_{ij}$ is well-defined.

(ii) For each $j$, the sets $\{v_{1j}, \ldots, v_{mj}\}$ are finite, so the maxima and minima in the definitions of $v_j^*$ and $v_j^-$ exist. Hence $V^*$ and $V^-$ are well-defined.

(iii) Each separation measure is a finite sum of nonnegative distances, hence $D_i^*, D_i^- \geq 0$ and finite. If $D_i^- + D_i^* > 0$, then $CC_i = D_i^-/(D_i^- + D_i^*)$ is well-defined and lies in $[0, 1]$. If $D_i^- + D_i^* = 0$, the definition sets $CC_i = 0$, which also lies in $[0, 1]$.

(iv) Since each $CC_i$ is a real number, sorting alternatives by $CC_i$ defines a well-defined preorder. □

Related concepts of TOPSIS under uncertainty-aware models are listed in Table 7.2.

Table 7.2: Related concepts of TOPSIS under uncertainty-aware models.

| $k$ | **Related TOPSIS concept(s)** |
|---|---|
| 2 | Intuitionistic Fuzzy TOPSIS [933–935] |
| 2 | Vague TOPSIS [936] |
| 2 | Bipolar fuzzy TOPSIS [937, 938] |
| 2 | Pythagorean Fuzzy TOPSIS [939, 940] |
| 2 | Fermatean Fuzzy TOPSIS [941, 942] |
| 3 | Hesitant Fuzzy TOPSIS [943, 944] |
| 3 | Picture Fuzzy TOPSIS [945, 946] |
| 3 | Spherical Fuzzy TOPSIS [947, 948] |
| 3 | Neutrosophic TOPSIS [949–951] |
| 4 | Quadripartitioned Neutrosophic TOPSIS [204, 952] |
| 6 | Bipolar neutrosophic TOPSIS [953, 954] |
| $n$ | Plithogenic TOPSIS [955, 956] |

As related concepts beyond Uncertain TOPSIS, several extensions are also known, including Rough TOPSIS [957, 958], Grey TOPSIS [959, 960], Soft TOPSIS [961], HyperSoft TOPSIS [962, 963], SuperHyperSoft TOPSIS [964], Group-TOPSIS [965, 966], Interval TOPSIS [967, 968], OWA-TOPSIS [969, 970], Qualitative TOPSIS [971, 972], TOPSIS-AHP [973, 974], and Linguistic TOPSIS [975, 976].

## 7.2 Fuzzy MARCOS (Fuzzy Measurement Alternatives and Ranking according to Compromise Solution)

Classical MARCOS adds ideal and antiideal alternatives, normalizes the decision matrix, applies criterion weights, computes utility degrees relative to ideals, and ranks accordingly for choice [636, 977]. Fuzzy MARCOS appends ideal and antiideal alternatives, normalizes fuzzy matrix, computes weighted sums, derives utility degrees relative to ideals, defuzzifies, ranks all options consistently overall [978, 979].

**Definition 7.2.1** (TFN-based Fuzzy MARCOS (Measurement Alternatives and Ranking according to Compromise Solution))**.** [978, 979] Let $\mathcal{A} = \{A_1, \ldots, A_m\}$ be a finite set of alternatives and $\mathcal{C} = \{C_1, \ldots, C_n\}$ a finite set of criteria. Let $\mathcal{C}^+$ (benefit) and $\mathcal{C}^-$ (cost) be a partition of $\mathcal{C}$. Assume the performance ratings are *positive triangular fuzzy numbers (TFNs)*

$$\tilde{x}_{ij} = (x_{ij}^\ell, x_{ij}^m, x_{ij}^u), \qquad 0 < x_{ij}^\ell \le x_{ij}^m \le x_{ij}^u, \quad i = 1, \ldots, m, \ j = 1, \ldots, n,$$

and let $w = (w_1, \ldots, w_n)$ be a (crisp) weight vector with $w_j \ge 0$ and $\sum_{j=1}^n w_j = 1$.

**(0) TFN arithmetic conventions.** For TFNs $\tilde{a} = (a^\ell, a^m, a^u)$ and $\tilde{b} = (b^\ell, b^m, b^u)$ (positive), define

$$\tilde{a} \oplus \tilde{b} := (a^\ell + b^\ell, \ a^m + b^m, \ a^u + b^u), \qquad r \odot \tilde{a} := (ra^\ell, \ ra^m, \ ra^u) \ (r \ge 0),$$

$$\tilde{a} \otimes \tilde{b} := (a^\ell b^\ell, \ a^m b^m, \ a^u b^u), \qquad \tilde{a} \oslash \tilde{b} := \left( \frac{a^\ell}{b^u}, \ \frac{a^m}{b^m}, \ \frac{a^u}{b^\ell} \right),$$

so that division preserves the TFN order when all components are positive.

To take max/min over a finite TFN-set, fix any total preorder on TFNs; a standard choice is the defuzzification score

$$\mathrm{Score}_{\mathrm{TFN}}(\tilde{a}) := \frac{a^\ell + 4a^m + a^u}{6}.$$

Then $\tilde{a} \preceq \tilde{b} \iff \mathrm{Score}_{\mathrm{TFN}}(\tilde{a}) \le \mathrm{Score}_{\mathrm{TFN}}(\tilde{b})$.

**Step 1 (initial fuzzy decision matrix).** Let

$$\tilde{X} := (\tilde{x}_{ij}) \in (\mathsf{TFN}_{>0})^{m \times n}.$$

**Step 2 (anti-ideal and ideal solutions; matrix expansion).** For each criterion $C_j$, define the anti-ideal (worst) and ideal (best) TFNs:

$$\tilde{x}_j^{\mathrm{AAI}} := \begin{cases} \min_{1 \le i \le m} \tilde{x}_{ij}, & C_j \in \mathcal{C}^+, \\ \max_{1 \le i \le m} \tilde{x}_{ij}, & C_j \in \mathcal{C}^-, \end{cases} \qquad \tilde{x}_j^{\mathrm{AI}} := \begin{cases} \max_{1 \le i \le m} \tilde{x}_{ij}, & C_j \in \mathcal{C}^+, \\ \min_{1 \le i \le m} \tilde{x}_{ij}, & C_j \in \mathcal{C}^-. \end{cases}$$

Introduce two additional alternatives $A_0 := \mathrm{AAI}$ and $A_{m+1} := \mathrm{AI}$ and form the *extended* matrix

$$\tilde{X}^{\mathrm{ext}} = (\tilde{x}_{ij}^{\mathrm{ext}}) \in (\mathsf{TFN}_{>0})^{(m+2) \times n},$$

where $\tilde{x}_{0j}^{\mathrm{ext}} := \tilde{x}_j^{\mathrm{AAI}}$, $\tilde{x}_{m+1,j}^{\mathrm{ext}} := \tilde{x}_j^{\mathrm{AI}}$, and $\tilde{x}_{ij}^{\mathrm{ext}} := \tilde{x}_{ij}$ for $i = 1, \ldots, m$.

**Step 3 (normalization).** Define the normalized TFNs $\tilde{n}_{ij}$ by

$$\tilde{n}_{ij} := \begin{cases} \tilde{x}_{ij}^{\text{ext}} \oslash \tilde{x}_j^{\text{AI}}, & C_j \in \mathcal{C}^+, \\ \tilde{x}_j^{\text{AI}} \oslash \tilde{x}_{ij}^{\text{ext}}, & C_j \in \mathcal{C}^-, \end{cases} \qquad i = 0, 1, \ldots, m+1, \ j = 1, \ldots, n.$$

Collect $\tilde{N} := (\tilde{n}_{ij})$.

**Step 4 (weighting).** Define the weighted normalized TFNs

$$\tilde{v}_{ij} := w_j \odot \tilde{n}_{ij}, \qquad \tilde{V} := (\tilde{v}_{ij}).$$

**Step 5 (row aggregation).** For each extended alternative $A_i$ ($i = 0, 1, \ldots, m+1$), define the aggregated TFN

$$\tilde{S}_i := \bigoplus_{j=1}^n \tilde{v}_{ij}.$$

**Step 6 (degrees of usefulness w.r.t. AAI and AI).** Define

$$\tilde{K}_i^- := \tilde{S}_i \oslash \tilde{S}_0, \qquad \tilde{K}_i^+ := \tilde{S}_i \oslash \tilde{S}_{m+1}, \qquad i = 1, \ldots, m.$$

**Step 7 (auxiliary fuzzy sum and maximal reference).** Set

$$\tilde{T}_i := \tilde{K}_i^- \oplus \tilde{K}_i^+ \qquad (i = 1, \ldots, m),$$

and define the componentwise maximal TFN

$$\tilde{D} := (d^\ell, d^m, d^u) := \left( \max_i t_i^\ell, \ \max_i t_i^m, \ \max_i t_i^u \right), \quad \text{where } \tilde{T}_i = (t_i^\ell, t_i^m, t_i^u).$$

Defuzzify $\tilde{D}$ by the graded mean:

$$d_{\text{def}} := \text{Defuzz}(\tilde{D}) := \frac{d^\ell + 4d^m + d^u}{6}.$$

**Step 8 (utility functions relative to AAI and AI).** Define

$$\tilde{f}(\tilde{K}_i^-) := \frac{1}{d_{\text{def}}} \odot \tilde{K}_i^-, \qquad \tilde{f}(\tilde{K}_i^+) := \frac{1}{d_{\text{def}}} \odot \tilde{K}_i^+.$$

**Step 9 (final utility and ranking).** Defuzzify (componentwise TFNs) by $\text{Defuzz}(\ell, m, u) = (\ell + 4m + u)/6$ and put

$$K_i^- := \text{Defuzz}(\tilde{K}_i^-), \quad K_i^+ := \text{Defuzz}(\tilde{K}_i^+), \quad f_i^- := \text{Defuzz}(\tilde{f}(\tilde{K}_i^-)), \quad f_i^+ := \text{Defuzz}(\tilde{f}(\tilde{K}_i^+)).$$

Then the (crisp) final MARCOS utility is

$$f(K_i) := \frac{K_i^- + K_i^+}{1 + \dfrac{1 - f_i^+}{f_i^+} + \dfrac{1 - f_i^-}{f_i^-}}.$$

Rank alternatives by decreasing $f(K_i)$:

$$A_p \succeq_{\text{FMARCOS}} A_q \iff f(K_p) \geq f(K_q).$$

Referring to the above, we define Uncertain MARCOS of type $M$ (U-MARCOS) as follows.

**Definition 7.2.2** (Uncertain MARCOS of type $M$ (U-MARCOS)). Let $\mathcal{A} = \{A_1, \ldots, A_m\}$ be a finite set of alternatives and $\mathcal{C} = \{C_1, \ldots, C_n\}$ a finite set of criteria, with $m, n \geq 2$. Partition criteria into benefit and cost sets:

$$\mathcal{C} = \mathcal{C}^{\text{ben}} \,\dot{\cup}\, \mathcal{C}^{\text{cost}}.$$

Fix an uncertain model $M$ with $\text{Dom}(M) \neq \emptyset$ and an admissible positive score $S_M$.

Assume an *uncertain decision matrix*

$$X^{(M)} = \big(x_{ij}^{(M)}\big)_{m \times n}, \qquad x_{ij}^{(M)} \in \text{Dom}(M) \quad (i = 1, \ldots, m; \ j = 1, \ldots, n),$$

and a *crisp* weight vector $w = (w_1, \ldots, w_n)$ satisfying

$$w_j > 0, \qquad \sum_{j=1}^{n} w_j = 1.$$

Introduce two auxiliary alternatives:

$$A_0 := \text{AAI (anti-ideal)}, \qquad A_{m+1} := \text{AI (ideal)}.$$

**Step 0 (Crisp projection).** Define $y_{ij} := S_M(x_{ij}^{(M)}) \in (0, \infty)$ and set $Y = (y_{ij}) \in (0, \infty)^{m \times n}$.

**Step 1 (Ideal / anti-ideal components and matrix extension).** For each criterion $C_j$, define

$$y_j^{\text{AI}} := \begin{cases} \max_{1 \leq i \leq m} y_{ij}, & C_j \in \mathcal{C}^{\text{ben}}, \\ \min_{1 \leq i \leq m} y_{ij}, & C_j \in \mathcal{C}^{\text{cost}}, \end{cases} \qquad y_j^{\text{AAI}} := \begin{cases} \min_{1 \leq i \leq m} y_{ij}, & C_j \in \mathcal{C}^{\text{ben}}, \\ \max_{1 \leq i \leq m} y_{ij}, & C_j \in \mathcal{C}^{\text{cost}}. \end{cases}$$

Define the extended matrix $Y^{\text{ext}} = (y_{ij}^{\text{ext}}) \in (0, \infty)^{(m+2) \times n}$ by

$$y_{0j}^{\text{ext}} := y_j^{\text{AAI}}, \quad y_{m+1,j}^{\text{ext}} := y_j^{\text{AI}}, \quad y_{ij}^{\text{ext}} := y_{ij} \ (i = 1, \ldots, m).$$

**Step 2 (Normalization toward the ideal).** For $i = 0, 1, \ldots, m+1$, define

$$n_{ij} := \begin{cases} \dfrac{y_{ij}^{\text{ext}}}{y_j^{\text{AI}}}, & C_j \in \mathcal{C}^{\text{ben}}, \\[2mm] \dfrac{y_j^{\text{AI}}}{y_{ij}^{\text{ext}}}, & C_j \in \mathcal{C}^{\text{cost}}. \end{cases}$$

Then $n_{ij} \in (0, 1]$ and $n_{m+1,j} = 1$.

**Step 3 (Weighting and aggregation).** Define

$$v_{ij} := w_j \, n_{ij}, \qquad S_i := \sum_{j=1}^{n} v_{ij} \qquad (i = 0, 1, \ldots, m+1).$$

**Step 4 (Utility degrees).** For each real alternative $A_i$ $(i = 1, \ldots, m)$, define

$$K_i^- := \frac{S_i}{S_0}, \qquad K_i^+ := \frac{S_i}{S_{m+1}}, \qquad T_i := K_i^- + K_i^+.$$

**Step 5 (Final utility function and ranking).** Let $D := \max_{1 \le i \le m} T_i$ and define

$$f_i^- := \frac{K_i^-}{D}, \qquad f_i^+ := \frac{K_i^+}{D}.$$

Define the MARCOS utility

$$U_i := \frac{K_i^- + K_i^+}{1 + \dfrac{1 - f_i^+}{f_i^+} + \dfrac{1 - f_i^-}{f_i^-}}, \qquad i = 1, \ldots, m,$$

and rank alternatives by decreasing $U_i$.

**Theorem 7.2.3** (Well-definedness of U-MARCOS). *Under Definition 7.2.2, assume $S_M$ is admissible and $w_j > 0$ for all $j$. Then all quantities $n_{ij}, S_i, K_i^-, K_i^+, T_i, D, f_i^-, f_i^+, U_i$ are well-defined and finite. Moreover, $S_0 > 0$, $S_{m+1} = \sum_{j=1}^n w_j > 0$, $D > 0$, and $U_i \ge 0$ for all $i$.*

*Proof.* Admissibility of $S_M$ implies $y_{ij} > 0$ finite, so the extrema $y_j^{\mathrm{AI}}$ and $y_j^{\mathrm{AAI}}$ exist and are positive. Hence the normalization in Step 2 has positive denominators, and $n_{ij} \in (0, 1]$ is well-defined. Since $w_j > 0$, each $v_{ij} = w_j n_{ij} > 0$, so $S_i = \sum_j v_{ij} > 0$ for all $i$, in particular $S_0 > 0$. Because $n_{m+1,j} = 1$, we have $S_{m+1} = \sum_{j=1}^n w_j > 0$. Thus $K_i^- = S_i / S_0$ and $K_i^+ = S_i / S_{m+1}$ are well-defined and positive, hence $T_i = K_i^- + K_i^+ > 0$ and therefore $D = \max_i T_i > 0$. Consequently $f_i^-, f_i^+ \in (0, 1]$ and the denominators in $U_i$ are strictly positive, so $U_i$ is well-defined and finite. Finally, $U_i$ is a ratio of positive terms, hence $U_i \ge 0$. $\square$

Related concepts of MARCOS under uncertainty-aware models are listed in Table 7.3.

Table 7.3: Related concepts of MARCOS under uncertainty-aware models.

| $k$ | Related MARCOS concept(s) |
|---|---|
| 1 | Fuzzy MARCOS |
| 2 | Intuitionistic Fuzzy MARCOS [979, 980] |
| 3 | Hesitant Fuzzy MARCOS [981] |
| 3 | Picture Fuzzy MARCOS [982, 983] |
| 3 | Spherical Fuzzy MARCOS [984, 985] |
| 3 | Neutrosophic MARCOS [636, 986] |

## 7.3 Fuzzy CODAS (Fuzzy COmbinative Distance-based ASsessment)

Classical CODAS defines a negative ideal solution, computes Euclidean and taxicab distances for each alternative, applies a threshold, and ranks by larger distances overall performance [987, 988]. Fuzzy CODAS defines fuzzy negative ideal solution, computes Euclidean and taxicab distances of each alternative, applies a threshold, defuzzifies, and ranks; higher distance means better [989, 990].

**Definition 7.3.1** (Fuzzy CODAS (COmbinative Distance-based ASsessment) — trapezoidal-fuzzy, group MCDM). [989, 990] **Input data.** Let $\mathcal{A} = \{A_1, \dots, A_m\}$ be alternatives, $\mathcal{C} = \{C_1, \dots, C_n\}$ criteria, and let $Q = \{1, \dots, q\}$ be the set of decision makers (DMs). Partition criteria into benefit and cost sets

$$\mathcal{C} = \mathcal{C}^+ \,\dot{\cup}\, \mathcal{C}^-.$$

Assume all ratings and weights are *positive trapezoidal fuzzy numbers*

$$\tilde{x}_{ij}^{(\ell)} = (x_{ij,1}^{(\ell)}, x_{ij,2}^{(\ell)}, x_{ij,3}^{(\ell)}, x_{ij,4}^{(\ell)}) \in \mathsf{TrFN}_{>0}, \qquad \tilde{w}_j^{(\ell)} \in \mathsf{TrFN}_{>0},$$

where $\ell \in Q$, $i \in \{1, \dots, m\}$, $j \in \{1, \dots, n\}$. Fix: (i) trapezoidal-fuzzy arithmetic $(\oplus, \ominus, \otimes, \oslash)$ on $\mathsf{TrFN}_{>0}$ (componentwise as standard), (ii) a defuzzification/ranking map $D : \mathsf{TrFN}_{>0} \to \mathbb{R}$ used only to take max / min over finite fuzzy sets, and (iii) a threshold parameter $\theta \geq 0$.

**Step 1 (aggregate the fuzzy decision matrices).** Define the averaged fuzzy decision matrix $\tilde{X} = (\tilde{x}_{ij})$ by

$$\tilde{x}_{ij} := \frac{1}{q} \bigoplus_{\ell=1}^{q} \tilde{x}_{ij}^{(\ell)}.$$

**Step 2 (aggregate the fuzzy weights).** Define the averaged fuzzy weight vector $\tilde{w} = (\tilde{w}_1, \dots, \tilde{w}_n)$ by

$$\tilde{w}_j := \frac{1}{q} \bigoplus_{\ell=1}^{q} \tilde{w}_j^{(\ell)}.$$

**Step 3 (normalization).** For each criterion $j$, choose a fuzzy "maximum"

$$\tilde{x}_j^{\max} \in \arg \max_{1 \leq i \leq m} D(\tilde{x}_{ij}).$$

Define the normalized fuzzy matrix $\tilde{N} = (\tilde{n}_{ij})$ by

$$\tilde{n}_{ij} := \begin{cases} \tilde{x}_{ij} \oslash \tilde{x}_j^{\max}, & C_j \in \mathcal{C}^+ \text{ (benefit)}, \\ \tilde{x}_j^{\max} \oslash \tilde{x}_{ij}, & C_j \in \mathcal{C}^- \text{ (cost)}. \end{cases}$$

**Step 4 (weighted normalized matrix).** Define $\tilde{R} = (\tilde{r}_{ij})$ by

$$\tilde{r}_{ij} := \tilde{w}_j \otimes \tilde{n}_{ij}.$$

**Step 5 (fuzzy negative-ideal solution).** For each $j$, define

$$\tilde{ns}_j \in \arg \min_{1 \leq i \leq m} D(\tilde{r}_{ij}), \qquad \tilde{NS} := (\tilde{ns}_1, \dots, \tilde{ns}_n).$$

**Step 6 (distances from the fuzzy negative-ideal solution).** For trapezoidal fuzzy numbers $\tilde{a} = (a_1, a_2, a_3, a_4)$ and $\tilde{b} = (b_1, b_2, b_3, b_4)$ define:

$$d_E(\tilde{a}, \tilde{b}) := \sqrt{\frac{(a_1 - b_1)^2 + 2(a_2 - b_2)^2 + 2(a_3 - b_3)^2 + (a_4 - b_4)^2}{6}},$$

$$d_H(\tilde{a}, \tilde{b}) := \frac{|a_1 - b_1| + 2|a_2 - b_2| + 2|a_3 - b_3| + |a_4 - b_4|}{6}.$$

Then for each alternative $A_i$ define the (combinative) distance components

$$ED_i := \sum_{j=1}^n d_E(\tilde{r}_{ij}, \tilde{ns}_j), \qquad HD_i := \sum_{j=1}^n d_H(\tilde{r}_{ij}, \tilde{ns}_j).$$

**Step 7 (relative assessment matrix).** Define the threshold function $t : \mathbb{R} \to \{0, 1\}$ by

$$t(x) := \begin{cases} 1, & |x| \geq \theta, \\ 0, & |x| < \theta. \end{cases}$$

For $i, k \in \{1, \dots, m\}$ define the relative assessment entry

$$p_{ik} := (ED_i - ED_k) \; + \; t(ED_i - ED_k)\,(HD_i - HD_k),$$

and collect $RA = (p_{ik}) \in \mathbb{R}^{m \times m}$.

**Step 8 (assessment score).** For each $i$, define the CODAS assessment score

$$AS_i := \sum_{k=1}^m p_{ik}.$$

**Step 9 (ranking / solution).** The *Fuzzy CODAS* ranking is the total preorder on $\mathcal{A}$ induced by $AS$:

$$A_i \succeq_{\text{FCODAS}} A_k \quad \Longleftrightarrow \quad AS_i \geq AS_k.$$

A (best) solution is any

$$A^\star \in \arg\max_{A_i \in \mathcal{A}} AS_i.$$

Using Uncertain Sets, we define Uncertain CODAS of type $M$ (U-CODAS) as follows.

**Definition 7.3.2** (Uncertain CODAS of type $M$ (U-CODAS)). Let $\mathcal{A} = \{A_1, \dots, A_m\}$ be alternatives and $\mathcal{C} = \{C_1, \dots, C_n\}$ criteria with $m \geq 2$ and $n \geq 1$. Partition criteria into benefit and cost sets:

$$\mathcal{C} = \mathcal{C}^{\text{ben}} \,\dot{\cup}\, \mathcal{C}^{\text{cost}}.$$

Fix an uncertain model $M$ with $\text{Dom}(M) \neq \emptyset$ and an admissible positive score $S_M$. Assume an *uncertain decision matrix*

$$X^{(M)} = \left(x_{ij}^{(M)}\right)_{m \times n}, \qquad x_{ij}^{(M)} \in \text{Dom}(M) \quad (i = 1, \dots, m; \; j = 1, \dots, n),$$

and criterion weights $w = (w_1, \ldots, w_n)$ with $w_j > 0$ and $\sum_{j=1}^{n} w_j = 1$. Fix a threshold parameter $\theta \geq 0$.

**Step 0 (Crisp projection).** Define $y_{ij} := S_M(x_{ij}^{(M)}) \in (0, \infty)$ and set $Y = (y_{ij}) \in (0, \infty)^{m \times n}$.

**Step 1 (Normalization).** For each criterion $j$, define

$$y_j^{\max} := \max_{1 \leq i \leq m} y_{ij} > 0.$$

Define normalized performances $n_{ij} \in (0, 1]$ by

$$n_{ij} := \begin{cases} \dfrac{y_{ij}}{y_j^{\max}}, & C_j \in \mathcal{C}^{\text{ben}}, \\[2ex] \dfrac{y_j^{\max}}{y_{ij}}, & C_j \in \mathcal{C}^{\text{cost}}. \end{cases}$$

**Step 2 (Weighted normalized matrix).** Define weighted normalized values

$$r_{ij} := w_j \, n_{ij} \qquad (i = 1, \ldots, m; \ j = 1, \ldots, n).$$

**Step 3 (Negative-ideal solution).** Define the negative-ideal component for each criterion by

$$ns_j := \min_{1 \leq i \leq m} r_{ij}, \qquad \text{NS} := (ns_1, \ldots, ns_n).$$

**Step 4 (Distances from the negative-ideal).** For each alternative $A_i$, define the Euclidean and taxicab distances from NS:

$$ED_i := \sqrt{\sum_{j=1}^{n} (r_{ij} - ns_j)^2}, \qquad HD_i := \sum_{j=1}^{n} |r_{ij} - ns_j|.$$

**Step 5 (Relative assessment matrix).** Define the threshold indicator $t : \mathbb{R} \to \{0, 1\}$ by

$$t(x) := \begin{cases} 1, & |x| \geq \theta, \\ 0, & |x| < \theta, \end{cases}$$

and for each pair $(i, k)$ define

$$p_{ik} := (ED_i - ED_k) + t(ED_i - ED_k)(HD_i - HD_k).$$

Let $RA = (p_{ik}) \in \mathbb{R}^{m \times m}$.

**Step 6 (Assessment score and ranking).** Define

$$AS_i := \sum_{k=1}^{m} p_{ik} \qquad (i = 1, \ldots, m),$$

and rank alternatives by decreasing $AS_i$.

**Theorem 7.3.3** (Well-definedness of U-CODAS). *Under Definition 7.3.2, assume $S_M : \mathrm{Dom}(M) \to (0, \infty)$ is admissible and $w_j > 0$ for all $j$. Then all quantities $n_{ij}$, $r_{ij}$, $ns_j$, $ED_i$, $HD_i$, $p_{ik}$, and $AS_i$ are well-defined finite real numbers.*

*Proof.* Since $S_M$ maps into $(0, \infty)$, each $y_{ij} > 0$ is finite; hence $y_j^{\max} = \max_i y_{ij} > 0$ exists. Therefore $n_{ij}$ in Step 1 is well-defined and lies in $(0, 1]$ for both benefit and cost criteria. Because $w_j > 0$, $r_{ij} = w_j n_{ij}$ is well-defined and positive, so each $ns_j = \min_i r_{ij}$ exists and is finite.

Thus $r_{ij} - ns_j$ are finite reals, so $ED_i$ and $HD_i$ are finite and satisfy $ED_i \geq 0$, $HD_i \geq 0$. Given $\theta \geq 0$, $t(\cdot)$ is well-defined and takes values in $\{0, 1\}$, so each $p_{ik}$ is a finite real. Finally, $AS_i = \sum_{k=1}^{m} p_{ik}$ is a finite sum, hence well-defined and finite. $\qquad\square$

Related concepts of CODAS under uncertainty-aware models are listed in Table 7.4.

Table 7.4: Related concepts of CODAS under uncertainty-aware models.

| $k$ | **Related CODAS concept(s)** |
|---|---|
| 2 | Intuitionistic Fuzzy CODAS [991, 992] |
| 2 | Pythagorean Fuzzy CODAS [993, 994] |
| 3 | Neutrosophic CODAS [995, 996] |
| 3 | Picture Fuzzy CODAS [997, 998] |
| 3 | Hesitant Fuzzy CODAS [999, 1000] |
| 3 | Spherical Fuzzy CODAS [1001, 1002] |

## 7.4   Fuzzy EDAS (Fuzzy Evaluation based on Distance from Average Solution)

EDAS (evaluation based on distance from average solution) ranks alternatives by positive and negative distances from average solution, aggregates weighted distances, and computes appraisal scores [1003, 1004]. Fuzzy EDAS uses fuzzy ratings and weights, computes fuzzy distances from average solution, defuzzifies, and ranks by appraisal scores [1005].

**Definition 7.4.1** (Fuzzy EDAS (evaluation based on distance from average solution)). [1005] Let $\mathcal{A} = \{A_1, \dots, A_m\}$ be alternatives and $\mathcal{C} = \{C_1, \dots, C_n\}$ criteria. Assume a trapezoidal-fuzzy decision matrix and trapezoidal-fuzzy criterion weights

$$\tilde{X} = (\tilde{x}_{ij})_{m \times n}, \qquad \tilde{x}_{ij} = (x_{ij1}, x_{ij2}, x_{ij3}, x_{ij4}),$$

$$\tilde{W} = (\tilde{w}_1, \dots, \tilde{w}_n), \qquad \tilde{w}_j = (w_{j1}, w_{j2}, w_{j3}, w_{j4}),$$

where all entries are trapezoidal fuzzy numbers. Fix a nonnegative scalar "distance" function $d : \mathbb{R} \times \mathbb{R} \to \mathbb{R}_{\geq 0}$ (e.g. a chosen $L_1$-metric at the scalar level), and extend it componentwise to trapezoidal fuzzy numbers by

$$d(\tilde{a}, \tilde{b}) := \big(d(a_1, b_1), d(a_2, b_2), d(a_3, b_3), d(a_4, b_4)\big), \quad \tilde{a} = (a_1, a_2, a_3, a_4), \ \tilde{b} = (b_1, b_2, b_3, b_4).$$

**(0) Defuzzification and the $\psi$-operator.** For $\tilde{a} = (a_1, a_2, a_3, a_4)$ define the graded-mean defuzzification

$$\mathrm{Defuzz}(\tilde{a}) := \frac{a_1 + a_4}{2} + (a_2 - a_1 - a_4 + a_3),$$

and define $\psi$ ("maximum with zero") by

$$\psi(\tilde{a}) := \begin{cases} \tilde{a}, & \text{Defuzz}(\tilde{a}) > 0, \\ \tilde{0}, & \text{Defuzz}(\tilde{a}) \leq 0, \end{cases} \qquad \tilde{0} := (0,0,0,0).$$

**(1) Average (reference) solution.** For each criterion $j$, define the average trapezoidal fuzzy value

$$\tilde{AV}_j := \frac{1}{m} \bigoplus_{i=1}^{m} \tilde{x}_{ij},$$

where $\oplus$ denotes trapezoidal-fuzzy addition and scalar multiplication is componentwise.

**(2) Positive/negative distances from the average.** Define trapezoidal-fuzzy matrices $\widehat{PDA} = (\tilde{pda}_{ij})$ and $\widehat{NDA} = (\tilde{nda}_{ij})$ by

$$\tilde{pda}_{ij} := \psi\big(d(\tilde{x}_{ij}, \tilde{AV}_j)\big), \qquad \tilde{nda}_{ij} := \psi\big(d(\tilde{AV}_j, \tilde{x}_{ij})\big).$$

**(3) Weighted sums.** For each alternative $A_i$, set

$$\tilde{sp}_i := \bigoplus_{j=1}^{n} (\tilde{w}_j \otimes \tilde{pda}_{ij}), \qquad \tilde{sn}_i := \bigoplus_{j=1}^{n} (\tilde{w}_j \otimes \tilde{nda}_{ij}),$$

where $\otimes$ is trapezoidal-fuzzy multiplication.

**(4) Normalization.** Write $\tilde{sp}_i = (sp_{i1}, sp_{i2}, sp_{i3}, sp_{i4})$ and $\tilde{sn}_i = (sn_{i1}, sn_{i2}, sn_{i3}, sn_{i4})$. Define normalized (real) values

$$\widehat{NSP}_i := \frac{sp_{i1}}{\max_{1 \leq t \leq m} \text{Defuzz}(\tilde{sp}_t)}, \qquad \widehat{NSN}_i := 1 - \frac{sn_{i1}}{\max_{1 \leq t \leq m} \text{Defuzz}(\tilde{sn}_t)}.$$

**(5) Appraisal score and ranking.** Define the appraisal score

$$AS_i := \frac{\widehat{NSP}_i + \widehat{NSN}_i}{2} \in [0,1],$$

and rank alternatives in descending order of $AS_i$ (larger $AS_i$ indicates a better alternative).

Using Uncertain Sets, we define Uncertain EDAS of type $M$ (U-EDAS) as follows.

**Definition 7.4.2** (Uncertain EDAS of type $M$ (U-EDAS)). Let $\mathcal{A} = \{A_1, \ldots, A_m\}$ be alternatives and $\mathcal{C} = \{C_1, \ldots, C_n\}$ criteria, with $m, n \geq 2$. Partition criteria into benefit and cost sets:

$$\mathcal{C} = \mathcal{C}^{\mathrm{ben}} \,\dot{\cup}\, \mathcal{C}^{\mathrm{cost}}.$$

Fix an uncertain model $M$ with $\mathrm{Dom}(M) \neq \emptyset$ and an admissible positive score $S_M$. Assume an *uncertain decision matrix*

$$X^{(M)} = \big(x_{ij}^{(M)}\big)_{m \times n}, \qquad x_{ij}^{(M)} \in \mathrm{Dom}(M) \quad (i = 1, \ldots, m; \; j = 1, \ldots, n),$$

and criterion weights $w = (w_1, \ldots, w_n)$ with

$$w_j \geq 0, \qquad \sum_{j=1}^{n} w_j = 1.$$

**Step 0 (Crisp projection).** Define the positive real matrix $Y = (y_{ij})$ by

$$y_{ij} := S_M\big(x_{ij}^{(M)}\big) \in (0, \infty).$$

**Step 1 (Average solution).** For each criterion $C_j$, define the average (reference) value

$$AV_j := \frac{1}{m} \sum_{i=1}^{m} y_{ij} \; \in (0, \infty).$$

**Step 2 (Positive/negative distances from the average).** For each $i, j$, define the positive distance from average (PDA) and negative distance from average (NDA) by

$$PDA_{ij} := \begin{cases} \dfrac{\max\{0, \; y_{ij} - AV_j\}}{AV_j}, & C_j \in \mathcal{C}^{\mathrm{ben}}, \\[2ex] \dfrac{\max\{0, \; AV_j - y_{ij}\}}{AV_j}, & C_j \in \mathcal{C}^{\mathrm{cost}}, \end{cases} \qquad NDA_{ij} := \begin{cases} \dfrac{\max\{0, \; AV_j - y_{ij}\}}{AV_j}, & C_j \in \mathcal{C}^{\mathrm{ben}}, \\[2ex] \dfrac{\max\{0, \; y_{ij} - AV_j\}}{AV_j}, & C_j \in \mathcal{C}^{\mathrm{cost}}. \end{cases}$$

Thus $PDA_{ij} \geq 0$ and $NDA_{ij} \geq 0$.

**Step 3 (Weighted sums).** For each alternative $A_i$, define

$$SP_i := \sum_{j=1}^{n} w_j \, PDA_{ij} \; \geq 0, \qquad SN_i := \sum_{j=1}^{n} w_j \, NDA_{ij} \; \geq 0.$$

**Step 4 (Normalization).** Let

$$SP^{\mathrm{max}} := \max_{1 \leq i \leq m} SP_i, \qquad SN^{\mathrm{max}} := \max_{1 \leq i \leq m} SN_i.$$

Define normalized values

$$NSP_i := \begin{cases} \dfrac{SP_i}{SP^{\max}}, & SP^{\max} > 0, \\ 0, & SP^{\max} = 0, \end{cases} \qquad NSN_i := \begin{cases} 1 - \dfrac{SN_i}{SN^{\max}}, & SN^{\max} > 0, \\ 1, & SN^{\max} = 0. \end{cases}$$

**Step 5 (Appraisal score and ranking).** Define the appraisal score

$$AS_i := \frac{NSP_i + NSN_i}{2} \in [0, 1],$$

and rank alternatives in descending order of $AS_i$ (ties allowed).

**Theorem 7.4.3** (Well-definedness and boundedness of U-EDAS). *Under Definition 7.4.2 (in particular, $m, n \geq 2$ and $S_M : \mathrm{Dom}(M) \to (0, \infty)$), all quantities in U-EDAS are well-defined and finite. Moreover,*

$$0 \leq PDA_{ij}, NDA_{ij} < \infty, \qquad 0 \leq SP_i, SN_i < \infty, \qquad 0 \leq NSP_i, NSN_i, AS_i \leq 1.$$

*Hence the ranking rule induced by $AS_i$ is well-defined.*

*Proof.* Since $S_M$ maps into $(0, \infty)$, each $y_{ij} > 0$ is finite, and thus $AV_j = \frac{1}{m} \sum_i y_{ij} > 0$ is finite. Therefore the ratios in the definitions of $PDA_{ij}$ and $NDA_{ij}$ have strictly positive denominators and finite numerators, so $PDA_{ij}$ and $NDA_{ij}$ are finite and nonnegative.

Because $w_j \geq 0$ and $\sum_j w_j = 1$, $SP_i$ and $SN_i$ are finite nonnegative weighted sums, so $SP^{\max}$ and $SN^{\max}$ exist (finite maxima over a finite set) and satisfy $SP^{\max} \geq 0$, $SN^{\max} \geq 0$. If $SP^{\max} > 0$, then $NSP_i = SP_i/SP^{\max} \in [0, 1]$; if $SP^{\max} = 0$, the definition sets $NSP_i = 0$. Similarly, if $SN^{\max} > 0$, then $SN_i/SN^{\max} \in [0, 1]$ so $NSN_i = 1 - SN_i/SN^{\max} \in [0, 1]$; if $SN^{\max} = 0$, the definition sets $NSN_i = 1 \in [0, 1]$. Finally, $AS_i$ is the average of two numbers in $[0, 1]$, hence $AS_i \in [0, 1]$. Thus sorting alternatives by $AS_i$ yields a well-defined preorder. $\qquad \square$

Related concepts of EDAS under uncertainty-aware models are listed in Table 7.5.

Table 7.5: Related concepts of EDAS under uncertainty-aware models.

| $k$ | Related EDAS concept(s) |
|---|---|
| 2 | Intuitionistic Fuzzy EDAS [70, 1006] |
| 3 | Hesitant Fuzzy EDAS [1007, 1008] |
| 3 | Spherical Fuzzy EDAS [1009, 1010] |
| 3 | Neutrosophic EDAS [1011, 1012] |

## 7.5 Uncertain VIKOR

VIKOR ranks alternatives by distances to ideal and anti-ideal points, computing group utility and individual regret, then a compromise index balancing both [1013, 1014]. Fuzzy VIKOR ranks alternatives under uncertainty using fuzzy ideal and nadir solutions, aggregated utility and regret measures, then a compromise index balancing group benefit and individual regret [892, 1015, 1016].

**Definition 7.5.1** (TFN-based Fuzzy VIKOR). [892, 1015, 1016] Let $A = \{A_1, \ldots, A_J\}$ be alternatives and $C = \{C_1, \ldots, C_n\}$ criteria. Let $I_b$ and $I_c$ be the index sets of benefit and cost criteria, with $I_b \dot{\cup} I_c = \{1, \ldots, n\}$. Assume TFN performances $\tilde{f}_{ij} = (l_{ij}, m_{ij}, r_{ij})$ and TFN weights $\tilde{w}_i = (l_i^w, m_i^w, r_i^w)$.

**(0) Conventions (defuzzification and max/min).** Use the "2nd weighted mean" defuzzification

$$\mathrm{Crisp}(l, m, r) := \frac{l + 2m + r}{4},$$

and compare TFNs by $\tilde{x} \preceq \tilde{y} \iff \mathrm{Crisp}(\tilde{x}) \leq \mathrm{Crisp}(\tilde{y})$. All TFN additions/scalar divisions below are componentwise; division is only by positive scalars.

**(1) Fuzzy ideal and nadir values.** For each criterion $i$ define

$$\tilde{f}_i^{\star} := \begin{cases} \max_{1 \leq j \leq J} \tilde{f}_{ij}, & i \in I_b, \\ \min_{1 \leq j \leq J} \tilde{f}_{ij}, & i \in I_c, \end{cases} \qquad \tilde{f}_i^{\ominus} := \begin{cases} \min_{1 \leq j \leq J} \tilde{f}_{ij}, & i \in I_b, \\ \max_{1 \leq j \leq J} \tilde{f}_{ij}, & i \in I_c, \end{cases}$$

where $\max, \min$ are taken w.r.t. $\preceq$. Write $\tilde{f}_i^{\star} = (l_i^{\star}, m_i^{\star}, r_i^{\star})$ and $\tilde{f}_i^{\ominus} = (l_i^{\ominus}, m_i^{\ominus}, r_i^{\ominus})$.

**(2) Normalized fuzzy deviations.** Define the (crisp) ranges

$$\Delta_i := \begin{cases} r_i^{\star} - l_i^{\ominus}, & i \in I_b, \\ r_i^{\ominus} - l_i^{\star}, & i \in I_c, \end{cases} \qquad (\Delta_i > 0),$$

and for each alternative $A_j$,

$$\tilde{d}_{ij} := \begin{cases} (\tilde{f}_i^{\star} - \tilde{f}_{ij})/\Delta_i, & i \in I_b, \\ (\tilde{f}_{ij} - \tilde{f}_i^{\star})/\Delta_i, & i \in I_c. \end{cases}$$

**(3) Group utility and individual regret (fuzzy).** For each alternative $A_j$ define TFNs

$$\tilde{S}_j := \sum_{i=1}^{n} \tilde{w}_i \, \tilde{d}_{ij}, \qquad \tilde{R}_j := \max_{1 \leq i \leq n} \left( \tilde{w}_i \, \tilde{d}_{ij} \right),$$

where products/sums are componentwise and max is w.r.t. $\preceq$.

**(4) Compromise index (fuzzy).** Let

$$\tilde{S} := \min_{1 \leq j \leq J} \tilde{S}_j, \qquad \tilde{R} := \min_{1 \leq j \leq J} \tilde{R}_j,$$

and set

$$S^{\max} := \max_{1 \leq j \leq J} r(\tilde{S}_j), \qquad R^{\max} := \max_{1 \leq j \leq J} r(\tilde{R}_j),$$

where $r(l, m, r) := r$ and $l(l, m, r) := l$ denote right/left endpoints. Fix $v \in [0, 1]$. Define

$$\tilde{Q}_j := v \, \frac{\tilde{S}_j - \tilde{S}}{S^{\max} - l(\tilde{S})} + (1 - v) \, \frac{\tilde{R}_j - \tilde{R}}{R^{\max} - l(\tilde{R})}.$$

**(5) Ranking and compromise solution (crisp).** Defuzzify

$$S_j := \mathrm{Crisp}(\tilde{S}_j), \qquad R_j := \mathrm{Crisp}(\tilde{R}_j), \qquad Q_j := \mathrm{Crisp}(\tilde{Q}_j),$$

and rank alternatives increasingly by $Q_j$. Let $A_{(1)}, A_{(2)}$ be the first two in the $Q$-ranking and $A_{(J)}$ the last. Set

$$DQ := \frac{1}{J-1}, \qquad \mathrm{Adv} := \frac{Q(A_{(2)}) - Q(A_{(1)})}{Q(A_{(J)}) - Q(A_{(1)})}.$$

Then $A_{(1)}$ is the (single) compromise solution if

$$\mathrm{Adv} \geq DQ \quad \text{and} \quad A_{(1)} \text{ is also best by } S \text{ and/or by } R.$$

If the stability condition fails, propose $\{A_{(1)}, A_{(2)}\}$. If the advantage condition fails, propose $\{A_{(1)}, \dots, A_{(M)}\}$ where $M$ is the largest index with $Q(A_{(M)}) - Q(A_{(1)}) < DQ$.

Using Uncertain Sets, we define Uncertain VIKOR of type $M$ (U-VIKOR) as follows.

**Definition 7.5.2** (Uncertain VIKOR of type $M$ (U-VIKOR)). Let $\mathcal{A} = \{A_1, \dots, A_m\}$ be alternatives and $\mathcal{C} = \{C_1, \dots, C_n\}$ criteria, with $m, n \geq 2$. Partition criteria into benefit and cost sets:

$$\mathcal{C} = \mathcal{C}^{\mathrm{ben}} \,\dot{\cup}\, \mathcal{C}^{\mathrm{cost}}.$$

Fix an uncertain model $M$ with $\mathrm{Dom}(M) \neq \emptyset$ and an admissible score $S_M$.

Assume an *uncertain decision matrix*

$$X^{(M)} = \big(x_{ij}^{(M)}\big)_{m \times n}, \qquad x_{ij}^{(M)} \in \mathrm{Dom}(M),$$

and criterion weights $w = (w_1, \dots, w_n)$ with

$$w_j > 0, \qquad \sum_{j=1}^{n} w_j = 1.$$

Fix the VIKOR compromise parameter $v \in [0, 1]$.

**Step 0 (Crisp projection).** Define $y_{ij} := S_M(x_{ij}^{(M)}) \in \mathbb{R}$.

**Step 1 (Best and worst values per criterion).** For each criterion $C_j$, define the best and worst (projected) values

$$f_j^* := \begin{cases} \max_{1 \leq i \leq m} y_{ij}, & C_j \in \mathcal{C}^{\mathrm{ben}}, \\ \min_{1 \leq i \leq m} y_{ij}, & C_j \in \mathcal{C}^{\mathrm{cost}}, \end{cases} \qquad f_j^- := \begin{cases} \min_{1 \leq i \leq m} y_{ij}, & C_j \in \mathcal{C}^{\mathrm{ben}}, \\ \max_{1 \leq i \leq m} y_{ij}, & C_j \in \mathcal{C}^{\mathrm{cost}}. \end{cases}$$

Define the range $\Delta_j := f_j^* - f_j^- \geq 0$.

**Step 2 (Normalized regret per criterion).** For each alternative $A_i$ and criterion $C_j$, define

$$d_{ij} := \begin{cases} \dfrac{f_j^* - y_{ij}}{\Delta_j}, & C_j \in \mathcal{C}^{\mathrm{ben}} \text{ and } \Delta_j > 0, \\[2ex] \dfrac{y_{ij} - f_j^*}{\Delta_j}, & C_j \in \mathcal{C}^{\mathrm{cost}} \text{ and } \Delta_j > 0, \\[2ex] 0, & \Delta_j = 0, \end{cases}$$

so that $d_{ij} \in [0, 1]$ and smaller is better (less regret).

**Step 3 (Group utility and individual regret measures).** Define

$$S_i := \sum_{j=1}^{n} w_j\, d_{ij}, \qquad R_i := \max_{1 \le j \le n}\,(w_j\, d_{ij}).$$

**Step 4 (Compromise index).** Let

$$S^* := \min_{1 \le i \le m} S_i, \quad S^- := \max_{1 \le i \le m} S_i, \qquad R^* := \min_{1 \le i \le m} R_i, \quad R^- := \max_{1 \le i \le m} R_i.$$

Define the VIKOR compromise index

$$Q_i := \begin{cases} v\, \dfrac{S_i - S^*}{S^- - S^*} + (1 - v)\, \dfrac{R_i - R^*}{R^- - R^*}, & S^- > S^* \text{ and } R^- > R^*, \\[2ex] \dfrac{S_i - S^*}{S^- - S^*}, & S^- > S^* \text{ and } R^- = R^*, \\[2ex] \dfrac{R_i - R^*}{R^- - R^*}, & S^- = S^* \text{ and } R^- > R^*, \\[2ex] 0, & S^- = S^* \text{ and } R^- = R^*. \end{cases}$$

Rank alternatives in ascending order of $Q_i$ (smaller $Q_i$ is better).

**(Optional) VIKOR compromise solution set.** Let $A_{(1)}, A_{(2)}, \ldots, A_{(m)}$ be the $Q$-sorted alternatives and set

$$DQ := \frac{1}{m - 1}.$$

One may apply the standard VIKOR acceptable-advantage and acceptable-stability rules on $(S_i, R_i, Q_i)$ to output either a single compromise solution or a small compromise set.

**Theorem 7.5.3** (Well-definedness of U-VIKOR). *Under Definition 7.5.2, assume $m, n \ge 2$, $\mathrm{Dom}(M) \ne \emptyset$, $S_M$ is admissible, and $w_j > 0$ with $\sum_{j=1}^{n} w_j = 1$. Then:*

*(i) For all $i, j$, the quantities $f_j^*, f_j^-, \Delta_j,$ and $d_{ij}$ are well-defined, with $0 \le d_{ij} \le 1$.*

*(ii) For each $i$, $S_i$ and $R_i$ are well-defined finite real numbers satisfying $0 \le S_i \le 1$ and $0 \le R_i \le 1$.*

*(iii) The compromise index $Q_i$ is well-defined and satisfies $0 \le Q_i \le 1$ for all $i$.*

*(iv) The ranking induced by $\{Q_i\}$ is well-defined (ties allowed).*

*Proof.* (i) Since $S_M$ is admissible, each $y_{ij}$ is finite. Because the sets $\{y_{1j}, \ldots, y_{mj}\}$ are finite, $f_j^*$ and $f_j^-$ exist for each $j$, hence $\Delta_j = f_j^* - f_j^- \ge 0$ is well-defined. If $\Delta_j > 0$, then $d_{ij}$ is defined by a ratio of finite numbers; if $\Delta_j = 0$, $d_{ij} = 0$ by definition. In either case, one has $0 \le d_{ij} \le 1$.

(ii) Since $0 \le d_{ij} \le 1$ and $\sum_j w_j = 1$, $S_i = \sum_j w_j d_{ij}$ is a convex combination, hence $0 \le S_i \le 1$. Also, $0 \le w_j d_{ij} \le w_j$, so $R_i = \max_j(w_j d_{ij})$ exists and satisfies $0 \le R_i \le \max_j w_j \le 1$.

(iii) The extrema $S^*, S^-, R^*, R^-$ exist because $m$ is finite. Each case in the definition of $Q_i$ avoids division by zero by switching to a single normalized component or setting $Q_i = 0$ when both ranges vanish. Whenever a fraction is used, its numerator lies between 0 and the corresponding denominator, hence the fraction lies in $[0, 1]$. Therefore $Q_i \in [0, 1]$.

(iv) Since each $Q_i$ is a real number, sorting alternatives by $Q_i$ defines a well-defined preorder. □

Related concepts of VIKOR under uncertainty-aware models are listed in Table 7.6.

Table 7.6: Related concepts of VIKOR under uncertainty-aware models.

| $k$ | Related VIKOR concept(s) |
|---|---|
| 2 | Intuitionistic Fuzzy VIKOR [1017, 1018] |
| 2 | Pythagorean Fuzzy VIKOR [1019, 1020] |
| 2 | Fermatean Fuzzy VIKOR [1021] |
| 2 | Bipolar fuzzy VIKOR [1022, 1023] |
| 3 | Hesitant Fuzzy VIKOR [1024, 1025] |
| 3 | Spherical Fuzzy VIKOR [1026–1028] |
| 3 | Neutrosophic VIKOR [1029, 1030] |
| $n$ | Plithogenic VIKOR [1031] |

Beyond Uncertain VIKOR, several related variants are also known, such as Grey VIKOR [1032], Group VIKOR [1033, 1034], Z-VIKOR [1035, 1036], VIKORSORT [1037, 1038], Rough VIKOR [1039, 1040], TOPSIS–VIKOR [1041, 1042], and Soft VIKOR [1043].

## 7.6 Uncertain MABAC (Multi-attributive border approximation area comparison)

MABAC ranks alternatives by distances from a border approximation area across criteria, then sums deviations to obtain final scores [1044–1046]. Fuzzy MABAC uses fuzzy ratings and weights, computes a fuzzy border approximation area, and ranks alternatives by defuzzified summed deviations [1047–1049].

**Definition 7.6.1** (TFN-based Fuzzy MABAC). [1047–1049] Let $\mathcal{A} = \{A_1, \ldots, A_m\}$ be alternatives and $\mathcal{C} = \{C_1, \ldots, C_n\}$ criteria. Partition criteria into benefit and cost types

$$\mathcal{C} = \mathcal{C}^+ \dot{\cup} \mathcal{C}^-,$$

and fix a crisp weight vector $\mathbf{w} = (w_1, \ldots, w_n)$ with $w_j \ge 0$ and $\sum_{j=1}^n w_j = 1$. Let $\tilde{X} = (\tilde{x}_{ij}) \in (\mathsf{TFN})^{m \times n}$ be the TFN decision matrix, where $\tilde{x}_{ij} = (x_{ij}^\ell, x_{ij}^m, x_{ij}^r)$.

**(0) TFN arithmetic used.** All operations are componentwise unless stated otherwise:

$$\tilde{x} \oplus \tilde{y} = (x^\ell + y^\ell, \, x^m + y^m, \, x^r + y^r), \qquad \alpha \odot \tilde{x} = (\alpha x^\ell, \, \alpha x^m, \, \alpha x^r) \; (\alpha \ge 0),$$

$$\tilde{x} \otimes \tilde{y} = (x^\ell y^\ell, \, x^m y^m, \, x^r y^r) \; \text{(nonnegative TFNs)}, \qquad \tilde{x}^p = \big((x^\ell)^p, \, (x^m)^p, \, (x^r)^p\big) \; (p > 0),$$

and we use the common triangular subtraction approximation

$$\tilde{x} \ominus \tilde{y} = (x^\ell - y^r,\ x^m - y^m,\ x^r - y^\ell).$$

**(1) Normalization.** For each criterion $j$, define

$$x_j^+ := \max_{1 \le i \le m} x_{ij}^r, \qquad x_j^- := \min_{1 \le i \le m} x_{ij}^\ell,$$

and the normalized TFNs

$$\tilde{t}_{ij} := \begin{cases} \dfrac{\tilde{x}_{ij} \ominus x_j^-}{x_j^+ - x_j^-}, & C_j \in \mathcal{C}^+, \\[2ex] \dfrac{\tilde{x}_{ij} \ominus x_j^+}{x_j^- - x_j^+}, & C_j \in \mathcal{C}^-, \end{cases}$$

where division is by a positive scalar.

**(2) Weighting.** Define $\tilde{v}_{ij} := w_j \odot \tilde{t}_{ij}$ and collect $\tilde{V} = (\tilde{v}_{ij})$.

**(3) Border approximation area (BAA).** For each criterion $j$, define the border approximation TFN

$$\tilde{g}_j := \left( \prod_{i=1}^m \tilde{v}_{ij} \right)^{1/m}, \qquad \tilde{G} := (\tilde{g}_1, \dots, \tilde{g}_n).$$

**(4) Signed deviations from the border and overall score.** Set

$$\tilde{q}_{ij} := \tilde{v}_{ij} \ominus \tilde{g}_j, \qquad \tilde{S}_i := \bigoplus_{j=1}^n \tilde{q}_{ij}.$$

(Heuristically, $\tilde{q}_{ij} > 0$ indicates $A_i$ lies in the upper approximation area for criterion $j$, while $\tilde{q}_{ij} < 0$ indicates the lower area; a crisp sign test can be implemented after defuzzification.)

**(5) Defuzzification and ranking.** For $\tilde{S}_i = (s_i^\ell, s_i^m, s_i^r)$, define

$$\mathrm{Defuzz}(\tilde{S}_i) := \frac{s_i^\ell + s_i^m + s_i^r}{3}.$$

Rank alternatives by $\mathrm{Defuzz}(\tilde{S}_i)$ in descending order:

$$A_p \succeq A_q \quad \Longleftrightarrow \quad \mathrm{Defuzz}(\tilde{S}_p) \ge \mathrm{Defuzz}(\tilde{S}_q).$$

The definition of Uncertain MABAC of type $M$ (U-MABAC) is given below.

**Definition 7.6.2** (Uncertain MABAC of type $M$ (U-MABAC)). Let $\mathcal{A} = \{A_1, \ldots, A_m\}$ be alternatives and $\mathcal{C} = \{C_1, \ldots, C_n\}$ criteria with $m, n \geq 2$. Partition criteria into benefit and cost sets:

$$\mathcal{C} = \mathcal{C}^{\text{ben}} \mathbin{\dot{\cup}} \mathcal{C}^{\text{cost}}.$$

Fix an uncertain model $M$ with $\mathrm{Dom}(M) \neq \emptyset$ and an admissible score $S_M$.

Assume an *uncertain decision matrix*

$$X^{(M)} = \big(x_{ij}^{(M)}\big)_{m \times n}, \qquad x_{ij}^{(M)} \in \mathrm{Dom}(M) \quad (i = 1, \ldots, m; \; j = 1, \ldots, n),$$

and criterion weights $w = (w_1, \ldots, w_n)$ with $w_j \geq 0$ and $\sum_{j=1}^{n} w_j = 1$.

**Step 0 (Crisp projection).** Define a real matrix $Y = (y_{ij})$ by

$$y_{ij} := S_M\big(x_{ij}^{(M)}\big) \in \mathbb{R}.$$

**Step 1 (Linear normalization to $[0,1]$).** For each criterion $j$, define

$$y_j^{\min} := \min_{1 \leq i \leq m} y_{ij}, \qquad y_j^{\max} := \max_{1 \leq i \leq m} y_{ij}, \qquad \Delta_j := y_j^{\max} - y_j^{\min} \geq 0.$$

Define normalized values $t_{ij} \in [0,1]$ by

$$t_{ij} := \begin{cases} \dfrac{y_{ij} - y_j^{\min}}{\Delta_j}, & C_j \in \mathcal{C}^{\text{ben}} \text{ and } \Delta_j > 0, \\[2ex] \dfrac{y_j^{\max} - y_{ij}}{\Delta_j}, & C_j \in \mathcal{C}^{\text{cost}} \text{ and } \Delta_j > 0, \\[2ex] 0, & \Delta_j = 0. \end{cases}$$

(Thus larger $t_{ij}$ always indicates better performance.)

**Step 2 (Weighting).** Define the weighted normalized matrix $V = (v_{ij})$ by

$$v_{ij} := w_j \, t_{ij} \qquad (i = 1, \ldots, m; \; j = 1, \ldots, n).$$

**Step 3 (Border approximation area).** For each criterion $j$, define the border approximation value (geometric mean):

$$g_j := \left( \prod_{i=1}^{m} v_{ij} \right)^{1/m} \geq 0.$$

Let $G := (g_1, \ldots, g_n)$.

**Step 4 (Signed deviations and overall score).** Define deviations

$$q_{ij} := v_{ij} - g_j, \qquad S_i := \sum_{j=1}^{n} q_{ij}.$$

Rank alternatives by decreasing $S_i$.

**Theorem 7.6.3** (Well-definedness of U-MABAC). *Under Definition 7.6.2, assume $m, n \geq 2$, $\mathrm{Dom}(M) \neq \emptyset$, and $S_M$ is admissible. Then all quantities $t_{ij}, v_{ij}, g_j, q_{ij}, S_i$ are well-defined finite real numbers.*

*Proof.* Because $S_M$ is admissible, each $y_{ij}$ is finite; hence for each $j$ the extrema $y_j^{\min}, y_j^{\max}$ exist and are finite, so $\Delta_j \geq 0$ is well-defined. If $\Delta_j > 0$, the normalization formulas define $t_{ij} \in [0, 1]$; if $\Delta_j = 0$, the definition sets $t_{ij} = 0$, so $t_{ij}$ is well-defined in all cases.

Since $w_j \geq 0$ and $t_{ij} \in [0, 1]$, each $v_{ij} = w_j t_{ij}$ is finite and nonnegative. Therefore the product $\prod_{i=1}^m v_{ij}$ is well-defined and nonnegative, so the geometric mean $g_j = (\prod_{i=1}^m v_{ij})^{1/m}$ is well-defined and finite (and equals 0 if any factor is 0). Consequently, $q_{ij} = v_{ij} - g_j$ and $S_i = \sum_{j=1}^n q_{ij}$ are finite sums/differences of finite reals, hence well-defined. □

Related concepts of MABAC are listed in Table 7.7.

Table 7.7: Related concepts of MABAC under uncertainty-aware models.

| $k$ | **Related MABAC concept(s)** |
|---|---|
| 2 | Intuitionistic Fuzzy MABAC [1050, 1051] |
| 3 | Spherical Fuzzy MABAC [43, 1052] |
| 3 | Hesitant Fuzzy MABAC [1053, 1054] |
| 3 | Picture Fuzzy MABAC [1055, 1056] |
| 3 | Neutrosophic MABAC [1057, 1058] |

As a related concept, Rough MABAC [1059, 1060] is also known.

## 7.7 Fuzzy MAIRCA (Fuzzy Multi-attributive ideal-real comparative analysis)

MAIRCA ranks alternatives by comparing theoretical ideal distribution of criterion importance with real performances, aggregating deviations to score [1061, 1062]. Fuzzy MAIRCA represents performances and possibly weights as fuzzy numbers, computes fuzzy ideal–real gaps, defuzzifies aggregated deviations for ranking [1063, 1064].

**Definition 7.7.1** (Fuzzy MAIRCA (TFN-based ideal–real comparative analysis)). [1063, 1064] Let $\mathcal{A} = \{A_1, \ldots, A_m\}$ be a finite set of alternatives and $\mathcal{C} = \{C_1, \ldots, C_n\}$ a finite set of criteria. Let $J^+$ and $J^-$ be the index sets of benefit and cost criteria, with $J^+ \dot\cup J^- = \{1, \ldots, n\}$.

Let $\mathsf{TFN}_{\geq 0} := \{(l, m, u) \in \mathbb{R}^3_{\geq 0} : l \leq m \leq u\}$. Assume a triangular-fuzzy decision matrix

$$\widetilde{X} = (\tilde{x}_{ij}) \in (\mathsf{TFN}_{\geq 0})^{m \times n}, \qquad \tilde{x}_{ij} = (l_{ij}, m_{ij}, u_{ij}),$$

and a (crisp) criterion-weight vector $w = (w_1, \ldots, w_n)$ with $w_j \geq 0$ and $\sum_{j=1}^n w_j = 1$. Fix a probability vector $p = (p_1, \ldots, p_m)$ with $p_i \geq 0$ and $\sum_{i=1}^m p_i = 1$ (typically $p_i = 1/m$ for all $i$).

**(0) TFN arithmetic and defuzzification.** For TFNs $\tilde{a} = (a^\ell, a^m, a^u)$ and $\tilde{b} = (b^\ell, b^m, b^u)$ and $\alpha \geq 0$, define

$$\tilde{a} \oplus \tilde{b} := (a^\ell + b^\ell, \ a^m + b^m, \ a^u + b^u), \qquad \alpha \odot \tilde{a} := (\alpha a^\ell, \ \alpha a^m, \ \alpha a^u),$$

and the standard TFN subtraction approximation

$$\tilde{a} \ominus \tilde{b} := (a^\ell - b^u, \ a^m - b^m, \ a^u - b^\ell).$$

Fix a defuzzification functional Defuzz : $\mathsf{TFN}_{\geq 0} \to \mathbb{R}_{\geq 0}$ (e.g. centroid)

$$\mathrm{Defuzz}(l, m, u) := \frac{l + m + u}{3}.$$

**Step 1 (matrix of theoretical weights).** Define the theoretical-weight matrix $T^{(p)} = (\tilde{t}_{ij}^{(p)}) \in (\mathsf{TFN}_{\geq 0})^{m \times n}$ by

$$\tilde{t}_{ij}^{(p)} := (p_i w_j, \ p_i w_j, \ p_i w_j), \qquad i = 1, \ldots, m, \ j = 1, \ldots, n.$$

(This is the "theoretical share" of criterion $j$ allocated to alternative $i$.)

**Step 2 (criterion-wise reference bounds).** For each criterion $j$, set the crisp bounds from TFN endpoints

$$x_j^- := \min_{1 \leq i \leq m} l_{ij}, \qquad x_j^+ := \max_{1 \leq i \leq m} u_{ij}, \qquad \Delta_j := x_j^+ - x_j^- > 0.$$

**Step 3 (fuzzy normalization factors).** Define normalized TFNs $\tilde{r}_{ij} \in \mathsf{TFN}_{\geq 0}$ by

$$\tilde{r}_{ij} := \begin{cases} \left( \dfrac{l_{ij} - x_j^-}{\Delta_j}, \ \dfrac{m_{ij} - x_j^-}{\Delta_j}, \ \dfrac{u_{ij} - x_j^-}{\Delta_j} \right), & j \in J^+, \\[4mm] \left( \dfrac{x_j^+ - u_{ij}}{\Delta_j}, \ \dfrac{x_j^+ - m_{ij}}{\Delta_j}, \ \dfrac{x_j^+ - l_{ij}}{\Delta_j} \right), & j \in J^-. \end{cases}$$

(Thus $\tilde{r}_{ij}$ encodes the normalized "real performance" on $[0, 1]$; for cost criteria the order is reversed to keep $l \leq m \leq u$.)

**Step 4 (matrix of real weights).** Define the real-weight matrix $T^{(r)} = (\tilde{t}_{ij}^{(r)}) \in (\mathsf{TFN}_{\geq 0})^{m \times n}$ by

$$\tilde{t}_{ij}^{(r)} := (p_i w_j) \odot \tilde{r}_{ij}, \qquad i = 1, \ldots, m, \ j = 1, \ldots, n.$$

**Step 5 (gap matrix).** Define the fuzzy gap matrix $\widetilde{G} = (\tilde{g}_{ij})$ by

$$\tilde{g}_{ij} := \tilde{t}_{ij}^{(p)} \ominus \tilde{t}_{ij}^{(r)}, \qquad i = 1, \ldots, m, \ j = 1, \ldots, n.$$

**Step 6 (criterion function and ranking).** For each alternative $A_i$, define its fuzzy criterion function

$$\tilde{Q}_i := \bigoplus_{j=1}^n \tilde{g}_{ij} \in \mathsf{TFN}_{\geq 0}, \qquad Q_i := \mathrm{Defuzz}(\tilde{Q}_i) \in \mathbb{R}_{\geq 0}.$$

The *Fuzzy MAIRCA ranking* is the preorder on $\mathcal{A}$ given by

$$A_i \succeq_{\mathrm{FMAIRCA}} A_k \iff Q_i \leq Q_k,$$

i.e., *smaller* $Q_i$ means a smaller ideal–real gap and hence a better alternative. A best alternative is any

$$A^\star \in \arg\min_{A_i \in \mathcal{A}} Q_i.$$

The definition of Uncertain MAIRCA of type $M$ (U-MAIRCA) is given below.

**Definition 7.7.2** (Uncertain MAIRCA of type $M$ (U-MAIRCA)). Let $\mathcal{A} = \{A_1, \ldots, A_m\}$ be a finite set of alternatives and $\mathcal{C} = \{C_1, \ldots, C_n\}$ a finite set of criteria, with $m, n \geq 2$. Let $J^+$ and $J^-$ be the index sets of benefit and cost criteria, with $J^+ \dot\cup J^- = \{1, \ldots, n\}$.

Fix an uncertain model $M$ with $\mathrm{Dom}(M) \neq \emptyset$ and an admissible positive score $S_M$. Assume an *uncertain decision matrix*

$$X^{(M)} = \big(x_{ij}^{(M)}\big)_{m \times n}, \qquad x_{ij}^{(M)} \in \mathrm{Dom}(M) \quad (i = 1, \ldots, m; \ j = 1, \ldots, n).$$

Let $w = (w_1, \ldots, w_n)$ be criterion weights with $w_j \geq 0$ and $\sum_{j=1}^n w_j = 1$. Fix a probability vector $p = (p_1, \ldots, p_m)$ with $p_i \geq 0$ and $\sum_{i=1}^m p_i = 1$ (typically $p_i = 1/m$).

**Step 0 (Crisp projection).** Define the positive real matrix $Y = (y_{ij})$ by

$$y_{ij} := S_M\big(x_{ij}^{(M)}\big) \in (0, \infty).$$

**Step 1 (Theoretical weight distribution).** Define the theoretical distribution matrix $T^{(p)} = (t_{ij}^{(p)}) \in \mathbb{R}_{\geq 0}^{m \times n}$ by

$$t_{ij}^{(p)} := p_i\, w_j, \qquad i = 1, \ldots, m, \ j = 1, \ldots, n.$$

**Step 2 (Min–max normalization of real performances).** For each criterion $j$, define

$$y_j^{\min} := \min_{1 \leq i \leq m} y_{ij}, \qquad y_j^{\max} := \max_{1 \leq i \leq m} y_{ij}, \qquad \Delta_j := y_j^{\max} - y_j^{\min} \geq 0.$$

Define normalized values $r_{ij} \in [0, 1]$ by

$$r_{ij} := \begin{cases} \dfrac{y_{ij} - y_j^{\min}}{\Delta_j}, & j \in J^+ \text{ and } \Delta_j > 0, \\[2ex] \dfrac{y_j^{\max} - y_{ij}}{\Delta_j}, & j \in J^- \text{ and } \Delta_j > 0, \\[2ex] 0, & \Delta_j = 0. \end{cases}$$

(Thus larger $r_{ij}$ always means better; degenerate criteria with $\Delta_j = 0$ contribute 0.)

**Step 3 (Real distribution matrix).** Define the real distribution matrix $T^{(r)} = (t_{ij}^{(r)})$ by

$$t_{ij}^{(r)} := t_{ij}^{(p)}\, r_{ij} = p_i\, w_j\, r_{ij}.$$

**Step 4 (Gap matrix and criterion function).** Define the gap matrix $G = (g_{ij})$ by

$$g_{ij} := t_{ij}^{(p)} - t_{ij}^{(r)} = p_i\, w_j\, (1 - r_{ij}) \ \geq \ 0.$$

For each alternative $A_i$, define its criterion function (overall gap)

$$Q_i := \sum_{j=1}^{n} g_{ij} = \sum_{j=1}^{n} p_i\, w_j\, (1 - r_{ij}) \;\geq\; 0.$$

**Ranking rule:** rank alternatives by ascending $Q_i$ (smaller gap is better).

**Theorem 7.7.3** (Well-definedness of U-MAIRCA)**.** *Under Definition 7.7.2, assume $m, n \geq 2$, $\mathrm{Dom}(M) \neq \emptyset$, and $S_M : \mathrm{Dom}(M) \to (0, \infty)$ is admissible. Then:*

    *(i) All quantities $y_{ij}$, $t_{ij}^{(p)}$, $r_{ij}$, $t_{ij}^{(r)}$, $g_{ij}$, and $Q_i$ are well-defined and finite.*

    *(ii) For all $i, j$, one has $0 \leq r_{ij} \leq 1$, hence $g_{ij} \geq 0$ and $Q_i \geq 0$.*

    *(iii) The ranking induced by $\{Q_i\}$ is well-defined (ties allowed).*

*Proof.* (i) Admissibility of $S_M$ implies $y_{ij} > 0$ finite. Since $p_i, w_j$ are finite and nonnegative, $t_{ij}^{(p)} = p_i w_j$ is finite. For each $j$, $y_j^{\min}, y_j^{\max}$ exist (finite index set) and $\Delta_j \geq 0$ is finite. If $\Delta_j > 0$, the min–max formulas define $r_{ij}$; if $\Delta_j = 0$, the definition sets $r_{ij} = 0$. Thus $r_{ij}$ is well-defined in all cases and finite. Then $t_{ij}^{(r)} = t_{ij}^{(p)} r_{ij}$ and $g_{ij} = t_{ij}^{(p)} - t_{ij}^{(r)}$ are finite, hence $Q_i = \sum_j g_{ij}$ is finite.

(ii) When $\Delta_j > 0$, min–max normalization yields $r_{ij} \in [0, 1]$; when $\Delta_j = 0$, $r_{ij} = 0 \in [0, 1]$. Therefore $1 - r_{ij} \in [0, 1]$, so $g_{ij} = p_i w_j (1 - r_{ij}) \geq 0$ and $Q_i \geq 0$.

(iii) Each $Q_i$ is a real number, so sorting alternatives by $Q_i$ defines a well-defined preorder. $\square$

Related concepts of MAIRCA under uncertainty-aware models are listed in Table 7.8.

Table 7.8: Related concepts of MAIRCA under uncertainty-aware models.

| $k$ | **Related MAIRCA concept(s)** |
|---|---|
| 1 | Fuzzy MAIRCA |
| 2 | Intuitionistic Fuzzy MAIRCA [1065, 1066] |
| 3 | Neutrosophic MAIRCA |

# Chapter 8

# Outranking (non-compensatory / dominance relations) Decision Methods

Outranking methods compare alternatives pairwise using concordance/discordance with thresholds and veto, yielding non-compensatory dominance relations where strong drawbacks cannot be offset by advantages. For convenience, a concise comparison of the representative outranking methods discussed in this chapter is presented in Table 8.1.

Table 8.1: A concise comparison of representative outranking decision methods.

| Method | Core comparison basis | Main mechanism | Primary output | Decision form |
|---|---|---|---|---|
| ELEC-TRE | Pairwise comparison of alternatives over concordant and discordant criteria | Builds concordance and discordance indices, applies thresholds (and possibly veto logic), and derives an outranking relation. | Outranking relation and kernel | Partial dominance / choice set |
| FlowSort | Comparison of each alternative with ordered limiting reference profiles | Computes preference flows against boundary profiles and assigns the alternative to the category whose profile interval contains its flow. | Category assignment | Sorting / ordered classification |
| PROMETHEE II | Pairwise criterionwise preference degrees between alternatives | Aggregates preference functions with criterion weights and computes positive, negative, and net outranking flows. | Net flow | Complete ranking |
| QUAL-IFLEX | Pairwise concordance of alternatives under each possible permutation | Evaluates all admissible permutations, aggregates weighted concordance/discordance over pairs, and selects the permutation with maximal comprehensive concordance. | Best permutation score | Complete ranking |

*Table 8.1 (continued).*

| Method | Core comparison basis | Main mechanism | Primary output | Decision form |
|--------|----------------------|----------------|----------------|---------------|
| ORESTE | Ordinal positions and distance-based preference intensities | Uses ordinal ranks of performances and criterion importance, computes distance-based scores and preference intensities, and derives weak and PIR-type relations. | Distance score and PIR relations | Weak ranking / outranking structure |

**Note.** All of these methods belong to the outranking family, but they differ in their decision logic. ELECTRE emphasizes non-compensatory dominance through concordance/discordance thresholds, FlowSort is primarily a sorting procedure based on reference profiles, PROMETHEE II yields a complete ranking through net preference flows, QUALIFLEX selects the best global permutation of alternatives, and ORESTE is especially suited to ordinal-information settings with weak preference, indifference, and incomparability-type structures.

## 8.1 Fuzzy ELECTRE (Fuzzy Elimination and choice translating reality)

ELECTRE is an outranking MCDM method that builds concordance/discordance indices, applies thresholds and veto, constructs an outranking relation, and eliminates dominated alternatives [1067, 1068]. Fuzzy ELECTRE is an outranking MCDM method using fuzzy evaluations to compute concordance/discordance and identify alternatives that dominate others under uncertainty [1069, 1070].

**Definition 8.1.1** (TFN-based Fuzzy ELECTRE). [1071, 1072] Let $A = \{A_1, \ldots, A_m\}$ be alternatives and $C = \{C_1, \ldots, C_n\}$ criteria. Let $B$ and $K$ be the index sets of benefit and cost criteria, with $B \,\dot\cup\, K = \{1, \ldots, n\}$. Assume a crisp decision matrix $X = (x_{ij}) \in \mathbb{R}_{>0}^{m \times n}$ and TFN criterion weights $\tilde{w}_j = (l_j, m_j, u_j)$.

**(0) Conventions.** For TFNs, use componentwise addition and scalar multiplication:

$$(l, m, u) \oplus (l', m', u') = (l + l', \ m + m', \ u + u'), \qquad \alpha \odot (l, m, u) = (\alpha l, \ \alpha m, \ \alpha u) \ (\alpha \geq 0).$$

**(1) Aggregate and normalize fuzzy weights (three components).** If $K$ decision makers provide numeric importance scores $y_{jk}$ for criterion $j$, set

$$\tilde{w}_j = (l_j, m_j, u_j) := \Big( \min_k y_{jk}, \ \frac{1}{K} \sum_{k=1}^{K} y_{jk}, \ \max_k y_{jk} \Big).$$

Normalize each component reciprocally:

$$w_{j1} := \frac{(1/l_j)}{\sum_{t=1}^{n}(1/l_t)}, \quad w_{j2} := \frac{(1/m_j)}{\sum_{t=1}^{n}(1/m_t)}, \quad w_{j3} := \frac{(1/u_j)}{\sum_{t=1}^{n}(1/u_t)}.$$

Let $\mathbf{w}^{(z)} = (w_{1z}, \ldots, w_{nz})$ for $z = 1, 2, 3$.

**(2) Normalize performances and build three weighted matrices.** Define the normalized matrix $R = (r_{ij})$ by

$$r_{ij} := \begin{cases} \dfrac{x_{ij}}{\sqrt{\sum_{p=1}^{m} x_{pj}^2}}, & j \in B, \\[3mm] \dfrac{(1/x_{ij})}{\sqrt{\sum_{p=1}^{m}(1/x_{pj})^2}}, & j \in K, \end{cases}$$

and for $z \in \{1, 2, 3\}$ define

$$v_{ij}^{(z)} := r_{ij}\, w_{jz}, \qquad V^{(z)} = (v_{ij}^{(z)})_{m \times n}.$$

**(3) Concordance/discordance sets and indices.** For $p \neq q$ and $z \in \{1, 2, 3\}$ define

$$C^{(z)}(p,q) := \{j : \ v_{pj}^{(z)} \geq v_{qj}^{(z)}\}, \qquad D^{(z)}(p,q) := \{1, \dots, n\} \setminus C^{(z)}(p,q),$$

$$C_{pq}^{(z)} := \sum_{j \in C^{(z)}(p,q)} w_{jz}, \qquad D_{pq}^{(z)} := \frac{\sum_{j \in D^{(z)}(p,q)} |v_{pj}^{(z)} - v_{qj}^{(z)}|}{\sum_{j=1}^{n} |v_{pj}^{(z)} - v_{qj}^{(z)}|}.$$

Defuzzify by summing components:

$$C_{pq}^{*} := \sum_{z=1}^{3} C_{pq}^{(z)}, \qquad D_{pq}^{*} := \sum_{z=1}^{3} D_{pq}^{(z)}.$$

**(4) Thresholds and outranking.** Let

$$\bar{C} := \frac{1}{m(m-1)} \sum_{\substack{p,q=1 \\ p \neq q}}^{m} C_{pq}^{*}, \qquad \bar{D} := \frac{1}{m(m-1)} \sum_{\substack{p,q=1 \\ p \neq q}}^{m} D_{pq}^{*}.$$

Define the fuzzy ELECTRE outranking relation by

$$A_p \succcurlyeq_{\mathrm{FE}} A_q \quad \Longleftrightarrow \quad (C_{pq}^{*} \geq \bar{C}) \ \wedge \ (D_{pq}^{*} \leq \bar{D}), \qquad (p \neq q).$$

**(5) Dominance matrix and kernel (best set).** Define

$$e_{pq} := \mathbb{1}[\, C_{pq}^{*} \geq \bar{C}\,], \qquad f_{pq} := \mathbb{1}[\, D_{pq}^{*} \leq \bar{D}\,], \qquad T := E \circ F,$$

so that $t_{pq} = 1$ iff $A_p \succcurlyeq_{\mathrm{FE}} A_q$. The kernel is

$$\mathcal{K} := \{A_p \in A : \ \nexists\, q \neq p \text{ with } t_{qp} = 1\}.$$

Using an Uncertain Set as the underlying extension framework, we define Uncertain ELECTRE of type $M$ (U-ELECTRE) as follows.

**Definition 8.1.2** (Uncertain ELECTRE of type $M$ (U-ELECTRE))**.** Let $A = \{A_1, \ldots, A_m\}$ be alternatives and $C = \{C_1, \ldots, C_n\}$ criteria, with $m \geq 2$ and $n \geq 1$. Let $B$ and $K$ be index sets of benefit and cost criteria with $B \dot\cup K = \{1, \ldots, n\}$. Fix an uncertain model $M$ with $\mathrm{Dom}(M) \neq \emptyset$ and an admissible score $S_M$.

Assume an *uncertain decision matrix*

$$X^{(M)} = \big(x_{ij}^{(M)}\big)_{m \times n}, \qquad x_{ij}^{(M)} \in \mathrm{Dom}(M),$$

and criterion weights $w = (w_1, \ldots, w_n)$ with $w_j \geq 0$ and $\sum_{j=1}^{n} w_j = 1$. (If uncertain weights are provided, score and normalize them to obtain such $w_j$.)

**Step 0 (Crisp projection and performance orientation).** Define $y_{ij} := S_M(x_{ij}^{(M)}) \in \mathbb{R}$. Convert all criteria to a "larger is better" orientation by defining

$$z_{ij} := \begin{cases} y_{ij}, & j \in B \quad \text{(benefit)}, \\ -y_{ij}, & j \in K \quad \text{(cost)}. \end{cases}$$

Let $Z = (z_{ij})$.

**Step 1 (Normalization).** For each criterion $j$, define

$$d_j := \sqrt{\sum_{p=1}^{m} z_{pj}^2} \ \geq \ 0, \qquad r_{ij} := \begin{cases} \dfrac{z_{ij}}{d_j}, & d_j > 0, \\ 0, & d_j = 0, \end{cases}$$

and the weighted normalized matrix

$$v_{ij} := w_j \, r_{ij}.$$

**Step 2 (Concordance and discordance sets).** For each ordered pair $(p, q)$ with $p \neq q$, define

$$\mathcal{C}(p, q) := \{\, j : \ v_{pj} \geq v_{qj} \,\}, \qquad \mathcal{D}(p, q) := \{1, \ldots, n\} \setminus \mathcal{C}(p, q).$$

**Step 3 (Concordance index).** Define the concordance index

$$C_{pq} := \sum_{j \in \mathcal{C}(p,q)} w_j \ \in [0, 1].$$

**Step 4 (Discordance index).** Define the discordance index by

$$D_{pq} := \begin{cases} \dfrac{\max\limits_{j \in \mathcal{D}(p,q)} |v_{pj} - v_{qj}|}{\max\limits_{1 \leq j \leq n} |v_{pj} - v_{qj}|}, & \max\limits_{1 \leq j \leq n} |v_{pj} - v_{qj}| > 0, \\ 0, & \max\limits_{1 \leq j \leq n} |v_{pj} - v_{qj}| = 0, \end{cases} \qquad (p \neq q).$$

(Thus $D_{pq} \in [0, 1]$.)

**Step 5 (Thresholds and outranking relation).** Define average thresholds

$$\bar{C} := \frac{1}{m(m-1)} \sum_{\substack{p,q=1 \\ p \neq q}}^{m} C_{pq}, \qquad \bar{D} := \frac{1}{m(m-1)} \sum_{\substack{p,q=1 \\ p \neq q}}^{m} D_{pq}.$$

Define the outranking relation $\succcurlyeq_{\mathrm{UE}}$ by

$$A_p \succcurlyeq_{\mathrm{UE}} A_q \quad \iff \quad \left(C_{pq} \geq \bar{C}\right) \wedge \left(D_{pq} \leq \bar{D}\right), \qquad (p \neq q).$$

**Step 6 (Kernel / best set).** Let $t_{pq} := \mathbf{1}[A_p \succcurlyeq_{\mathrm{UE}} A_q]$ and define the kernel

$$\mathcal{K} := \{A_p \in A : \ \nexists \, q \neq p \text{ with } t_{qp} = 1\}.$$

**Theorem 8.1.3** (Well-definedness of U-ELECTRE). *Under Definition 8.1.2, assume $m \geq 2$, $\mathrm{Dom}(M) \neq \emptyset$, and $S_M$ is admissible. Then all quantities $z_{ij}, d_j, r_{ij}, v_{ij}$, the sets $\mathcal{C}(p,q), \mathcal{D}(p,q)$, the indices $C_{pq}, D_{pq}$, the thresholds $\bar{C}, \bar{D}$, and the kernel $\mathcal{K}$ are well-defined. Moreover,*

$$0 \leq C_{pq} \leq 1, \qquad 0 \leq D_{pq} \leq 1.$$

*Proof.* Since $S_M$ is admissible, each $y_{ij}$ is finite, hence $z_{ij}$ is finite. Thus each $d_j = \sqrt{\sum_{p=1}^{m} z_{pj}^2}$ is well-defined and finite, and $r_{ij}$ is well-defined by case distinction ($d_j > 0$ yields a valid division; $d_j = 0$ sets $r_{ij} = 0$). Then $v_{ij} = w_j r_{ij}$ is well-defined.

For each pair $(p, q)$, the set $\mathcal{C}(p, q)$ is determined by finitely many comparisons $v_{pj} \geq v_{qj}$, hence is well-defined, and so is $\mathcal{D}(p, q)$. Because $w_j \geq 0$ and $\sum_j w_j = 1$, the concordance index $C_{pq} = \sum_{j \in \mathcal{C}(p,q)} w_j$ is well-defined and lies in $[0, 1]$.

For discordance, the quantities $|v_{pj} - v_{qj}|$ are finite. If their maximum over $j$ is positive, then the ratio defining $D_{pq}$ is valid and lies in $[0, 1]$ because the numerator is a maximum over a subset of indices. If the maximum is 0, then $v_{pj} = v_{qj}$ for all $j$ and the definition sets $D_{pq} = 0 \in [0, 1]$. Hence $D_{pq}$ is always well-defined and in $[0, 1]$.

The thresholds $\bar{C}$ and $\bar{D}$ are averages over finitely many well-defined indices, hence well-defined. Therefore the outranking relation is well-defined, as is the kernel $\mathcal{K}$ defined by a finite quantification over $q \neq p$. $\square$

Related concepts of ELECTRE under uncertainty-aware models are listed in Table 8.2.

In addition to Uncertain ELECTRE, several other extensions are also known, such as ELECTRE TRI [1084,1085], Rough ELECTRE [1086,1087], Extended ELECTRE [1088,1089], ELECTRE IS [1090], Group-ELECTRE [1091,1092], ELECTRE TRI-nC [1093,1094], Grey ELECTRE [1095], and Soft ELECTRE [1096,1097].

Table 8.2: Related concepts of ELECTRE under uncertainty-aware models.

| $k$ | Related ELECTRE concept(s) |
|---|---|
| 2 | Intuitionistic Fuzzy ELECTRE [1073, 1074] |
| 2 | Pythagorean Fuzzy ELECTRE [1075] |
| 2 | Fermatean Fuzzy ELECTRE [1076, 1077] |
| 3 | Neutrosophic ELECTRE [1078–1080] |
| 3 | Spherical Fuzzy ELECTRE [1081] |
| 3 | Picture Fuzzy ELECTRE [1082] |
| 3 | Hesitant Fuzzy ELECTRE [1083] |

## 8.2 Fuzzy FlowSort

FlowSort classifies alternatives into ordered categories using PROMETHEE outranking flows versus reference profiles, applying assignment rules by flows [1098, 1099]. Fuzzy FlowSort extends FlowSort to fuzzy evaluations and weights, computing fuzzy outranking degrees, defuzzified flows, then assigning categories [1100, 1101].

**Definition 8.2.1** (Fuzzy FlowSort (F-FlowSort) with limiting profiles). [1100, 1101] Let $\mathcal{A} = \{a_1, \ldots, a_m\}$ be a finite set of alternatives and $\mathcal{G} = \{g_1, \ldots, g_n\}$ a finite set of criteria. Assume $K \geq 2$ ordered categories

$$\mathcal{C} = \{C_1 \succ C_2 \succ \cdots \succ C_K\},$$

described by $(K+1)$ *limiting reference profiles*

$$\mathcal{R} = \{r_1, \ldots, r_{K+1}\} \quad \text{(intended order: } r_1 \succ r_2 \succ \cdots \succ r_{K+1}),$$

where category $C_k$ is bounded (in performance) between $r_k$ (upper) and $r_{k+1}$ (lower).

**Fuzzy performance data.** For each criterion $g_j$ and each $x \in \mathcal{A} \cup \mathcal{R}$, the evaluation is a triangular fuzzy number (TFN)

$$\tilde{g}_j(x) = (\ell_j(x), m_j(x), u_j(x)), \qquad \ell_j(x) \leq m_j(x) \leq u_j(x),$$

and the criterion weight may be crisp ($w_j \in [0,1]$) or fuzzy ($\tilde{w}_j$), with the standard normalization condition $\sum_{j=1}^{n} w_j = 1$ (or an agreed defuzzified normalization if $\tilde{w}_j$ are used).

**Fuzzy preference modelling.** For each criterion $j$, fix a (PROMETHEE-type) unicriterion preference function

$$P_j : \mathbb{R} \to [0,1]$$

(e.g., usual, U-shape, V-shape, level, Gaussian), together with any thresholds (indifference, preference) needed by that type. For $x, y \in \mathcal{A} \cup \mathcal{R}$ define the *fuzzy deviation*

$$\widetilde{\Delta}_j(x,y) := \tilde{g}_j(x) \ominus \tilde{g}_j(y)$$

(using TFN subtraction), and define the *fuzzy unicriterion preference* $\widetilde{P}_j(x,y)$ by applying $P_j$ to $\widetilde{\Delta}_j(x,y)$ via the chosen TFN calculus (e.g., interval/vertex evaluation or LR-form).

**Aggregated (fuzzy) outranking degree.** For $x, y \in \mathcal{A} \cup \mathcal{R}$ define the aggregated fuzzy preference

$$\widetilde{\pi}(x,y) := \sum_{j=1}^{n} w_j \otimes \widetilde{P}_j(x,y),$$

where $\otimes$ is scalar–TFN multiplication (or TFN multiplication if $\tilde{w}_j$ are used).  Choose a defuzzification operator

$$\mathrm{Def}: \ \mathrm{TFN} \to \mathbb{R}$$

(e.g., Yager-type, centroid, mean of maxima) and set the *crisp* outranking degree

$$\pi_d(x,y) := \mathrm{Def}\big(\widetilde{\pi}(x,y)\big) \in [0,1].$$

**Local comparison set and PROMETHEE-like flows.**  For each alternative $a_i \in \mathcal{A}$ define the local set

$$\mathcal{R}_i^* := \{a_i\} \cup \mathcal{R}.$$

For any $x \in \mathcal{R}_i^*$ define its *positive*, *negative*, and *net* flows (within $\mathcal{R}_i^*$) by

$$\varphi_i^+(x) := \frac{1}{|\mathcal{R}_i^*| - 1} \sum_{\substack{y \in \mathcal{R}_i^* \\ y \neq x}} \pi_d(x,y), \qquad \varphi_i^-(x) := \frac{1}{|\mathcal{R}_i^*| - 1} \sum_{\substack{y \in \mathcal{R}_i^* \\ y \neq x}} \pi_d(y,x),$$

$$\varphi_i(x) := \varphi_i^+(x) - \varphi_i^-(x).$$

**Category assignment rules.**  For each $a_i \in \mathcal{A}$ and each $k \in \{1, \dots, K\}$ define:

(R$^+$)  (positive-flow rule)   $C_{\varphi^+}(a_i) = C_k$ if

$$\varphi_i^+(r_k) > \varphi_i^+(a_i) \ \geq \ \varphi_i^+(r_{k+1}).$$

(R$^-$)  (negative-flow rule)   $C_{\varphi^-}(a_i) = C_k$ if

$$\varphi_i^-(r_k) \ \leq \ \varphi_i^-(a_i) \ < \ \varphi_i^-(r_{k+1}).$$

(R)  (net-flow rule)   $C_{\varphi}(a_i) = C_k$ if

$$\varphi_i(r_k) > \varphi_i(a_i) \ \geq \ \varphi_i(r_{k+1}).$$

A *Fuzzy FlowSort classification* is any mapping $\sigma : \mathcal{A} \to \mathcal{C}$ obtained by choosing one of the rules (R$^+$), (R$^-$), (R), or by a deterministic fusion policy, e.g.

$$\sigma(a_i) = C_{k^*} \quad \text{where} \quad k^* = \mathrm{Mode}\big\{k^+, k^-, k\big\}$$

with $C_{\varphi^+}(a_i) = C_{k^+}$, $C_{\varphi^-}(a_i) = C_{k^-}$, $C_{\varphi}(a_i) = C_k$, and a fixed tie-breaking convention (e.g., choose the worst index among ties).

**Proposition 8.2.2** (Basic well-definedness).  *Assume $\pi_d(x,y) \in [0,1]$ for all relevant $(x,y)$ and let $N_i := |\mathcal{R}_i^*|$.  Then $\varphi_i^+(x), \varphi_i^-(x) \in [0,1]$ and $\varphi_i(x) \in [-1,1]$ for all $x \in \mathcal{R}_i^*$.  Moreover, if for a fixed rule (R$^+$) (resp. (R$^-$), (R)) the profile flows satisfy*

$$\varphi_i^+(r_1) > \varphi_i^+(r_2) > \cdots > \varphi_i^+(r_{K+1}) \quad \text{(resp. similarly for } \varphi_i^- \text{ or } \varphi_i\text{),}$$

*then the corresponding assignment produces a unique category for each $a_i$.*

*Proof.* Since $\pi_d(\cdot, \cdot) \in [0, 1]$, each flow is an average of numbers in $[0, 1]$, hence lies in $[0, 1]$. The net flow is a difference of two $[0, 1]$ numbers, hence in $[-1, 1]$. Strict monotonicity of profile flows yields a partition of the real line into disjoint half-open intervals, so exactly one $k$ satisfies the relevant inequality. □

Using an Uncertain Set as the underlying extension framework, we define Uncertain FlowSort of type $M$ (U-FlowSort) as follows.

**Definition 8.2.3** (Uncertain FlowSort of type $M$ (U-FlowSort) with limiting profiles)**.** Let $\mathcal{A} = \{a_1, \dots, a_m\}$ be a finite set of alternatives and $\mathcal{G} = \{g_1, \dots, g_n\}$ a finite set of criteria, with $m \geq 1$ and $n \geq 1$. Fix $K \geq 2$ ordered categories

$$\mathcal{C} = \{C_1 \succ C_2 \succ \cdots \succ C_K\},$$

described by $(K + 1)$ limiting reference profiles

$$\mathcal{R} = \{r_1, \dots, r_{K+1}\} \quad \text{(intended order: } r_1 \succ r_2 \succ \cdots \succ r_{K+1}\text{)},$$

where category $C_k$ is bounded by $r_k$ (upper) and $r_{k+1}$ (lower).

Fix an uncertain model $M$ with $\mathrm{Dom}(M) \neq \emptyset$ and an admissible score $S_M$. Assume uncertain evaluations

$$x_j^{(M)}(x) \in \mathrm{Dom}(M) \qquad \text{for all } x \in \mathcal{A} \cup \mathcal{R}, \ j = 1, \dots, n,$$

and crisp criterion weights $w = (w_1, \dots, w_n)$ with

$$w_j \geq 0, \qquad \sum_{j=1}^{n} w_j = 1.$$

**Step 1 (Crisp projection).** Define crisp performances

$$y_j(x) := S_M\big(x_j^{(M)}(x)\big) \in \mathbb{R} \qquad (x \in \mathcal{A} \cup \mathcal{R}, \ j = 1, \dots, n).$$

**Step 2 (Unicriterion preference functions).** For each criterion $j$, fix a PROMETHEE-type preference function

$$P_j : \mathbb{R} \to [0, 1],$$

possibly with thresholds (indifference/preference) implicit in its definition. For $x, y \in \mathcal{A} \cup \mathcal{R}$ define the unicriterion preference degree

$$\pi_j(x, y) := P_j\big(y_j(x) - y_j(y)\big) \in [0, 1].$$

**Step 3 (Aggregated preference index).** Define the aggregated outranking (preference) index

$$\pi(x, y) := \sum_{j=1}^{n} w_j \, \pi_j(x, y) \in [0, 1], \qquad x, y \in \mathcal{A} \cup \mathcal{R}.$$

**Step 4 (Local comparison set and flows).** For each alternative $a_i \in \mathcal{A}$ define the local set

$$\mathcal{S}_i := \{a_i\} \cup \mathcal{R}, \qquad N_i := |\mathcal{S}_i| = K + 2.$$

For any $x \in \mathcal{S}_i$ define the positive, negative, and net flows:

$$\varphi_i^+(x) := \frac{1}{N_i - 1} \sum_{\substack{y \in \mathcal{S}_i \\ y \neq x}} \pi(x, y), \qquad \varphi_i^-(x) := \frac{1}{N_i - 1} \sum_{\substack{y \in \mathcal{S}_i \\ y \neq x}} \pi(y, x),$$

$$\varphi_i(x) := \varphi_i^+(x) - \varphi_i^-(x).$$

**Step 5 (Assignment rules).** For each $a_i \in \mathcal{A}$, assign a category using one of the following deterministic rules:

*($R^+$) Positive-flow rule:* choose the unique $k \in \{1, \ldots, K\}$ such that

$$\varphi_i^+(r_k) > \varphi_i^+(a_i) \ \geq \ \varphi_i^+(r_{k+1}),$$

and set $\sigma(a_i) := C_k$.

*($R^-$) Negative-flow rule:* choose the unique $k \in \{1, \ldots, K\}$ such that

$$\varphi_i^-(r_k) \ \leq \ \varphi_i^-(a_i) \ < \ \varphi_i^-(r_{k+1}),$$

and set $\sigma(a_i) := C_k$.

*(R) Net-flow rule:* choose the unique $k \in \{1, \ldots, K\}$ such that

$$\varphi_i(r_k) > \varphi_i(a_i) \ \geq \ \varphi_i(r_{k+1}),$$

and set $\sigma(a_i) := C_k$.

If uniqueness fails due to ties, adopt any fixed tie-breaking rule (e.g. choose the worst category among eligible ones). The resulting mapping $\sigma : \mathcal{A} \to \mathcal{C}$ is called an *Uncertain FlowSort classification*.

**Theorem 8.2.4** (Well-definedness of U-FlowSort). *Under Definition 8.2.3, assume $S_M$ is admissible and each preference function $P_j$ maps $\mathbb{R}$ into $[0, 1]$. Then:*

*(i) The unicriterion indices $\pi_j(x, y)$ and aggregated index $\pi(x, y)$ are well-defined and lie in $[0, 1]$.*

*(ii) For each $i$ and each $x \in \mathcal{S}_i$, the flows satisfy*

$$0 \leq \varphi_i^+(x) \leq 1, \qquad 0 \leq \varphi_i^-(x) \leq 1, \qquad -1 \leq \varphi_i(x) \leq 1.$$

*(iii) The assignment mapping $\sigma : \mathcal{A} \to \mathcal{C}$ is well-defined once a deterministic rule ($R^+$), ($R^-$), or (R) and a tie-breaking convention are fixed.*

*(iv) If, for a chosen rule, the corresponding profile flows are strictly decreasing (e.g. $\varphi_i^+(r_1) > \cdots > \varphi_i^+(r_{K+1})$ for $(R^+)$), then each $a_i$ is assigned to a* unique *category without tie-breaking.*

*Proof.* (i) Since $S_M$ is admissible, each $y_j(x)$ is finite, hence each difference $y_j(x) - y_j(y)$ is finite. By assumption $P_j : \mathbb{R} \to [0,1]$, so $\pi_j(x,y)$ is well-defined in $[0,1]$. Because $w_j \geq 0$ and $\sum_j w_j = 1$, the aggregated index $\pi(x,y) = \sum_j w_j \pi_j(x,y)$ is a convex combination, hence lies in $[0,1]$ and is well-defined.

(ii) For fixed $i$ and $x \in S_i$, $\varphi_i^+(x)$ is an average of $N_i - 1$ numbers in $[0,1]$, hence lies in $[0,1]$. The same holds for $\varphi_i^-(x)$. Therefore $\varphi_i(x) = \varphi_i^+(x) - \varphi_i^-(x) \in [-1,1]$.

(iii) With flows well-defined, each assignment rule compares finitely many real numbers. A deterministic rule plus a fixed tie-break produces a unique category, hence $\sigma$ is well-defined.

(iv) Under strict decrease of the profile flows, the real line is partitioned into disjoint half-open intervals $\left[\varphi(r_{k+1}), \varphi(r_k)\right)$ (or the analogous form), so exactly one $k$ satisfies the defining inequalities. $\qquad\square$

Related concepts of FlowSort under uncertainty-aware models are listed in Table 8.3.

Table 8.3: Related concepts of FlowSort under uncertainty-aware models.

| $k$ | **Related FlowSort concept(s)** |
| --- | --- |
| 1 | Fuzzy FlowSort |
| 2 | Intuitionistic Fuzzy FlowSort |
| 3 | Neutrosophic FlowSort |

## 8.3 Uncertain PROMETHEE (Preference Ranking Organization METhod for Enrichment of Evaluations)

PROMETHEE ranks alternatives using preference functions on pairwise criterion differences, aggregates weighted preferences, and computes positive, negative, and net outranking flows [1102–1104]. Fuzzy PROMETHEE models ratings and/or weights as fuzzy numbers, defuzzifies or compares them, then applies PROMETHEE preference aggregation and flow calculations under uncertainty [1105–1107].

**Definition 8.3.1** (Fuzzy PROMETHEE II (TFN inputs + Yager defuzzification)). (cf. [1105–1107]) Let $A = \{a_1, \ldots, a_m\}$ be alternatives and let $C = \{1, \ldots, n\}$ be criteria. Assume each criterion $j \in C$ is either a *benefit* criterion ($\uparrow$) or a *cost* criterion ($\downarrow$). Let $\tilde{f}_j(a) \in \mathbb{R}^3$ be the (triangular) fuzzy performance of alternative $a \in A$ on criterion $j$, and let $\tilde{w}_j \in \mathbb{R}^3$ be the (triangular) fuzzy weight of criterion $j$. Fix preference functions (generalized criteria)

$$p_j : \mathbb{R} \to [0,1] \qquad (j = 1, \ldots, n),$$

which convert a performance deviation into a preference degree.

**(1) Yager defuzzification.** For a TFN $\tilde{x} = (l, m, u)$, define its Yager magnitude (center/centroid score) by

$$\mathrm{Yag}(\tilde{x}) := \frac{l + m + u}{3}.$$

Define defuzzified performances

$$g_j(a) := \begin{cases} \mathrm{Yag}(\tilde{f}_j(a)), & j \text{ is benefit } (\uparrow), \\ -\mathrm{Yag}(\tilde{f}_j(a)), & j \text{ is cost } (\downarrow), \end{cases}$$

and defuzzified weights $\hat{w}_j := \mathrm{Yag}(\tilde{w}_j)$. Normalize the weights (optional but standard) by

$$w_j := \frac{\hat{w}_j}{\sum_{t=1}^{n} \hat{w}_t} \qquad (j = 1, \ldots, n).$$

**(2) Pairwise preferences.** For $a, b \in A$, define the deviation on criterion $j$ by

$$d_j(a, b) := g_j(a) - g_j(b),$$

the unicriterion preference degree by

$$P_j(a, b) := p_j\big(d_j(a, b)\big) \in [0, 1],$$

and the aggregated (global) preference index by

$$\pi(a, b) := \sum_{j=1}^{n} w_j \, P_j(a, b) \in [0, 1].$$

**(3) Outranking flows and complete ranking (PROMETHEE II).** Define the positive, negative, and net flows for each $a \in A$ by

$$\varphi^+(a) := \frac{1}{m-1} \sum_{\substack{b \in A \\ b \neq a}} \pi(a, b), \qquad \varphi^-(a) := \frac{1}{m-1} \sum_{\substack{b \in A \\ b \neq a}} \pi(b, a), \qquad \varphi(a) := \varphi^+(a) - \varphi^-(a).$$

The *Fuzzy PROMETHEE II* ranking is the complete preorder on $A$ induced by $\varphi$:

$$a \succeq b \iff \varphi(a) \geq \varphi(b).$$

Using an Uncertain Set as the underlying extension framework, we define Uncertain PROMETHEE II of type $M$ (U-PROMETHEE) as follows.

**Definition 8.3.2** (Uncertain PROMETHEE II of type $M$ (U-PROMETHEE)). Let $A = \{a_1, \ldots, a_m\}$ be a finite set of alternatives with $m \geq 2$, and let $C = \{1, \ldots, n\}$ be criteria. Partition criteria into benefit and cost sets $B, K \subseteq C$ with $B \dot{\cup} K = C$. Fix an uncertain model $M$ with $\mathrm{Dom}(M) \neq \emptyset$ and an admissible score $S_M$.

Assume uncertain performances $x_j^{(M)}(a) \in \mathrm{Dom}(M)$ for each $a \in A$ and criterion $j \in C$. Let $w = (w_1, \ldots, w_n)$ be criterion weights with

$$w_j \geq 0, \qquad \sum_{j=1}^{n} w_j = 1.$$

Fix PROMETHEE preference functions (generalized criteria)

$$p_j : \mathbb{R} \to [0,1] \qquad (j = 1, \ldots, n),$$

which convert a criterion deviation into a preference degree.

**Step 1 (Crisp projection and orientation).** For each $a \in A$ and $j \in C$, define

$$g_j(a) := \begin{cases} S_M\big(x_j^{(M)}(a)\big), & j \in B \quad \text{(benefit)}, \\ -S_M\big(x_j^{(M)}(a)\big), & j \in K \quad \text{(cost)}, \end{cases}$$

so that larger $g_j(a)$ always means better.

**Step 2 (Pairwise unicriterion preferences).** For $a, b \in A$ and $j \in C$, define

$$d_j(a,b) := g_j(a) - g_j(b) \in \mathbb{R}, \qquad P_j(a,b) := p_j\big(d_j(a,b)\big) \in [0,1].$$

**Step 3 (Aggregated preference index).** Define the global preference index

$$\pi(a,b) := \sum_{j=1}^{n} w_j \, P_j(a,b) \in [0,1], \qquad a, b \in A.$$

**Step 4 (PROMETHEE II flows).** For each $a \in A$, define the positive, negative, and net flows

$$\varphi^+(a) := \frac{1}{m-1} \sum_{\substack{b \in A \\ b \neq a}} \pi(a,b), \qquad \varphi^-(a) := \frac{1}{m-1} \sum_{\substack{b \in A \\ b \neq a}} \pi(b,a), \qquad \varphi(a) := \varphi^+(a) - \varphi^-(a).$$

**Ranking rule (PROMETHEE II).** Define the complete preorder on $A$ by

$$a \succeq_{\text{UP}} b \quad \Longleftrightarrow \quad \varphi(a) \geq \varphi(b).$$

**Theorem 8.3.3** (Well-definedness and bounds of U-PROMETHEE). *Under Definition 8.3.2, assume $m \geq 2$, $\text{Dom}(M) \neq \emptyset$, $S_M$ is admissible, and each $p_j$ maps $\mathbb{R}$ into $[0,1]$. Then:*

*(i) For all $a, b \in A$, $\pi(a,b)$ is well-defined and lies in $[0,1]$.*

*(ii) For all $a \in A$, the flows satisfy*

$$0 \leq \varphi^+(a) \leq 1, \qquad 0 \leq \varphi^-(a) \leq 1, \qquad -1 \leq \varphi(a) \leq 1.$$

*(iii) The ranking relation $\succeq_{\text{UP}}$ is well-defined (ties allowed).*

*Proof.* (i) Admissibility of $S_M$ implies each $g_j(a)$ is finite, hence each deviation $d_j(a, b)$ is finite. Since $p_j : \mathbb{R} \to [0, 1]$, $P_j(a, b)$ is well-defined in $[0, 1]$. With $w_j \geq 0$ and $\sum_j w_j = 1$, $\pi(a, b) = \sum_j w_j P_j(a, b)$ is a convex combination, hence lies in $[0, 1]$.

(ii) For fixed $a$, $\varphi^+(a)$ is the average of $m-1$ numbers in $[0, 1]$, hence $\varphi^+(a) \in [0, 1]$. Similarly, $\varphi^-(a) \in [0, 1]$. Therefore $\varphi(a) = \varphi^+(a) - \varphi^-(a) \in [-1, 1]$.

(iii) Since each $\varphi(a)$ is a real number, sorting alternatives by $\varphi(a)$ defines a complete preorder (ties allowed), hence $\succeq_{\mathrm{UP}}$ is well-defined. $\qquad\square$

For reference, related concepts of PROMETHEE under uncertainty-aware models are listed in Table 8.4.

Table 8.4: Related concepts of PROMETHEE under uncertainty-aware models.

| $k$ | Related **PROMETHEE** concept(s) |
|---|---|
| 1 | Fuzzy PROMETHEE |
| 2 | Intuitionistic Fuzzy PROMETHEE [1106, 1108] |
| 2 | Bipolar fuzzy PROMETHEE [1109, 1110] |
| 2 | Pythagorean fuzzy PROMETHEE [1111, 1112] |
| 2 | Fermatean fuzzy PROMETHEE [1113, 1114] |
| 3 | Picture Fuzzy PROMETHEE [1115, 1116] |
| 3 | Hesitant Fuzzy PROMETHEE [503, 1117] |
| 3 | Spherical Fuzzy PROMETHEE [1118, 1119] |
| 3 | Neutrosophic PROMETHEE [1120, 1121] |

Linguistic PROMETHEE [1122, 1123], Soft PROMETHEE [1112, 1124], Grey PROMETHEE [1125, 1126], Group-PROMETHEE [1127, 1128], Extended PROMETHEE [1129, 1130], PROMETHEE-GAIA [1113, 1131], SMAA-PROMETHEE [1132, 1133], PROMETHEE-MD-2T [1134], and Rough PROMETHEE [1135, 1136] are also known as related variants.

## 8.4  Fuzzy QUALIFLEX

QUALIFLEX ranks alternatives by evaluating all permutations, aggregating concordance of pairwise criterion preferences, selecting best ordering [1137, 1138]. Fuzzy QUALIFLEX encodes preferences with fuzzy numbers or linguistic terms, aggregates fuzzy concordance across permutations, yielding robust ranking [1139, 1140].

**Definition 8.4.1** (Fuzzy QUALIFLEX (permutation-based outranking)). [1139,1140] Let $\mathcal{A} = \{A_1, \ldots, A_m\}$ be a finite set of alternatives and $\mathcal{C} = \{C_1, \ldots, C_n\}$ a finite set of criteria. Let $\mathbb{F}$ be a chosen family of fuzzy evaluations (e.g. triangular/trapezoidal fuzzy numbers, hesitant fuzzy numbers, etc.). Assume a fuzzy decision matrix

$$\widetilde{X} = (\widetilde{x}_{ij}) \in \mathbb{F}^{m \times n}, \qquad \widetilde{x}_{ij} \in \mathbb{F} \text{ is the evaluation of } A_i \text{ under } C_j.$$

Let $w = (w_1, \ldots, w_n) \in [0, 1]^n$ be criterion weights with $\sum_{j=1}^n w_j = 1$. For each criterion $C_j$, fix:

- an *ideal* fuzzy value $\widetilde{u}_j^+ \in \mathbb{F}$ (best attainable on $C_j$; for cost criteria, choose the minimum-type ideal);

- a real-valued *closeness/score functional*

$$r_j : \mathbb{F} \to \mathbb{R}, \qquad r_j(\widetilde{x}) := d_j(\widetilde{x}, \widetilde{u}_j^+),$$

where $d_j$ is a distance (or dissimilarity) on $\mathbb{F}$.

(Thus, smaller $r_j(\widetilde{x})$ means "closer to ideal" on criterion $C_j$.)

Let $\mathfrak{S}_m$ denote the set of all permutations of $\{1, \ldots, m\}$. Each $\pi \in \mathfrak{S}_m$ represents a candidate complete ranking

$$P_\pi = \big(A_{\pi(1)}, A_{\pi(2)}, \ldots, A_{\pi(m)}\big), \quad \text{where } A_{\pi(p)} \text{ is ranked not worse than } A_{\pi(q)} \text{ if } p < q.$$

For any ordered pair $(p, q)$ with $1 \le p < q \le m$, define the *criterion-wise concordance/discordance index* by

$$\varphi_j^\pi\big(A_{\pi(p)}, A_{\pi(q)}\big) := r_j(\widetilde{x}_{\pi(q)j}) - r_j(\widetilde{x}_{\pi(p)j}).$$

Hence:

$\varphi_j^\pi > 0 \Rightarrow$ concordance (the higher-ranked alternative is closer to ideal), $\quad \varphi_j^\pi = 0 \Rightarrow$ ex aequo, $\quad \varphi_j^\pi < 0 \Rightarrow$ discordan

Aggregate across criteria (weighted) for each pair:

$$\varphi^\pi\big(A_{\pi(p)}, A_{\pi(q)}\big) := \sum_{j=1}^n w_j \, \varphi_j^\pi\big(A_{\pi(p)}, A_{\pi(q)}\big).$$

Define the *comprehensive concordance/discordance index* of the permutation $P_\pi$ by

$$\Phi(\pi) := \sum_{1 \le p < q \le m} \varphi^\pi\big(A_{\pi(p)}, A_{\pi(q)}\big).$$

A *Fuzzy QUALIFLEX solution* is any permutation $\pi^\star \in \mathfrak{S}_m$ achieving

$$\pi^\star \in \arg\max_{\pi \in \mathfrak{S}_m} \Phi(\pi),$$

and the induced ranking

$$A_{\pi^\star(1)} \succeq A_{\pi^\star(2)} \succeq \cdots \succeq A_{\pi^\star(m)}.$$

Using an Uncertain Set, we define Uncertain QUALIFLEX of type $M$ (U-QUALIFLEX) as follows.

**Definition 8.4.2** (Uncertain QUALIFLEX of type $M$ (U-QUALIFLEX)). Let $\mathcal{A} = \{A_1, \ldots, A_m\}$ be a finite set of alternatives and $\mathcal{C} = \{C_1, \ldots, C_n\}$ a finite set of criteria, with $m \ge 2$ and $n \ge 1$. Partition criteria into benefit and cost sets:

$$\mathcal{C} = \mathcal{C}^{\text{ben}} \,\dot\cup\, \mathcal{C}^{\text{cost}}.$$

Fix an uncertain model $M$ with $\text{Dom}(M) \neq \emptyset$ and an admissible score $S_M$. Assume an *uncertain decision matrix*

$$X^{(M)} = \big(x_{ij}^{(M)}\big)_{m \times n}, \qquad x_{ij}^{(M)} \in \text{Dom}(M),$$

and criterion weights $w = (w_1, \ldots, w_n)$ with $w_j \ge 0$ and $\sum_{j=1}^n w_j = 1$.

**Step 0 (Crisp projection and orientation).** Define $y_{ij} := S_M(x_{ij}^{(M)}) \in \mathbb{R}$ and convert all criteria to a "larger is better" form:

$$z_{ij} := \begin{cases} y_{ij}, & C_j \in \mathcal{C}^{\mathrm{ben}}, \\ -y_{ij}, & C_j \in \mathcal{C}^{\mathrm{cost}}. \end{cases}$$

Let $Z = (z_{ij}) \in \mathbb{R}^{m \times n}$.

**Permutation set.** Let $\mathfrak{S}_m$ be the set of all permutations of $\{1, \ldots, m\}$. Each $\pi \in \mathfrak{S}_m$ encodes a candidate complete ranking

$$P_\pi = \big(A_{\pi(1)}, A_{\pi(2)}, \ldots, A_{\pi(m)}\big), \quad \text{where } p < q \;\Rightarrow\; A_{\pi(p)} \text{ is ranked not worse than } A_{\pi(q)}.$$

**Step 1 (Pairwise concordance contribution under a permutation).** For any permutation $\pi \in \mathfrak{S}_m$ and any ordered pair $(p, q)$ with $1 \leq p < q \leq m$, define the criterionwise concordance contribution

$$\varphi_j^\pi\big(A_{\pi(p)}, A_{\pi(q)}\big) := z_{\pi(p)j} - z_{\pi(q)j} \in \mathbb{R}, \qquad j = 1, \ldots, n.$$

Thus $\varphi_j^\pi > 0$ means the higher-ranked alternative is better on criterion $j$.

**Step 2 (Weighted pairwise concordance index).** Define the weighted pairwise concordance index

$$\varphi^\pi\big(A_{\pi(p)}, A_{\pi(q)}\big) := \sum_{j=1}^n w_j \, \varphi_j^\pi\big(A_{\pi(p)}, A_{\pi(q)}\big) \in \mathbb{R}.$$

**Step 3 (Comprehensive concordance index of a permutation).** Define the comprehensive concordance of permutation $\pi$ by

$$\Phi(\pi) := \sum_{1 \leq p < q \leq m} \varphi^\pi\big(A_{\pi(p)}, A_{\pi(q)}\big) \in \mathbb{R}.$$

**Step 4 (Optimal permutation and ranking).** Any maximizer

$$\pi^\star \in \arg\max_{\pi \in \mathfrak{S}_m} \; \Phi(\pi)$$

is called an *Uncertain QUALIFLEX solution*, and the induced ranking is

$$A_{\pi^\star(1)} \succeq_{\mathrm{UQF}} A_{\pi^\star(2)} \succeq_{\mathrm{UQF}} \cdots \succeq_{\mathrm{UQF}} A_{\pi^\star(m)}.$$

**Theorem 8.4.3** (Well-definedness of U-QUALIFLEX). *Under Definition 8.4.2, assume $m \geq 2$, $\mathrm{Dom}(M) \neq \emptyset$, and $S_M$ is admissible. Then:*

(i) *For every $\pi \in \mathfrak{S}_m$, the value $\Phi(\pi)$ is a well-defined finite real number.*

(ii) *The maximizer set $\arg\max_{\pi \in \mathfrak{S}_m} \Phi(\pi)$ is nonempty, hence an optimal permutation $\pi^\star$ exists.*

(iii) *The induced ranking by $\pi^\star$ is well-defined (ties correspond to multiple maximizers).*

*Proof.* (i) Admissibility of $S_M$ implies each $y_{ij}$ is finite; hence each oriented value $z_{ij}$ is finite. For any $\pi$ and any $(p,q)$, each $\varphi_j^\pi = z_{\pi(p)j} - z_{\pi(q)j}$ is a finite real. Since $w_j \geq 0$ and $\sum_j w_j = 1$, $\varphi^\pi = \sum_j w_j \varphi_j^\pi$ is finite, and $\Phi(\pi)$ is a finite sum over the $\binom{m}{2}$ pairs, hence finite.

(ii) The set $\mathfrak{S}_m$ is finite with $|\mathfrak{S}_m| = m!$. Therefore the finite set $\{\Phi(\pi) : \pi \in \mathfrak{S}_m\} \subset \mathbb{R}$ attains a maximum, so $\arg\max_{\pi \in \mathfrak{S}_m} \Phi(\pi) \neq \emptyset$.

(iii) Any maximizer $\pi^\star$ defines an ordering of the alternatives, hence a ranking relation. If multiple maximizers exist, multiple optimal rankings are possible; selecting any $\pi^\star$ yields a well-defined ranking. $\square$

As related concepts, QUALIFLEX variants under uncertainty-aware models are listed in Table 8.5.

Table 8.5: Related concepts of QUALIFLEX under uncertainty-aware models.

| $k$ | Related QUALIFLEX concept(s) |
|---|---|
| 2 | Intuitionistic Fuzzy QUALIFLEX [1141] |
| 3 | Hesitant Fuzzy QUALIFLEX [1139, 1140] |
| 3 | Neutrosophic QUALIFLEX [1142–1144] |

## 8.5 Fuzzy ORESTE (Fuzzy Organization Rangement Et Synthese De Donnees Relationnelles)

ORESTE derives rankings from ordinal information using preference intensities and distance-based aggregation, suitable when precise data unavailable [1145, 1146]. Fuzzy ORESTE represents ordinal judgments as fuzzy ranks, aggregates fuzzy preference distances, producing rankings under vague, imprecise assessments [76, 1147, 1148].

**Definition 8.5.1** (Fuzzy ORESTE (distance-based outranking)). [76, 1147, 1148] Let $\mathcal{A} = \{a_1, \ldots, a_m\}$ be a finite set of alternatives and $\mathcal{C} = \{c_1, \ldots, c_n\}$ a finite set of criteria, partitioned as $\mathcal{C} = \mathcal{C}^{\mathrm{ben}} \sqcup \mathcal{C}^{\mathrm{cost}}$ (benefit vs. cost).

Let $\mathbb{F}$ denote a chosen family of fuzzy numbers on $\mathbb{R}$ (e.g. triangular, trapezoidal, or general normal convex fuzzy numbers). A *fuzzy decision matrix* is

$$\widetilde{X} = (\widetilde{x}_{ij}) \in \mathbb{F}^{m \times n}, \qquad \widetilde{x}_{ij} \in \mathbb{F} \text{ encodes the (fuzzy) performance of } a_i \text{ on } c_j.$$

A *fuzzy importance profile* is

$$\widetilde{w} = (\widetilde{w}_1, \ldots, \widetilde{w}_n) \in \mathbb{F}^n, \qquad \widetilde{w}_j \in \mathbb{F} \text{ encodes the (fuzzy) importance of } c_j.$$

Assume:

(A1) A ranking functional $\mathrm{Rank} : \mathbb{F} \to \mathbb{R}$ that induces a total preorder on $\mathbb{F}$ (used to define max / min of fuzzy numbers by comparing Rank values).

(A2) A (normalized) distance $d_{\mathbb{F}} : \mathbb{F} \times \mathbb{F} \to [0, 1]$ (any metric/pseudometric suitable for $\mathbb{F}$, e.g. via $\alpha$-cuts).

**Step 1 (Ideal points).** Define the criterion-wise fuzzy ideal value $\widetilde{x}_j^+ \in \mathbb{F}$ by

$$\widetilde{x}_j^+ := \begin{cases} \arg \max\limits_{i \in \{1,\dots,m\}} \mathrm{Rank}(\widetilde{x}_{ij}), & c_j \in \mathcal{C}^{\mathrm{ben}}, \\ \arg \min\limits_{i \in \{1,\dots,m\}} \mathrm{Rank}(\widetilde{x}_{ij}), & c_j \in \mathcal{C}^{\mathrm{cost}}. \end{cases}$$

Also define the most important fuzzy weight

$$\widetilde{w}^+ := \arg \max_{j \in \{1,\dots,n\}} \mathrm{Rank}(\widetilde{w}_j).$$

**Step 2 (Distances / coordinates).** For each $(i,j)$ set

$$d_{ij} := d_{\mathbb{F}}(\widetilde{x}_{ij}, \widetilde{x}_j^+) \in [0,1], \qquad d_j := d_{\mathbb{F}}(\widetilde{w}_j, \widetilde{w}^+) \in [0,1].$$

Interpret $(d_{ij}, d_j)$ as the coordinate of the pair "$(a_i, c_j)$" relative to the ideal.

**Step 3 (Global preference score).** Fix a tradeoff parameter $\xi \in [0,1]$. The *global preference score* of $a_i$ w.r.t. $c_j$ is

$$D_{ij} := \left( \xi \, d_{ij}^2 + (1-\xi) \, d_j^2 \right)^{1/2} \in [0,1],$$

where *smaller $D_{ij}$ means closer to the ideal and thus better*. The *average preference score* of $a_i$ is

$$D_i := \frac{1}{n} \sum_{j=1}^{n} D_{ij} \in [0,1].$$

The *weak ranking* (preorder) is induced by $D_i$:

$$a_i \succ a_k \iff D_i < D_k, \qquad a_i \sim a_k \iff D_i = D_k.$$

**Step 4 (Preference intensities).** For paired comparison $(a_i, a_k)$ define the criterion-wise preference intensity

$$T_j(a_i, a_k) := \max\{D_{kj} - D_{ij}, \, 0\} \in [0,1],$$

the average preference intensity

$$T(a_i, a_k) := \frac{1}{n} \sum_{j=1}^{n} T_j(a_i, a_k) \in [0,1],$$

and the net preference intensity

$$\Delta T(a_i, a_k) := T(a_i, a_k) - T(a_k, a_i) \in [-1, 1].$$

**Step 5 (P/I/R test and strong ranking).** Fix thresholds $\mu \in (0,1]$ (preference threshold) and $\sigma \in (0,1]$ (indifference bound). Define the PIR relation between $a_i$ and $a_k$ by:

$$a_i \, I \, a_k \iff \big| \Delta T(a_i, a_k) \big| < \mu \text{ and } T(a_i, a_k) < \sigma \text{ and } T(a_k, a_i) < \sigma,$$

$$a_i \, R \, a_k \iff \big|\Delta T(a_i, a_k)\big| < \mu \text{ and } \max\{T(a_i, a_k), T(a_k, a_i)\} \geq \sigma,$$

$$a_i \, P \, a_k \iff \big|\Delta T(a_i, a_k)\big| \geq \mu \text{ and } \Delta T(a_i, a_k) > 0.$$

(The case $\big|\Delta T(a_i, a_k)\big| \geq \mu$ and $\Delta T(a_i, a_k) < 0$ yields $a_k P a_i$.)

A *Fuzzy ORESTE output* is the weak ranking induced by $(D_i)_{i=1}^m$ together with the PIR structure $(P, I, R)$ on $\mathcal{A} \times \mathcal{A}$. A *best* alternative (distance-to-ideal optimum) is any

$$a^\star \in \arg\min_{a_i \in \mathcal{A}} D_i.$$

Using an Uncertain Set, we define Uncertain ORESTE of type $M$ (U-ORESTE) as follows.

**Definition 8.5.2** (Uncertain ORESTE of type $M$ (U-ORESTE)). Let $\mathcal{A} = \{a_1, \ldots, a_m\}$ be a finite set of alternatives and $\mathcal{C} = \{c_1, \ldots, c_n\}$ a finite set of criteria, with $m, n \geq 2$. Partition criteria into benefit and cost sets:

$$\mathcal{C} = \mathcal{C}^{\text{ben}} \,\dot\cup\, \mathcal{C}^{\text{cost}}.$$

Fix an uncertain model $M$ with $\mathrm{Dom}(M) \neq \emptyset$ and an admissible score $S_M$.

Assume uncertain performances $x_{ij}^{(M)} \in \mathrm{Dom}(M)$ for each $(i, j)$ and uncertain criterion-importance degrees $w_j^{(M)} \in \mathrm{Dom}(M)$ for each $j$.

Fix an admissible score $S_M$ and define crisp performances and importance scores

$$y_{ij} := \begin{cases} S_M(x_{ij}^{(M)}), & c_j \in \mathcal{C}^{\text{ben}}, \\ -\, S_M(x_{ij}^{(M)}), & c_j \in \mathcal{C}^{\text{cost}}, \end{cases} \qquad \beta_j := S_M(w_j^{(M)}).$$

Thus larger $y_{ij}$ is better and larger $\beta_j$ means more important.

**Step 1 (Ordinal ranks from projected scores).** Define the performance rank of alternative $a_i$ under criterion $c_j$ by

$$r_{ij} := 1 + \big|\{t \in \{1, \ldots, m\} : \ y_{tj} > y_{ij}\}\big| \in \{1, \ldots, m\},$$

and define the importance rank of criterion $c_j$ by

$$s_j := 1 + \big|\{t \in \{1, \ldots, n\} : \ \beta_t > \beta_j\}\big| \in \{1, \ldots, n\}.$$

(Thus smaller $r_{ij}$ and smaller $s_j$ indicate higher desirability/importance; ties allowed.)

**Step 2 (Besson–rank distance / global preference coordinates).** Fix a tradeoff parameter $\xi \in [0, 1]$. Define the ORESTE global distance score

$$D_{ij} := \sqrt{\xi \left(\frac{r_{ij}}{m}\right)^2 + (1 - \xi) \left(\frac{s_j}{n}\right)^2} \in \left[0, \sqrt{\xi + (1 - \xi)}\right] = [0, 1],$$

and the average distance (global score) of alternative $a_i$:

$$D_i := \frac{1}{n} \sum_{j=1}^n D_{ij} \in [0, 1].$$

**Step 3 (Weak ranking).** Define the weak preorder on $\mathcal{A}$ by

$$a_i \succeq_{\mathrm{UO}} a_k \iff D_i \le D_k \quad \text{(smaller distance is better)}.$$

**Step 4 (Preference intensity and PIR structure).** For $a_i, a_k \in \mathcal{A}$ define the criterionwise preference intensity

$$T_j(a_i, a_k) := \max\{D_{kj} - D_{ij},\, 0\} \in [0, 1],$$

the average preference intensity

$$T(a_i, a_k) := \frac{1}{n} \sum_{j=1}^{n} T_j(a_i, a_k) \in [0, 1],$$

and the net intensity

$$\Delta T(a_i, a_k) := T(a_i, a_k) - T(a_k, a_i) \in [-1, 1].$$

Fix thresholds $\mu \in (0, 1]$ and $\sigma \in (0, 1]$ and define relations $P, I, R$ by

$$a_i \, I \, a_k \iff |\Delta T(a_i, a_k)| < \mu \text{ and } T(a_i, a_k) < \sigma \text{ and } T(a_k, a_i) < \sigma,$$

$$a_i \, R \, a_k \iff |\Delta T(a_i, a_k)| < \mu \text{ and } \max\{T(a_i, a_k), T(a_k, a_i)\} \ge \sigma,$$

$$a_i \, P \, a_k \iff |\Delta T(a_i, a_k)| \ge \mu \text{ and } \Delta T(a_i, a_k) > 0,$$

(with $a_k P a_i$ when $|\Delta T(a_i, a_k)| \ge \mu$ and $\Delta T(a_i, a_k) < 0$).

**Theorem 8.5.3** (Well-definedness of U-ORESTE)**.** *Under Definition 8.5.2, assume $m, n \ge 2$, $\mathrm{Dom}(M) \ne \emptyset$, and $S_M$ is admissible. Then:*

*(i) The ranks $r_{ij} \in \{1, \ldots, m\}$ and $s_j \in \{1, \ldots, n\}$ are well-defined (ties allowed).*

*(ii) $D_{ij}$ and $D_i$ are well-defined and satisfy $0 \le D_{ij} \le 1$ and $0 \le D_i \le 1$.*

*(iii) For all $a_i, a_k$, $T_j(a_i, a_k), T(a_i, a_k) \in [0, 1]$ and $\Delta T(a_i, a_k) \in [-1, 1]$ are well-defined.*

*(iv) The weak ranking $\succeq_{\mathrm{UO}}$ and the PIR relations $(P, I, R)$ are well-defined.*

*Proof.* (i) Since all $y_{ij}$ and $\beta_j$ are finite real numbers, the sets $\{t : y_{tj} > y_{ij}\}$ and $\{t : \beta_t > \beta_j\}$ are finite and have well-defined cardinalities. Therefore $r_{ij} = 1 + |\{t : y_{tj} > y_{ij}\}| \in \{1, \ldots, m\}$ and $s_j = 1 + |\{t : \beta_t > \beta_j\}| \in \{1, \ldots, n\}$.

(ii) Because $1 \le r_{ij} \le m$ and $1 \le s_j \le n$, one has $0 < r_{ij}/m \le 1$ and $0 < s_j/n \le 1$. With $\xi \in [0, 1]$, the expression under the square root in $D_{ij}$ lies in $[0, 1]$, hence $D_{ij} \in [0, 1]$. Then $D_i$ is an average of $n$ numbers in $[0, 1]$, so $D_i \in [0, 1]$.

(iii) Since $D_{ij} \in [0, 1]$, we have $D_{kj} - D_{ij} \in [-1, 1]$, hence $T_j(a_i, a_k) = \max\{D_{kj} - D_{ij}, 0\} \in [0, 1]$. Averaging yields $T(a_i, a_k) \in [0, 1]$. Therefore $\Delta T(a_i, a_k)$ is a difference of two numbers in $[0, 1]$, so it lies in $[-1, 1]$.

(iv) The weak ranking uses comparisons of real numbers $D_i$, hence is well-defined. The relations $P, I, R$ are defined by finite inequalities involving $\Delta T$ and $T$ and fixed thresholds $\mu, \sigma$, so they are well-defined. $\square$

As related concepts, ORESTE variants under uncertainty-aware models are listed in Table 8.6.

Table 8.6: Related concepts of ORESTE under uncertainty-aware models.

| $k$ | Related ORESTE concept(s) |
|---|---|
| 2 | Intuitionistic Fuzzy ORESTE [1149] |
| 3 | Hesitant Fuzzy ORESTE [1150] |
| 3 | Spherical Fuzzy ORESTE [1151] |
| 3 | Neutrosophic ORESTE [1152, 1153] |

# Chapter 9

# Rule induction / learning / evidence / sequential decision methods

This section describes decision-making methods from the perspectives of rule induction, learning, evidence aggregation, and sequential decision-making. For convenience, the main concepts introduced in this chapter are compared in Table 9.1.

Table 9.1: Concise comparison of representative rule-induction, learning, evidence-aggregation, and sequential-decision methods.

| Method | Type | Basic input structure | Core mechanism | Primary output | Decision role |
|---|---|---|---|---|---|
| Decision Tree | Fuzzy | Fuzzy-valued attributes, fuzzy partitions, or soft split functions | Recursively partitions the instance space by fuzzy tests and propagates memberships through branches | Predicted class, value, or leaf-based membership | Rule induction / classification |
| Decision Tree | Uncertain | Uncertain-valued features scored by an admissible map into $\mathbb{R}$ | Converts uncertain features to crisp scores, then performs hard or soft tree splitting on measurable tests. | Deterministic prediction or aggregated soft output | Rule induction / classification |
| DRSA | Fuzzy | Ordered decision classes and fuzzy dominance relations on criteria | Builds fuzzy lower and upper approximations of class unions through fuzzy dominance and fuzzy quantifiers. | Fuzzy approximations and graded decision rules | Preference rule induction |
| DRSA | Uncertain | Uncertain-valued decision table projected to crisp criterion scores | Defines crisp dominance cones from scored uncertain data and computes lower/upper approximations of class unions. | Approximation system and dominance-based rules | Preference rule induction |

*Table 9.1 (continued).*

| Method | Type | Basic input structure | Core mechanism | Primary output | Decision role |
|--------|------|----------------------|----------------|----------------|---------------|
| Markov Decision Process | Fuzzy | States, actions, fuzzy transitions, and fuzzy rewards | Propagates fuzzy states and evaluates discounted fuzzy returns, typically through fuzzy Bellman-type updates. | Fuzzy value function and policy | Sequential decision-making |
| Markov Decision Process | Uncertain | Uncertain transition degrees and uncertain reward degrees | Scores and normalizes uncertain transitions into probabilities, scores rewards, and solves the induced crisp MDP. | Value function and optimal policy | Sequential decision-making |
| Evidential Reasoning | Fuzzy | Fuzzy belief rules, matching degrees, and belief distributions over evaluation grades | Activates rules, constructs evidence masses, and aggregates them by ER/Dempster–Shafer style combination. | Aggregated belief distribution | Evidence aggregation / assessment |
| Evidential Reasoning | Uncertain | Uncertain belief degrees and uncertain reliabilities scored into $[0, 1]$ | Converts uncertain evidence into crisp masses and combines them sequentially under ER aggregation formulas. | Synthesized belief distribution with incompleteness degree | Evidence aggregation / assessment |

**Note.** Decision trees are primarily local partition-based predictive models; DRSA is an approximation- and dominance-based rule-induction framework; MDPs are sequential optimization models under state transitions; and evidential reasoning is an evidence-fusion framework over evaluation grades. In the uncertain versions, the common structural principle is to start from uncertain-valued inputs in $\mathrm{Dom}(M)$, project them through an admissible score, and then apply a well-defined crisp decision mechanism.

## 9.1 Fuzzy Decision Tree

A decision tree is a rule-based model splitting data by feature tests, forming branches to predict classes or values [1154, 1155]. A fuzzy decision tree replaces crisp splits with fuzzy membership functions, enabling soft rules, smoother boundaries, and robustness to noise [1156–1158].

**Definition 9.1.1** (Fuzzy decision tree (attribute-value / fuzzy-partition form)). [1156–1158] Let $X \neq \varnothing$ be a finite set of training objects, and let

$$\mathcal{F}(X) := \{ A : X \to [0, 1] \}$$

denote the family of all fuzzy subsets of $X$. For $A \in \mathcal{F}(X)$, define its (sigma-count) fuzzy cardinality by

$$M(A) := \sum_{x \in X} A(x).$$

Fix $m \geq 1$ attributes. For each attribute $i \in \{1, \ldots, m\}$, let $X_i \subseteq \mathcal{F}(X)$ be the (finite) set of fuzzy attribute-values (linguistic values), and assume $|X_i| \geq 1$. Fix a $t$-norm $T$ (e.g. $T = \min$) and define the fuzzy intersection

$$(A \wedge_T B)(x) := T\big(A(x), B(x)\big) \qquad (x \in X).$$

A *fuzzy decision tree* is a rooted directed tree $\mathcal{T}$ whose nodes are fuzzy subsets of $X$ (i.e. elements of $\mathcal{F}(X)$) such that:

(a) Every node $N$ of $\mathcal{T}$ satisfies $N \in \mathcal{F}(X)$.

(b) For every non-leaf node $N$, there exists an attribute index $i \in \{1, \ldots, m\}$ for which the set of children of $N$ is exactly
$$\mathrm{ch}(N) = \{\, A \wedge_T N \ : \ A \in X_i \,\} \subseteq \mathcal{F}(X).$$

(c) Each leaf is assigned one or more class labels (classification decisions).

**Definition 9.1.2** (Soft/complete fuzzy decision tree (membership-propagation semantics)). Let $U$ be a universe of objects (instances). A *soft (fuzzy) decision tree* for a (crisp or fuzzy) target class $C$ consists of a rooted directed tree whose nodes $v$ correspond to fuzzy subsets $S_v \subseteq U$ with membership $\mu_v : U \to [0,1]$, and:

- each internal node $v$ selects an attribute $a$ and a *discriminator* (soft split) $\sigma_v : U \to [0,1]$ (e.g. a piecewise-linear function governed by a cut-point $\alpha$ and width $\beta$), and for a binary split defines the children $v_L, v_R$ by
$$\mu_{v_L}(o) := \mu_v(o)\,\sigma_v(o), \qquad \mu_{v_R}(o) := \mu_v(o)\,\big(1 - \sigma_v(o)\big) \qquad (o \in U).$$

- each leaf $\ell$ carries a numeric label $L_\ell \in \mathbb{R}$ (often $L_\ell \in [0,1]$ for class-membership prediction).

For an input object $o \in U$, the tree output (estimated class-membership) is the leaf-aggregation
$$\widehat{\mu}_C(o) := \frac{\displaystyle\sum_{\ell \in \mathrm{Leaves}} \mu_\ell(o)\,L_\ell}{\displaystyle\sum_{\ell \in \mathrm{Leaves}} \mu_\ell(o)},$$
with the usual convention that the denominator is positive (e.g. $\mu_{\mathrm{root}}(o) = 1$).

Using an Uncertain Set, we define an Uncertain decision tree (U-tree) as follows.

**Definition 9.1.3** (Uncertain decision tree (U-tree) as a measurable partition rule). Let $(\Omega, \mathcal{F})$ be a measurable space of instances and let $Y$ be a label space (e.g. $Y = \{1, \ldots, K\}$ for classification or $Y = \mathbb{R}$ for regression). Fix an uncertain model $M$ with $\mathrm{Dom}(M) \neq \emptyset$ and an admissible score $S_M$.

Assume $d$ uncertain-valued features
$$X_j : \Omega \to \mathrm{Dom}(M) \qquad (j = 1, \ldots, d),$$
and let $\mathcal{S} = \{S_v\}_{v \in V}$ be the node set of a finite rooted directed tree $\mathcal{T}$ with root $r$. An *uncertain decision tree* is specified by:

- a finite rooted tree $\mathcal{T} = (V, E)$;

- for each internal node $v \in V$ a *split rule* determined by an index $j(v) \in \{1, \ldots, d\}$ and a measurable set $B_v \subseteq \mathbb{R}$, producing two children $v_L, v_R$ and the measurable tests

$$o \in S_{v_L} \iff S_M(X_{j(v)}(o)) \in B_v, \qquad o \in S_{v_R} \iff S_M(X_{j(v)}(o)) \notin B_v;$$

- for each leaf $\ell$ a prediction value $h(\ell) \in Y$.

The induced predictor $H : \Omega \to Y$ is

$$H(o) := h(\ell(o)),$$

where $\ell(o)$ is the unique leaf reached by starting at the root and applying the tests along the unique root-to-leaf path.

**Definition 9.1.4** (Soft uncertain decision tree (membership-propagation form)). Let $(\Omega, \mathcal{F})$ be a measurable space. Fix an uncertain model $M$ and an admissible score $S_M$. A *soft uncertain decision tree* is a finite rooted tree $\mathcal{T} = (V, E)$ with:

- a root membership $\mu_r : \Omega \to [0, 1]$ (typically $\mu_r \equiv 1$);

- for each internal node $v$, a feature index $j(v)$ and a measurable *soft split* function

$$\sigma_v : \Omega \to [0, 1], \qquad \sigma_v(o) = \Sigma_v\big(S_M(X_{j(v)}(o))\big)$$

for some measurable $\Sigma_v : \mathbb{R} \to [0, 1]$ (e.g. a sigmoid or triangular membership around a cut-point); with children $v_L, v_R$ defined by the membership propagation

$$\mu_{v_L}(o) := \mu_v(o)\, \sigma_v(o), \qquad \mu_{v_R}(o) := \mu_v(o)\,\big(1 - \sigma_v(o)\big);$$

- for each leaf $\ell$, a numeric label $L_\ell \in \mathbb{R}$ (or a class-probability vector in $\Delta^{K-1}$).

The tree output is the leaf aggregation

$$\widehat{H}(o) := \begin{cases} \dfrac{\sum_{\ell \in \mathrm{Leaves}} \mu_\ell(o)\, L_\ell}{\sum_{\ell \in \mathrm{Leaves}} \mu_\ell(o)}, & \sum_\ell \mu_\ell(o) > 0, \\ 0, & \sum_\ell \mu_\ell(o) = 0, \end{cases}$$

(with 0 replaced by any fixed default output when the denominator is 0).

**Theorem 9.1.5** (Well-definedness of uncertain decision trees). *In Definitions 9.1.3 and 9.1.4, assume:*

*(i) $S_M$ is admissible (finite everywhere);*

*(ii) all sets $B_v \subseteq \mathbb{R}$ and all functions $\Sigma_v : \mathbb{R} \to [0, 1]$ are measurable;*

*(iii) $\mathcal{T}$ is finite.*

*Then:*

(a) *The hard uncertain decision tree predictor $H : \Omega \to Y$ is well-defined.*

(b) *In the soft uncertain decision tree, each node membership $\mu_v$ is well-defined and satisfies $0 \leq \mu_v \leq 1$.*

(c) *For every $o \in \Omega$, the soft output $\widehat{H}(o)$ is well-defined as a real number under the stated denominator convention.*

*Proof.* (a) Since $\mathcal{T}$ is a rooted tree, each instance $o$ follows a unique path from the root to a leaf by repeatedly applying the binary test "$S_M(X_{j(v)}(o)) \in B_v$". Because $S_M(X_{j(v)}(o))$ is a finite real and $B_v$ is a fixed set, the test outcome is defined at each internal node. Finiteness of $\mathcal{T}$ implies the path terminates at a leaf $\ell(o)$ in finitely many steps. Thus $H(o) = h(\ell(o))$ is defined for all $o$.

(b) By induction on depth. At the root, $\mu_r(o) \in [0,1]$. Suppose $\mu_v(o) \in [0,1]$ for some node $v$. Since $\sigma_v(o) = \Sigma_v(S_M(X_{j(v)}(o))) \in [0,1]$, we have $\mu_{v_L}(o) = \mu_v(o)\sigma_v(o) \in [0,1]$ and $\mu_{v_R}(o) = \mu_v(o)(1-\sigma_v(o)) \in [0,1]$. Thus all node memberships are well-defined and bounded in $[0,1]$.

(c) For each $o$, the set of leaves is finite and each $\mu_\ell(o) \in [0,1]$, so the sums $\sum_\ell \mu_\ell(o)$ and $\sum_\ell \mu_\ell(o)L_\ell$ are finite real numbers. If the denominator is positive, the ratio is well-defined; if it is 0, the definition assigns a fixed default value, so $\widehat{H}(o)$ is well-defined for all $o$. $\square$

Table 9.2: Related concepts of decision trees under uncertainty-aware models.

| $k$ | Related decision tree concept(s) |
|---|---|
| 2 | Intuitionistic Fuzzy Decision Trees [1159] |
| 3 | Hesitant Fuzzy Decision Trees [1160, 1161] |
| 3 | Neutrosophic Decision Trees [1162, 1163] |

Besides Uncertain Decision Trees, several other related decision-tree concepts are also known, including Boosted Decision Trees [1164, 1165], Decision Hypertrees [1166], Rough Decision Trees [1167, 1168], Directed Decision Trees [1169, 1170], Complete Decision Trees [1171, 1172], Phonetic Decision Trees [1173, 1174], and Multi-Criteria Decision Trees [1175, 1176].

## 9.2 Fuzzy DRSA (Fuzzy Dominance-based rough approximation)

Dominance-based rough set is a rough-set approach for preference-ordered data that uses dominance relations to approximate decision classes in multicriteria decision analysis [1177–1179]. DRSA uses dominance relations on ordered criteria to approximate decision classes, inducing if–then decision rules and identifying reducts without requiring globally fixed cutoffs in advance [1180, 1181]. Fuzzy DRSA softens dominance and class boundaries using membership grades, producing fuzzy lower/upper approximations and graded decision rules that better capture vague preferences in practice [1182, 1183].

**Definition 9.2.1** (Fuzzy DRSA (fuzzy extensions of the dominance-based rough set approach))**.** [1182] Let

$$S = \langle U, Q \cup \{d\}, V, f \rangle$$

be a (finite) decision table, where $U$ is the universe of objects, $Q = \{q_1, \ldots, q_m\}$ is the set of condition criteria, $d$ is the decision criterion, and $f : U \times (Q \cup \{d\}) \to V$ is the information function. Assume that the decision attribute induces an ordered family of decision classes

$$\mathrm{Cl}_1, \ldots, \mathrm{Cl}_k, \qquad \text{with higher index meaning a (weakly) better class.}$$

Fix fuzzy logic connectives: a $t$-norm $T$, a $t$-conorm $S$, and an implicator $I$ (typically the $R$-implicator of a left-continuous $t$-norm).

A *Fuzzy DRSA model* consists of the following ingredients.

**(1) Fuzzy (criterion-wise) ordering relations and fuzzy dominance.** For each criterion $q \in Q$, let

$$R_q : U \times U \to [0,1]$$

be a fuzzy preorder (reflexive and $T$-transitive). For a nonempty set of criteria $P \subseteq Q$, define the *fuzzy dominance relation*

$$D_P(u,v) := T_{q \in P}\big(R_q(u,v)\big) \in [0,1], \qquad u,v \in U.$$

(Thus $D_P(u,v)$ is the degree to which "$u$ is at least as good as $v$ w.r.t. all criteria in $P$".) Write $D := D_P$ when $P$ is fixed.

**(2) Fuzzy decision classes and fuzzy class unions.** In fuzzy DRSA, each class is allowed to be a fuzzy set

$$\mathrm{Cl}_t : U \to [0,1], \qquad \mathrm{Cl}_t(u) \text{ is the credibility that } u \text{ belongs to class } \mathrm{Cl}_t.$$

Define the *fuzzy upward* and *fuzzy downward* unions (one standard choice) by

$$\mathrm{Cl}_t^{\geq}(u) := \max_{s \geq t} \mathrm{Cl}_s(u), \qquad \mathrm{Cl}_t^{\leq}(u) := \max_{s \leq t} \mathrm{Cl}_s(u), \qquad u \in U, \ t = 1, \dots, k.$$

**(3) Fuzzified quantifiers.** Let $\mathrm{qua}_{\forall}$ and $\mathrm{qua}_{\exists}$ be fuzzy quantifiers used to aggregate truth degrees over $U$. Two canonical choices are:

$$(\mathrm{qua}_{\forall}, \mathrm{qua}_{\exists}) = (T, S) \quad \text{(finite-$U$ aggregation)} \qquad \text{or} \qquad (\mathrm{qua}_{\forall}, \mathrm{qua}_{\exists}) = (\inf, \sup) \quad \text{(general)}.$$

If $(\mathrm{qua}_{\forall}, \mathrm{qua}_{\exists}) = (T, S)$, interpret

$$\mathrm{qua}_{\forall}(a_v; \ v \in U) = T_{v \in U} a_v, \qquad \mathrm{qua}_{\exists}(a_v; \ v \in U) = S_{v \in U} a_v.$$

**(4) Fuzzy DRSA lower/upper approximations.** For each $t \in \{1, \dots, k\}$, define fuzzy lower and upper approximations of the unions as fuzzy sets on $U$ with membership degrees in $[0,1]$:

$$\underline{\mathrm{apr}}_D^{\mathrm{qua}_{\forall}, I}\big(\mathrm{Cl}_t^{\geq}\big)(u) := \mathrm{qua}_{\forall}\Big(I\big(D(v,u),\, \mathrm{Cl}_t^{\geq}(v)\big); \ v \in U\Big),$$

$$\overline{\mathrm{apr}}_D^{\mathrm{qua}_{\exists}, T}\big(\mathrm{Cl}_t^{\geq}\big)(u) := \mathrm{qua}_{\exists}\Big(T\big(D(u,v),\, \mathrm{Cl}_t^{\geq}(v)\big); \ v \in U\Big),$$

$$\underline{\mathrm{apr}}_D^{\mathrm{qua}_{\forall}, I}\big(\mathrm{Cl}_t^{\leq}\big)(u) := \mathrm{qua}_{\forall}\Big(I\big(D(u,v),\, \mathrm{Cl}_t^{\leq}(v)\big); \ v \in U\Big),$$

$$\overline{\mathrm{apr}}_D^{\mathrm{qua}_{\exists}, T}\big(\mathrm{Cl}_t^{\leq}\big)(u) := \mathrm{qua}_{\exists}\Big(T\big(D(v,u),\, \mathrm{Cl}_t^{\leq}(v)\big); \ v \in U\Big).$$

We define Uncertain DRSA of type $M$ (U-DRSA) as follows.

**Definition 9.2.2** (Uncertain DRSA of type $M$ (U-DRSA)). Let

$$\mathsf{DT} = \langle U,\, Q \cup \{d\},\, V,\, f \rangle$$

be a finite decision table, where $U$ is a finite universe of objects, $Q = \{q_1, \ldots, q_m\}$ is the set of condition criteria, $d$ is the decision criterion, and $f : U \times (Q \cup \{d\}) \to V$ is the information function.

Fix an uncertain model $M$ with $\mathrm{Dom}(M) \neq \emptyset$ and an admissible score $S_M$.

**(A) Uncertain-valued criteria and their crisp projections.** Assume that each condition criterion $q \in Q$ and the decision criterion $d$ take values in $\mathrm{Dom}(M)$:

$$f(u, q) \in \mathrm{Dom}(M) \quad (q \in Q), \qquad f(u, d) \in \mathrm{Dom}(M) \quad (u \in U).$$

Define the projected (crisp) criterion values

$$g_q(u) := S_M\big(f(u, q)\big) \in \mathbb{R} \quad (q \in Q), \qquad g_d(u) := S_M\big(f(u, d)\big) \in \mathbb{R} \quad (u \in U).$$

**(B) Ordered decision classes and unions.** Assume the decision attribute induces an ordered family of decision classes

$$\mathrm{Cl}_1, \ldots, \mathrm{Cl}_K \subseteq U, \qquad \text{with higher index meaning a (weakly) better class.}$$

Define the upward and downward unions

$$\mathrm{Cl}^{\geq t} := \bigcup_{s=t}^{K} \mathrm{Cl}_s, \qquad \mathrm{Cl}^{\leq t} := \bigcup_{s=1}^{t} \mathrm{Cl}_s, \qquad t = 1, \ldots, K.$$

**(C) Dominance relations on criteria.** For any nonempty subset $P \subseteq Q$, define the (crisp) dominance relation

$$D_P := \{(u, v) \in U \times U : \ g_q(u) \geq g_q(v) \ \ \forall q \in P\}.$$

Equivalently, $(u, v) \in D_P$ means that $u$ dominates $v$ on all criteria in $P$. For $u \in U$ define the dominance cones

$$D_P^+(u) := \{v \in U : \ (v, u) \in D_P\} \quad \text{(objects dominating } u\text{)},$$

$$D_P^-(u) := \{v \in U : \ (u, v) \in D_P\} \quad \text{(objects dominated by } u\text{)}.$$

**(D) DRSA lower/upper approximations of unions.** For each $t \in \{1, \ldots, K\}$ define:

$$\underline{P}\big(\mathrm{Cl}^{\geq t}\big) := \{u \in U : \ D_P^+(u) \subseteq \mathrm{Cl}^{\geq t}\}, \qquad \overline{P}\big(\mathrm{Cl}^{\geq t}\big) := \{u \in U : \ D_P^-(u) \cap \mathrm{Cl}^{\geq t} \neq \varnothing\},$$

$$\underline{P}\big(\mathrm{Cl}^{\leq t}\big) := \{u \in U : \ D_P^-(u) \subseteq \mathrm{Cl}^{\leq t}\}, \qquad \overline{P}\big(\mathrm{Cl}^{\leq t}\big) := \{u \in U : \ D_P^+(u) \cap \mathrm{Cl}^{\leq t} \neq \varnothing\}.$$

The quadruple of approximations

$$\Big( \underline{P}(\mathrm{Cl}^{\geq t}), \overline{P}(\mathrm{Cl}^{\geq t}), \underline{P}(\mathrm{Cl}^{\leq t}), \overline{P}(\mathrm{Cl}^{\leq t}) \Big)_{t=1}^{K}$$

is called the *Uncertain DRSA approximation system* associated with $(\mathsf{DT}, M, S_M, P)$.

**Theorem 9.2.3** (Well-definedness of U-DRSA). *Under Definition 9.2.2, assume $U$ is finite, $\mathrm{Dom}(M) \neq \emptyset$, and $S_M$ is admissible. Then, for any nonempty $P \subseteq Q$ and any $t \in \{1, \ldots, K\}$:*

*(i) The dominance relation $D_P \subseteq U \times U$ is well-defined and is a preorder (reflexive and transitive).*

*(ii) The dominance cones $D_P^+(u)$ and $D_P^-(u)$ are well-defined subsets of $U$ for all $u \in U$.*

*(iii) The sets $\underline{P}(\mathrm{Cl}^{\geq t})$, $\overline{P}(\mathrm{Cl}^{\geq t})$, $\underline{P}(\mathrm{Cl}^{\leq t})$, and $\overline{P}(\mathrm{Cl}^{\leq t})$ are well-defined subsets of $U$.*

*Proof.* (i) Since $S_M$ is admissible, each $g_q(u) = S_M(f(u, q))$ is a finite real number, so the comparisons $g_q(u) \geq g_q(v)$ are well-defined. Hence $D_P$ is well-defined. Reflexivity holds because $g_q(u) \geq g_q(u)$ for all $q \in P$, so $(u, u) \in D_P$. Transitivity: if $(u, v) \in D_P$ and $(v, w) \in D_P$, then for each $q \in P$, $g_q(u) \geq g_q(v) \geq g_q(w)$, hence $(u, w) \in D_P$. Thus $D_P$ is a preorder.

(ii) Given $D_P$, the cones $D_P^+(u) = \{v : (v, u) \in D_P\}$ and $D_P^-(u) = \{v : (u, v) \in D_P\}$ are defined by set comprehension over the finite universe $U$, hence are well-defined subsets of $U$.

(iii) Each approximation set is defined using standard set operations (subset test, intersection, emptiness) applied to the well-defined sets $D_P^+(u), D_P^-(u)$ and unions $\mathrm{Cl}^{\geq t}, \mathrm{Cl}^{\leq t}$, hence is well-defined. □

## 9.3 Fuzzy Markov Decision Process

A Markov decision process models sequential decisions with states, actions, transition probabilities, and rewards; dynamic programming (value/policy iteration) yields an optimal policy maximizing expected return [1184, 1185]. A fuzzy MDP represents transitions, rewards, or states with fuzzy sets when probabilities are imprecise; fuzzy Bellman updates compute fuzzy value functions and robust policies [1186, 1187].

**Definition 9.3.1** (Fuzzy Markov Decision Process (FMDP)). [1186, 1187] Let $(S, \Sigma_S)$ be a (finite or measurable) state space and $(A, \Sigma_A)$ an action space. For each $s \in S$, let $A(s) \subseteq A$ be the (nonempty) feasible action set. Fix a discount factor $\gamma \in [0, 1)$.

A *fuzzy Markov decision process* (in the possibilistic/max–min sense) is a tuple

$$\mathcal{M} = \big(S, \{A(s)\}_{s \in S}, \widetilde{P}, \widetilde{r}, \gamma\big),$$

where:

1. **Fuzzy transition kernel (possibility kernel).** For each $(s, a) \in S \times A(s)$, the mapping

$$\widetilde{P}(\,\cdot\,|s, a) : S \to [0, 1], \qquad s' \mapsto \widetilde{P}(s'|s, a),$$

is a *normal* fuzzy set on $S$ (i.e. $\sup_{s' \in S} \widetilde{P}(s'|s, a) = 1$). It induces a possibility measure on $(S, \Sigma_S)$ by

$$\Pi(B \,|\, s, a) := \sup_{s' \in B} \widetilde{P}(s'|s, a), \qquad B \in \Sigma_S.$$

2. **Fuzzy reward.** For each $(s, a) \in S \times A(s)$, the immediate reward is a fuzzy set

$$\widetilde{r}(s, a) \in \mathcal{F}(\mathbb{R}),$$

often restricted to fuzzy numbers (normal, upper semicontinuous, and with bounded support).

3. **Policies.** A (deterministic) *stationary policy* is a map $f : S \to A$ with $f(s) \in A(s)$. More generally, a (possibly randomized) history-dependent policy is a family $\pi = \{\pi_t\}_{t \geq 0}$ with $\pi_t(\cdot \mid h_t)$ a probability distribution over $A(s_t)$ given a history $h_t = (s_0, a_0, \ldots, s_t)$.

**Possibilistic state evolution (fuzzy state / possibility distribution).** A fuzzy state (possibility distribution) is $\widetilde{x} \in \mathcal{F}(S)$, i.e. a map $\widetilde{x} : S \to [0, 1]$ with $\sup_{s \in S} \widetilde{x}(s) = 1$. Given a chosen action $a_t \in A(\cdot)$ at time $t$, the next fuzzy state is defined by the max–min composition

$$\widetilde{x}_{t+1}(s') = \bigvee_{s \in S} \left( \widetilde{x}_t(s) \wedge \widetilde{P}(s' \mid s, a_t) \right), \qquad s' \in S,$$

where $\vee = \max$ and $\wedge = \min$.

**Discounted fuzzy return.** Let $\oplus$ and $\odot$ denote fuzzy addition and scalar multiplication (e.g. defined via Zadeh's extension principle; equivalently, via $\alpha$-cuts when $\widetilde{r}(s, a)$ are fuzzy numbers):

$$(\widetilde{u} \oplus \widetilde{v})_\alpha = (\widetilde{u})_\alpha + (\widetilde{v})_\alpha, \qquad (c \odot \widetilde{u})_\alpha = c \, (\widetilde{u})_\alpha \quad (c \geq 0).$$

For a fixed policy $\pi$ and initial fuzzy state $\widetilde{x}_0$, define the discounted total fuzzy reward (formally) by

$$\widetilde{G}^\pi(\widetilde{x}_0) := \bigoplus_{t=0}^{\infty} \gamma^t \odot \widetilde{r}(s_t, a_t),$$

where $(s_t, a_t)$ are generated under $\pi$ and the transition kernel $\widetilde{P}$. (For bounded rewards and $\gamma < 1$, this series is well-defined under standard choices of fuzzy arithmetic, typically by checking convergence on $\alpha$-cuts.)

The definition of Uncertain Markov decision process of type $M$ (U-MDP) is given below.

**Definition 9.3.2** (Uncertain Markov decision process of type $M$ (U-MDP))**.** Let $S$ be a finite nonempty state set and $A$ a finite nonempty action set. For each $s \in S$, let $A(s) \subseteq A$ be a nonempty feasible action set. Fix a discount factor $\gamma \in [0, 1)$.

Fix an uncertain model $M$ with $\mathrm{Dom}(M) \neq \emptyset$ and admissible scores $S_M^P, S_M^R$. An *uncertain Markov decision process of type $M$* is a tuple

$$\mathcal{M}_M = \big( S, \{A(s)\}_{s \in S}, P_M, R_M, \gamma; S_M^P, S_M^R \big),$$

where:

- **Uncertain transition degrees:** $P_M : S \times A \times S \to \mathrm{Dom}(M)$ assigns an uncertainty degree tuple $P_M(s' \mid s, a) \in \mathrm{Dom}(M)$ to each $(s, a, s')$ with $a \in A(s)$.

- **Uncertain reward degrees:** $R_M : S \times A \to \mathrm{Dom}(M)$ assigns an uncertainty degree tuple $R_M(s,a) \in \mathrm{Dom}(M)$ to each $(s,a)$ with $a \in A(s)$.

**Induced (crisp) MDP via scoring and normalization.** Define the (strictly positive) transition scores

$$\alpha(s'|s,a) := S_M^P\big(P_M(s'|s,a)\big) \in (0,\infty), \qquad a \in A(s),$$

and the normalizing constant

$$Z(s,a) := \sum_{t \in S} \alpha(t|s,a) \in (0,\infty).$$

Define the induced transition probabilities

$$p(s'|s,a) := \frac{\alpha(s'|s,a)}{Z(s,a)} \in (0,1), \qquad \sum_{s' \in S} p(s'|s,a) = 1.$$

Define the induced immediate reward

$$r(s,a) := S_M^R\big(R_M(s,a)\big) \in \mathbb{R}.$$

The induced (crisp) MDP is $\big(S, \{A(s)\}, p, r, \gamma\big)$.

**Policies and value functions.** A (deterministic) stationary policy is a map $\pi : S \to A$ with $\pi(s) \in A(s)$. For such $\pi$, define the value function $V^\pi : S \to \mathbb{R}$ by the discounted expectation

$$V^\pi(s) := \mathbb{E}^\pi \left[ \sum_{t=0}^{\infty} \gamma^t \, r(S_t, \pi(S_t)) \ \Big| \ S_0 = s \right],$$

where $(S_t)_{t \geq 0}$ evolves according to $p(\cdot|s, \pi(s))$.

The optimal value function is

$$V^\star(s) := \sup_\pi V^\pi(s),$$

and an optimal policy $\pi^\star$ satisfies $V^{\pi^\star} = V^\star$.

**Proposition 9.3.3** (Probability well-definedness). *Under Definition 9.3.2, for every $(s,a)$ with $a \in A(s)$, the induced transition probabilities $p(\cdot|s,a)$ are well-defined and satisfy*

$$p(s'|s,a) \geq 0, \qquad \sum_{s' \in S} p(s'|s,a) = 1.$$

*Proof.* $\alpha(s'|s,a) = S_M^P(P_M(s'|s,a)) > 0$ for every $s'$. Hence $Z(s,a) = \sum_{t \in S} \alpha(t|s,a) > 0$, so $p(s'|s,a) = \alpha(s'|s,a)/Z(s,a)$ is well-defined and nonnegative. Summing over $s'$ yields $\sum_{s'} p(s'|s,a) = Z(s,a)/Z(s,a) = 1$. $\qquad \square$

**Theorem 9.3.4** (Well-definedness of discounted values and Bellman optimality). *Assume the setting of Definition 9.3.2 and suppose the induced rewards are bounded: there exists $R_{\max} < \infty$ such that $|r(s,a)| \leq R_{\max}$ for all feasible $(s,a)$. Then:*

(i) *For every stationary policy $\pi$, the value function $V^\pi$ is well-defined and bounded:*

$$|V^\pi(s)| \le \frac{R_{\max}}{1-\gamma} \qquad (s \in S).$$

(ii) *The Bellman expectation operator $T^\pi : \mathbb{R}^S \to \mathbb{R}^S$,*

$$(T^\pi V)(s) := r(s, \pi(s)) + \gamma \sum_{s' \in S} p(s'|s, \pi(s))\, V(s'),$$

*is a $\gamma$-contraction in the sup norm $\|\cdot\|_\infty$ and has a unique fixed point equal to $V^\pi$.*

(iii) *The Bellman optimality operator $T^\star : \mathbb{R}^S \to \mathbb{R}^S$,*

$$(T^\star V)(s) := \max_{a \in A(s)} \left[ r(s, a) + \gamma \sum_{s' \in S} p(s'|s, a)\, V(s') \right],$$

*is also a $\gamma$-contraction and has a unique fixed point $V^\star$. Moreover, there exists an optimal deterministic stationary policy $\pi^\star$ attaining the maximizers in $T^\star$.*

*Proof.* (i) Since $|r(S_t, \pi(S_t))| \le R_{\max}$ almost surely, the discounted series is absolutely bounded by

$$\sum_{t=0}^{\infty} \gamma^t R_{\max} = \frac{R_{\max}}{1-\gamma},$$

so the expectation defining $V^\pi(s)$ is finite and the bound holds.

(ii) For any $V, W \in \mathbb{R}^S$ and any $s \in S$,

$$|(T^\pi V)(s) - (T^\pi W)(s)| = \gamma \left| \sum_{s'} p(s'|s, \pi(s))\, (V(s') - W(s')) \right| \le \gamma \sum_{s'} p(s'|s, \pi(s))\, \|V - W\|_\infty = \gamma \|V - W\|_\infty.$$

Taking $\sup_s$ gives $\|T^\pi V - T^\pi W\|_\infty \le \gamma \|V - W\|_\infty$, so $T^\pi$ is a contraction. By the Banach fixed-point theorem on the complete metric space $(\mathbb{R}^S, \|\cdot\|_\infty)$, $T^\pi$ has a unique fixed point. Standard MDP arguments show that this fixed point equals the discounted value function $V^\pi$.

(iii) For any $V, W$ and any $s$,

$$|(T^\star V)(s) - (T^\star W)(s)| \le \max_{a \in A(s)} \gamma \left| \sum_{s'} p(s'|s, a)(V(s') - W(s')) \right| \le \gamma \|V - W\|_\infty,$$

hence $\|T^\star V - T^\star W\|_\infty \le \gamma \|V - W\|_\infty$, so $T^\star$ is a contraction and has a unique fixed point $V^\star$. Since $A(s)$ is finite, the maximum in $(T^\star V)(s)$ is attained for each $s$; choosing an argmax action defines a deterministic stationary policy $\pi^\star$, and the fixed-point identity $V^\star = T^\star V^\star$ implies optimality. $\qquad \square$

Related concepts of Markov decision processes under uncertainty-aware models are listed in Table 9.3.

Table 9.3: Related concepts of Markov decision processes under uncertainty-aware models.

| $k$ | Related Markov decision process concept(s) |
|---|---|
| 1 | Fuzzy Markov Decision Processes |
| 3 | Neutrosophic Markov Decision Processes [1188, 1189] |

## 9.4 Fuzzy Evidential Reasoning

Classical Evidential Reasoning assigns belief degrees to evaluation grades, combines multiple criteria using evidential reasoning aggregation based on Dempster Shafer calculus, then computes expected utilities [1190, 1191]. Fuzzy Evidential Reasoning represents each criterion as fuzzy belief degrees over evaluation grades, combines evidence across criteria using ER rules, then defuzzifies utilities for ranking [1192, 1193].

**Definition 9.4.1** (Fuzzy Evidential Reasoning (Fuzzy ER)). [1194, 1195] Let $\Theta = \{H_1, \ldots, H_N\}$ be a finite set of *evaluation grades* (hypotheses). Let $X_i$ be the domain of the $i$-th antecedent attribute ($i = 1, \ldots, M$). For each $i$, fix a linguistic term set $\mathcal{A}_i = \{A_{i,1}, \ldots, A_{i,J_i}\}$, where each $A_{i,j}$ is represented by a fuzzy set on $X_i$ with membership function $\mu_{i,j} : X_i \to [0, 1]$.

A *belief-rule base* (fuzzy rule base with belief structure) is a collection $\mathcal{R} = \{R_k\}_{k=1}^{L}$ of rules of the form

$$R_k : \quad \textbf{IF } x_1 \text{ is } A_1^k \wedge \cdots \wedge x_M \text{ is } A_M^k \textbf{ THEN } \big\{(H_j, \beta_{k,j})\big\}_{j=1}^{N},$$

where $A_i^k \in \mathcal{A}_i$ is the antecedent linguistic term used in rule $k$ for attribute $i$, and the *belief degrees* satisfy

$$\beta_{k,j} \in [0, 1], \qquad \sum_{j=1}^{N} \beta_{k,j} \leq 1 \quad (k = 1, \ldots, L).$$

(The inequality allows *incompleteness* of the consequent belief assignment.)

Given an observation (possibly fuzzy) $\widetilde{x}_i \in \mathcal{F}(X_i)$ for each attribute $i$, define the *matching degree* between $\widetilde{x}_i$ and the antecedent term $A_i^k$ by the Max–Min similarity

$$\alpha_i^k := \max_{t \in X_i} \min\big(\mu_{\widetilde{x}_i}(t), \mu_{A_i^k}(t)\big) \in [0, 1], \qquad i = 1, \ldots, M, \ k = 1, \ldots, L.$$

(If the input is crisp $x_i \in X_i$, one may take $\mu_{\widetilde{x}_i}(t) = \mathbb{1}_{\{x_i\}}(t)$, yielding $\alpha_i^k = \mu_{A_i^k}(x_i)$.)

Let $\omega_k \geq 0$ denote an optional *intrinsic rule weight* (importance of rule $k$). Define the *activation weight* (normalized firing strength) of rule $k$ by

$$\theta_k := \frac{\omega_k \prod_{i=1}^{M} \alpha_i^k}{\sum_{\ell=1}^{L} \omega_\ell \prod_{i=1}^{M} \alpha_i^\ell} \quad \text{so that} \quad \theta_k \geq 0, \quad \sum_{k=1}^{L} \theta_k = 1.$$

**Evidence-mass construction.** For each rule $k$, define basic probability masses on $\Theta \cup \{D\}$ by

$$m_k(H_j) := \theta_k \, \beta_{k,j} \qquad (j = 1, \ldots, N),$$

$$m_k(D) := 1 - \sum_{j=1}^{N} m_k(H_j) = 1 - \theta_k \sum_{j=1}^{N} \beta_{k,j},$$

and split the unassigned mass as

$$\overline{m}_k(D) := 1 - \theta_k, \qquad \widetilde{m}_k(D) := \theta_k\Big(1 - \sum_{j=1}^{N} \beta_{k,j}\Big), \quad \text{so that } m_k(D) = \overline{m}_k(D) + \widetilde{m}_k(D).$$

**Evidential reasoning aggregation.** Initialize $m^{(1)}(\cdot) = m_1(\cdot)$. For $k = 1, \ldots, L-1$, combine $m^{(k)}$ and $m_{k+1}$ by

$$m^{(k+1)}(H_j) = K_{k+1}\Big( m^{(k)}(H_j) m_{k+1}(H_j) + m^{(k)}(H_j) m_{k+1}(D) + m^{(k)}(D) m_{k+1}(H_j) \Big),$$

$$\widetilde{m}^{(k+1)}(D) = K_{k+1}\Big( \widetilde{m}^{(k)}(D) \widetilde{m}_{k+1}(D) + \widetilde{m}^{(k)}(D) \overline{m}_{k+1}(D) + \overline{m}^{(k)}(D) \widetilde{m}_{k+1}(D) \Big),$$

$$\overline{m}^{(k+1)}(D) = K_{k+1}\, \overline{m}^{(k)}(D)\, \overline{m}_{k+1}(D), \qquad m^{(k+1)}(D) = \widetilde{m}^{(k+1)}(D) + \overline{m}^{(k+1)}(D),$$

where the normalization factor $K_{k+1}$ is

$$K_{k+1} := \Big( 1 - \sum_{j=1}^{N} \sum_{\substack{t=1 \\ t \neq j}}^{N} m^{(k)}(H_j) m_{k+1}(H_t) \Big)^{-1}.$$

**Fuzzy ER output (belief distribution).** The *Fuzzy ER* synthesis result is the belief distribution $\boldsymbol{\beta} = (\beta_1, \ldots, \beta_N, \beta_D)$ defined by

$$\beta_j := \frac{m^{(L)}(H_j)}{1 - \overline{m}^{(L)}(D)} \quad (j = 1, \ldots, N), \qquad \beta_D := \frac{\widetilde{m}^{(L)}(D)}{1 - \overline{m}^{(L)}(D)}.$$

Here $\beta_D$ quantifies the remaining (normalized) incompleteness after aggregation. The mapping

$$\text{ER} : \big\{ (\theta_k, \beta_{k,1}, \ldots, \beta_{k,N}) \big\}_{k=1}^{L} \longmapsto (\beta_1, \ldots, \beta_N, \beta_D)$$

is called *fuzzy evidential reasoning*.

Using an Uncertain Set, we define Uncertain Evidential Reasoning as follows.

**Definition 9.4.2** (Uncertain Evidential Reasoning of type $M$ (U-ER))**.** Let $\Theta = \{H_1, \ldots, H_N\}$ be a finite set of evaluation grades (hypotheses), with $N \geq 2$. Let $L \geq 1$ be the number of evidence sources (e.g. criteria, rules, experts).

Fix an uncertain model $M$ with $\text{Dom}(M) \neq \emptyset$ and an admissible score $S_M : \text{Dom}(M) \to [0, 1]$.

For each evidence source $\ell \in \{1, \ldots, L\}$ assume:

- uncertain belief degrees $b_{\ell j}^{(M)} \in \text{Dom}(M)$ for grades $H_j$ $(j = 1, \ldots, N)$,

- an uncertain reliability/activation degree $\theta_\ell^{(M)} \in \mathrm{Dom}(M)$.

Define their crisp projections

$$\beta_{\ell j} := S_M\big(b_{\ell j}^{(M)}\big) \in [0,1], \qquad \theta_\ell := S_M\big(\theta_\ell^{(M)}\big) \in [0,1].$$

**(A) Consistency (incomplete belief allowed).** Assume for each $\ell$,

$$\sum_{j=1}^{N} \beta_{\ell j} \le 1.$$

Define the incompleteness (ignorance) for source $\ell$:

$$\beta_{\ell D} := 1 - \sum_{j=1}^{N} \beta_{\ell j} \in [0,1].$$

**(B) Basic probability masses.** For each $\ell$, define a basic probability assignment (BPA) on $\Theta \cup \{D\}$ by

$$m_\ell(H_j) := \theta_\ell\, \beta_{\ell j} \quad (j = 1, \ldots, N), \qquad m_\ell(D) := 1 - \sum_{j=1}^{N} m_\ell(H_j) = 1 - \theta_\ell \sum_{j=1}^{N} \beta_{\ell j}.$$

Split $m_\ell(D)$ into

$$\overline{m}_\ell(D) := 1 - \theta_\ell, \qquad \widetilde{m}_\ell(D) := \theta_\ell\, \beta_{\ell D}, \qquad m_\ell(D) = \overline{m}_\ell(D) + \widetilde{m}_\ell(D).$$

**(C) ER aggregation (sequential combination).** Set $m^{(1)} := m_1$, $\overline{m}^{(1)} := \overline{m}_1$, $\widetilde{m}^{(1)} := \widetilde{m}_1$. For $k = 1, \ldots, L-1$, combine $(m^{(k)}, \overline{m}^{(k)}, \widetilde{m}^{(k)})$ with $(m_{k+1}, \overline{m}_{k+1}, \widetilde{m}_{k+1})$ by

$$K_{k+1} := \left( 1 - \sum_{j=1}^{N} \sum_{\substack{t=1 \\ t \ne j}}^{N} m^{(k)}(H_j)\, m_{k+1}(H_t) \right)^{-1},$$

$$m^{(k+1)}(H_j) := K_{k+1}\Big( m^{(k)}(H_j)m_{k+1}(H_j) + m^{(k)}(H_j)m_{k+1}(D) + m^{(k)}(D)m_{k+1}(H_j) \Big),$$

$$\overline{m}^{(k+1)}(D) := K_{k+1}\, \overline{m}^{(k)}(D)\, \overline{m}_{k+1}(D),$$

$$\widetilde{m}^{(k+1)}(D) := K_{k+1}\Big( \widetilde{m}^{(k)}(D)\widetilde{m}_{k+1}(D) + \widetilde{m}^{(k)}(D)\overline{m}_{k+1}(D) + \overline{m}^{(k)}(D)\widetilde{m}_{k+1}(D) \Big),$$

$$m^{(k+1)}(D) := \overline{m}^{(k+1)}(D) + \widetilde{m}^{(k+1)}(D).$$

**(D) Final belief distribution.** Let the aggregated masses be $m^{(L)}$ with split $\overline{m}^{(L)}(D), \widetilde{m}^{(L)}(D)$. Define the normalized belief degrees

$$\beta_j := \frac{m^{(L)}(H_j)}{1 - \overline{m}^{(L)}(D)} \quad (j = 1, \ldots, N), \qquad \beta_D := \frac{\widetilde{m}^{(L)}(D)}{1 - \overline{m}^{(L)}(D)}.$$

The vector $(\beta_1, \ldots, \beta_N, \beta_D)$ is called the *U-ER synthesis result*.

**Theorem 9.4.3** (Well-definedness of U-ER). *Under Definition 9.4.2, assume:*

(i) $S_M : \mathrm{Dom}(M) \to [0,1]$ *is admissible;*

(ii) *for every $\ell$, $\sum_{j=1}^{N} \beta_{\ell j} \leq 1$ (so $\beta_{\ell D} \in [0,1]$);*

(iii) *for each aggregation step $k \to k+1$, the* conflict *satisfies*

$$\kappa_{k+1} := \sum_{j=1}^{N} \sum_{\substack{t=1 \\ t \neq j}}^{N} m^{(k)}(H_j)\, m_{k+1}(H_t) \; < \; 1;$$

(iv) *the final normalization denominator is positive:*

$$1 - \overline{m}^{(L)}(D) > 0.$$

*Then:*

(a) *All masses $m_\ell(H_j), m_\ell(D), \overline{m}_\ell(D), \widetilde{m}_\ell(D)$ are well-defined in $[0,1]$.*

(b) *Each intermediate factor $K_{k+1}$ is well-defined and finite, and all aggregated masses remain in $[0,1]$.*

(c) *The final belief degrees $\beta_j, \beta_D$ are well-defined, satisfy $0 \leq \beta_j, \beta_D \leq 1$, and*

$$\sum_{j=1}^{N} \beta_j + \beta_D = 1.$$

*Proof.* (a) Since $\theta_\ell, \beta_{\ell j} \in [0,1]$, $m_\ell(H_j) = \theta_\ell \beta_{\ell j} \in [0,1]$. Also $m_\ell(D) = 1 - \theta_\ell \sum_j \beta_{\ell j} \in [0,1]$ because $\sum_j \beta_{\ell j} \leq 1$. The split satisfies $\overline{m}_\ell(D) = 1 - \theta_\ell \in [0,1]$ and $\widetilde{m}_\ell(D) = \theta_\ell \beta_{\ell D} \in [0,1]$.

(b) By assumption, $\kappa_{k+1} < 1$, hence $K_{k+1} = (1 - \kappa_{k+1})^{-1}$ is well-defined and finite. Each update formula is a finite sum/product of nonnegative terms multiplied by $K_{k+1}$, hence defines finite nonnegative masses. (These are the standard ER/Dempster-style normalized combinations, which preserve total mass 1 on $\Theta \cup \{D\}$ under the same condition.)

(c) Since $1 - \overline{m}^{(L)}(D) > 0$, the normalization defining $\beta_j$ and $\beta_D$ is valid. Nonnegativity follows from (b). Finally,

$$\sum_{j=1}^{N} \beta_j + \beta_D = \frac{\sum_{j=1}^{N} m^{(L)}(H_j) + \widetilde{m}^{(L)}(D)}{1 - \overline{m}^{(L)}(D)} = \frac{1 - \overline{m}^{(L)}(D)}{1 - \overline{m}^{(L)}(D)} = 1,$$

because $m^{(L)}(D) = \overline{m}^{(L)}(D) + \widetilde{m}^{(L)}(D)$ and the total mass on $\Theta \cup \{D\}$ equals 1. $\qquad\square$

Related concepts of evidential reasoning under uncertainty-aware models are listed in Table 9.4.

Table 9.4: Related concepts of evidential reasoning under uncertainty-aware models.

| $k$ | Related evidential reasoning concept(s) |
|---|---|
| 2 | Intuitionistic Fuzzy Evidential Reasoning [1196, 1197] |
| 2 | Pythagorean Fuzzy Evidential Reasoning [1198, 1199] |
| 3 | Hesitant Fuzzy Evidential Reasoning [1200] |
| 3 | Spherical Fuzzy Evidential Reasoning [1201] |
| 3 | Picture Fuzzy Evidential Reasoning [1202] |
| 3 | Neutrosophic Evidential Reasoning [1203, 1204] |
| $n$ | Plithogenic Evidential Reasoning [1205] |

# Chapter 10

# Other Related Decision Methods

In this section, we describe other decision-making methods and related tools. For convenience, the main concepts introduced in this chapter are compared in Table 10.1.

Table 10.1: Concise comparison of representative other related decision methods.

| Method | Type | Basic input structure | Core mechanism | Primary output | Decision role |
|---|---|---|---|---|---|
| Goal Programming | Fuzzy | Feasible set, goal functions, aspiration levels, and fuzzy tolerances | Defines goal-satisfaction memberships and maximizes the common satisfaction level over all goals | Compromise solution and satisfaction level | Goal-based optimization |
| Goal Programming | Uncertain | Feasible set, real-valued goals, and uncertain targets/tolerances scored into crisp values | Projects uncertain targets and tolerances, builds membership-style satisfaction functions, and solves a max–min compromise model | U-GP compromise solution | Goal-based optimization |
| Data Envelopment Analysis | Fuzzy | DMUs with fuzzy inputs and fuzzy outputs | Extends DEA efficiency analysis through fuzzy arithmetic or $\alpha$-cuts around the empirical frontier | Fuzzy efficiency score or interval efficiency | Efficiency evaluation |
| Data Envelopment Analysis | Uncertain | DMUs with uncertain input/output degrees scored into positive crisp data | Converts uncertain observations into crisp DEA data and solves the induced CCR-type efficiency program | Crisp efficiency scores | Efficiency evaluation |
| Social Choice | Fuzzy | Individuals, alternatives, and fuzzy preference relations | Aggregates graded preferences into a collective fuzzy preference and selects a socially preferred alternative | Collective choice or collective preference | Collective decision-making |

*Table 10.1 (continued).*

| Method | Type | Basic input structure | Core mechanism | Primary output | Decision role |
|---|---|---|---|---|---|
| Social Choice | Uncertain | Uncertain preference relations scored into $[0, 1]$ and an aggregation rule | Scores uncertain pairwise preferences, aggregates them into a collective relation, and applies a score-based social choice rule | Chosen alternative | Collective decision-making |
| Decision Table | Fuzzy | Fuzzy condition states, action states, and rule-table entries | Matches inputs to fuzzy rule columns and selects or aggregates actions by graded firing strengths | Action configuration or rule output | Rule-based decision support |
| Decision Table | Uncertain | Uncertain condition states and crisp or uncertain action values | Scores uncertain condition states, evaluates rule firing by a $t$-norm, and selects an action via deterministic tie-breaking | Action configuration | Rule-based decision support |
| Decision Matrix | Fuzzy | Alternatives, fuzzy states/events, and fuzzy evaluations | Propagates fuzzy payoffs through fuzzy arithmetic or $\alpha$-cut computations to obtain aggregate characteristics | Fuzzy expected evaluation and related indices | Matrix-based assessment |
| Decision Matrix | Uncertain | Alternatives, criteria, and uncertain evaluation entries | Scores uncertain entries into a crisp matrix and supports normalization, weighting, and aggregation operations | Scored decision matrix | Matrix-based assessment |
| Ultrafilter | Fuzzy | Fuzzy acceptance degrees on subsets of a universe | Assigns graded acceptance to subsets while preserving monotonicity, intersection, and ultrafilter-type dichotomy | Fuzzy ultrafilter / accepted family | Decisive-choice structure |
| Ultrafilter | Uncertain | Uncertain acceptance degrees on subsets with an admissible score | Converts uncertain subset-acceptance into scored acceptance satisfying ultrafilter axioms at the crisp level | Uncertain ultrafilter and induced crisp ultrafilter | Decisive-choice structure |
| SWOT Analysis | Fuzzy | Internal and external factors represented by fuzzy numbers or memberships | Aggregates internal–external factor pairs, assigns strategy quadrants, and prioritizes strategies by closeness or scoring rules | Prioritized SWOT strategies | Strategic planning |
| SWOT Analysis | Uncertain | Internal and external factors assessed by uncertain degrees and importance weights | Scores uncertain factors, assigns SO/ST/WO/WT quadrants, and ranks factor pairs by normalized interaction strength | Prioritized annotated strategies | Strategic planning |

*Table 10.1 (continued).*

| Method | Type | Basic input structure | Core mechanism | Primary output | Decision role |
|---|---|---|---|---|---|
| Cost–Benefit Analysis | Fuzzy | Fuzzy benefit and cost cashflows, possibly fuzzy discount rates | Computes fuzzy present worths and fuzzy benefit–cost measures through fuzzy arithmetic or $\alpha$-cuts | Fuzzy NPV or fuzzy BCR | Economic evaluation |
| Cost–Benefit Analysis | Uncertain | Uncertain benefits, costs, and discount-rate degrees scored into crisp quantities | Converts uncertain cashflows into crisp present worths and evaluates projects by NPV or BCR | NPV, BCR, and project ranking | Economic evaluation |
| Decision Curve Analysis | Fuzzy | Binary outcomes, fuzzy predicted risks, thresholds, and harm function | Uses threshold-exceedance degrees from fuzzy risks to compute fuzzy net benefit across thresholds | Fuzzy decision curve and optimal strategy set | Threshold-based model evaluation |
| Decision Curve Analysis | Uncertain | Binary outcomes, uncertain predicted risks, threshold-exceedance functional, and harm function | Evaluates uncertain risk exceedance at each threshold and computes net benefit for competing strategies | Uncertain decision curve and optimal strategy set | Threshold-based model evaluation |
| Rational Choice | Fuzzy | Alternatives and a fuzzy binary preference relation | Selects alternatives whose preference degree against every feasible competitor exceeds a fixed threshold | Fuzzy choice correspondence | Preference-based choice |
| Rational Choice | Uncertain | Alternatives, uncertain binary preference relation, threshold functional, and cutoff level | Scores uncertain pairwise preference comparisons through a threshold functional and induces a choice correspondence | Uncertain choice correspondence | Preference-based choice |

**Note.** Goal programming and cost–benefit analysis are optimization- and evaluation-oriented methods; data envelopment analysis focuses on relative efficiency; social choice and rational choice aggregate or compare preferences; decision tables and decision matrices provide structured rule or data representations; ultrafilters formalize decisive acceptance structures; SWOT supports strategic prioritization; and decision curve analysis evaluates threshold-based decision strategies. In the uncertain variants, the common construction is to begin with uncertain-valued inputs in $\mathrm{Dom}(M)$, project them through an admissible score or threshold functional, and then apply a well-defined crisp decision mechanism.

## 10.1   Fuzzy Goal programming

Goal programming is an optimization approach that seeks a compromise solution by minimizing weighted deviations from multiple target goals under given constraints [1206, 1207]. Fuzzy goal programming extends this idea by representing goals and constraints as fuzzy sets and then minimizing degrees of dissatisfaction, enabling robust optimization when aspiration levels or requirements are imprecise [1208, 1209].

**Definition 10.1.1** (Fuzzy goal programming (membership-based GP)). [1208, 1209] Let $X \subseteq \mathbb{R}^n$ be a nonempty feasible set and let

$$f_i : X \to \mathbb{R} \qquad (i = 1, \ldots, p)$$

be objective (goal) functions with aspiration levels (targets) $g_i \in \mathbb{R}$. Assume each goal has an admissible (tolerance) deviation $\Delta_i > 0$. Define the *goal-satisfaction membership function* $\mu_i : X \to [0,1]$ by the (symmetric) triangular form

$$\mu_i(x) := \begin{cases} 0, & f_i(x) \le g_i - \Delta_i, \\ \dfrac{f_i(x) - (g_i - \Delta_i)}{\Delta_i}, & g_i - \Delta_i \le f_i(x) \le g_i, \\ \dfrac{(g_i + \Delta_i) - f_i(x)}{\Delta_i}, & g_i \le f_i(x) \le g_i + \Delta_i, \\ 0, & f_i(x) \ge g_i + \Delta_i. \end{cases}$$

(Other one-sided or nonlinear membership profiles can be used, depending on the decision-maker's preference.)

The *fuzzy goal programming (FGP) model* selects a compromise solution by maximizing the common satisfaction level:

$$\max_{x \in X, \; \lambda \in [0,1]} \lambda \quad \text{subject to} \quad \lambda \le \mu_i(x) \;\; (i = 1, \ldots, p).$$

Any optimizer $x^\star$ is called an *FGP solution*.

Using an Uncertain Set, we define Uncertain goal programming as follows.

**Definition 10.1.2** (Uncertain goal programming of type $M$ (U-GP)). Let $X \subseteq \mathbb{R}^n$ be a nonempty feasible set, and let

$$f_i : X \to \mathbb{R} \qquad (i = 1, \ldots, p)$$

be goal functions with $p \ge 1$. Fix aspiration (target) degrees

$$g_i^{(M)} \in \text{Dom}(M) \qquad (i = 1, \ldots, p),$$

and (strictly positive) tolerance degrees

$$\Delta_i^{(M)} \in \text{Dom}(M) \qquad (i = 1, \ldots, p),$$

together with an uncertain model $M$ and admissible scores

$$S_M : \text{Dom}(M) \to \mathbb{R}, \qquad S_M^+ : \text{Dom}(M) \to (0, \infty).$$

Define the projected targets and tolerances

$$g_i := S_M(g_i^{(M)}) \in \mathbb{R}, \qquad \Delta_i := S_M^+(\Delta_i^{(M)}) \in (0, \infty).$$

Define the (membership-style) goal satisfaction for each $i$ by the triangular profile

$$\mu_i(x) := \max\left\{ 0, \; 1 - \frac{|f_i(x) - g_i|}{\Delta_i} \right\} \in [0, 1], \qquad x \in X.$$

(Thus $\mu_i(x) = 1$ at the target $f_i(x) = g_i$ and decreases linearly to 0 at distance $\Delta_i$.)

The *uncertain goal programming problem* is

$$\max_{x \in X, \ \lambda \in [0,1]} \lambda \quad \text{subject to} \quad \lambda \le \mu_i(x) \ \ (i = 1, \ldots, p).$$

Any optimizer $x^\star$ is called a *U-GP compromise solution.*

**Theorem 10.1.3** (Well-definedness and feasibility of U-GP)**.** *Under Definition 10.1.2, assume:*

(i) $X \ne \varnothing$;

(ii) $S_M$ *is admissible and* $S_M^+$ *is admissible positive (i.e.* $\Delta_i > 0$ *for all* $i$);

(iii) *each* $f_i$ *is well-defined as a real-valued function on* $X$.

*Then:*

(a) *Each membership function* $\mu_i : X \to [0,1]$ *is well-defined.*

(b) *The U-GP feasible set is nonempty (hence the optimization problem is well-defined).*

(c) *Any optimal value satisfies* $0 \le \lambda^\star \le 1$.

*Proof.* (a) Fix $i$ and $x \in X$. Since $f_i(x), g_i \in \mathbb{R}$, the quantity $|f_i(x) - g_i|$ is finite. Because $\Delta_i > 0$, the ratio $|f_i(x) - g_i|/\Delta_i$ is well-defined and nonnegative. Hence $1 - |f_i(x) - g_i|/\Delta_i$ is a real number, and taking $\max\{0, \cdot\}$ yields $\mu_i(x) \ge 0$. Moreover, since $|f_i(x) - g_i|/\Delta_i \ge 0$, we have $1 - |f_i(x) - g_i|/\Delta_i \le 1$, so $\mu_i(x) \le 1$. Thus $\mu_i(x) \in [0,1]$ and $\mu_i$ is well-defined.

(b) Because $X \ne \varnothing$, choose any $\bar{x} \in X$. By (a), $\mu_i(\bar{x}) \in [0,1]$ for all $i$. Let $\bar{\lambda} := 0$. Then $\bar{\lambda} \le \mu_i(\bar{x})$ holds for every $i$, and $\bar{\lambda} \in [0,1]$. Thus $(\bar{x}, \bar{\lambda})$ is feasible, so the feasible set is nonempty.

(c) By construction $\lambda \in [0,1]$ is enforced, hence any optimum satisfies $0 \le \lambda^\star \le 1$. $\qquad \square$

Related concepts of goal programming under uncertainty-aware models are listed in Table 10.2.

As related concepts other than Uncertain Goal Programming, Multi-Goal Programming [1218, 1219], Interactive Goal Programming [1220, 1221], Meta-Goal Programming [1222, 1223], Weighted Goal Programming [1224, 1225], Lexicographic Goal Programming [1226, 1227], Chebyshev Goal Programming [1228, 1229], Group-Goal Programming [1230], Fractional Goal Programming [1231, 1232], and Multi-Choice Goal Programming [1233, 1234] are also known.

Table 10.2: Related concepts of goal programming under uncertainty-aware models.

| $k$ | Related goal programming concept(s) |
|---|---|
| 1 | Fuzzy Goal Programming |
| 2 | Intuitionistic Fuzzy Goal Programming [1210, 1211] |
| 2 | Pythagorean Fuzzy Goal Programming [1212, 1213] |
| 3 | Neutrosophic Goal Programming [1214, 1215] |
| 3 | Picture Fuzzy Goal Programming [1216] |
| 3 | Spherical Fuzzy Goal Programming [1217] |

## 10.2   Fuzzy DEA (Fuzzy Data Envelopment Analysis)

DEA evaluates relative efficiency of decision-making units using linear programming, comparing multiple inputs and outputs against an empirical frontier simultaneously [1235, 1236]. Fuzzy DEA extends DEA by modeling inputs, outputs, or data uncertainty as fuzzy numbers, yielding possibilistic efficiency ranges robustly overall [1237, 1238].

**Definition 10.2.1** (Fuzzy Data Envelopment Analysis (Fuzzy DEA / FDEA))**.** [1239, 1240] Fix integers $n \geq 1$ (DMUs), $m \geq 1$ (inputs), and $s \geq 1$ (outputs). Let $\mathcal{J} := \{1, \ldots, n\}$ be the index set of decision making units (DMUs). For each $j \in \mathcal{J}$, DMU $j$ consumes $m$ inputs and produces $s$ outputs.

**Fuzzy observations.** For every input $i \in \{1, \ldots, m\}$ and DMU $j \in \mathcal{J}$, the input is a fuzzy number $\tilde{x}_{ij} \in \mathcal{F}(\mathbb{R}_+)$. For every output $r \in \{1, \ldots, s\}$ and DMU $j \in \mathcal{J}$, the output is a fuzzy number $\tilde{y}_{rj} \in \mathcal{F}(\mathbb{R}_+)$. (Thus the classical crisp data $(x_{ij}, y_{rj})$ are replaced by fuzzy input/output variables.)

**Underlying crisp DEA map (CCR, input-oriented).** Given any *crisp realization* $X = (x_{ij}) \in \mathbb{R}_+^{m \times n}$ and $Y = (y_{rj}) \in \mathbb{R}_+^{s \times n}$, define the (input-oriented) CCR efficiency of DMU $p \in \mathcal{J}$ as

$$\theta_p(X, Y) := \min_{\theta, \lambda} \; \theta$$

$$\text{s.t.} \quad \sum_{j=1}^n \lambda_j x_{ij} \leq \theta \, x_{ip} \; (i = 1, \ldots, m),$$

$$\sum_{j=1}^n \lambda_j y_{rj} \geq y_{rp} \; (r = 1, \ldots, s), \quad \lambda_j \geq 0 \; (j = 1, \ldots, n),$$

where $\lambda = (\lambda_1, \ldots, \lambda_n)$ are the intensity variables. (Other standard DEA variants such as BCC/VRS are obtained by adding $\sum_{j=1}^n \lambda_j = 1$.)

**Fuzzy efficiency (extension principle).** The *fuzzy CCR efficiency* of DMU $p$ is the fuzzy number $\tilde{\theta}_p \in \mathcal{F}((0, 1])$ induced from the fuzzy data $\tilde{X} = (\tilde{x}_{ij})$ and $\tilde{Y} = (\tilde{y}_{rj})$ by Zadeh's extension principle, i.e. $\tilde{\theta}_p := \text{Ext}(\theta_p)(\tilde{X}, \tilde{Y})$.

Equivalently (and most commonly in FDEA practice), $\tilde{\theta}_p$ is characterized by $\alpha$-cuts: for each $\alpha \in [0, 1]$, define the $\alpha$-cut intervals

$$(\tilde{x}_{ij})_\alpha = [x_{ij}^L(\alpha), x_{ij}^U(\alpha)], \qquad (\tilde{y}_{rj})_\alpha = [y_{rj}^L(\alpha), y_{rj}^U(\alpha)],$$

and the corresponding set of admissible crisp datasets

$$\mathcal{D}_\alpha := \Big(\prod_{i,j}[x_{ij}^L(\alpha), x_{ij}^U(\alpha)]\Big) \times \Big(\prod_{r,j}[y_{rj}^L(\alpha), y_{rj}^U(\alpha)]\Big).$$

Then the $\alpha$-cut of the fuzzy efficiency is the interval

$$(\tilde{\theta}_p)_\alpha = [\theta_p^L(\alpha), \theta_p^U(\alpha)], \qquad \theta_p^L(\alpha) := \inf_{(X,Y)\in\mathcal{D}_\alpha} \theta_p(X,Y), \quad \theta_p^U(\alpha) := \sup_{(X,Y)\in\mathcal{D}_\alpha} \theta_p(X,Y).$$

Hence FDEA yields, for each $\alpha$, an *interval efficiency score* rather than a single crisp score, by transforming a fuzzy DEA model into crisp linear programs on $\alpha$-cut intervals.

Finally, the membership function of $\tilde{\theta}_p$ can be recovered from $\alpha$-cuts via

$$\mu_{\tilde{\theta}_p}(t) = \sup\{\alpha \in [0,1] \ : \ t \in [\theta_p^L(\alpha), \theta_p^U(\alpha)]\}.$$

**Fuzzy DEA (as a method).** The collection $\{\tilde{\theta}_p\}_{p\in\mathcal{J}}$ is called the *Fuzzy DEA evaluation* of the $n$ DMUs. Ranking/selection is then performed by a chosen fuzzy-number ranking/defuzzification rule applied to $\tilde{\theta}_p$ (or by comparing the $\alpha$-level intervals).

Using an Uncertain Set, the definition of Uncertain DEA of type M is given below.

**Definition 10.2.2** (Uncertain DEA of type $M$ (U-DEA), CCR input-oriented). Fix integers $n \geq 1$ (DMUs), $m \geq 1$ (inputs), and $s \geq 1$ (outputs). Let $\mathcal{J} := \{1,\ldots,n\}$ index the DMUs. Fix an uncertain model $M$ with $\mathrm{Dom}(M) \neq \emptyset$ and an admissible positive score $S_M$.

**Uncertain observations.** For each input $i = 1,\ldots,m$ and DMU $j \in \mathcal{J}$, assume an uncertain input degree $x_{ij}^{(M)} \in \mathrm{Dom}(M)$; for each output $r = 1,\ldots,s$ and DMU $j \in \mathcal{J}$, assume an uncertain output degree $y_{rj}^{(M)} \in \mathrm{Dom}(M)$.

Define the induced positive crisp data by scoring:

$$x_{ij} := S_M(x_{ij}^{(M)}) \in (0,\infty), \qquad y_{rj} := S_M(y_{rj}^{(M)}) \in (0,\infty).$$

**CCR efficiency (envelopment form).** For a fixed DMU $p \in \mathcal{J}$, define its (input-oriented) CCR efficiency as the optimal value

$$\theta_p := \min_{\theta,\lambda} \ \theta$$

subject to

$$\sum_{j=1}^n \lambda_j x_{ij} \leq \theta \, x_{ip} \quad (i=1,\ldots,m), \qquad \sum_{j=1}^n \lambda_j y_{rj} \geq y_{rp} \quad (r=1,\ldots,s), \qquad \lambda_j \geq 0 \ (j=1,\ldots,n), \qquad \theta \geq 0.$$

The collection $\{\theta_p\}_{p\in\mathcal{J}}$ is called the *U-DEA (CCR) efficiency evaluation.* (For the BCC/VRS model add $\sum_{j=1}^n \lambda_j = 1$.)

**Theorem 10.2.3** (Well-definedness and feasibility of U-DEA). *Under Definition 10.2.2, assume $S_M$ is admissible positive and hence all induced data satisfy $x_{ij} > 0$ and $y_{rj} > 0$. Then for each $p \in \mathcal{J}$:*

*(i) The CCR feasible set is nonempty, so $\theta_p$ is well-defined as an extended real number.*

*(ii) The optimal value satisfies $0 \leq \theta_p \leq 1$.*

*(iii) In particular, $\theta_p$ is a finite real number (hence U-DEA is well-defined).*

*Proof.* (i) Feasibility: choose $\lambda_p := 1$ and $\lambda_j := 0$ for $j \neq p$, and set $\theta := 1$. Then for each input $i$,

$$\sum_{j=1}^{n} \lambda_j x_{ij} = x_{ip} \leq 1 \cdot x_{ip},$$

and for each output $r$,

$$\sum_{j=1}^{n} \lambda_j y_{rj} = y_{rp} \geq y_{rp}.$$

Also $\lambda_j \geq 0$ and $\theta \geq 0$. Hence the feasible set is nonempty.

(ii) Since $\theta \geq 0$ is a constraint, $\theta_p \geq 0$. Because $(\theta, \lambda) = (1, e_p)$ is feasible, the minimum satisfies $\theta_p \leq 1$.

(iii) Combining (i) and (ii), the optimal value exists and obeys $0 \leq \theta_p \leq 1$, so it is finite and real. $\square$

For reference, related concepts of data envelopment analysis under uncertainty-aware models are listed in Table 10.3.

Table 10.3: Related concepts of data envelopment analysis under uncertainty-aware models.

| $k$ | **Related data envelopment analysis concept(s)** |
|---|---|
| 2 | Intuitionistic Fuzzy Data Envelopment Analysis [1241, 1242] |
| 2 | Bipolar Fuzzy Data Envelopment Analysis [1243, 1244] |
| 2 | Pythagorean Fuzzy Data Envelopment Analysis [1245] |
| 3 | Hesitant Fuzzy Data Envelopment Analysis [1246] |
| 3 | Spherical Fuzzy Data Envelopment Analysis [1247, 1248] |
| 3 | Neutrosophic Data Envelopment Analysis [1249–1251] |
| $n$ | Plithogenic Data Envelopment Analysis [1252–1254] |

Moreover, besides Uncertain Data Envelopment Analysis, a very large number of other related concepts are also known, including Two-Stage Data Envelopment Analysis [1255, 1256], Three-Stage Data Envelopment Analysis [1257, 1258], Four-Stage Data Envelopment Analysis [1259], Global Data Envelopment Analysis [1260, 1261], Rough Data Envelopment Analysis [1262, 1263], Grey Data Envelopment Analysis [1264, 1265], and Multi-Criteria Data Envelopment Analysis [1266, 1267].

### 10.3   Fuzzy Social Choice

Social choice aggregates individual preferences into collective decisions, using voting or welfare rules while addressing fairness, strategy, and consistency [1268]. Fuzzy social choice uses graded preference relations, aggregates them into a collective fuzzy relation, then selects an alternative via defuzzification [1269–1271].

**Definition 10.3.1** (Fuzzy social choice (via a fuzzy social choice function)). (cf. [1269–1271]) Let $X = \{x_1, \ldots, x_m\}$ be a finite set of alternatives with $|X| \geq 3$, and let $N = \{1, \ldots, n\}$ be a finite set of individuals with $n \geq 2$. Fix a nonempty domain $T$ of admissible *fuzzy preference relations* on $X$.

**(i) Individual fuzzy preference relations.** For each $i \in N$, an individual preference is a fuzzy binary relation

$$R_i : X \times X \longrightarrow [0, 1],$$

where $R_i(x, y)$ is interpreted as the degree to which "$x$ is at least as preferred as $y$". A *preference profile* is the $n$-tuple

$$R^N := (R_1, \ldots, R_n) \in T^n.$$

(One common choice is to take $T = H$, the set of *fuzzy orderings*, i.e. relations that are reflexive, connected, and max–min transitive.)

**(ii) Fuzzy social choice function.** A *fuzzy social choice function* (FSCF) is a mapping

$$f : T^n \longrightarrow X$$

that assigns a *single* collective choice $f(R^N) \in X$ to each profile $R^N$ of individuals' fuzzy preference relations. The resulting collective alternative $f(R^N)$ is called the *fuzzy social choice*.

**Definition 10.3.2** (Uncertain preference relations of type $M$). Let $X$ be a finite set of alternatives with $|X| \geq 3$ and let $M$ be an uncertain model. An *uncertain preference relation of type $M$* on $X$ is a mapping

$$R_M : \ X \times X \longrightarrow \mathrm{Dom}(M),$$

where $R_M(x, y)$ encodes the (uncertain) degree to which "$x$ is at least as preferred as $y$". Given an admissible score $S_M : \mathrm{Dom}(M) \to [0, 1]$, its induced *scored preference relation* is

$$r(x, y) := S_M\big(R_M(x, y)\big) \in [0, 1].$$

We define Uncertain social choice function of type $M$ (U-SCF) as follows.

**Definition 10.3.3** (Uncertain social choice function of type $M$ (U-SCF)). Let $X$ be a finite set of alternatives with $|X| \geq 3$ and let $N = \{1, \ldots, n\}$ be voters with $n \geq 2$. Fix an uncertain model $M$ with $\mathrm{Dom}(M) \neq \emptyset$ and an admissible score $S_M : \mathrm{Dom}(M) \to [0, 1]$.

Let $T_M$ be a nonempty domain of admissible uncertain preference relations on $X$. A *preference profile* is an $n$-tuple

$$\mathbf{R} := (R_M^{(1)}, \ldots, R_M^{(n)}) \in (T_M)^n.$$

**Step 1 (Aggregate uncertain preferences to a collective scored relation).** Fix an aggregation operator

$$\mathrm{Agg} : \ [0, 1]^n \to [0, 1]$$

(e.g. weighted mean, OWA, min/max, or any monotone aggregator). For each pair $(x, y) \in X \times X$, define the collective scored relation

$$r^*(x, y) := \mathrm{Agg}\Big(S_M(R_M^{(1)}(x, y)), \dots, S_M(R_M^{(n)}(x, y))\Big) \in [0, 1].$$

**Step 2 (Social choice by a score-based rule).** Define the *collective dominance score* of an alternative $x \in X$ by

$$\mathrm{Score}(x) := \sum_{y \in X \setminus \{x\}} r^*(x, y) \ \in [0, m-1].$$

An *uncertain social choice function* (U-SCF) is the mapping

$$f : (T_M)^n \longrightarrow X, \qquad f(\mathbf{R}) \in \arg\max_{x \in X} \ \mathrm{Score}(x),$$

with a fixed tie-breaking convention if the argmax is not unique. The chosen alternative $f(\mathbf{R})$ is called the *uncertain social choice* (U-social choice).

**Theorem 10.3.4** (Well-definedness of U-social choice)**.** *Under Definition 10.3.3, assume:*

*(i)* $T_M \neq \emptyset$ *and* $\mathrm{Dom}(M) \neq \emptyset$;

*(ii)* $S_M : \mathrm{Dom}(M) \to [0, 1]$ *is admissible;*

*(iii)* $\mathrm{Agg} : [0, 1]^n \to [0, 1]$ *is well-defined;*

*(iv)* *a deterministic tie-breaking rule is fixed.*

*Then the mapping* $f : (T_M)^n \to X$ *is well-defined.*

*Proof.* Fix a profile $\mathbf{R} \in (T_M)^n$. For each $(x, y)$ and each voter $i$, $R_M^{(i)}(x, y) \in \mathrm{Dom}(M)$ by definition, so $S_M(R_M^{(i)}(x, y)) \in [0, 1]$ is well-defined. Hence the $n$-tuple of scored inputs to Agg lies in $[0, 1]^n$, so $r^*(x, y) = \mathrm{Agg}(\cdots) \in [0, 1]$ is well-defined for every $(x, y) \in X \times X$.

Therefore, for each $x \in X$, $\mathrm{Score}(x) = \sum_{y \neq x} r^*(x, y)$ is a finite sum of numbers in $[0, 1]$, hence a well-defined real number. Since $X$ is finite, the set $\{\mathrm{Score}(x) : x \in X\}$ attains a maximum, so $\arg\max_{x \in X} \mathrm{Score}(x) \neq \emptyset$. With a fixed tie-breaking convention, $f(\mathbf{R})$ is uniquely determined. Thus $f$ is well-defined on $(T_M)^n$. $\quad\square$

## 10.4 Fuzzy decision tables

Decision tables map combinations of condition states to actions, enabling transparent rule-based decisions, consistency checks, and maintenance in complex systems [1272]. Fuzzy decision tables extend decision tables with fuzzy sets and linguistic terms, handling uncertainty via graded matching and t-norm aggregation [1273, 1274].

**Definition 10.4.1** (Fuzzy decision table (FDT)). [1273, 1274] Let $c \geq 1$ and $a \geq 1$. For each condition index $i \in \{1, \ldots, c\}$ let $CS_i$ be a *condition subject* with *condition domain* $CD_i$ (universe of discourse). For each action index $j \in \{1, \ldots, a\}$ let $AS_j$ be an *action subject* with *action domain* $AD_j$.

**(1) Fuzzy condition states.** For each $i$, fix a finite nonempty set of linguistic *condition states*

$$CT_i = \{\widetilde{C}_{i1}, \ldots, \widetilde{C}_{in_i}\}, \qquad n_i \geq 1,$$

where each $\widetilde{C}_{ik}$ is a fuzzy set on $CD_i$, i.e.

$$\widetilde{C}_{ik} : CD_i \to [0, 1].$$

Write $CT := \{CT_1, \ldots, CT_c\}$.

**(2) Fuzzy action states (two standard forms).**

*Form 1 (crisp execution flags, fuzzy only in conditions).* For each $j$, let

$$AV_j := \{\text{true}\,(x), \text{false}\,(-), \text{nil}\,(\cdot)\},$$

and define the action-configuration space $AV := \prod_{j=1}^{a} AV_j$.

*Form 2 (fuzzy/linguistic action values allowed).* For each $j$, fix a finite nonempty set of *action states*

$$AT_j = \{\widetilde{A}_{j1}, \ldots, \widetilde{A}_{jm_j}\}, \qquad m_j \geq 1,$$

where each $\widetilde{A}_{j\ell}$ is a fuzzy set on $AD_j$: $\widetilde{A}_{j\ell} : AD_j \to [0, 1]$. Equivalently, one may define an action-value set

$$AV_j := \{\,\widetilde{a} \mid \widetilde{a} \in \mathcal{F}(AD_j)\,\} \quad \text{or a designated linguistic subset of it,}$$

and again set $AV := \prod_{j=1}^{a} AV_j$.

**(3) Fuzzy decision table as a single-hit function.** A *fuzzy decision table* is a (single-valued) mapping

$$F : CT_1 \times \cdots \times CT_c \longrightarrow AV$$

such that *each possible condition-state combination is mapped into exactly one action configuration.* Equivalently, viewing condition/action combinations as entries, an FDT is a function from the condition-entry space to the action-entry space:

$$F : SPACE(C) \to SPACE(A),$$

where $SPACE(C) := CT_1 \times \cdots \times CT_c$ and $SPACE(A) := AV$ (Form 1 or Form 2).

**(4) Rule (column) interpretation.** Each column of the table corresponds to a rule

**IF** $CS_1$ is $\widetilde{C}_{1k_1} \wedge \cdots \wedge CS_c$ is $\widetilde{C}_{ck_c}$ **THEN** $(AS_1, \ldots, AS_a)$ takes configuration $F(\widetilde{C}_{1k_1}, \ldots, \widetilde{C}_{ck_c})$, where the antecedent is evaluated by a chosen $t$-norm (e.g. min) when consulting the table.

Using an Uncertain Set, we define Uncertain decision table of type $M$ (U-DT) as follows.

**Definition 10.4.2** (Uncertain decision table of type $M$ (U-DT))**.** Let $c \geq 1$ and $a \geq 1$.

**Subjects and domains.** For each condition index $i \in \{1, \ldots, c\}$, let $CS_i$ be a condition subject with a (nonempty) condition domain $CD_i$. For each action index $j \in \{1, \ldots, a\}$, let $AS_j$ be an action subject with an action domain $AD_j$.

**Uncertain condition states (linguistic states).** Fix an uncertain model $M$ with $\mathrm{Dom}(M) \neq \emptyset$ and an admissible score $S_M$. For each condition $i$, fix a finite nonempty set of linguistic condition states

$$CT_i = \{C_{i1}^{(M)}, \ldots, C_{in_i}^{(M)}\}, \qquad n_i \geq 1,$$

where each state is an uncertain set (degree map) on $CD_i$:

$$C_{ik}^{(M)} : \ CD_i \to \mathrm{Dom}(M).$$

Its induced (scored) membership is

$$\mu_{ik}(t) := S_M\big(C_{ik}^{(M)}(t)\big) \in [0, 1], \qquad t \in CD_i.$$

**Action values.** Choose either of the following common forms.

*Form 1 (crisp execution flags).* For each $j$, let

$$AV_j := \{\mathrm{true}, \mathrm{false}, \mathrm{nil}\}, \qquad AV := \prod_{j=1}^{a} AV_j.$$

*Form 2 (uncertain/linguistic action values).* For each $j$, fix a finite nonempty set of action states

$$AT_j = \{A_{j1}^{(M)}, \ldots, A_{jm_j}^{(M)}\}, \qquad m_j \geq 1,$$

where each $A_{j\ell}^{(M)} : AD_j \to \mathrm{Dom}(M)$ is an uncertain set on $AD_j$. Set $AV_j := AT_j$ and $AV := \prod_{j=1}^{a} AV_j$.

**Decision table as a single-hit mapping.** Define the condition-entry space

$$SPACE(C) := CT_1 \times \cdots \times CT_c.$$

An *uncertain decision table* (U-DT) is a mapping

$$F : \ SPACE(C) \longrightarrow AV,$$

so that each condition-state combination is mapped to exactly one action configuration.

**Consultation semantics (graded matching).** Fix a $t$-norm $T : [0,1]^c \to [0,1]$ (e.g. min or product). For an observed input $o = (t_1, \ldots, t_c) \in CD_1 \times \cdots \times CD_c$ and a rule-entry $(C_{1k_1}^{(M)}, \ldots, C_{ck_c}^{(M)}) \in SPACE(C)$, define its firing strength by

$$\alpha_{k_1,\ldots,k_c}(o) := T\big(\mu_{1k_1}(t_1), \ldots, \mu_{ck_c}(t_c)\big) \in [0,1].$$

A standard deterministic selection rule is to choose a maximizer

$$(k_1^\star, \ldots, k_c^\star) \in \arg \max_{(k_1,\ldots,k_c)} \alpha_{k_1,\ldots,k_c}(o),$$

with a fixed tie-breaking convention, and output the action

$$\mathrm{Act}(o) := F\big(C_{1k_1^\star}^{(M)}, \ldots, C_{ck_c^\star}^{(M)}\big) \in AV.$$

**Theorem 10.4.3** (Well-definedness of U-DT and its consultation rule)**.** *Under Definition 10.4.2, assume:*

*(i) $\mathrm{Dom}(M) \neq \emptyset$ and $S_M : \mathrm{Dom}(M) \to [0,1]$ is admissible;*

*(ii) each $CT_i$ is finite and nonempty, and $AV$ is nonempty;*

*(iii) the $t$-norm $T$ is a well-defined map $[0,1]^c \to [0,1]$;*

*(iv) a deterministic tie-breaking convention is fixed for $\arg \max$.*

*Then:*

*(a) The mapping $F : SPACE(C) \to AV$ is well-defined as a function (single-hit table).*

*(b) For every observation $o \in CD_1 \times \cdots \times CD_c$, all firing strengths $\alpha_{k_1,\ldots,k_c}(o)$ are well-defined in $[0,1]$.*

*(c) The consultation output $\mathrm{Act}(o) \in AV$ is well-defined for every observation $o$.*

*Proof.* (a) By definition, $SPACE(C)$ is a Cartesian product of nonempty sets, hence nonempty, and $F$ is assumed to be a single-valued mapping into $AV$.

(b) Fix $o = (t_1, \ldots, t_c)$. For any rule-entry $(k_1, \ldots, k_c)$, each $\mu_{ik_i}(t_i) = S_M(C_{ik_i}^{(M)}(t_i)) \in [0,1]$ is well-defined by admissibility of $S_M$. Since $T$ maps $[0,1]^c$ to $[0,1]$, $\alpha_{k_1,\ldots,k_c}(o) \in [0,1]$ is well-defined.

(c) The set of all indices $(k_1, \ldots, k_c)$ is finite because each $CT_i$ is finite; hence the maximum of the finite set of values $\{\alpha_{k_1,\ldots,k_c}(o)\}$ is attained, so $\arg \max$ is nonempty. With a fixed tie-breaking rule, a unique maximizer is selected, and then $\mathrm{Act}(o)$ is defined by applying $F$ to that entry. $\square$

Related concepts of decision tables under uncertainty-aware models are listed in Table 10.4.

Table 10.4: Related concepts of decision tables under uncertainty-aware models.

| $k$ | Related decision table concept(s) |
|---|---|
| 1 | Fuzzy Decision Tables |
| 2 | Intuitionistic Fuzzy Decision Tables [1275, 1276] |
| 3 | Neutrosophic Decision Tables [1277, 1278] |

## 10.5 Fuzzy decision matrices

Decision matrices tabulate alternatives versus criteria or states, record payoffs or ratings, and support selection via scoring, weighting, dominance checks, or expected-value calculations [1279, 1280]. Fuzzy decision matrices replace payoffs or ratings with fuzzy numbers or fuzzy events, propagate uncertainty through fuzzy arithmetic or $\alpha$-cuts, and rank alternatives robustly [1281–1283].

**Definition 10.5.1** (Fuzzy decision matrix (with fuzzy states of the world and fuzzy evaluations))**.** Let $(\Omega, \mathcal{A}, P)$ be a probability space. Let $X = \{x_1, \ldots, x_n\}$ be a finite set of alternatives and let $m \geq 1$.

**(1) Fuzzy states of the world.** For $j = 1, \ldots, m$, let $S_j$ be a *fuzzy event* on $\Omega$, i.e. a fuzzy set $S_j \in \mathcal{F}(\Omega)$ with membership function $\mu_{S_j} : \Omega \to [0, 1]$ whose $\alpha$-cuts $(S_j)_\alpha = \{\omega \in \Omega : \mu_{S_j}(\omega) \geq \alpha\}$ belong to $\mathcal{A}$ (for all $\alpha \in (0, 1]$). Assume that $(S_1, \ldots, S_m)$ forms a *fuzzy partition* of $\Omega$:

$$\sum_{j=1}^{m} \mu_{S_j}(\omega) = 1 \qquad (\forall\, \omega \in \Omega).$$

**(2) Zadeh probabilities of fuzzy states.** Define the (Zadeh) probability of a fuzzy event $A$ by

$$P_Z(A) := \int_\Omega \mu_A(\omega)\, dP(\omega),$$

and set

$$p_{Zj} := P_Z(S_j) \qquad (j = 1, \ldots, m).$$

**(3) Fuzzy evaluations.** For each alternative $x_i$ and each fuzzy state $S_j$, let $H_{i,j}$ be a fuzzy number representing the evaluation (consequence) of choosing $x_i$ when $S_j$ occurs. Thus $H_{i,j} \in \mathcal{F}_N(\mathbb{R})$ and its $\alpha$-cut is an interval

$$(H_{i,j})_\alpha = \left[ h_{i,j}^L(\alpha),\, h_{i,j}^U(\alpha) \right] \qquad (\alpha \in (0, 1]).$$

**(4) The fuzzy decision matrix.** The *fuzzy decision matrix* is the data structure

$$\mathcal{D}_F = \Big( X,\ (S_1, \ldots, S_m),\ (p_{Z1}, \ldots, p_{Zm}),\ (H_{i,j})_{1 \leq i \leq n,\, 1 \leq j \leq m} \Big),$$

often displayed as a table whose columns are the fuzzy states $S_j$ (with probabilities $p_{Zj}$) and whose entries are the fuzzy evaluations $H_{i,j}$.

**(5) Induced fuzzy characteristics (expected value and variance).** Using fuzzy arithmetic (e.g. via $\alpha$-cuts), the *fuzzy expected evaluation* of $x_i$ is

$$(F)\mathbb{E}H_i := \sum_{j=1}^m p_{Zj} \cdot H_{i,j},$$

so that for each $\alpha \in (0,1]$,

$$\big((F)\mathbb{E}H_i\big)_\alpha = \left[\sum_{j=1}^m p_{Zj}\, h_{i,j}^L(\alpha),\ \sum_{j=1}^m p_{Zj}\, h_{i,j}^U(\alpha)\right].$$

A commonly used (but dependency-ignoring) fuzzy variance is

$$(F)\mathrm{var}^t(H_i) := \sum_{j=1}^m p_{Zj}\left(H_{i,j} - (F)\mathbb{E}H_i\right)^2.$$

A dependency-respecting variance can be defined by $\alpha$-cuts through optimization: for $\alpha \in (0,1]$,

$$(F)\mathrm{var}^s(H_i)_\alpha = \left[\mathrm{var}_i^{s,L}(\alpha),\ \mathrm{var}_i^{s,U}(\alpha)\right],$$

where

$$\mathrm{var}_i^{s,L}(\alpha) = \min\left\{\sum_{j=1}^m p_{Zj}\left(h_{i,j} - \sum_{k=1}^m p_{Zk}h_{i,k}\right)^2 \,\middle|\, h_{i,j} \in (H_{i,j})_\alpha,\ j = 1,\dots,m\right\},$$

$$\mathrm{var}_i^{s,U}(\alpha) = \max\left\{\sum_{j=1}^m p_{Zj}\left(h_{i,j} - \sum_{k=1}^m p_{Zk}h_{i,k}\right)^2 \,\middle|\, h_{i,j} \in (H_{i,j})_\alpha,\ j = 1,\dots,m\right\}.$$

**Definition 10.5.2** (Uncertain decision matrix of type $M$ (U-DM)). Let $\mathcal{A} = \{A_1, \dots, A_m\}$ be a finite set of alternatives and $\mathcal{C} = \{C_1, \dots, C_n\}$ a finite set of criteria, with $m, n \geq 1$. Fix an uncertain model $M$ with $\mathrm{Dom}(M) \neq \emptyset$ and an admissible score $S_M$.

An *uncertain decision matrix of type $M$* is a triple

$$\widetilde{X}^{(M)} = \big(\mathcal{A}, \mathcal{C}, X^{(M)}\big), \qquad X^{(M)} = \big(x_{ij}^{(M)}\big)_{m \times n}, \qquad x_{ij}^{(M)} \in \mathrm{Dom}(M),$$

where $x_{ij}^{(M)}$ encodes the uncertain evaluation (performance/payoff) of alternative $A_i$ under criterion $C_j$.

The induced *scored (crisp) decision matrix* is

$$X = \big(x_{ij}\big)_{m \times n}, \qquad x_{ij} := S_M(x_{ij}^{(M)}) \in \mathbb{R}.$$

If $S_M$ is admissible positive, then $X \in (0, \infty)^{m \times n}$.

**Definition 10.5.3** (Basic transformations on a U-DM). Let $\widetilde{X}^{(M)} = (\mathcal{A}, \mathcal{C}, X^{(M)})$ be a U-DM and let $X = S_M(X^{(M)})$ be its scored matrix.

- A *weight vector* is any $w = (w_1, \dots, w_n)$ with $w_j \geq 0$ and $\sum_{j=1}^n w_j = 1$.

- A *benefit/cost orientation* is a partition $\mathcal{C} = \mathcal{C}^{\text{ben}} \dot{\cup} \mathcal{C}^{\text{cost}}$, often converted to "larger is better" by replacing $x_{ij}$ with $-x_{ij}$ on $\mathcal{C}^{\text{cost}}$.

- A *normalized matrix* $R = (r_{ij})$ is any transformation $R = \mathcal{N}(X)$ such that $r_{ij}$ is finite (common choices: min–max, vector normalization, ratio normalization).

**Theorem 10.5.4** (Well-definedness of uncertain decision matrices). *Under Definition 10.5.2, assume* $\text{Dom}(M) \neq \emptyset$ *and* $S_M$ *is admissible. Then:*

(i) *The scored decision matrix* $X = (S_M(x_{ij}^{(M)}))$ *is well-defined and belongs to* $\mathbb{R}^{m \times n}$.

(ii) *Any finite aggregation of entries of* $X$ *(e.g. weighted sums* $\sum_j w_j x_{ij}$ *or pairwise differences* $x_{ij} - x_{kj}$*) is well-defined as a real number.*

(iii) *If* $S_M$ *is admissible positive, then all ratio-based normalizations of the form*

$$r_{ij} = \frac{x_{ij}}{\max_p x_{pj}}, \qquad r_{ij} = \frac{\min_p x_{pj}}{x_{ij}}, \qquad r_{ij} = \frac{x_{ij}}{\sqrt{\sum_p x_{pj}^2}}$$

*are well-defined (denominators are strictly positive).*

*Proof.* (i) Since $x_{ij}^{(M)} \in \text{Dom}(M)$ and $S_M$ is admissible, each $x_{ij} = S_M(x_{ij}^{(M)})$ is a finite real number. Thus $X \in \mathbb{R}^{m \times n}$ is well-defined.

(ii) Finite sums and differences of real numbers are well-defined, so any finite aggregation (including weighted sums with $w_j \geq 0$ and $\sum_j w_j = 1$) is well-defined.

(iii) If $S_M$ is admissible positive, then $x_{ij} > 0$ for all $i, j$. Hence $\max_p x_{pj} > 0$, $\min_p x_{pj} > 0$, and $\sum_p x_{pj}^2 > 0$ for each fixed $j$, so all displayed ratio-based normalizations are well-defined. $\qquad\square$

## 10.6   Fuzzy Ultrafilter

An ultrafilter is a maximal filter on a set: for every subset, either it or its complement belongs to the ultrafilter, never both [1284]. An ultrafilter relates to decision-making by modeling decisive coalitions in collective choice: under Arrow-type axioms, the sets able to determine social outcomes form an ultrafilter. Fuzzy ultrafilters assign each subset a membership degree in $[0, 1]$, preserving monotonicity and intersection, with maximal top-level acceptance (cf. [1285–1287]).

**Definition 10.6.1** (Filter and Ultrafilter). (cf. [1284, 1288, 1289]) Let $X$ be a nonempty set.

A *(classical) filter* on $X$ is a nonempty family $\mathcal{F} \subseteq \mathcal{P}(X)$ such that

1. $\varnothing \notin \mathcal{F}$ and $X \in \mathcal{F}$;

2. (Upward closed) if $A \in \mathcal{F}$ and $A \subseteq B \subseteq X$, then $B \in \mathcal{F}$;

   3. (Closed under finite intersections) if $A, B \in \mathcal{F}$, then $A \cap B \in \mathcal{F}$.

A *(classical) ultrafilter* on $X$ is a filter $\mathcal{U}$ on $X$ which is maximal under inclusion: whenever $\mathcal{F}$ is a filter on $X$ with $\mathcal{U} \subseteq \mathcal{F}$, one has $\mathcal{F} = \mathcal{U}$.

Equivalently, a filter $\mathcal{U}$ on $X$ is an ultrafilter if and only if it satisfies the *ultrafilter dichotomy*:

$$\forall A \subseteq X, \qquad A \in \mathcal{U} \ \text{ or } \ X \setminus A \in \mathcal{U}.$$

**Definition 10.6.2** (Principal Ultrafilter). Let $X$ be a nonempty set and fix a point $x \in X$. The *principal ultrafilter at $x$* is the family

$$\mathcal{U}_x := \{A \subseteq X : x \in A\}.$$

Equivalently, $\mathcal{U}_x$ is the unique ultrafilter on $X$ satisfying

$$A \in \mathcal{U}_x \iff x \in A \qquad (\forall A \subseteq X).$$

Thus $\mathcal{U}_x$ expresses the policy "select the specific element $x$ throughout," because every set is accepted exactly when it contains $x$.

**Definition 10.6.3** (Fuzzy Ultrafilter on a Set). (cf. [1285–1287]) Let $X$ be a nonempty set. A *fuzzy ultrafilter* on $X$ is a map

$$\widetilde{\mathcal{U}} : \mathcal{P}(X) \longrightarrow [0,1]$$

satisfying, for all $A, B \subseteq X$:

   1. (Normalization) $\widetilde{\mathcal{U}}(X) = 1$ and $\widetilde{\mathcal{U}}(\varnothing) = 0$.

   2. (Monotonicity) If $A \subseteq B$, then $\widetilde{\mathcal{U}}(A) \leq \widetilde{\mathcal{U}}(B)$.

   3. (Meet as intersection) $\widetilde{\mathcal{U}}(A \cap B) = \min\{\widetilde{\mathcal{U}}(A), \widetilde{\mathcal{U}}(B)\}$.

   4. (Ultrafilter maximality) $\max\{\widetilde{\mathcal{U}}(A), \widetilde{\mathcal{U}}(X \setminus A)\} = 1$.

**Definition 10.6.4** (Principal Fuzzy Ultrafilter). [1290] Let $X$ be a nonempty set and fix $x \in X$. The *principal fuzzy ultrafilter at $x$* is the fuzzy ultrafilter

$$\widetilde{\mathcal{U}}_x : \mathcal{P}(X) \longrightarrow [0,1], \qquad \widetilde{\mathcal{U}}_x(A) := \begin{cases} 1, & x \in A, \\ 0, & x \notin A. \end{cases}$$

In particular,

$$\widetilde{\mathcal{U}}_x(A) = 1 \iff x \in A,$$

so $\widetilde{\mathcal{U}}_x$ encodes "a fixed element $x$ is (fully) selected" in a fuzzy-valued format.

**Definition 10.6.5** (Uncertain ultrafilter of type $M$). Let $X$ be a nonempty set and fix an uncertain model $M$ with $\mathrm{Dom}(M) \neq \emptyset$. Fix an admissible score $S_M : \mathrm{Dom}(M) \to [0,1]$.

An *uncertain ultrafilter of type $M$ (relative to $S_M$)* is a mapping

$$\mathcal{U}_M : \mathcal{P}(X) \longrightarrow \mathrm{Dom}(M)$$

whose *scored acceptance* map

$$u : \mathcal{P}(X) \to [0,1], \qquad u(A) := S_M\big(\mathcal{U}_M(A)\big)$$

satisfies, for all $A, B \subseteq X$:

(U1) **(Normalization)** $u(X) = 1$ and $u(\varnothing) = 0$.

(U2) **(Monotonicity)** $A \subseteq B \ \Rightarrow \ u(A) \leq u(B)$.

(U3) **(Meet = intersection)** $u(A \cap B) = \min\{u(A), u(B)\}$.

(U4) **(Ultrafilter dichotomy at level** 1**)** $\max\{u(A), u(X \setminus A)\} = 1$.

**Theorem 10.6.6** (Well-definedness and induced crisp ultrafilter). *Let $\mathcal{U}_M : \mathcal{P}(X) \to \mathrm{Dom}(M)$ satisfy Definition 10.6.5. Then:*

(i) *The scored acceptance map $u = S_M \circ \mathcal{U}_M$ is well-defined and satisfies $u(A) \in [0, 1]$ for all $A \subseteq X$.*

(ii) *The family*
$$\mathcal{F}_1 := \{A \subseteq X : \ u(A) = 1\}$$
*is a (classical) ultrafilter on $X$.*

*Proof.* (i) Since $S_M$ is admissible, $S_M(d) \in [0, 1]$ for every $d \in \mathrm{Dom}(M)$; hence $u(A) = S_M(\mathcal{U}_M(A)) \in [0, 1]$ for all $A$.

(ii) First, $\varnothing \notin \mathcal{F}_1$ because $u(\varnothing) = 0$ by (U1), and $X \in \mathcal{F}_1$ because $u(X) = 1$. If $A \in \mathcal{F}_1$ and $A \subseteq B$, then $u(B) \geq u(A) = 1$ by (U2), hence $u(B) = 1$ and $B \in \mathcal{F}_1$ (upward closed). If $A, B \in \mathcal{F}_1$, then by (U3),
$$u(A \cap B) = \min\{u(A), u(B)\} = \min\{1, 1\} = 1,$$
so $A \cap B \in \mathcal{F}_1$ (closed under finite intersections). Thus $\mathcal{F}_1$ is a filter.

To show maximality, let $A \subseteq X$. By (U4), $\max\{u(A), u(X \setminus A)\} = 1$, so either $u(A) = 1$ or $u(X \setminus A) = 1$, i.e., $A \in \mathcal{F}_1$ or $X \setminus A \in \mathcal{F}_1$. Hence $\mathcal{F}_1$ satisfies the ultrafilter dichotomy, and therefore is a classical ultrafilter. □

**Definition 10.6.7** (Principal uncertain ultrafilter). Let $X$ be a nonempty set, fix $x \in X$, and fix an uncertain model $M$ with an admissible score $S_M$. Assume there exist two degree tuples $d^1, d^0 \in \mathrm{Dom}(M)$ such that
$$S_M(d^1) = 1, \qquad S_M(d^0) = 0.$$
Define $\mathcal{U}_{M,x} : \mathcal{P}(X) \to \mathrm{Dom}(M)$ by
$$\mathcal{U}_{M,x}(A) := \begin{cases} d^1, & x \in A, \\ d^0, & x \notin A. \end{cases}$$
Then $\mathcal{U}_{M,x}$ is called the *principal uncertain ultrafilter at $x$*.

**Proposition 10.6.8** (Well-definedness of the principal uncertain ultrafilter). *Under Definition 10.6.7, $\mathcal{U}_{M,x}$ is an uncertain ultrafilter of type $M$.*

*Proof.* Let $u(A) = S_M(\mathcal{U}_{M,x}(A)) \in \{0, 1\}$. Then $u(X) = 1$, $u(\varnothing) = 0$. If $A \subseteq B$ and $u(A) = 1$ then $x \in A \subseteq B$ so $u(B) = 1$; otherwise $u(A) = 0 \leq u(B)$, proving monotonicity. Also $u(A \cap B) = 1$ iff $x \in A$ and $x \in B$, i.e. iff $u(A) = u(B) = 1$, so $u(A \cap B) = \min\{u(A), u(B)\}$. Finally, for any $A$, either $x \in A$ or $x \in X \setminus A$, hence $\max\{u(A), u(X \setminus A)\} = 1$. Thus (U1)–(U4) hold. □

Related concepts of ultrafilters under uncertainty-aware models are listed in Table 10.5.

Besides Uncertain Ultrafilters, several other related concepts are also known, including Weak Ultrafilters [1299, 1300], Soft Ultrafilters [1288, 1289], and HyperSoft Ultrafilters [1301].

Table 10.5: Related concepts of ultrafilters under uncertainty-aware models.

| $k$ | Related ultrafilter concept(s) |
|---|---|
| 2 | Intuitionistic Fuzzy Ultrafilter [1291–1293] |
| 3 | Neutrosophic Ultrafilter [1294–1297] |
| 3 | Picture Fuzzy Ultrafilter [1298] |
| $n$ | Plithogenic Ultrafilter [1297] |

## 10.7 Fuzzy SWOT Analysis

SWOT analysis identifies strengths, weaknesses, opportunities, and threats, organizes internal-external factors, and derives strategies for planning decisions [1302, 1303]. Fuzzy SWOT analysis represents SWOT factors with fuzzy numbers or linguistic terms, aggregates uncertainties, and prioritizes strategies using fuzzy ranking [1304, 1305]. SWOT supports decision-making by structuring internal and external factors, generating strategic alternatives, prioritizing options with weights/scores, and justifying choices transparently under uncertainty.

**Definition 10.7.1** (Fuzzy SWOT Analysis (FSWOT)). [1304, 1305] Let $I = \{I_1, \ldots, I_{n_i}\}$ be the set of *internal factors* and $E = \{E_1, \ldots, E_{n_e}\}$ the set of *external factors*. Fix the SWOT coordinate scale

$$\mathcal{X} := [-10, 10] \quad \text{(internal axis)}, \qquad \mathcal{Y} := [-10, 10] \quad \text{(external axis)},$$

where $x < 0$ represents *weakness*, $x > 0$ *strength*, and $y < 0$ *threat*, $y > 0$ *opportunity*.

A *Fuzzy SWOT analysis* is a procedure that maps the factor sets $(I, E)$ into a prioritized list of strategies by the following mathematical components.

**(1) Fuzzification of factors.** Each internal factor $I_u$ is represented by a membership function $\mu_{I_u} : \mathcal{X} \to [0, 1]$, and each external factor $E_v$ by $\mu_{E_v} : \mathcal{Y} \to [0, 1]$. A common parametrization is the triangular form $I_u \equiv (x_u^p, x_u^m, x_u^o)$ and $E_v \equiv (y_v^p, y_v^m, y_v^o)$, with

$$\mu_{I_u}(x) = \text{tri}(x; x_u^p, x_u^m, x_u^o), \qquad \mu_{E_v}(y) = \text{tri}(y; y_v^p, y_v^m, y_v^o),$$

where for $a < b < c$,

$$\text{tri}(t; a, b, c) := \begin{cases} 0, & t < a, \\ \dfrac{t - a}{b - a}, & a \le t \le b, \\ \dfrac{c - t}{c - b}, & b \le t \le c, \\ 0, & t > c. \end{cases}$$

**(2) Aggregation surface (fuzzy SWOT surface).** Fix a t-norm $T : [0, 1]^2 \to [0, 1]$ (often $T = \min$). For each pair $(I_u, E_v)$ define the aggregated membership surface

$$\mu_{u,v} : \mathcal{X} \times \mathcal{Y} \to [0, 1], \qquad \mu_{u,v}(x, y) := T\big(\mu_{I_u}(x), \mu_{E_v}(y)\big).$$

**(3) $\alpha$-cut defuzzification and fuzzy area.** For a chosen $\alpha \in (0, 1]$, define the $\alpha$-cut region (fuzzy area)

$$A_{u,v}(\alpha) := \{(x, y) \in \mathcal{X} \times \mathcal{Y} : \mu_{u,v}(x, y) \ge \alpha\}.$$

This set encodes the interaction between $I_u$ and $E_v$ at uncertainty level $\alpha$.

**(4) Quadrant assignment (SO/ST/WO/WT).** Let

$$Q_{SO} = \{(x,y) : x \geq 0, \ y \geq 0\}, \quad Q_{ST} = \{(x,y) : x \geq 0, \ y \leq 0\},$$

$$Q_{WO} = \{(x,y) : x \leq 0, \ y \geq 0\}, \quad Q_{WT} = \{(x,y) : x \leq 0, \ y \leq 0\}.$$

If $A_{u,v}(\alpha)$ intersects multiple quadrants, define its quadrant by maximal area:

$$\mathrm{quad}_{u,v}(\alpha) \in \arg \max_{Q \in \{Q_{SO}, Q_{ST}, Q_{WO}, Q_{WT}\}}$$

$$\mathrm{Area}\big(A_{u,v}(\alpha) \cap Q\big).$$

**(5) Prioritization via ideal-point closeness (one standard choice).** Assume $A_{u,v}(\alpha)$ has positive area. Let its centroid (center of gravity) be

$$(x_{u,v}(\alpha), y_{u,v}(\alpha)) := \frac{1}{\mathrm{Area}(A_{u,v}(\alpha))} \iint_{A_{u,v}(\alpha)} (x,y)\, dx\, dy.$$

Define the positive and negative ideal points

$$z^+ = (10,10), \qquad z^- = (-10,-10),$$

the Euclidean distances

$$d^+_{u,v}(\alpha) := \big\|(x_{u,v}(\alpha), y_{u,v}(\alpha)) - z^+\big\|_2, \qquad d^-_{u,v}(\alpha) := \big\|(x_{u,v}(\alpha), y_{u,v}(\alpha)) - z^-\big\|_2,$$

and the *closeness coefficient*

$$\mathrm{cc}_{u,v}(\alpha) := \frac{d^-_{u,v}(\alpha)}{d^-_{u,v}(\alpha) + d^+_{u,v}(\alpha)} \in [0,1].$$

Larger $\mathrm{cc}_{u,v}(\alpha)$ indicates higher priority for the pair $(I_u, E_v)$ at level $\alpha$.

**(6) Strategy extraction.** A (crisp) strategy candidate is an ordered triple

$$\sigma = (I_u, E_v, \mathrm{quad}_{u,v}(\alpha)),$$

interpreted as a strategy generated from the internal factor $I_u$ and external factor $E_v$ according to the chosen quadrant. The prioritized strategy list at level $\alpha$ is obtained by sorting candidates by $\mathrm{cc}_{u,v}(\alpha)$ (or another chosen ranking functional).

**(7) Multi-$\alpha$ synthesis (optional).** For $\alpha_1, \ldots, \alpha_K \in (0,1]$ and weights $w_k \geq 0$ with $\sum_{k=1}^{K} w_k = 1$, define a final score

$$\mathrm{Score}_{u,v} := \sum_{k=1}^{K} w_k \, \mathrm{cc}_{u,v}(\alpha_k),$$

and rank strategies by $\mathrm{Score}_{u,v}$.

**Definition 10.7.2** (Uncertain SWOT analysis of type $M$ (U-SWOT)). Let $I = \{I_1, \ldots, I_{n_i}\}$ be internal factors and $E = \{E_1, \ldots, E_{n_e}\}$ external factors. Fix an uncertain model $M$ with $\text{Dom}(M) \neq \emptyset$ and an admissible score $S_M$.

Assume each internal factor $I_u$ and external factor $E_v$ is assessed by uncertain degrees

$$\iota_u^{(M)} \in \text{Dom}(M) \qquad (u = 1, \ldots, n_i), \qquad \epsilon_v^{(M)} \in \text{Dom}(M) \qquad (v = 1, \ldots, n_e).$$

Let the internal and external signed scores be

$$x_u := S_M(\iota_u^{(M)}) \in \mathbb{R}, \qquad y_v := S_M(\epsilon_v^{(M)}) \in \mathbb{R}.$$

Interpretation (convention):

$$x_u > 0: \text{ strength}, \quad x_u < 0: \text{ weakness}, \qquad y_v > 0: \text{ opportunity}, \quad y_v < 0: \text{ threat}.$$

(If a different sign convention is preferred, replace $x_u, y_v$ by affine rescalings.)

Fix a nonnegative *importance weight* for each factor:

$$\alpha_u \geq 0 \ (u = 1, \ldots, n_i), \qquad \beta_v \geq 0 \ (v = 1, \ldots, n_e),$$

and, for normalization convenience, assume $\sum_u \alpha_u = 1$ and $\sum_v \beta_v = 1$.

**(1) Strategy candidates (pairing).** Each pair $(I_u, E_v)$ generates a candidate strategy token

$$\sigma_{uv} := (u, v).$$

**(2) Quadrant (SO/ST/WO/WT) assignment.** Define

$$\text{quad}(u, v) := \begin{cases} SO, & x_u \geq 0, \ y_v \geq 0, \\ ST, & x_u \geq 0, \ y_v < 0, \\ WO, & x_u < 0, \ y_v \geq 0, \\ WT, & x_u < 0, \ y_v < 0. \end{cases}$$

**(3) Pair priority score.** Define the (nonnegative) interaction magnitude

$$\kappa_{uv} := \alpha_u \, \beta_v \, |x_u| \, |y_v| \ \in \mathbb{R}_{\geq 0}.$$

Normalize to obtain a comparable priority score:

$$\text{Score}_{uv} := \begin{cases} \dfrac{\kappa_{uv}}{\sum\limits_{p=1}^{n_i} \sum\limits_{q=1}^{n_e} \kappa_{pq}}, & \sum_{p,q} \kappa_{pq} > 0, \\[2ex] \dfrac{1}{n_i n_e}, & \sum_{p,q} \kappa_{pq} = 0, \end{cases} \qquad \text{Score}_{uv} \in [0, 1].$$

**(4) U-SWOT output.** The U-SWOT output is the prioritized list of annotated strategies

$$\Big\{ \big( \sigma_{uv}, \text{quad}(u, v), \text{Score}_{uv} \big) \Big\}_{u,v}$$

sorted by decreasing $\text{Score}_{uv}$ (ties allowed).

**Theorem 10.7.3** (Well-definedness of U-SWOT). *Under Definition 10.7.2, assume* $\mathrm{Dom}(M) \neq \emptyset$ *and* $S_M$ *is admissible. Then:*

    *(i) The signed scores* $x_u, y_v$ *are well-defined real numbers.*

    *(ii) For every* $(u, v)$, *the quadrant label* $\mathrm{quad}(u, v) \in \{SO, ST, WO, WT\}$ *is well-defined.*

    *(iii) The scores* $\kappa_{uv} \geq 0$ *and* $\mathrm{Score}_{uv} \in [0, 1]$ *are well-defined, and*

$$\sum_{u=1}^{n_i} \sum_{v=1}^{n_e} \mathrm{Score}_{uv} = 1.$$

    *(iv) Consequently, the prioritized strategy list (sorted by* $\mathrm{Score}_{uv}$*) is well-defined.*

*Proof.* (i) Since $\iota_u^{(M)}, \epsilon_v^{(M)} \in \mathrm{Dom}(M)$ and $S_M$ is admissible, $x_u = S_M(\iota_u^{(M)})$ and $y_v = S_M(\epsilon_v^{(M)})$ are finite real numbers.

(ii) Each quadrant condition is a Boolean combination of comparisons of real numbers ($x_u \geq 0$, $y_v \geq 0$ etc.), so exactly one of the four cases applies; thus $\mathrm{quad}(u, v)$ is well-defined.

(iii) By assumption $\alpha_u, \beta_v \geq 0$, so $\kappa_{uv} = \alpha_u \beta_v |x_u||y_v| \geq 0$ is well-defined. If $\sum_{p,q} \kappa_{pq} > 0$, then $\mathrm{Score}_{uv} = \kappa_{uv} / \sum_{p,q} \kappa_{pq}$ is well-defined and lies in $[0, 1]$, and summing over $(u, v)$ yields 1. If $\sum_{p,q} \kappa_{pq} = 0$, the definition sets $\mathrm{Score}_{uv} = 1/(n_i n_e)$, which also lies in $[0, 1]$ and sums to 1.

(iv) The set of candidates is finite, hence sorting by the real scores $\mathrm{Score}_{uv}$ yields a well-defined ranking (with ties allowed). □

Related concepts of SWOT under uncertainty-aware models are listed in Table 10.6.

Table 10.6: Related concepts of SWOT under uncertainty-aware models.

| $k$ | **Related SWOT concept(s)** |
|---|---|
| 1 | Fuzzy SWOT |
| 2 | Intuitionistic Fuzzy SWOT [1306] |
| 3 | Hesitant Fuzzy SWOT [1307, 1308] |
| 3 | Neutrosophic SWOT [1309–1311] |

As extensions other than Uncertain SWOT Analysis, several related concepts have been proposed, including the TOWS Matrix [1312, 1313], SWOT-AHP [1314, 1315], ANP-SWOT [1316, 1317], Quantified SWOT [1318, 1319], SWOT-QSPM [1320, 1321], and SOAR [1322, 1323].

## 10.8 Fuzzy Cost-Benefit Analysis

Cost-benefit analysis compares monetized benefits and costs over time, discounts cashflows, computes net present value, and guides project selection decisions [1324, 1325]. Fuzzy cost-benefit analysis models uncertain benefits, costs, or discount rates as fuzzy numbers, aggregates via $\alpha$-cuts, ranks alternatives robustly overall [1326–1328].

**Definition 10.8.1** (Fuzzy Cost–Benefit Analysis (Fuzzy CBA)). [1326–1328] Fix a finite planning horizon $T \in \mathbb{N}$ and let $\mathbb{R}_+ := [0, \infty)$. Let $\mathcal{F}(\mathbb{R}_+)$ denote the set of fuzzy numbers on $\mathbb{R}_+$ (normal, convex, upper semicontinuous, compact support).

Consider a project (or design alternative) $x$. For each time $t = 0, 1, \ldots, T$, model the (uncertain) benefit and cost cash-flows by fuzzy numbers

$$\widetilde{B}_t(x) \in \mathcal{F}(\mathbb{R}_+), \qquad \widetilde{C}_t(x) \in \mathcal{F}(\mathbb{R}_+).$$

Let the discount rate be either crisp $r \in [0, \infty)$ or fuzzy $\widetilde{r} \in \mathcal{F}([0, \infty))$. Define the (crisp) discount factor $d_t(r) := (1 + r)^{-t}$ and, when $r$ is fuzzy, treat $d_t(\widetilde{r})$ via the extension principle (equivalently, $\alpha$-cuts).

**(1) Fuzzy present worth of benefits and costs.** Using fuzzy arithmetic (e.g. via Zadeh's extension principle; equivalently via $\alpha$-cuts), define the fuzzy present worths

$$\widetilde{\mathrm{PW}}_B(x) := \bigoplus_{t=0}^{T} \widetilde{B}_t(x) \odot d_t(r), \qquad \widetilde{\mathrm{PW}}_C(x) := \bigoplus_{t=0}^{T} \widetilde{C}_t(x) \odot d_t(r),$$

when $r$ is crisp; and replace $d_t(r)$ by $d_t(\widetilde{r})$ when $r$ is fuzzy. Here $\oplus$ and $\odot$ denote fuzzy addition and scalar multiplication.

**$\alpha$-cut characterization (crisp $r$).** If $r$ is crisp and all cash-flows are nonnegative fuzzy numbers, then for each $\alpha \in (0, 1]$, writing $(\widetilde{B}_t(x))_\alpha = [B_t^L(\alpha; x), B_t^U(\alpha; x)]$ and $(\widetilde{C}_t(x))_\alpha = [C_t^L(\alpha; x), C_t^U(\alpha; x)]$,

$$\left(\widetilde{\mathrm{PW}}_B(x)\right)_\alpha = \left[\sum_{t=0}^{T} \frac{B_t^L(\alpha; x)}{(1+r)^t}, \ \sum_{t=0}^{T} \frac{B_t^U(\alpha; x)}{(1+r)^t}\right], \quad \left(\widetilde{\mathrm{PW}}_C(x)\right)_\alpha = \left[\sum_{t=0}^{T} \frac{C_t^L(\alpha; x)}{(1+r)^t}, \ \sum_{t=0}^{T} \frac{C_t^U(\alpha; x)}{(1+r)^t}\right].$$

**(2) Fuzzy benefit–cost ratio (fuzzy CBF/BCR).** Assume $\widetilde{\mathrm{PW}}_C(x)$ is strictly positive in the sense that $\inf(\widetilde{\mathrm{PW}}_C(x))_\alpha > 0$ for all $\alpha \in (0, 1]$. Define the *fuzzy benefit–cost ratio* (also called fuzzy cost–benefit function)

$$\widetilde{\mathrm{BCR}}(x) := \widetilde{\mathrm{PW}}_B(x) \oslash \widetilde{\mathrm{PW}}_C(x),$$

where $\oslash$ is fuzzy division. With nonnegative $\alpha$-cut intervals $(\widetilde{\mathrm{PW}}_B(x))_\alpha = [PW_B^L(\alpha; x), PW_B^U(\alpha; x)]$ and $(\widetilde{\mathrm{PW}}_C(x))_\alpha = [PW_C^L(\alpha; x), PW_C^U(\alpha; x)]$ and $PW_C^L(\alpha; x) > 0$, one obtains

$$\left(\widetilde{\mathrm{BCR}}(x)\right)_\alpha = \left[\frac{PW_B^L(\alpha; x)}{PW_C^U(\alpha; x)}, \ \frac{PW_B^U(\alpha; x)}{PW_C^L(\alpha; x)}\right].$$

Classically, BCR > 1 indicates cost-effectiveness; fuzzy CBA keeps this rule but interprets it through a fuzzy comparison/ranking.

**(3) Decision rule (one mathematically standard option).** Let $\widetilde{Z}$ be a fuzzy number on $\mathbb{R}$ and define the possibility and necessity of an event $\{\widetilde{Z} \geq z_0\}$ by

$$\Pi(\widetilde{Z} \geq z_0) := \sup_{z \geq z_0} \mu_{\widetilde{Z}}(z), \qquad N(\widetilde{Z} \geq z_0) := 1 - \sup_{z < z_0} \mu_{\widetilde{Z}}(z).$$

For a confidence level $\eta \in [0, 1]$, accept project $x$ if

$$N(\widetilde{\mathrm{BCR}}(x) \geq 1) \geq \eta \quad \text{(conservative)}, \qquad \text{or} \qquad \Pi(\widetilde{\mathrm{BCR}}(x) \geq 1) \geq \eta \quad \text{(optimistic)}.$$

For ranking multiple alternatives, compare $\widetilde{\mathrm{BCR}}(x)$ by any fixed fuzzy-number ranking functional (e.g. centroid defuzzification).

**(4) Incremental fuzzy CBA for mutually exclusive alternatives.** Given two alternatives $x$ (challenger) and $y$ (defender), define incremental fuzzy present worths

$$\Delta\widetilde{\mathrm{PW}}_B := \widetilde{\mathrm{PW}}_B(x) \ominus \widetilde{\mathrm{PW}}_B(y), \qquad \Delta\widetilde{\mathrm{PW}}_C := \widetilde{\mathrm{PW}}_C(x) \ominus \widetilde{\mathrm{PW}}_C(y),$$

and the incremental ratio

$$\Delta\widetilde{\mathrm{BCR}} := \Delta\widetilde{\mathrm{PW}}_B \oslash \Delta\widetilde{\mathrm{PW}}_C.$$

Then select $x$ over $y$ if (in the chosen fuzzy-comparison sense) $\Delta\widetilde{\mathrm{BCR}} > 1$; iterate this pairwise elimination to obtain a final choice.

**Definition 10.8.2** (Uncertain cost–benefit analysis of type $M$ (U-CBA)). Fix a finite planning horizon $T \in \mathbb{N}$ and a finite set of projects (alternatives) $\mathcal{X}$. Fix an uncertain model $M$ with $\mathrm{Dom}(M) \neq \emptyset$ and admissible scores $S_M^{\$}, S_M^{+}$.

For each project $x \in \mathcal{X}$ and each time $t = 0, 1, \ldots, T$, assume uncertain benefit and cost degrees

$$B_t^{(M)}(x) \in \mathrm{Dom}(M), \qquad C_t^{(M)}(x) \in \mathrm{Dom}(M),$$

and assume an uncertain discount-rate degree $r^{(M)} \in \mathrm{Dom}(M)$. Define the induced (crisp) nonnegative cashflows and discount rate by scoring:

$$B_t(x) := S_M^{\$}(B_t^{(M)}(x)) \in [0, \infty), \qquad C_t(x) := S_M^{\$}(C_t^{(M)}(x)) \in [0, \infty),$$

$$r := S_M^{+}(r^{(M)}) - 1 \ \in [0, \infty).$$

(Equivalently, one may define $S_M^r : \mathrm{Dom}(M) \to [0, \infty)$ directly; the above form enforces $1 + r > 0$.)

Define the discount factor

$$d_t(r) := (1 + r)^{-t} \in (0, 1] \qquad (t = 0, \ldots, T).$$

**(1) Present worths.** Define the present worth (PW) of benefits and costs:

$$\mathrm{PW}_B(x) := \sum_{t=0}^{T} B_t(x)\, d_t(r), \qquad \mathrm{PW}_C(x) := \sum_{t=0}^{T} C_t(x)\, d_t(r).$$

**(2) Net present value and benefit–cost ratio.** Define

$$\mathrm{NPV}(x) := \mathrm{PW}_B(x) - \mathrm{PW}_C(x) \in \mathbb{R},$$

and, provided $\mathrm{PW}_C(x) > 0$, define the benefit–cost ratio

$$\mathrm{BCR}(x) := \frac{\mathrm{PW}_B(x)}{\mathrm{PW}_C(x)} \in [0, \infty).$$

**(3) Decision rules (two standard options).**

- *Feasibility/acceptance:*  accept $x$ if $\mathrm{NPV}(x) \geq 0$ (or equivalently $\mathrm{BCR}(x) \geq 1$ when $\mathrm{PW}_C(x) > 0$).
- *Ranking:*  rank projects by decreasing $\mathrm{NPV}(x)$ (or decreasing $\mathrm{BCR}(x)$).

**Theorem 10.8.3** (Well-definedness of U-CBA). *Under Definition 10.8.2, assume:*

(i) *$S_M^\$$ is admissible monetary (finite on $\mathrm{Dom}(M)$) and $S_M^+$ is admissible positive;*

(ii) *$T < \infty$ and $\mathcal{X}$ is finite.*

*Then for every project $x \in \mathcal{X}$:*

(a) *$\mathrm{PW}_B(x)$ and $\mathrm{PW}_C(x)$ are well-defined finite nonnegative real numbers.*

(b) *$\mathrm{NPV}(x)$ is well-defined finite real.*

(c) *If $\mathrm{PW}_C(x) > 0$, then $\mathrm{BCR}(x)$ is well-defined and finite.*

*Moreover, if there exists at least one $x$ with $\mathrm{PW}_C(x) > 0$, then ranking by $\mathrm{BCR}$ is well-defined (up to ties) on that subset.*

*Proof.* (a) For each $t$, $B_t(x), C_t(x) \in [0, \infty)$ by admissibility of $S_M^\$$. Also $r \geq 0$ implies $1 + r > 0$ and hence $d_t(r) = (1 + r)^{-t} \in (0, 1]$ is well-defined. Therefore each term $B_t(x)d_t(r)$ and $C_t(x)d_t(r)$ is finite and nonnegative. Since $T$ is finite, the sums defining $\mathrm{PW}_B(x)$ and $\mathrm{PW}_C(x)$ are finite and nonnegative.

(b) $\mathrm{NPV}(x)$ is the difference of two finite reals, hence finite.

(c) If $\mathrm{PW}_C(x) > 0$, then division by $\mathrm{PW}_C(x)$ is valid and yields a finite real $\mathrm{BCR}(x) \geq 0$.

Finally, ranking by $\mathrm{BCR}$ on the subset $\{x : \mathrm{PW}_C(x) > 0\}$ is well-defined because it compares real numbers (and ties are allowed). $\qquad\square$

Related concepts of cost-benefit analysis under uncertainty-aware models are listed in Table 10.7.

Table 10.7: Related concepts of cost-benefit analysis under uncertainty-aware models.

| $k$ | Related cost-benefit analysis concept(s) |
|---|---|
| 1 | Fuzzy Cost-Benefit Analysis |
| 2 | Intuitionistic Fuzzy Cost-Benefit Analysis |
| 3 | Neutrosophic Cost-Benefit Analysis (cf. [1329]) |
| 3 | Spherical Fuzzy Cost-Benefit Analysis [1330, 1331] |

## 10.9 Fuzzy Decision curve analysis

Decision curve analysis evaluates prediction models across threshold probabilities by net benefit, balancing true positives against false positives to identify clinically useful decision strategies [1332, 1333]. Fuzzy decision curve analysis extends decision curve analysis by using fuzzy risk estimates or thresholds, modeling uncertainty in predictions while preserving net-benefit-based strategy comparison.

Let

$$\mathcal{FN}_{[0,1]}$$

denote a fixed class of fuzzy numbers supported on the interval $[0,1]$. Thus, each $\tilde{p} \in \mathcal{FN}_{[0,1]}$ is a fuzzy set on $[0,1]$ with membership function $\mu_{\tilde{p}} : [0,1] \to [0,1]$, and is interpreted as a fuzzy predicted risk / fuzzy probability.

**Definition 10.9.1** (Fuzzy Decision Curve Analysis (FDCA)). Let

$$\Omega = \{1, \ldots, N\}$$

be a finite set of cases (patients, objects, or decision units), and let

$$y_i \in \{0, 1\} \qquad (i \in \Omega)$$

denote the true binary outcome, where $y_i = 1$ means "event/disease present" and $y_i = 0$ means "event/disease absent."

Assume that a given prediction strategy $S$ provides, for each case $i \in \Omega$, a fuzzy predicted risk

$$\tilde{p}_i^{(S)} \in \mathcal{FN}_{[0,1]}.$$

Fix a threshold probability

$$t \in (0, 1),$$

and let

$$H : [0, 1) \to [0, \infty)$$

be a prescribed *test-harm function* (possibly $H \equiv 0$).

For each case $i$, define the *fuzzy treatment-recommendation degree* at threshold $t$ by

$$r_i^{(S)}(t) := \Pi(\tilde{p}_i^{(S)} \geq t) := \sup_{u \in [t,1]} \mu_{\tilde{p}_i^{(S)}}(u) \in [0, 1].$$

Hence $r_i^{(S)}(t)$ is the possibility degree that the fuzzy predicted risk of case $i$ reaches or exceeds $t$.

Define the corresponding fuzzy-weighted true-positive and false-positive counts by

$$\text{TP}_F^{(S)}(t) := \sum_{i=1}^N y_i\, r_i^{(S)}(t), \qquad \text{FP}_F^{(S)}(t) := \sum_{i=1}^N (1-y_i)\, r_i^{(S)}(t).$$

The *fuzzy net benefit* of strategy $S$ at threshold $t$ is then defined by

$$\text{NB}_F^{(S)}(t) := \frac{1}{N}\,\text{TP}_F^{(S)}(t) - \frac{1}{N}\,\text{FP}_F^{(S)}(t)\,\frac{t}{1-t} - H(t).$$

The mapping

$$\text{NB}_F^{(S)} : (0,1) \to \mathbb{R}, \qquad t \longmapsto \text{NB}_F^{(S)}(t),$$

is called the *fuzzy decision curve* of strategy $S$.

If $\mathscr{S}$ is a finite family of competing strategies, then the *FDCA-optimal strategy set* at threshold $t$ is

$$\text{Opt}_F(t) := \arg\max_{S \in \mathscr{S}} \text{NB}_F^{(S)}(t).$$

A strategy $S_1$ is said to *dominate* $S_2$ on a threshold region $T \subset (0,1)$ if

$$\text{NB}_F^{(S_1)}(t) \geq \text{NB}_F^{(S_2)}(t) \quad \text{for all } t \in T,$$

with strict inequality for at least one $t \in T$.

**Proposition 10.9.2** (Well-definedness of FDCA)**.** *Assume that, for a fixed strategy $S$,*

$$\tilde{p}_i^{(S)} \in \mathcal{FN}_{[0,1]} \qquad (i = 1, \ldots, N),$$

*that $t \in (0,1)$, and that $H(t) < \infty$. Then:*

1. *$r_i^{(S)}(t) \in [0,1]$ is well-defined for every $i$;*

2. *$\text{TP}_F^{(S)}(t)$ and $\text{FP}_F^{(S)}(t)$ are well-defined real numbers satisfying*

$$0 \leq \text{TP}_F^{(S)}(t) \leq N, \qquad 0 \leq \text{FP}_F^{(S)}(t) \leq N;$$

3. *$\text{NB}_F^{(S)}(t) \in \mathbb{R}$ is well-defined;*

4. *if $\mathscr{S}$ is finite and nonempty, then $\text{Opt}_F(t) \neq \varnothing$.*

*Proof.* Since each $\tilde{p}_i^{(S)}$ is a fuzzy number on $[0,1]$, its membership function $\mu_{\tilde{p}_i^{(S)}} : [0,1] \to [0,1]$ is well-defined. Therefore

$$r_i^{(S)}(t) = \sup_{u \in [t,1]} \mu_{\tilde{p}_i^{(S)}}(u)$$

is well-defined and belongs to $[0,1]$, proving (1).

Because $y_i \in \{0,1\}$ and $0 \leq r_i^{(S)}(t) \leq 1$, each summand in

$$\mathrm{TP}_F^{(S)}(t) = \sum_{i=1}^{N} y_i \, r_i^{(S)}(t)$$

and

$$\mathrm{FP}_F^{(S)}(t) = \sum_{i=1}^{N} (1-y_i) \, r_i^{(S)}(t)$$

lies in $[0,1]$. Hence both sums are well-defined real numbers in $[0,N]$, proving (2).

Since $t \in (0,1)$, the factor $t/(1-t)$ is a finite real number. Together with $H(t) < \infty$, this implies that

$$\mathrm{NB}_F^{(S)}(t) = \frac{1}{N}\,\mathrm{TP}_F^{(S)}(t) - \frac{1}{N}\,\mathrm{FP}_F^{(S)}(t)\frac{t}{1-t} - H(t)$$

is a well-defined real number. Thus (3) holds.

Finally, if $\mathscr{S}$ is finite and nonempty, then the set

$$\left\{ \mathrm{NB}_F^{(S)}(t) : S \in \mathscr{S} \right\} \subset \mathbb{R}$$

is finite and nonempty, and therefore attains its maximum. Hence

$$\mathrm{Opt}_F(t) = \arg\max_{S \in \mathscr{S}} \mathrm{NB}_F^{(S)}(t) \neq \varnothing.$$

This proves (4). $\qquad\square$

**Proposition 10.9.3** (Reduction to classical DCA)**.** *Assume that each fuzzy predicted risk degenerates to a crisp singleton:*

$$\tilde{p}_i^{(S)} = \chi_{\{p_i^{(S)}\}}, \qquad p_i^{(S)} \in [0,1].$$

*Then*

$$r_i^{(S)}(t) = \mathbf{1}[p_i^{(S)} \geq t],$$

*and consequently*

$$\mathrm{TP}_F^{(S)}(t) = \#\{\, i : \; y_i = 1, \; p_i^{(S)} \geq t \,\},$$
$$\mathrm{FP}_F^{(S)}(t) = \#\{\, i : \; y_i = 0, \; p_i^{(S)} \geq t \,\}.$$

*Hence $\mathrm{NB}_F^{(S)}(t)$ reduces exactly to the usual decision-curve net benefit*

$$\mathrm{NB}^{(S)}(t) = \frac{\mathrm{TP}^{(S)}(t)}{N} - \frac{\mathrm{FP}^{(S)}(t)}{N}\frac{t}{1-t} - H(t).$$

*Proof.* If $\tilde{p}_i^{(S)} = \chi_{\{p_i^{(S)}\}}$, then

$$\mu_{\tilde{p}_i^{(S)}}(u) = \begin{cases} 1, & u = p_i^{(S)}, \\ 0, & u \neq p_i^{(S)}. \end{cases}$$

Therefore

$$r_i^{(S)}(t) = \sup_{u \in [t,1]} \mu_{\tilde{p}_i^{(S)}}(u) = \begin{cases} 1, & p_i^{(S)} \geq t, \\ 0, & p_i^{(S)} < t, \end{cases}$$

which is exactly $\mathbf{1}[p_i^{(S)} \geq t]$. The remaining statements follow immediately by substitution into the definitions of $\mathrm{TP}_F^{(S)}(t)$, $\mathrm{FP}_F^{(S)}(t)$, and $\mathrm{NB}_F^{(S)}(t)$. $\qquad\square$

We now define *Uncertain Decision Curve Analysis* (UDCA) by extending fuzzy / probabilistic decision curve analysis to a general uncertain model $M$.

**Definition 10.9.4** (Uncertain Decision Curve Analysis (UDCA) of type $M$)**.** Let $M$ be an uncertain model with degree-domain

$$\mathrm{Dom}(M) \subseteq [0,1]^k$$

for some integer $k \geq 1$.

Let

$$\Omega = \{1, \dots, N\}$$

be a finite set of cases, and let

$$y_i \in \{0,1\} \qquad (i = 1, \dots, N)$$

denote the true binary outcome, where $y_i = 1$ means that the event/disease is present and $y_i = 0$ means that it is absent.

Let $\mathscr{S}$ be a finite nonempty family of candidate decision strategies. For each strategy $S \in \mathscr{S}$, assume that each case $i \in \Omega$ is assigned an *uncertain predicted risk*

$$\rho_i^{(S)} \in \mathrm{Dom}(M).$$

Fix a threshold probability

$$t \in (0,1),$$

and let

$$H : (0,1) \to [0, \infty)$$

be a prescribed *test-harm function*.

Assume further that a model-dependent *threshold-exceedance functional*

$$\mathrm{Exc}_M : \mathrm{Dom}(M) \times (0,1) \longrightarrow [0,1]$$

is given. For $d \in \mathrm{Dom}(M)$ and $t \in (0,1)$, the value

$$\mathrm{Exc}_M(d, t)$$

is interpreted as the degree to which the uncertain predicted risk $d$ reaches or exceeds the treatment threshold $t$.

For each strategy $S \in \mathscr{S}$, case $i \in \Omega$, and threshold $t \in (0,1)$, define the *uncertain treatment-recommendation degree*

$$r_{i,M}^{(S)}(t) := \mathrm{Exc}_M(\rho_i^{(S)}, t) \in [0,1].$$

Define the *uncertain true-positive count* and *uncertain false-positive count* of $S$ at threshold $t$ by

$$\mathrm{TP}_M^{(S)}(t) := \sum_{i=1}^{N} y_i\, r_{i,M}^{(S)}(t), \qquad \mathrm{FP}_M^{(S)}(t) := \sum_{i=1}^{N} (1 - y_i)\, r_{i,M}^{(S)}(t).$$

The *uncertain net benefit* of strategy $S$ at threshold $t$ is defined by

$$\mathrm{NB}_M^{(S)}(t) := \frac{1}{N}\,\mathrm{TP}_M^{(S)}(t) - \frac{1}{N}\,\mathrm{FP}_M^{(S)}(t)\,\frac{t}{1-t} - H(t).$$

The mapping

$$\mathrm{NB}_M^{(S)} : (0,1) \to \mathbb{R}, \qquad t \longmapsto \mathrm{NB}_M^{(S)}(t),$$

is called the *uncertain decision curve* of strategy $S$.

The induced preference relation among strategies at threshold $t$ is

$$S_1 \succeq_{M,t} S_2 \quad \Longleftrightarrow \quad \mathrm{NB}_M^{(S_1)}(t) \geq \mathrm{NB}_M^{(S_2)}(t).$$

The *UDCA-optimal strategy set* at threshold $t$ is

$$\mathrm{Opt}_M(t) := \arg\max_{S \in \mathscr{S}} \mathrm{NB}_M^{(S)}(t).$$

A strategy $S_1$ is said to *dominate* $S_2$ on a threshold region $T \subset (0,1)$ if

$$\mathrm{NB}_M^{(S_1)}(t) \geq \mathrm{NB}_M^{(S_2)}(t) \quad \text{for all } t \in T,$$

with strict inequality for at least one $t \in T$.

**Theorem 10.9.5** (Well-definedness of UDCA). *Let $M$ be an uncertain model, and let*

$$\mathfrak{U}_{\mathrm{DCA}} = (\Omega,\, (y_i)_{i=1}^N,\, \mathscr{S},\, (\rho_i^{(S)})_{i,S},\, \mathrm{Exc}_M,\, H)$$

*be a UDCA instance as in Definition 10.9.4. Assume:*

*(A1)  $N \geq 1$, $\Omega = \{1, \ldots, N\}$, and $y_i \in \{0,1\}$ for all $i \in \Omega$;*

*(A2)  $\mathscr{S}$ is finite and nonempty;*

*(A3)  for every $S \in \mathscr{S}$ and $i \in \Omega$,*

$$\rho_i^{(S)} \in \mathrm{Dom}(M);$$

*(A4)*

$$\mathrm{Exc}_M : \mathrm{Dom}(M) \times (0,1) \to [0,1]$$

    *is a total map;*

*(A5)*

$$H : (0,1) \to [0,\infty)$$

    *is a total map.*

*Then, for every threshold $t \in (0,1)$ and every strategy $S \in \mathscr{S}$, the following objects are well-defined:*

(i) *the uncertain treatment-recommendation degrees*

$$r_{i,M}^{(S)}(t) = \text{Exc}_M(\rho_i^{(S)}, t) \in [0,1] \qquad (i = 1, \ldots, N);$$

(ii) *the uncertain true-positive and false-positive counts*

$$\text{TP}_M^{(S)}(t), \ \text{FP}_M^{(S)}(t) \in [0, N];$$

(iii) *the uncertain net benefit*

$$\text{NB}_M^{(S)}(t) \in \mathbb{R};$$

(iv) *the induced preference relation $\succeq_{M,t}$ on $\mathscr{S}$, which is a total preorder;*

(v) *the optimal strategy set*

$$\text{Opt}_M(t) = \arg\max_{S \in \mathscr{S}} \text{NB}_M^{(S)}(t),$$

*which is nonempty.*

*Hence Uncertain Decision Curve Analysis of type $M$ is well-defined.*

*Proof.* Fix $t \in (0,1)$ and $S \in \mathscr{S}$.

By (A3), for each $i \in \Omega$,

$$\rho_i^{(S)} \in \text{Dom}(M).$$

Since $t \in (0,1)$ and $\text{Exc}_M$ is total by (A4), the value

$$r_{i,M}^{(S)}(t) = \text{Exc}_M(\rho_i^{(S)}, t)$$

is well-defined and belongs to $[0,1]$. This proves (i).

Next, because $y_i \in \{0,1\}$ by (A1) and $r_{i,M}^{(S)}(t) \in [0,1]$, each summand in

$$\text{TP}_M^{(S)}(t) = \sum_{i=1}^{N} y_i \, r_{i,M}^{(S)}(t)$$

and

$$\text{FP}_M^{(S)}(t) = \sum_{i=1}^{N} (1 - y_i) \, r_{i,M}^{(S)}(t)$$

belongs to $[0,1]$. Hence both sums are well-defined real numbers, and moreover

$$0 \leq \text{TP}_M^{(S)}(t) \leq N, \qquad 0 \leq \text{FP}_M^{(S)}(t) \leq N.$$

Thus (ii) holds.

Since $t \in (0,1)$, the quantity

$$\frac{t}{1-t}$$

is a finite real number. Also, $H(t) \in [0, \infty)$ is defined by (A5). Therefore

$$\mathrm{NB}_M^{(S)}(t) = \frac{1}{N}\,\mathrm{TP}_M^{(S)}(t) - \frac{1}{N}\,\mathrm{FP}_M^{(S)}(t)\frac{t}{1-t} - H(t)$$

is a well-defined real number. This proves (iii).

Now define

$$S_1 \succeq_{M,t} S_2 \iff \mathrm{NB}_M^{(S_1)}(t) \geq \mathrm{NB}_M^{(S_2)}(t).$$

Because $\geq$ on $\mathbb{R}$ is reflexive, transitive, and total, the induced relation $\succeq_{M,t}$ on $\mathscr{S}$ is also reflexive, transitive, and total. Hence it is a total preorder. This proves (iv).

Finally, since $\mathscr{S}$ is finite and nonempty by (A2), the set

$$\left\{ \mathrm{NB}_M^{(S)}(t) : S \in \mathscr{S} \right\} \subset \mathbb{R}$$

is finite and nonempty, so it attains a maximum. Therefore

$$\mathrm{Opt}_M(t) = \arg\max_{S \in \mathscr{S}} \mathrm{NB}_M^{(S)}(t) \neq \varnothing.$$

This proves (v), and hence UDCA is well-defined. $\qquad\square$

**Proposition 10.9.6** (Reduction to classical DCA). *Assume that $M$ is the crisp probabilistic model with*

$$\mathrm{Dom}(M) = [0, 1],$$

*and let the threshold-exceedance functional be*

$$\mathrm{Exc}_M(p, t) := \mathbf{1}[p \geq t] \qquad (p \in [0,1],\ t \in (0,1)).$$

*Then, for every strategy $S$, we have*

$$r_{i,M}^{(S)}(t) = \mathbf{1}[\rho_i^{(S)} \geq t],$$

*and consequently*

$$\mathrm{TP}_M^{(S)}(t) = \#\{\, i:\ y_i = 1,\ \rho_i^{(S)} \geq t \,\},$$

$$\mathrm{FP}_M^{(S)}(t) = \#\{\, i:\ y_i = 0,\ \rho_i^{(S)} \geq t \,\}.$$

*Hence $\mathrm{NB}_M^{(S)}(t)$ reduces exactly to the usual classical decision-curve net benefit formula.*

*Proof.* Immediate from the definition of $\mathrm{Exc}_M$ and substitution into the formulas of Definition 10.9.4. $\quad\square$

## 10.10   Fuzzy Rational Choice

Rational choice selects the most preferred feasible alternative according to a consistent preference relation, assuming decision makers compare options coherently and choose utility-maximizing outcomes [1334, 1335]. Fuzzy rational choice extends rational choice by using fuzzy preference degrees, allowing vague or partial comparisons, and selecting alternatives that sufficiently dominate competitors [1336–1338].

**Definition 10.10.1** (Fuzzy Rational Choice)**.** [1339,1340] Let $X$ be a finite nonempty set of alternatives, and let

$$\mathcal{P}^*(X) := \{A \subseteq X : A \neq \varnothing\}$$

denote the family of all nonempty subsets of $X$.

A *fuzzy preference relation* on $X$ is a mapping

$$r : X \times X \longrightarrow [0,1],$$

where $r(a,b)$ represents the degree to which $a$ is at least as good as $b$.

For each $a \in X$, define its 1/2-upper contour set by

$$R_{1/2}(a) := \{x \in X : r(a,x) \geq \tfrac{1}{2}\}.$$

The *choice correspondence induced by $r$* is the mapping

$$C_r : \mathcal{P}^*(X) \longrightarrow \mathcal{P}(X)$$

given by

$$C_r(A) := \{a \in A : A \subseteq R_{1/2}(a)\} = \{a \in A : r(a,x) \geq \tfrac{1}{2} \text{ for all } x \in A\}, \qquad A \in \mathcal{P}^*(X).$$

A choice function

$$C : \mathcal{P}^*(X) \longrightarrow \mathcal{P}(X)$$

is called a *fuzzy rational choice function* if there exists a fuzzy preference relation $r : X \times X \to [0,1]$ such that

$$C(A) = C_r(A) \qquad \text{for all } A \in \mathcal{P}^*(X).$$

In this case, $r$ is said to *rationalize $C$*.

We now define *Uncertain Rational Choice* by extending fuzzy rational choice from the fuzzy degree-domain $[0,1]$ to a general uncertain model $M$ with degree-domain $\mathrm{Dom}(M) \subseteq [0,1]^k$.

**Definition 10.10.2** (Uncertain binary preference relation of type $M$)**.** Let $X$ be a finite nonempty set of alternatives, and let $M$ be an uncertain model with degree-domain

$$\mathrm{Dom}(M) \subseteq [0,1]^k$$

for some integer $k \geq 1$.

An *uncertain binary preference relation of type $M$* on $X$ is a mapping

$$R_M : X \times X \longrightarrow \mathrm{Dom}(M).$$

For $a,b \in X$, the value

$$R_M(a,b) \in \mathrm{Dom}(M)$$

encodes the uncertain degree to which $a$ is at least as good as $b$, according to the model $M$.

**Definition 10.10.3** (Threshold functional and induced upper contour set)**.** Let $R_M : X \times X \to \mathrm{Dom}(M)$ be an uncertain binary preference relation. Fix a total map

$$\Theta_M : \mathrm{Dom}(M) \longrightarrow [0,1]$$

and a threshold level

$$\lambda \in (0,1].$$

For each $a \in X$, define the $(\Theta_M, \lambda)$-*upper contour set* of $a$ by

$$U_{M,\Theta,\lambda}(a) := \big\{ x \in X : \ \Theta_M(R_M(a,x)) \geq \lambda \big\}.$$

Thus $x \in U_{M,\Theta,\lambda}(a)$ means that, after evaluating the uncertain comparison $R_M(a,x)$ through the threshold functional $\Theta_M$, the alternative $a$ is judged to dominate $x$ at level at least $\lambda$.

**Definition 10.10.4** (Choice correspondence induced by an uncertain preference relation)**.** Let

$$\mathcal{P}^*(X) := \{ A \subseteq X : \ A \neq \varnothing \}$$

be the family of all nonempty subsets of $X$. Under the assumptions of Definition 10.10.3, define

$$C_{M,\Theta,\lambda} : \mathcal{P}^*(X) \longrightarrow \mathcal{P}(X)$$

by

$$C_{M,\Theta,\lambda}(A) := \big\{ a \in A : \ A \subseteq U_{M,\Theta,\lambda}(a) \big\} \qquad (A \in \mathcal{P}^*(X)).$$

Equivalently,

$$C_{M,\Theta,\lambda}(A) = \Big\{ a \in A : \ \Theta_M(R_M(a,x)) \geq \lambda \text{ for all } x \in A \Big\}.$$

This set is called the *uncertain choice correspondence induced by $R_M$*.

**Definition 10.10.5** (Uncertain rational choice function)**.** A mapping

$$C : \mathcal{P}^*(X) \longrightarrow \mathcal{P}(X)$$

is called an *uncertain rational choice function of type $M$* if there exist

- an uncertain binary preference relation

$$R_M : X \times X \to \mathrm{Dom}(M),$$

- a total threshold functional

$$\Theta_M : \mathrm{Dom}(M) \to [0,1],$$

- and a threshold level $\lambda \in (0,1]$,

such that

$$C(A) = C_{M,\Theta,\lambda}(A) \qquad \text{for all } A \in \mathcal{P}^*(X).$$

In this case, we say that $C$ is *rationalized* by the triple

$$(R_M, \Theta_M, \lambda).$$

**Theorem 10.10.6** (Well-definedness of uncertain rational choice). *Let $X$ be a finite nonempty set, let $M$ be an uncertain model, and let*

$$R_M : X \times X \longrightarrow \mathrm{Dom}(M)$$

*be an uncertain binary preference relation. Assume that*

$$\Theta_M : \mathrm{Dom}(M) \longrightarrow [0,1]$$

*is a total map and that $\lambda \in (0,1]$.*

*Then the induced mapping*

$$C_{M,\Theta,\lambda} : \mathcal{P}^*(X) \longrightarrow \mathcal{P}(X)$$

*defined by*

$$C_{M,\Theta,\lambda}(A) = \Big\{ a \in A : \ \Theta_M(R_M(a,x)) \geq \lambda \ \text{for all } x \in A \Big\}$$

*is well-defined. More precisely, for every $A \in \mathcal{P}^*(X)$,*

1. *$C_{M,\Theta,\lambda}(A) \subseteq A$;*

2. *$C_{M,\Theta,\lambda}(A)$ is uniquely determined;*

3. *$C_{M,\Theta,\lambda}(A) \in \mathcal{P}(X)$.*

*Hence $C_{M,\Theta,\lambda}$ is a well-defined set-valued choice correspondence on $X$.*

*Moreover, if the following* existence condition *holds:*

$$\forall A \in \mathcal{P}^*(X) \ \exists a \in A \ \forall x \in A, \ \Theta_M(R_M(a,x)) \geq \lambda,$$

*then*

$$C_{M,\Theta,\lambda}(A) \neq \varnothing \qquad \text{for all } A \in \mathcal{P}^*(X),$$

*so $C_{M,\Theta,\lambda}$ is a genuine choice function in the usual nonempty-valued sense.*

*Proof.* Fix any $A \in \mathcal{P}^*(X)$. Since $A \subseteq X$ and $A \neq \varnothing$, every element $a \in A$ and every $x \in A$ belong to $X$. Therefore, because

$$R_M : X \times X \to \mathrm{Dom}(M),$$

the quantity

$$R_M(a,x) \in \mathrm{Dom}(M)$$

is defined for every pair $(a,x) \in A \times A$.

Since $\Theta_M$ is total on $\mathrm{Dom}(M)$, the value

$$\Theta_M(R_M(a,x)) \in [0,1]$$

is defined for every $(a,x) \in A \times A$. Because $\lambda \in (0,1]$, the statement

$$\Theta_M(R_M(a,x)) \geq \lambda$$

has a definite truth value for every such pair.

Hence the set

$$\left\{ a \in A : \ \Theta_M(R_M(a,x)) \geq \lambda \text{ for all } x \in A \right\}$$

is well-defined by separation from the finite set $A$. By construction, every element of this set belongs to $A$, so

$$C_{M,\Theta,\lambda}(A) \subseteq A.$$

This proves (1).

The defining condition depends only on the already fixed data

$$A, \ R_M, \ \Theta_M, \ \lambda,$$

so the resulting subset $C_{M,\Theta,\lambda}(A)$ is uniquely determined. This proves (2).

Since $C_{M,\Theta,\lambda}(A) \subseteq A \subseteq X$, it follows that

$$C_{M,\Theta,\lambda}(A) \in \mathcal{P}(X).$$

This proves (3).

Therefore $C_{M,\Theta,\lambda}$ is a well-defined mapping from $\mathcal{P}^*(X)$ into $\mathcal{P}(X)$.

For the final statement, assume the existence condition

$$\forall A \in \mathcal{P}^*(X) \ \exists a \in A \ \forall x \in A, \ \Theta_M(R_M(a,x)) \geq \lambda.$$

Fix $A \in \mathcal{P}^*(X)$. Then there exists some $a^\star \in A$ such that

$$\Theta_M(R_M(a^\star,x)) \geq \lambda \qquad \text{for all } x \in A.$$

By definition of $C_{M,\Theta,\lambda}(A)$, this implies

$$a^\star \in C_{M,\Theta,\lambda}(A).$$

Hence $C_{M,\Theta,\lambda}(A) \neq \varnothing$. Since $A$ was arbitrary, the conclusion holds for all nonempty feasible sets $A$.   $\square$

**Proposition 10.10.7** (Reduction to fuzzy rational choice). *Let $M$ be the fuzzy model with*

$$\mathrm{Dom}(M) = [0,1].$$

*Take*

$$\Theta_M(u) := u \qquad (u \in [0,1]),$$

*and set*

$$\lambda = \frac{1}{2}.$$

*Then, for every fuzzy binary relation*

$$r : X \times X \to [0,1],$$

*the induced uncertain choice correspondence becomes*

$$C_{M,\Theta,\lambda}(A) = \{ \, a \in A : \ r(a,x) \geq \tfrac{1}{2} \text{ for all } x \in A \, \},$$

*which is exactly the standard 1/2-cut fuzzy rational choice rule.*

*Proof.* Under $\mathrm{Dom}(M) = [0,1]$ and $\Theta_M(u) = u$, we have

$$\Theta_M(R_M(a,x)) \geq \lambda \iff r(a,x) \geq \frac{1}{2}.$$

Substituting this into Definition 10.10.4 yields the stated formula.   $\square$

# Chapter 11

# Applications of Decision-Making

In this chapter, we briefly review application areas of decision-making.

## 11.1  Applications of Computer Science and Engineering

This section presents representative applications of decision-making in computer science and engineering.

- Software requirements prioritization and release planning: Software requirements prioritization and release planning rank features by value, cost, risk, and dependencies to choose optimal release contents (cf. [1341]). Analyses have been conducted using Fuzzy MADM [1342, 1343] and multi-person decision-making [1344].

- IT service management (ITSM) improvement prioritization: ITSM improvement prioritization ranks process and tool initiatives by incident reduction, customer impact, effort, risk, compliance, and ROI (cf. [1345]). Analyses have been conducted using fuzzy sets and neutrosophic sets [1346, 1347].

- Cybersecurity control prioritization and risk-treatment planning: Ranks security controls by risk reduction, cost, and feasibility, guiding treatment plans, budgeting, and implementation sequencing against evolving threats effectively (cf. [1348]). Analyses have been conducted using fuzzy decision-making [1349, 1350] and neutrosophic decision-making [1351, 1352].

- Cloud migration and system architecture alternative selection: Evaluates cloud migration and architecture options by cost, latency, security, scalability, compliance, and operational complexity to choose best-fit design. Analyses have been conducted using fuzzy logic [1353].

## 11.2  Applications of Manufacturing

This section presents representative applications of decision-making in Manufacturing.

- Production planning and assembly line balancing: Production planning and assembly line balancing schedule tasks and assign workloads to stations, minimizing cycle time, costs, and bottlenecks (cf. [1354]). Cases have been analyzed using fuzzy decision-making [1355, 1356] and neutrosophic decision-making [1357, 1358].

- Maintenance strategy selection (preventive, predictive, condition-based): Maintenance strategy selection chooses preventive, predictive, or corrective policies by balancing reliability, downtime, cost, risk, and resource constraints (cf. [1359]). Cases have been analyzed using fuzzy MCDM [1360, 1361].

- Prioritization in power grids and industrial assets: Prioritization in power grids and industrial assets ranks maintenance and investment actions by reliability impact, risk, cost, urgency, and safety constraints. Cases have been analyzed using fuzzy MCDM [1362].

- Project selection and portfolio optimization (including R&D portfolios): Project selection and portfolio optimization choose and balance projects under budget and risk constraints to maximize strategic value and returns (cf. [1363]). Analyses have been conducted using a variety of methods, including Fuzzy MCDM [1364, 1365], Fuzzy MULTIMOORA [1366], Fuzzy TOPSIS [1367–1369], Neutrosophic MCDM [1370], Neutrosophic AHP [1371], and neutrosophic TOPSIS [1372].

## 11.3 Applications of Finance

This section presents representative applications of decision-making in Finance.

- Credit scoring and loan approval (financial decision-making): Credit scoring and loan approval assess borrower risk using financial and behavioral indicators, assigning ratings to decide credit limits and approvals (cf. [1373]). Analyses have been conducted using fuzzy decision-making [1374, 1375], intuitionistic fuzzy decision-making [1376], and neutrosophic decision-making [1377, 1378].

## 11.4 Applications of Business

This section presents representative applications of decision-making in Business.

- Supplier selection and procurement evaluation: Supplier selection and procurement evaluation ranks vendors by cost, quality, delivery, risk, sustainability, and strategic fit, using multi-criteria analysis to support sourcing decisions (cf. [1379]). A wide range of analytical approaches has been investigated, including Fuzzy MCDM [1380, 1381], Fuzzy AHP [1382], Fuzzy TOPSIS [1383, 1384], and Neutrosophic TOPSIS [1385–1387].

- Logistics hub / warehouse location selection: Logistics hub / warehouse location selection chooses optimal facility sites by balancing cost, service coverage, demand proximity, capacity, risk, and sustainability objectives under multiple criteria (cf. [1388]). Studies have explored analyses using methods such as Fuzzy MCDM [1389], Fuzzy TOPSIS [1390], and Fuzzy BWM [1391].

- Transportation mode and route selection in supply chains: Chooses transport modes and routes by balancing cost, transit time, reliability, capacity, emissions, and risk across the supply chain. It has been studied using fuzzy MCDM approaches [1392, 1393] and fuzzy AHP methods [1394–1396], among others.

- Human resource selection, performance evaluation, and assignment decisions: Selects and assigns personnel by evaluating skills, experience, performance metrics, availability, and fit, balancing fairness, cost, and organizational objectives under uncertainty. Analyses using methods such as fuzzy AHP have been reported [1397, 1398].

- Marketing strategy selection (budget allocation and campaign prioritization): Selects marketing strategies by comparing channels and campaigns on ROI, reach, timing, risk, and budget constraints to prioritize spending. Analyses have been conducted using fuzzy ANP [1399, 1400], fuzzy AHP [1401], and fuzzy DEMATEL [1402].

## 11.5 Applications of Medicine

This section presents representative applications of decision-making in Medicine.

- Medical Diagnosis: Interprets symptoms, history, tests, and imaging to infer disease likelihoods, confirm causes, and guide treatment decisions under uncertainty (cf. [1403]). Several studies have been conducted within frameworks such as fuzzy decision-making [1404, 1405], intuitionistic fuzzy decision-making [87, 1406], and neutrosophic decision-making [1407, 1408].

- Treatment option selection and clinical decision support: Ranks treatment alternatives using patient data and criteria, supporting clinicians with transparent recommendations under uncertainty, constraints, and trade-offs. Analyses using methods such as fuzzy PROMETHEE have been reported [1409].

- Bed/ICU allocation and surge planning under uncertain demand and length of stay: Allocates beds and ICU capacity by forecasting uncertain demand and stay lengths, optimizing staffing, resources, triage, and surge responses (cf. [1410]). Analyses using fuzzy MCDM [1411] and fuzzy AHP [1412, 1413] have been reported.

## 11.6 Applications of Social Science

This section presents representative applications of decision-making in Social Science.

- Environmental impact assessment and mitigation option selection: Assesses environmental impacts and selects mitigation options by comparing severity, likelihood, cost, compliance, and stakeholder priorities under uncertainty. Analyses have been conducted using methods such as fuzzy AHP [1414].

- Renewable energy site selection (wind/solar) and technology choice: Selects wind/solar sites by evaluating resource potential, grid access, land constraints, environmental impacts, costs, and uncertainty (cf. [1415]). It has been analyzed using fuzzy MCDM methods [1416, 1417].

- Weather prediction: Forecasts future weather by combining observations and models to estimate temperature, precipitation, wind, and uncertainty over time and space (cf. [1418]). It has been analyzed using fuzzy MCDM methods [1419] and neutrosophic MCDM approaches [1420].

- Smartphone selection: Chooses smartphones by comparing price, performance, camera, battery, display, software support, durability, and user preferences under constraints. It has been studied using fuzzy MCDM [1421], fuzzy WASPAS [1422], fuzzy AHP [31], and fuzzy TOPSIS [1423], among others.

# Chapter 12

# Discussions: New Decision-Making Methods

In this chapter, we examine several new decision-making methods. For convenience, the main concepts introduced in this chapter are compared in Table 12.1.

Table 12.1: Concise comparison of the new decision-making methods introduced in this chapter.

| Method | Type | Basic structure | Core mechanism | Primary output | Decision role |
|---|---|---|---|---|---|
| Unified Decisional Structure (UDS) | Crisp meta-framework | Alternatives, criteria, evaluation domains, decision data, weights, preprocessing, kernel, and decision map | Processes raw evaluation data through Prep → Ker → Dec to produce an admissible decision object | Score, ranking, choice set, or sorting/classification | Method-unifying decision architecture |
| Unified Uncertain Decisional Structure (UUDS) | Uncertain meta-framework | Alternatives, criteria, uncertainty algebras, uncertain evaluations, weights, and representation-invariant operators | Applies uncertainty-aware preprocessing, kernel construction, and decision mapping in a representation-independent way | Representation-invariant score, ranking, choice set, or sorting/classification | Uncertainty-unifying decision architecture |
| Iterated Multi-Criteria Decision-Making (I-MCDM) | Hierarchical scoring method | Criterion tree, leaf scores, and internal MCDM scoring operators with local weights | Recursively aggregates child criterion scores upward until a root score is obtained | Final score vector and induced ranking | Hierarchical MCDM |
| Iterated Multi-Attribute Decision-Making (I-MADM) | Hierarchical attribute aggregation | Attribute tree, atomic attribute scores, and internal MADM operators with local weights | Recursively combines lower-level attribute scores into higher-level attribute scores | Final root score and ranking | Hierarchical MADM |
| Iterated Multi-Objective Decision-Making (I-MODM) | Hierarchical optimization | Feasible set, objective tree, leaf objectives, aggregation operators, and a solver | Recursively aggregates objectives into a final scalar objective and then solves it | Final scalar objective and solution set | Hierarchical MODM |

*Table 12.1 (continued).*

| Method | Type | Basic structure | Core mechanism | Primary output | Decision role |
|---|---|---|---|---|---|
| Analytic Hyper-Network Process (AHNP) | Hypernetwork prioritization | Directed hypernetwork, pairwise-comparison matrices, local priorities, and a hypersupermatrix | Computes Perron priority vectors on hyperarcs and derives global priorities from a stationary distribution | Global priority vector and ranking | Network-based prioritization |
| Analytic SuperHyperNetwork Process (ASHNP) | Superhypernetwork prioritization | Directed $n$-superhypernetwork, pairwise-comparison matrices on supernodes, and a superhypersupermatrix | Extends AHNP from ordinary nodes to $n$-supernodes and computes stationary global priorities | Global priority vector and ranking on supernodes | Hierarchical network prioritization |
| Analytic Recursive SuperHyperNetwork Process (ARSHNP) | Recursive superhypernetwork prioritization | Directed recursive superhypernetwork, support-based pairwise matrices, and a recursive hypersupermatrix | Flattens recursive structures through supports, computes local priorities, and extracts global priorities by a stationary vector | Global priority vector and ranking | Recursive hierarchical prioritization |
| SuperHyperDecision-Making (SHDM) | Reachability-based decision system | Working universe in $\mathcal{P}^n(D)$, superhypercombination rule, seed set, and preorder | Generates the least reachable closed set by iterative consequence expansion and selects maximal reachable states | Set of maximal reachable decision states | Reachability-based choice |
| Uncertain SuperHyperDecision-Making (USHDM) | Uncertain reachability-based decision system | SHDM structure together with an uncertainty structure and uncertain valuation map | Combines reachable-state generation with uncertainty-valued evaluation and score-induced maximality | Set of maximal uncertain decision outcomes | Uncertainty-aware reachability-based choice |

**Note.** UDS and UUDS are general decision architectures rather than single concrete methods. I-MCDM, I-MADM, and I-MODM are recursive hierarchical aggregation schemes on trees of criteria, attributes, or objectives. AHNP, ASHNP, and ARSHNP are network-based prioritization models using stationary supermatrix ideas on increasingly richer structures. SHDM and USHDM are closure- and reachability-based decision systems, where the final decision is obtained from maximal reachable states, with uncertainty incorporated in the latter case.

## 12.1 Unified Decisional Structure

The Unified Decisional Structure is a framework that maps decision data through preprocessing, a decision kernel, and a decision map to produce scores, rankings, choices, or classifications across methods.

**Definition 12.1.1** (Decision tasks and admissible outputs). Let $\mathcal{A}$ be a finite nonempty set of alternatives. A *decision output* on $\mathcal{A}$ is one of the following objects:

1. (Scoring) a map $u : \mathcal{A} \to \mathbb{R}$;

2. (Ranking) a total preorder $\succeq$ on $\mathcal{A}$ (reflexive, transitive, and total);

3. (Choice) a nonempty subset $\mathcal{A}^\star \subseteq \mathcal{A}$;

4. (Sorting / classification) a map $\sigma : \mathcal{A} \to \mathcal{K}$ into a finite ordered set of categories $\mathcal{K} = \{K_1 \succ K_2 \succ \cdots \succ K_L\}$.

Denote by $\mathfrak{O}(\mathcal{A})$ the (disjoint) collection of all such admissible outputs.

**Definition 12.1.2** (Unified Decisional Structure (UDS)). A *Unified Decisional Structure* (UDS) is a tuple

$$\mathsf{UDS} := \Big( \mathcal{A}, \mathcal{C}, (\mathbb{E}_j)_{j=1}^n, X, w, \mathsf{Prep}, \mathsf{Ker}, \mathsf{Dec} \Big)$$

satisfying:

1. (Alternatives and criteria) $\mathcal{A} = \{A_1, \ldots, A_m\}$ and $\mathcal{C} = \{C_1, \ldots, C_n\}$ are finite nonempty sets.

2. (Evaluation domains) For each criterion $C_j$, $\mathbb{E}_j$ is a nonempty set (e.g. $\mathbb{R}$, an interval scale, an ordinal scale).

3. (Decision data) The *evaluation map* is a total function

$$X : \mathcal{A} \times \mathcal{C} \to \bigsqcup_{j=1}^n \mathbb{E}_j, \qquad X(A_i, C_j) =: x_{ij} \in \mathbb{E}_j.$$

4. (Weights) The *weight vector* is $w = (w_1, \ldots, w_n) \in [0, 1]^n$ with

$$\sum_{j=1}^n w_j = 1.$$

5. (Preprocessing operator) $\mathsf{Prep}$ is a specification of maps (possibly depending on $w$ and/or on the full matrix $(x_{ij})$) producing a *processed representation* $\mathsf{Prep}(X, w) \in \mathcal{D}$, where $\mathcal{D}$ is a fixed data space. This step may include normalization, benefit/cost conversion, scaling, reference-profile construction, etc.

6. (Decision kernel) $\mathsf{Ker} : \mathcal{D} \to \mathcal{Z}$ is a total function that maps processed data to an *evidence object* $z \in \mathcal{Z}$ (e.g. a utility vector, a distance-to-ideal vector, an outranking matrix, a dominance matrix, flows, etc.).

7. (Decision map) $\mathsf{Dec} : \mathcal{Z} \to \mathfrak{O}(\mathcal{A})$ is a total function that converts evidence into an admissible output (score/ranking/choice/sorting).

The induced decision procedure is the composite map

$$\mathsf{DM}_{\mathsf{UDS}} := \mathsf{Dec} \circ \mathsf{Ker} \circ \mathsf{Prep}.$$

**Theorem 12.1.3** (Well-definedness of UDS)**.** *Every* $\mathsf{UDS}$ *in Theorem 12.1.2 induces a unique decision procedure*

$$\mathsf{DM}_{\mathsf{UDS}} : (X, w) \longmapsto \mathsf{DM}_{\mathsf{UDS}}(X, w) \in \mathfrak{O}(\mathcal{A}),$$

*i.e. the output exists and is uniquely determined by* $(X, w)$.

*Proof.* By Theorem 12.1.2, $\mathsf{Prep}$ is a total mapping from inputs $(X, w)$ to $\mathcal{D}$, $\mathsf{Ker}$ is a total mapping from $\mathcal{D}$ to $\mathcal{Z}$, and $\mathsf{Dec}$ is a total mapping from $\mathcal{Z}$ to $\mathfrak{O}(\mathcal{A})$. Therefore the composition $\mathsf{Dec} \circ \mathsf{Ker} \circ \mathsf{Prep}$ is a total function from $(X, w)$ to $\mathfrak{O}(\mathcal{A})$. Function values are unique, hence the induced output is unique. $\qquad\square$

## 12.2   Unified Uncertain Decisional Structure

Unified Uncertain Decisional Structure extends UDS with uncertainty algebras and representation-invariant operators, yielding method-agnostic decisions from fuzzy/rough/grey/neutrosophic evaluations.

**Definition 12.2.1** (Uncertainty algebra with representation equivalence). An *uncertainty algebra* is a tuple

$$\mathbb{U} = (U, \sim, \{\omega_\alpha\}_{\alpha \in \mathcal{I}}, \ \mathrm{sc})$$

where:

1. $U$ is a nonempty set of uncertain values (e.g. fuzzy numbers, rough intervals, grey numbers, neutrosophic triples, intuitionistic pairs, linguistic terms, etc.).

2. $\sim$ is an equivalence relation on $U$ encoding *representational equality* (e.g. two different parameterizations describing the same membership function).

3. Each $\omega_\alpha$ is a (possibly multi-ary) operation on $U$ used in the decision pipeline, and *compatibility* holds: if $u_k \sim u'_k$ for all arguments, then

$$\omega_\alpha(u_1, \ldots, u_p) \sim \omega_\alpha(u'_1, \ldots, u'_p) \quad \text{(whenever both sides are defined)}.$$

4. $\mathrm{sc} : U \to \mathbb{R}^q$ is a *score (crispification) map* that is constant on equivalence classes:

$$u \sim v \implies \mathrm{sc}(u) = \mathrm{sc}(v).$$

**Definition 12.2.2** (Unified Uncertain Decisional Structure (UUDS)). A *Unified Uncertain Decisional Structure* (UUDS) is a tuple

$$\mathsf{UUDS} := \left( \mathcal{A}, \mathcal{C}, (\mathbb{U}_j)_{j=1}^n, \widetilde{X}, w, \widetilde{\mathsf{Prep}}, \widetilde{\mathsf{Ker}}, \widetilde{\mathsf{Dec}} \right)$$

such that:

1. $\mathcal{A} = \{A_1, \ldots, A_m\}$ and $\mathcal{C} = \{C_1, \ldots, C_n\}$ are finite nonempty.

2. For each $j$, $\mathbb{U}_j = (U_j, \sim_j, \{\omega_{\alpha,j}\}_{\alpha \in \mathcal{I}_j}, \mathrm{sc}_j)$ is an uncertainty algebra in the sense of Theorem 12.2.1.

3. The uncertain evaluation map is total:

$$\widetilde{X} : \mathcal{A} \times \mathcal{C} \to \bigsqcup_{j=1}^n U_j, \qquad \widetilde{X}(A_i, C_j) =: \tilde{x}_{ij} \in U_j.$$

4. Weights satisfy $w_j \in [0, 1]$ and $\sum_{j=1}^n w_j = 1$.

5. The preprocessing operator $\widetilde{\mathsf{Prep}}$ maps $(\widetilde{X}, w)$ to a processed space $\widetilde{\mathcal{D}}$, and is *representation-invariant*: if $\tilde{x}_{ij} \sim_j \tilde{x}'_{ij}$ for all $(i, j)$, then

$$\widetilde{\mathsf{Prep}}(\widetilde{X}, w) = \widetilde{\mathsf{Prep}}(\widetilde{X}', w).$$

6. The kernel $\widetilde{\mathsf{Ker}} : \widetilde{\mathcal{D}} \to \widetilde{\mathcal{Z}}$ is total and representation-invariant (with respect to the induced equivalence on $\widetilde{\mathcal{D}}$ coming from the $\sim_j$-compatibility of all used operations).

7. The decision map $\widetilde{\mathsf{Dec}} : \widetilde{\mathcal{Z}} \to \mathcal{O}(\mathcal{A})$ is total and depends on uncertain objects only via class-invariant quantities (typically via score maps $\mathrm{sc}_j$ or their aggregates).

The induced uncertain decision procedure is

$$\mathsf{DM}_{\mathsf{UUDS}} := \widetilde{\mathsf{Dec}} \circ \widetilde{\mathsf{Ker}} \circ \widetilde{\mathsf{Prep}}.$$

**Theorem 12.2.3** (Well-definedness of UUDS). *Let* UUDS *be as in Theorem 12.2.2. Then:*

1. *The output* $\mathsf{DM}_{\mathsf{UUDS}}(\widetilde{X}, w)$ *exists and is unique for every input* $(\widetilde{X}, w)$.

2. *The output is* representation-independent*: if* $\widetilde{X}$ *and* $\widetilde{X}'$ *satisfy* $\tilde{x}_{ij} \sim_j \tilde{x}'_{ij}$ *for all* $(i, j)$, *then*

$$\mathsf{DM}_{\mathsf{UUDS}}(\widetilde{X}, w) = \mathsf{DM}_{\mathsf{UUDS}}(\widetilde{X}', w).$$

*Proof.* (1) By Theorem 12.2.2, $\widetilde{\mathsf{Prep}}$ is total from inputs $(\widetilde{X}, w)$ to $\widetilde{\mathcal{D}}$, $\widetilde{\mathsf{Ker}}$ is total from $\widetilde{\mathcal{D}}$ to $\widetilde{\mathcal{Z}}$, and $\widetilde{\mathsf{Dec}}$ is total from $\widetilde{\mathcal{Z}}$ to $\mathfrak{O}(\mathcal{A})$. Hence the composite map exists and is a function; therefore its value is unique.

(2) Assume $\tilde{x}_{ij} \sim_j \tilde{x}'_{ij}$ entrywise. By the representation-invariance requirement in Theorem 12.2.2(5),

$$\widetilde{\mathsf{Prep}}(\widetilde{X}, w) = \widetilde{\mathsf{Prep}}(\widetilde{X}', w).$$

Applying the representation-invariant kernel Theorem 12.2.2(6) yields

$$\widetilde{\mathsf{Ker}}\left(\widetilde{\mathsf{Prep}}(\widetilde{X}, w)\right) = \widetilde{\mathsf{Ker}}\left(\widetilde{\mathsf{Prep}}(\widetilde{X}', w)\right).$$

Finally, $\widetilde{\mathsf{Dec}}$ depends only on class-invariant quantities (e.g. score maps constant on $\sim_j$-classes), so applying $\widetilde{\mathsf{Dec}}$ preserves equality and gives $\mathsf{DM}_{\mathsf{UUDS}}(\widetilde{X}, w) = \mathsf{DM}_{\mathsf{UUDS}}(\widetilde{X}', w)$. $\qquad\square$

## 12.3 Iterated Multi-Criteria Decision-Making

Iterated Multi-Criteria Decision-Making recursively composes MCDM scoring operators across a hierarchical criterion tree, aggregating leaf criterion scores into a final root score ranking alternatives.

**Definition 12.3.1** (MCDM scoring operator). Let $\mathcal{A} = \{A_1, \ldots, A_m\}$ be a finite nonempty set of alternatives and let $k \in \mathbb{N}$. A *(crisp) MCDM scoring operator* of arity $k$ is a total function

$$\mathcal{M} : \mathbb{R}^{m \times k} \times \Delta_k \longrightarrow \mathbb{R}^m,$$

where $\Delta_k := \{ w \in [0, 1]^k : \sum_{\ell=1}^k w_\ell = 1 \}$ is the simplex of weights. Given a performance matrix $X \in \mathbb{R}^{m \times k}$ and weights $w \in \Delta_k$, the vector $\mathcal{M}(X, w) = (u(A_1), \ldots, u(A_m))^\top$ is interpreted as a *score* on $\mathcal{A}$.

(Examples: weighted sum, TOPSIS-score, VIKOR-index, MOORA-score, TODIM-value, PROMETHEE net-flow, etc., provided they output a real-valued score for each alternative.)

**Definition 12.3.2** (Criterion tree). A *criterion tree* is a finite rooted tree $T = (V, E, r)$. For a node $v \in V$, let $\mathrm{Ch}(v) \subseteq V$ be the (finite) set of children of $v$. A node is a *leaf* if $\mathrm{Ch}(v) = \varnothing$, and an *internal node* otherwise. The *height* $\mathrm{ht}(T)$ is the maximum number of edges on a directed path from the root to a leaf.

**Definition 12.3.3** (Iterated Multi-Criteria Decision-Making (I-MCDM)). Let $\mathcal{A} = \{A_1, \ldots, A_m\}$ be alternatives and let $T = (V, E, r)$ be a criterion tree. An *Iterated Multi-Criteria Decision-Making* (I-MCDM) instance is a collection

$$\text{I-MCDM} := \Big( \mathcal{A},\, T,\, \{u_\ell\}_{\ell \in L},\, \{\mathcal{M}_v, w^{(v)}\}_{v \in V \setminus L} \Big),$$

where:

1. $L \subseteq V$ is the set of leaves. Each leaf $\ell \in L$ is equipped with a *base (atomic) score*

   $$u_\ell : \mathcal{A} \to \mathbb{R}.$$

   (Equivalently, $u_\ell$ may be obtained from raw measurements/linguistic assessments by any fixed pre-processing and crispification; the present definition only requires the resulting map $\mathcal{A} \to \mathbb{R}$.)

2. Each internal node $v \in V \setminus L$ with $k_v := |\operatorname{Ch}(v)|$ is equipped with

   $$\mathcal{M}_v : \ \mathbb{R}^{m \times k_v} \times \Delta_{k_v} \to \mathbb{R}^m \quad \text{(an MCDM scoring operator)}$$

   and a weight vector $w^{(v)} \in \Delta_{k_v}$ over its children.

**Recursive semantics ("multi-criteria of multi-criteria of $\cdots$").** Define, for every node $v \in V$, a score function $u_v : \mathcal{A} \to \mathbb{R}$ recursively by:

(S1) If $v$ is a leaf, set $u_v := u_v$ given above (base score).

(S2) If $v$ is internal, enumerate its children as $\operatorname{Ch}(v) = \{c_1, \ldots, c_{k_v}\}$ and form the *child-score matrix* $X^{(v)} \in \mathbb{R}^{m \times k_v}$ by

$$X^{(v)}_{i\ell} := u_{c_\ell}(A_i) \qquad (i = 1, \ldots, m, \ \ell = 1, \ldots, k_v).$$

Then define the parent score vector by

$$\big( u_v(A_1), \ldots, u_v(A_m) \big)^\top := \mathcal{M}_v\Big( X^{(v)},\, w^{(v)} \Big).$$

The *final I-MCDM score* is $u_r : \mathcal{A} \to \mathbb{R}$ at the root. A final ranking can be induced by $A_i \succeq A_k \iff u_r(A_i) \geq u_r(A_k)$.

**Theorem 12.3.4** (Well-definedness of I-MCDM). *Let* I-MCDM *be as in Theorem 12.3.3. Assume that $T$ is finite and every $\mathcal{M}_v$ is a total function on its stated domain. Then:*

1. *For every node $v \in V$, the recursively defined map $u_v : \mathcal{A} \to \mathbb{R}$ exists and is unique.*

2. *In particular, the final score $u_r$ and the induced ranking are uniquely determined by the instance data.*

*Proof.* Because $T$ is a finite rooted tree, nodes admit a well-founded order by depth-from-leaves (equivalently, by decreasing distance to leaves). Proceed by induction on this order.

*Base step (leaves).* If $v$ is a leaf, $u_v$ is given as part of the instance, hence exists and is unique.

*Inductive step (internal nodes).* Assume that for all children $c \in \mathrm{Ch}(v)$, the functions $u_c$ exist and are unique. Then the child-score matrix $X^{(v)}$ is uniquely determined, since each entry $X^{(v)}_{i\ell} = u_{c_\ell}(A_i)$ is uniquely determined. Because $\mathcal{M}_v$ is total and $w^{(v)} \in \Delta_{k_v}$ is fixed, the vector $\mathcal{M}_v(X^{(v)}, w^{(v)}) \in \mathbb{R}^m$ exists and is unique, hence defines a unique function $u_v$.

By finiteness, every node is reached after finitely many steps, so all $u_v$ exist and are unique. Applying this to the root $r$ yields existence and uniqueness of $u_r$ and thus of the induced ranking. $\qquad\square$

## 12.4 Iterated Multi-Attribute Decision-Making

Iterated Multi-Attribute Decision-Making composes MADM scoring operators along an attribute hierarchy, recursively aggregating lower-level attribute scores into a root score for ranking alternatives.

**Definition 12.4.1** (Attribute/objective tree). A *finite rooted tree* $T = (V, E, r)$ is used to represent nested *attributes* or *objectives*. For each node $v \in V$, let $\mathrm{Ch}(v)$ be its (finite) set of children. Leaves are nodes with $\mathrm{Ch}(v) = \varnothing$.

**Definition 12.4.2** (MADM scoring operator). Let $\mathcal{A} = \{A_1, \ldots, A_m\}$ be a finite nonempty set of alternatives and let $k \in \mathbb{N}$. A *MADM scoring operator* of arity $k$ is a total function

$$\mathcal{M} : \ \mathbb{R}^{m \times k} \times \Delta_k \longrightarrow \mathbb{R}^m, \qquad \Delta_k := \Big\{ w \in [0,1]^k : \sum_{\ell=1}^{k} w_\ell = 1 \Big\}.$$

Given an attribute-performance matrix $X \in \mathbb{R}^{m \times k}$ and weights $w \in \Delta_k$, $\mathcal{M}(X, w)$ returns one real score per alternative.

**Definition 12.4.3** (Iterated Multi-Attribute Decision-Making (I-MADM)). Let $\mathcal{A} = \{A_1, \ldots, A_m\}$ be alternatives and let $T = (V, E, r)$ be an *attribute tree*. An *Iterated Multi-Attribute Decision-Making* (I-MADM) instance is a collection

$$\text{I-MADM} := \Big( \mathcal{A}, \ T, \ \{u_\ell\}_{\ell \in L}, \ \{\mathcal{M}_v, w^{(v)}\}_{v \in V \setminus L} \Big),$$

where $L \subseteq V$ is the set of leaves, such that:

1. Each leaf $\ell \in L$ (an atomic attribute) has a base score

$$u_\ell : \mathcal{A} \to \mathbb{R}.$$

2. Each internal node $v \in V \setminus L$ with $k_v := |\mathrm{Ch}(v)|$ has an MADM scoring operator

$$\mathcal{M}_v : \mathbb{R}^{m \times k_v} \times \Delta_{k_v} \to \mathbb{R}^m$$

and a weight vector $w^{(v)} \in \Delta_{k_v}$ on its children.

**Recursive semantics.** Define scores $u_v : \mathcal{A} \to \mathbb{R}$ for all nodes $v \in V$ recursively:

1. If $v$ is a leaf, $u_v$ is given (base attribute score).

2. If $v$ is internal with $\mathrm{Ch}(v) = \{c_1, \ldots, c_{k_v}\}$, form $X^{(v)} \in \mathbb{R}^{m \times k_v}$ by $X_{i\ell}^{(v)} = u_{c_\ell}(A_i)$ and set

$$(u_v(A_1), \ldots, u_v(A_m))^\top := \mathcal{M}_v(X^{(v)}, w^{(v)}).$$

The final I-MADM score is $u_r$ at the root, yielding a ranking by $A_i \succeq A_k \iff u_r(A_i) \geq u_r(A_k)$.

**Theorem 12.4.4** (Well-definedness of I-MADM). *Under Theorem 12.4.3, if $T$ is finite and each $\mathcal{M}_v$ is total, then for every node $v \in V$ the score $u_v$ exists and is unique; in particular, $u_r$ is unique.*

*Proof.* Proceed by induction from leaves to the root. Leaf scores are given. Assuming child scores $u_c$ are uniquely determined, the matrix $X^{(v)}$ is uniquely determined. Totality of $\mathcal{M}_v$ implies $\mathcal{M}_v(X^{(v)}, w^{(v)}) \in \mathbb{R}^m$ exists uniquely, hence defines $u_v$ uniquely. Finiteness of $T$ ensures termination at the root. $\square$

## 12.5   Iterated Multi-Objective Decision-Making

Iterated Multi-Objective Decision-Making composes objective-aggregation operators along a hierarchical objective tree, recursively producing a final scalar objective, then solving it to obtain optimal decisions.

**Definition 12.5.1** (MODM objective-aggregation operator). Let $\mathcal{X} \subseteq \mathbb{R}^d$ be a nonempty feasible decision set. A *MODM objective-aggregation operator* of arity $k$ is a total function

$$\mathcal{G} : (\mathbb{R}^{\mathcal{X}})^k \times \Delta_k \longrightarrow \mathbb{R}^{\mathcal{X}},$$

where $(\mathbb{R}^{\mathcal{X}})^k$ denotes $k$-tuples of real-valued functions on $\mathcal{X}$. Given objective functions $f_1, \ldots, f_k : \mathcal{X} \to \mathbb{R}$ and weights $w \in \Delta_k$, $\mathcal{G}(f_1, \ldots, f_k; w)$ returns an aggregated objective $F : \mathcal{X} \to \mathbb{R}$ (to be minimized or maximized as specified).

**Definition 12.5.2** (Iterated Multi-Objective Decision-Making (I-MODM)). Let $\mathcal{X} \subseteq \mathbb{R}^d$ be a nonempty feasible set and let $T = (V, E, r)$ be an *objective tree*. An *Iterated Multi-Objective Decision-Making* (I-MODM) instance is a collection

$$\textsf{I-MODM} := \Big( \mathcal{X}, \, T, \, \{f_\ell\}_{\ell \in L}, \, \{\mathcal{G}_v, w^{(v)}\}_{v \in V \setminus L}, \, \textsf{Solve} \Big),$$

where $L$ is the leaf set, such that:

1. Each leaf $\ell \in L$ has an atomic objective $f_\ell : \mathcal{X} \to \mathbb{R}$.

2. Each internal node $v$ with $k_v := |\mathrm{Ch}(v)|$ has an objective-aggregation operator

$$\mathcal{G}_v : (\mathbb{R}^{\mathcal{X}})^{k_v} \times \Delta_{k_v} \to \mathbb{R}^{\mathcal{X}}$$

and weights $w^{(v)} \in \Delta_{k_v}$.

3. Solve is a fixed optimization rule that maps a scalar objective $F : \mathcal{X} \to \mathbb{R}$ to a nonempty solution set $\mathcal{X}^\star(F) \subseteq \mathcal{X}$ (e.g. argmin or argmax), whenever such a set exists.

**Recursive semantics.** Define aggregated objectives $F_v : \mathcal{X} \to \mathbb{R}$ for all $v \in V$ recursively:

1. If $v$ is a leaf, set $F_v := f_v$.

2. If $v$ is internal with $\mathrm{Ch}(v) = \{c_1, \ldots, c_{k_v}\}$, set

$$F_v := \mathcal{G}_v(F_{c_1}, \ldots, F_{c_{k_v}}; w^{(v)}).$$

The final scalar objective is $F_r$ at the root, and the I-MODM output is the solution set

$$\mathcal{X}^\star := \mathsf{Solve}(F_r) \subseteq \mathcal{X}.$$

**Theorem 12.5.3** (Well-definedness of I-MODM)**.** *Under Theorem 12.5.2, if $T$ is finite, each $\mathcal{G}_v$ is total, and* $\mathsf{Solve}(F_r)$ *is defined (i.e. returns a nonempty set), then:*

1. *each aggregated objective $F_v$ exists and is unique for all $v \in V$;*

2. *the final objective $F_r$ is unique; hence the output solution set $\mathcal{X}^\star$ is uniquely determined.*

*Proof.* Induct from leaves upward. Leaves give objectives $F_v$ uniquely. Assuming objectives $F_c$ are uniquely defined for all children $c \in \mathrm{Ch}(v)$, totality of $\mathcal{G}_v$ gives a unique function $F_v = \mathcal{G}_v(F_{c_1}, \ldots, F_{c_{k_v}}; w^{(v)})$. Finiteness of $T$ ensures the recursion terminates at the root, yielding a unique $F_r$. Applying the fixed rule Solve to this unique $F_r$ yields a uniquely determined output set. □

## 12.6 Analytic SuperHyperNetwork Process (ASHNP)

In this section, we consider extending the Analytic Network Process (ANP) by using hypergraphs and superhypergraphs. Hypergraph generalizes graphs: vertices with hyperedges as nonempty vertex subsets, enabling single edges to connect multiple vertices, modeling higher-order relations [1424–1426]. Hypernetwork is a (weighted) hypergraph: nodes and hyperedges with a weight/attribute function on hyperedges, representing multi-node interactions with strengths or confidences [1427]. $n$-SuperHyperGraph uses vertices drawn from the $n$-th iterated powerset of a base set; edges are nonempty families of such supervertices, capturing hierarchy [215, 1428–1432]. Related concepts, such as directed superhypergraphs [1433, 1434] and Meta-SuperHyperGraph [1435–1437] are also known. $n$-SuperHyperNetwork is a weighted $n$-SuperHyperGraph: supernodes from the $n$-th powerset and weighted hyperedges over them, encoding hierarchical interactions and strengths [1438, 1439]. We also discuss in the Appendix a concept that can generalize arbitrary structures such as graphs and hypergraphs; please refer to it as needed.

Analytic HyperNetwork Process (AHNP) extends ANP to directed hypernetworks; pairwise comparisons yield local priorities, assembled into a hyper-supermatrix whose stationary distribution provides global node weights. ASHNP generalizes AHNP to n-supernodes from iterated powersets; superhyperarcs encode hierarchical interactions; eigenvector priorities build a superhyper-supermatrix, producing global importance rankings.

**Definition 12.6.1** (Iterated powerset and iterated nonempty powerset). (cf. [1440]) Let $H$ be a set and let $\mathcal{P}(H)$ denote its (ordinary) powerset.

(1) Iterated powerset. Define the iterated powerset operator $\mathcal{P}^n(H)$ for $n \in \mathbb{N}_0$ by

$$\mathcal{P}^0(H) := H, \qquad \mathcal{P}^{n+1}(H) := \mathcal{P}\big(\mathcal{P}^n(H)\big) \quad (n \in \mathbb{N}_0).$$

(2) Iterated nonempty powerset. Let $\mathcal{P}_*(X) := \mathcal{P}(X) \setminus \{\varnothing\}$ be the nonempty powerset of a set $X$. Define $\mathcal{P}_*^n(H)$ for $n \in \mathbb{N}_0$ by

$$\mathcal{P}_*^0(H) := H, \qquad \mathcal{P}_*^{n+1}(H) := \mathcal{P}_*\big(\mathcal{P}_*^n(H)\big) = \mathcal{P}\big(\mathcal{P}_*^n(H)\big) \setminus \{\varnothing\} \quad (n \in \mathbb{N}_0).$$

**Definition 12.6.2** (Hypergraph [1425, 1441]). A *(finite) hypergraph* is an ordered pair $H = (V(H), E(H))$ where

- $V(H)$ is a finite nonempty set of *vertices*;

- $E(H) \subseteq \mathcal{P}(V(H)) \setminus \{\varnothing\}$ is a finite set of *hyperedges*.

Thus each hyperedge $e \in E(H)$ is a nonempty subset of $V(H)$ (allowing one edge to connect any number of vertices).

**Definition 12.6.3** ($n$-SuperHyperGraph [215, 1442]). Let $V_0$ be a finite nonempty *base* vertex set and let $n \in \mathbb{N}_0$. Using the iterated powerset from the previous definition, an *$n$-SuperHyperGraph over $V_0$* is a pair

$$\mathbf{SHG}^{(n)} = (V, E)$$

such that

$$\varnothing \neq V \subseteq \mathcal{P}^n(V_0) \quad \text{and} \quad E \subseteq \mathcal{P}(V) \setminus \{\varnothing\}.$$

Elements of $V$ are called *$n$-supervertices*, and elements of $E$ are called *$n$-superedges*.

Special cases. For $n = 0$, one recovers an ordinary hypergraph on a subset of the base set: $V \subseteq V_0$ and $E \subseteq \mathcal{P}(V) \setminus \{\varnothing\}$. For $n = 1$, vertices are subsets of $V_0$ and edges are nonempty families of such subsets.

Remark. Since $E \subseteq \mathcal{P}(V) \subseteq \mathcal{P}(\mathcal{P}^n(V_0)) = \mathcal{P}^{n+1}(V_0)$, every $n$-superedge can be viewed (canonically) as an element of the $(n + 1)$-st iterated powerset of $V_0$.

**Definition 12.6.4** (Hypernetwork [1443, 1444]). A *(weighted) hypernetwork* is a triple

$$\mathcal{N} = (V, \mathcal{E}, w)$$

where

- $V$ is a finite nonempty set of *nodes*;

- $\mathcal{E} \subseteq \mathcal{P}(V) \setminus \{\varnothing\}$ is a finite set of *hyperedges*;

- $w : \mathcal{E} \to \mathbb{R}_{\geq 0}$ is a *weight/attribute* function on hyperedges (omit $w$ for the unweighted case).

A *directed hypernetwork* can be modeled by replacing $\mathcal{E}$ with a set of ordered pairs

$$\mathcal{E} \subseteq \big(\mathcal{P}(V) \setminus \{\varnothing\}\big) \times \big(\mathcal{P}(V) \setminus \{\varnothing\}\big),$$

where each directed hyperedge is of the form $(T, H)$ (tail/head), or by any equivalent head–tail partition formalism. Optional node and hyperedge labelings $\ell_V : V \to L_V$ and $\ell_{\mathcal{E}} : \mathcal{E} \to L_{\mathcal{E}}$ may be added to encode types.

**Definition 12.6.5** ($n$-SuperHypernetwork [1444–1446])**.** Let $V_0$ be a finite nonempty base node set and let $n \in \mathbb{N}_0$. An *$n$-superhypernetwork* is a triple

$$\mathcal{N}^{(n)} = (V, \mathcal{E}, w)$$

where

- $V \subseteq \mathcal{P}^n(V_0)$ is a finite nonempty set of *$n$-supernodes*;

- $\mathcal{E} \subseteq \mathcal{P}(V) \setminus \{\varnothing\}$ is a finite set of *$n$-superhyperedges*;

- $w : \mathcal{E} \to \mathbb{R}_{\geq 0}$ is an optional weight function.

Equivalently, an $n$-superhypernetwork is an $n$-SuperHyperGraph endowed with a hyperedge-weight function. Moreover, $\mathcal{E} \subseteq \mathcal{P}(V) \subseteq \mathcal{P}^{n+1}(V_0)$ holds automatically.

**Definition 12.6.6** (Positive reciprocal pairwise-comparison matrix)**.** A matrix $M = (m_{pq}) \in \mathbb{R}_{>0}^{k \times k}$ is called *positive reciprocal* if

$$m_{pp} = 1, \qquad m_{pq} = \frac{1}{m_{qp}} \quad (p \neq q).$$

**Lemma 12.6.7** (Perron priority vector)**.** *Let $M \in \mathbb{R}_{>0}^{k \times k}$ be a positive matrix (in particular, any matrix in Theorem 12.6.6 is positive). Then there exists an eigenvalue $\lambda_{\max} > 0$ and an eigenvector $v \in \mathbb{R}_{>0}^k$ such that $Mv = \lambda_{\max} v$. Moreover, $v$ is unique up to a positive scalar factor. Hence the normalized vector*

$$p(M) := \frac{v}{\mathbf{1}^\top v} \in \Delta_k, \qquad \Delta_k := \{x \in \mathbb{R}_{\geq 0}^k : \ \mathbf{1}^\top x = 1\},$$

*is uniquely determined by $M$.*

*Proof.* This is a standard consequence of the Perron–Frobenius theorem for positive matrices: a positive matrix has a unique (up to scaling) positive eigenvector associated with its spectral radius. Normalization by $\mathbf{1}^\top v$ fixes the scaling uniquely. $\qquad \square$

**Lemma 12.6.8** (Unique stationary distribution for primitive column-stochastic matrices)**.** *Let $W \in \mathbb{R}_{\geq 0}^{m \times m}$ be column-stochastic, i.e. $\mathbf{1}^\top W = \mathbf{1}^\top$, and assume $W$ is primitive (there exists $t \in \mathbb{N}$ such that $W^t$ has all entries strictly positive). Then there exists a unique vector $\pi \in \Delta_m$ such that*

$$W\pi = \pi,$$

*and for every $x \in \Delta_m$ one has $W^k x \to \pi$ as $k \to \infty$.*

*Proof.* Primitivity implies, by Perron–Frobenius, that eigenvalue 1 is simple and has a unique positive right eigenvector. Normalizing it to sum to 1 yields the unique $\pi \in \Delta_m$. Convergence $W^k x \to \pi$ for all $x \in \Delta_m$ follows from the Perron decomposition for primitive stochastic matrices. □

**Definition 12.6.9** (Directed hypernetwork). A *directed hypernetwork* is a triple

$$\mathcal{H} = (V, \mathcal{E}, \gamma),$$

where $V$ is a finite nonempty set of nodes, $\mathcal{E}$ is a finite set of *directed hyperarcs*

$$e = (T_e, H_e), \qquad \varnothing \neq T_e \subseteq V, \ \varnothing \neq H_e \subseteq V,$$

and $\gamma : \mathcal{E} \to \mathbb{R}_{>0}$ is an optional hyperarc weight (omit $\gamma$ if unweighted). Intuitively, the tail set $T_e$ influences the head set $H_e$.

**Definition 12.6.10** (Analytic HyperNetwork Process (AHNP)). Let $\mathcal{A} = \{A_1, \ldots, A_m\}$ be a finite set of decision elements (criteria, subcriteria, alternatives, or any mix) represented as nodes of a directed hypernetwork $\mathcal{H} = (V, \mathcal{E}, \gamma)$ with $V = \mathcal{A}$.

An *Analytic HyperNetwork Process (AHNP)* instance consists of:

1. A directed hypernetwork $\mathcal{H} = (V, \mathcal{E}, \gamma)$ on $V = \mathcal{A}$.

2. For each hyperarc $e = (T_e, H_e) \in \mathcal{E}$ and each head node $h \in H_e$, a positive reciprocal matrix

$$M^{(e,h)} \in \mathbb{R}_{>0}^{|T_e| \times |T_e|}$$

   whose entries compare the relative influence of tail nodes in $T_e$ *with respect to* the head node $h$.

3. A fixed rule to assign a nonnegative *arc–head mixing weight* $\alpha_{e,h} \geq 0$ for each $(e, h)$ with $h \in H_e$. (For example, $\alpha_{e,h}$ may be set proportional to $\gamma(e)$ and then normalized per head node.)

**Local priorities.** For each $(e, h)$, define the *local priority vector*

$$p^{(e,h)} := p\Big(M^{(e,h)}\Big) \in \Delta_{|T_e|}$$

as in Theorem 12.6.7. Index $p^{(e,h)}$ by tail nodes in $T_e$.

**Global influence (hyper-supermatrix).** Define the raw influence matrix $S = (s_{ih})_{i,h \in V} \in \mathbb{R}_{\geq 0}^{m \times m}$ by

$$s_{ih} := \sum_{\substack{e = (T_e, H_e) \in \mathcal{E} \\ h \in H_e, \ i \in T_e}} \alpha_{e,h} \, p^{(e,h)}(i),$$

where $p^{(e,h)}(i)$ denotes the component corresponding to $i \in T_e$.

To obtain a column-stochastic *hyper-supermatrix* $W = (w_{ih})$, set for each column $h$:

$$w_{ih} := \begin{cases} \dfrac{s_{ih}}{\sum_{u \in V} s_{uh}}, & \text{if } \sum_{u \in V} s_{uh} > 0, \\ \delta_{ih}, & \text{if } \sum_{u \in V} s_{uh} = 0, \end{cases}$$

where $\delta_{ih}$ is the Kronecker delta (a self-loop fallback to avoid a zero column).

**Final priorities and output.** Define the AHNP priority vector $\pi \in \Delta_m$ as any solution of

$$W\pi = \pi.$$

If a distinguished subset $V_{\text{alt}} \subseteq V$ represents alternatives, rank alternatives by descending values of $\pi$ restricted to $V_{\text{alt}}$ (renormalize on $V_{\text{alt}}$ if desired).

**Theorem 12.6.11** (Well-definedness of AHNP). *Assume an AHNP instance as in Theorem 12.6.10. Then:*

1. *All local priority vectors $p^{(e,h)}$ are well-defined and unique.*

2. *The hyper-supermatrix $W$ is well-defined and column-stochastic.*

3. *If $W$ is primitive, then the stationary priority vector $\pi \in \Delta_m$ exists, is unique, and $W^k x \to \pi$ for every $x \in \Delta_m$.*

*Proof.* (1) Each $M^{(e,h)}$ is positive reciprocal, hence positive; apply Theorem 12.6.7.

(2) The entries $s_{ih}$ are finite sums of nonnegative terms, hence well-defined and nonnegative. Column normalization defines $w_{ih} \geq 0$ and ensures $\sum_i w_{ih} = 1$ for each $h$; if the raw column sum is 0, the fallback column $\delta_{ih}$ also sums to 1. Thus $\mathbf{1}^\top W = \mathbf{1}^\top$.

(3) If $W$ is primitive, apply Theorem 12.6.8 to obtain existence/uniqueness of $\pi$ and convergence of powers. $\square$

**Definition 12.6.12** (Analytic SuperHyperNetwork Process (ASHNP)). Fix a finite nonempty base node set $V_0$ and an integer $n \in \mathbb{N}_0$. Let $V \subseteq \mathcal{P}^n(V_0)$ be a finite nonempty set of $n$-*supernodes*. Let $\mathcal{H}^{(n)} = (V, \mathcal{E}, \gamma)$ be a directed hypernetwork on the supernode set $V$ (in the sense of Theorem 12.6.9), whose hyperarcs are of the form $e = (T_e, H_e)$ with $\varnothing \neq T_e, H_e \subseteq V$.

An *Analytic SuperHyperNetwork Process (ASHNP)* instance consists of:

1. The directed superhypernetwork $\mathcal{H}^{(n)} = (V, \mathcal{E}, \gamma)$.

2. For each $e = (T_e, H_e) \in \mathcal{E}$ and each head supernode $h \in H_e$, a positive reciprocal matrix

$$M^{(e,h)} \in \mathbb{R}_{>0}^{|T_e| \times |T_e|}$$

comparing supernodes in $T_e$ *with respect to* the head supernode $h$.

3. Nonnegative mixing weights $\alpha_{e,h} \geq 0$ assigned to each $(e, h)$ with $h \in H_e$.

**Computation.** Define local priorities $p^{(e,h)} := p(M^{(e,h)})$, build the raw influence matrix $S = (s_{ih})_{i,h \in V}$ by

$$s_{ih} := \sum_{\substack{e=(T_e, H_e) \in \mathcal{E} \\ h \in H_e, \ i \in T_e}} \alpha_{e,h} \, p^{(e,h)}(i),$$

and column-normalize it (with the same zero-column fallback) to obtain a column-stochastic superhyper-supermatrix $W$ on the index set $V$. Define the ASHNP priority vector $\pi \in \Delta_{|V|}$ by $W\pi = \pi$ and rank the designated alternative supernodes accordingly.

**Theorem 12.6.13** (Well-definedness of ASHNP). *Assume an ASHNP instance as in Theorem 12.6.12. Then:*

1. *All local priority vectors $p^{(e,h)}$ are well-defined and unique.*

2. *The superhyper-supermatrix $W$ is well-defined and column-stochastic.*

3. *If $W$ is primitive, then the stationary priority vector $\pi \in \Delta_{|V|}$ exists, is unique, and $W^k x \to \pi$ for every $x \in \Delta_{|V|}$.*

*Proof.* Identical to the proof of Theorem 12.6.11, since the construction depends only on the finiteness of the node set and positivity/reciprocity of the pairwise-comparison matrices. The fact that $V \subseteq \mathcal{P}^n(V_0)$ changes only the interpretation of nodes, not the algebra. □

## 12.7 Analytic Recursive SuperHyperNetwork Process (ARSHNP)

Recursive SuperHyperGraph allows supervertices from an iterated powerset, while edges are recursive set-objects drawn from a depth-$k$ powerset universe, capturing nested, multi-level hyper-relations without membership cycles [1447–1449]. Recursive SuperHyperNetwork is a weighted recursive SuperHyperGraph: it attaches nonnegative weights/attributes to recursive superhyperedges, enabling quantified strengths or confidences for hierarchical, nested interactions among supernodes. Analytic Recursive SuperHyperNetwork Process (ARSHNP) extends AHNP to recursive superhypernetworks: pairwise comparisons on supports yield local priorities, assembled into a column-stochastic supermatrix whose stationary vector provides global node importances.

**Definition 12.7.1** (Atomization convention). Throughout, the node set $V$ is treated as a set of *atoms* (urelements): elements of $V$ are not decomposed by the membership relation $\in$ in the recursive constructions below. Equivalently, one may replace $V$ by a tagged copy

$$\mathsf{At}(V) := \{(0, v) \mid v \in V\},$$

and work over $\mathsf{At}(V)$. In either case, all recursive set-constructions are built *over* these atoms.

**Definition 12.7.2** (Depth-$k$ powerset universe). (cf. [1450,1451]) Let $V$ be a nonempty set of atoms and let $k \in \mathbb{N}_0$. Define $(S_i)_{i \geq 0}$ by

$$S_0 := V, \qquad S_i := \mathcal{P}\Big(\bigcup_{j=0}^{i-1} S_j\Big) \quad (i \geq 1),$$

and define the *object universe up to depth $k$* by

$$\mathsf{Obj}(V,k) := \bigcup_{i=0}^{k} S_i.$$

The *depth-$k$ powerset universe* generated by $V$ is

$$2_{V,k} \; := \; \mathcal{P}\big(\mathsf{Obj}(V,k)\big).$$

**Lemma 12.7.3** (Finite depth and canonical support). *Let $V$ be a set of atoms and $k \in \mathbb{N}_0$. Every $x \in \mathsf{Obj}(V,k)$ admits a finite structural depth $\mathrm{d}(x) \in \{0, 1, \ldots, k\}$ and a well-defined support $\mathrm{Supp}(x) \subseteq V$.*

(i) Atoms. *For $v \in V$, define*

$$\mathrm{d}(v) := 0, \qquad \mathrm{Supp}(v) := \{v\}.$$

(ii) Set-objects. *For $x \in \mathsf{Obj}(V,k) \setminus V$ (so $x$ is a set of objects), define*

$$\mathrm{d}(x) := 1 + \max\Big(\big\{\, \mathrm{d}(y) \mid y \in x \,\big\} \cup \{0\}\Big), \qquad \mathrm{Supp}(x) := \bigcup_{y \in x} \mathrm{Supp}(y).$$

*Then $\mathrm{d}(x) \leq k$. Moreover, $\mathrm{Supp}(x) \neq \varnothing$ whenever $x \neq \varnothing$.*

*Proof.* By construction, $S_0 = V$ consists of atoms, hence has depth 0. If $x \in S_i$ for some $i \geq 1$, then $x \subseteq \bigcup_{j=0}^{i-1} S_j$, so every $y \in x$ lies in $S_{j(y)}$ for some $j(y) \leq i - 1$. Thus the recursion is well-founded, and $\mathrm{d}(y)$ is defined for all $y \in x$ before defining $\mathrm{d}(x)$. Induction on the least $i$ with $x \in S_i$ yields $\mathrm{d}(x) \leq i \leq k$. If $x \neq \varnothing$, it contains some $y$, and $\mathrm{Supp}(y) \neq \varnothing$; hence $\mathrm{Supp}(x) = \bigcup_{y \in x} \mathrm{Supp}(y) \neq \varnothing$. $\qquad \square$

**Definition 12.7.4** (Support of a recursive set). For any $X \in 2_{V,k}$ (so $X \subseteq \mathsf{Obj}(V,k)$), define its support by

$$\mathrm{Supp}(X) := \bigcup_{x \in X} \mathrm{Supp}(x) \subseteq V,$$

where $\mathrm{Supp}(x)$ is as in Theorem 12.7.3.

**Definition 12.7.5** ($k$-recursive hypergraph). (cf. [1450, 1451]) Let $V$ be a finite nonempty set of atoms and let $k \in \mathbb{N}_0$. A *$k$-recursive hypergraph* is a pair

$$H = (V, E)$$

such that

$$E \subseteq 2_{V,k} \setminus \{\varnothing\},$$

where $2_{V,k}$ is the depth-$k$ powerset universe from Theorem 12.7.2. For $k = 0$, one has $2_{V,0} = \mathcal{P}(V)$, hence $E \subseteq \mathcal{P}(V) \setminus \{\varnothing\}$ and $H$ is an ordinary hypergraph.

**Definition 12.7.6** ($(n,k)$-recursive SuperHyperGraph). Let $V_0$ be a finite nonempty base set and let $n, k \in \mathbb{N}_0$. Define iterated powersets by

$$\mathcal{P}^0(V_0) := V_0, \qquad \mathcal{P}^{n+1}(V_0) := \mathcal{P}\big(\mathcal{P}^n(V_0)\big).$$

A *$(n,k)$-recursive SuperHyperGraph* is a pair

$$\mathrm{RSHG}^{(n,k)} = (V, E)$$

such that

$$\varnothing \neq V \subseteq \mathcal{P}^n(V_0) \quad \text{(finite)}, \qquad E \subseteq 2_{V,k} \setminus \{\varnothing\} \quad \text{(finite)},$$

where $2_{V,k}$ is generated from $V$ as in Theorem 12.7.2. In addition, $V$ is interpreted as a set of atoms in the sense of Theorem 12.7.1.

**Definition 12.7.7** (($n, k$)-Recursive SuperHyperNetwork). Let $V_0$ be a finite nonempty base set and let $n, k \in \mathbb{N}_0$. Let $\mathrm{RSHG}^{(n,k)} = (V, E)$ be an $(n, k)$-recursive SuperHyperGraph as in Theorem 12.7.6. A *($n, k$)-recursive SuperHyperNetwork* is a triple

$$\mathcal{N}^{(n,k)} = (V, E, w)$$

where $w : E \to \mathbb{R}_{\geq 0}$ is an optional weight/attribute function (omit $w$ if unweighted).

**Theorem 12.7.8** (Well-definedness of recursive SuperHyperNetworks). *Let $\mathcal{N}^{(n,k)} = (V, E, w)$ be as in Theorem 12.7.7. Then:*

1. *For every recursive superhyperedge $e \in E$, the support $\mathrm{Supp}(e) \subseteq V$ is well-defined.*

2. *The* flattened *(ordinary) weighted hypernetwork*

$$\underline{\mathcal{N}} := (V, \underline{E}, \underline{w})$$

   *is well-defined, where*

$$\underline{E} := \{\mathrm{Supp}(e) \mid e \in E\} \subseteq \mathcal{P}(V) \setminus \{\varnothing\}, \qquad \underline{w}(F) := \sum_{\substack{e \in E \\ \mathrm{Supp}(e) = F}} w(e) \quad (F \in \underline{E}).$$

*Proof.* (1) Since $e \in 2_{V,k}$, one has $e \subseteq \mathsf{Obj}(V, k)$, hence $\mathrm{Supp}(x)$ is defined for each $x \in e$ by Theorem 12.7.3. Thus $\mathrm{Supp}(e) = \bigcup_{x \in e} \mathrm{Supp}(x)$ is well-defined and is a subset of $V$. If $e \neq \varnothing$, then by Theorem 12.7.3 the support of some element is nonempty, so $\mathrm{Supp}(e) \neq \varnothing$.

(2) Therefore each $\mathrm{Supp}(e)$ lies in $\mathcal{P}(V) \setminus \{\varnothing\}$, so $\underline{E}$ is well-defined and finite. For each $F \in \underline{E}$, $\underline{w}(F)$ is a finite sum of nonnegative reals, hence well-defined. $\qquad \square$

**Definition 12.7.9** (Positive reciprocal matrix and Perron priority). Let $r \in \mathbb{N}$. A matrix $M = (m_{pq}) \in \mathbb{R}_{>0}^{r \times r}$ is *positive reciprocal* if

$$m_{pp} = 1, \qquad m_{pq} = \frac{1}{m_{qp}} \ (p \neq q).$$

Let $\Delta_r := \{x \in \mathbb{R}_{\geq 0}^r : \mathbf{1}^\top x = 1\}$. For such $M$, let $p(M) \in \Delta_r$ denote the normalized Perron (principal) right eigenvector; it exists and is unique after normalization (Perron–Frobenius).

**Definition 12.7.10** (Directed recursive superhypernetwork). A *directed recursive superhypernetwork* is a triple

$$\overrightarrow{\mathcal{N}}^{(n,k)} = (V, \overrightarrow{E}, \gamma),$$

where $V$ is as in Theorem 12.7.7,

$$\overrightarrow{E} \subseteq \left( 2_{V,k} \setminus \{\varnothing\} \right) \times \left( 2_{V,k} \setminus \{\varnothing\} \right)$$

is a finite set of *directed recursive superhyperarcs* $e = (T_e, H_e)$, and $\gamma : \overrightarrow{E} \to \mathbb{R}_{>0}$ is an optional arc weight. Define the *tail support* and *head support* by

$$\underline{T}_e := \mathrm{Supp}(T_e) \subseteq V, \qquad \underline{H}_e := \mathrm{Supp}(H_e) \subseteq V,$$

which are nonempty by Theorem 12.7.3 and Theorem 12.7.4.

**Definition 12.7.11** (Analytic Recursive SuperHyperNetwork Process (ARSHNP)). Let $\overrightarrow{\mathcal{N}}^{(n,k)} = (V, \overrightarrow{E}, \gamma)$ be a directed recursive superhypernetwork. An *ARSHNP* instance consists of:

1. For each arc $e = (T_e, H_e) \in \overrightarrow{E}$ and each head node $h \in \underline{H}_e$, a positive reciprocal pairwise-comparison matrix
$$M^{(e,h)} \in \mathbb{R}_{>0}^{|\underline{T}_e| \times |\underline{T}_e|}$$
comparing tail nodes in $\underline{T}_e$ *with respect to* head node $h$.

2. Nonnegative mixing weights $\alpha_{e,h} \geq 0$ for each $(e,h)$ with $h \in \underline{H}_e$ (e.g., proportional to $\gamma(e)$ and normalized per head node).

**Local priorities.** For each $(e,h)$, define $p^{(e,h)} := p(M^{(e,h)}) \in \Delta_{|\underline{T}_e|}$ and index its components by nodes in $\underline{T}_e$.

**Recursive hyper-supermatrix.** Define $S = (s_{ih})_{i,h \in V} \in \mathbb{R}_{\geq 0}^{|V| \times |V|}$ by
$$s_{ih} := \sum_{\substack{e \in \overrightarrow{E} \\ h \in \underline{H}_e, \ i \in \underline{T}_e}} \alpha_{e,h} \, p^{(e,h)}(i).$$

Column-normalize $S$ to obtain a column-stochastic matrix $W = (w_{ih})$:
$$w_{ih} := \begin{cases} \dfrac{s_{ih}}{\sum_{u \in V} s_{uh}}, & \text{if } \sum_{u \in V} s_{uh} > 0, \\ \delta_{ih}, & \text{if } \sum_{u \in V} s_{uh} = 0, \end{cases}$$
where $\delta_{ih}$ is the Kronecker delta.

**Global priorities.** Any $\pi \in \Delta_{|V|}$ satisfying
$$W\pi = \pi$$
is called an *ARSHNP priority vector*. If a distinguished subset $V_{\text{alt}} \subseteq V$ represents alternatives, rank them by the corresponding components of $\pi$ (renormalizing on $V_{\text{alt}}$ if desired).

**Theorem 12.7.12** (Well-definedness of ARSHNP). *Assume an ARSHNP instance as in Theorem 12.7.11. Then:*

1. *Tail/head supports $\underline{T}_e, \underline{H}_e$ are well-defined nonempty subsets of $V$ for all arcs $e \in \overrightarrow{E}$.*

2. *Each local priority vector $p^{(e,h)}$ is well-defined and unique.*

3. *The matrix $W$ is well-defined, nonnegative, and column-stochastic.*

4. *There exists at least one $\pi \in \Delta_{|V|}$ such that $W\pi = \pi$. If, in addition, $W$ is primitive, then this $\pi$ is unique and strictly positive.*

*Proof.* (1) Since $T_e, H_e \in 2_{V,k} \setminus \{\varnothing\}$, supports $\mathrm{Supp}(T_e), \mathrm{Supp}(H_e)$ are defined by Theorem 12.7.4. Nonemptiness follows from $T_e, H_e \neq \varnothing$ and Theorem 12.7.3.

(2) Each $M^{(e,h)}$ is positive reciprocal, hence positive. By Perron–Frobenius, it has a unique normalized Perron vector, so $p^{(e,h)}$ is well-defined and unique.

(3) Each $s_{ih}$ is a finite sum of nonnegative reals, hence well-defined and $s_{ih} \geq 0$. Column-normalization gives $w_{ih} \geq 0$ and $\sum_{i \in V} w_{ih} = 1$ for each $h$; the fallback column $\delta_{ih}$ also sums to 1. Thus $W$ is column-stochastic.

(4) Since $W$ is column-stochastic, $\|W\|_1 = 1$, hence its spectral radius satisfies $\rho(W) \leq 1$. Also $\mathbf{1}^\top W = \mathbf{1}^\top$ implies 1 is an eigenvalue of $W^\top$, hence 1 is an eigenvalue of $W$. Therefore $\rho(W) = 1$, and by Perron–Frobenius for nonnegative matrices, $W$ admits a nonnegative right eigenvector for eigenvalue 1. Normalizing it to sum to 1 yields $\pi \in \Delta_{|V|}$ with $W\pi = \pi$. If $W$ is primitive, the Perron eigenvector is unique up to scaling and strictly positive; normalization makes it unique and strictly positive. $\qquad\square$

## 12.8 SuperHyperDecision-Making

SuperHyperDecision-Making defines hierarchical decision states drawn from an n-th powerset, generates reachable sets via set-valued combination rules, and selects maximal outcomes under a preorder.

**Definition 12.8.1** (SuperHyperdecision-making system (level $n$))**.** Fix an integer $n \geq 1$ and a nonempty set $D$ of atomic decision items. Let $\Omega \subseteq \mathcal{P}^n(D)$ be a nonempty *working universe* (typically finite), and fix an arity $k \geq 1$.

A *level-n superhypercombination* is a set-valued map

$$\star : \Omega^k \longrightarrow \mathcal{P}(\Omega) \setminus \{\varnothing\}.$$

Given a nonempty seed $S_0 \subseteq \Omega$, define the *immediate-consequence operator* $T_\star : \mathcal{P}(\Omega) \to \mathcal{P}(\Omega)$ by

$$T_\star(S) := S \ \cup \ \bigcup \Big\{ \star(X_1, \ldots, X_k) \ \Big| \ X_1, \ldots, X_k \in S \Big\}, \qquad S \subseteq \Omega,$$

and define the *reachable (least $\star$-closed) set* containing $S_0$ by

$$S^* := \bigcup_{t=0}^{\infty} T_\star^t(S_0), \qquad T_\star^0(S_0) := S_0, \ \ T_\star^{t+1}(S_0) := T_\star\big(T_\star^t(S_0)\big).$$

Let $\succeq$ be a preorder on $\Omega$ (reflexive and transitive). The set of *SuperHyperdecision outcomes* is

$$\mathrm{Opt}_{\star,\succeq}(S_0) := \mathrm{Max}_{\succeq}(S^*) := \big\{ X \in S^* \ \big| \ \forall Y \in S^* : \ X \succeq Y \big\}.$$

**Theorem 12.8.2** (Well-definedness of SuperHyperdecision-making)**.** *Assume Theorem 12.8.1 and, in addition, that $\Omega$ is finite. Then:*

1. *$T_\star : \mathcal{P}(\Omega) \to \mathcal{P}(\Omega)$ is well-defined and monotone (i.e. $S \subseteq S' \Rightarrow T_\star(S) \subseteq T_\star(S')$).*

2. *The reachable set $S^*$ exists, is nonempty, satisfies $S_0 \subseteq S^* \subseteq \Omega$, and is the least $T_\star$-closed subset of $\Omega$ containing $S_0$:*

$$T_\star(S^*) = S^*, \qquad (\forall S \subseteq \Omega) \ (S_0 \subseteq S \ \& \ T_\star(S) \subseteq S) \Rightarrow S^* \subseteq S.$$

*Moreover the chain stabilizes after finitely many steps: there exists $N \leq |\Omega|$ such that $T_\star^N(S_0) = S^*$.*

3. *For every preorder $\succeq$ on $\Omega$, the outcome set $\mathrm{Opt}_{\star,\succeq}(S_0)$ is well-defined and nonempty.*

*Proof.* (1) Let $S \subseteq \Omega$. For any $(X_1, \ldots, X_k) \in S^k$, the value $\star(X_1, \ldots, X_k)$ is a nonempty subset of $\Omega$ by definition, hence

$$\bigcup \{\star(X_1, \ldots, X_k) \mid X_1, \ldots, X_k \in S\} \subseteq \Omega.$$

Therefore $T_\star(S) \subseteq \Omega$, so $T_\star$ is well-defined. If $S \subseteq S'$, then $S^k \subseteq (S')^k$, hence the union taken over $S^k$ is contained in the union taken over $(S')^k$, giving $T_\star(S) \subseteq T_\star(S')$.

(2) The sequence $S_0 \subseteq T_\star(S_0) \subseteq T_\star^2(S_0) \subseteq \cdots$ is increasing by (1), so $S^*$ is well-defined as its union and is nonempty because it contains $S_0$. Clearly $S^* \subseteq \Omega$ since each iterate is a subset of $\Omega$. To show $T_\star(S^*) = S^*$, use monotonicity: $T_\star(T_\star^t(S_0)) = T_\star^{t+1}(S_0) \subseteq S^*$ for all $t$, so $T_\star(S^*) \subseteq S^*$. Conversely, $S^* \subseteq T_\star(S^*)$ because $T_\star$ is extensive ($S \subseteq T_\star(S)$ by definition), hence equality. Minimality follows because any $T_\star$-closed $S$ containing $S_0$ contains all iterates $T_\star^t(S_0)$ by induction, hence contains their union $S^*$. If $\Omega$ is finite, the increasing chain stabilizes in at most $|\Omega|$ strict growth steps, so $T_\star^N(S_0) = S^*$ for some $N \leq |\Omega|$.

(3) Since $\Omega$ is finite, $S^*$ is finite and nonempty. Every preorder on a finite nonempty set admits at least one maximal element. Hence $\mathrm{Max}_\succeq(S^*)$ is nonempty and well-defined. $\qquad \square$

## 12.9 Uncertain SuperHyperDecision-Making

Uncertain SuperHyperDecision-Making extends this framework by valuing reachable states with uncertain degrees and score-induced preorders, yielding representation-invariant maximal outcomes under uncertainty.

**Definition 12.9.1** (Uncertain-model-induced uncertainty structure)**.** Let $M$ be an uncertain model with degree-domain $\mathrm{Dom}(M) \subseteq [0,1]^k$. Fix either:

(i) a preorder $\preceq_M$ on $\mathrm{Dom}(M)$, or (ii) a score map $\mathrm{sc}_M : \mathrm{Dom}(M) \to \mathcal{S}$ into a preordered set $(\mathcal{S}, \preceq_{\mathcal{S}})$.

Define an uncertainty structure

$$\mathbb{U}_M = (U, \sim, \mathcal{S}, \preceq_{\mathcal{S}}, \mathrm{sc})$$

by

$$U := \mathrm{Dom}(M), \qquad \sim := \text{equality on } U,$$

and:

- If (i) is chosen, set $\mathcal{S} := U$, $\preceq_{\mathcal{S}} := \preceq_M$, and $\mathrm{sc} := \mathrm{id}_U$.

- If (ii) is chosen, set sc := $sc_M$.

Then any U-set $\mathcal{U} = (X, \mu_M)$ induces a preorder on $X$ via

$$x \succeq y \iff sc(\mu_M(x)) \succeq_{\mathcal{S}} sc(\mu_M(y)).$$

**Definition 12.9.2** (Uncertain SuperHyperdecision-making (USHDM)). An *Uncertain SuperHyperdecision-making* (USHDM) instance is a tuple

$$\text{USHDM} := \big(D, n, \Omega, k, \star, S_0, \mathbb{U}, \text{Val}\big)$$

where $(D, n, \Omega, k, \star, S_0)$ satisfy Theorem 12.8.1, $\mathbb{U} = (U, \sim, \mathcal{S}, \preceq_{\mathcal{S}})$ is an uncertainty structure, and Val : $\Omega \to U$ is an uncertain evaluation map. The reachable set $S^*$ is computed exactly as in Theorem 12.8.1, and the *uncertain superhyperdecision outcomes* are defined by

$$\text{Opt}^{\text{unc}}_{\star, \text{Val}}(S_0) := \text{Max}_{\succeq_{\text{Val}}}(S^*),$$

where $\succeq_{\text{Val}}$ is the induced preorder.

**Theorem 12.9.3** (Well-definedness and representation invariance of USHDM). *Assume Theorem 12.9.2 and that $\Omega$ is finite. Then:*

1. *The set $\text{Opt}^{\text{unc}}_{\star, \text{Val}}(S_0)$ is well-defined and nonempty.*

2. *(*Representation invariance*) If* Val, Val' : $\Omega \to U$ *satisfy* Val$(X) \sim$ Val'$(X)$ *for all* $X \in \Omega$*, then*

$$\text{Opt}^{\text{unc}}_{\star, \text{Val}}(S_0) = \text{Opt}^{\text{unc}}_{\star, \text{Val}'}(S_0).$$

*Proof.* (1) By Theorem 12.8.2(2), the reachable set $S^*$ is a finite nonempty subset of $\Omega$. $\succeq_{\text{Val}}$ is a preorder (pullback of a preorder), so by the same finiteness argument as in Theorem 12.8.2(3), $\text{Max}_{\succeq_{\text{Val}}}(S^*)$ exists and is nonempty.

(2) If Val$(X) \sim$ Val'$(X)$ for all $X$, then sc(Val$(X)$) = sc(Val'$(X)$) by class-invariance of sc. Hence $\succeq_{\text{Val}}$ and $\succeq_{\text{Val}'}$ coincide on $\Omega$, so their maximal sets over the same $S^*$ coincide, proving the claim. $\qquad\square$

# Chapter 13

# Conclusion

In this book, we conducted a comprehensive survey of methods for uncertain decision-making, with particular attention to the roles of uncertainty modeling, weight elicitation, structural and causal analysis, compensatory and reference-based ranking schemes, outranking procedures, and related rule-based or sequential decision frameworks. We also discussed how diverse uncertainty paradigms—including fuzzy, intuitionistic fuzzy, neutrosophic, plithogenic, rough, soft, and other generalized structures—can be incorporated into decision-making processes in a systematic manner.

Overall, the survey shows that uncertain decision-making is not a single method, but a broad methodological family in which the representation of uncertainty and the choice of decision mechanism must be designed in a mutually consistent way. In particular, the selection of a suitable framework depends on the problem structure, the type of available information, the interaction among criteria, and the decision context, such as individual, group, dynamic, multi-stage, or multi-scenario settings.

We expect that future work will advance in several directions. First, further computational and experimental studies are needed in order to compare the stability, robustness, interpretability, and computational complexity of different uncertain decision-making methods under common benchmark settings. Second, more case-study–driven investigations should be developed in practical domains such as engineering design, manufacturing, finance, medicine, logistics, and social decision analysis, so that the theoretical advantages of these methods can be evaluated in realistic decision environments. Third, promising research directions include deeper integration with machine learning, intelligent optimization, expert systems, and broader decision-support frameworks, especially in settings where uncertain, incomplete, inconsistent, or multi-source information must be processed.

More generally, future studies may also focus on the development of unified mathematical frameworks capable of comparing, generalizing, and connecting existing uncertain decision-making methods. Such work may help clarify the relationships among current models and may support the design of new methods that are both theoretically well-founded and practically applicable.

# Disclaimer

**Funding**

This study was conducted without any financial support from external organizations or grants.

**Acknowledgments**

We would like to express our sincere gratitude to everyone who provided valuable insights, support, and encouragement throughout this research. We also extend our thanks to the readers for their interest and to the authors of the referenced works, whose scholarly contributions have greatly influenced this study. Lastly, we are deeply grateful to the publishers and reviewers who facilitated the dissemination of this work.

**Data Availability**

Since this research is purely theoretical and mathematical, no empirical data or computational analysis was utilized. Researchers are encouraged to expand upon these findings with data-oriented or experimental approaches in future studies.

**Ethical Statement**

As this study does not involve experiments with human participants or animals, no ethical approval was required.

**Conflicts of Interest**

The authors declare that they have no conflicts of interest related to the content or publication of this book.

**Code Availability**

No code or software was developed for this study.

## Clinical Trial

This study did not involve any clinical trials.

## Consent to Participate

Not applicable.

## Use of Generative AI and AI-Assisted Tools

I use generative AI and AI-assisted tools for tasks such as English grammar checking, and I do not employ them in any way that violates ethical standards.

## Disclaimer (Others)

This work presents theoretical ideas and frameworks that have not yet been empirically validated. Readers are encouraged to explore practical applications and further refine these concepts. Although care has been taken to ensure accuracy and appropriate citations, any errors or oversights are unintentional. The perspectives and interpretations expressed herein are solely those of the authors and do not necessarily reflect the viewpoints of their affiliated institutions.

# Appendix A

# Graphic Structure and Uncertain Graphic Structure

In this appendix, we briefly examine *graphic structures* and *uncertain graphic structures*. A graphic structure is a unified incidence-based framework consisting of a vertex set and a collection of edge objects, where each edge is associated with a nonempty finite set of incident vertices. An uncertain graphic structure is a graphic structure equipped with an uncertainty-degree map that assigns to every vertex and edge a degree tuple from a chosen uncertainty model, enabling graded and multi-component descriptions.

## A.1 Graphic structure: a unifying incidence-based model

The definition of a graphic structure is given below.

**Definition A.1.1** (Iterated universe (for nested / super-hyper entities)). Let $X_0$ be a nonempty set ("atomic" entities). Define iterated power sets by

$$\mathcal{P}^{(0)}(X_0) := X_0, \qquad \mathcal{P}^{(i+1)}(X_0) := \mathcal{P}\big(\mathcal{P}^{(i)}(X_0)\big) \quad (i \geq 0),$$

and the depth-$n$ universe by

$$\Omega^{(n)}(X_0) := \bigcup_{i=0}^{n} \mathcal{P}^{(i)}(X_0).$$

Thus $\Omega^{(0)}(X_0) = X_0$ (ordinary vertices), while $\Omega^{(n)}(X_0)$ ($n \geq 1$) contains set-valued and nested entities, enabling super/higher-order vertex types.

**Definition A.1.2** (Graphic structure of depth $n$). Fix a base set $X_0 \neq \emptyset$ and an integer $n \geq 0$. A *graphic structure of depth $n$ over $X_0$* is a tuple

$$\mathsf{GS} = \big(V, E, \mathrm{Inc}, \mathrm{Tail}, \mathrm{Head}\big)$$

satisfying:

- **Vertex set:** $V \subseteq \Omega^{(n)}(X_0)$ is a nonempty set of vertices (entities). Elements of $V$ may be atomic ($\in X_0$) or nested (set-valued), depending on $n$.

- **Edge set:** $E$ is a (possibly empty) set of edge-objects, disjoint from $V$.

- **Incidence:** $\mathrm{Inc} : E \to \mathcal{P}^+_{\mathrm{fin}}(V)$ assigns to each edge $e \in E$ its (finite, nonempty) incident vertex set $\mathrm{Inc}(e) \subseteq V$.

- **Optional direction data:** $\mathrm{Tail}, \mathrm{Head} : E \to \mathcal{P}_{\mathrm{fin}}(V)$ are maps such that, for every $e \in E$,

$$\mathrm{Tail}(e) \cap \mathrm{Head}(e) = \emptyset, \qquad \mathrm{Tail}(e) \cup \mathrm{Head}(e) = \mathrm{Inc}(e).$$

  (If direction is not needed, one may set $\mathrm{Tail}(e) = \mathrm{Head}(e) = \emptyset$ for all $e$.)

**Remark A.1.3** (Standard graph notions as special cases). Within Definition A.1.2, many familiar objects are recovered by imposing additional constraints:

- **(Undirected) simple graph:** $n = 0$, $\mathrm{Inc}(e)$ has cardinality 2 for all $e \in E$, and $\mathrm{Tail} = \mathrm{Head} = \emptyset$.

- **Directed graph:** $n = 0$, $|\mathrm{Inc}(e)| = 2$ and $|\mathrm{Tail}(e)| = |\mathrm{Head}(e)| = 1$ for all $e$ (so each edge has a unique tail and head).

- **Hypergraph:** $n = 0$, $\mathrm{Inc}(e) \in \mathcal{P}^+_{\mathrm{fin}}(V)$ arbitrary (no restriction to size 2).

- **Directed hypergraph:** $n = 0$, $\mathrm{Tail}(e), \mathrm{Head}(e)$ nonempty (typically), disjoint, and $\mathrm{Tail}(e) \cup \mathrm{Head}(e) = \mathrm{Inc}(e)$.

- **SuperHyperGraph (nested vertices):** $n \geq 1$ with $V \subseteq \Omega^{(n)}(X_0)$, allowing vertices that are sets (or sets of sets, etc.); edges are still incidence sets of such vertices via $\mathrm{Inc}$.

- **$r$-regular graph (property):** in the undirected graph case, define

$$\deg(v) := \big| \{\, e \in E : \ v \in \mathrm{Inc}(e) \,\} \big| \quad (v \in V),$$

  and require $\deg(v) = r$ for all $v \in V$.

Hence Definition A.1.2 can be viewed as a common "incidence calculus" for graph-like objects.

## A.2  Uncertain graphic structure

The definition of an uncertain graphic structure is given below. Throughout this subsection, let $M$ be an *uncertain model* with nonempty degree-domain $\mathrm{Dom}(M) \subseteq [0,1]^k$ (as defined previously in the uncertain-set framework).

**Definition A.2.1** (Uncertain graphic structure of type $M$). Let $\mathsf{GS} = (V, E, \mathrm{Inc}, \mathrm{Tail}, \mathrm{Head})$ be a graphic structure. An *uncertain graphic structure of type $M$* is a pair

$$\mathsf{UGS}_M = \big( \mathsf{GS}, \mu_M \big),$$

where

$$\mu_M : V \cup E \longrightarrow \mathrm{Dom}(M)$$

is an uncertainty-degree map assigning a degree tuple in $\mathrm{Dom}(M)$ to every vertex and every edge.

**Optional (model-dependent) consistency.** Depending on the intended specialization (fuzzy, intuitionistic fuzzy, neutrosophic, plithogenic, etc.), one may additionally impose constraints linking edge-degrees to incident vertex-degrees. For example, in the fuzzy case $\mathrm{Dom}(M) = [0,1]$, a common constraint is $\mu_M(e) \leq \min\{\mu_M(u), \mu_M(v)\}$ for an undirected edge $e$ incident to $\{u, v\}$. Such constraints are not fixed at the level of this general definition; they are selected according to the chosen model $M$ and the application semantics.

**Remark A.2.2** (Underlying crisp structure and reduction to uncertain sets)**.** There is an obvious *forgetful projection*

$$\pi : \; \mathsf{UGS}_M = \big((V, E, \mathrm{Inc}, \mathrm{Tail}, \mathrm{Head}), \mu_M\big) \; \longmapsto \; (V, E, \mathrm{Inc}, \mathrm{Tail}, \mathrm{Head}),$$

which discards uncertainty degrees and returns the underlying graphic structure. If $E = \emptyset$, then $\mathsf{UGS}_M$ reduces to an uncertain set on $V$ via the restriction $\mu_M|_V : V \to \mathrm{Dom}(M)$.

**Theorem A.2.3** (Well-definedness and existence)**.** *Fix a depth $n \geq 0$, a base set $X_0 \neq \emptyset$, and an uncertain model $M$ with $\mathrm{Dom}(M) \neq \emptyset$.*

  *(i) For every graphic structure $\mathsf{GS}$ of depth $n$ over $X_0$, there exists an uncertain graphic structure $\mathsf{UGS}_M = (\mathsf{GS}, \mu_M)$ of type $M$.*

  *(ii) The definition of $\mathsf{UGS}_M$ is well-defined: namely, for any choice of map $\mu_M : V \cup E \to \mathrm{Dom}(M)$, the pair $(\mathsf{GS}, \mu_M)$ is a valid uncertain graphic structure of type $M$.*

  *(iii) Every classical graph-like object (graph, directed graph, hypergraph, superhypergraph, etc.) can be embedded as a special case of $\mathsf{UGS}_M$ by choosing $\mathsf{GS}$ the corresponding constraints from Remark A.1.3 and taking $\mu_M$ to be constant (or crisp-valued, when appropriate).*

*Proof.* (i) Since $\mathrm{Dom}(M) \neq \emptyset$, choose any fixed degree tuple $d_0 \in \mathrm{Dom}(M)$. Define $\mu_M : V \cup E \to \mathrm{Dom}(M)$ by the constant map $\mu_M(z) := d_0$ for all $z \in V \cup E$. Then $\mu_M$ is a well-typed function with codomain $\mathrm{Dom}(M)$, so $\mathsf{UGS}_M = (\mathsf{GS}, \mu_M)$ satisfies Definition A.2.1.

(ii) Let $\mu_M : V \cup E \to \mathrm{Dom}(M)$ be any function. Definition A.2.1 requires only that $\mathsf{GS}$ is a graphic structure and that $\mu_M$ maps each vertex/edge into the degree-domain $\mathrm{Dom}(M)$. Both hold by assumption, hence $(\mathsf{GS}, \mu_M)$ is a valid uncertain graphic structure.

(iii) Take any classical structure (e.g., a graph, directed graph, hypergraph, or a nested-vertex superhypergraph). Select $n$ and $X_0$ so that its vertex objects lie in $\Omega^{(n)}(X_0)$, and encode it as a graphic structure $\mathsf{GS}$ by imposing the appropriate constraints from Remark A.1.3 on $\mathrm{Inc}, \mathrm{Tail}, \mathrm{Head}$. Finally, apply (i) to obtain $\mathsf{UGS}_M = (\mathsf{GS}, \mu_M)$ (e.g., using a constant $\mu_M$), which realizes the classical object as a special case within the uncertain graphic-structure framework. $\qquad \square$

# Appendix (List of Tables)

*

# Appendix (List of Figures)

\*

Real-world decisions rarely rely on perfectly precise information.

This book surveys major methodologies for decision-making under uncertainty, including fuzzy sets, neutrosophic sets, plithogenic models, and multi-criteria decision frameworks.

It provides a structured taxonomy of methods, computational procedures, and applications across science, engineering, and social systems.

- Fuzzy Set Theories
- Neutrosophic and Plithogenic Models
- Multi-Criteria Decision-Making (MCDM)
- Weight Elicitation Methods
- Consensus and Dynamic Decision Frameworks

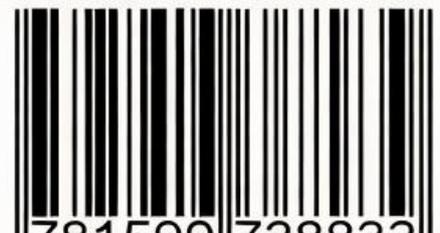

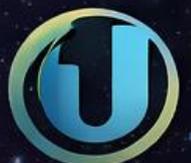